# Pattern-Based Constraint Satisfaction
# and Logic Puzzles

Denis Berthier

Institut Mines Télécom

# Pattern-Based Constraint Satisfaction
# and Logic Puzzles

Denis Berthier

# Pattern-Based Constraint Satisfaction
# and Logic Puzzles

9 8 7 6 5 4 3 2 1

# Table of Contents

# Foreword

*Motivations for the approach of the present book*

Since the 1970s, when it was identified as a class of problems with its own specificities, Constraint Satisfaction has quickly evolved into a major area of Artificial Intelligence (AI). Two broad families of very efficient algorithms (with many freely available implementations) have become widely used for solving its instances: general purpose structured search of the "problem space" (e.g. depth-first, breadth-first) and more specialised "constraint propagation" (that must generally be combined with search according to various recipes).

One may therefore wonder why they would use the computationally much harder techniques inherent in the approach introduced in the present book. It should be clear from the start that there is no reason at all if speed is the first or only criterion, as may legitimately be the case in such a typical Constraint Satisfaction Problem (CSP) as scene labelling.

But, instead of just wanting a final result obtained by any available and/or efficient method, one can easily imagine additional requirements of various types and one may thus be interested in how the solution was reached, i.e. by the *resolution path*. Whatever meaning is associated with the quoted words below, there are several inter-related families of requirements one can consider:

– the solution must be built by "constructive" methods, with no "guessing";
– the solution must be obtained by "pure logic";
– the solution must be "pattern-based", "rule-based";
– the solution must be "understandable", "explainable";
– the solution must be the "simplest" one.

Vague as they may be, such requirements are quite natural for logic puzzles and in many other conceivable situations, e.g. when one wants to ask explanations about the solution or parts of it.

Starting from the above vague requirements, Part I of this book will elaborate a formal interpretation of the first three, leading to a very general, pattern-based resolution paradigm belonging to the classical "progressive domain restriction" family and resting on the notions of a *resolution rule* and a *resolution theory*.

Then, in relation with the last purpose of finding the "simplest" solution, it will introduce ideas that, if read in an algorithmic perspective, should be considered as defining a new kind of search, "*simplest-first search*" – indeed various versions of it based on different notions of logical simplicity. However, instead of such an algorithmic view (or at least before it), a pure logic one will systematically be adopted, because:

– it will be consistent with the previous purposes,

– it will convey clear non-ambiguous semantics (and it will therefore include a unique complete specification for possibly multiple types of implementation),

– it will allow a deeper understanding of the general idea of "simplest-first search", in particular of how there can be various underlying concrete notions of logical simplicity and how these have to be defined by different kinds of resolution rules associated with different types of chain patterns. At this point, it may be useful to notice that the classical structured search algorithms are not compatible with pure logic definitions (as will be explained in the text).

*Simplest-first search and the rating of instances*

In this context, there will appear the question of rating and/or classifying the instances of a (fixed size) CSP according to their "difficulty". This is a much more difficult topic than just solving them. The families of resolution rules introduced in this book (by order of increasing complexity) will go by couples (corresponding to two kinds of chains with no OR-branching but with different linking properties, namely T-whips and T-braids); for each couple, there will be two ratings, defined in pure logic ways:

– one based on T-braids, allowing a smooth theoretical development and having good abstract computational properties; we shall devote much time to prove the *confluence property* of all the braid and T-braid resolution theories, because it justifies a "*simplest-first*" *resolution strateg*y (and the associated "simplest-first search" algorithms that may implement it) and it allows to find the "simplest" resolution path and the corresponding rating by trying *only one path*;

– one based on T-whips, providing in practice an easier to compute good approximation of the first when it is combined with the "simplest-first" strategy. (The quality of the approximation can be studied in detail and precisely quantified in the Sudoku case, but it will also appear in intuitive form in all our other examples.)

We shall explain in which restricted sense all these ratings are compatible. But we shall also show that each of them corresponds to a different legitimate pure logic view of simplicity.

In chapter 11, we shall analyse the scope of the previously defined resolution rules in terms of a search procedure with no guessing, Trial-and-Error (T&E), and of the depth of T&E necessary to solve an instance. There are universal ratings, respectively the B and the BB ratings, for instances in T&E(1) and T&E(2) (i.e. requiring no more than one or two levels of Trial-and-Error). Universality must be understood in the sense that they assign a finite rating to all of these instances, but not in the sense that they could provide a unique notion of simplicity. For instances beyond T&E(2), it is questionable whether a "pure logic" solution, with all the complex and boring steps that it would involve, would be of any interest; moreover, it appears that there may be many different incompatible notions of "simplest"; in chapter 12, we shall introduce the notion of a pattern of proof and, based on it, we shall re-assess our initial requirements. The main purpose is to provide hints about the scope of practical validity of our approach.

*Examples from logic puzzles*

Mainly because they can be described shortly and they are easy to understand with no previous knowledge, all the examples dealt with in this book will be logic puzzles: Latin Squares, Sudoku, N-Queens…, with a special status granted to Sudoku for reasons that will be explained in the Introduction. But they have been selected in such a way that they make us tackle very different types of constraints, so that this choice should not suggest a lack of generality in our approach: *transitive* constraints in Futoshiki, *non-binary arithmetic* constraints in Kakuro, *topological* and *geometric* constraints in Map colouring or path finding (Numbrix® and Hidato®).

In several places, we shall even give results that are only valid for 9×9 Sudoku (e.g. the unbiased whip classification results of minimal instances in chapter 6 and the analysis of extreme instances in chapter 11), for the purpose of illustrating with precise quantitative data questions that cannot yet be tackled with such detail in other CSPs and that call for further studies, such as:

– the difficulty (much beyond what one may imagine) of finding uncorrelated unbiased samples of minimal instances of a CSP, a pre-requisite for any statistical analysis; the way we present it shows that it is likely to appear in many CSPs; the final chapters on various other CSPs show that this is indeed true for them; (a related well known problem is that of finding the hardest instances of a CSP);

– the surprisingly high resolution power of short whips for instances in T&E(1);

– the concrete application of various classification principles to the extreme instances.

*The "Hidden Logic of Sudoku" heritage [mainly for the readers of* **HLS***]*

The origins of the work reported in this book can be traced back to my choice of Sudoku as a topic of practical classes for an introductory course in Artificial Intelligence (AI) and Rule-Based Systems in early 2006. As I was formalising for myself the simplest classical techniques (Subset rules, xy-chains) before submitting them as exercises to my students, I had two ideas that kept me interested in this game longer than I had first expected: logical symmetries between three well-known types of Subset rules (Naked, Hidden and Super-Hidden, the last of which are commonly known as "Fish") and a simple non-reversible extension (xyt-chains) of the well-known reversible xy-chains. As time passed, the short article I had planned to write grew to the size of a 430-page book: *The Hidden Logic of Sudoku – HLS* in the sequel (first edition, *HLS1*, May 2007; second edition, *HLS2*, November 2007).

The present book inherits many of the ideas I first introduced in *HLS* but it extends them to any finite CSP. Based on the classical idea of *candidate elimination*, *HLS* provided a clear logical status for the notion of a candidate (which does not pertain to the original problem formulation) and it introduced the notions of a resolution rule and a resolution theory. All the concepts were strictly formalised in Predicate Logic (FOL) – more precisely in Multi Sorted First Order Logic (MS-FOL) – which (surprisingly) was a new idea: previously, all the books and Web forums had always considered that Propositional Logic was enough. Indeed, *HLS* had to make a further step, because intuitionistic (or, equivalently, constructive) logic is necessary for the proper formalisation of the notion of a candidate.

Notwithstanding the more general formulation, the "pattern-based" conceptual framework developed in this book is very close to that of *HLS*. From the start, the framework of *HLS* was intended as a formalisation of what had always been looked for when it was said that a "pure logic solution" was wanted. The basic concepts appearing in the resolution rules introduced in *HLS* were grounded in the most elementary notions used to propose or solve a puzzle (numbers, rows, columns, blocks, …); the more elaborate ones (the various types of chain patterns) were progressively introduced and strictly defined from the basic ones. Because the concepts of a *candidate* and of a *link* between two candidates were enough to formulate most of the resolution rules, extending them to any CSP was almost straightforward. The additional requirement that appeared in *HLS* in relation with the idea of rating, that of finding the *simplest* resolution path, is also tackled here according to the same general principles as in *HLS*.

On the practical puzzle solving side, *HLS1* introduced new resolution rules, based on natural generalisations of the famous xy-chains, such as xyt-, xyz- and zyzt- chains; contrary to those proposed in the current Sudoku literature, these were not based on "Subsets" (or almost locked sets – "ALS") and most of these chains were not "reversible"; the systematic clarification and exploitation of all the generalised symmetries of the game and the combination of my first two initial ideas had also led me to the "hidden" counterparts of the previous chains (hxy-, hxyt- hxyzt- chains). Later, I found further generalisations (nrczt- chains and lassoes), pushing the idea of supersymmetry to its maximal extent and allowing to solve almost any puzzle with short chain patterns. Giving a more systematic presentation of these new "3D" chain rules was the main reason for the second edition (*HLS2*).

Still later, I introduced (on Sudoku forums) other generalisations (that, in the simplified terminology of the present book and in a formulation meaningful for any CSP, will appear as whips, braids, g-whips, $S_p$-whips, $W_p$-whips, …). These may have justified a third edition of *HLS*, but I have just added a few pages to my *HLS* website instead – concentrating my work on another type of generalisation.

It appeared to me that most of what I had done for Sudoku could be generalised to any finite CSP [Berthier 2008a, 2008b, 2009]. But, once more, as I found further generalisations and as the analysis of additional CSPs with different characteristics was necessary to guarantee that my definitions were not too restrictive, the normal size of journal articles did not fit the purposes of a clear and systematic exposition; this is how this work grew into a new book, "*Constraint Resolution Theories*" (*CRT*, November 2011).

As for the resolution rules themselves, whereas *HLS* proceeded by successive generalisations of well-known elementary rules for Sudoku into more complex ones, in *CRT* and in the present book, we start (in Part II) from powerful rules meaningful in any CSP (whips, in chapter 5) equivalent (in the Sudoku case) to those that were only reached at the end of *HLS2* (nrczt- chains and lassoes).

As a result, in this book, patterns such as Subsets, with much less resolution power than whips of same size and with more complex definitions in the general CSP than in Sudoku, come after bivalue-chains, whips and braids, and also after their "grouped" versions, g-whips and g-braids. Moreover, Subsets are introduced here with purposes very different from those in *HLS*:
1) providing them with a definition meaningful in any CSP (in particular, independent of any underlying grid structure);
2) showing that whips subsume most cases of Subsets in any CSP;
3) illustrating by Sudoku examples how, in rare cases, Subset rules can nevertheless simplify the resolution paths obtained with whips;
4) defining in any CSP a "grouped" version of Subsets, g-Subsets; surprisingly, in the Sudoku case, g-Subsets do not lead to new rules, but they give a new perspective of the well-known Franken and Mutant Fish; this could be useful for the purposes of classifying these patterns (which has always been a very obscure topic);
5) showing that, in any CSP, the basic principles according to which whips are built can be generalised to allow the insertion of Subsets into them (obtaining $S_p$-whips), thus extending the resolution power of whips towards the exceptionally hard instances.

***What is new with respect to "Constraint Resolution Theories" [mainly for the readers of* CRT]**

This book can be considered as the second, revised and largely extended edition of *Constraint Resolution Theories* (*CRT*). Following a colleague's advice, we changed the title (which seemed too technical) so that it includes the "Constraint Satisfaction" key phrase referring to its global domain; "Pattern-Based" was then a natural choice for qualifying our approach, while the explicit reference to "Logic Puzzles" became almost necessary with the addition of all the examples in part IV to the already existing Sudoku content. Apart from this cosmetic change, there are three different degrees of newness with respect to *CRT*, in increasing magnitude.

Firstly, this book corrects a few typos and errors that remained in *CRT* in spite of careful re-readings; in several places, it also marginally improves or completes the wording and it adds a few remarks or comments; moreover:

– z-chains are no longer included in the analysis of loops in sections 5.8.1 to 5.8.3; instead, the obvious and simpler fact that z-whips subsume z-chains with a global loop is mentioned;

– an unnecessary restriction in the definition of a g-label (section 7.1.1.1) has been eliminated, without modifying the notion of a g-link; this leaves unchanged the definitions of a g-candidate and of predicate "g-linked" (relating a g-candidate and a candidate); as before, these two definitions refer to the full underlying g-label and label (this is why the restriction was unnecessary); nothing else had to be changed in chapter 7 or in any place where g-labels are dealt with; in particular, this does not change the sets of g-labels of the various examples already tackled by *CRT*; however, the restriction made it impossible to apply the initial definition given in *CRT* to g-labels in Futoshiki (see chapter 14);

– the "saturation" or "local maximality" condition in the definition of a g-label has been broadened for an easier applicability to new examples; it has also been isolated by splitting the initial definition into two parts; as it was there only for efficiency purposes, but it had no impact on theoretical analyses, this entails no other changes; however, the efficiency purposes should not be underestimated: section 15.5 shows how essential this condition is in practice in Kakuro;

– section 11.4 of *CRT* (bi-whips, bi-braids, W*-whips and B*-braids) has been significantly reworded, corrected and extended, giving rise to a new chapter of its own (chapter 12);

– a section (17.4) describing our general pattern-based CSP-Rules solver, used for all the examples presented in this book, has been introduced.

Secondly, this book adds a few new results, mainly to the W-whip and B-braid patterns and/or to the Sudoku CSP case study. The following list is not exhaustive:

– very instructive whip[2] examples are given in section 8.8.1; they are the key for understanding why whips can be more powerful than Subsets of same size;

– an example of a non-whip braid[3] in Sudoku is given in section 5.10.5;

– a new graphico-symbolic representation of W-whips is introduced in section 11.2.9, based on the analogy between whips and Subsets;

– the most recent collections of extreme puzzles, harder than most of those already considered in *CRT*, published in the meantime by various puzzle creators, are analysed and their $B_7B$ classifications are given in section 11.4; these new results show that a few puzzles (we have found only three in these collections) require $B_7$-braids and they provide very strong support to our old conjecture that all the 9×9 Sudoku puzzles can be solved by T&E(2) and to our new one that they can all be solved by $B_7$-braids;

– occasionally, larger sized Sudoku grids are considered; this allows in particular to show that the universal solvability by T&E(2) is not true for them.

Thirdly and most importantly, chapter 12 and part IV about modelling various logic puzzles are almost completely new; in particular:

– chapter 12, revolving around the notion of a *pattern of proof*, shows that our initial simplicity and understandability requirements may be at variance for instances beyond T&E(1) or gT&E(1); it discusses various options for their interpretation, such as B*-braid solutions; it shows that a pure logic approach is still possible in theory, although the computational complexity may be much higher, depending on which *patterns of proof* one is ready to accept;

– chapter 13, via an illustrative example (the sk-loop in Sudoku), tackles general questions about *modelling resolution rules*; these arise when one wants to formalise new (possibly application-specific) techniques; although part of the material in it has been available for several years on the Sudoku part of our website in a rather technical form, subtle changes (making the presentation much simpler and slightly more general) appear here for the first time;

– chapter 14 on *transitive constraints* and the *Futoshiki* CSP concretely shows how the general concepts and resolution rules defined in this book can be applied to a CSP with significantly different types of constraints (inequalities) than the symmetric ones considered in the LatinSquare, Sudoku and N-Queens examples; it also shows that the few known, apparently application-specific, resolution rules of Futoshiki (ascending chains, hills and valleys) are special cases of these general rules; finally, it indicates how our controlled-bias approach to puzzle generation, at the basis of any unbiased statistical results, can be adapted to it in a straightforward way;

– chapter 15 on *non-binary arithmetic constraints* and the *Kakuro* CSP may be the most important one among our non-Sudoku examples, as it shows that the binary constraints restriction of our approach can be relaxed not only in theory but also in practice and that non-binary constraints can be efficiently managed in application-specifc ways (better than by relying on the standard general replacement method);

– chapter 16 deals with some *topological and geometric constraints* associated with *map colouring* and *path finding* (in Numbrix® and Hidato®); together with chapters 14 and 15, it confirms that our generalisations from Sudoku to the general CSP work concretely – a point in which *CRT* was partially lacking.

# 1. Introduction

## 1.1. The general Constraint Satisfaction Problem (CSP)

Many real world problems, such as resource allocation, temporal reasoning, activity scheduling, scene labelling…, naturally appear as Constraint Satisfaction Problems (CSP) [Guesguen et al. 1992, Tsang 1993]. Many theoretical problems and many logic games are also natural examples of CSPs: graph colouring, graph matching, cryptarithmetic, N-Queens, Latin Squares, Sudoku and its innumerable variants, Futoshiki, Kakuro and many other logic games (or logic puzzles).

In the past decades, the study of such problems has evolved into a main sub-area of Artificial Intelligence (AI) with its own specialised techniques. Research has concentrated on finding efficient algorithms, which was a necessity for dealing with large scale applications. As a result, one aspect of the problem has been almost completely overlooked: producing readable solutions. This aspect will be the main topic of the present book.

### 1.1.1. Statement of the Constraint Satisfaction Problem

A CSP is defined by:
– a set of variables $X_1, X_2, … X_n$, the "CSP variables", each with values in a given domain $Dom(X_1), Dom(X_2), …, Dom(X_n)$,
– a set of constraints (i.e. of relations) these variables must satisfy.

The problem consists of assigning a value from its domain to each of these variables, such that these values satisfy all the constraints. Later (in Chapter 3), we shall show that a CSP can easily be re-written as a theory in First Order Logic.

As in many studies of CSPs, all the CSPs we shall consider in this book will be finite, i.e. the number of variables, each of their domains and the number of constraints will all be finite. When we write "CSP", it should therefore always be read as "finite CSP".

Also, we shall consider only CSPs with binary constraints. One can always tackle unary constraints by an appropriate choice of the domains. And, for k > 2, a k-ary constraint between a subset of k variables $(X_{n1}, .., X_{nk})$ can always be replaced by k binary constraints between each of these $X_{ni}$ and an additional variable representing the original k-ary constraint; although this new variable has a large domain and this may be a very inefficient way of dealing with the given k-ary constraint, this is a very standard approach (for details, see [Tsang 1993]). With the Kakuro CSP, chapter 15 will show an example of how this can be done in practice, using application specific techniques more efficient than the general method.

Moreover, a binary CSP can always be represented as a (generally large) labelled undirected graph: a node (or vertex) of this graph, called a *label*, is a couple < CSP variable, possible value for it > (or, in our approach, an equivalence class of such couples); given two nodes in this graph, each binary constraint not satisfied by this pair of labels (including the "strong" constraints induced by CSP variables, i.e. all the contradictions between different values for the same CSP variable) gives rise to an arc (or edge) between them, labelled by the name of the constraint and representing it. We shall call this graph *the CSP graph*. (Notice that this is different from what is usually called the constraint graph.) The CSP graph expresses all the direct contradictions between any two labels (whereas the constraint graph usually considered in the CSP literature expresses their compatibilities).

### 1.1.2. *The Sudoku example*

As explained in the foreword, Sudoku has been at the origin of our work on CSPs. In this book, we shall keep it as our main example for illustrating the techniques we introduce, even though we shall also deal with other CSPs in order to palliate its specificities (for other detailed examples, see chapters 14 to 16).

Let us start with the usual formulation of the problem (with its own, self-explanatory vocabulary in italics): given a 9×9 *grid*, partially filled with *numbers* from 1 to 9 (the *givens* of the problem, also called the *clues* or the *entries*), *complete* it with *numbers* from 1 to 9 in such a way that in each of the nine *rows*, in each of the nine *columns* and in each of the nine disjoint *blocks* of 3×3 contiguous *cells*, the following property holds:

– there is *at most* one occurrence of each of these numbers.

Although this defining condition could be replaced by either of the following two, which are obviously equivalent to it, we shall stick to the first formulation, for reasons that will appear later:

– there is *at least* one occurrence of each of these numbers,

– there is *exactly* one occurrence of each of these numbers.

Figure 1.1 shows the standard presentations of a *problem grid* (also called a *Sudoku puzzle*) and of a *solution grid* (also called a *complete Sudoku grid*).

Since rows, columns and blocks play similar roles in the defining constraints, they will naturally appear to do so in many other places and a word that makes no difference between them is widely used in the Sudoku world: a *unit* is either a row or a column or a block. And one says that two cells *share a unit*, or that they *see* each other, if they are different and they are either in the same row or in the same column or in the same block (where "or" is non exclusive). We shall also say that these two cells are *linked*. It should be noticed that this (symmetric) relation between two different cells, whichever of the three equivalent names it is given, does not depend on the content of these cells but only on their place in the grid; it is therefore a straightforward and quasi physical notion.

|   |   |   |   |   |   |   | 1 | 2 |
|---|---|---|---|---|---|---|---|---|
|   |   |   | 3 | 5 |   |   |   |   |
|   |   | 6 |   |   |   | 7 |   |   |
| 7 |   |   |   |   | 3 |   |   |   |
|   |   |   | 4 |   | 8 |   |   |   |
| 1 |   |   |   |   |   |   |   |   |
|   |   |   | 1 | 2 |   |   |   |   |
|   | 8 |   |   |   |   | 4 |   |   |
|   | 5 |   |   |   | 6 |   |   |   |

| 6 | 7 | 3 | 8 | 9 | 4 | 5 | 1 | 2 |
|---|---|---|---|---|---|---|---|---|
| 9 | 1 | 2 | 7 | 3 | 5 | 4 | 8 | 6 |
| 8 | 4 | 5 | 6 | 1 | 2 | 9 | 7 | 3 |
| 7 | 9 | 8 | 2 | 6 | 1 | 3 | 5 | 4 |
| 5 | 2 | 6 | 4 | 7 | 3 | 8 | 9 | 1 |
| 1 | 3 | 4 | 5 | 8 | 9 | 2 | 6 | 7 |
| 4 | 6 | 9 | 1 | 2 | 8 | 7 | 3 | 5 |
| 2 | 8 | 7 | 3 | 5 | 6 | 1 | 4 | 9 |
| 3 | 5 | 1 | 9 | 4 | 7 | 6 | 2 | 8 |

***Figure 1.1.*** *A puzzle (Royle17#3) and its solution*

As appears from the definition, a Sudoku grid is a special case of a Latin Square. Latin Squares must satisfy the same constraints as Sudoku, except the condition on blocks. Following *HLS1*, the logical relationship between the two theories will be fully clarified in chapters 3 and 4.

What we need now is to see how the above natural language formulation of the Sudoku problem can be re-written as a CSP. In Chapter 2, the essential question of modelling in general and its practical implications on how to deal with a CSP will be raised and we shall see that the following formalisation is neither the only one nor the best one. But, for the time being, we only want to write the most straightforward one.

For each row r and each column c, introduce a variable $X_{rc}$ with domain the set of digits {1, 2, 3, 4, 5, 6, 7, 8, 9}. Then the general Sudoku problem can be expressed as a CSP for these variables, with the following set of (binary) constraints:
$X_{rc} \neq X_{r'c'}$ for all the pairs {rc, r'c'} such that the cells rc and r'c' share a unit, and a particular puzzle will add to these binary constraints the set of unary constraints fixing the values of the $X_{rc}$ variables corresponding to the givens.

Notice that the natural language phrase "complete the grid" in the original formulation has naturally been understood as "assign one and only one value to each of the cells" – which has then been translated into "assign a value to each of the $X_{rc}$ variables" in the CSP formulation.

**1.2. Paradigms of resolution**

A CSP states the constraints a solution must satisfy, i.e. it says *what* is desired. But it does not say anything about *how* a solution can be obtained; this is the job of resolution methods, the choice of which will depend on the various purposes one may have in addition to merely finding a solution. A particular class of resolution methods, based on *resolution rules*, will be the main topic of this book.

*1.2.1. Various purposes and methods*

If one's only goal is to get a solution by any available means, very efficient general-purpose algorithms have been known for a long time [Kumar 1992, Tsang 1993]; they guarantee that they will either find one solution or all the solutions (according to what is desired) or find a contradiction in the givens; they have lots of more recent variants and refinements. Most of these algorithms involve the combination of two very different techniques: some direct propagation of constraints between variables (in order to progressively reduce their sets of possible values) and some kind of structured search with "backtracking" (depth-first, breadth-first, …, possibly with some forms of look-ahead); they consist of trying (recursively if necessary) a value for a variable and propagating (based on the constraints) the consequences of this tentative choice as restrictions on other variables; eventually, either a solution or a contradiction will be reached; the latter case allows to conclude that this value (or this combination of values simultaneously tried in the recursive case) is impossible and it restricts the possibilities for this (subset of) variables(s).

But, in some cases, such blind search is not possible for practical reasons (e.g. one is not in a simulator but in real life) or not allowed (for *a priori* theoretical or æsthetic reasons), or one wants to simulate human behaviour, or one wants to "understand" or to be able to "explain" each step of the resolution process (as is generally the case with logic puzzles), or one wants a "constructive" solution (with no "guessing") or one wants a "pure logic" or a "pattern-based" or a "rule-based" or the "simplest" solution, whatever meaning they associate with the quoted words.

Contrary to the current CSP literature, this book will only deal with the latter cases and more attention will be paid to the resolution path than to the final solution itself. Indeed, it can also be considered as an informal reflection on how notions such as "no guessing", "a constructive solution", "a pure logic solution", "a pattern-based solution", "an understandable proof of the solution", "an explanation of the solution" and "the simplest solution" can be defined (but we shall only be able to say more on this topic in the retrospective "final remarks" chapter). It does not mean that efficiency questions are not relevant to our approach, but they are not our primary goal, they are conditioned by such higher-level requirements. Without these additional requirements, there is no reason to use techniques computationally much harder (probably exponentially much harder) than the general-purpose algorithms.

In such situations, it is convenient to introduce the notion of a *candidate*, i.e. of a "still possible" value for a variable. As this intuitive notion does not pertain to the CSP itself, it must first be given a clear definition and a logical status. When this is done (in chapter 4), one can define the concepts of a *resolution rule* (a logical formula in the "condition $\Rightarrow$ action" form, which says what to do in some factual, observable situation described by the condition *pattern*), a *resolution theory* (a set of such rules), a *resolution strategy* (a particular way of using the rules in a resolution theory). One can then study the relationship between the original CSP and several of its resolution theories. One can also introduce several properties a resolution theory can have, such as confluence and completeness (contrary to general purpose algorithms, a resolution theory cannot in general solve all the instances of a given CSP; evaluating its scope is thus a new topic in its own; one can also study its statistical resolution power in specific CSP cases).

This "rule-based" or "pattern-based" approach was first introduced in *HLS1*, in the limited context of Sudoku. It is the purpose of this book to show that it is indeed very general and chapters 14 to 16 will concretely show that it does apply to the very different types of constraints appearing in Futoshiki, Kakuro and map colouring, but let us first illustrate how these ideas work for Sudoku.

### *1.2.2. Candidates and candidate elimination in Sudoku*

The process of solving a Sudoku puzzle "by hand" is generally initialised by defining the "candidates" for each cell. For later formalisation, one must be careful with this notion: if one analyses the natural way of using it, it appears that, *at any stage of the resolution process, a candidate for a cell is a number that has not yet been explicitly proven to be an impossible value for this cell*.

Usually, candidates for a cell are displayed in the grid as smaller and/or clearer digits in this cell (as in Figure 1.2). Similarly, at any stage, a *decided value* is a number that has been explicitly proven to be the only possible value for this cell; it is written in big fonts, like the givens.

At the start of the game, one possibility is to consider that any cell with no input value admits all the numbers from 1 to 9 as candidates – but more subtle initialisations are possible (e.g. as shown in Figure 1.2) and a slightly different, more symmetric, view of candidates can be introduced (see chapter 2).

Then, according to the formalisation introduced in *HLS1*, a resolution process that corresponds to the vague requirement of a "pure logic" solution is a sequence of steps consisting of repeatedly applying "resolution rules" of the general condition-action type: if some *pattern* – i.e. configuration of cells, possible cell-values, links, decided values, candidates and non-candidates – defined by the condition part of the rule, is effectively present in the grid, then carry out the action(s) specified by the action part of the rule. Notice that any such pattern always has a purely "physical", invariant part (which may be called its "physical" or "structural" support), defined by conditions on possible cell-values and on links between them, and an additional part, related to the actual presence/absence of decided values and/or candidates in these cells in the current situation. (Again, this will be generalised in chapter 2 with the four "2D" views.)

|    | c1 | c2 | c3 | c4 | c5 | c6 | c7 | c8 | c9 |    |
|----|----|----|----|----|----|----|----|----|----|----|
| r1 | 3456 89 | 3 4 6 7 9 | 3 4 5 6 7 8 9 | 7 8 9 | 4 7 8 9 | 4 7 8 9 | 4 5 9 | 1 | 2 | r1 |
| r2 | 2 4 6 8 9 | 1 2 4 6 7 9 | 1 2 4 6 7 8 9 | 2 7 8 9 | 3 | 5 | 4 9 | 6 8 9 | 4 6 8 9 | r2 |
| r3 | 2 3 4 5 8 9 | 1 2 3 4 9 | 1 2 3 4 5 8 9 | 6 | 1 4 8 9 | 1 2 4 8 9 | 4 5 9 | 7 | 3 4 5 8 9 | r3 |
| r4 | 7 | 2 4 6 9 | 2 4 5 6 8 9 | 2 5 8 9 | 1 5 6 8 9 | 1 2 6 8 9 | 3 | 2 5 6 9 | 1 4 5 6 9 | r4 |
| r5 | 2 3 5 6 9 | 2 3 6 9 | 2 3 5 6 9 | 4 | 1 5 6 7 9 | 1 2 3 6 7 9 | 8 | 2 5 6 9 | 1 5 6 7 9 | r5 |
| r6 | 1 | 2 3 4 6 9 | 2 3 4 5 6 8 9 | 2 3 5 7 8 9 | 5 6 7 8 9 | 2 3 6 7 8 9 | 2 4 5 7 9 | 2 5 6 9 | 4 5 6 7 9 | r6 |
| r7 | 3 4 6 9 | 3 4 6 7 9 | 3 4 6 7 9 | 1 | 2 | 3 4 6 7 8 9 | 5 7 9 | 3 5 8 9 | 3 5 7 8 9 | r7 |
| r8 | 2 3 6 9 | 8 | 1 2 3 6 7 9 | 3 5 7 9 | 5 6 7 9 | 3 6 7 9 | 1 2 5 7 9 | 4 | 1 3 5 7 9 | r8 |
| r9 | 2 3 4 9 | 5 | 1 2 3 4 7 9 | 3 7 8 9 | 4 7 8 9 | 4 7 8 9 | 6 | 2 3 8 9 | 1 3 7 8 9 | r9 |
|    | c1 | c2 | c3 | c4 | c5 | c6 | c7 | c8 | c9 |    |

**Figure 1.2.** *Grid Royle17#3 of Figure 1.1, with the candidates remaining after the elementary constraints for the givens have been propagated*

Depending on the type of their action part, such resolution rules can be classified into two categories (assertion type and elimination type):

– either they assert a decided value for a cell (e.g. the Single rule: if it is proven that there is only one possibility left for it); there are very few such assertion rules;

– or they eliminate some candidate(s) (which we call the *target(s)* of the pattern); as appears from a quick browsing of the available literature, almost all the classical Sudoku resolution rules are of this type (and, apart from Singles, the few rules that seem to be of the assertion type can be reduced to elimination rules); they express elaborated forms of constraints propagation; their general form is: if such pattern is present, then it is impossible for some number(s) to be in some cell(s) and the target candidates must therefore be deleted; for the general CSP also, all the rules we shall meet in this book, apart from Singles, will be of the elimination type.

The interpretation of the above resolution rules, whatever their type, should be clear: none of them claims that there is a solution with such value asserted or such candidate deleted. Rather, it must be interpreted as saying: "from the current situation it can be asserted that any solution, if there is any, must satisfy the conclusion of this rule".

From both theoretical and practical points of view, it is also important to notice that, as one proceeds with resolution, candidates form a monotone decreasing set and decided values form a monotone increasing set. Whereas the notion of a candidate is the intuitive one for players, what is classical in logic is increasing monotonicity (what is known / what has been proven can only increase with time); but this is not a real problem, as it could easily be restored by considering non-candidates instead (i.e. what has been erased instead of what is still present).

For some very difficult puzzles, it seems necessary to (recursively) make a hypothesis on the value of a cell, to analyse its consequences and to eliminate it if it leads to a contradiction; techniques of this kind do not fit *a priori* the above condition-action form; they are proscribed by purists (for the main reason that they often make the game totally uninteresting) and they are assigned the infamous, though undefined, name of Trial-and-Error. As shown in *HLS* and in the statistics of chapter 6, they are needed in only extremely rare cases if one admits the kinds of chain rules (whips) that will be introduced in chapter 5.

### *1.2.3. Extension of this model of resolution to the general CSP*

It appears that the above ideas can be generalised from Sudoku to any CSP. Candidate elimination corresponds to the now classical idea of domain restriction in CSPs. What has been called a candidate above is related to the notion of a *label* in the CSP world, a name coming from the domain of scene labelling, which historically led to identifying the general Constraint Satisfaction Problem. However, contrary to labels that can be given a very simple set theoretic definition based on the data defining the CSP, the status of a candidate is not *a priori* clear from the point of view of mathematical logic, because this notion does not pertain *per se* to the CSP formulation, nor to its direct logic transcription.

In chapter 4, we shall show that a formal definition of a candidate must rely on intuitionistic logic and we shall introduce more formally our general model of resolution. Then we shall define the notion of a resolution theory and we shall show that, for each CSP, a Basic Resolution Theory can be defined. Even though this Basic Theory may not be very powerful, it will be the basis for defining more elaborate ones; it is therefore "basic" in the two meanings of the word.

**1.3. Parameters and instances of a CSP; minimal instances; classification**

Generally, a CSP defines a whole family of problem instances.

Typically, there is an integer parameter that splits this family into subclasses. A good example of such a parameter is the size of the grid in N-Queens, Latin Squares, Sudoku or Futoshiki; in Kakuro, it could be the number of white cells. In the resource allocation problem, it could be some combination of the number of resources and the number of tasks competing for them. In graph colouring and graph matching, it could be the size of the graph (e.g. the number of vertices or some combination of the number of vertices and the number of edges).

*1.3.1. Minimal instances*

Typically also, once this main parameter has been fixed, there remains a whole family of instances of the CSP. In 9×9 Sudoku, an instance is defined by a set of givens. In N-Queens, although the usual presentation of the problem starts from an empty grid and asks for all the solutions, we shall adopt for our purposes another view of this CSP; it consists of setting a few initial entries and asking for a solution or a "readable" proof that there is none. In "pure" Futoshiki, an instance is defined by a set of inequalities between adjacent cells; in Kakuro by a set of sum constraints in horizontal or vertical sectors. In graph colouring, the possibilities are still more open: there may be lots of graphs of a given size and, once such a graph has been chosen, it may also be required to have predefined colours for some subsets of vertices (although this is a non-standard requirement in graph theory). The same remarks apply to graph matching, where one may want to have predefined correspondences between some vertices (and/or edges) of the two graphs.

In such cases, classifying all the instances of a CSP or doing statistics on the difficulty of solving them meets problems of two kinds. Firstly, lots of instances will have very easy solutions: if givens are progressively added to an instance, until only the values of few variables remain non given, the problem becomes easier and easier to solve. Conversely, if there are so few instances that the problem has several solutions, some of these may be much easier to find than others. These two types of situations make statistics on all the instances somewhat irrelevant. This is the motivation for the following definition (inherited from the Sudoku classics).

Definition: an instance of a CSP is called *minimal* if it has one and only one solution and any instance obtained from it by eliminating any of its givens has more than one solution. [This is a notion of *local* minimality.]

For the above-mentioned reasons, all our statistical analyses of a CSP (and only the statistical ones!) will be restricted to the set of its minimal instances.

### *1.3.2. Rating and the complexity distribution of instances*

Classically, the complexity of a CSP is studied with respect to its main size parameter and one relies on a worst case (or more rarely on a mean case) analysis. It often reaches conclusions such as "this CSP is NP-complete" – as is the case for Sudoku(n) or LatinSquare(n), considered as depending on grid size n.

The questions about complexity that we shall tackle in this book are of a very different kind; they will not be based on the main size parameter. Instead, they will be about the statistical complexity distribution of instances of a fixed size CSP.

This supposes that we define a measure of complexity for instances of a CSP. We shall therefore introduce several ratings (starting in chapter 5) that are meaningful for the general CSP. And we shall be able to give detailed results (in chapter 6) for the standard (i.e. 9×9) Sudoku case. In trying to do so, the problem arises of creating unbiased samples of minimal instances and it appears to be very much harder than one may expect. We shall be able to show this in full detail only for the particular Sudoku case, but our approach is sufficiently general to suggest that the same kind of problem is very likely to arise in any CSP; moreover, the final chapters on different logic puzzles will show that they do face the same problem.

Indeed, we shall define measures of complexity associated with various families of resolution rules. For each of them, the complexity of a CSP instance will be defined as the complexity of the hardest rule in this family necessary to solve it, which is also the complexity of the hardest step of the "simplest" resolution path using only rules from this family. Sudoku examples show that a given set of rules can solve puzzles whose full resolution paths vary largely in intuitive complexity (whatever intuitive notion of complexity one adopts for the paths), but the hardest step rating is *statistically* meaningful; moreover, there is currently no idea about how to formally define the complexity of a full path, i.e. of how to combine in a consistent way the complexities of a sequence of individual steps.

The main advantage of considering ratings of the hardest step type is that, for each family of rules, an associated rank can be defined in a very simple, pure logic way. This naturally leads to an interpretation of our initial "simplest solution" requirement and to the notion of a "simplest-first strategy".

**1.4. The basic and the more complex resolution theories of a CSP**

Following the definition of the CSP graph in section 1.1.1, we say that two candidates are linked by a direct contradiction, or simply *linked*, if there is a constraint making them incompatible (including the obvious "strong" constraints, usually not explicitly stated as such, that different values for a CSP variable are incompatible).

*1.4.1. Universal elementary resolution rules and their limitations*

Every CSP has a Basic Resolution Theory: BRT(CSP). The simplest elimination rule (obviously valid for any CSP) is the direct translation of the initial problem formulation into operational rules for managing candidates. We call it the "elementary constraints propagation rule" (ECP):

– ECP: if a value is asserted for a CSP variable (as is the case for the givens), then remove any candidate that is linked to this value by a direct contradiction.

The simplest assertion rule (also obviously valid) is called Single (S):

– S: if a CSP variable has only one candidate left, then assert it as the only possible value of this variable.

There is also an obvious Contradiction Detection rule (CD):

– CD: if a CSP variable has no decided value and no candidate left, then conclude that the problem has no solution.

Together, the "elementary rules" ECP, S and CD constitute the Basic Resolution Theory of the CSP, BRT(CSP).

In Sudoku, novice players may think that these three elementary rules express the whole problem and that applying them repeatedly is therefore enough to solve any puzzle. If such were the case, one would probably never have heard of Sudoku, because it would amount to mere paper scratching and it would soon become boring. Anyway, as they get stuck in situations in which they cannot apply any of these rules, they soon discover that, except for the easiest puzzles, this is very far from being sufficient. The puzzle in Figure 1.1 is a very simple illustration of how one gets stuck if one only knows and uses the elementary rules: the resulting situation is shown in Figure 1.2, in which none of these rules can be applied. For this puzzle, modelling considerations related to symmetry (chapter 2) lead to "Hidden Single" rules allowing to solve it, but even this is generally very far from being enough.

*1.4.2. Derived constraints and more complex resolution theories*

As we shall see later, there are lots of puzzles that require resolution rules of a much higher complexity than those in the Basic Resolution Theory in order to be solved. And this is why Sudoku has become so popular: all but the easiest puzzles need a particular combination of neuron-titillating techniques and they may even suggest the discovery of as yet unknown ones.

In any CSP, the general reason for the limited resolution power of its Basic Resolution Theory can be explained as follows. Given a set of constraints, there are usually many "derived" or "implied" constraints not immediately obvious from the original ones. Many resolution rules can be considered as a way of expliciting some of the derived unary constraints. As we shall see that very complex resolution rules are needed to solve some instances of a CSP, this will show not only that derived constraints cannot be reduced to the elementary rules of the Basic Resolution Theory (which constitute the most straightforward operationalization of the axioms) but also that they can be unimaginably more complex than the initial constraints.

With all our examples being minimal instances, secondary questions about multiple or inexistent solutions can be discarded. From an epistemological point of view, the gap between the *what* (the initial constraints) and the *how* (the resolution rules necessary to solve an instance) is thus exhibited in all its purity, in a concrete way understandable by anyone. [In spite of my formal logic background and of my familiarity with all the well-known mathematical ideas more or less related to it (culminating in deterministic chaos), this gap has always been for me a subject of much wonder. It is undoubtedly one of the main reasons why I kept interested in the Sudoku CSP for much longer than I expected when I first chose it as a topic for practical classes in AI.]

All the families of resolution rules defined in this book can be seen as different ways of exploring this gap – and the consideration of derived binary constraints and/or larger Sudoku grids shows that the gap can be still much larger or deeper than shown by the standard 9×9 case.

### *1.4.3. Resolution rules and resolution strategies; the confluence property*

One last point can now be clarified: the difference between a resolution theory (a set of resolution rules) and a resolution strategy. Everywhere in this book, a *resolution strategy* must be understood in the following extra-logical sense:

– a set of *resolution rules*, i.e. a *resolution theory*, plus

– a *non-strict precedence ordering* of these rules. Non-strict means that two rules can have the same precedence (for instance, in Sudoku, there is no reason to give a rule higher precedence than a rule obtained from it by transposing rows and columns or by any of the generalised symmetries explained in chapter 2).

As a consequence of this definition, several resolution strategies can be based on the same resolution theory with different partial orderings of its rules and they may lead to different resolution paths for a given instance.

Moreover, with every resolution strategy one can associate several deterministic procedures for solving instances of the CSP, as given by the following (sketchy) pseudo-code.

As a preamble (each of the following choices will generate a different procedure):
- list all the resolution rules in a way compatible with their precedence ordering (i.e. among the different possibilities of doing so, choose one);
- list all the labels in a predefined order or take them in random order.

Given an instance P, loop until a solution of P is found (or until all the solutions are found or until it is proven that P has no solution):
|   Do until a rule can effectively be applied:
|   |   Take the first rule not yet tried in the list
|   |   Do until its condition pattern is effectively active:
|   |   |   Try to apply all the possible mappings of the condition pattern of this rule
|   |   |   to subsets of labels, according to their order in the list of labels
|   |   End do
|   End do
|   Apply the rule to the selected matching pattern
End loop

In this context, a natural question arises: given a resolution theory T, can different resolution procedures built on T lead to an instance being finally solved by some of them and unsolved by others? The answer lies in the *confluence property* of a resolution theory, to be explained in chapter 5; this fundamental property implies that the order in which the rules of T are applied is irrelevant as long as we are only interested in solving instances (but it can still be relevant when we also consider the efficiency of the procedure): all the resolution paths will lead to the same final state.

This apparently abstract confluence property (first introduced in *HLS1*) has very practical consequences when it holds in a resolution theory T. It allows any opportunistic strategy, such as applying a rule as soon as a pattern instantiating it is found (e.g. instead of waiting to have found all the potential instantiations of rules with the same precedence before choosing which should be applied first). Most importantly, it also allows to define a "simplest first" strategy that is guaranteed to produce a correct rating of an instance with respect to T after following a single resolution path (with the easy to imagine computational consequences).

**1.5. The roles of logic, AI, Sudoku and other examples**

As its organisation shows, this book about the general CSP has a large part (about a quarter) dedicated to illustrating the abstract concepts with a detailed case study of Sudoku; to a lesser extent, it also provides examples from various other logic puzzles. It can be considered as an exercise in either logic or AI or any of these games. Let us clarify the roles we grant each of these topics.

### *1.5.1. The role of logic*

Throughout this book, the main function of logic will be to provide a rigorous framework for the precise definitions of our basic concepts (such as a "candidate", a "resolution rule" and a "resolution theory"). Apart from the formalisation of the CSP itself, the simplest and most striking example is the formalisation (in section 4.3) of the CSP Basic Resolution Theory informally defined in section 1.4.1 and of all the forthcoming more complex resolution theories. Logic will also be used as a compact notational tool for expressing some resolution rules in a non-ambiguous way. In the Sudoku example, it will also be a very useful tool for expliciting the precise symmetry relationships between different "Subset rules" (in chapter 8).

For better readability, the rules we introduce are always formulated first in plain English and their validity is only established by elementary non-formal means. The non-mathematically oriented reader should thus not be discouraged by the logical formalism. Moreover, all the types of chain rules we shall consider will always be represented in a very intuitive, almost graphical formalism.

As a fundamental and practical application of our strict logical foundations to the Sudoku CSP, its natural symmetry properties can be transposed into three formal meta-theorems allowing one to deduce systematically new rules from given ones (see chapter 2 and sections 3.6 and 4.7). In *HLS*, this allowed us to introduce chain rules of completely new types (e.g. "hidden chains"). It also allowed the statement of a clear logical relationship between Sudoku and Latin Squares.

Finally, the other role assigned to logic is that of a mediator between the intuitive formulation of the resolution rules and their implementation in an AI program (e.g. our general purpose CSP-Rules solver). This is a methodological point for AI (or software engineering in general): no program development should ever be started before precise definitions of its components are given (though not necessarily in strict logical form) – a commonsense principle that is very often violated, especially by those who consider it as obvious [this is the teacher speaking!]. Notice however that the logical formalism is only one among other preliminaries to implementation (even in the form of rules of an inference engine) and that it does not dispense with the need for some design work (be it only for efficiency matters!).

### *1.5.2. The role of AI*

The role we assign to AI in this book is mainly that of providing a quick testbed for the general ideas developed in the theoretical part. The main rules have been implemented in our general CSP-Rules solver. This was initially designed for Sudoku only (and accordingly named SudoRules), with input and output functions dedicated to Sudoku, but the hard core (CSP-Rules) can be applied to any CSP and all the examples of chapters 14 to 16 also rely on it. See section 17.4 for more about CSP-Rules and the specific CSPs that have already been interfaced to it.

One important facet of the rules introduced in this book is their resolution power. This can only be tested on specific examples but the resolution of each instance by a human solver needs a significant amount of time and the number of instances that can be tested "by hand" against any resolution method is very limited. On the contrary, implementing our resolution rules in a solver allowed us to test about ten millions of Sudoku puzzles (see chapter 6). This also gave us indications of the relative efficiency of different rules. It is not mere chance that the writing of *HLS, CRT* and the present book occurred in parallel with successive versions of (SudoRules and) CSP-Rules. Abstract definitions of the relative complexities of rules were checked against our puzzle collections for their resolution times and for their memory requirements (in terms of the number of partial chains generated).

This book can also be considered as the basis for a long exercise in AI. Many computer science departments in universities have used Sudoku for various projects. According to our personal experience, it is a most welcome topic for student projects in computer science or AI. This is also true of the other types of puzzles introduced in chapters 14 to 16. Trying to implement some rules, even the "simple" Subset rules of chapter 8 and even in an application-specific way, shows how re-ordering the conditions can drastically change the behaviour of a knowledge-based system: without care, Quads can easily lead to memory overflow problems. (We give detailed formulations for Subset rules in Sudoku, also valid for games based on similar square grids, so that they can be used for such exercises without too long preliminaries.) Trying to implement $S_p$-whips or $W_p$-whips is a real challenge.

### 1.5.3. *The role of Sudoku*

Because some parts of this book related to the general CSP may seem abstract to the non-mathematician reader (e.g. chapters 3 and 4) or technical (e.g. chapters 9 to 11), a detailed case study was needed to show progressively how the general concepts work in practice. It is also necessary to show how the general theory can easily be adapted, in the most important initial modelling phase, for dealing more efficiently or more naturally with each specific case. Choosing Sudoku for these purposes was for us a natural consequence of the historical development of the techniques described here, both the general approach and all the types of resolution rules. But there are many other reasons why it is an excellent example for the general CSP.

A fast browsing of this book shows that examples from the Sudoku CSP appear in many chapters (generally at the end, in order not to overload the main text with long resolution paths) and we keep our *HLS* constraint that all of them should originate in a real minimal puzzle. But it should be clear for the readers of *HLS* that the purpose here is very different: we have no goal of illustrating with a Sudoku example each of the rules we introduce (for this, there is *HLS*).

Each example is chosen to satisfy a precise function with respect to the general Constraint Satisfaction Problem, such as providing a counter-example to some conjecture. As a result, most of our Sudoku examples will be exceptional cases, with very long resolution paths – which (without this warning) could give a very bad idea of how difficult the resolution paths look for the vast majority of instances; the statistics in chapter 6 will give a much better idea: most of the time, the chains used and the paths are short.

*1.5.3.1. Why Sudoku is a good example*

Sudoku is known to be NP-complete [Gary & al. 1979]; more precisely, the CSP family Sudoku(n) on square grids of sizes n×n for all n is NP-complete. As we fix n = 9, this should not have any impact on our analyses. But the Sudoku case will exemplify very clearly (in chapter 6) that, for fixed n, the instances of an NP-complete problem often have a broad spectrum of complexity. It will also show that standard analyses, only based on worst case (worst instances) or (more rarely) mean case, can be very far from reflecting the realities of a CSP.

For fixed n = 9, Sudoku is much easier to study than other readily formalised problems such as Chess or Go or any "real world" example. But it keeps enough structure so that it is not obvious.

Sudoku is a particular case of Latin Squares. Latin Squares are more elegant (and somehow more "respectable") from a mathematical point of view, because they enjoy a complete symmetry of all the types of variables: numbers, rows, columns. In Sudoku, the constraint on blocks introduces some apparently mild complexity that makes it more exciting for players. But this lack of full symmetry also makes it much more interesting from a theoretical point of view. In particular, it allows to introduce the notion of a grouped label (g-label), not present in Latin Squares, and new resolution rules based on it: g-whips and g-braids (see chapter 7). It is noticeable that, with the proper definition of these patterns, they appear (in very different guises) in many other CSPs.

There are millions of Sudoku players all around the world and many forums where the rules defined in *HLS* have been the topic of much debate. A huge amount of invaluable experience has been cumulated and is available – including generators of random (but biased) puzzles, collections of puzzles with very specific properties (fish patterns, symmetry properties, …) and other collections of extremely hard puzzles. The lack of similar collections and of generators of minimal instances is a strong limitation for the detailed analysis of other CSPs.

*1.5.3.2. Origin of our Sudoku examples*

Most of our Sudoku examples rely on the following sets of minimal puzzles:

– the *Sudogen0* collection consists of 1,000,000 puzzles randomly generated by us with the top-down suexg generator (http://magictour.free.fr/suexco.txt), with seed 0 for the random numbers generator; puzzle number n is named Sudogen0#n;

– the *cb* collection consists of 5,926,343 puzzles we produced with a new kind of generator, the controlled-bias generator (we first introduced it on the late Sudoku Player's Forum; see also [Berthier 2009] and chapter 6 below); it is still biased, but much less than the previously existing ones and in a precisely known way, so that it allows to compute unbiased statistics; puzzle number n is named cb#n;

– the *Magictour* collection of 1,465 puzzles considered to be the hardest (at the time of its publishing); puzzle number n is named Magictour-top1465#n;

– the *gsf* collection of 8,152 puzzles considered to contain the hardest puzzles (at the time of its publishing); puzzle number n is named gsf-top8152 #n;

– the recent *eleven* collection of 26,370 puzzles not solvable by T&E($S_4$); puzzle number n is named eleven#n; we occasionally refer to complementary collections so as to deal with all the known hardest puzzles (see chapter 11).

*1.5.4. The role of non Sudoku examples*

Although Sudoku is a very good CSP example, it has a few specificities, such as (the major one of) having only "strong" constraints (i.e. all its constraints are defined by CSP variables). With other examples (e.g. N-Queens), we shall show that these specificities have no negative impact on our general theory: the main resolution rules (for whips, g-whips, Subsets, $S_p$-whips, $W_p$-whips, braids, …) can effectively be applied to other CSPs; we shall also illustrate how different these patterns may look in these cases.

We are aware that many more examples should be granted as much consideration as Sudoku. We hope that the final chapters partially palliate this shortcoming by considering CSPs based on constraints of very different kinds (*transitive* in Futoshiki, *non-binary arithmetic* in Kakuro, *topological* and *geometric* in Map colouring, Numbrix® and Hidato®). We also hope that this book will motivate more research for applications to other CSPs.

*1.5.5. Uniform presentation of all the examples*

If we displayed the full resolution path of an instance, it would generally take several pages, most of which would describe obvious or uninteresting steps. We shall skip most of these steps, by adopting the following conventions (the same as in *HLS*):

– elementary constraint propagation rules (ECP) will never be displayed;

– as the final rules that apply to any instance are always ECP and Singles (at least when these rules are given higher priority than more complex ones – which is a natural choice), they will be omitted from the end of the path.

All our examples respect the following uniform format. After an introductory text explaining the purpose of the example, the resolution theory T applied to it and/or comments on some particular point, a row of two (sometimes three) grids is displayed: the original puzzle (sometimes an intermediate state) and its solution. Then comes *the resolution path, a proof of the solution within theory T, where "proof" is meant in the strict sense of intuitionistic/constructive logic*.

Each line in the resolution path consists of the name of the rule applied, followed by: the description of how the rule is "instantiated" (i.e. how the condition part is satisfied), the "==>" sign, the conclusion allowed by the "action" part. The conclusion is always either that a candidate can be eliminated (symbolically written as r4c8 ≠ 6 in Sudoku) or that a value must be asserted (symbolically written as r4c8 = 5). When the same rule instantiation justifies several conclusions, they are written on the same line, separated by commas: e.g. r4c8 ≠ 8, r5c8 ≠ 8. Occasionally, the detailed situation at some point in the resolution path (the "resolution state") is displayed so that the presence of the pattern under discussion can be directly checked, but, due to place constraints, this cannot be systematic.

All the resolution paths given in this second edition were obtained with version 1.2 of our general pattern-based CSP solver: *CSP-Rules*[1] (with occasional hand editing for a shorter and/or cleaner appearance), using the CLIPS inference engine (release 6.30), on a MacPro® 2006 running at 2.66 GHz. It was easily supplemented with inpout/output functions specific to Sudoku (making it correspond to version 15d.1.12 of our SudoRules solver), Futoshiki, Kakuro, Map colouring, Numbrix® and Hidato®.

### 1.6. Notations

Throughout this book, we consider an arbitrary, but fixed, finite Constraint Satisfaction Problem. We call it CSP, generically. BRT(CSP) or simply BRT (when there is no ambiguity) refers to its Basic Resolution Theory, RT to any of its resolution theories, $W_n$ [respectively $B_n$, $gW_n$, $gB_n$, $S_pW_n$, $S_pB_n$, $B_pB_n$, …] to its nth whip [respectively braid, g-whip, g-braid, $S_p$-whip, $S_p$-braid, $B_p$-braid, …] resolution theory. The same letters, with no *n* subscript, are used for the associated ratings.

---

[1] See section 17.4 for more information about CSP-Rules.

**Part One**

# LOGICAL FOUNDATIONS

# 2. The role of modelling, illustrated with Sudoku

Before we start with the logical formalisation of a general CSP, the main purpose of this chapter is to show in detail, using the Sudoku example, how some initial modelling choices and/or associated mental or graphical representations can radically change our view of a CSP. Together with consequences of several non-standard modelling choices that will appear throughout this book, it will also illustrate the general epistemological principle that changing our representations of a problem can drastically change its apparent complexity. Almost all of the material here was first introduced in *HLS1*.

It may seem strange to start a part on the "logical foundations" with a chapter on modelling that is almost only about Sudoku. But we mean to insist that, in CSP as in any other domain, modelling choices are the starting point of any good application of any general theory. And most of such choices can only be application specific.

Complementary considerations on modelling a CSP will appear in section 5.11, when we introduce the N-Queens and the N-SudoQueens CSPs, after we have defined our general logical framework and our first resolution rules. See also chapters 14 to 16 for other detailed examples (Futoshiki, Kakuro, Map colouring…).

## 2.1. Symmetries, analogies and supersymmetries

### 2.1.1. Symmetries

Throughout this book, the word "symmetry" is used in the general abstract mathematical sense. A Sudoku symmetry, or symmetry for short, is a transformation that, when applied to *any* valid Sudoku grid, produces a valid Sudoku grid. Any combination of symmetries is a symmetry, there is a null symmetry (that does not change anything) and every symmetry has a reverse; therefore symmetries form a group (in the usual mathematical sense).

Two grids (completed or not) that are related by some symmetry are said to be *essentially equivalent*. The reason is that when the first is solved, its solution and its resolution path can be transposed by the same symmetry to a solution and a resolution path for the second. These abstract notions become very concrete and intuitive as soon as a set of generators for the whole group of symmetries is given.

By definition, any symmetry is then composed of a finite sequence of these generating ones. The simplest set of generators one can consider is composed of two different types of obvious symmetries (see e.g. [Russell 2005]):

– permutations of the numbers: the numerical values of the numbers used to fill the grid are totally irrelevant; they could indeed be replaced by arbitrary symbols; any permutation of the digits (which is just a relabeling of the entries) defines a symmetry of the game; there are obviously 9! = 362,880 such symmetries.

– "geometrical" symmetries of the grid:
  - permutations of individual rows 1, 2, 3;
  - permutations of individual rows 4, 5, 6;
  - permutations of individual rows 7, 8, 9;
  - permutations of triplets of rows ("floors") 1-2-3, 4-5-6 and 7-8-9;
  - symmetry relative to the first diagonal (row-column symmetry).

From these primary geometrical symmetries, others can be deduced:
  - permutations of individual columns 1, 2, 3;
  - permutations of individual columns 4, 5, 6;
  - permutations of individual columns 7, 8, 9;
  - permutations of triplets of columns ("towers") 1-2-3, 4-5-6 and 7-8-9;
  - reflection (left-right symmetry);
  - up-down symmetry;
  - symmetry relative to the second diagonal;
  - ± 90° rotation,
  - and, more generally, any combination of symmetries in the generating set.

As of the writing of *HLS1*, the above-mentioned symmetries had been used mainly to count the number of essentially non-equivalent grids. Expressed in terms of elementary symmetries, two grids (completed or not) are essentially equivalent if there is a sequence of elementary symmetries such that the second is obtained from the first by application of this sequence.

Thus, it has been shown in [Russell 2005] that the number of non-essentially equivalent complete Sudoku grids is 5,472,730,538 – much less than the *a priori* possibly different 6,670,903,752,021,072,936,960 complete grids. But the number of essentially different minimal puzzles is still much greater, its exact value being still unknown (however, see our estimate in chapter 6: $2.55 \times 10^{25}$). The point is that each complete grid is, in the mean, the solution for $4.67 \times 10^{15}$ minimal puzzles.

Later we shall formulate axioms for Sudoku in a logical language and in a way that exhibits all the previous symmetries. In turn, such symmetries in the axioms will lead to symmetries in the logical formulation of our resolution rules. But all the types of symmetries will not be expressed in the same way in these axioms or rules.

Primary symmetries other than row-column will be totally transparent, in that they will make use of variable names (for numbers, rows, columns…) but they will refer to no specific values of these entities.

As for row-column symmetry, in elementary resolution rules, our formalisation will stick to their classical formulation and it will be expressed by the presence of two similar axioms or rules, each of which can be obtained from the other by a simple permutation of the words "row" and "column". As a consequence of this symmetry in the axioms, there will be a meta-symmetry in the theorems and the resolution rules, as expressed by the following intuitively obvious

***meta-theorem 2.1 (informal): for any valid Sudoku resolution rule, the rule deduced from it by permuting systematically the words "row" and "column" is valid and it obviously has the same logical complexity as the original. We shall express this as: the set of valid Sudoku resolution rules is closed under row-column symmetry.***

In more evolved resolution rules, in particular in chain rules, we shall show that a more powerful approach consists of building them only on primary predicates that already take all the symmetries into account.

### 2.1.2. The two canonical coordinate systems on a grid

Let the nine rows be numbered 1, 2, …, 9 from top to bottom. Let the nine columns be numbered 1, 2, …, 9 from left to right. Let the nine blocks and the nine squares inside any fixed block be numbered according to the same scheme, as follows:

$$\begin{array}{ccc} 1 & 2 & 3 \\ 4 & 5 & 6 \\ 7 & 8 & 9 \end{array}$$

Any cell, in "natural" row-column space, can be unambiguously located on the grid via either of its two pairs of coordinates (row, column) or [block, square]. One can therefore consider two coordinate systems on the grid. We call them the two canonical coordinate systems and we write the coordinates of a cell in each of them as (r, c) or as [b, s], respectively.

Change of coordinates F: (r, c) → [b, s] is defined by the following formulæ:
b = block (r, c) = 1 + 3×IP((r – 1)/3) + IP((c - 1)/3);
s = square(r, c) = 1 + 3×mod((r + 2), 3) + mod((c + 2), 3).

Conversely, change of coordinates [b, s] → (r, c) is defined by:
r = row(b, s) = 1 + 3×IP((b - 1)/3) + IP((s - 1)/3);
c = column(b, s) = 1 + 3×mod((b + 2), 3) + mod((s + 2), 3),
where "IP" stands for "integer part" and "mod" for "modulo".

Notice that transformation F: (r, c) → [b, s]: is involutive, i.e. $F^{-1} = F$ or $F \bullet F = Id$ (the identity), where "$F^{-1}$" denotes as usual the inverse of F and "•" denotes function composition.

### 2.1.3. Coordinates and names

Coordinates should not be confused with the various names that can be given to the rows, columns, blocks, squares and cells for displaying purposes. Various displaying conventions can be used (e.g. the chess convention: A1, A2, … G8, G9), but we shall systematically stick to the following one, which we have found the most convenient and which is easier to generalise to any CSP:

– rows are named: r1, r2, r3, r4, r5, r6, r7, r8, r9;

– columns are named: c1, c2, c3, c4, c5, c6, c7, c8, c9;

– cells in natural rc-space are named accordingly, in the obvious way: r1c1, r1c2, …, r9c9;

– blocks are named: b1, b2, b3, b4, b5, b6, b7, b8, b9;

– squares in a block are named: s1, s2, s3, s4, s5, s6, s7, s8, s9;

– as a result, cells in rc-space can also be named: b1s1, b1s2, …, b9s9;

– when needed, numbers are named n1, n2, n3, n4, n5, n6, n7, n8, n9; this will be useful in the next sections when we consider "abstract spaces": row-number, column-number and block-number and we want to name cells in these spaces: r1n1, r1n2… in rn-space; c1n1, c1n2,… in cn-space; b1n1, b1n2,… in bn-space; the reason is that r11, r12… or c11, c12… would be rather obscure and confusing.

Notice that the same lower case letters as for constants will be used for naming variables, but with subscripts, e.g. $r_1$, $b_3$, …; these close conventions should not lead to any confusion between variables and constants. In any case, the risk of confusion is very limited: no variable symbol can appear in the description of any real fact on a real grid and no constant symbol will ever appear in an axiom (except of course in the axioms corresponding to the givens of the puzzle) or a resolution rule.

### 2.1.4. Supersymmetries

Up to now, symmetries relative to the entries (numbers) and "geometrical" symmetries relative to the grid have been considered separately. One of the results of *HLS1* was the elicitation of other symmetries (named *supersymmetries*) that mix numbers, rows and columns. It showed how they translate into relationships between some of the constraints propagation rules, how they entail a new logical classification of these rules, how this allows clearer definitions of the rules themselves and how this leads to introduce new types of chains ("hidden" chains and "supersymmetric" chains) and associated rules.

The main reason for our interest in supersymmetry is the following:

***meta-theorem 2.2 (informal): for any valid Sudoku resolution rule mentioning only numbers, rows and columns (i.e. neither blocks nor squares nor any property referring to such objects), any rule deduced from it by any systematic permutation of the words "number", "row" and "column" is valid and it obviously has the same logical complexity as the original. We shall express this as: the set of valid Sudoku resolution rules is closed under supersymmetry***.

Meta-theorem 2.2 is not intuitively as obvious as meta-theorem 2.1. From a logical point of view, it is nevertheless a straightforward consequence of the subsequent logical formulation of the problem in Multi-Sorted First Order Logic (more on this in chapters 3 and 4). And, from a practical point of view, subtle correspondences between Subset rules become explicit (see chapter 8). If we consider the LatinSquare CSP, the above theorem has a much simpler formulation:
***for any valid LatinSquare resolution rule, any rule deduced from it by a systematic permutation of the words "number", "row" and "column" is valid***.

### 2.1.5. Analogies

Analogies should not be confused with symmetries. There are analogies between rows and blocks (or between columns and blocks) but there is no real symmetry.

This is related to the fact that the two canonical coordinate systems do not share the same properties with respect to the rules of Sudoku. There is a symmetry between the coordinates in the first system (rows and columns) and, relying explicitly on this symmetry, many axioms and rules exist by pairs; but there is no symmetry between the coordinates in the second system (blocks and squares) so that transposing rules from the first system to the second would be meaningless.

There is nevertheless a partial analogy between rows (or columns) and blocks, captured by the following informal

***meta-theorem 2.3 (informal): for any valid Sudoku resolution rule mentioning only numbers, rows and columns (i.e. neither blocks nor squares nor any property referring to such objects), if this rule displays a systematic symmetry between rows and columns but it can be proved without using the axiom on columns, then the rule deduced from it by systematically replacing the word "row" by "block" and the word "column" by "square" is valid and it obviously has the same logical complexity as the original one. We shall express this as: the set of valid Sudoku resolution rules is closed under analogy.***

What the phrases "systematic symmetry between rows and columns" and "proved without using the axiom on columns" mean will be defined precisely in chapter 3.

**2.2. Introducing the four 2D spaces: rc, rn, cn and bn**

To better visualise the symmetries, supersymmetries and analogies defined in the previous section, we introduce three 2D spaces and their graphical representations. The latter can be grouped with the usual one to form an extended Sudoku board (Figure 2.3). These new representations were first introduced in *HLS1*. How to build and use them was explained in detail in *HLS2*; we do not repeat it here.

In the Subset rules of chapter 8, they will be used to illustrate how apparently complex familiar rules (such as X-wing, Swordfish or Jellyfish) are no more than the supersymmetric versions of obvious ones (Naked-Pairs, Naked-Triplets and Naked-Quads, respectively); all this was already in *HLS1*, where they have also been the basis for the notion of hidden chains and associated resolution rules.

In this book, however, the main role of these new spaces and representations will be to justify intuitively the introduction of additional CSP variables.

*2.2.1. Additional graphical representations of a puzzle*

In addition to the standard "natural" row-column space (or rc-space), we consider three new "abstract" spaces: row-number, column-number and block-number. In the sequel, these four spaces will also be called respectively rc-space, rn-space, cn-space and bn-space and "cells" in these four spaces will be called rc-cells, rn-cells, cn-cells and bn-cells. As for their graphical representations, when they are displayed together, they are aligned so that rows in the first two coincide and columns in the first and the third coincide (cn space is thus displayed as nc).

When it comes to candidates, the reason for considering rn-cell with coordinates (r, n) in rn-space is that it will contain all the possibilities (all the possible columns) for the unique instance of number n that must occur in row r; similarly, the reason for considering cn-cell with coordinates (c, n) in cn-space is that it will contain all the possibilities (all the possible rows) for the unique instance of number n that must occur in column c; finally, the reason for considering bn-cell with coordinates (b, n) in bn-space is that it will contain all the possibilities (all the possible squares) for the unique instance of number n that must occur in block b.

At any point in the resolution process, all the data in the grid (values and candidates) can be displayed in any of these four representations. We insist that each of them displays exactly the same logical information content – or, to say it more formally: they correspond to the same underlying set of ground atomic formulæ in the (basically 3D) logical language that will be introduced later. They should be considered only as different visual supports for symmetry, supersymmetry and analogy, in the sense that it is easier to detect some patterns in some representations than in others, as illustrated by several chapters in this book and in *HLS*.

The correspondences are straightforward and are given by the equivalences:
– Boolean symbol True is present in nrc-cell (n, r, c), (3D view, to be discussed in section 2.4),
  – number n is present in rc-cell (r, c), (standard view),
  – column c is present in rn-cell (r, n),
  – row r is in present in cn-cell (c, n),
  – square s is in present in bn-cell (b, n), where (r, c) = [b, s].

Notice that pseudo blocks (i.e. groups of 3×3 rn, cn or bn cells) have no meaning in the new rn, cn or bn representations (this is why we do not mark them with thick borders): only constraints valid for Latin Squares can be directly propagated in rn or cn spaces (as will be proved in chapter 3). Moreover, links in bn-space cannot use the number coordinate.

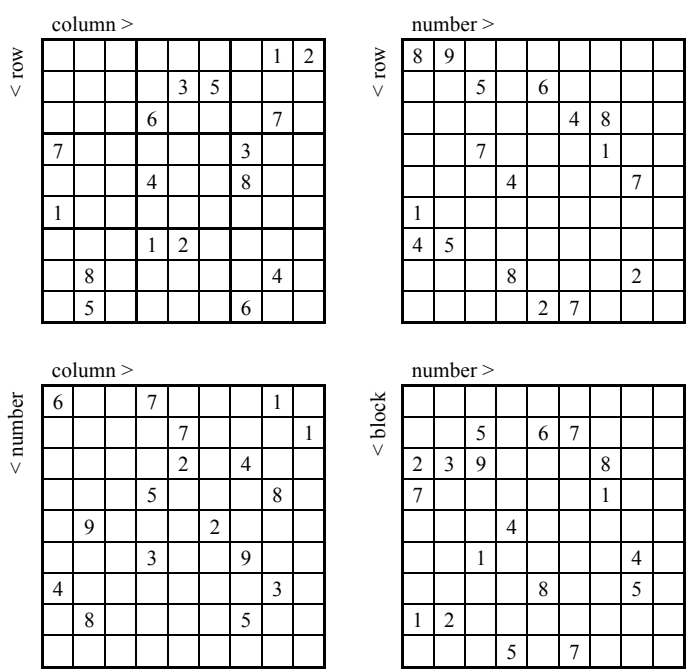

***Figure 2.1.*** *Same puzzle Royle17#3 as in Figure 1.1, but viewed in the four different representation spaces (rc, rn, cn, bn)*

Generating these new grid representations by hand is easy as long as we consider only values, as in Figure 2.1, but it is tedious when it comes to the candidates.

Nevertheless, with some practice, it is relatively simple to apply the above stated equivalences (see *HLS*). Moreover, programming a spreadsheet computing the three new grids and their candidates automatically from the first is an easy exercise.

Let us illustrate these new representations with the example given in Figure 1.1 (puzzle Royle17#3). Starting from the standard form of the puzzle, we can first display its entries in the standard grid and in the three new grids of Figure 2.1. After applying all the elementary constraints propagation rules in rc-space, we get the usual representation of the resolution state in rc-space (Figure 1.2).

Now, suppose we generate the full rn, cn and bn representations with candidates. For our puzzle, there is nothing particularly appealing in the rn and bn representations, so we skip them. But a surprise is awaiting us with its cn representation (Figure 2.2). It makes it obvious that there is a cn-cell (c7n1) with only one possibility left: the unique instance of number 1 that must appear somewhere in column 7 is in fact confined to row 8 (i.e. cn-cell c7n1 has only one row candidate: r8).

|    | c1 | c2 | c3 | c4 | c5 | c6 | c7 | c8 | c9 |    |
|----|----|----|----|----|----|----|----|----|----|----|
| n1 | r6 | r2 r3 r8 r9 | r2 r3 r8 r9 | r7 | r3 r4 r5 | r3 r4 r5 | **r8** | r1 | r4 r5 r8 r9 | n1 |
| n2 | r2 r3 r5 r8 r9 | r2 r3 r4 r5 r6 | r2 r3 r4 r5 r6 r8 r9 | r2 r4 r6 | r7 | r3 r4 r5 r6 | r6 r8 | r4 r5 r6 r9 | r1 | n2 |
| n3 | r1 r3 r5 r7 r8 r9 | r1 r3 r5 r6 r7 | r1 r3 r5 r6 r7 r8 r9 | r6 r8 r9 | r2 | r5 r6 r7 r8 r9 | r4 | r7 r9 | r3 r7 r8 r9 | n3 |
| n4 | r1 r2 r3 r7 r9 | r1 r2 r3 r4 r6 r7 | r1 r2 r3 r4 r6 r7 r9 | r5 | r1 r3 r9 | r1 r3 r7 r9 | r1 r2 r3 r6 | r8 | r2 r3 r4 r6 | n4 |
| n5 | r1 r3 r5 | r9 | r1 r3 r4 r5 r6 | r4 r6 r8 | r4 r5 r6 r8 | r2 | r1 r3 r6 r7 r8 | r4 r5 r6 r7 | r3 r4 r5 r6 r7 r8 | n5 |
| n6 | r1 r2 r5 r7 r8 | r1 r2 r4 r5 r6 r7 | r1 r2 r4 r5 r6 r7 r8 | r3 | r4 r5 r6 r8 | r4 r5 r6 r7 r8 | r9 | r2 r4 r5 r6 | r2 r4 r5 r6 | n6 |
| n7 | r4 | r1 r2 r7 | r1 r2 r7 r8 r9 | r1 r2 r6 r8 r9 | r1 r5 r6 r8 r9 | r1 r5 r6 r7 r8 r9 | r6 r7 r8 | r3 | r5 r6 r7 r8 r9 | n7 |
| n8 | r1 r2 r3 | r8 | r1 r2 r3 r4 r6 | r1 r2 r4 r6 r9 | r1 r3 r4 r6 r9 | r1 r3 r4 r6 r7 r9 | r5 | r2 r7 r9 | r2 r3 r7 r9 | n8 |
| n9 | r1 r2 r3 r5 r7 r8 r9 | r1 r2 r3 r4 r5 r6 r7 | r1 r2 r3 r4 r5 r6 r7 r8 r9 | r1 r2 r4 r8 r9 | r1 r3 r4 r5 r6 r8 r9 | r1 r3 r4 r5 r6 r7 r8 r9 | r1 r2 r3 r6 r7 r8 | r2 r4 r5 r6 r7 | r2 r3 r4 r5 r6 r7 r8 r9 | n9 |
|    | c1 | c2 | c3 | c4 | c5 | c6 | c7 | c8 | c9 |    |

**Figure 2.2.** *Same puzzle Royle17#3 as in Figure 1.2, but viewed in cn-space*

As an example that the groups of 3×3 contiguous cn-cells have no meaning, we can see that there are many of these pseudo-blocks in which the same candidate (row) appears two or more times.

Now, it appears that, if we had considered more attentively the standard rc representation with candidates (Figure 1.2 of the Introduction), we could have seen that, in column c7, there is only one row (row r8) having number 1 among its candidates. Therefore, the unique instance of number 1 that must be found somewhere in column c7 has only one possibility left of finding its place in this column and that is in row r8. But the difference is, this cannot be seen in rc-space by looking only at one rc-cell (namely r8c7) since it still has five candidates: 1, 2, 5, 7 and 9. What the representation in cn-space provides is the possibility of *detecting locally* this forced value by looking at a single cn-cell, while in "natural" rc-space we must examine all the nine rc-cells of column c7. This is a very elementary example of how rn, cn or bn spaces can be used in practice.

This is our first example of a "Hidden-Single" (HS) in a column. Notice that the phrase "hidden single in a column" suggests properly that, in column c7, cell r8c7 has a single possible value but that this fact is hidden, i.e. is not visible by looking only at the candidates for this cell in the usual rc-representation. Of course, one can also find Hidden-Singles in rows or in blocks. Actually, this Royle17#3 puzzle can be solved using only these types of Hidden-Singles (in addition, of course, to Naked Singles and the elementary constraints propagation rules).

*Graphically, in the standard rc representation, spotting a Hidden-Single-in-a-row [respectively in-a-column, in-a-block] for some Number n supposes that one checks that the other eight cells in this row [resp. this column, this block] do not contain n among their candidates. In the new rn [resp. cn, bn] representation, all that is needed is checking that one cell has a single possibility left.* Thus, even in very elementary cases, the new representations simplify the detection job.

Now, a few comments about these new graphical representations are in order. Should one consider them as a practical basis for human solving? There will probably never be any general agreement on this point. Our personal opinion is that, given the additional paperwork needed for building and maintaining the four representations in parallel, they are not very useful for easy puzzles; but, one can easily imagine a computerised interface that maintains the coherency between the four grids (any time a candidate is eliminated from one of them or a value is asserted in one of them, this information is transferred to the others). Moreover, there are many difficult puzzles that become easier to solve if we use such representations (and rules based on them): see *HLS*, a significant part of which was based on symmetries, supersymmetries and "hidden" structures.

Anyway, in the present book, they will mainly be considered as a step towards the introduction of new CSP variables and as a representation system for them.

|    | c1 | c2 | c3 | c4 | c5 | c6 | c7 | c8 | c9 |    |
|----|----|----|----|----|----|----|----|----|----|----|
| r1 | n1 n2 n3 n4 n5 n6 n7 n8 n9 | n1 n2 n3 n4 n5 n6 n7 n8 n9 | n1 n2 n3 n4 n5 n6 n7 n8 n9 | n1 n2 n3 n4 n5 n6 n7 n8 n9 | n1 n2 n3 n4 n5 n6 n7 n8 n9 | n1 n2 n3 n4 n5 n6 n7 n8 n9 | n1 n2 n3 n4 n5 n6 n7 n8 n9 | n1 n2 n3 n4 n5 n6 n7 n8 n9 | n1 n2 n3 n4 n5 n6 n7 n8 n9 | r1 |
| r2 | n1 n2 n3 n4 n5 n6 n7 n8 n9 | n1 n2 n3 n4 n5 n6 n7 n8 n9 | n1 n2 n3 n4 n5 n6 n7 n8 n9 | n1 n2 n3 n4 n5 n6 n7 n8 n9 | n1 n2 n3 n4 n5 n6 n7 n8 n9 | n1 n2 n3 n4 n5 n6 n7 n8 n9 | n1 n2 n3 n4 n5 n6 n7 n8 n9 | n1 n2 n3 n4 n5 n6 n7 n8 n9 | n1 n2 n3 n4 n5 n6 n7 n8 n9 | r2 |
| r3 | n1 n2 n3 n4 n5 n6 n7 n8 n9 | n1 n2 n3 n4 n5 n6 n7 n8 n9 | n1 n2 n3 n4 n5 n6 n7 n8 n9 | n1 n2 n3 n4 n5 n6 n7 n8 n9 | n1 n2 n3 n4 n5 n6 n7 n8 n9 | n1 n2 n3 n4 n5 n6 n7 n8 n9 | n1 n2 n3 n4 n5 n6 n7 n8 n9 | n1 n2 n3 n4 n5 n6 n7 n8 n9 | n1 n2 n3 n4 n5 n6 n7 n8 n9 | r3 |
| r4 | n1 n2 n3 n4 n5 n6 n7 n8 n9 | n1 n2 n3 n4 n5 n6 n7 n8 n9 | n1 n2 n3 n4 n5 n6 n7 n8 n9 | n1 n2 n3 n4 n5 n6 n7 n8 n9 | n1 n2 n3 n4 n5 n6 n7 n8 n9 | n1 n2 n3 n4 n5 n6 n7 n8 n9 | n1 n2 n3 n4 n5 n6 n7 n8 n9 | n1 n2 n3 n4 n5 n6 n7 n8 n9 | n1 n2 n3 n4 n5 n6 n7 n8 n9 | r4 |
| r5 | n1 n2 n3 n4 n5 n6 n7 n8 n9 | n1 n2 n3 n4 n5 n6 n7 n8 n9 | n1 n2 n3 n4 n5 n6 n7 n8 n9 | n1 n2 n3 n4 n5 n6 n7 n8 n9 | n1 n2 n3 n4 n5 n6 n7 n8 n9 | n1 n2 n3 n4 n5 n6 n7 n8 n9 | n1 n2 n3 n4 n5 n6 n7 n8 n9 | n1 n2 n3 n4 n5 n6 n7 n8 n9 | n1 n2 n3 n4 n5 n6 n7 n8 n9 | r5 |
| r6 | n1 n2 n3 n4 n5 n6 n7 n8 n9 | n1 n2 n3 n4 n5 n6 n7 n8 n9 | n1 n2 n3 n4 n5 n6 n7 n8 n9 | n1 n2 n3 n4 n5 n6 n7 n8 n9 | n1 n2 n3 n4 n5 n6 n7 n8 n9 | n1 n2 n3 n4 n5 n6 n7 n8 n9 | n1 n2 n3 n4 n5 n6 n7 n8 n9 | n1 n2 n3 n4 n5 n6 n7 n8 n9 | n1 n2 n3 n4 n5 n6 n7 n8 n9 | r6 |
| r7 | n1 n2 n3 n4 n5 n6 n7 n8 n9 | n1 n2 n3 n4 n5 n6 n7 n8 n9 | n1 n2 n3 n4 n5 n6 n7 n8 n9 | n1 n2 n3 n4 n5 n6 n7 n8 n9 | n1 n2 n3 n4 n5 n6 n7 n8 n9 | n1 n2 n3 n4 n5 n6 n7 n8 n9 | n1 n2 n3 n4 n5 n6 n7 n8 n9 | n1 n2 n3 n4 n5 n6 n7 n8 n9 | n1 n2 n3 n4 n5 n6 n7 n8 n9 | r7 |
| r8 | n1 n2 n3 n4 n5 n6 n7 n8 n9 | n1 n2 n3 n4 n5 n6 n7 n8 n9 | n1 n2 n3 n4 n5 n6 n7 n8 n9 | n1 n2 n3 n4 n5 n6 n7 n8 n9 | n1 n2 n3 n4 n5 n6 n7 n8 n9 | n1 n2 n3 n4 n5 n6 n7 n8 n9 | n1 n2 n3 n4 n5 n6 n7 n8 n9 | n1 n2 n3 n4 n5 n6 n7 n8 n9 | n1 n2 n3 n4 n5 n6 n7 n8 n9 | r8 |
| r9 | n1 n2 n3 n4 n5 n6 n7 n8 n9 | n1 n2 n3 n4 n5 n6 n7 n8 n9 | n1 n2 n3 n4 n5 n6 n7 n8 n9 | n1 n2 n3 n4 n5 n6 n7 n8 n9 | n1 n2 n3 n4 n5 n6 n7 n8 n9 | n1 n2 n3 n4 n5 n6 n7 n8 n9 | n1 n2 n3 n4 n5 n6 n7 n8 n9 | n1 n2 n3 n4 n5 n6 n7 n8 n9 | n1 n2 n3 n4 n5 n6 n7 n8 n9 | r9 |

|    | c1 | c2 | c3 | c4 | c5 | c6 | c7 | c8 | c9 |    |
|----|----|----|----|----|----|----|----|----|----|----|
| n1 | r1 r2 r3 r4 r5 r6 r7 r8 r9 | r1 r2 r3 r4 r5 r6 r7 r8 r9 | r1 r2 r3 r4 r5 r6 r7 r8 r9 | r1 r2 r3 r4 r5 r6 r7 r8 r9 | r1 r2 r3 r4 r5 r6 r7 r8 r9 | r1 r2 r3 r4 r5 r6 r7 r8 r9 | r1 r2 r3 r4 r5 r6 r7 r8 r9 | r1 r2 r3 r4 r5 r6 r7 r8 r9 | r1 r2 r3 r4 r5 r6 r7 r8 r9 | n1 |
| n2 | r1 r2 r3 r4 r5 r6 r7 r8 r9 | r1 r2 r3 r4 r5 r6 r7 r8 r9 | r1 r2 r3 r4 r5 r6 r7 r8 r9 | r1 r2 r3 r4 r5 r6 r7 r8 r9 | r1 r2 r3 r4 r5 r6 r7 r8 r9 | r1 r2 r3 r4 r5 r6 r7 r8 r9 | r1 r2 r3 r4 r5 r6 r7 r8 r9 | r1 r2 r3 r4 r5 r6 r7 r8 r9 | r1 r2 r3 r4 r5 r6 r7 r8 r9 | n2 |
| n3 | r1 r2 r3 r4 r5 r6 r7 r8 r9 | r1 r2 r3 r4 r5 r6 r7 r8 r9 | r1 r2 r3 r4 r5 r6 r7 r8 r9 | r1 r2 r3 r4 r5 r6 r7 r8 r9 | r1 r2 r3 r4 r5 r6 r7 r8 r9 | r1 r2 r3 r4 r5 r6 r7 r8 r9 | r1 r2 r3 r4 r5 r6 r7 r8 r9 | r1 r2 r3 r4 r5 r6 r7 r8 r9 | r1 r2 r3 r4 r5 r6 r7 r8 r9 | n3 |
| n4 | r1 r2 r3 r4 r5 r6 r7 r8 r9 | r1 r2 r3 r4 r5 r6 r7 r8 r9 | r1 r2 r3 r4 r5 r6 r7 r8 r9 | r1 r2 r3 r4 r5 r6 r7 r8 r9 | r1 r2 r3 r4 r5 r6 r7 r8 r9 | r1 r2 r3 r4 r5 r6 r7 r8 r9 | r1 r2 r3 r4 r5 r6 r7 r8 r9 | r1 r2 r3 r4 r5 r6 r7 r8 r9 | r1 r2 r3 r4 r5 r6 r7 r8 r9 | n4 |
| n5 | r1 r2 r3 r4 r5 r6 r7 r8 r9 | r1 r2 r3 r4 r5 r6 r7 r8 r9 | r1 r2 r3 r4 r5 r6 r7 r8 r9 | r1 r2 r3 r4 r5 r6 r7 r8 r9 | r1 r2 r3 r4 r5 r6 r7 r8 r9 | r1 r2 r3 r4 r5 r6 r7 r8 r9 | r1 r2 r3 r4 r5 r6 r7 r8 r9 | r1 r2 r3 r4 r5 r6 r7 r8 r9 | r1 r2 r3 r4 r5 r6 r7 r8 r9 | n5 |
| n6 | r1 r2 r3 r4 r5 r6 r7 r8 r9 | r1 r2 r3 r4 r5 r6 r7 r8 r9 | r1 r2 r3 r4 r5 r6 r7 r8 r9 | r1 r2 r3 r4 r5 r6 r7 r8 r9 | r1 r2 r3 r4 r5 r6 r7 r8 r9 | r1 r2 r3 r4 r5 r6 r7 r8 r9 | r1 r2 r3 r4 r5 r6 r7 r8 r9 | r1 r2 r3 r4 r5 r6 r7 r8 r9 | r1 r2 r3 r4 r5 r6 r7 r8 r9 | n6 |
| n7 | r1 r2 r3 r4 r5 r6 r7 r8 r9 | r1 r2 r3 r4 r5 r6 r7 r8 r9 | r1 r2 r3 r4 r5 r6 r7 r8 r9 | r1 r2 r3 r4 r5 r6 r7 r8 r9 | r1 r2 r3 r4 r5 r6 r7 r8 r9 | r1 r2 r3 r4 r5 r6 r7 r8 r9 | r1 r2 r3 r4 r5 r6 r7 r8 r9 | r1 r2 r3 r4 r5 r6 r7 r8 r9 | r1 r2 r3 r4 r5 r6 r7 r8 r9 | n7 |
| n8 | r1 r2 r3 r4 r5 r6 r7 r8 r9 | r1 r2 r3 r4 r5 r6 r7 r8 r9 | r1 r2 r3 r4 r5 r6 r7 r8 r9 | r1 r2 r3 r4 r5 r6 r7 r8 r9 | r1 r2 r3 r4 r5 r6 r7 r8 r9 | r1 r2 r3 r4 r5 r6 r7 r8 r9 | r1 r2 r3 r4 r5 r6 r7 r8 r9 | r1 r2 r3 r4 r5 r6 r7 r8 r9 | r1 r2 r3 r4 r5 r6 r7 r8 r9 | n8 |
| n9 | r1 r2 r3 r4 r5 r6 r7 r8 r9 | r1 r2 r3 r4 r5 r6 r7 r8 r9 | r1 r2 r3 r4 r5 r6 r7 r8 r9 | r1 r2 r3 r4 r5 r6 r7 r8 r9 | r1 r2 r3 r4 r5 r6 r7 r8 r9 | r1 r2 r3 r4 r5 r6 r7 r8 r9 | r1 r2 r3 r4 r5 r6 r7 r8 r9 | r1 r2 r3 r4 r5 r6 r7 r8 r9 | r1 r2 r3 r4 r5 r6 r7 r8 r9 | n9 |

***Figure 2.3.*** *The Extended Sudoku Board, with the four rc, rn, cn and bn spaces; each cell in this Extended Board represents a CSP variable of the extended list.*

|    | n1 | n2 | n3 | n4 | n5 | n6 | n7 | n8 | n9 |    |
|----|----|----|----|----|----|----|----|----|----|----|
| r1 | c1 c2 c3 c4 c5 c6 c7 c8 c9 | c1 c2 c3 c4 c5 c6 c7 c8 c9 | c1 c2 c3 c4 c5 c6 c7 c8 c9 | c1 c2 c3 c4 c5 c6 c7 c8 c9 | c1 c2 c3 c4 c5 c6 c7 c8 c9 | c1 c2 c3 c4 c5 c6 c7 c8 c9 | c1 c2 c3 c4 c5 c6 c7 c8 c9 | c1 c2 c3 c4 c5 c6 c7 c8 c9 | c1 c2 c3 c4 c5 c6 c7 c8 c9 | r1 |
| r2 | c1 c2 c3 c4 c5 c6 c7 c8 c9 | c1 c2 c3 c4 c5 c6 c7 c8 c9 | c1 c2 c3 c4 c5 c6 c7 c8 c9 | c1 c2 c3 c4 c5 c6 c7 c8 c9 | c1 c2 c3 c4 c5 c6 c7 c8 c9 | c1 c2 c3 c4 c5 c6 c7 c8 c9 | c1 c2 c3 c4 c5 c6 c7 c8 c9 | c1 c2 c3 c4 c5 c6 c7 c8 c9 | c1 c2 c3 c4 c5 c6 c7 c8 c9 | r2 |
| r3 | c1 c2 c3 c4 c5 c6 c7 c8 c9 | c1 c2 c3 c4 c5 c6 c7 c8 c9 | c1 c2 c3 c4 c5 c6 c7 c8 c9 | c1 c2 c3 c4 c5 c6 c7 c8 c9 | c1 c2 c3 c4 c5 c6 c7 c8 c9 | c1 c2 c3 c4 c5 c6 c7 c8 c9 | c1 c2 c3 c4 c5 c6 c7 c8 c9 | c1 c2 c3 c4 c5 c6 c7 c8 c9 | c1 c2 c3 c4 c5 c6 c7 c8 c9 | r3 |
| r4 | c1 c2 c3 c4 c5 c6 c7 c8 c9 | c1 c2 c3 c4 c5 c6 c7 c8 c9 | c1 c2 c3 c4 c5 c6 c7 c8 c9 | c1 c2 c3 c4 c5 c6 c7 c8 c9 | c1 c2 c3 c4 c5 c6 c7 c8 c9 | c1 c2 c3 c4 c5 c6 c7 c8 c9 | c1 c2 c3 c4 c5 c6 c7 c8 c9 | c1 c2 c3 c4 c5 c6 c7 c8 c9 | c1 c2 c3 c4 c5 c6 c7 c8 c9 | r4 |
| r5 | c1 c2 c3 c4 c5 c6 c7 c8 c9 | c1 c2 c3 c4 c5 c6 c7 c8 c9 | c1 c2 c3 c4 c5 c6 c7 c8 c9 | c1 c2 c3 c4 c5 c6 c7 c8 c9 | c1 c2 c3 c4 c5 c6 c7 c8 c9 | c1 c2 c3 c4 c5 c6 c7 c8 c9 | c1 c2 c3 c4 c5 c6 c7 c8 c9 | c1 c2 c3 c4 c5 c6 c7 c8 c9 | c1 c2 c3 c4 c5 c6 c7 c8 c9 | r5 |
| r6 | c1 c2 c3 c4 c5 c6 c7 c8 c9 | c1 c2 c3 c4 c5 c6 c7 c8 c9 | c1 c2 c3 c4 c5 c6 c7 c8 c9 | c1 c2 c3 c4 c5 c6 c7 c8 c9 | c1 c2 c3 c4 c5 c6 c7 c8 c9 | c1 c2 c3 c4 c5 c6 c7 c8 c9 | c1 c2 c3 c4 c5 c6 c7 c8 c9 | c1 c2 c3 c4 c5 c6 c7 c8 c9 | c1 c2 c3 c4 c5 c6 c7 c8 c9 | r6 |
| r7 | c1 c2 c3 c4 c5 c6 c7 c8 c9 | c1 c2 c3 c4 c5 c6 c7 c8 c9 | c1 c2 c3 c4 c5 c6 c7 c8 c9 | c1 c2 c3 c4 c5 c6 c7 c8 c9 | c1 c2 c3 c4 c5 c6 c7 c8 c9 | c1 c2 c3 c4 c5 c6 c7 c8 c9 | c1 c2 c3 c4 c5 c6 c7 c8 c9 | c1 c2 c3 c4 c5 c6 c7 c8 c9 | c1 c2 c3 c4 c5 c6 c7 c8 c9 | r7 |
| r8 | c1 c2 c3 c4 c5 c6 c7 c8 c9 | c1 c2 c3 c4 c5 c6 c7 c8 c9 | c1 c2 c3 c4 c5 c6 c7 c8 c9 | c1 c2 c3 c4 c5 c6 c7 c8 c9 | c1 c2 c3 c4 c5 c6 c7 c8 c9 | c1 c2 c3 c4 c5 c6 c7 c8 c9 | c1 c2 c3 c4 c5 c6 c7 c8 c9 | c1 c2 c3 c4 c5 c6 c7 c8 c9 | c1 c2 c3 c4 c5 c6 c7 c8 c9 | r8 |
| r9 | c1 c2 c3 c4 c5 c6 c7 c8 c9 | c1 c2 c3 c4 c5 c6 c7 c8 c9 | c1 c2 c3 c4 c5 c6 c7 c8 c9 | c1 c2 c3 c4 c5 c6 c7 c8 c9 | c1 c2 c3 c4 c5 c6 c7 c8 c9 | c1 c2 c3 c4 c5 c6 c7 c8 c9 | c1 c2 c3 c4 c5 c6 c7 c8 c9 | c1 c2 c3 c4 c5 c6 c7 c8 c9 | c1 c2 c3 c4 c5 c6 c7 c8 c9 | r9 |
|    | n1 | n2 | n3 | n4 | n5 | n6 | n7 | n8 | n9 |    |

|    | n1 | n2 | n3 | n4 | n5 | n6 | n7 | n8 | n9 |    |
|----|----|----|----|----|----|----|----|----|----|----|
| b1 | s1 s2 s3 s4 s5 s6 s7 s8 s9 | s1 s2 s3 s4 s5 s6 s7 s8 s9 | s1 s2 s3 s4 s5 s6 s7 s8 s9 | s1 s2 s3 s4 s5 s6 s7 s8 s9 | s1 s2 s3 s4 s5 s6 s7 s8 s9 | s1 s2 s3 s4 s5 s6 s7 s8 s9 | s1 s2 s3 s4 s5 s6 s7 s8 s9 | s1 s2 s3 s4 s5 s6 s7 s8 s9 | s1 s2 s3 s4 s5 s6 s7 s8 s9 | b1 |
| b2 | s1 s2 s3 s4 s5 s6 s7 s8 s9 | s1 s2 s3 s4 s5 s6 s7 s8 s9 | s1 s2 s3 s4 s5 s6 s7 s8 s9 | s1 s2 s3 s4 s5 s6 s7 s8 s9 | s1 s2 s3 s4 s5 s6 s7 s8 s9 | s1 s2 s3 s4 s5 s6 s7 s8 s9 | s1 s2 s3 s4 s5 s6 s7 s8 s9 | s1 s2 s3 s4 s5 s6 s7 s8 s9 | s1 s2 s3 s4 s5 s6 s7 s8 s9 | b2 |
| b3 | s1 s2 s3 s4 s5 s6 s7 s8 s9 | s1 s2 s3 s4 s5 s6 s7 s8 s9 | s1 s2 s3 s4 s5 s6 s7 s8 s9 | s1 s2 s3 s4 s5 s6 s7 s8 s9 | s1 s2 s3 s4 s5 s6 s7 s8 s9 | s1 s2 s3 s4 s5 s6 s7 s8 s9 | s1 s2 s3 s4 s5 s6 s7 s8 s9 | s1 s2 s3 s4 s5 s6 s7 s8 s9 | s1 s2 s3 s4 s5 s6 s7 s8 s9 | b3 |
| b4 | s1 s2 s3 s4 s5 s6 s7 s8 s9 | s1 s2 s3 s4 s5 s6 s7 s8 s9 | s1 s2 s3 s4 s5 s6 s7 s8 s9 | s1 s2 s3 s4 s5 s6 s7 s8 s9 | s1 s2 s3 s4 s5 s6 s7 s8 s9 | s1 s2 s3 s4 s5 s6 s7 s8 s9 | s1 s2 s3 s4 s5 s6 s7 s8 s9 | s1 s2 s3 s4 s5 s6 s7 s8 s9 | s1 s2 s3 s4 s5 s6 s7 s8 s9 | b4 |
| b5 | s1 s2 s3 s4 s5 s6 s7 s8 s9 | s1 s2 s3 s4 s5 s6 s7 s8 s9 | s1 s2 s3 s4 s5 s6 s7 s8 s9 | s1 s2 s3 s4 s5 s6 s7 s8 s9 | s1 s2 s3 s4 s5 s6 s7 s8 s9 | s1 s2 s3 s4 s5 s6 s7 s8 s9 | s1 s2 s3 s4 s5 s6 s7 s8 s9 | s1 s2 s3 s4 s5 s6 s7 s8 s9 | s1 s2 s3 s4 s5 s6 s7 s8 s9 | b5 |
| b6 | s1 s2 s3 s4 s5 s6 s7 s8 s9 | s1 s2 s3 s4 s5 s6 s7 s8 s9 | s1 s2 s3 s4 s5 s6 s7 s8 s9 | s1 s2 s3 s4 s5 s6 s7 s8 s9 | s1 s2 s3 s4 s5 s6 s7 s8 s9 | s1 s2 s3 s4 s5 s6 s7 s8 s9 | s1 s2 s3 s4 s5 s6 s7 s8 s9 | s1 s2 s3 s4 s5 s6 s7 s8 s9 | s1 s2 s3 s4 s5 s6 s7 s8 s9 | b6 |
| b7 | s1 s2 s3 s4 s5 s6 s7 s8 s9 | s1 s2 s3 s4 s5 s6 s7 s8 s9 | s1 s2 s3 s4 s5 s6 s7 s8 s9 | s1 s2 s3 s4 s5 s6 s7 s8 s9 | s1 s2 s3 s4 s5 s6 s7 s8 s9 | s1 s2 s3 s4 s5 s6 s7 s8 s9 | s1 s2 s3 s4 s5 s6 s7 s8 s9 | s1 s2 s3 s4 s5 s6 s7 s8 s9 | s1 s2 s3 s4 s5 s6 s7 s8 s9 | b7 |
| b8 | s1 s2 s3 s4 s5 s6 s7 s8 s9 | s1 s2 s3 s4 s5 s6 s7 s8 s9 | s1 s2 s3 s4 s5 s6 s7 s8 s9 | s1 s2 s3 s4 s5 s6 s7 s8 s9 | s1 s2 s3 s4 s5 s6 s7 s8 s9 | s1 s2 s3 s4 s5 s6 s7 s8 s9 | s1 s2 s3 s4 s5 s6 s7 s8 s9 | s1 s2 s3 s4 s5 s6 s7 s8 s9 | s1 s2 s3 s4 s5 s6 s7 s8 s9 | b8 |
| b9 | s1 s2 s3 s4 s5 s6 s7 s8 s9 | s1 s2 s3 s4 s5 s6 s7 s8 s9 | s1 s2 s3 s4 s5 s6 s7 s8 s9 | s1 s2 s3 s4 s5 s6 s7 s8 s9 | s1 s2 s3 s4 s5 s6 s7 s8 s9 | s1 s2 s3 s4 s5 s6 s7 s8 s9 | s1 s2 s3 s4 s5 s6 s7 s8 s9 | s1 s2 s3 s4 s5 s6 s7 s8 s9 | s1 s2 s3 s4 s5 s6 s7 s8 s9 | b9 |
|    | n1 | n2 | n3 | n4 | n5 | n6 | n7 | n8 | n9 |    |

*2.2.2. Extended Sudoku Board*

As several examples in *HLS* have shown, especially when we deal with chains, the rn, cn and bn spaces allow to describe simple "hidden" patterns and rules that would need much more complex descriptions in the standard rc-space. In order to facilitate their use, the rn, cn and bn representations can be grouped with the standard one into the Extended Sudoku Board of Figure 2.3. Notice that these representations do not replace the standard one; they are added to it, so that the four representations, when placed in the proper relative positions, form an extended board. In order to avoid confusion between numbers, rows and columns, in this extended board we tend to use systematically their full names: n1, n2, …; r1, r2, …; c1, c2, … But, when an example uses only the rc-space, we may be lax on this.

## 2.3. CSP variables associated with the rc, rn, cn and bn cells

What is more important for the present book is that, ***corresponding to the full set of four 2D views, one can define an extended set of CSP variables*** (with cardinality 324 instead of 81): in addition to all the $Xr°c°$ as before, one can now introduce all the $Xr°n°$, $Xc°n°$ and $Xb°n°$ for n° in {n1, n2, n3, n4, n5, n6, n7, n8, n9}, r° in {r1, r2, r3, r4, r5, r6, r7, r8, r9}, c° in {c1, c2, c3, c4, c5, c6, c7, c8, c9} and b° in {b1, b2, b3, b4, b5, b6, b7, b8, b9}. And one has the following obvious interpretation:

***The Extended Sudoku Board represents the extended set of CSP variables for Sudoku; and, at any stage in the resolution process, the content of each cell represents the set of still possible values (the candidates) for the corresponding CSP variable***.

***The original CSP can now be reformulated in a very different way: find a value for each of these 324 CSP variables such that***, for each n°, r°, c°, b°, s° with (r°, c°) = [b°, s°], one has: $Xr°c° = n° \Leftrightarrow Xr°n° = c° \Leftrightarrow Xc°n° = r° \Leftrightarrow Xb°n° = s°$.

From a logical point of view, there is nothing really new, only obvious rewritings of the initial natural language constraints with redundant CSP variables. One may therefore wonder whether introducing such new variables and constraints can be of any practical use. All this book will show that it is, but part of the answer is already given, at the most intuitive and elementary level, by our analysis of the Hidden Single rule in the example of Figure 2.1: written with the new variables, this rule appears as a mere Naked Single rule. Thus, a very straightforward extension of the original set of CSP variables is enough to suggest new resolution rules or to extend the scope of the existing ones.

Moreover, ***this apparently innocuous method is indeed very powerful***, even at this basic level: only very few minimal Sudoku puzzles can be solved using

Elementary Constraints Propagation and Naked Singles; but 29% of the minimal puzzles (in unbiased statistics) can be solved if we add Hidden Singles (for detailed statistics, see *HLS* or chapter 6 of this book for a better version).

**2.4. Introducing the 3D nrc-space**

Can one go further? Could the above 2D representations be a mere stage towards a more abstract, more synthetic, 3D representation? Instead of considering the four 2D spaces, one could consider a 3D space, with coordinates n, r, c. In the nrc-cell with coordinates (n, r, c), one would put the Boolean True (or a 1, or a dot, or any arbitrarily chosen sign) if n is present in rc-cell (r, c). The 2D spaces would then appear as the 2D projections of the 3D nrc-space.

Corresponding to this 3D view, there would be a still larger set (of cardinality $2 \times 9^3 = 1458$) of possible CSP variables: all the $Xn°r°c°$ and $Xn°b°s°$ for all the constants n°, r°, c°, b°, s° as above. Each of these CSP variables would take Boolean values (True or False). The constraints would then have to be re-written in a different, more complex way:
$Xn°r°c° \wedge Xn°'r°'c°' = $ False, for all the pairs {n°r°c°, n°'r°'c°'} such that
 – either n° = n°' and the rc-cells r°c° and r°'c°' share a unit;
 – or n° ≠ n°' and r°c° = r°'c°';
together with similar constraints for the $Xn°b°s°$. Moreover, obvious relationships could be written between these "3D" CSP variables and the "2D" CSP variables of the previous section: $Xn°r°c° = $ True $\Leftrightarrow Xr°c° = n° \Leftrightarrow Xr°n° = c° \Leftrightarrow Xc°n° = r° \Leftrightarrow Xb°n° = s°$ whenever (r°, c°) = [b°, s°].

However, considered as CSP variables, these "3D" variables would not bring anything new (with respect to the four sets of "2D" CSP variables), because all the "strong" CSP constraints they would allow to write can already be written in the four sets of "2D" CSP variables. Actually, Sudoku has no "3D diagonal" constraints. Rejecting the adoption of the "3D" variables as CSP variables is thus a form of Occam's razor principle.

Nevertheless, the 3D view will not be completely forgotten: each of these non-CSP-variables will reappear later as a "label" (see section 3.2.1), i.e. as a name n°r°c° or (n°, r°, c°) for the set of four equivalent possibilities: {$Xr°c° = n°$, $Xr°n° = c°$, $Xc°n° = r°$, $Xb°n° = s°$}. And the 3D nrc-space will reappear as a representation of the set of these labels.

# 3. The logical formalisation of a CSP

Although this book may be used as a support for exercises in Logic or AI and it must therefore adopt a clear and non ambiguous formalism, it is not intended to be an introductory textbook on these disciplines and it also aims at defining resolution techniques readable with no pre-requisite. The non-mathematically oriented reader should not be discouraged by the formalism introduced in this chapter: apart from the proof (in chapter 4) of meta-theorems 2.1, 2.2 and 2.3 and some local remarks, it will be used mainly as a general background for our resolution paradigm. On the practical side of things, starting with Part II, the resolution rules will always be formulated in plain English, so that it will be possible to skip the logical version, if it is ever written. Moreover, most of the resolution rules (and, in particular, the chain rules of the various types considered in this book) will also be displayed in very simple, intuitive, quasi-graphical representations. As for the Sudoku example, the Sudoku Grid Theory (SGT) and Sudoku Theory (ST) introduced in section 3.5 below can be considered as completely obvious from an intuitive point of view (so that this chapter and the next can be skipped or kept for later reading).

**3.1. A quick introduction to Multi-Sorted First Order Logic (MS-FOL)**

In order to have a logical formalism as concrete and intuitive as possible, we want our formulæ to be simple and compact; we shall therefore use Multi-Sorted First Order Logic with equality (MS-FOL). A theory in formal logic always deals with some limited topic and it does this in a well-defined language adapted to its purpose. The distinctive feature of MS-FOL consists of assuming that the topic of interest has different types of objects, called *sorts*.

From a theoretical point of view, such logic is known to be formally equivalent to standard First Order Logic with equality (FOL): formulæ, theories and proofs in MS-FOL translate easily to and from formulæ, theories and proofs in FOL. But, for practical purposes, the natural expressive power of MS-FOL is much greater, i.e. things are generally much easier to write. For a more extensive introduction to MS-FOL and an easy but technical proof of its equivalence with FOL, see e.g. [Meinke et al. 1993].

In most of the real world applications of logic and in computer science (where modern languages are typed – and even object oriented), MS-FOL rather than FOL is the natural reference, whether or not any kind of variant or extension (intuitionistic, modal, temporal, dynamic and so on) is required. This is not to suggest that the specific sorts needed for an application are in any way "natural"; they can only be the result of a modelling process, as shown in the previous chapter.

Our introduction to MS-FOL follows the standard lines of any introduction to logic. It is here only for purposes of (almost) self-containment of this book. It also introduces a few unusual but intuitive and useful abbreviations.

### *3.1.1. The language of a theory in MS-FOL*

Every theory in FOL or MS-FOL is defined by a specific language reflecting the concepts and only the concepts pertaining to the underlying domain or "universe of discourse" (its "vocabulary"); but the syntax or "grammar" of all these specific languages is built according to universal principles.

#### *3.1.1.1. Specific sorts, constants and variables*

First is given a set Sort of *sorts*; these are merely abstract symbols (generally written as Greek letters or with a capital first letter), naming the various types of objects of the application. Attached to each sort $\sigma$, there are two disjoint sets of symbols: $ct(\sigma)$ for naming *constants* of this sort and $var(\sigma)$ for naming *variables* of this sort. Moreover, the sets attached to two different sorts are disjoint (unless one sort is a sub-sort of the other). When a variable appears anywhere (e.g. after a quantifier), its sort does not have to be further specified: it is known from its name.

#### *3.1.1.2. Specific predicates and functions*

In FOL, *predicate symbols* (also called relation symbols) are names used to express either properties of objects or relations between objects they relate. A predicate symbol has an "arity": an integer number defining the number of arguments it takes. In MS-FOL, it also has a "signature": a sequence of sorts, the length of its arity, specifying that each of the arguments of this predicate must be of the sort corresponding to the place it occupies in it.

One generally considers theories with equality. In this case, for each sort $\sigma$, there is an equality predicate: "$=_\sigma$" (= with subscript $\sigma$) expressing equality between objects of the same sort $\sigma$. "$=_\sigma$" has arity 2 and signature $(\sigma, \sigma)$. We shall also use $\neq_\sigma$ to express non-equality: if $x_1$ and $x_2$ are variables of sort $\sigma$, then $x_1 \neq_\sigma x_2$ is an abbreviation for $\neg(x_1 =_\sigma x_2)$. As sorts are known from the names of the variables, a loose notation with = instead of $=_\sigma$ is generally used.

Similarly, a *function symbol* is a name used to refer to a function. In MS-FOL, it has a sort (the sort of the result), an arity and a signature (specifying respectively the number and the sequence of sorts of its arguments).

*3.1.1.3. Terms and atomic formulæ*

From now on, we describe general principles (the "grammar" or syntax of MS-FOL) for building formulæ (the "sentences" of MS-FOL) from the above-defined specific "vocabulary".

*Terms* of sort σ are defined recursively:
– if "a" is a symbol for a constant of sort σ, then it is a term of sort σ;
– if "x" is a symbol for a variable of sort σ, then it is a term of sort σ;
– if f is a symbol for a function of sort σ, arity n and signature $(\sigma_1, …, \sigma_n)$, and if $t_1, …, t_n$ are terms of respective sorts $\sigma_1, …, \sigma_n$, then $f(t_1, …, t_n)$ is a term of sort σ.

An *atomic formula* is the standard means for expressing elementary relations between its arguments. Atomic formulæ are defined as follows:
– if R is a symbol for a predicate of arity n and signature $(\sigma_1, …, \sigma_n)$, and if $t_1, …, t_n$ are terms of respective sorts $\sigma_1, …, \sigma_n$, then $R(t_1, …, t_n)$ is an atomic formula.

An atomic formula $R(t_1, …, t_n)$ is said to be *ground* if for every i from 1 to n, $t_i$ contains no variable symbol. Such a formula expresses a relation between constants.

*3.1.1.4. Logical connectives (or logical operators)*

The language of MS-FOL has the standard logical connectives of FOL:
– "∧", "&" or "and" are used indifferently to express conjunction;
– "∨" or "or" are used indifferently to express disjunction;
– "¬" or "not" are used indifferently to express negation;
– "⇒" expresses logical implication;
– "∀x" expresses universal quantification over objects of the sort of x;
– "∃x" expresses existential quantification over objects of the sort of x.

We shall also make an extensive use of the following (not all very standard) abbreviations (especially for the formal expression of the chain rules in chapter 5 and of the Subset rules in chapter 8), where F is any formula:
– "∃!xF(x)" expresses that "there exists one and only one x such that F(x)";
– "∀x≠$x_1,x_2,…,x_n$F" expresses a single quantification over x; by definition, it will mean: $\forall x[x=x_1 \vee x=x_2 \vee … \vee x=x_n \vee F]$;
– "∀≠$(x_1,x_2,…,x_n)$F" expresses n universal quantifications for n different objects of the same sort; it should not be confused with the previous abbreviation; by definition, it will mean:
$\forall x_1 \forall x_2 … \forall x_n[x_2=x_1 \vee x_3=x_1 \vee x_3=x_2 \vee … \vee x_n=x_1 \vee x_n=x_2 \vee … \vee x_n=x_{n-1} \vee F]$;
– "∀x∈$\{x_1,x_2,…x_n\}$F(x)" does not surreptitiously introduce set theory; it merely expresses the conjunction of n non quantified formulæ: $F(x_1) \wedge F(x_2) \wedge … \wedge F(x_n)$;

– similarly, "∃x∈{x$_1$,x$_2$,…,x$_n$}F(x)" merely expresses the disjunction of n non-quantified formulæ: F(x$_1$) ∨ F(x$_2$) ∨ … ∨ F(x$_n$).

*3.1.1.4 Formulæ*

Formulæ of an MS-FOL theory are defined recursively:

– if R(t$_1$, …, t$_n$) is an atomic formula, then it is a formula;

– Boolean combinations of formulæ are formulæ: if F and G are formulæ, then ¬F (also written "not F"), F ∧ G (also written "F & G" or "F and G") , F ∨ G (also written "F or G") and F ⇒ G are formulæ;

– if F is a formula and x is a variable of any sort, then ∀xF and ∃xF are formulæ.

A variable x appearing in a formula is called free if it is not in the scope of a ∀x or ∃x quantifier. A formula with no free variables is called closed (all its variables are quantified); otherwise, the formula is called open. An open formula may have quantifiers (when only some but not all of its variables are quantified).

### *3.1.2. General logic axioms and inference rules*

Notice that, up to this point, no notion of truth has been introduced: a formula is only a syntactic construct. Provability (rather than truth) will be defined via axioms and rules of inference. As we shall need classical logic to formulate the CSP problem in the rest of this chapter and intuitionistic logic to define the CSP resolution theories in chapter 4, we shall introduce these axioms in a way that allows a clear separation between classical and intuitionistic logic.

*3.1.2.1 Gentzen's "natural logic"*

There are two main formulations of logic. Hilbert's is probably the most familiar one (it is the one we adopted in *HLS*). Here, we shall prefer Gentzen's "natural logic" [Gentzen 1934], for three reasons:

– it makes no formal distinction between an axiom (such as: A ∧ B ⇒ A) and a rule of inference (such as *Modus Ponens*: from A and A ⇒ B, infer B);

– each logical connective is defined in itself by two complementary and very intuitive rules of elimination and introduction (whereas some of Hilbert's axioms mix several connectives and they can have many equivalent formulations);

– in many occasions, proofs can be made recursively by following the structure of a formula; a separate rule for each axiom makes this easier; in particular, our three meta-theorems will be shown to be obvious.

Gentzen's formulation is a set of rules in the form: $\dfrac{\text{premises}}{\text{conclusion}}$ (name of the rule),

more precisely: 
$$\frac{\Gamma_1 \vdash \phi_1, \quad \Gamma_2 \vdash \phi_2, \quad \Gamma_3 \vdash \phi_3, \ldots}{\Delta \vdash \psi} \quad \text{(name of the rule)}$$

$\Gamma \vdash \phi$ is interpreted as: $\phi$ can be deduced from $\Gamma$;
the whole rule is interpreted as: if $\phi_i$ can be deduced from $\Gamma_i$, for i = 1, 2, 3, …, then $\psi$ can be deduced from $\Delta$;
here $\phi_i$ and $\psi$ are formulæ, $\Gamma_i$ and $\Delta$ are finite sets of formulæ (sets, not sequences – the order of their elements is irrelevant).

This formalism is the same for classical and intuitionistic logic, but the intended meaning of "can be deduced from" is stronger in intuitionistic logic: it means that there is an effective, constructive proof (in particular, not only a proof by contradiction). Whereas the classical interpretations are in terms of True and False (i.e. $\phi$ means that $\phi$ is True), the intuitionistic ones are in terms of Provable and Contradictory (i.e. $\phi$ means that $\phi$ is provable; $\phi_1 \wedge \phi_2$ means that $\phi_1$ is provable and $\phi_2$ is provable; $\phi_1 \vee \phi_2$ means that $\phi_1$ is provable or $\phi_2$ is provable).

*3.1.2.2 Propositional axioms common to intuitionistic and classical logic*

Most of the rules for the various connectives go by pairs (E for elimination, I for introduction). We use the standard abbreviations such as: $\Gamma, \phi_1, \phi_2$ for $\Gamma \cup \{\phi_1, \phi_2\}$; we also use the symbol $\bot$ for the absurd, considered as a proposition always false.

– *Implication*:

$$\frac{\Gamma \vdash \phi \Rightarrow \psi \quad \Gamma \vdash \phi}{\Gamma \vdash \psi} \; (\Rightarrow E) \qquad\qquad \frac{\Gamma, \phi \vdash \psi}{\Gamma \vdash \phi \Rightarrow \psi} \; (\Rightarrow I)$$

($\Rightarrow$ E) is the way *Modus Ponens* is expressed in Gentzen's natural logic.

– *Conjunction* (there are two elimination rules, one for each conjunct):

$$\frac{\Gamma \vdash \phi_1 \wedge \phi_2}{\Gamma \vdash \phi_i} \; (\wedge E_i) \qquad\qquad \frac{\Gamma \vdash \phi_1 \quad \Gamma \vdash \phi_2}{\Gamma \vdash \phi_1 \wedge \phi_2} \; (\wedge I)$$

– *Disjunction* (there are two introduction rules, one for each disjunct):

$$\frac{\Gamma \vdash \phi_1 \vee \phi_2 \quad \Gamma, \phi_1 \vdash \psi \quad \Gamma, \phi_2 \vdash \psi}{\Gamma \vdash \psi} \; (\vee E) \qquad \frac{\Gamma \vdash \phi_i}{\Gamma \vdash \phi_1 \vee \phi_2} \; (\vee I_i)$$

– *Negation*: there is no rule for negation, ¬ϕ is considered as an abbreviation for ϕ ⇒ ⊥. Instead there is an elimination rule for the absurd:

– *Absurd*:

$$\frac{\Gamma \vdash \bot}{\Gamma \vdash \phi} \quad (\bot E)$$

The meaning of rule (⊥ E) is that anything can be deduced from the absurd. Contrary to the other connectives, there is (fortunately) no rule (⊥ I) for introducing the absurd.

*3.1.2.3 Propositional axioms specific to classical logic: "the excluded middle"*

These are four *intuitionistically equivalent* forms of the only law specific to classical logic, the "law of the excluded middle":
  – Excluded middle: ⊢ A ∨ ¬A
  – *Reductio ad absurdum* (reduction to the absurd): ⊢ ¬¬A ⇒ A
  – Contraposition: ⊢ (¬B ⇒ ¬A) ⇒ (A ⇒ B)
  – Material implication: ⊢ (A ⇒ B) ⇔ (¬A ∨ B)

*3.1.2.4 Axioms on quantifiers*

They can also be written as natural deductions:

– *Universal quantification*:

$$\frac{\Gamma, \phi[t/x] \vdash \psi}{\Gamma, \forall x\phi \vdash \psi} \quad (\forall E) \qquad \frac{\Gamma \vdash \phi}{\Gamma \vdash \forall x\phi} \quad (\forall I)$$

– *Existential quantification*:

$$\frac{\Gamma, \phi \vdash \psi}{\Gamma, \exists x\phi \vdash \psi} \quad (\exists E) \qquad \frac{\Gamma \vdash \phi[t/x]}{\Gamma \vdash \exists x\phi} \quad (\exists I)$$

In these rules, ϕ[t/x] is the formula obtained by replacing every free occurrence of variable x in ϕ(x) by term t (where t does not contain variables present in ϕ). Notice that, in intuitionistic logic, contrary to classical logic, ∃x is not equivalent to ¬∀x¬. This is usually interpreted by saying that proofs of existence by the absurd are not allowed; proofs of existence must be constructive; they must explicitly exhibit the object whose existence is asserted.

*3.1.3. Theory specific axioms, proofs and theorems in an MS-FOL theory*

In any logic, an axiom is defined as a closed formula and a theory as a set of axioms including the general logic axioms. In Gentzen's natural logic, an axiom appears as a rule with no premise and with empty set Γ. In short notation, it can be written, as: │── A (as we did in section 3.1.2.3).

A proof is a sequence of expressions of the form Γ │── φ, each of which is either an axiom or the conclusion of a logic rule with premises equal to previous expressions in the sequence. A theorem is the last expression of a proof, with empty set Γ.

*3.1.4. Model theory, consistency and completeness theorems*

In this section, we shall consider classical logic only. Models of intuitionistic logic will be introduced in chapter 4.

Definition: an *interpretation* of a theory T is a set of disjoint sets (unless one sort is a subsort of another), one for each sort (more precisely, it is a functor i from Sort to Set, i.e. to the category of sets), together with:

– for each sort σ, an application from ct(σ) into i(σ);

– for each n-ary function symbol f with sort σ and signature $(\sigma_1, \ldots \sigma_n)$, a function i(f): $i(\sigma_1) \times \ldots \times i(\sigma_n) \rightarrow i(\sigma)$;

– for each n-ary predicate symbol R with signature $(\sigma_1, \ldots \sigma_n)$, a subset i(R) of $i(\sigma_1) \times \ldots \times i(\sigma_n)$.

An interpretation i of a theory T can be extended to any formula of T in an obvious way, following the recursive definition of formulæ. If i is an interpretation of T and F is a formula, we introduce the symbol "│=" (read satisfies) and the expression i │= F to mean that i satisfies F.

Definition: a *model* of T is an interpretation i of T such that its extension satisfies all the axioms of T.

The most basic theorems of logic (proven in any logic textbook) are Gödel's consistency and completeness theorems. They establish the correspondence between syntax and semantics, i.e. between formal proof and set theoretic interpretations:

 – **Consistency theorem: a formula provable in T is valid in any model of T**;

 – **Completeness theorem: a formula valid in any model of T is provable in T**.

*3.1.5. Non uniqueness of models of an MS-FOL theory*

In FOL or MS-FOL, there is no general means of specifying that a theory has a unique model. For theories with an infinite model, it is even the contrary that is true:

due to the "compactness" theorem, there are always infinitely many models and there are models of arbitrarily large infinite cardinality.

**3.2. The formalisation of a CSP in MS-FOL: T(CSP)**

The CSP axioms can generally be classified into four general categories: CSP sort axioms (defining the domain of the variables, e.g. rows, columns, …), CSP background axioms (expliciting general structural properties of the problem, e.g. the structure of the Sudoku grid), CSP constraints axioms (the core content of the CSP, e.g. the famous four Sudoku axioms), CSP instance axioms (relative to each instance of the CSP, e.g. the entries of a puzzle).

*3.2.1. Sorts and predicates of the CSP*

There are many ways a CSP could be expressed as a logical theory T(CSP). Some of them may be simpler than the one proposed here, but our universal formalisation is mainly intended to be a step towards the introduction of CSP resolution theories.

Our approach will be based on the following two remarks. Firstly, as mentioned in the Introduction, any non-unary constraint (including the implicit "strong" constraints between different values for the same variable) is supposed to be re-written as a set of binary constraints and we can thus suppose that our CSP is binary.

Secondly, the notion of a *label* will play a central role. Labels will be the basis for a proper definition of candidates in chapter 4. Our non standard definition of a label (as an equivalence class of pre-labels) may seem a little convoluted, but it provides for the possibility of having multiple representations of the same basic facts without confusing the underlying CSP variables. As shown in chapter 2 with the four "2D" spaces in Sudoku, multiple representations are very useful in practice.

From a set theoretic point of view, a binary constraint c between two CSP variables $X_1$ and $X_2$ (which may be the same one) is the subset of pairs in $Dom(X_1) \times Dom(X_2)$ satisfying this constraint; equivalently, it is also a symmetric subset of
$[\{X_1\} \times Dom(X_1) \oplus \{X_2\} \times Dom(X_2)] \times [\{X_1\} \times Dom(X_1) \oplus \{X_2\} \times Dom(X_2)])$, which is itself a symmetric subset of P×P (where P is the set of pre-labels, defined below).

The complement of this set in P×P is a symmetric subset DC(c) of P×P; it is obviously equivalent to a set of pairwise c-links between pre-labels, if we say that there is a c-link between two pre-labels $p_1$ and $p_2$ if and only if $(p_1, p_2) \in DC(c)$, i.e. if they are contradictory with respect to constraint c. The following definitions make this more formal.

Definition: in a CSP, a *pre-label* is a <variable, value> pair, i.e. a pair <X°, x°>, where X° is a CSP variable and x° ∈ Dom(X°). The set P of pre-labels is thus the disjoint union (the "direct sum", the ⊕) of the domains of the variables. Informally, this can also be viewed as the union of all the elements of all the domains, after each element has been subscripted by the name of the variable.

Definition: in a CSP, two pre-labels <X°, x°> and <X°', x°'> are equivalent if equalities X° = x° and X°' = x°' are equivalent as a direct effect of the definitions. Equivalence is the result of a modelling decision. It entails that the two equivalent pre-labels are related to any other pre-labels by exactly the same constraints.

Definition: a *label* is a name for an equivalence class of pre-labels (with respect to the above defined equivalence relation). If l° is a label and <X°, x°> is an element of this class, i.e. if <X°, x°> ∈ l°, we often use <X°, x°> to mean l°, by abuse of language. It should be noted that, given a CSP variable X° and a value x° in its domain, there is a unique label associated with the <X°, x°> pair. But, conversely, due to our approach of introducing several redundant representations in the modelling process, given a label, there will generally be several elements in its equivalence class.

Given a label l° and a CSP variable X°, there are only two possibilities: either there is one and only one value x° in Dom(X°) such that <X°, x°> ∈ l° (in which case we say that *<X°, x°> is a representative of l°* and that *l° is a label for X°*) or there is no such x° (in which case we say that *l° is not a label for X°*).

Definition: two different labels $l_1$ and $l_2$ are *linked by constraint c* if there are representatives $p_1$ = <$X_1$, $x_1$> of $l_1$ and $p_2$ = <$X_2$, $x_2$> of $l_2$ such that ($p_1$, $p_2$) ∈ DC(c). "linked-by c" is a symmetric (but neither reflexive nor transitive) relation. This definition entails that ($p_1$, $p_2$) ∈ DC(c) for any representatives $p_1$ of $l_1$ and $p_2$ of $l_2$. By abuse of language, we sometimes write that ($l_1$, $l_2$) ∈ DC(c).

Definition: two different labels $l_1$ and $l_2$ are *linked by some constraint* or simply *linked* if ($l_1$, $l_2$) ∈ DC(c) for some c. "linked" is a symmetric (but neither reflexive nor transitive) relation.

Pre-labels are used as a technical tool for the definition of labels. From now on, we shall meet mainly CSP variables, values and labels.

We can now define the logical language of T(CSP). Basically, it has the following *sorts*, sort constants and sort variables:

– for each CSP variable X, there is a sort X; for CSP variable X, for each element in Dom(X), there is a constant symbol of sort X (considered as a name for this possible value of X); variables of sort X are: x, x', $x_1$, $x_2$, …;

– a sort Label; for each element in the set of labels, there is a constant symbol of sort Label (the name of this label); variables of sort Label are: l, l', $l_1$, $l_2$, … but also (because it will be convenient when we define chains) $z, z', z_1, z_2, \ldots$ and $r, r', r_1, r_2, \ldots$; sometimes, we shall also use capital letters for labels;

– a sort Constraint; for each constraint in the CSP, there is a constant symbol of sort Constraint (the name of this constraint); variables of sort Constraint are $c, c', c_1, c_2, \ldots$; [additionally, or alternatively when each constraint can be defined in a unique way by a label and a constraint type (as in the Sudoku or the N-Queens cases), one may have a sort Constraint-Type; modifying accordingly the general theory and all the resolution rules defined later in this book is straightforward];

– a sort CSP-Variable; for each CSP variable X, there is a constant symbol X of sort CSP-Variable (CSP variables are considered to be their own name); variables of sort CSP-Variable are $V, V', V_1, V_2, \ldots$; **CSP-Variable is considered as a sub-sort of Constraint**; [one could also have CSP-Variable-Type, a sub-sort of Constraint-Type];

– a sort Value; for each value in the (ordinary, set theoretic) union of the domains of the CSP variables, there is a constant symbol; variables of sort Value are $v, v', v_1, v_2, \ldots$

The logical language of the CSP has only the following four *predicates*:

– a unary predicate: value, with signature (Label); the intended meaning of value(l) is that, if <X, x> is any representative of l, then x is the value of variable X;

– a ternary predicate: linked-by, with signature (Label, Label, Constraint); the intended meaning is that the first two arguments, labels $l_1$ and $l_2$, are linked by the constraint given in the third argument, i.e. they are incompatible for this constraint;

– a binary predicate: linked, with signature (Label, Label); the intended meaning is that the two arguments, labels $l_1$ and $l_2$, are linked by some of the constraints.

For technical reasons, it also has the following *predicate*:

– a ternary predicate: label, with signature (Label, CSP-Variable, Value); the intended meaning of label(l, X, x) is that l is the label of the <variable, value> pair <X, x>.

Notice that, contrary to the sorts Label, Constraint [and/or Constraint-Type] and CSP-Variable [and/or CSP-Variable-Type] that will play a major theoretical role in the formulation of the resolution rules, sort Value and associated predicate "label" will appear mainly for the technical purpose of specifying the correspondence between labels and <variable, value> pairs (see the "meaning of labels" axiom below) and for formulating the completeness of the solution (see the eponym axiom below). In applications, there may be simpler, perhaps implicit ways of specifying this correspondence and of writing this axiom (see section 3.5).

Optionally, the language of the CSP may include additional sorts useful for formulating certain types of rules or for interacting with the outer world in natural terms; in some cases, the general sorts above may be defined from these additional sorts. For details about this, see the Sudoku example (section 3.5).

What is most important here is that:

– the universal language necessary to formulate the general CSP theory is very restricted;

– with the mere addition of a single predicate "candidate" in the CSP resolution theories (in chapter 4), this language will be enough to define very general and powerful resolution rules valid for any CSP.

*3.2.2. Implicit CSP sort axioms*

In MS-FOL, sort axioms do not have to be written explicitly, as would be the case in FOL, because they are considered as part of the definition of sorts. For each sort X, implicit sort axioms for a finite CSP would be of two kinds: exhaustiveness of domain constants (the domain of X has no other value than those corresponding to constants of this sort) and unique names assumption (two different constants for X name two different objects of sort X). Notice that, contrary to constants, there is no unique names assumption on variables: two variables (of same sort) can designate the same object (of this sort); when one wants to specify that they refer to different objects, this must be stated explicitly.

*3.2.3. CSP background axioms*

Until now, we have defined sorts, predicates and functions and we have given their intended meaning. But we have written nothing that would formally ensure that they really have this meaning. The role of the following background axioms is to express the fixed structure of the problem and its translation into a graph of labels, independently of any values; they deal with correspondences between original <variable, value> pairs and labels, and with the re-writing of the original constraints into symmetric links between labels:

**meaning of labels:** for each CSP variable $X°$, for each $x°$ in $Dom(X°)$, if $l°$ is the (unique) label of $<X°, x°>$, the axiom defined by the ground atomic formula: **label($l°, X°, x°$)**;

**re-writing of each constraint as a set of links:** for each constraint $c°$, for each pair of labels $l°_1$ and $l°_2$ such that $(l°_1, l°_2) \in DC(c°)$, the axiom defined by the ground atomic formula: **linked-by($l°_1, l°_2, c°$)**;

**symmetry of links:** $\forall c\ \forall l_1\ \forall l_2$ {**linked-by($l_1, l_2, c$)** $\Leftrightarrow$ **linked-by($l_2, l_1, c$)**}; (this is normally useless, because it should be ensured by the modelling process);

**exhaustiveness of constraints:** $\forall l_1 \forall l_2$ {**linked($l_1, l_2$)** $\Leftrightarrow \exists c$ **linked-by($l_1, l_2, c$)**}.

This is the general, slightly artificial, formulation of background axioms for any CSP. In each particular CSP, the concrete expression of these axioms may be adapted to the specificities of the problem. They may even be partly implicit in the definition of the "technical sorts". This will appear clearly in the Sudoku example.

### 3.2.4. CSP constraints axioms

It is not enough to associate a link with each constraint; the fact that these links really stand for constraints must also be written. We can now state what could be called the "core" CSP axioms (the background ones being only technicalities):

**Meaning of links as constraints:** $\forall l_1 \forall l_2 \{value(l_1) \wedge linked(l_1, l_2) \Rightarrow \neg value(l_2)\}$;

**Completeness of solution:** $\forall V \, \exists!v \, \exists l \, [label(l, V, v) \wedge value(l)]$.

We have written the first axiom in an asymmetrical way that will make the transition to CSP resolution theories more natural. As for the second axiom, it can be read as: each CSP variable has one and only one value. Notice that this does not mean that the CSP has a unique solution; it only means that, in any solution, there is one and only one value for each CSP variable.

### 3.2.5. Logical theory of the CSP: T(CSP)

Finally, define the Theory of the CSP, **T(CSP)**, as the MS-FOL theory written in the above defined language and consisting of (the implicit sort axioms,) the CSP background axioms and the CSP constraints axioms.

### 3.2.6. CSP instance axioms

A given corresponds to the assertion of a value for a label: $value(l^0)$. An instance P of the CSP is specified by a set of n givens $l^0_1, \ldots, l^0_n$ (where all the $l^0_i$ are meta-symbols for – i.e. they stand for – constant label symbols) and it thus corresponds to the conjunction:

$value(l^0_1) \wedge \ldots \wedge value(l^0_n)$. We name it indifferently E(P) or $E_P$ (E for "entries").

Finally, we have the obvious ***theorem: there is a natural correspondence between a solution of the original CSP instance P and a model of its logical theory T(CSP) $\cup$ $E_P$.***

Consequence: as a logical theory can only prove properties that are true in all its models, the CSP Theory for a given instance can only prove values that are common to all the solutions of this instance, if there is at least one (it can prove anything if there is no solution, i.e. if the instance axioms are inconsistent).

### 3.3. Remarks on the existence and uniqueness of a solution

Notice that, given any instance P, the axioms of T(CSP) together with $E_P$ *a priori* imply neither the existence nor the uniqueness of a solution for P. Concerning the existence, this may seem to contradict the axiom of completeness, but this axiom only puts a condition on a solution, it does not assert that there is a solution (i.e. that $E_P$ is consistent with T(CSP)). Indeed, any axiom that would assert the existence of a solution for any P would be trivially inconsistent. Let us consider the Sudoku example (see section 3.5 for the specific notations).

In this case, no set of *a priori* conditions on the entries of an instance P is known that would ensure that P has a solution (at least one). Obviously, some trivial necessary conditions for existence can be written (such as not having the same entry twice in a row, a column or a block) but they are very far from being sufficient.

As for uniqueness, for any puzzle P and corresponding axiom $E_P$, one may think that it could be expressed by the following additional axiom:

– ST-U: there is at most one solution:

$$\forall r \forall c \forall n_{rc} \forall n'_{rc} [\text{value}(n_{rc}, r, c) \wedge \text{value}(n'_{rc}, r, c) \Rightarrow n_{rc} = n'_{rc}].$$

But this is not true: such an axiom for uniqueness cannot imply that the solution is unique. It can only imply that, if the solution is not unique, then $E_P$ contradicts this axiom; i.e. theory ST ∪ ST-U ∪{$E_P$} is inconsistent. This is why we prefer to speak of the *assumption* rather than the *axiom* of uniqueness. Whereas the Sudoku axioms are constraints the player must satisfy, the assumption of uniqueness puts a constraint on the puzzle creator; a player may choose to believe it or not; if he does, it amounts to accepting an oracle.

Uniqueness of a solution is a very delicate question (see also section 3.1.5). As was the case for existence, some trivial necessary conditions on the givens can be written for uniqueness (such as having entries for at least eight different numbers – otherwise, given any solution, one could get a different one by merely permuting two of the remaining numbers) but, again, they are very far from being sufficient.

Uniqueness of the solution (i.e. of a model of the puzzle theory) can only be a consequence of the givens. But is it possible to write a formula U(P) that would be equivalent to the uniqueness of the solution if the set of givens of P satisfies it? It is likely that this problem is much more difficult than solving the puzzle.

There are famous examples of puzzles that have been proposed and asserted as having a unique solution and that have indeed several. Many of the resolution rules that have been proposed to take uniqueness into account have been used inconsistently to *conclude* that some puzzle has a unique solution. Moreover, the uniqueness of a solution for a given puzzle can be asserted only if it has already been proven – which supposes that there exists some means for proving it. In our approach, unless explicitly stated otherwise, we shall never take the uniqueness of a solution as granted and we therefore do not adopt this assumption for any CSP.

**3.4. Operationalizing the axioms of a CSP Theory**

From a logical point of view, the above-defined theory T(CSP) is necessary and sufficient to define the CSP: given any instance P (with axiom $E_P$ corresponding to its entries) and any complete solution G of P, the following are equivalent:

– G is a solution (in the intuitive sense) of instance P of the CSP;

– G is a *model* of T(CSP) ∪ {$E_P$} (in the standard sense of mathematical logic introduced in section 3.1);

– G satisfies the axioms of T(CSP) ∪{$E_P$}.

T(CSP) is therefore theoretically perfect: for any instance of the CSP, its formal and intuitive meanings coincide. The only problem with it is practical: it does not give any indication on *how* to build a solution.

From an operational point of view, the "meaning of links as constraints" axioms could be considered as a set of contradiction detection rules. For instance, they could be re-written in the following operational form: if, at some point in the resolution process of an instance, we reach a situation in which two different values should be assigned to the same variable, then we can conclude that this instance has no solution (the entries of this instance are contradictory with the axioms). This is, somehow, an operational form of these axioms. But do these forms express all the operational consequences of the original formulæ? Actually, the developments in chapter 4 will show that they do not (and they are indeed very far from doing so). The situation for the "completeness of a solution" axiom is still worse, since it does not tell anything about how it can be used in practice.

Vague as this may remain, let us define the aim we shall pursue with CSP Resolution Theories: we want to replace the above axioms by another set of axioms that could easily be interpreted as (or transformed into) a set of operational rules for building a solution. And, since most known resolution rules in the Sudoku case and in many logic puzzles are based on the notion of a candidate and on the progressive elimination of candidates, and since this idea corresponds to the common one of domain restriction in the general CSP, we want to write rules explicitly designed for this purpose. The problem is that, unless one admits recursive search (which is not a rule), no theory of this kind is known that would be equivalent to T(CSP).

This book can thus be considered as being about the operationalization of the axioms of a CSP Theory – or about its replacement by a set of axioms that can be used in a constructive way.

**3.5. Example: Sudoku Theory, T(Sudoku) or ST**

The rest of this chapter illustrates the abstract general theory with the Sudoku case. T(Sudoku) is written ST for short. With the detailed Sudoku example, our goal is to illustrate simultaneously the above formalism and the ways of taking some liberty with it in order to simplify it in any specific case. For this purpose, we start with the "natural" formalisation of Sudoku and we show how it can be made compliant with the above general approach. For the most part, at the cost of some redundancy, the following sections are designed in such a way that they can be read independently of the previous ones or before them, for readers who do not like the abstract technicalities of formal logic.

*3.5.1 Sudoku background axioms: Sudoku Grid Theory, SGT*

The minimal underlying framework of Sudoku – the minimal support necessary for the representation of any Sudoku puzzle and any intermediate state in the resolution process – is a 9×9 grid composed of nine disjoint square blocks of 3×3 contiguous cells. Therefore, whichever formulation one chooses for the constraints (in rows, columns and blocks) defining the game, any theory of Sudoku must include an appropriate theory of such a grid. In the sequel, (our version of) this theory will be called 9-Sudoku Grid Theory (or simply Sudoku Grid Theory or SGT); it will contain all the general and "static" or "structural" knowledge about grids and only this knowledge, i.e. all the knowledge that does not depend on any particular entries for a puzzle and that does not change throughout the resolution process.

*3.5.1.1. Sorts*

In the limited world of SGT (and of ST in the next section), we shall consider the following sorts:

– Number: "Number" is the type of the objects intended to fill up the rc-cells of a grid; when, outside of the formal ST world, we need to refer to other kinds of numbers, we shall use their standard specific mathematical type: for instance, integers from 0 to infinity are simply called integers; the subscripts appearing in variables of any sort are integers, not Numbers; we have chosen to introduce the sort Number, because Sudoku is generally expressed in terms of digits, but one could introduce instead a sort Symbol, with nine arbitrary constant symbols;
  - constant symbols: n1, n2, n3, n4, n5, n6, n7, n8, n9;
  - variable symbols: n, n', n'', $n_0$, $n_1$, $n_2$, …;
– Row:
  - constant symbols: r1, r2, r3, r4, r5, r6, r7, r8, r9;
  - variable symbols: r, r', r'', $r_0$, $r_1$, $r_2$, …;

- Column:
  - constant symbols: c1, c2, c3, c4, c5, c6, c7, c8, c9;
  - variable symbols: c, c', c'', $c_1$, $c_2$, …;
- Block:
  - constant symbols: b1, b2, b3, b4, b5, b6, b7, b8, b9;
  - variable symbols: b, b', b'', $b_0$, $b_1$, $b_2$, …;
- Square:
  - constant symbols: s1, s2, s3, s4, s5, s6, s7, s8, s9;
  - variable symbols: s, s', s'', $s_0$, $s_1$, $s_2$, …;
- Label: we define Label as a sort with domain the 729 elements (n°, r°, c°) such that n° is a Number constant, r° is a Row constant and c° is a Column constant; each label (n°, r°, c°) will be the label for four different <variable, value> pairs, one associated with each of the four groups of CSP-Variables, namely: (n°, r°, c°) = {<Xr°c°, n°>, <Xr°n°, c°>, <Xc°n°, r°>, <Xb°n°, s°>}, where [b°, s°] = (r°, c°); labels can be assimilated with cells in 3D space; we sometimes use a loose notation n°r°c° for (n°, r°, c°);
  - constant symbols: (n1, r1, c1), … (n9, r9, c9); sometimes also written in a loose notation: n1r1c1, … n9r9c9;
  - variable symbols: l, l', …, r, r',… , z, z';
- Constraint-Type (and CSP-Variable-Type):
  - constant symbols: rc, rn, cn, bn; notice that we use only four symbols corresponding to the four original types of constraints (*a* number in *a* cell, *a* row, *a* column or *a* block), not to specific constraints (e.g. a given number in a given row);
  - variable symbols: lk, lk', lk'', $lk_0$, $lk_1$, $lk_2$, … ("lk" instead of "c" in the general theory, because symbol "c" is used for columns in Sudoku; we choose the "lk" symbol because constraint types are used to *link* candidates).

As the variable symbols explicitly carry their sort with the first letter(s) of their name, they can be used straightforwardly in quantifiers or in equality with no further specification. For instance:

- ∀r always means "for all rows r",
- ∀c always means "for all columns c",
- ∃n always means "there exists a number n",
- = can only be used with objects of the same sort, so that writing r = c is not allowed; to be more formal, the = sign should also be subscripted according to the type of objects it relates; for instance, to assert that two rows $r_1$ and $r_2$ are equal, we should use a specific equality symbol $=_r$ and write $r_1 =_r r_2$ (but we shall be lax on this notation also, since no confusion can arise from it).

Here is a very simple example of how MS-FOL simplifies formulæ: one can write ∀rF instead of what could only be written in FOL with an additional "row" predicate, something like ∀r[row(r) ⇒ F]. In longer formulæ, this may lead to drastic simplifications.

Remark on Constraint versus Constraint-Type: while the four elements of Constraint-Type correspond to the four 2D-spaces, the elements of Constraint (if we used this sort instead of Constraint-Type) would be represented by the 324 2D-cells of these four 2D spaces. Given any label l = (n°, r°, c°) and any constraint type lk, there is one and only one constraint of type lk "passing through l".

*3.5.1.2. Function and predicate symbols*

The SGT language has the "label" predicate necessary to specify all the correspondences between each label n°r°c° and its four <Xr°c°, n°>, <Xr°n°, c°>, <Xc°n°, r°>, <Xb°n°, s°> representatives. It also has the following functions: block and square [both with signature (Row, Column) and with respective sorts Block and Square], row and column [both with signature (Block, Square) and with respective sorts Row and Column], establishing the correspondences between the two coordinate systems: (r, c) and [b, s]. See sections 2.3 and 2.4 for details.

*3.5.1.3. Background axioms (Axioms of Sudoku Grid Theory: SGT)*

SGT has all the axioms asserting the equivalences stated in section 2.3.5, but they are now written in the form specified by the general theory (meaning of labels), i.e. for each Number constant n°, for each Row constant r°, for each Column constant c°, for each Block constant b° and for each Square constant s° such that [b°, s°] = (r°, c°), the following four ground atomic formulæ are axioms of SGT:
label(n°r°c°, r°c°, n°), label(n°r°c°, r°n°, c°), label(n°r°c°, c°n°, r°),
label(n°r°c°, b°n°, s°).

*3.5.1.4. Block-free Grid Theory, LatinSquare Grid Theory (LSGT)*

The Sudoku Grid Theory defined above can be simplified according to the following principles:
 – forget the sorts Block and Square,
 – forget all the functions and predicates referring to the above sorts.

What is thus obtained is a theory of grids that does not mention blocks and that is appropriate for Latin Squares: LSGT.

***Theorem 3.1: There is a one-to-one correspondence between the models of SGT and the models of LSGT with added functions defining the proper correspondence between the two coordinate systems.***

Proof: the proof involves some easy but tedious technicalities concerning the correspondence between theories in MS-FOL and in FOL (along the lines of [Meinke & al. 1993]). Given a model of SGT, just forget anything about blocks and squares to get a model of LSGT. Conversely, given a model of LSGT, the key is that the added functions can be used to define new predicates for blocks and squares and that these predicates can, in turn, be used to introduce the new sorts Block and Square. Details of the proof are left as an exercise for the motivated reader.

### 3.5.2. Sudoku axioms, Sudoku Theory (ST)

With a proper choice of the sorts, Sudoku Theory (ST) can be axiomatised as a mere transliteration of the naive problem formulation. ST is an extension of Sudoku Grid Theory (SGT).

*3.5.2.1. The sorts, functions and predicates of Sudoku Theory*

ST has the same sorts, functions and axioms as SGT.

In addition, in conformance with the general theory, ST also has a predicate **value** with signature (Number, Row, Column). We define an auxiliary predicate **value'** with signature (Number, Block, Square) by the change-of-coordinates axiom:

**CC: $\forall n \forall b \forall s$ {value'[n, b, s] $\Leftrightarrow$ value(n, row(b, s), column(b, s))}.**

*3.5.2.2. The axioms of Sudoku Theory*

The only point in stating the ST axioms is that we must be careful if we want to guarantee the best possible proximity with the resolution theories to be defined later. For instance, if we write that there must be one value for each cell (*in fine* an inescapable condition of the problem), this precludes all intermediate states from satisfying this axiom; we therefore try to limit the number of such assertions: indeed it will appear in only one axiom (ST-C). All the other general conditions in the statement of the problem can be expressed as "single occupancy" or "mutual exclusion" axioms – this is why, anticipating on the present formalisation, we adopted the first presentation of the game in the Introduction.

ST is defined as the specialisation of SGT (i.e. it has all the axioms of SGT) with CC and the following additional five axioms.

The first four axioms, "meaning of links as constraints axioms" are the quasi direct transliteration of the English formulation of the problem, as given in the Introduction:

– **ST-rc**: in natural rc-space, every rc-cell has at most one number as its value (i.e. given any rc-cell, it can have at most one value):

$\forall r \forall c \forall n_1 \forall n_2$ {value($n_1$, r, c) $\wedge$ $n_1 \neq n_2$ $\Rightarrow$ ¬value($n_2$, r, c)};

notice that the condition linked-by($l_1$, $l_2$, rc) of the general theory is here written more explicitly by giving the same values to the r and c components of both labels $l_1 = n_1 rc$ and $l_2 = n_2 rc$ and different values to their n components; the same remark applies to the next three axioms;

– **ST-rn**: in abstract rn-space, every rn-cell has at most one column as its value (i.e. given a row, a given number can appear in it in at most one column):

$$\forall r \forall n \forall c_1 \forall c_2 \{\text{value}(n, r, c_1) \wedge c_1 \neq c_2 \Rightarrow \neg\text{value}(n, r, c_2)\};$$

– **ST-cn**: in abstract cn-space, every cn-cell has at most one row as its value (i.e. given a column, a given number can appear in it in at most one row):

$$\forall c \forall n \forall r_1 \forall r_2 \{\text{value}(n, r_1, c) \wedge r_1 \neq r_2 \Rightarrow \neg\text{value}(n, r_2, c)\};$$

– **ST-bn**: in abstract bn-space, every bn-cell has at most one square as its value (i.e. given a block, a given number can appear in it in at most one square):

$$\forall b \forall n \forall s_1 \forall s_2 \{\text{value'}[n, b, s_1] \wedge s_1 \neq s_2 \Rightarrow \neg\text{value'}[n, b, s_2]\};$$

As in the general theory, the last axiom of ST says that the grid is complete:

– **ST-C**: the grid must be complete:

$$\forall r \forall c \exists n \, \text{value}(n, r, c).$$

At this point, it is important to notice that the first three of these axioms exhibit the symmetries and supersymmetries reviewed in chapter 2 (and they are block-free according to the definition in the next section), while the fourth exhibits analogy with the second and the third (and it is not block-free).

To better explicit the link with the general theory, let us introduce the following auxiliary predicate, with arity 7 and signature (Number, Row, Column, Number, Row, Column, Constraint-Type):
linked-by($n_1$, $r_1$, $c_1$, $n_2$, $r_2$, $c_2$, lk) is defined as a shorthand for:
[lk = rc $\wedge$ $r_1 = r_2$ $\wedge$ $c_1 = c_2$ $\wedge$ $n_1 \neq n_2$] $\vee$
[lk = rn $\wedge$ $r_1 = r_2$ $\wedge$ $n_1 = n_2$ $\wedge$ $c_1 \neq c_2$] $\vee$
[lk = cn $\wedge$ $c_1 = c_2$ $\wedge$ $n_1 = n_2$ $\wedge$ $r_1 \neq r_2$] $\vee$
[lk = bn $\wedge$ block($r_1$, $c_1$) = block($r_2$, $c_2$) $\wedge$ $n_1 = n_2$ $\wedge$ square($r_1$, $c_1$) $\neq$ square($r_2$, $c_2$)].

Then predicate "linked" of the general theory, with arity 6 and signature (Number, Row, Column, Number, Row, Column), is obviously equivalent to:
[$n_1 \neq n_2$ $\wedge$ $r_1 = r_2$ $\wedge$ $c_1 = c_2$] $\vee$ [$n_1 = n_2$ $\wedge$ share-a-unit($r_1$, $c_1$, $r_2$, $c_2$)]
with auxiliary predicate share-a-unit($r_1$, $c_1$, $r_2$, $c_2$) defined as:
[$r_1 = r_2$ $\vee$ $c_1 = c_2$ $\vee$ block($r_1$, $c_1$) = block($r_2$, $c_2$)] $\wedge$ [$r_1 \neq r_2$ $\vee$ $c_1 \neq c_2$].

*3.5.2.3. The axioms of LatinSquare Theory: LST*

One can define LatinSquare Theory (LST) as the Theory obtained from ST by forgetting any sort, function, predicate and axiom mentioning blocks and/or squares. In formal logic, we should normally have started with LST and specialised it to ST, but we are more interested in ST than in LST.

*3.5.3 Instance specific axioms (specifying the entries of a given puzzle)*

In order to be potentially consistent with any set of entries, ST includes no axioms on specific values. With any specific puzzle P we can associate the axiom $E_P$ defined as the finite conjunction of the set of all the ground atomic formulæ **value($n_k$, $r_i$, $c_j$)** such that there is an entry of P asserting that number $n_k$ must occupy rc-cell ($r_i$, $c_j$). Then, when added to the axioms of ST, axiom $E_P$ defines the theory of the specific puzzle P.

**3.6. Formalising the Sudoku symmetries**

In this section, we introduce the concept of a block-free formula and we define three transformations on formulæ (in the language of ST) that will be used in chapter 4 to state and prove the formal versions of the intuitive meta-theorems 2.1, 2.2 and 2.3. We also prove a theorem that may be interesting in its own respect: it states that if a block-free formula (a formula that does not mention blocks or squares) can be proved in ST, then it can be proved without axiom ST-bn. As a result, a block-free formula is true for Sudoku (i.e. in ST) if and only if it is true for Latin Squares (i.e. in LST).

*3.6.1. Block-free predicates and formulæ*

The notion of a block-free formula is the formalisation of the natural language phrase ("mentioning only numbers, rows and columns") that we used in chapter 2 to express informally our Sudoku meta-theorems. Block-free formulæ play a major role in all that is related to Sudoku, because they are the formulæ to which these meta-theorems can be applied.

Definition: a function or predicate is called *block-free* if the sorts Block and Square do not appear in its sort or signature. "$=_n$", "$=_r$" and "$=_c$" are block-free predicates, and so are "label" and "value", whereas "$=_b$" and "$=_s$" are not.

Definition: a formula is called block-free if it is built only on block-free functions and predicates and it does not contain the bn constant (of Constraint-Type). For instance, "value" is block-free but "value' " is not.

### 3.6.2. The $S_{rc}$, $S_{rn}$ and $S_{cn}$ transformations of a block-free formula

In order to deal properly with the different kinds of symmetries reviewed in chapter 2, we need the following definitions. For any block-free formula F, we define inductively the three block-free formulæ $S_{rc}(F)$, $S_{rn}(F)$ and $S_{cn}(F)$. These formulæ have the same arity as F but they have different signatures.

Before giving the formal definitions, notice that they are just a pompous way of saying what was said informally in chapter 2, so that they can be skipped as technicalities of secondary interest:

– $S_{rc}(F)$ is the formula obtained from F by permuting systematically the words "row" and "column",

– $S_{rn}(F)$ is the formula obtained from F by permuting systematically the words "row" and "number",

– $S_{cn}(F)$ is the formula obtained from F by permuting systematically the words "column" and "number".

As is usual in logic, the formal definitions of $S_{rc}(F)$, $S_{rn}(F)$ and $S_{cn}(F)$ are given recursively, following the general construction of a formula:

– block-free terms (notice that the sorts cannot be permuted in functions, but the subscripts on the variables are permuted instead; this is technically important, especially when we deal with transformations of formulæ with different numbers of variables of different sorts):

| F | $S_{rc}(F)$ | $S_{rn}(F)$ | $S_{cn}(F)$ |
|---|---|---|---|
| $f(n_i, r_j, c_k)$ | $f(n_i, r_k, c_j)$ | $f(n_j, r_i, c_k)$ | $f(n_k, r_j, c_i)$ |

– block-free atomic formulæ (as in functions, the sorts cannot be permuted in predicate "value", but the subscripts on the variables are permuted instead):

| F | $S_{rc}(F)$ | $S_{rn}(F)$ | $S_{cn}(F)$ |
|---|---|---|---|
| $n_i =_n n_j$ | $n_i =_n n_j$ | $r_i =_r r_j$ | $c_i =_c c_j$ |
| $r_i =_r r_j$ | $c_i =_c c_j$ | $n_i =_n n_j$ | $r_i =_r r_j$ |
| $c_i =_c c_j$ | $r_i =_r r_j$ | $c_i =_c c_j$ | $n_i =_n n_j$ |
| lk = rc | lk = rc | lk = cn | lk = rn |
| lk = rn | lk = cn | lk = rn | lk = rc |
| lk = cn | lk = rn | lk = rc | lk = cn |
| $value(n_i, r_j, c_k)$ | $value(n_i, r_k, c_j)$ | $value(n_j, r_i, c_k)$ | $value(n_k, r_j, c_i)$ |

– logical connectives: each of the logical connectives merely commutes with each of $S_{rc}$, $S_{rn}$, $S_{cn}$;

– quantifiers: they partly commute, with quantified variables exchanged:

| F | $S_{rc}(F)$ | $S_{rn}(F)$ | $S_{cn}(F)$ |
|---|---|---|---|
| $\forall n_i F, \exists n_i F$ | $\forall n_i S_{rc}(F), \exists n_i S_{rc}(F)$ | $\forall r_i S_{rn}(F), \exists r_i S_{rn}(F)$ | $\forall c_i S_{cn}(F), \exists c_i S_{cn}(F)$ |
| $\forall r_i F, \exists r_i F$ | $\forall c_i S_{rc}(F), \exists c_i S_{rc}(F)$ | $\forall n_i S_{rn}(F), \exists n_i S_{rn}(F)$ | $\forall r_i S_{cn}(F), \exists r_i S_{cn}(F)$ |
| $\forall c_i F, \exists c_i F$ | $\forall r_i S_{rc}(F), \exists r_i S_{rc}(F)$ | $\forall c_i S_{rn}(F), \exists c_i S_{rn}(F)$ | $\forall n_i S_{cn}(F), \exists n_i S_{cn}(F)$ |

Notice that the three transformations are involutive, i.e. for any block-free formula F, one has $S_{rc} \bullet S_{rc}(F) = F$, $S_{rn} \bullet S_{rn}(F) = F$ and $S_{cn} \bullet S_{cn}(F) = F$.

### 3.6.3. $S_{rcbs}$ transformation of a block-free formula

For a block-free formula F, its $S_{rcbs}$ transform is also defined recursively by:

– block-free terms (notice again that the sorts cannot be permuted in the functions, but the subscripts on the variables are permuted instead; this is technically important, especially when we deal with transformations of formulæ with different numbers of variables of different sorts):

| F | $S_{rcbs}(F)$ |
|---|---|
| $f(n_i, r_j, c_k)$ | $f(n_i, b_j, s_k)$ |

– block-free atomic formulæ:

| F | $S_{rcbs}(F)$ |
|---|---|
| $n_i =_n n_j$ | $n_i =_n n_j$ |
| $r_i =_r r_j$ | $b_i =_b b_j$ |
| $c_i =_c c_j$ | $s_i =_s s_j$ |
| lk = rc | lk = rc |
| lk = rn | lk = bn |
| lk = cn | $\bot$ |
| $value(n_i, r_j, c_k)$ | $value'[n_i, b_j, s_k]$ |

– logical connectives: all of them merely commute with $S_{rcbs}$;

– quantifiers: they partly commute, (r, c) variables being changed to [b, s]:

| F | $S_{rcbs}(F)$ |
|---|---|
| $\forall n_i F, \exists n_i F$ | $\forall n_i S_{rcbs}(F), \exists n_i S_{rcbs}(F)$ |
| $\forall r_i F, \exists r_i F$ | $\forall b_i S_{rcbs}(F), \exists b_i S_{rcbs}(F)$ |
| $\forall c_i F, \exists c_i F$ | $\forall s_i S_{rcbs}(F), \exists s_i S_{rcbs}(F)$ |

### *3.6.4. Formal symmetries between the ST axioms*

Using the above definitions, figure 3.1 shows all the symmetry, supersymmetry and analogy relationships between the four main axioms of ST.

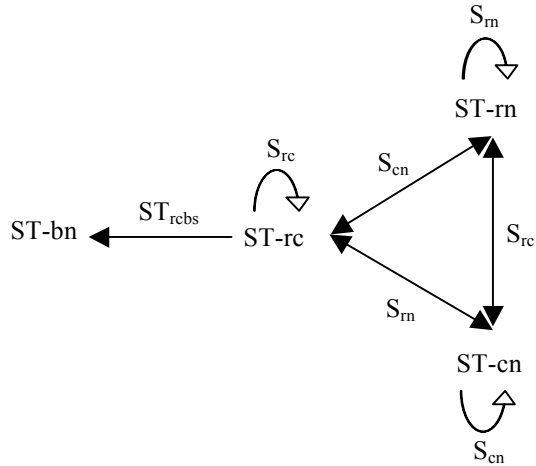

*Figure 3.1. The symmetry relationships between the ST axioms*

### **3.7. Formal relationship between Sudoku and Latin Squares**

### *3.7.1. Block-free transform of a formula*

With any formula G (not necessarily block-free) one can associate a well-defined block-free formula BF(G), called its block-free transform. It is defined recursively:
 – if G is a block-free atomic formulæ, then BF(G) is F;
 – if G is a non block-free atomic formulæ, then BF(G) is $\bot$;
 – logical connectives ¬, ∧, ∨, and ⇒ merely commute with BF;

– if G is $\forall x G_1$, then BF(G) is $\forall x BF(G_1)$ if x is a block-free variable and it is merely $G_1$ if x is a non block-free variable;

– if G is $\exists x G_1$, then BF(F) is $\exists x BF(G_1)$ if x is a block-free variable and it is merely $G_1$ if x is a non block-free variable.

Remarks:

– the last two conditions are justified by the fact that non block-free variables are eliminated together with the non block-free atomic formulæ containing them;

– for any formula G (and not only the atomic ones), if G is block-free, then BF(G) is merely G.

### 3.7.2. Formal relationship between Sudoku and Latin Squares

***Theorem 3.2: a block-free formula that is valid in ST has a block-free proof.***

As an obvious corollary, we have:

***Theorem 3.3: a block-free formula is valid for Sudoku (i.e. is a theorem of ST) if and only if it is valid for Latin Squares (i.e. it is a theorem of LST).***

Proof of theorem 3.2: Remember the standard definition of a proof of F: it is a sequence of formulæ ending with F, where each formula in the sequence either is a logical axiom or is an axiom of ST or can be deduced from the previous ones by the rules of natural deduction.

Let F be a block-free formula and consider a proof of it in ST. It suffices to show that, if we apply BF to any step in this proof, we get a block-free proof of BF(F). This is an advantage of the Gentzen's formulation of logic adopted in this book: it is obvious (though tedious to check in detail) that all the rules of natural deduction in section 3.1 are stable under the BF transformation. (See *HLS1* for a slightly less obvious proof based on Hilbert's formalism instead of Gentzen's). The proof therefore reduces to the following obvious relationship between the sets of axioms of ST and LST: BF(ST) = LST.

# 4. CSP Resolution Theories

Before we try to capture CSP Resolution Theories in a logical formalism, we must establish a clear distinction between a logical theory of the CSP itself (as it has been formulated in chapter 3, with no reference to candidates) and theories related to the resolution methods (which we consider from now on as being based on the progressive elimination of candidates). These two kinds of theories correspond to two options: are we just interested in formulating a set of axioms describing the constraints a solution of a given CSP instance (if it has any) must satisfy or do we want a theory that somehow applies to intermediate states in the resolution process? To maintain this distinction as clearly as possible, we shall consistently use the expressions "*CSP Theory*" for the first type and "*CSP Resolution Theory*" for the second type. Section 4.1 elaborates on this distinction. Since it has been shown in chapter 3 that formulating the first theory is straightforward, theories of the second kind will remain as our main topic of interest in the present book. Nevertheless, it will be necessary to clarify the relationship between the two types of theories and between their respective basic notions ("value" and "candidate").

In section 4.2, we formalise the notion of a "resolution state". This provides the intuitive notion of a candidate with a clear logical status allowing to define precise relationships between the basic formal predicates "value" and "candidate".

As the first illustration of our logical formalism, section 4.3 shows that any CSP has a minimal CSP Resolution Theory (its Basic Resolution Theory or BRT(CSP)) and it expresses its axioms in this formalism. Here, "minimal" means that all the other resolution theories introduced in this book will be obtained by adding axioms to BRT(CSP) (logically speaking, they will thus be specialisations of BRT(CSP)). Section 4.4 then defines the general concepts of a CSP Resolution Theory. Section 4.5 defines a very important property a resolution theory can have (or not), the confluence property, and it shows that BRT(CSP) has it in any CSP.

Finally, sections 4.6 and 4.7 deal with the Sudoku example. The latter proves the formal versions of the informally stated meta-theorems 2.1, 2.2 and 2.3. It also proves an extension of theorem 2.3 that will be very useful when we want to apply it in practice. Notice that, even without understanding the technicalities of their proofs, one can consider these meta-theorems as simple heuristics suggesting new potential rules and one can prove directly all the resolution rules deduced from them (this will generally be very easy).

### 4.1. CSP Theory vs CSP Resolution Theories; resolution rules

As our first approximation, we could say that a CSP Theory is about *what* we want (a complete assignment of values to the CSP variables satisfying the general CSP constraints and the specific givens), with no consideration at all for the way it can be obtained, whereas a CSP Resolution Theory is about *how* we can reach this desired final state; but then we must correct the resulting erroneous suggestion that a theory of this second kind would be mainly concerned with resolution *processes*.

To state it formally, throughout this book, the status we grant a CSP Resolution Theory is *logical*, not *operational*; and we make a clear distinction between a Resolution Theory and possible *resolution methods* that may be built as operational counterparts or algorithmic computer implementations of it (e.g. by superimposing priorities on the pure logic of the resolution rules). Such resolution methods may themselves be considered from different points of view and different kinds of logic may be used to express these. For instance, one might be interested in the dynamics of the resolution processes associated with the method, in which case one could use temporal or dynamic logic for modelling them. This is not the point of view chosen in this book, where we consider a resolution method from the point of view of the "resolution states" underlying it and we adopt modal logic (logic of necessity and possibility) to model these. However, whereas the main part of this book deals with resolution theories themselves, these theories can have properties, such as confluence, that will be shown (in chapter 5) to be very important when one wants to define and implement specific resolution methods based on them.

Then, from a logical standpoint, the only purpose of a Resolution Theory is to restrict the number of resolution states compatible with the axioms (i.e. the number of partial solutions, expressed in terms of values and candidates) and the relationships that exist between them. From an operational standpoint, it can be used as a reference for defining a resolution method that will dynamically modify the current information content; but, before a resolution theory can be used this way, there must be some operationalization process. This distinction is essential (and very classical in Artificial Intelligence) because a given set of logical axioms (a Resolution Theory) can often be operationalized in many different ways. (To be more specific: it can, for instance, be expressed as very different sets of rules in an inference engine; but it can also be implemented as a classical C program.)

Whereas CSP Theories, as developed in chapter 3 are very simple, CSP Resolution Theories require a more complex approach. All the CSP Resolution Theories should be restricted to satisfy two obvious general requirements: a) any of their rules should be a consequence of the CSP theory (under conditions, to be defined, on the relationship between values and candidates); b) they should apply to any set of givens. This is very far from being enough to constrain the possible theories of interest. But, as a consequence of these broad requirements, some aspects of CSP solving are excluded from our considerations, such as any form of psychological bias: in Sudoku, we do not take into account the physical proximity of rows or columns, although it is probably easier to see Hidden Triplets in three contiguous cells in a row than in three cells disseminated in this row; in map colouring, we forget the real shapes of the regions, although complicated shapes may make some adjacency relations more difficult to see.

## 4.2. The logical nature of CSP Resolution Theories

The analyses in this section constitute the central part of this chapter and they are the key to understanding the logical foundations of this book: given that the naive notion of a candidate is the basis for the various popular resolution rules in many logic puzzles and that it will also be the basis for the formulation of any resolution rule for any CSP, can one grant it a well defined logical status? Another point to be considered here is the relationship between the CSP Theory T(CSP), which does not use this notion, and related CSP Resolution Theories, which are based on it.

### *4.2.1. On the (non existent) problem on non-monotonicity*

Let us first clarify the following point. One apparent problem in choosing the notion of a candidate as the basis for a logical formulation is that the set of candidates for any CSP variable is monotonically decreasing throughout the resolution process, whereas logic is usually associated with monotonically increasing sets: starting from what is initially assumed to be true (the axioms), each step in a proof adds new assertions to what has been proven to be true in the previous steps; there is no possibility in standard logic for removing anything.

Do we therefore need to use some sort of non-monotonic logic, as is often the case with AI problems? Not really: instead of considering candidates for a variable, we can consider the complementary set of "not-candidates" or excluded values, i.e. values that are effectively proven to be incompatible with all that is already known (in the Sudoku case, the crossed or erased candidates in the grid on the paper sheet) – and this is a monotonically increasing set. By "effectively proven", one should understand "proven by admissible reasoning techniques" (and the sequel will show that the informal word "admissible" must in turn be understood technically as "intuitionistically valid" or, equivalently, "constructively valid").

What is really important in logic is that the abstract information content is monotone increasing with the development of the proof. (One should not confuse this information content with possibly varied representations of it.) In the sequel, when we write resolution rules, we shall conform to what we have done in *HLS* for Sudoku and we shall refer to candidates, but we must keep in mind that, when expressed with not-candidates, the underlying logic is always monotone increasing.

*4.2.2. Resolution states and resolution models*

Notwithstanding the above remarks on the informal notion of a candidate, can we grant it a precise logical status allowing us to use it consistently in the expression of the resolution rules? But, first of all, how is it related to the primary predicate "value"? Notice the vocabulary we used spontaneously: a value is asserted as being true, while a candidate is proven (or not proven) to be incompatible with all that is already proven. The most straightforward way of interpreting this is as an indication that the underlying logic of any CSP Resolution Theory based on candidates should be modal: it should be a logic of possibility/necessity as opposed to a logic of truth (such as standard logic or MS-FOL).

Before entering into the formal details, let us define the notions of a resolution state and of a resolution model. Defining the model theoretic aspects before the syntactic aspects is not the usual way to proceed in logic, but it is more intuitive.

*4.2.2.1. Resolution states*

Definitions (here, meta-variable $l°$ designates a constant symbol for a label):

– a value datum is any ground atomic formula of the kind value($l°$);

– a candidate datum is any ground atomic formula of the kind candidate($l°$);

– a *resolution state* RS is any set of value data, of candidate data and of negated candidate data; it is not necessarily devoid of (implicit) contradictions with respect to the CSP constraints, but it cannot contain both candidate($l°$) and ¬candidate($l°$) for the same label $l°$; we shall write RS $\models$ value($l°$), RS $\models$ candidate($l°$) and RS $\models$ ¬candidate($l°$) to mean respectively that the value datum is present in RS, that the candidate datum or the negated candidate datum is present in RS;

– for a resolution state RS and a label $l°$, if RS $\models$ candidate($l°$) [respectively RS $\models$ ¬candidate($l°$), RS $\models$ value($l°$)], we say informally that $l°$ is a candidate [resp. is not a candidate, is a value] in RS.

Notice that:
a) we need not consider negated value data, because value data can only be asserted;
b) instead of considering the absence of a candidate from RS (which could have an ambiguous interpretation), we consider the presence of its negation (the positive fact that the candidate has been "effectively eliminated" from RS).

Any resolution state is a finite set and the whole set **RS** of resolution states is therefore finite (and independent of any particular instance of the CSP) although very large.

As suggested in part by the name, a resolution state is intended to represent the totality of the ground atomic facts and their negations (in terms of value and candidate predicates) that are present in some possible state of reasoning for some instance of the CSP. This is what we called informally the information content of this state – in which all the "static" knowledge about the CSP, such as links between labels, is considered as background knowledge and is not explicitly listed, but is implicitly present. In the Sudoku CSP, a resolution state is a straightforward abstraction for something very concrete: the set of decided values, of candidates still present on the sheet of paper used to solve a puzzle and of candidates erased or crossed. (And the structure of the grid remains implicit.)

Vocabulary: if RS is a resolution state, "*a candidate l in RS*" is an informal way of saying "a label l such that RS $\models$ candidate(l)". Similarly, "a value in RS" is a way of saying "a label l such that RS $\models$ value(l)".

*4.2.2.2. Resolution models*

In order to be able to give the above interpretation of a resolution state in a way that respects our resolution paradigm, we must add some structure on the set **RS** of all the resolution states and on the way they are related. On **RS**, we define a natural partial order relation: $RS_1 \leq RS_2$ if and only if, for any constant symbol l° for a label, one has:

– if $RS_1 \models$ value(l°), then $RS_2 \models$ value(l°), (assertion/addition of a value is not reversible),

– if $RS_1 \models \neg$candidate(l°), then $RS_2 \models \neg$candidate(l°) (negation/deletion of a candidate is not reversible),

– if $RS_2 \models$ candidate(l°), then $RS_1 \models$ candidate(l°) (new candidates cannot appear or re-appear in a posterior resolution state).

Thus, the intended meaning of $RS_1 \leq RS_2$ is that when one passes from one resolution state to a "greater" or "posterior" one (according to this abstract order relation), the information content can only increase – the negation of a candidate being considered as an increase of this information content. The last condition says that no candidate absent from a resolution state can (re-)appear in a posterior one. In practical terms, it also means that $RS_2$ is closer to a solution (or to the detection of a contradiction) than $RS_1$ is.

Now, with any instance P of the CSP (considered as defined by a set of labels), one can associate a unique well-defined resolution state $RS_P$, called the initial resolution state of P, in which:

– for every given l° in P, $RS_P \models$ value(l°),

– for every label $l^1$ which has no direct contradiction with any of the givens l° of P, i.e. such that linked(l°, $l^1$) is not in the background axioms for any given l° of P, $RS_P \models$ candidate($l^1$),

– $RS_P$ contains no other value or candidate data than those defined above (in particular, it contains no negated candidate data).

The resolution model of an instance P is then defined as the subset $\mathbf{RS_P}$ of $\mathbf{RS}$ (together with the order relation induced by $\mathbf{RS}$) consisting of all the resolution states RS such that $RS_P \leq RS$. When trying to solve P, one can never escape $\mathbf{RS_P}$, at least as long as one reasons consistently. Any solution of P must be in $\mathbf{RS_P}$ and it can only be a maximally consistent element of $\mathbf{RS_P}$. But, conversely, a maximally consistent element of $\mathbf{RS_P}$ is not necessarily a solution (especially in case there is no solution). By exploring systematically all the states in $\mathbf{RS_P}$, one is certain either to prove that P has no solution or to find all the solutions of P, if P has any. Of course, to find a solution, one does not have to explore all of $\mathbf{RS_P}$. In some sense, the purpose of a resolution theory is to define a smart way of reducing $\mathbf{RS_P}$ to a relevant part as small as possible (without excluding any parts that may lead to a solution).

Our definition of $\mathbf{RS_P}$ already includes the deletion of candidates obviously contradictory with the givens of the problem instance. This amounts to restricting from the start the resolution model $\mathbf{RS_P}$ of P to a relevant part.

*4.2.2.3. Remarks on the notions of a resolution state and a resolution model*

Notice that the above notions of a resolution state and a resolution model are very narrow. For instance, a resolution state does not include any "mental" component such as having identified a pattern corresponding to the preconditions of a resolution rule. Similarly, the resolution model $\mathbf{RS_P}$ of an instance P defines only an abstract order relation on the set of resolution states reachable from the initial state $RS_P$, it does not indicate *how* to pass from one state to a posterior one. But this is the only way one can build a consistent semantics in case an instance has zero or several solutions.

Simplistic as they may seem, the above-defined notions allow us to state precisely what kind of resolution rules we are looking for. Given a resolution theory T, the application of any resolution rule R in T to an instance P should lead from one resolution state in $\mathbf{RS_P}$ to a posterior one, with the following interpretation: if, starting from a resolution state RS in $\mathbf{RS_P}$, we notice a pattern (or configuration) of labels, links, values and candidates, satisfying the condition part of R, then R can be applied to this pattern; and, if we apply it, then, in the resulting resolution state $RS_1$ and in all the subsequent ones (still in $\mathbf{RS_P}$), the value(s) and candidate(s) specified in the action part of R will respectively be asserted and negated (in a resolution rule, values can only be asserted, candidates can only be negated). Notice that the whole process of detecting a pattern, applying a rule and passing from RS to $RS_1$ is superimposed on $\mathbf{RS_P}$ but is not part of this abstract static model.

Now, still starting from the same resolution state RS, if we notice that the conditions of another resolution rule R' in T are also satisfied in RS and if we apply R' instead of R, we usually reach a resolution state $RS_2$ (still in $\mathbf{RS_P}$) different from $RS_1$. For a real understanding of what a resolution theory is and is not, it is crucial to remark that the (relatively informal) definition we have just given does not *a priori* imply that the two states $RS_1$ and $RS_2$ are T-compatible, in the sense that there would be a resolution state $RS_3$ posterior to both $RS_1$ and $RS_2$ (i.e. such that $RS_1 \leq RS_3$, $RS_2 \leq RS_3$) and accessible from each of $RS_1$ and $RS_2$ *via rules in T* (see Figure 4.1). This is related to the fundamental question of the confluence property of a resolution theory T (see section 4.5 for a definition and an example of a theory with the confluence property).

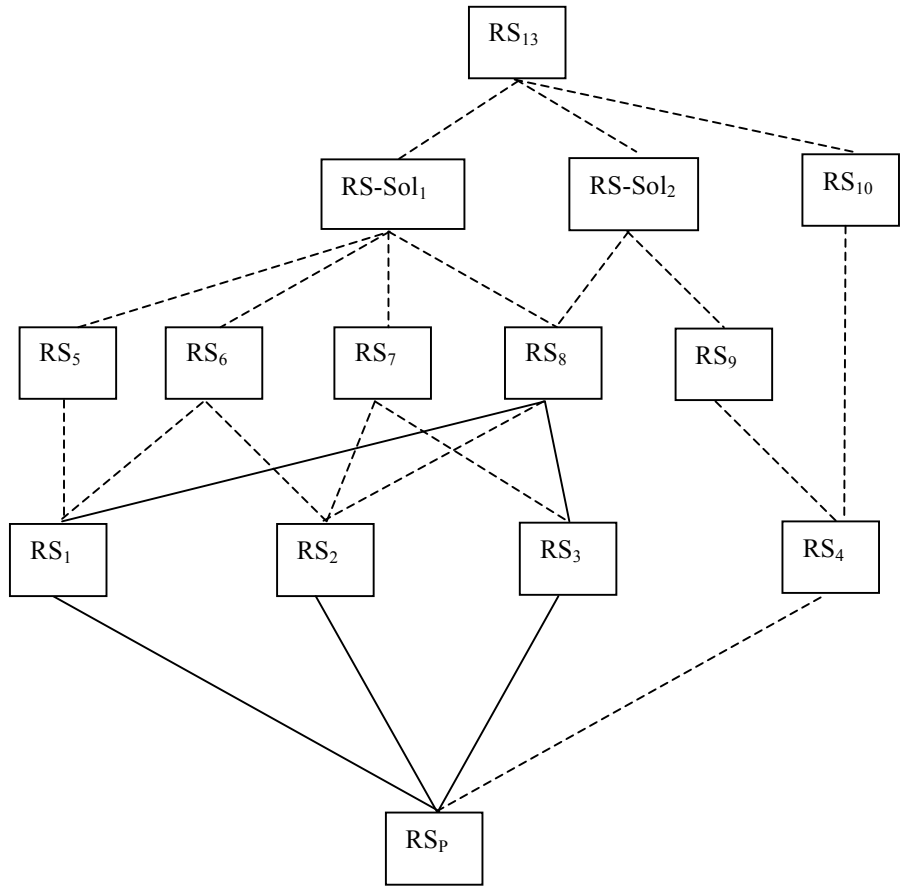

***Figure 4.1.*** *The resolution model **RS$_P$** of an instance P with two solutions (RS-Sol$_1$ and RS-Sol$_2$) and the part of it accessible by some Resolution Theory T (full lines). Notice that the resolution states RS$_1$ and RS$_2$ (or RS$_2$ and RS$_3$) are not T-compatible, but RS$_1$ and RS$_3$ are.*

*4.2.3. Logical interpretations of a resolution model*

There are two possible logical interpretations of the above notions. The most straightforward one is in terms of modal logic. [In *HLS*, we used epistemic, instead of modal, logic; but the final interpretation of resolution theories (intuitionistic or constructive logic) is the same.]

*4.2.3.1. The modal interpretation of a resolution model*

Our notions of a resolution state and a resolution model appear to be a special case of the classical notions of a possible world and a Kripke model in modal logic.

In modal logic, there is a modal operator "□" of necessity (and a modal operator "◊" of possibility, which does not always appear explicitly, because it is equivalent to ¬□¬ in the most common modal theories); for any formula A, □A and ◊A are intended to mean respectively "A is necessary" and "A is possible".

Our notion of a resolution model coincides with that of a canonical Kripke model and the order relation we have defined on the set of resolution states corresponds to the accessibility relation between possible worlds in this model ([Kripke 1963]). We can apply Hintikka's interpretation of "□" ([Hintikka 1962]): RS |= □A if and only if RS' |= A for any possible world RS' accessible from RS (i.e., in our resolution model, such that RS ≤ RS').

Which (propositional) logical axioms for the modal operator □ should one adopt? This is the subject of much philosophical and scientific debate. It concerns the general relationship between truth, necessity and possibility and the axioms expressing this relationship. There are several modal theories in competition, the most classical of which are, in increasing order of strength: S4 < S4.2 < S4.3 < S4.4 < S5 (on this point and the following, see e.g. [Feys 1965], [Fitting et al. 1999] or the Stanford Encyclopaedia of Philosophy: http://plato.stanford.edu/entries/logic-modal/).

Moreover, it is known that there is a correspondence between the axioms on □ and the properties of the accessibility relation between possible worlds (this is a form of the classical relationship between syntax and semantics). A very general expression of this correspondence was obtained by [Lemmon et al. 1977].

Here, we shall adopt the following rule of inference and set of axioms (in addition to the usual axioms of classical logic), which constitute the (most commonly used) propositional system S4 (we give them the names they are classically given in modal logic and, because of axioms M and 4, we write the accessibility relation "≤"):

– (Necessitation Rule) if A is a theorem, then so is □A;
– (Distribution Axiom) □(A ⇒ B) ⇒ (□A ⇒ □B);

4. CSP Resolution Theories                                                             83

– (axiom M) □A ⇒ A: "if a proposition is necessary then it is true" or "only true propositions can be necessary"; this axiom corresponds to the accessibility relation being reflexive (for all RS in **RS**, one has: RS ≤ RS);

– (axiom 4, reflection) □A ⇒ □□A: if a proposition is necessary then it is necessarily necessary; this axiom corresponds to the accessibility relation being transitive (for all $RS_1$, $RS_2$ and $RS_3$ in **RS**, one has: if $RS_1 ≤ RS_2$ and $RS_2 ≤ RS_3$, then $RS_1 ≤ RS_3$).

From our definition of a resolution model, it can easily be checked that it satisfies all the axioms of S4.

As for the predicate calculus part of our logic, quantifiers are generally a big problem in modal logic. But we must notice that in our CSPs we deal only with fixed domains; there is therefore no problem with quantifiers: we can merely adopt as axioms both Barcan Formula (BF) and its converse (CBF) ([Barcan 1946a and 1946b]), namely:

– (BF) ∀x□A ⇒ □∀xA,

– (CBF) □∀xA ⇒ ∀x□A.

One final thing should be noted: in modal logic, for any *ground atomic* formula A, "A ∨ ¬A" is true in any resolution state and it is also necessarily true, i.e. one always has RS |= □( A ∨ ¬A), but this is not the case for "□A ∨ □¬A". For instance, given some definite place in space-time, it is always true that either it is raining (A) or it is not raining (¬A) at this place, and this is necessarily true (□(A ∨ ¬A)). But it is not true that either it is necessarily raining (□A) or it is necessarily not raining (□¬A) at this place: the weather may change at this place. Said otherwise, "□¬A" (A is necessarily false) and "¬□A" (A is not necessarily true) are very different things and the first is much stronger than the second.

*4.2.3.2. The intuitionistic interpretation of a resolution model*

So far so good; but we are not very enthusiastic with the prospect of having to overload the formulation of our resolution rules with modal operators. Let us try to do one more step.

There is a well-known correspondence ([Fitting 1969]) between modal logic S4 and intuitionistic or constructive logic ([Bridges et al. 2006]). The language of a theory in intuitionistic logic is the same as in classical logic (there is no □ or ◊ logical operator). Given a formula A in intuitionistic logic, one can define a formula M(A) in S4 recursively by:

– for A atomic: M(A) = □A,

– M(A ∧ B) = M(A) ∧ M(B),

– M(A ∨ B) = M(A) ∨ M(B),

– $M(\neg A) = \Box \neg M(A)$,
– $M(A \Rightarrow B) = \Box(M(A) \Rightarrow M(B))$,
– $M(\forall x A) = \forall x M(A)$.

Then, for every formula F with no modal operator, one has the well-known correspondence theorem (proven in any textbook on modal logic): ***F is a theorem in intuitionistic logic if and only if M(F) is a theorem in modal logic S4***.

In intuitionistic logic, although the formulæ are the same as in classical logic, their informal interpretation is different:

– A means that A is effectively proven;

– ¬A means that A is effectively proven to be contradictory;

– ¬¬A is not equivalent to A; it is weaker than A; it means that it is not effectively proven that A is contradictory (which does not imply that A is proven).

One main difference with classical logic is the "law of the excluded middle": $A \lor \neg A$ is not valid (when A is atomic, it corresponds to formula $\Box A \lor \Box \neg A$ in S4). $A \lor \neg A$ would mean that either A is proven or ¬A is proven. But there are propositions for which this is not true. Similarly, $\exists x A$ is stronger than $\neg \forall x \neg A$; $\exists x A$ means that a proof has effectively produced some x and it has shown that it satisfies A; $\neg \forall x \neg A$ only supposes that $\forall x \neg A$ leads to a contradiction.

The question for us is now: can we adopt intuitionistic instead of modal logic? It amounts to: can each of our resolution rules be written in the form M(A) for some formula A without modal operators? This raises the question of the intended meaning of the resolution rules.

### *4.2.4. Resolution theories are intuitionistic*

Anticipating on our resolution rules (which will not refer explicitly to resolution states), in their naive formulations, their (non static) conditions will bear on the presence of some candidates and on the absence of others and their conclusions will always be the assertion of a value or the elimination of a candidate.

#### *4.2.4.1. Analysing the intended meaning of resolution rules*

Let us see how this can be used in the formulation of a CSP resolution theory:

– first, the entries of a CSP instance P, which are axioms, can be understood as necessarily true (in a formal way by the Necessitation rule, or in a semantic way because they will be present in all the resolution states): □value; this can be written as M(value), because "value" is atomic; intuitionistically, this is merely the tautology that axioms of T are effectively proven in T.

As for the resolution rules themselves:

– as links are part of the CSP structural background, they are also axioms of any Resolution Theory and a condition on the presence of a link between two labels can be understood as necessarily true (by the Necessitation rule): □linked-by($l_1$, $l_2$, c); this can be written as M(linked-by($l_1$, $l_2$, c)), because "linked-by" is atomic; using Barcan formula, the same conclusion is valid for predicate "linked";

– a negative condition on a candidate [i.e. a condition ¬candidate(l)] in a resolution state RS implies that it is negated in any posterior resolution state; semantically, it must therefore be interpreted as: □¬candidate(l); this can be written as M(¬candidate(l)); intuitionistically, this means that this candidate has effectively been proven to be contradictory;

– a positive condition on a candidate in a resolution state RS could be intended to mean (in the modal sense) that "this label is still a possible value in RS": ◊value(l); but one should here anticipate on the final intended intuitionistic meaning: "this label l has not yet been effectively proven to be an impossible value"; therefore, one should rather interpret such a condition in the sense of ¬¬value(l) (in the intuitionistic meaning of it); in relation to the modal setting, this would appear to have for M transform the stronger □¬□¬value(l) or □◊value(l); (see section 4.2.4.3 below for comments);

– any ∧ and ∨ combination of such conditions remains of the form M(some formula with no □ symbol);

– a conclusion on the assertion of a value is intended to mean that the value becomes necessarily true: □value; this can be written as M(value), because "value" is atomic;

– a conclusion on the elimination of a candidate is intended to mean that this candidate becomes necessarily contradictory: □¬candidate; this can be written as M(¬candidate);

– any ∧ combination of such conclusions remains of the form M(some formula with no □ symbol);

– again by the Necessitation rule, the implication sign appearing in a resolution rule Cond ⇒ Act (which is an axiom in a Resolution Theory) can be understood as necessary: □(Cond ⇒ Act); this can be written as M(Cond ⇒ Act).

– finally, if the whole resolution rule ∀xR is surrounded by ∀ quantifiers, where R = M(A), it can be written as M(∀xA).

*4.2.4.2. Resolution rules pertain to intuitionistic instead of classical logic*

The above analysis shows that a resolution rule will always be of the form M(F) with no □ symbol in F. The general conclusion of all this is that a resolution rule is always the M transform of an MS-FOL formula and the MS-FOL formula can be used instead of the modal form, provided that we consider that ***Resolution Theories pertain to intuitionistic (or constructive) logic***.

*4.2.4.3. The meaning of positive conditions on candidates in resolution rules*

Our interpretation of a positive condition on a candidate in the condition part of a resolution rule is worth some discussion. Our intuitionistic interpretation of "candidate" as "¬ ¬ value", corresponding to the modal interpretation $\Box\Diamond$value, rather than adopting the seemingly more natural (from the modal point of view) $\Diamond$value, is consistent with our definition of the order relation on **RS**: once a candidate has been eliminated, it can no longer re-appear in a posterior resolution state. So that, for any label l and resolution state RS, one can have RS $\models$ $\Diamond$candidate(l) only if RS $\models$ candidate(l), i.e. if l is effectively present in RS as a candidate, which in turn implies that RS $\models$ $\Box\Diamond$candidate(l).

Notice that the definition of **RS** and this interpretation together put a strong restriction on how resolution rules can be applied in a resolution state RS: a pattern mentioning non-negated candidates may only be instantiated if such candidates are effectively present in this resolution state. The condition part of the rule thus means: the pattern defined by this rule can be considered as present in RS only if the following candidates are still present in RS (i.e. have not yet been proven to be contradictory) and the other conditions of the rule are satisfied. From a computational point of view, the positive aspect is that, as candidates are progressively eliminated, it puts stronger and stronger conditions on patterns and it makes their potential number decrease while the resolution process goes on.

**4.3. The Basic Resolution Theory of a CSP: BRT(CSP)**

We can now define formally the Basic Resolution Theory of any CSP: BRT(CSP). Its logical language is an extension of the language defined in section 3.2 for the CSP Theory T(CSP). In addition to it, it has only:

– two 0-ary predicates: solution-found and contradiction-found,

– a unary predicate: candidate, with signature (Label).

As for the axioms of BRT(CSP), they include all (the implicit sort axioms and) the background axioms of the CSP Theory defined in section(s) (3.2.2 and) 3.2.3. They cannot include the CSP constraint axioms of section 3.2.4 because these do not have the structure required of resolution rules: "meaning of links as constraints" is of the condition-action type, but it has the negation of a value in its conclusion (in a resolution rule, a value can never be negated); "completeness of solution" is not of the condition-action type. Instead, they contain the following:

– **ECP (Elementary Constraints Propagation)**: "if a value is asserted for a CSP variable (as is initially the case for the givens), then remove any candidate that is linked to this value by a direct contradiction":

**ECP:** ∀l₁ ∀l₂ {value(l₁) ∧ linked(l₁, l₂) ⇒ ¬ candidate(l₂)};
this is very close to "meaning of links as constraints", but the conclusion is about a candidate instead of a value;

– **S (Single)**: "if a CSP variable V has only one candidate <V, v> left, then assert it as the value of this variable":

**S:** ∀l ∀V ∀v { [label(l, V, v) ∧ candidate(l)
          ∧ ∀v'≠v ∀l'≠l (¬ label(l', V, v') ∨ ¬ candidate(l'))] ⇒ value(l) };

this rule has no equivalent in the CSP Theory.

Axioms ECP and S together establish the correspondence between predicates "value" and "candidate". We define the set of value-candidate relationship axioms as **VCR = ECP ∪ S**.

BRT(CSP) also has a few technical axioms:
– **OOS (Only One Status)**: "when a label is asserted as a value, it is no longer a candidate" (this rule has no equivalent in the CSP Theory):

**OOS:** ∀l {value(l) ⇒ ¬ candidate(l)};

– **SD (Solution Detection)**: "if all the CSP variables have a unique decided value, then the problem is solved":

**SD:** ∀V ∃!v ∃l {[label(l, V, v) ∧ value(l)] ⇒ solution-found()};

– **CD (Contradiction Detection)**: "if there is a CSP variable with no decided value and no candidate left, then the problem has no solution":

**CD:** ∃V ∀v ∃l {[label(l, V, v) ∧ ¬ value(l) ∧ ¬ candidate(l)]
          ⇒ contradiction-found()}.

Predicates "solution-found" and "contradiction-found" as well as rules SD and CD are not strictly necessary, but they illustrate how such situations can be written as resolution rules. They can be considered as hooks for external non-logical actions (such as displaying the solution). ⊥ could be used instead of contradiction-found.

Finally, we define:

**BRT(CSP) = {background axioms} ∪ ECP ∪ S ∪ {OOS, SD, CD}.**

Two questions immediately come to mind. Can one solve all the instances of the CSP with only BRT(CSP)? No. How powerful is this Basic Resolution Theory? Just to give an idea, in Sudoku (with the strongest formulation including all the $X_{rc}$, $X_{rn}$, $X_{cn}$, $X_{bn}$ variables), it allows to solve about 29% of the minimal puzzles; notice that, if we considered only the $X_{rc}$ variables, very few minimal puzzles could be solved.

**4.4. Formalising the general concept of a Resolution Theory of a CSP**

Let us now state our final formal definitions. Given a CSP:

– a formula in the language of the CSP Basic Resolution Theory defined above, BRT(CSP), is said to be in the ***restricted condition-action form*** if it is written as $A \Rightarrow B$, possibly surrounded with universal quantifiers, where formula A does not contain the "$\Rightarrow$" sign and formula B is either value(z) or ¬candidate(z) for some variable z of sort Label, called the target of the rule, that already appears in the condition part (one can act only on what has been previously identified);

– a ***resolution rule*** is a formula written *in the restricted condition-action form*, with no constant symbols other than those already present in the constraint axioms of T(CSP), if any, and ***provable in the intuitionistic theory T(CSP) ∪ {ECP, S}***, i.e. the union of the CSP Theory (now considered as an intuitionistic theory) and the axioms on the value-candidate relationship;

– a resolution rule is ***instantiated*** in some resolution state RS when a value has been assigned to each of its variables in such a way that RS satisfies all the conditions of this rule; the rule can thus be applied; after its action part has been applied, another resolution state is reached in which its conclusion is valid;

– the condition part of a resolution rule is composed of two subparts: the pattern-conditions and the target-conditions;

– the ***pattern-conditions*** describe (in terms of labels, of well defined links between some of these labels and of value and candidate predicates for these labels) a factual situation that may occur in a resolution state (some of these conditions may depend on the target z);

– the ***target-conditions*** bear on label variable z; they always include the actual presence of this candidate in the resolution state (one cannot assert or eliminate something that is not present as a candidate; said otherwise, it is absurd to assert something that has already been proven to be impossible and it is useless to negate something that has already been negated); expressed in terms of its links with other labels mentioned in the pattern, they specify the conditions under which, in the action part of the rule, this candidate can be negated or asserted as a value;

– a ***Resolution Theory*** for a CSP is a *specialisation of its Basic Resolution Theory* in which all the additional axioms are *resolution rules*; it must be understood as a theory in intuitionistic logic.

In order to be concretely used to solve some instance of a CSP, a Resolution Theory must be completed with the same instance axioms as the corresponding

T(CSP) theory (see section 3.5). Nothing guarantees that a resolution theory can solve all the instances of the CSP, not even those that have a unique solution.

One immediate consequence of this definition is that the general-purpose search algorithms – depth-first seach (DFS), breadth-first search (BFS), etc. – which are guaranteed to find a solution or to prove a contradiction, cannot in general be replaced by any "equivalent" resolution theory, i.e. one that would always produce the same results. The reason is obvious if one considers instances of the CSP that have multiple solutions: DFS or BFS will always find a solution, whereas a logical theory can only prove properties (here value assertions and candidate eliminations) that are true in all its models and it cannot therefore find a solution.

## 4.5. The confluence property of resolution theories

The confluence property is one of the most useful properties a resolution theory T can have. It justifies our principle according to which the instantiation of a rule in some resolution state RS depends on the effective presence of some candidates in RS (instead of depending only on relations between underlying labels); moreover, it allows to superimpose on T different resolution strategies.

### 4.5.1. Definition of the confluence property

Given a resolution theory T, consider all the strategies that can be built on it, e.g. by defining various implementations with different priorities on the rules in T. Given an instance P of the CSP and starting from the corresponding resolution state $RS_P$, the resolution process associated with a strategy S built on T consists of repeatedly applying resolution rules from T according to the additional conditions (e.g. the priorities) introduced by S. Considering that, at any point in the resolution process, different rules from T may be applicable (and different rules will be applied) depending on the chosen strategy S, we may obtain different resolution paths starting from $RS_P$ when we vary S.

Definition: a CSP Resolution Theory T has the *confluence property* if, for any instance P of the CSP, any two resolution paths in T can be extended in T to meet in a common resolution state.

When a resolution theory has the confluence property, all the resolution paths starting from $RS_P$ and associated with all the strategies built on T will lead to the same final state in **$RS_P$** (all explicitly inconsistent states are considered as identical; they mean contradictory constraints). If a resolution theory T does not have the confluence property, one must be careful about the order in which they apply the resolution rules (and they must try all the resolution paths if they want to find the "simplest"). But if T has this property, one may choose any resolution strategy, which makes finding a solution much easier, and one can define "simplest first" strategies if they want to find the simplest solution (see chapters 5 and 7).

Equivalent definitions:

– for any instance P of the CSP and any two resolution states $RS_1$ and $RS_2$ of P reachable from $RS_P$ by resolution rules in T, there is a resolution state $RS_3$ such that $RS_3$ is reachable independently from both $RS_1$ and $RS_2$ by resolution rules in T;

– for any instance P of the CSP, the subset of **$RS_P$** consisting of the resolution states for P reachable by resolution rules in T, ordered by the reachability relation defined by T, is a DAG (Directed Acyclic Graph).

Consequence: if a resolution theory T has the confluence property, then for any instance P of the CSP, there is a single final state reachable by rules in T and all the resolution paths lead to this state. In particular, if T solves P, one cannot miss the solution by choosing to apply the "wrong" rule at any time.

The following property, *a priori* stronger than confluence, will often be useful to prove the confluence property of a resolution theory.

Definition: a CSP resolution theory T is *stable for confluence* if for any instance P of the CSP, for any resolution state $RS_1$ of P and for any resolution rule R in T applicable in state $RS_1$ for an elimination of a candidate Z, if any set Y of consistency preserving assertions and/or eliminations is done before R is applied, leading to a resolution state $RS_2$, and if it destroys the pattern of R (R can therefore no longer be applied to eliminate Z), then, there always exists a sequence of rules in T that will eliminate Z starting from $RS_2$ (if Z is still in $RS_2$). (Remark: in this definition, the assertions or eliminations in Y are not necessarily done by rules in T.)

It is obvious that: ***if T is stable for confluence, then T has the confluence property***. A result that will be useful in Part III is the following (obvious):

***Lemma 4.1: Let $T_1$ and $T_2$ be two resolution theories. If $T_1$ and $T_2$ are stable for confluence, then the union of $T_1$ and $T_2$ (considered as sets of rules) is stable for confluence (and therefore it has the confluence property).***

*4.5.2. The confluence property of BRT(CSP)*

The following obvious case will be useful in many places, e.g. for defining T&E in section 5.5.

***Theorem 4.1: The Basic Resolution Theory of any CSP, BRT(CSP), is stable for confluence and it has the confluence property***.

*4.5.3. Resolution strategies and the strategic level*

There are the resolution theories defined above and there are the many ways one can use them in practice to solve real instances of a CSP. From a strict logical standpoint, all the rules in a resolution theory are on an equal footing, which leaves no possibility for ordering them. But, when it comes to the practical exploitation of resolution theories and in particular to their implementation, e.g. in an inference engine (as in our general CSP-Rules solver) or in any procedural algorithm, one question remains unanswered: can superimposing some ordering on the set of rules (using priorities or "saliences") prevent us from reaching a solution that the choice of another ordering might have made accessible? With resolution theories that have the confluence property, such problems cannot appear and one can take advantage of this to define different resolution strategies.

Indeed, the confluence property allows to define a *strategic level* above the *logic level* (the level of the resolution rules) – which is itself above the implementation level in case the rules are implemented in a computer program of any kind.

*Resolution strategies* based on a resolution theory T can be defined in different ways and may correspond to different goals:

– implementation efficiency (in terms of speed, memory, …);

– giving a preference to some patterns over other ones: preference for bivalue-chains over whips, for whips over braids (see chapter 5 for the definitions);

– allowing the use of heuristics, such as focusing the attention on the elimination of some candidates (e.g. because they correspond to a bivalue variable or because they seem to be the key for further eliminations); but good heuristics are hard to define (in particular, the popular, intuitively natural heuristics consisting of focusing the attention on bivalue variables is blatantly unfit for hard Sudoku puzzles);

– finding the "simplest" resolution path and computing the rating of the instance according to some rating system; this will be the justification for the "simplest-first" resolution strategies we shall introduce later; notice that this goal will in general be in strong opposition to a goal of pure implementation efficiency.

**4.6. Example: the Basic Sudoku Resolution Theory (BSRT)**

After all the above general considerations, time has come to turn to the concrete Sudoku example and to its Basic Resolution Theory, hereafter named BSRT. It will follow the general theory above, with the same adaptations as in ST for taking the basic sorts and their symmetries into better account.

*4.6.1. Sorts, functions and predicates*

As in the above general theory, the logical language of BSRT has the same sorts, functions and predicates as ST. In addition, it has predicates "solution-found", "contradiction-found" and "candidate". Indeed, as in the case of "value" in ST, we introduce a predicate **candidate** with signature (Number, Row, Column) and an auxiliary predicate **candidate'** with signature (Number, Block, Square) defined by the "change-of-coordinates axiom":

**CC': $\forall n \forall b \forall s$ [candidate'[n, b, s] $\Leftrightarrow$ candidate(n, row(b,s), column(b,s))].**

As can be seen from the signatures of predicates "value" and "candidate", they will be the basic support for the quasi-automatic expression of symmetry and super-symmetry in the Sudoku Theory and in all the Sudoku Resolution Theories.

*4.6.2. The axioms of Basic Sudoku Resolution Theory (BSRT)*

BSRT is defined *a priori* as being composed of the axioms of SGT plus CC, CC' and the following fourteen resolution rules.

The first group of four axioms expresses the mutual exclusion conditions on cells, rows, columns and blocks. They correspond to the ECP rule of the general theory (cut into four parts according to the type of constraint: rc, rn, cn or bn). These four rules, the elementary constraints propagation rules, can be considered as the direct operational transpositions of axioms ST-rc to ST-bn of ST. They can be used in practice to eliminate candidates as soon as a value is asserted. In this respect, they will be much more useful than rules such as ST-rc to ST-bn could be:

– **ECP(cell)**: unique value in a cell: if a number is effectively proven to be the value of a cell, then any other number is effectively proven to be excluded for this cell:

$\forall r \forall c \forall n \forall n'\{$value(n, r, c) $\wedge$ n'$\neq$n $\Rightarrow$ $\neg$ candidate(n', r, c)$\}$;

– **ECP(row)**: unique value in a row: if a number is effectively proven to be the value of a cell, then it is effectively proven to be excluded for any other cell in this row:

$\forall r \forall n \forall c \forall c'\{$value(n, r, c) $\wedge$ c'$\neq$c $\Rightarrow$ $\neg$ candidate(n, r, c')$\}$;

– **ECP(col)**: unique value in a column: if a number is effectively proven to be the value of a cell, then it effectively proven to be excluded for any other cell in this column:

$\forall c \forall n \forall r \forall r'\{$value(n, r, c) $\wedge$ r'$\neq$r $\Rightarrow$ $\neg$ candidate(n, r', c)$\}$;

– **ECP(blk)**: unique value in a block: if a number is effectively proven to be the value of a cell, then it is effectively proven to be excluded for any other cell in this block:

$$\forall b \forall n \forall s \forall s' \{\text{value'}[n, b, s] \wedge s' \neq s \Rightarrow \neg \text{candidate'}[n, b, s']\}.$$

The second group of four axioms corresponds to the S rule of the general theory (again cut into four parts according to the type of constraint: rc, rn, cn or bn):

– **NS** or Naked-Single: assert a value whenever there is a unique possibility in an rc-cell:
$$\forall r \forall c \forall n \{[\text{candidate}(n, r, c) \wedge \forall n' \neq n \neg \text{candidate}(n', r, c)] \Rightarrow \text{value}(n, r, c)\};$$

– **HS(row)** or Naked-Single-in-a-row: assert a value whenever there is a unique possibility in an rn-cell:
$$\forall r \forall n \forall c \{[\text{candidate}(n, r, c) \wedge \forall c' \neq c \neg \text{candidate}(n, r, c')] \Rightarrow \text{value}(n, r, c)\};$$

– **HS(col)** or Naked-Single-in-a-column: assert a value whenever there is a unique possibility in a cn-cell:
$$\forall c \forall n \forall r \{[\text{candidate}(n, r, c) \wedge \forall r' \neq r \neg \text{candidate}(n, r', c)] \Rightarrow \text{value}(n, r, c)\};$$

– **HS(blk)** or Naked-Single-in-a-block: assert a value whenever there is a unique possibility in a bn-cell:
$$\forall b \forall n \forall s \{[\text{candidate'}[n, b, s] \wedge \forall s' \neq s \neg \text{candidate'}[n, b, s']] \Rightarrow \text{value'}[n, b, s]\}.$$

The ninth axiom is the general axiom about uniqueness of status:

– **OOS (Only One Status)**: "when a label is asserted as a value, it is no longer a candidate":
$$\forall n \forall r \forall c \{\text{value}(n, r, c)] \Rightarrow \neg \text{candidate}(n, r, c)\};$$

The tenth axiom expresses solution detection (there could also be four axioms):

– **SD**: if every rc-cell has a value assigned, then the problem is solved:
$$\forall r \forall c \exists n \, \text{value}(n, r, c) \Rightarrow \text{solution-found}();$$

The last group of four axioms expresses contradiction detection (these axioms are redundant, but it is easier to have them all if we want to apply to Sudoku the general correspondence between braids and T&E in section 5.7):

– **CD-rc**: if there is an rc-cell such that all the numbers are proven to be excluded values for it, then the puzzle has no solution:
$$\exists r \exists c \forall n \, [\neg \text{value}(n, r, c) \wedge \neg \text{candidate}(n, r, c)] \Rightarrow \text{contradiction-found}();$$

– **CD-rn**: if there is an rn-cell such that all the columns are proven to be excluded values for it, then the puzzle has no solution:
**∃r∃n∀c [¬ value(n, r, c) ∧ ¬ candidate(n, r, c)] ⇒ contradiction-found()**;

– **CD-cn**: if there is a cn-cell such that all the rows are proven to be excluded values for it, then the puzzle has no solution:
**∃c∃n∀r [¬ value(n, r, c) ∧ ¬ candidate(n, r, c)] ⇒ contradiction-found()**;

– **CD-bn**: if there is a bn-cell such that all the squares are proven to be excluded values for it, then the puzzle has no solution:
**∃b∃n∀s (¬ value'[n, b, s] ∧ ¬ candidate'[n, b, s)]) ⇒ contradiction-found()**.

Finally, we define the same sets of axioms as in the general theory (plus those associated with the existence of a double coordinate system):
ECP = {ECP(cell), ECP(row), ECP(col), ECP(blk)},
S = {NS, HS(row), HS(col), HS(blk)},
CD = {CD-rc, CD-rn, CD-cn, CD-bn},
VCR = ECP ∪ S (the value-candidate relationship axioms),
BSRT = SGT ∪ {CC, CC'} ∪ ECP ∪ S ∪ CD ∪ {OOS, SD}.

### 4.6.3. The axiom associated with the entries of a puzzle

As was the case for Sudoku Theory ST, with any specific puzzle P we can associate the axiom $E_P$ defined as the finite conjunction of all the formulæ of type value($n_k$, $r_i$, $c_j$) corresponding to each entry of P. Then, when added to the axioms of BSRT (or any extension of it), axiom $E_P$ defines a Sudoku Resolution Theory for the specific puzzle P.

### 4.6.4. The Basic LatinSquare Resolution Theory: BLSRT

Let us define the following sets of block-free axioms:
B(ECP) = {ECP(cell), ECP(row), ECP(col)},
B(S) = {NS, HS(row), HS(col)},
B(VCR) = B(ECP) ∪ B(S) (the block-free value-candidate relationship axioms),
BLSRT = LSGT ∪ B(ECP) ∪ B(S) ∪ {OOS, CD, SD}.

BLSRT is the Basic LatinSquare Resolution Theory: BRT(LatinSquare)

### 4.7. Sudoku symmetries and the three fundamental meta-theorems

Let us first extend the definition of the $S_{rc}$, $S_{rn}$, $S_{cn}$ and $S_{rcbs}$ transforms to predicate "candidate" and therefore to the whole language of BSRT:

| F | $S_{rc}(F)$ | $S_{rn}(F)$ | $S_{cn}(F)$ |
|---|---|---|---|
| candidate $(n_i, r_j, c_k)$ | candidate $(n_i, r_k, c_j)$ | candidate $(n_j, r_i, c_k)$ | candidate $(n_k, r_j, c_i)$ |

| F | $S_{rcbs}(F)$ |
|---|---|
| candidate$(n_i, r_j, c_k)$ | candidate'$[n_i, b_j, s_k]$ |

We now have all the technical tools necessary for stating and proving our three fundamental meta-theorems.

### 4.7.1. Formal statement and proof of meta-theorem 2.1

*Meta-theorem 4.1 (formal version of 2.1): if R is a resolution rule, then $S_{rc}(F)$ is a resolution rule (and it obviously has the same logical complexity as R). We shall express this as: the set of resolution rules is closed under symmetry.*

Proof: If R is a resolution rule, then (by definition) R has a formal proof in ST ∪ VCR. From such a proof of R, a proof of $S_{rc}(R)$ in ST ∪ VCR can be obtained by replacing successively each step in the first proof (axioms included) by its transformation under $S_{rc}$. This is legitimate since:
– the set of axioms in ST ∪ VCR is invariant under $S_{rc}$ symmetry;
– any application of a logical rule can be transposed.

The only technicality is that $S_{rc}$ must be extended to non block-free formulæ. This is easily done by letting unchanged anything that is not of sort Row or Column.

### 4.7.2. Formal statement and proof of meta-theorem 2.2

*Meta-theorem 4.2 (formal version of 2.2): if R is a block-free resolution rule, then $S_{rn}(R)$ and $S_{cn}(R)$ are resolution rules (and they obviously have the same logical complexity as R). We shall express this as: the set of resolution rules is closed under supersymmetry.*

Proof: the proof (for $S_{rn}$) is similar to that of meta-theorem 4.1. By definition, R has a formal proof in ST ∪ VCR. Let T be the block-free theory consisting of the axioms in B(ST ∪ VCR) = B(ST) ∪ B(VCR) = LST ∪ B(VCR). Following the same lines as in the proof of theorem 3.2, there is a (second) proof of R, this time in LST ∪ B(VCR). From such a proof, a proof of $S_{rn}(R)$ in LST ∪ B(VCR) can be obtained by replacing successively each step in the second proof (axioms included) by its transformation under $S_{rn}$. This will also be a proof of $S_{rn}(R)$ in ST ∪ VCR.

### 4.7.3. Formal statement and proof of meta-theorem 2.3

Formally stating and proving meta-theorem 2.3 is done along the same lines as we did for meta-theorems 2.1 and 2.2.

***Meta-theorem 4.3 (formal version of 2.3): if a block-free resolution rule R can be proved without using axiom ST-cn, then $S_{rcbs}(R)$ is a resolution rule (and it obviously has the same logical complexity as R). We shall express this as: the set of resolution rules is closed under analogy.***

Proof: after the proof of theorem 4.2, there is a proof of R in LST ∪ B(VCR). This is not enough for our purpose, but the proof of theorem 4.2 can be transposed to show that there is a proof of R in LST ∪ B(VCR) that does not use axiom ST-cn (the transposition done in the proof of theorem 4.2 does not introduce axiom ST-cn if it was not used in the first proof); it is therefore a proof of R using only the axioms in the set {ST-rc, ST-rn, ST-C} ∪ B(VCR). From this proof of R, a proof of $S_{rcbs}(R)$ using only the axioms in the set {ST-rc, ST-bn, ST-C} ∪ B(VCR) is obtained by replacing each step in the first proof by its transformation under $S_{rcbs}$.

### 4.7.4. Symmetries, analogies and supersymmetries in BSRT

The above theorems are illustrated in Figure 4.2 with the various relationships existing between Singles. Similar figures could be drawn for ECP or CD rules.

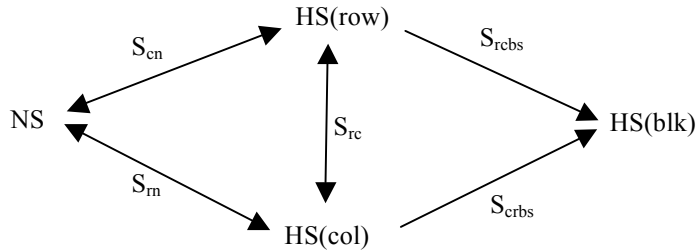

*Figure 4.2. Symmetries, analogies and supersymmetries for Singles*

*4.7.5. Extension of meta-theorem 4.2*

Finally, meta-theorem 4.2 can be modified and extended to a wider class of resolution rules by defining the notion of a block-positive formula. For an easier formulation, let us consider formulæ written without the logical symbol for implication ("⇒"), i.e. written with only the following logical symbols: ∧, ∨, ¬, ∀, ∃. Remember that the condition part of any resolution rule satisfies this restriction.

Definitions: A formula F is *block-positive* if it does not contain the logical symbol for implication ("⇒") and if any of its non block-free primary predicates is in the scope of an even number of negations (i.e. of "¬" symbols). A resolution rule A⇒B is said to be block-positive if B is block-free and A is block-positive.

***Theorem 4.4: if F is a block-positive formula, then the validity of BF(F) entails the validity of F; in particular, if R is a block-positive resolution rule, then BF(R) is a resolution rule.***

The proof of the first part is obvious. Notice that BF(R) is weaker than R, since it has stronger conditions; it might therefore be considered as totally uninteresting. But BF(R) is block-free and it can be submitted to meta-theorem 4.3. This is the way how, when we dealt with chains in *HLS1*, counterparts of all the chain rules in natural rc-space could be defined in rn- and cn-spaces, leading to entirely new types of chains (hidden xy-chains, hidden xyzt-chains, …).

***Meta-theorem 4.5 (formal, extended version of 4.2): if R is a block-positive resolution rule, then $S_{rn} \bullet BF(R)$ and $S_{cn} \bullet BF(R)$ are resolution rules.***

**Part Two**

# GENERAL CHAIN RULES

# 5. Bivalue-chains, whips and braids

Now that our logical framework is completely set, this chapter – the central one of this book as for the types of resolution rules we shall meet – introduces very general types of chain patterns (of increasing complexity) giving rise to resolution rules for any CSP: bivalue chains and whips (together with a few intermediate cases). Braids, a pattern more general than chains, are also defined. We review a few properties of these patterns and of resolution theories based on them. All the examples studied in this book will show that whips are very powerful.

In this chapter, we give only examples related to the subsumption relationships between the whip and braid resolution theories. In the Sudoku case, many specialisations of the patterns introduced here (such as 2D chains and hidden chains) and many more examples can be found in *HLS*. In order not to overload the main text with long resolution paths, these are all grouped in the final section.

Let us now introduce the basic definitions needed for all the rules of this chapter.

Definition: in a resolution state RS, a *chain* is a finite *sequence* of candidates (it is thus linearly ordered) such that any two consecutive candidates in the sequence are *linked* (we call this the "continuity condition" of chains; it implies that consecutive candidates are different).

Remarks:
– non consecutive candidates are not *a priori* forbidden to be identical, so that a chain may contain inner loops; for some specific types of chains, one can discard such loops as being "unproductive", an idea that will be explained in section 5.9;

– in case we need to specify the length of a chain, we shall speak of a chain[3], a chain[4], a chain[5]…, according to half the number of candidates it contains; if the number of candidates is odd, we round to the integer above (these conventions will be justified later);

– sequentiality (or linearity) and continuity are the two characteristic properties of all our types of chains; but chains must satisfy additional conditions in order to be usable for eliminations, such as given by the following definition.

Definition: in a resolution state RS, *a regular sequence of length n associated with a sequence $(V_1, ... V_n)$ of CSP variables* is a sequence of 2n or 2n-1 candidates $(L_1, R_1, L_2, R_2, …. L_n, [R_n])$ such that:

– any two consecutive candidates in the sequence are different;

– $L_n$ is a label for $V_n$: $L_n = <V_n, l_n>$; if $R_n$ is present in the sequence, it is also a label for $V_n$: $R_n = <V_n, r_n>$;

– for any $1 \leq k < n$, both $L_k$ and $R_k$ are labels for $V_k$: $<V_k, l_k>$ and $<V_k, r_k>$; this condition, which we call "the strong $L_k$ to $R_k$ continuity condition", implies the "$L_k$ to $R_k$ continuity condition" of chains, i.e. that, for any $1 \leq k < n$, $L_k$ and $R_k$ are linked. The $L_k$'s are called the *left-linking candidates* and the $R_k$'s the *right-linking candidates*.

Definition: in a resolution state RS, a *regular chain* is a regular sequence that satisfies all the $R_{k-1}$ to $L_k$ continuity conditions of chains (i.e. $L_k$ is linked to $R_{k-1}$ for all k). It is thus a particular kind of chain.

**5.1 Bivalue chains**

Bivalue chains are the most basic chains that can be defined for any CSP.

Definition: a CSP variable V is said to be *bivalue* in a resolution state RS if it has exactly two candidates in RS. This could be formally defined by the auxiliary predicate "bivalue", with signature (CSP-Variable):

bivalue(V) ≡ $\exists_{\neq}(v_1, v_2)$ $\exists_{\neq}(l_1, l_2)$ { label($l_1$, V, $v_1$) ∧ candidate($l_1$)
$\qquad\qquad\qquad\qquad$ ∧ label($l_2$, V, $v_2$) ∧ candidate($l_2$)
$\qquad\qquad\qquad\qquad$ ∧ $\forall v \neq (v_1, v_2)$ $\forall l$ ¬[label(l, V, v) ∧ candidate(l)]}.

Definition: in any CSP and resolution state RS, a *bivalue-chain of length n* ($n \geq 1$) is a *regular chain* ($L_1, R_1, L_2, R_2, \ldots L_n, R_n$) associated with CSP variables ($V_1, \ldots V_n$) such that, for any $1 \leq k \leq n$, $V_k$ is bivalue in RS ($L_k$ and $R_k$ are thus the only two candidates for $V_k$ in RS).

Definition: in a resolution state RS, a target of a bivalue chain is a candidate Z that does not belong to the chain and that is linked to both its endpoints ($L_1$ and $R_n$). Notice that these conditions imply that Z is a label for none of the CSP variables $V_k$.

***Theorem 5.1 (bivalue-chain rule for a general CSP): in any resolution state of any CSP, if Z is a target of a bivalue-chain, then it can be eliminated (formally, this rule concludes ¬candidate(Z)).***

Proof: the proof is short and obvious but it will be the basis for the proofs of all our forthcoming chain, whip and braid rules.

If Z was True, then $L_1$ would be eliminated by ECP; therefore $R_1$ would be asserted by S; but then $L_2$ would be eliminated by ECP and $R_2$ would be asserted by S… ; finally $R_n$ would be asserted by S; which would contradict the hypothesis that Z was True. Therefore Z can only be False. qed.

Notation: a bivalue-chain of length n, together with a potential target elimination, is written symbolically as:
**biv-chain[n]: {L$_1$ R$_1$} – {L$_2$ R$_2$} – …… – {L$_n$ R$_n$} ⇒ ¬ candidate(Z)**,
where the curly brackets recall that the two candidates inside have representatives with the same CSP variable.

Re-writing the candidates as <variable, value> pairs and "factoring" the CSP variables out of the pairs, a bivalue chain will also be written symbolically in either of the more explicit forms:

**biv-chain[n]: V$_1$**{**l$_1$ r$_1$**} **– V$_2$**{**l$_2$ r$_2$**} **– …… – V$_n$**{**l$_n$ r$_n$**} **⇒ ¬ candidate(Z)**, or:
**biv-chain[n]: V$_1$**{**l$_1$ r$_1$**} **– V$_2$**{**l$_2$ r$_2$**} **– …… – V$_n$**{**l$_n$ r$_n$**} **⇒ V$_Z$ ≠ v$_Z$**.

### 5.2 z-chains, t-whips and zt-whips (or whips)

*5.2.1 Definitions*

The definition of a bivalue-chain can be extended in different ways (z-extension, t-extension and zt-extension), as follows. We first introduced the following generalisations of bivalue-chains in *HLS*, in the Sudoku context. But everything works similarly for any CSP (see [Berthier 2008b]). It is convenient to start with the definitions: a label C is *compatible* with a set S of labels if it does not belong to S and it is not linked to any element of S; notice that this is a structural property, independent of any resolution state. In a resolution state RS, a candidate is compatible with a set S of candidates if its underlying label is compatible with all the underlying labels of the elements of S.

Definition: in a resolution state RS, given a candidate Z (which will be the target), a *z-chain* of length n (n ≥ 1) built on Z is a *regular chain* (L$_1$, R$_1$, L$_2$, R$_2$, …. L$_n$, R$_n$) associated with a sequence (V$_1$, … V$_n$) of CSP variables, such that:
– Z does not belong to {L$_1$, R$_1$, L$_2$, R$_2$, …. L$_n$, R$_n$};
– L$_1$ is linked to Z;
– for any 1 ≤ k < n, R$_k$ is the only candidate for V$_k$ compatible with Z, apart possibly for L$_k$;
– Z is not a label for V$_n$;
– L$_n$ is the only candidate for V$_n$ possibly compatible with Z (but V$_n$ has more than one candidate – this is a non-degeneracy condition); in particular R$_n$ is linked to Z.

***Theorem 5.2 (z-chain rule for a general CSP [Berthier 2008b]): in any resolution state of any CSP, any target of a z-chain can be eliminated (formally, this rule concludes ¬ candidate(Z)).***

For the following "t-extension" of bivalue chains, it is natural to introduce *whips* instead of chains. Whips are also more general, because they are able to catch more contradictions than chains. A *target of a whip* is required to be linked to its first candidate, not necessarily to its last; the condition on the last variable is changed so that the final contradiction can occur with previous right-linking candidates.

Definition: in a resolution state RS, given a candidate Z (which will be the target), a *t-whip* of length n (n ≥ 1) built on Z is a *regular chain* ($L_1$, $R_1$, $L_2$, $R_2$, …. $L_n$) [notice there is no $R_n$] associated with a sequence ($V_1$, … $V_n$) of CSP variables, such that:

– Z does not belong to {$L_1$, $R_1$, $L_2$, $R_2$, …. $L_n$};

– $L_1$ is linked to Z;

– for any 1 ≤ k < n, $R_k$ is the only candidate for $V_k$ compatible with all the previous right-linking candidates (i.e. with all the $R_i$ for i ≤ k), [in t-whips, compatibility with Z does not have to be checked for intermediate candidates; it allows to build t-whips before the target is fixed; this is a computational advantage over the zt-whips defined below, but they have a weaker resolution power];

– Z is not a label for $V_n$;

– $V_n$ has no candidate compatible with Z and with all the previous right-linking candidates (but $V_n$ has more than one candidate – this is a non-degeneracy condition).

Definition: in a resolution state RS, given a candidate Z (which will be the target), a *zt-whip* (in short a *whip*) of length n (n ≥ 1) built on Z is a *regular chain* ($L_1$, $R_1$, $L_2$, $R_2$, …. $L_n$) [notice there is no $R_n$] associated with a sequence ($V_1$, … $V_n$) of CSP variables, such that:

– Z does not belong to {$L_1$, $R_1$, $L_2$, $R_2$, …. $L_n$};

– $L_1$ is linked to Z;

– for any 1 ≤ k < n, $R_k$ is the only candidate for $V_k$ compatible with Z and with all the previous right-linking candidates (i.e. with Z and with all the $R_i$, 1 ≤ i < k);

– Z is not a label for $V_n$;

– $V_n$ has no candidate compatible with Z and with all the previous right-linking candidates (but $V_n$ has more than one candidate – this is a non-degeneracy condition).

Definition: in any of the above defined chains and whips (and in the forthcoming braids), a candidate other than $L_k$ or $R_k$ for any of the CSP variables $V_k$ is called a *t-candidate* [respectively a *z-candidate*] if it is incompatible with a previous right-linking candidate [resp. with the target]. Notice that a candidate can be z- and t- at the same time and that the t- and z- candidates are not considered as being part of the pattern.

***Theorem 5.3 (t- and zt-whip rules for a general CSP [Berthier 2008b]): in any resolution state of any CSP, if Z is a target of a t- or a zt- whip, then it can be eliminated (formally, this rule concludes ¬ candidate(Z)).***

Proof: the proof is a simple adaptation of that for bivalue-chains. If Z was True, then all the z- candidates would be eliminated by ECP and, progressively: all the left-linking candidates and all the t- candidates would be eliminated by ECP and all the right-linking ones would be asserted by S. The end is slightly different: the last condition on the whip entails that there would be no possible value for the last variable $V_n$ (because it is not a CSP-Variable for Z), a contradiction.

Although these new chains or whips seem to be straightforward generalisations of bivalue-chains, their solving potential is much higher. In chapter 6, we shall give detailed statistics illustrating this in the Sudoku case.

Notation:
– a z-chain is written symbolically in the same two ways as a bivalue chain, but with prefix "z-chain" instead of "biv-chain";
– (a t-whip or) a whip of length n, together with a potential target elimination, is written symbolically as:
(t-)whip[n]: $\{L_1\ R_1\} - \{L_2\ R_2\} - \ldots - \{L_n\ .\} \Rightarrow \neg$ **candidate(Z)**,
where the curly brackets recall that the two candidates inside them are relative to the same CSP variable; the dot inside the last curly brackets means the absence of a compatible candidate; as in the bivalue chains case, and for the same reasons, we shall also write whips in the form:

whip[n]: $V_1\{l_1\ r_1\} - V_2\{l_2\ r_2\} - \ldots - V_n\{l_n\ .\} \Rightarrow \neg$ **candidate(Z)**, or:
whip[n]: $V_1\{l_1\ r_1\} - V_2\{l_2\ r_2\} - \ldots - V_n\{l_n\ .\} \Rightarrow V_Z \neq v_Z$.

Remarks:
– an alternative equivalent definition of a whip is available in section 11.1;
– as a consequence of the definition, Z is a label for none of the CSP variables in the whip;
– another consequence is that all the CSP variables of the whip are different;
– particular attention should be given to the whip[1] case; a given CSP may have whips of length 1 or not (without prejudice for longer ones – see section 5.11.7): Sudoku, N-Queens, N-SudoQueens, Futoshiki and Kakuro have, LatinSquare does not; having whips of length one has many consequences for the resolution theories of the CSP; chapter 7 will be entirely dedicated to CSPs having such whips;
– very instructive whip[2] examples can be found in sections 8.7.1.1 and 8.8.1, where it is shown that they cannot be considered as $S_2$-subsets.

*5.2.2. Formal definitions*

As a mere exercise in logic, let us write the formal definitions of whips. We leave it as an exercise for the reader to write similar formulæ for all the other kinds of chains introduced above (and for the forthcoming braids).

Let us first introduce two auxiliary predicates "Linked-by" and "Linked" (with capital "L"), with respective signatures (Label, Label, Constraint) and (Label, Label):
Linked-by($l_1$, $l_2$, c) ≡ linked-by($l_1$, $l_2$, c) ∧ candidate($l_1$) ∧ candidate($l_2$)
Linked($l_1$, $l_2$) ≡ linked($l_1$, $l_2$) ∧ candidate($l_1$) ∧ candidate($l_2$)
         ≡ ∃c Linked-by($l_1$, $l_2$, c)

Recalling that CSP-Variable is a sub-sort of Constraint, we can now define a whip of length 1 based on target z by predicate whip[1] with signature (Label, Label, CSP-Variable):
whip[1](z, $l_1$, $V_1$) ≡ Linked(z, $l_1$) ∧ ∀l ¬[Linked-by($l_1$, l, $V_1$) ∧ ¬linked(z, l)].

Notice that the third argument of predicate whip[1] is restricted to be of sort CSP-Variable in order to conform to our definition of whips.

The second condition above says that any candidate l linked to $l_1$ by CSP variable $V_1$ (i.e. any candidate value l≠$l_1$ for CSP variable $V_1$) must be linked to z (by some constraint); notice that this condition could not be written as
∀l [¬Linked-by($l_1$, l, $V_1$) ∨ linked(z, l)], because negating Linked-by($l_1$, l, $V_1$) may negate a condition which is not the one we want, e.g. that l is a candidate.

For longer whips, we need auxiliary predicates whip[n] and partial-whip[n] with signatures (Label, [Label, Label, CSP-Variable]$^{n-1 \text{ times}}$, Label, CSP-Variable) and (Label, [Label, Label, CSP-Variable]$^{n \text{ times}}$) respectively; we define partial-whips and whips by simultaneous induction on n:
partial-whip[1](z, $l_1$, $r_1$, $V_1$) ≡
  Linked(z, $l_1$) ∧ Linked-by($l_1$, $r_1$, $V_1$) ∧ $r_1$≠z
  ∧ ¬whip[1](z, $l_1$, $V_1$)
  ∧ ∀l≠$r_1$ ¬[Linked-by($l_1$, l, $V_1$) ∧ ¬linked(z, l)];

whip[n+1](z, $l_1$, $r_1$, $V_1$, $l_2$, $r_2$, $V_2$, …, $l_n$, $r_n$, $V_n$, $l_{n+1}$, $V_{n+1}$) ≡
  partial-whip[n](z, $l_1$, $r_1$, $V_1$, $l_2$, $r_2$, $V_2$, …, $l_n$, $r_n$, $V_n$) ∧
  ∧ Linked($r_n$, $l_{n+1}$) ∧ $l_{n+1}$≠z
  ∧ ∀l ¬[Linked-by($l_{n+1}$, l, $V_{n+1}$)
          ∧ ¬linked(z, l) ∧ ¬ linked(z, $r_1$) ∧ ¬ … ∧ ¬ linked(z, $r_n$)];

partial-whip[n+1](z, $l_1$, $r_1$, $V_1$, $l_2$, $r_2$, $V_2$, …, $l_n$, $r_n$, $V_n$, $l_{n+1}$, $r_{n+1}$, $V_{n+1}$) ≡
  partial-whip[n](z, $l_1$, $r_1$, $V_1$, $l_2$, $r_2$, $V_2$, …, $l_n$, $r_n$, $V_n$)
  ∧ Linked($r_n$, $l_{n+1}$) ∧ $l_{n+1}$≠z ∧ Linked-by($l_{n+1}$, $r_{n+1}$, $V_{n+1}$) ∧ $r_{n+1}$≠z

∧ ¬whip[n+1](z, $l_1$, $r_1$, $V_1$, $l_2$, $r_2$, $V_2$, …, $l_n$, $r_n$, $V_n$, $l_{n+1}$, $V_{n+1}$)
∧ ∀l≠ $r_{n+1}$ ¬[Linked-by($l_{n+1}$, l, $V_{n+1}$)
∧¬linked(z, l) ∧¬ linked(z, $r_1$) ∧¬ … ∧¬ linked(z, $r_n$)].

Notice that, in these definitions, a whip is minimal, i.e. no initial segment is a shorter whip, due to the condition "¬whip[n+1](z, …" in partial-whip[n+1].

### 5.2.3. A typical moderately hard example with bivalue-chains and whips

The resolution path of the Sudoku puzzle in Figure 5.1 is typical of how the above defined resolution rules allow to solve a moderately difficult instance.

|   | 3 | 4 | 5 |   |   | 8 |   |   |
|---|---|---|---|---|---|---|---|---|
| 4 | 6 |   |   |   |   |   |   |   |
|   |   | 1 |   |   |   |   | 6 | 5 |
|   | 1 |   | 5 |   | 8 |   | 7 |   |
|   |   |   | 3 |   |   |   | 9 | 1 |
| 9 |   |   |   |   |   |   |   |   |
|   |   | 8 |   |   |   |   |   |   |
| 6 |   |   |   |   | 7 |   | 3 |   |
|   | 1 |   | 3 |   |   | 5 | 7 |   |

| 1 | 2 | 3 | 4 | 5 | 6 | 7 | 8 | 9 |
|---|---|---|---|---|---|---|---|---|
| 4 | 5 | 6 | 7 | 8 | 9 | 1 | 2 | 3 |
| 7 | 8 | 9 | 1 | 2 | 3 | 4 | 6 | 5 |
| 2 | 1 | 4 | 5 | 9 | 8 | 3 | 7 | 6 |
| 5 | 6 | 7 | 3 | 4 | 2 | 8 | 9 | 1 |
| 9 | 3 | 8 | 6 | 7 | 1 | 5 | 4 | 2 |
| 3 | 7 | 2 | 8 | 6 | 5 | 9 | 1 | 4 |
| 6 | 4 | 5 | 9 | 1 | 7 | 2 | 3 | 8 |
| 8 | 9 | 1 | 2 | 3 | 4 | 6 | 5 | 7 |

***Figure 5.1.** A moderately difficult Sudoku puzzle (cb#7) and its solution*

\*\*\*\*\*  SudoRules 16.2 based on CSP-Rules 1.2, config: W  \*\*\*\*\*
25 givens, 224 candidates, 1648 csp-links and 1648 links. Initial density = 1.65
singles: r7c6 = 5, r5c1 = 5, r6c7 = 5, r8c3 = 5, r2c2 = 5, r2c5 = 8, r6c6 = 1, r4c5 = 9, r3c7 = 4, r3c6 = 3, r1c6 = 6, r1c1 = 1
whip[1]: r4n6{c7 .} ==> r6c9 ≠ 6, r5c7 ≠ 6
whip[1]: r3n9{c3 .} ==> r1c2 ≠ 9
whip[1]: r1n9{c9 .} ==> r2c7 ≠ 9, r2c9 ≠ 9
biv-chain[2]: r4n4{c3 c9} - c8n4{r6 r7} ==> r7c3 ≠ 4
whip[1]: b7n4{r9c2 .} ==> r6c2 ≠ 4, r5c2 ≠ 4
biv-chain[3]: b6n8{r5c7 r6c9} – r6n3{c9 c2} – c2n6{r6 r5} ==> r5c2 ≠ 8
biv-chain[3]: r4c1{n2 n3} – r6n3{c2 c9} – r2c9{n3 n2} ==> r4c9 ≠ 2
biv-chain[3]: r1c2{n2 n7} – c1n7{r3 r7} – r7n3{c1 c2} ==> r7c2 ≠ 2
whip[3]: r6n3{c2 c9} – r2c9{n3 n2} – r1n2{c7 .} ==> r6c2 ≠ 2
**whip[4]: b3n7{r1c7 r2c7} – c7n3{r2 r4} – c1n3{r4 r7} – c1n7{r7 .} ==> r1c2 ≠ 7**
singles ==> r1c2 = 2, r1c9 = 9, r1c7 = 7, r2c4 = 7, r3c5 = 2, r2c6 = 9
whip[3]: b8n2{r9c6 r8c4} – b8n9{r8c4 r9c4} – r9n6{c4 .} ==> r9c7 ≠ 2
whip[3]: r9c1{n8 n2} – c6n2{r9 r5} – r5c7{n2 .} ==> r9c7 ≠ 8
whip[1]: b9n8{r8c7 .} ==> r8c2 ≠ 8
biv-chain[3]: r8c2{n4 n9} – r8c4{n9 n2} – r9c6{n2 n4} ==> r9c2 ≠ 4
singles: r9c6 = 4, r8c5 = 1, r7c5 = 6, r5c6 = 2, r6c4 = 6, r5c7 = 8, r8c9 = 8, r8c2 = 4, r5c2 = 6, r9c7 = 6, r4c9 = 6, r4c3 = 4, r5c3 = 7, r5c5 = 4, r6c5 = 7

biv-chain[2]: r7n3{c1 c2} – r7n7{c2 c1} ==> r7c1 ≠ 2
biv-chain[2]: b7n2{r7c3 r9c1} – r4n2{c1 c7} ==> r7c7 ≠ 2
whip[2]: b7n3{r7c2 r7c1} – b7n7{r7c1 .} ==> r7c2 ≠ 9
biv-chain[3]: r3c3{n8 n9} – b7n9{r7c3 r9c2} – r9n8{c2 c1} ==> r3c1 ≠ 8
singles to the end
GRID SOLVED. rating-type = W, MOST COMPLEX RULE = Whip[4]

### 5.3 Braids

We now introduce braids, a further generalisation of whips. Whereas whips have a sequential *and* continuous structure (a chain structure), braids still have a sequential structure but it is discontinuous (in restricted ways). In any CSP, braids are interesting for three reasons:

– they have an *a priori* greater solving potential than whips (at the cost of a more complex logical structure and *a priori* higher computational complexity);

– resolution theories based on them can be proven to have the very important confluence property, allowing to superimpose on them various resolution strategies (see section 5.5);

– their scope can be defined very precisely by a simple procedure: they can eliminate any candidate that can be eliminated by pure Trial-and-Error (T&E); they can therefore solve any instance that can be solved by T&E (and conversely – see section 5.6).

Definition: in a resolution state RS, given a candidate Z (which will be the target), a *zt-braid* (in short a *braid*) of length n (n ≥ 1) built on Z is a *regular sequence* ($L_1$, $R_1$, $L_2$, $R_2$, …. $L_n$) [notice that there is no $R_n$] associated with a sequence ($V_1$, … $V_n$) of CSP variables, such that:

– Z does not belong to {$L_1$, $R_1$, $L_2$, $R_2$, …. $L_n$};

– $L_1$ is linked to Z;

– for any 1 < k ≤ n, $L_k$ is linked either to a previous right-linking candidate (some $R_i$, i < k) or to the target; this is the only (but major) structural difference with whips (for which the only linking possibility is $R_{k-1}$); the $R_{k-1}$ to $L_k$ continuity condition of chains is not satisfied by braids (a braid is defined as a regular *sequence*, a whip as a regular *chain*);

– for any 1 ≤ k < n, $R_k$ is the only candidate for $V_k$ compatible with Z and with all the previous right-linking candidates (i.e. with Z and with all the $R_i$, 1 ≤ i < k);

– Z is not a label for $V_n$;

– $V_n$ has no candidate compatible with the target and with all the previous right-linking candidates (but $V_n$ has more than one candidate – this is a non-degeneracy condition).

Remarks:
– an alternative equivalent definition is available in section 11.1;
– as in the case of whips, the t- and z- candidates are not considered as being part of the braid;
– in order to show the kind of restriction this definition implies on the nettish structure of a braid, the first of the following two structures can be part of a braid starting with $\{L_1\ R_1\} - \{L_2\ R_2\} - ...$ , whereas the second cannot:
$\{L_1\ R_1\} - \{L_2\ R_2\ A_2\} - ...$ where $A_2$ is linked to $R_1$ (or to Z);
$\{L_1\ R_1\ A_1\} - \{L_2\ R_2\ A_2\} - ...$ where $A_1$ is linked to $R_2$ and $A_2$ is linked to $R_1$ but none of them is linked to Z. The only thing that could be concluded from this pattern if Z was True is $(R_1 \wedge R_2) \vee (A_1 \wedge A_2)$, whereas a braid should allow to conclude $R_1 \wedge R_2$.

The proof of the following theorem is almost the same as for whips, because the condition replacing $R_{k-1}$ to $L_k$ continuity still allows the elimination of $L_k$ by ECP.

***Theorem 5.4 (braid rule for a general CSP [Berthier 2008b]): in any resolution state of any CSP, if Z is a target of a braid, then it can be eliminated (formally, this rule concludes ¬candidate(Z)).***

Notation: a braid is written symbolically in exactly the same ways as a whip, with prefix "braid" instead of "whip", but the "–" symbol must be interpreted differently:

**braid[n]: $\{L_1\ R_1\} - \{L_2\ R_2\} - ...... - \{L_n\ .\} \Rightarrow $ ¬ candidate(Z)**, or
**braid[n]: $V_1\{l_1\ r_1\} - V_2\{l_2\ r_2\} - ...... - V_n\{l_n\ .\} \Rightarrow $ ¬ candidate(Z)**, or:
**braid[n]: $V_1\{l_1\ r_1\} - V_2\{l_2\ r_2\} - ...... - V_n\{l_n\ .\} \Rightarrow V_Z \neq v_Z$**.

Notice the double role played by the prefix in all of the above-defined notations:
– it indicates how the curly brackets must be understood (pure bivalue or bivalue "modulo" the previous right-linking candidates and/or the target);
– it also indicates how the link symbol "–" must be understood.

The prefix of each resolution rule applied to solve any instance of the CSP should therefore always appear explicitly in any resolution path.

### 5.4. Whip and braid resolution theories; the W and B ratings

*5.4.1. Whip resolution theories in a general CSP; the W rating*

We are now in a position to define an increasing sequence of resolution theories based on whips. As there can be no confusion, we shall always use the same name for a resolution theory and for the set of instances it can solve. Recall that BRT(CSP) is the Basic Resolution Theory of the CSP, as defined in section 4.3.

Definition: for any n ≥ 0, let $W_n$ be the following resolution theory:
- $W_0$ = BRT(CSP),
- $W_1 = W_0 \cup$ {rules for whips of length 1},
- ....
- $W_n = W_{n-1} \cup$ {rules for whips of length n},
- $W_\infty = \cup_{n \geq 0} W_n$.

Definition : the **W-rating** of an instance P of the CSP, noted W(P), is the smallest n ≤ ∞ such that P can be solved within $W_n$. An instance P has W rating n [i.e. W(P) = n] if it can be solved using only whips of length no more than n but it cannot be solved using only whips of length strictly smaller than n. By convention, W(P) = ∞ means that P cannot be solved by whips.

The W rating has some good properties one can expect of a rating:

– it is defined in a purely logical way, independent of any implementation; the W rating of an instance P is an intrinsic property of P;

– in the Sudoku case, it is invariant under symmetry and supersymmetry ; similar symmetry properties will be true for any CSP, if it has symmetries of any kind and they are properly formalised in the definition of its CSP variables;

– in the Sudoku case, it is well correlated with familiar (though informal) measures of complexity.

### 5.4.2. Braid resolution theories in a general CSP; the B rating

One can define a similar increasing sequence of resolution theories, now based on braids.

Definition: for any n ≥ 0, let $B_n$ be the following resolution theory:
- $B_0$ = BRT(CSP) = $W_0$,
- $B_1 = B_0 \cup$ {rules for braids of length 1} = $W_1$ (obviously),
- $B_2 = B_1 \cup$ {rules for braids of length 2},
- ....
- $B_n = B_{n-1} \cup$ {rules for braids of length n},
- $B_\infty = \cup_{n \geq 0} B_n$.

Definition : the **B-rating** of an instance P of the CSP, noted B(P), is the smallest n ≤ ∞ such that P can be solved within $B_n$. By convention, B(P) = ∞ means that P cannot be solved by braids.

The B rating has all the good properties one can expect of a rating:

– it is defined in a purely logical way, independent of any implementation; the B rating of an instance P is an intrinsic property of P;

– as will be shown in the next section, it is based on an increasing sequence ($B_n$, n≥0) of resolution theories with the confluence property; this ensures *a priori* better computational properties; in particular, one can define a "simplest first" resolution strategy able to find the B rating after following a single resolution path;

– in the Sudoku case, it is invariant under symmetry and supersymmetry ; similar symmetry properties will be true for any CSP, if it has symmetries of any kind and they are properly formalised in the definition of its CSP variables;

– in the Sudoku case, it is well correlated with familiar (though informal) measures of complexity.

### 5.4.3. Comparison of whip and braid resolution theories (and ratings)

Notice first that both the W and B ratings are measures of the hardest step in the simplest resolution paths, they do not take into account any combination of steps in the whole path. An instance P with W(P) = 12 having a single step with such a long whip may be simpler (in some different, intuitive sense) than an instance Q with W(Q) = 11 but that has many steps with whips of length 11.

As a whip is a particular case of a braid, one has ***$W_n \subseteq B_n$ and B(P) ≤ W(P) for any CSP, any instance P and any n ≥ 1***. Moreover, as braids have a much more complex structure than whips, one may expect that the two ratings are very different in general. However, in the Sudoku case, it will be shown in chapter 6 that (although whip theories do not have the confluence property, they are not far from having it and) the W rating, when it is finite, is an excellent approximation of the B rating (fairly good approximations of W are easier to compute than the real value of B).

One has $W_n \subseteq B_n$ for any n and any CSP, but the converse is not true in general, except for ***$B_1 = W_1$*** (obviously) and ***$B_2 = W_2$*** (proof below): braids are a true generalisation of whips. Firstly, there are Sudoku puzzles (e.g. the example in section 5.10.1) with W(P) = 5 and B(P) = 4. Secondly, even in the Sudoku case (for which whips solve almost any puzzle), examples can be given (see one in section 5.10.2) of puzzles that can be solved with braids but not with whips, i.e. $W_\infty$ is strictly included in $B_\infty$.

The case n = 3 remains open for the general CSP. We have no example in Sudoku with B(P) = 3 and W(P) > 3, although there exist braids[3] that are not whips[3] (see an example in section 5.10.5). In section 7.4.2, we shall show that, for any CSP, one has ***$gW_3 = gB_3$*** and therefore ***$W_3 \subseteq B_3 \subseteq gW_3$***, where $gW_n$ (respectively $gB_n$) is the resolution theory for g-whips (resp. g-braids) of length ≤ n.

***Theorem 5.5: In any CSP, any elimination done by a braid of length 2 can be done by a whip of same or shorter length; as a result, $B_2 = W_2$.***

Proof: Let $B = V_1\{l_1\ r_1\} - V_2\{l_2\ .\} \Rightarrow V_z \neq v_z$ be a braid[2] with target $Z = <V_Z, r_Z>$ in some resolution state RS.

If variable $V_2$ has a candidate $<V_2, v'>$ (it may be $<V_2, l_2>$) such that $<V_2, v'>$ is linked to $<V_1, r_1>$, then $V_1\{l_1\ r_1\} - V_2\{v'\ .\} \Rightarrow V_z \neq v_z$ is a whip[2] with target Z. Otherwise, $<V_2, l_2>$ can only be linked to $<V_z, v_z>$ and $V_2\{l_2\ .\} \Rightarrow V_z \neq v_z$ is a shorter whip[1] with target Z.

## 5.5. Confluence of the $B_n$ resolution theories; resolution strategies

We now consider the braid resolution theories $B_n$ defined in section 5.4.2 and we prove that they have the confluence property. As a result, we can define a "simplest first strategy" allowing more efficient ways of computing the B rating of instances.

### 5.5.1. The confluence property of braid resolution theories

***Theorem 5.6 [Berthier 2008b]: each of the $B_n$ resolution theories, $0 \leq n \leq \infty$, is stable for confluence; therefore it has the confluence property***.

Before proving this theorem, we must recall a convention about candidates. When one is asserted, its status changes: it becomes a value and it is "eliminated" (i.e. negated) as a candidate (axiom OOS). (This convention is very important for minimising the number of useless patterns, but the theorem does not really depend on it; the proof would only have to be slightly modified with other conventions.)

Let $n<\infty$ be fixed (the case $n=\infty$ is a corollary to all the cases $n<\infty$). We must show that, if an elimination of a candidate Z could have been done in a resolution state $RS_1$ by a braid B of length $m \leq n$ and with target Z, it will always still be possible, starting from any further state $RS_2$ obtained from $RS_1$ by consistency preserving assertions and eliminations, if we use a sequence of rules from $B_n$. Let B be: $\{L_1\ R_1\} - \{L_2\ R_2\} - …. - \{L_p\ R_p\} - \{L_{p+1}\ R_{p+1}\} - … - \{L_m\ .\}$, with target Z.

Consider first the state $RS_3$ obtained from $RS_2$ by applying repeatedly the rules in BRT until quiescence. As BRT has the confluence property (theorem 4.1), this state is uniquely defined, independently of the way we apply the BRT rules.

If target Z has been eliminated in $RS_3$, there remains nothing to prove. If target Z has been asserted, then the instance of the CSP is contradictory; if not yet detected in $RS_3$, this contradiction can be detected by CD in a state posterior to $RS_3$, reached by a series of applications of rules from BRT, following the braid structure of B.

Otherwise, we must consider all the elementary events related to B that can have happened between $RS_1$ and $RS_3$ (all the possibilities are marked by a letter for reference in further proofs). For this, we start from B' = what remains of B in $RS_3$. At this point, B' may not be a braid in $RS_3$. We repeat the following procedure, for p = 1 to p = m, producing in the end a new (possibly shorter) braid B' in $RS_3$ with target Z. All the references below are to the current B'.

a) If, in $RS_3$, the left-linking or any t- or z- candidate of CSP variable $V_p$ has been asserted, then Z and/or the previous $R_k$('s) to which $L_p$ is linked must have been eliminated by ECP in the passage from $RS_2$ to $RS_3$ (if it was not yet eliminated in $RS_2$); if Z is among these eliminations, there remains nothing to prove; otherwise, the procedure has already been successfully terminated by case f of the first such k.

b) If, in $RS_3$, left-linking candidate $L_p$ has been eliminated (but not asserted) (it can therefore no longer be used as a left-linking candidate in a braid) and if CSP variable $V_p$ still has a z- or a t- candidate $C_p$, then replace $L_p$ by $C_p$; now, up to $C_p$, B' is a partial braid in $RS_3$ with target Z. Notice that, even if $L_p$ was linked to $R_{p-1}$ (as it would if B was a whip), this may not be the case for $C_p$; therefore trying to prove a similar theorem for whips would fail here (see section 5.10.3 for an example of non-confluence of the $W_n$ theories). [As it missed this point, the proof given for zt-chains in *HLS1* was not correct.]

c) If, in $RS_3$, any t- or z- candidate of $V_p$ has been eliminated (but not asserted), this has not changed the basic structure of B (at stage p). Continue with the same B'.

d) If, in $RS_3$, right-linking candidate $R_p$ has been asserted (p can therefore not be the last index of B'), it can no longer be used as an element of a braid, because it is no longer a candidate. Notice that all the left-linking and t- candidates for CSP variables of B after p that were incompatible in B with $R_p$, i.e. linked to it, if still present in $RS_2$, must have been eliminated by ECP somewhere between $RS_2$ and $RS_3$. But, considering the braid structure of B upwards from p, more eliminations and assertions must have been done by rules from BRT between $RS_2$ and $RS_3$.

Let q be the smallest number strictly greater than p such that, in $RS_3$, CSP variable $V_q$ still has a (left-linking, t- or z-) candidate $C_q$ that is not linked to any of the $R_i$ for $p \leq i < q$ (by definition of a braid, $C_q$ is therefore linked to Z or to some $R_i$ with i < p). Between $RS_2$ and $RS_3$, the following rules from BRT must have been applied for each of the CSP variables $V_u$ of B with index u increasing from p+1 to q-1 included: eliminate its left-linking candidate ($L_u$) by ECP, assert its right-linking candidate ($R_u$) by S, eliminate by ECP all the left-linking and t-candidates for CSP variables after u that were incompatible in B with the newly asserted candidate ($R_u$).

In $RS_3$, excise from B' the part related to CSP variables p to q-1 (included) and (if $L_q$ has been eliminated in the passage from $RS_1$ to $RS_3$) replace $L_q$ by $C_q$; for each integer s ≥ p, decrease by q-p the index of CSP variable $V_s$ and of its candidates in B'; in $RS_3$, B' is now, up to p (the ex q), a partial braid in $B_n$ with target Z.

e) If, in $RS_3$, left-linking candidate $L_p$ has been eliminated (but not asserted) and if CSP variable $V_p$ has no t- or z- candidate in $RS_3$ (complementary to case b), then $V_p$ has only one possible value in $RS_3$, namely $R_p$; $R_p$ must therefore have been asserted by S somewhere between $RS_1$ and $RS_3$; this case has therefore been dealt with by case d (because the assertion of $R_p$ also entails the elimination of $L_p$).

f) If, in $RS_3$, right-linking candidate $R_p$ of B has been eliminated (but not asserted), in which case p cannot be the last index of B', then replace B' by its initial part: $\{L_1\ R_1\} – \{L_2\ R_2\} – …. – \{L_p\ .\}$. At this stage, B' is in $RS_3$ a shorter braid with target Z. Return B' and stop.

Notice that this proof works only because the notion of being linked does not depend on the resolution state.

Notice also that what we have proven is indeed the following: given $RS_1$, B and $RS_2$ as above, if $RS_3$ is the resolution state obtained from $RS_2$ by the repeated application of rules from BRT until quiescence, then:

– either a contradiction has been detected by CD somewhere between $RS_2$ and $RS_3$ (and, due to consistency preservation between $RS_1$ and $RS_2$, it can only be because a contradiction inherent in the givens of P has been made manifest by CD);

– or Z has been eliminated by ECP somewhere between $RS_2$ and $RS_3$;

– or Z can be eliminated in $RS_3$ by a braid B' possibly shorter than B, with target Z, with CSP variables a sub-sequence W' of those of B, with right-linking candidates those of B belonging to the sub-sequence W', with left-linking candidates those of B belonging to the sub-sequence W', each of them possibly replaced by a t-candidate of B for the same CSP variable.

### 5.5.2. Braid resolution strategies consistent with the B rating

As explained in section 4.5.3, we can take advantage of the confluence property of braid resolution theories to define a "simplest first" strategy that will always find the simplest solution, in terms of the maximum length of the braids it will use. As a result, it will also compute the B rating of an instance after following a single resolution path. The following precedence order satisfies this requirement:
ECP > S > biv-chain[1] > z-chain[1] > t-whip[1] > whip[1] > braid[1] > … >
biv-chain[k] > z-chain[k] > t-whip[k] > whip[k] > braid[k] >
biv-chain[k+1] > z-chain[k+1] > t-whip[k+1] > whip[k+1] > braid[k+1] > …

Notice that bivalue-chains, z-chains, t-whips and whips being special cases of braids of same length, their explicit presence in the set of rules does not change the final result. We put them here because when we look at a resolution path, it may be nicer to see simple patterns appear instead of more complex ones (braids). Also, it shows (in the Sudoku case) that braids that are not whips appear only rarely.

The above ordering defines a "simplest first" resolution strategy. It does not completely define a deterministic procedure: it does not set any precedence between different chains of same type and length. This could be done by using an ordering of the candidates instantiating them, based e.g. on their lexicographic order. But one can also decide that, for all practical purposes, which of these equally prioritised rule instantiations should be "fired" first will be chosen randomly (as in CSP-Rules).

### 5.6. The "T&E vs braids" theorem

For braids, the following "T&E vs braids" theorem is second in importance only to the confluence property. As it is easy to program very fast implementations of the T&E procedure, it allows to check quickly if a given instance P will be solvable by braids. This may be very useful: in case the answer is negative, we may not want to waste computation time on P. In case it is positive, it does not produce an explicit resolution path with braids and, even if we build one from the trace of this procedure, it will not be one with the shortest braids and it will not provide the B rating; but the computations with braids will then be guaranteed to give a solution.

#### 5.6.1. Definition of the Trial-and-Error procedure T&E(T, P)

The following definition of the Trial-and-Error (T&E) procedure is intimately related to the informal idea that the solution should be obtained with "no guessing". Indeed, in our view, it is the only proper formalisation of the vague "no guessing" requirement. In standard search algorithms (depth-first, beadth-first, …), if a path in the search graph leads to a solution, this result is accepted. In T&E, this would be considered as arbitrary, i.e. as "guessing"; it must be shown that there can be no other solution (see section 5.6.3 for more detailed comments).

Definition: given a resolution theory T with the confluence property, a resolution state RS and a candidate Z in RS, *T&E(T, Z, RS) or Trial-and-Error based on T for Z in RS*, is the following procedure (notice: a procedure, not a resolution rule):
- make a copy RS' of RS; in RS', delete Z as a candidate and assert it as a value;
- in RS', apply repeatedly all the rules in T until quiescence;
- if RS' has become a contradictory state, then delete Z from RS (*sic*: RS, not RS'); else do nothing (in particular if a solution is obtained in RS', merely forget it);
- return the (possibly) modified RS state.

Notice that this definition is meaningful only if T has the confluence property: otherwise, the result of "applying repeatedly in RS' all the rules in T until quiescence" may not be uniquely defined.

Definition: given a resolution theory T with the confluence property and a resolution state RS, we define the *T&E(T, RS)* procedure as follows:
a) in RS, apply the rules in T until quiescence; if the resulting RS is a solution or a contradictory state, then return it and stop;
b) mark all the candidates remaining in RS as "not-tried";
c) choose some "not-tried" candidate Z, un-mark it and apply T&E(T, Z, RS);
d) if Z has been eliminated from RS by step c,
   then goto a
   else if there remains at least one "not-tried" candidate in RS
        then goto c else return RS and stop.

Definition: given a resolution theory T with the confluence property and an instance P with initial resolution state $RS_P$, we define *T&E(T, P)* as T&E(T, $RS_P$).

Notice that this procedure always stays at depth 1 (i.e. only one candidate is tested at a time) but that a candidate Z may be tried several times for T&E(T, Z, $RS_i$) in different resolution states $RS_i$. This is normal, because the result may be different if other candidates have been eliminated in the meanwhile. This also guarantees that the result of this procedure does not depend on the order in which remaining candidates are "tried".

We say that P can be solved by *T&E(T)*, or that P is in T&E(T), if T&E(T, P) produces a solution for P. When T is the Basic Resolution Theory of a CSP (which is known to always have the confluence property), we simply write *T&E* instead of T&E(BRT(CSP))).

### 5.6.2. The "T&E vs braids" theorem

Consider the simplest resolution theory T = BRT(CSP). It is obvious that any elimination that can be done by a braid B can be done by T&E (by applying rules from BRT following the structure of B). The converse is more interesting:

**Theorem 5.7: for any instance of any CSP, any elimination that can be done by T&E can be done by a braid. Any instance of a CSP that can be solved by T&E can be solved by braids.**

Proof: Let RS be a resolution state and let Z be a candidate eliminated by T&E(BRT, Z, RS) using some auxiliary resolution state RS'. Following the steps of BRT in RS', we progressively build a braid in RS with target Z. First, remember that BRT contains three types of rules: ECP (which eliminates candidates), S (which asserts a value for a CSP variable) and CD (which detects a contradiction on a CSP variable).

Consider the first step of BRT in RS' that is an application of rule S, asserting some label $R_1$ as a value. As $R_1$ was not a value in RS, there must have been in RS' some elimination of a candidate, say $L_1$, for a CSP variable $V_1$ of which $R_1$ is a candidate, and the elimination of $L_1$ (which made the assertion of $R_1$ by S possible in RS') can only have been made possible in RS' by the assertion of Z. But if $L_1$ has been eliminated in RS', it can only be by ECP and because it is linked to Z. Then $\{L_1\ R_1\}$ is the first pair of candidates of our braid in RS and $V_1$ is its first CSP variable. (Notice that there may be other z-candidates for $V_1$, but this is pointless, we can choose any of them as $L_1$ and consider the remaining ones as z-candidates).

The sequel is done by recursion. Suppose we have built a braid in RS corresponding to the part of the BRT resolution in RS' up to its k-th assertion step. Let $R_{k+1}$ be the next candidate asserted by BRT in RS'. As $R_{k+1}$ was not a value in RS, there must have been in RS' some elimination of a candidate, say $L_{k+1}$, for a CSP variable $V_{k+1}$ of which $R_{k+1}$ is a candidate, and the elimination of $L_{k+1}$ (which made the assertion of $R_{k+1}$ possible in RS') can only have been made possible in RS' by the assertion of Z and/or of some of the previous $R_i$. But if $L_{k+1}$ has been eliminated in RS', it can only be by ECP and because it is linked to Z or to some of the previous $R_i$, say C. Then our partial braid in RS can be extended to a longer one, with $\{L_{k+1}\ R_{k+1}\}$ added to its candidates, $L_{k+1}$ linked to C, and $V_{k+1}$ added to its sequence of CSP variables.

End of the procedure: as Z is supposed to be eliminated by T&E(Z, RS), a contradiction must have been obtained by BRT in RS'. As, in BRT, only ECP can eliminate a candidate, a contradiction is obtained if a value asserted in RS', i.e. Z or one of the $R_i$, i<n, eliminates in RS' (via ECP) a candidate, say $L_n$, that was the last one for a corresponding variable $V_n$ and that is linked to Z or one of the $R_i$, i<n. $L_n$ and $V_n$ are thus the last left-linking candidate and CSP variable of the braid we were looking for in RS.

Here again (as in the proof of confluence), this proof works only because the existence of a link between two candidates does not depend on the resolution state. Finally, notice that it is very unlikely that the T&E procedure followed by the construction in this proof would produce the shortest available braid in resolution state RS (and this intuition is confirmed by experience).

### 5.6.3. Comments on T&E and on the "T&E vs braids" theorem

As using T&E(T) leads to examining arbitrary hypotheses for the creation of auxiliary resolution states, it could be considered as blind search. Nevertheless, as T has the confluence property, the final result of T&E(T) applied to any instance does not depend in any way on the sequence of tested candidates.

#### 5.6.3.1. T&E versus structured search: no-guessing

Moreover, it is essential for our purposes and for our vague initial "no guessing" requirement to notice that, contrary to the usual structured search algorithms [e.g.

depth-first or breadth-first search, with search paths pruned by the rules in T – DFS(T) or BFS(T)], T&E(T) includes no "guessing": if a solution is obtained in an auxiliary state RS', then it is not taken into account. This notion of "guessing" is inherent to the DFS or BFS procedures. Closely related to it is the idea of a "backdoor" of an instance (see section 11.5.3): a set of labels of minimal cardinality such that adding them as values to the instance would give a solution within T, i.e. with no search at all. But this idea is totally alien to T&E(T).

As a result of the "no guessing" and no recursion, there is a major difference between T&E(T) and general DFS(T) and BFS(T): whereas, given any instance P, the latter algorithms can always find a solution (if there is any) or prove that it has none, T&E(T) cannot: if P has multiple solutions, T&E(T) can only find what is common to all its solutions. Given the "T&E(T) vs T-braids" theorem (this theorem will be proved for many resolution theories T) and the correspondence between a solution of P in a resolution theory T' and a model of T' ∪ $E_P$, this is just the basic fact that what can be proved in a FOL theory (here T' = T-braids) is (and can only be) what is true in all the models of this theory. Notice that another consequence of this basic property of FOL is that, given T, there cannot exist any resolution theory TT such that one would get a "DFS(T) vs TT" or a "BFS(T) vs TT" theorem.

*5.6.3.2. Comments on the "acceptability" of braids*

In the Sudoku community, T&E (which had always been the topic of heated debates, although it had never been precisely defined before *HLS*) is generally not accepted by advocates of "pattern-based" solutions. But the above "T&E vs braids" theorem shows that a solution based on T&E can always be replaced by a rule-based solution, more precisely by a solution based on braids. The question naturally arises, for any CSP: can one reject T&E and nevertheless accept solutions based on braids? There are three main reasons for a positive answer, both related to the goals one pursues.

Firstly, as shown in section 5.5, resolution theories based on braids have the confluence property and many different resolution strategies can be super-imposed on them. One may prefer a solution with the shortest braids and adopt the "simplest first" strategy defined in section 5.5. The T&E procedure cannot provide this (unless it is drastically modified, in ways that would make it computationally very inefficient).

Secondly, in each of the $B_n$ resolution theories based on braids, one can add rules corresponding to special cases, such as whips or bivalue-chains of same length, and one can decide to give a natural preference to such special cases. This is still a "simplest first" principle. In Sudoku (and in most of the other examples we have analysed), this entails that non-whip braids appear very rarely in the solution of randomly generated puzzles; in a sense, this is a measure of how powerful whips are: although they are structurally much more "beautiful" and simpler (they are continuous chains with no "branching") and computationally much better than braids, they can solve almost all the puzzles that can be solved by T&E (i.e. almost all the randomly generated ones). One could say that the "T&E vs braids" theorem (together with the statistical results of chapter 6 and the subsumption results of chapter 8) is the best advertisement for whips.

Thirdly, in spite of what some Sudoku addicts would like to believe or make believe, the reality is that most of the Sudoku players (and, more generally, players of logic puzzles) heavily rely on T&E as their main and most natural resolution strategy for the non-trivial instances. Trying to find a braid or a whip justifying an elimination in a simpler way than what they have first found by T&E may thus be an entertaining idea. The same remarks may be applied to any CSP.

### 5.7. The objective properties of chains and braids

Chains should not be confused with chain rules. A chain rule can only be valid or not valid, which depends neither on the way it has been proven nor on any of the properties defined below for the underlying chain. A non-valid chain rule is merely useless. But a valid chain rule can be more a less general (giving rise to subsumption relationships), more or less useful, easy to apply, acceptable. As (apart from the first) these are purely subjective criteria, they can only lead to confusion if they cannot be grounded in objective ones.

We have therefore devised a few, purely objective (or descriptive, or factual) properties of chains that may be relevant to estimate their usefulness, desirability or understandability. Even these objective properties can give rise to much debate when it comes to subjectively evaluating their impact on usefulness or acceptability; it all depends on which criteria of acceptability are adopted.

#### 5.7.1. Linearity (sequentiality)

We use the words *linearity* or *sequentiality* as synonyms to mean that the candidates composing the pattern are sequentially ordered; it is supposed that this order is essential in the definition of the pattern (i.e. not arbitrarily super-imposed on it) and in the proof of the associated resolution rule. Linearity is what makes the difference with a net: a net has a DAG (directed acyclic graph) structure; in a net, only a partial ordering of the candidates is required, while there may be branching and merging of different paths. Both whips and braids are linear.

#### 5.7.2. Continuity

*Continuity* supposes linearity and means that consecutive candidates are linked. In this definition, possible additional t- or z- candidates of whips and braids, which are not considered as part of the pattern, do not alleviate in any way this requirement (they are considered as inessential). Continuity is what distinguishes whips from braids: braids satisfy linearity but not continuity.

### 5.7.3. Homogeneity

*Homogeneous* means that the pattern is a sequence of similar bricks. This vague property is obvious for all the chains and braids introduced here.

### 5.7.4. Reversibility

Although it had never been defined before *HLS2*, the word "reversibility" has been the pretext for the most poisonous debates on Sudoku Web forums. There is nevertheless an obvious definition, valid for any CSP:

– given a sequential pattern, the reversed pattern is the sequential pattern obtained by reversing the order of the candidates; in the process, when used in the definition of some types of chains, left- [respectively right-] linking candidates become right- [resp. left-] linking candidates;

– a given type of sequential pattern is called *reversible* if for any pattern of this type, the reversed pattern is of this type.

These definitions will be extended in chapters 9 and 10 to sequential patterns with more general right-linking objects.

**Theorem 5.8: bivalue-chains and z-chains are reversible.**

Proof: obvious (left and right-linking candidates are interchanged).

The advantage of reversibility is that, in general, one can find other chains by "circulating along the chain" (i.e. making circular permutations of the candidates and changing the endpoints accordingly); these often allow other eliminations.

Notice that chains (and braids) using the t-extension are not reversible. This is a weak point for them. But the sequel will show that they satisfy properties (left-extendibility and composability) that partially palliate this weakness.

### 5.7.5. Non anticipativeness (or no look-ahead)

Definition: a given type of sequential pattern is called *non-anticipative* or *no look-ahead* if, when a pattern of this type is built from left to right, all that needs be checked when the next candidate is added depends only on the previous candidates (and not on the potential future ones) and possibly on the target (for patterns that have to be built around a target, such as whips or braids). Notice that this does not imply that adding a candidate will always allow to finally get a full pattern of this type, but it guarantees that, up to the new candidate added, the pattern satisfies the conditions on patterns of this type whatever will be added to it later.

Comment: this seems to be a strong criterion for acceptability of sequential patterns, from both points of view of human solvers and programmers, because it is the practical condition necessary for being able to build the pattern progressively from left to right, instead of having to spot it globally at once. It is a major computational property, the opposite of which is look-ahead.

***Theorem 5.9: a reversible chain is non-anticipative.***

***Theorem 5.10: all the sequential patterns defined in this chapter, from bivalue-chains to whips and braids, are non-anticipative.***

Proofs: obvious. Indeed, we had implicitly this condition in mind when we introduced the first types of chains in *HLS1*.

### 5.7.6. Left-extendibility and composability

Definition: a given type of sequential pattern is called *left-extendable* if, when given a partial pattern of this type, candidates can be added not only to its right but also to its left (of course, respecting the linking conditions on left- and right- linking candidates for patterns of this type at the junction and having the same target in case they are built around a target).

***Theorem 5.11: a reversible chain is left-extendable.***

***Theorem 5.12: a non-anticipative chain is left-extendable.***

***Theorem 5.13: all the sequential patterns defined in this chapter, from bivalue-chains to whips and braids, are left-extendable.***

Proof: obvious. The idea is that, when the presence of a t-candidate can be justified by previous right-linking candidates in a partial chain, it will remain justified by them if we add candidates to the left of this partial chain (and justifications of z-candidates will not be changed). This notion and theorem 5.13 were first suggested by Mike Barker.

Definition: a given type of sequential pattern is called *composable* if, when two partial patterns of this type are given, they can be combined into a single pattern of this type (of course, respecting the linking conditions on left- and right- linking candidates for chains of this type at the junction and having the same target in case they are built around a target).

***Theorem 5.14: all the sequential patterns defined in this chapter, from bivalue-chains to whips and braids, are composable.***

The practical impact of this theorem is mainly for sequential patterns with the t-extension (t-whips, zt-whips, t-braids and zt-braids): when t-candidates are justified by previous right-linking candidates of a partial pattern, they will still be justified by the same candidates if another partial pattern of the same type is added to its left. Of course, not all the sequential patterns with the t-extension can be obtained by combining shorter patterns of the same type, but looking first for combinations of such shorter sub-patterns before patterns with longer distance t-interactions may be a valuable strategy, different from the one described in section 5.5.2 (and it can also be combined with it in order to keep taking advantage of the confluence property).

### 5.7.7. No OR-branching

All the chain/whip/braid patterns introduced in this chapter and all their extensions that will appear later on in this book have two essential properties in common:

– they involve no OR-branching,

– they involve only structured AND-branching.

*5.7.7.1. How do you branch?: AND-branching vs OR-branching*

Originating in the theorem proving literature and the associated backwards-chaining view (and closely related to PROLOG-like languages), there is a classical distinction in AI between AND-branching and OR-branching. Whereas the conditions of only one of the branches are required to be satisfied at any OR-branching point, AND-branching is much more complex because the conditions of all the branches are required to be simultaneously satisfied at any AND-branching point.

Transposed to the forward-chaining view that better applies to our approach, OR-branching becomes the most complex of the two. OR-branching corresponds to patterns where alternative possibilities would be allowed to appear, namely, instead of having only one right-linking candidate (or, anticipating on later chapters, right-linking pattern), one would have several. From the point of view of logic, OR-branching in forward-chaining is equivalent to reasoning by cases, which is perfectly valid in theory (even in intuitionistic logic): if one has $A_1 \Rightarrow B$, $A_2 \Rightarrow B$, … and $A_n \Rightarrow B$, then one can conclude $A_1 \vee A_2 \vee \ldots \vee A_n \Rightarrow B$.

But, in practice, mathematicians do not like reasoning by cases very much and it is relatively important for the current discussion to understand why. In addition to the often invoked reason that it is inelegant, especially if it is repeated several times in a proof (subcases of subcases of …), there is always the suspicion that it fails to find deeper properties common to all the cases. This is true even without recursion, as shown by the most famous example of an extensive use of reasoning by cases, the proof of the four-colour theorem ("every planar graph is 4-colourable" or, more informally, "every map can be coloured by only four colours"). In 1976, Appel and Haken proposed a proof reducing the theorem to 1,936 particular cases [twenty years later, this number was brought down to "only" 633, but this is irrelevant here] and they proved all these cases separately by a computer program. There have been many arguments against this type of proof: 1) the final step (the 1,936 cases) was done by a computer program, which could always be suspected of having bugs; 2) there are so many cases (even in the improved version) that it is impossible for a human being to check them all. But, in our view, the most powerful objection does not bear on validity; it is that this final part of the proof is meaningless, it does not teach us anything general, it involves no general mathematical knowledge – and this objection would remain relevant even if there were only a dozen cases.

It should now be stressed that ***none of the patterns introduced in this book involve OR-branching*** – except the forcing-whips and forcing-braids quickly mentioned below in section 5.9. Even forcing-bi-braids (see chapter 12), if properly construed as B*-braids[1], do not rely on OR-branching. g-candidates (chapter 7) or inner Subsets (chapter 8) could be considered as involving a form of OR-branching, but they are wrapped in such a way in the S-braids, S-whips, g-braids or g-whips that this pseudo OR-branching is limited to one step and can only merge in predefined labels.

*5.7.7.2. Structured AND-branching vs free AND-branching*

As for AND-branching, we said that all the patterns introduced in this book involve only *structured* forms of it. There are actually only two forms:

– in both whips and braids: from the target or a right-linking candidate (or object) to a left-linking candidate and to the associated z- and t- candidates;

– in braids: from a left-linking candidate to possibly several right-linking ones.

*5.7.8. Complexity*

As whips or braids are much more general than bivalue-chains, the search for whips or braids in a real resolution state of a real instance of a CSP is likely to be more difficult than the search for the simplest bivalue-chains of same length. The counterpart is, the former can solve many more instances (see chapter 6).

Unfortunately, defining an objective complexity measure for the instances of a CSP is a very difficult task. Whereas worst case analysis is not very meaningful, mean case analysis is more meaningful but is very difficult in practice, as will be illustrated by the Sudoku case in chapter 6, where the $W_n$ and $B_n$ ratings will be shown to be reasonable measures of complexity.

**5.8. About loops in bivalue-chains, in whips and in braids**

We say that there is a loop in a sequential pattern if it has two identical candidates. In this section, we review the usefulness of accepting various kinds of loops in the different chains or braids introduced in this chapter.

*5.8.1 Global loops are useless in bivalue-chains*

Define a global loop as a chain with same first and last candidates; this is the broadest definition of a global loop one can give for a chain.

Consider a bivalue-chain $C = \{L_1 R_1\} - \{L_2 R_2\} - \ldots - \{L_n R_n\}$ with target Z and with a global loop, i.e. $R_n = L_1$. (Notice that this situation is globally contradictory and that such a pattern could be used to detect contradictory instances of a CSP, but this is not the question we want to deal with here.) We shall show that Z can be eliminated by rules from BRT and by a shorter bivalue-chain with no loop.

The bivalue-chain obtained by excising the last pair of candidates from C, i.e. $\{L_1 R_1\} - \{L_2 R_2\} - \ldots - \{L_{n-1} R_{n-1}\}$, admits $L_n$ as a target: $L_n$ is linked to its first candidate (because $L_1 = R_n$) and to its other endpoint ($R_{n-1}$). $L_n$ can therefore be eliminated by this shorter bivalue-chain with no global loop. After this, $R_n$ can be asserted by rule S (because the CSP variable $V_n$ of $\{L_n R_n\}$ in C was bivalue); and Z can be deleted by rule ECP.

As a result, we have:

***Theorem 5.15: Any elimination that could be done by a bivalue-chain with a global loop can be done by BRT(CSP) and by a shorter bivalue-chain with no global loop. Practical statement: global loops are useless in bivalue-chains.***

*5.8.2. Inner loops are useless in bivalue-chains*

We say that a chain has an inner loop if it has two equal candidates, but at most one of them is an endpoint.

Let $\{L_1 R_1\} - \{L_k R_k\} - \ldots - \{L_p R_p\} - \ldots - \{L_n R_n\}$ be a bivalue-chain and suppose it has an inner loop. Let Z be a target. There are two possibilities for an inner loop.

The first possibility is the equality of two left-linking candidates or of two right-linking candidates: $L_k = L_p$ or $R_k = R_p$. Then, by excision of the inner loop, we get a shorter bivalue-chain with Z as a target:
$\{L_1 R_1\} - \{L_{k-1} R_{k-1}\} - \{L_p R_p\} - \ldots - \{L_n R_n\}$ or
$\{L_1 R_1\} - \{L_k R_k\} - \{L_{p+1} R_{p+1}\} - \ldots - \{L_n R_n\}$.

The second possibility is the equality of a right-linking candidate with a subsequent or a previous left-linking candidate, corresponding respectively to the two cases $R_k = L_p$ and $L_k = R_p$.

In the first case, by excision of the extremities, we get a shorter bivalue-chain: $\{L_{k+1}\ R_{k+1}\} - \ldots - \{L_{p-1}\ R_{p-1}\}$ with $R_k = L_p$ as a target. Once it has been used to eliminate $L_p$, rules S and ECP from BRT(CSP) will progressively assert all the right-linking candidates and eliminate all the left-linking candidates after p. After $R_n$ has been asserted by S, Z will be eliminated by ECP.

The second case can be dealt with in exactly the same way, after reversing the original chain (which reverses the role of candidates: left-linking become right-linking and conversely).

In case the original chain had several inner loops, all these reductions can be applied iteratively to as many subparts of the chain as necessary; every iteration eliminates one loop, until there remains none. Finally, we get:

***Theorem 5.16: Any elimination that could be done by a bivalue-chain with inner loops can be done by BRT(CSP) and by a shorter bivalue-chain with no inner loop. Practical statement: inner loops are useless in bivalue-chains.***

*5.8.3. Bivalue-chains should have no loops*

As a general conclusion of all the preceding cases, we have:

***Theorem 5.17: resolution rules that might be obtained from bivalue-chains with global or inner loops are subsumed by BRT(CSP) together with rules for shorter bivalue-chains with no loops. Practical statement: bivalue-chains should have no such loops.***

*5.8.4. Should one allow loops in whips and braids?*

In whips or braids, equality of a left-linking and a right-linking candidate is the closest notion we can have of a loop; but such equality would produce the final whip contradiction. In particular, a z-chain with a global loop would merely be a z-whip.

As for other kinds of inner loops in whips (equality of two left-linking or of two right-linking candidates), nothing allows to eliminate them. *A priori*, one can consider that, as we go forward along a whip, we accumulate knowledge about the consequences of assuming its target, which allows more possibilities of extending it; allowing such inner loops could therefore lead to accumulate more knowledge and to find more whips.

However, although the general definition of a whip does not exclude loops, experience with the Sudoku example shows that they do not bring much more generality but they bring more computational complexity. Moreover, in any CSP, whips with loops are subsumed by braids and it may be more interesting to use braids than whips with inner loops. Therefore, we do not add any *a priori* no-loop condition in the general definition of a whip, but, unless otherwise stated, all the whips we shall consider will be loopless. In particular, the statistical results for Sudoku in chapter 6 are about loopless whips.

As for braids, the notion of an inner loop is pointless: the same left-linking or right-linking candidate can be used several times for sprouting new branches, which has the same "accumulation" result as loops, but without the useless parts that may be needed to join the endpoints of a loop; i.e. for any possible inner loop, there is always, obviously, a shorter braid without this loop.

**5.9. Forcing whips and braids, a bad idea?**

Consider a bivalue variable (in any resolution state), with its two possible values $x_1$ and $x_2$ corresponding to candidates $Z_1$ and $Z_2$. Suppose there are two partial whips/braids, say $W_1$ and $W_2$, one with target $Z_1$, the other with target $Z_2$. In case $W_1$ and $W_2$ share a left-linking candidate L [respectively a right-linking candidate R], L can be deleted [resp. R can be asserted]: this is reasoning by cases, which is perfectly valid in intuitionistic logic. Do we get interesting new patterns this way (to be called forcing whips / forcing braids because they *force* a conclusion that could not be obtained by a single whip/braid pattern)? Obviously, the answer would be negative for reversible chains: it would suffice to reverse one of the chains and to link the two by the bivalue variable in order to obtain a single chain of same type as the two given ones. Notice that in this process the chain thus obtained should be assigned length $n_1+n_2+1$, where the $n_i$ are the lengths of the two chains.

But as whips and braids are not reversible, it seems one could get new patterns, more general than whips and braids. What is the resolution power of such patterns? We have no general answer. But, in the Sudoku case, if these patterns are assigned length $n_1+n_2+1$ and given smaller priority than whips/braids of same length, we have found no occurrence in a random sample of 1,300 puzzles. As the memory requirements for such combinations are very high, we did not try on larger samples.

There is still the possibility of starting from trivalue variables and considering three whips/braids instead of two, but the complexity increases accordingly.

See chapter 12 for the definition of a much broader type of patterns based on similar ideas, but with drastically increased resolution power.

**5.10. Exceptional examples**

All the resolution rules defined in this chapter have been implemented in our CSP-Rules solver in a way valid for any CSP. Each of them can be activated or de-activated independently. Different strategies can be chosen. In the examples below, we systematically apply the "simplest first" resolution strategy defined in section 5.5.2, with whips [respectively whips and braids] activated, in order to get the W [respectively the B] rating.

The longest whip(s) or braid(s) of each resolution path appears in bold characters. As for the notation (the "nrc notation"), it is self explaining and consistent with the general representation for whips and braids introduced in sections 5.2 and 5.3; the only adaptations are: 1) outside curly brackets, CSP Variables $X_{rc}$, $X_{rn}$, … are merely written as rc, rn, …; 2) within curly brackets, the s value of an $X_{bn}$ variable is replaced by its equivalent in rc-coordinates; the reason is better readability on the standard rc-grid. Apart from some hand editing, the following is the raw output from SudoRules. Handmade changes (in addition to those mentioned in the Introduction) have the only purpose of using less paper; they consist mainly of writing several whips in the same line (even if, because they have different targets, the justifications of their z-candidates may be different).

A general warning is in order about our Sudoku examples: because they are intended to illustrate exceptional properties, most of them are much more difficult than the vast majority of puzzles (see the classification results in chapter 6); as a result, they have exceptionally long resolution paths with exceptionally long whips/braids and they may give a very wrong idea of the much simpler typical resolution paths. A "normal" puzzle can be solved with only a few rule applications (not counting ECP), often even less than in the example of Figure 5.1.

***5.10.1. Proof of $B_4 \ne W_4$: an instance with $W(P) = 5$ and $B(P) = 4$***

|   | 2 |   | 4 |   |   | 7 |   |   |
|---|---|---|---|---|---|---|---|---|
|   |   |   |   | 8 | 9 | 1 |   |   |
|   |   |   |   |   |   |   | 6 | 5 |
|   |   | 4 | 8 |   |   |   |   |   |
| 3 |   |   | 9 |   |   |   |   | 1 |
|   | 9 | 5 |   |   | 1 |   | 7 |   |
|   | 7 |   | 3 |   |   |   | 1 | 2 |
| 6 | 3 |   |   |   |   |   |   |   |
|   |   | 2 |   | 1 |   |   |   | 8 |

| 1 | 2 | 3 | 4 | 5 | 6 | 7 | 8 | 9 |
|---|---|---|---|---|---|---|---|---|
| 4 | 5 | 6 | 7 | 8 | 9 | 1 | 2 | 3 |
| 7 | 8 | 9 | 1 | 2 | 3 | 4 | 6 | 5 |
| 2 | 1 | 4 | 8 | 7 | 5 | 3 | 9 | 6 |
| 3 | 6 | 7 | 9 | 4 | 2 | 8 | 5 | 1 |
| 8 | 9 | 5 | 6 | 3 | 1 | 2 | 7 | 4 |
| 5 | 7 | 8 | 3 | 6 | 4 | 9 | 1 | 2 |
| 6 | 3 | 1 | 2 | 9 | 8 | 5 | 4 | 7 |
| 9 | 4 | 2 | 5 | 1 | 7 | 6 | 3 | 8 |

**Figure 5.2.** *A puzzle P with B(P) = 4 and W(P) = 5*

The example in Figure 5.2 is one of the rare (in percentage) puzzles with a B rating smaller than its W rating.

1) The resolution path with whips shows that W(P) = 5:

***** SudoRules 16.2 based on CSP-Rules 1.2, config: W *****
26 givens, 196 candidates, 1151 csp-links and 1151 links. Initial density = 1.51
singles ==> r8c9 = 7, r8c3 = 1, r1c1 = 1, r4c2 = 1, r3c4 = 1, r8c6 = 8
whip[1]: c9n6{r6 .} ==> r4c7 ≠ 6, r5c7 ≠ 6, r6c7 ≠ 6
whip[1]: r1n5{c5 .} ==> r2c4 ≠ 5
whip[1]: c4n5{r9 .} ==> r8c5 ≠ 5, r7c6 ≠ 5, r7c5 ≠ 5, r9c6 ≠ 5
whip[1]: b4n6{r5c2 .} ==> r5c6 ≠ 6, r5c5 ≠ 6
whip[2]: c8n8{r1 r5} – c2n8{r5 .} ==> r1c3 ≠ 8
hidden-single-in-a-row ==> r1c8 = 8
whip[3]: r9c2{n4 n5  r7n5{c1 c7} – b9n6{r7c7 .} ==> r9c7 ≠ 4
whip[4]: r6c1{n2 n8} – r5n8{c3 c7} – r5n2{c7 c8} – r2n2{c8 .} ==> r6c4 ≠ 2
singles ==> r6c4 = 6, r4c9 = 6, r1c9 = 9
whip[1]: r2n6{c2 .} ==> r1c3 ≠ 6
naked-single ==> r1c3 = 3
whip[1]: r2n3{c9 .} ==> r3c7 ≠ 3
whip[4]: r6n8{c7 c1} – r6n2{c1 c5} – c6n2{r4 r3} – r3c7{n2 .} ==> r6c7 ≠ 4
whip[4]: b9n3{r9c8 r9c7} – r9n6{c7 c6} – r7c6{n6 n4} – r8n4{c5 .} ==> r9c8 ≠ 4
whip[4]: r6n4{c5 c9} – c8n4{r5 r2} – r2n2{c8 c4} – b8n2{r8c4 .} ==> r8c5 ≠ 4
whip[1]: r8n4{c8 .} ==> r7c7 ≠ 4
whip[4]: b9n4{r8c8 r8c7} – r3c7{n4 n2} – r2n2{c8 c4} – r8c4{n2 .} ==> r8c8 ≠ 5
whip[4]: b8n9{r7c5 r8c5} – r8c8{n9 n4} – r8c7{n4 n5} – r7n5{c7 .} ==> r7c1 ≠ 9
whip[4]: b7n9{r9c1 r7c3} – b7n8{r7c3 r7c1} – r7n5{c1 c7} – b9n6{r7c7 .} ==> r9c7 ≠ 9
whip[4]: r9n9{c1 c8} – b9n3{r9c8 r9c7} – b9n6{r9c7 r7c7} – r7n5{c7 .} ==> r9c1 ≠ 5
;;; Resolution state RS$_1$

**whip[5]: c2n5{r2 r9} – r9c4{n5 n7} – r2c4{n7 n2} – b3n2{r2c8 r3c7} – b3n4{r3c7 .} ==> r2c2 ≠ 4**
whip[4]: c2n8{r5 r3} – c1n8{r3 r7} – b7n5{r7c1 r9c2} – c2n4{r9 .} ==> r5c3 ≠ 8
whip[2]: c3n9{r3 r7} – c3n8{r7 .} ==> r3c3 ≠ 7
whip[3]: r3c7{n2 n4} – r3c2{n4 n8} – r5n8{c2 .} ==> r5c7 ≠ 2
whip[3]: c6n2{r4 r3} – r2n2{c4 c8} – r5n2{c8 .} ==> r4c5 ≠ 2
whip[3]: c6n2{r5 r3} – r2n2{c4 c8} – r5n2{c8 .} ==> r6c5 ≠ 2
whip[2]: r6c9{n3 n4} – r6c5{n4 .} ==> r6c7 ≠ 3
whip[4]: b6n9{r4c7 r4c8} – c8n2{r4 r2} – c8n3{r2 r9} – c7n3{r9 .} ==> r4c7 ≠ 2
whip[4]: b9n5{r9c7 r9c8} – r9c2{n5 n4} – r3c2{n4 n8} – r5n8{c2 .} ==> r5c7 ≠ 5
whip[3]: r3c7{n4 n2} – r6c7{n2 n8} – r5c7{n8 .} ==> r8c7 ≠ 4
hidden-single-in-a-block ==> r8c8 = 4
whip[3]: r3c2{n4 n8} – r5n8{c2 c7} – c7n4{r5 .} ==> r3c1 ≠ 4
whip[3]: b1n5{r2c1 r2c2} – r9c2{n5 n4} – c1n4{r9 .} ==> r2c1 ≠ 7
whip[2]: c5n7{r4 r3} – c1n7{r3 .} ==> r4c6 ≠ 7
whip[4]: b7n9{r9c1 r7c3} – b7n8{r7c3 r7c1} – r7n5{c1 c7} – r8c7{n5 .} ==> r9c8 ≠ 9
singles ==> r4c8 = 9, r9c1 = 9, r7c3 = 8, r3c3 = 9

whip[3]: r4c7{n3 n5} – r5c8{n5 n2} – b5n2{r5c5 .} ==> r4c6 ≠ 3
hidden-single-in-a-column ==> r3c6 = 3
whip[1]: c6n2{r5 .} ==> r5c5 ≠ 2
whip[2]: c3n7{r5 r2} – b2n7{r2c4 .} ==> r5c5 ≠ 7
whip[3]: c7n4{r5 r3} – c7n2{r3 r6} – r5n2{c8 .} ==> r5c6 ≠ 4
whip[1]: c6n4{r9 .} ==> r7c5 ≠ 4
whip[3]: c8n5{r5 r9} – r9c4{n5 n7} – c6n7{r9 .} ==> r5c6 ≠ 5
whip[3]: r4c6{n5 n2} – r5n2{c6 c8} – b6n5{r5c8 .} ==> r4c5 ≠ 5
whip[3]: r9n6{c7 c6} – r1c6{n6 n5} – r4n5{c6 .} ==> r9c7 ≠ 5
whip[4]: r9c2{n4 n5} – r9c8{n5 n3} – b3n3{r2c8 r2c9} – b3n4{r2c9 .} ==> r3c2 ≠ 4
singles to the end

2) The resolution path with braids shows that B(P) = 4:

***** SudoRules 16.2 based on CSP-Rules 1.2, config: B *****
;;; same path up to $RS_1$ (no braid appears before); after, the two paths diverge:
**braid[4]: r2c4{n7 n2} – r4c1{n7 n2} – c8n2{r2 r5} – c6n2{r5 .} ==> r2c1 ≠ 7**
whip[2]: c5n7{r4 r3} – c1n7{r3 .} ==> r4c6 ≠ 7
whip[3]: c2n4{r2 r9} – c2n5{r9 r2} – r2c1{n5 .} ==> r3c1 ≠ 4
whip[4]: b1n6{r2c2 r2c3} – r2n7{c3 c4} – r9c4{n7 n5} – c2n5{r9 .} ==> r2c2 ≠ 4
whip[4]: c2n8{r5 r3} – c1n8{r3 r7} – b7n5{r7c1 r9c2} – c2n4{r9 .} ==> r5c3 ≠ 8
whip[2]: c3n9{r3 r7} – c3n8{r7 .} ==> r3c3 ≠ 7
whip[3]: r3c7{n2 n4} – r3c2{n4 n8} – r5n8{c2 .} ==> r5c7 ≠ 2
whip[3]: c6n2{r4 r3} – r2n2{c4 c8} – r5n2{c8 .} ==> r4c5 ≠ 2
whip[3]: c6n2{r5 r3} – r2n2{c4 c8} – r5n2{c8 .} ==> r6c5 ≠ 2
whip[2]: r6c9{n3 n4} – r6c5{n4 .} ==> r6c7 ≠ 3
whip[4]: b6n9{r4c7 r4c8} – c8n2{r4 r2} – c8n3{r2 r9} – c7n3{r9 .} ==> r4c7 ≠ 2
whip[4]: b9n5{r9c7 r9c8} – r9c2{n5 n4} – r3c2{n4 n8} – r5n8{c2 .} ==> r5c7 ≠ 5
whip[3]: r3n4{c7 c2} – c2n8{r3 r5} – r5c7{n8 .} ==> r8c7 ≠ 4
hidden-single-in-a-block ==> r8c8 = 4
whip[4]: b7n9{r9c1 r7c3} – b7n8{r7c3 r7c1} – r7n5{c1 c7} – r8c7{n5 .} ==> r9c8 ≠ 9
singles ==> r4c8 = 9, r9c1 = 9, r7c3 = 8, r3c3 = 9
whip[3]: r4c7{n3 n5} – r5c8{n5 n2} – b5n2{r5c5 .} ==> r4c6 ≠ 3
hidden-single-in-a-column ==> r3c6 = 3
whip[1]: c6n2{r5 .} ==> r5c5 ≠ 2
whip[2]: c3n7{r5 r2} – b2n7{r2c4 .} ==> r5c5 ≠ 7
whip[3]: c7n4{r5 r3} – c7n2{r3 r6} – r5n2{c8 .} ==> r5c6 ≠ 4
whip[1]: c6n4{r9 .} ==> r7c5 ≠ 4
whip[3]: c8n5{r5 r9} – r9c4{n5 n7} – c6n7{r9 .} ==> r5c6 ≠ 5
whip[3]: r4c6{n5 n2} – r5n2{c6 c8} – b6n5{r5c8 .} ==> r4c5 ≠ 5
whip[3]: r9n6{c7 c6} – r1c6{n6 n5} – r4n5{c6 .} ==> r9c7 ≠ 5
whip[4]: r2c9{n3 n4} – r2c1{n4 n5} – r7n5{c1 c7} – r4c7{n5 .} ==> r6c9 ≠ 3
singles to the end

### *5.10.2. Proof of $B_\infty \neq W_\infty$ : an instance with $W(P) = \infty$ and $B(P) = 12$*

After the previous example, one may still wonder: if a puzzle can be solved by braids, cannot one always find whips, though longer than the braids, such that they will also solve it? Said otherwise, is not $B_\infty$ equal to $W_\infty$? The answer is negative; there are puzzles that can be solved by braids but not by whips of any length. The example in Figure 5.3 is one of the exceptional (in percentage) puzzles in this case (see statistics in chapter 6); it is the only one in the whole "Magictour top 1465" collection; its B rating is 12 but its W rating is ∞.

|   |   | 3 |   | 5 |   |   |   |
|---|---|---|---|---|---|---|---|
|   | 5 |   | 1 |   |   | 3 |   |
|   |   | 7 |   | 4 |   |   | 1 |
| 2 |   |   |   | 4 |   |   |   |
|   | 6 |   | 9 |   |   |   |   |
|   | 1 |   |   | 6 |   |   | 2 |
| 8 |   | 7 |   | 2 |   |   |   |
|   | 9 |   | 8 |   | 5 |   |   |
|   | 5 |   | 9 |   |   | 7 |   |

| 9 | 1 | 4 | 3 | 7 | 8 | 5 | 2 | 6 |
|---|---|---|---|---|---|---|---|---|
| 6 | 5 | 8 | 9 | 1 | 2 | 7 | 3 | 4 |
| 3 | 2 | 7 | 5 | 6 | 4 | 9 | 8 | 1 |
| 2 | 8 | 9 | 1 | 3 | 7 | 4 | 6 | 5 |
| 4 | 6 | 3 | 2 | 9 | 5 | 1 | 7 | 8 |
| 5 | 7 | 1 | 8 | 4 | 6 | 3 | 9 | 2 |
| 8 | 4 | 6 | 7 | 5 | 3 | 2 | 1 | 9 |
| 7 | 9 | 2 | 4 | 8 | 1 | 6 | 5 | 3 |
| 1 | 3 | 5 | 6 | 2 | 9 | 8 | 4 | 7 |

**Figure 5.3.** *Puzzle Magictour top 1465 #89 and its solution; W = ∞ and B = 12*

Although the following resolution paths are exceptionally long, they have a feature typical of what one gets with the "simplest first" strategy: braids that are not whips appear much less often than whips. For puzzles P solvable by whips, if both whips and braids are activated, braids appear even more rarely – and they very rarely change the rating, i.e. $W(P) = B(P)$ most of the time. In both resolution paths below, one can also notice the long streaks of eliminations necessary before a new value can be asserted.

1) The resolution path with whips shows that $W(P) = \infty$ ; it also gives an example of a very long whip[18] (but there are much longer ones in other puzzles):

***** SudoRules 16.2 based on CSP-Rules 1.2, config: W *****
24 givens, 218 candidates, 1379 csp-links and 1379 links. Initial density = 1.46
hidden-single-in-a-row ==> r8c1 = 7
whip[2]: r5n7{c8 c6} – r2n7{c6 .} ==> r6c7 ≠ 7
whip[3]: c2n2{r1 r9} – c5n2{r9 r3} – b3n2{r3c8 .} ==> r1c3 ≠ 2
whip[3]: c5n2{r1 r9} – c2n2{r9 r3} – b3n2{r3c8 .} ==> r1c6 ≠ 2
whip[3]: c1n9{r2 r6} – c7n9{r6 r3} – r1n9{c8 .} ==> r2c3 ≠ 9
whip[3]: r2n2{c6 c3} – c2n2{r1 r9} – c5n2{r9 .} ==> r3c4 ≠ 2
whip[3]: b6n5{r4c9 r5c9} – b4n5{r5c1 r6c1} – b4n9{r6c1 .} ==> r4c9 ≠ 9
whip[3]: b7n2{r9c2 r8c3} – r2n2{c3 c6} – b5n2{r5c6 .} ==> r9c4 ≠ 2
whip[4]: b6n6{r4c8 r4c9} – b6n5{r4c9 r5c9} – b4n5{r5c1 r6c1} – b4n9{r6c1 .} ==> r4c8 ≠ 9

hidden-single-in-a-row ==> r4c3 = 9
whip[7]: b9n9{r7c9 r7c8} – r1n9{c8 c1} – r3n9{c1 c4} – b2n5{r3c4 r3c5} – b8n5{r7c5 r7c6} – r7n1{c6 c2} – c1n1{r9 .} ==> r2c9 ≠ 9
;;; Resolution state RS$_1$
whip[9]: c7n7{r2 r5} – c8n7{r4 r1} – r1n9{c8 c1} – c1n1{r1 r9} – c7n1{r9 r8} – b8n1{r8c6 r7c6} – b8n5{r7c6 r7c5} – b2n5{r3c5 r3c4} – b2n9{r3c4 .} ==> r2c7 ≠ 9
whip[11]: b3n7{r2c7 r1c8} – r1c6{n7 n8} – c3n8{r1 r5} – c9n8{r5 r4} – c8n8{r4 r9} – c8n4{r9 r7} – b9n9{r7c8 r7c9} – r1n9{c9 c1} – c1n1{r1 r9} – r7c2{n1 n3} – c3n3{r8 .} ==> r2c7 ≠ 8
**whip[18]: r1c6{n8 n7} – r2n7{c6 c7} – r5n7{c7 c8} – r6c8{n7 n9} – c7n9{r6 r3} – r1n9{c9 c1} – c1n1{r1 r9} – c2n1{r7 r1} – r1n2{c2 c5} – r2n2{c4 c3} – c3n8{r2 r5} – c9n8{r5 r4} – c7n8{r5 r9} – c7n6{r9 r8} – b7n6{r8c3 r7c3} – c3n3{r7 r8} – b9n3{r8c7 r7c9} – b9n9{r7c9 .} ==> r1c8 ≠ 8**

After this very long whip, there is no more elimination. (Whips are programmed up to length 36 in CSP-Rules and there is a mechanism for detecting the need for longer ones – it never fired! The same programmed maximum length is true of the braids and of the g-whips and g-braids to be introduced in chapter 7.)

2) The resolution path with braids shows that B(P) = 12:

***** SudoRules 16.2 based on CSP-Rules 1.2, config: B *****
;;; same path up to resolution state RS$_1$
;;; the next two eliminations were done by slightly longer whips (length +1) in the previous path
braid[8]: r1n9{c9 c1} – r3n9{c1 c4} – b2n5{r3c4 r3c5} – b8n5{r7c5 r7c6} – c1n1{r1 r9} – c7n7{r2 r5} – c7n1{r5 r8} – b8n1{r9c4 .} ==> r2c7 ≠ 9
braid[10]: b3n7{r2c7 r1c8} – r1c6{n7 n8} – c3n8{r1 r5} – b9n8{r9c7 r9c8} – c8n4{r1 r7} – b9n9{r7c8 r7c9} – r1n9{c8 c1} – c1n1{r1 r9} – r7c2{n1 n3} – c3n3{r8 .} ==> r2c7 ≠ 8
;;; now the two paths diverge completely
braid[11]: c9n9{r7 r1} – c7n9{r3 r6} – b6n3{r6c7 r5c7} – c3n3{r5 r8} – c3n2{r8 r2} – c7n7{r5 r2} – r2c6{n2 n8} – r1c6{n8 n7} – r5n7{c6 c8} – r6c8{n7 n8} – c9n8{r5 .} ==> r7c9 ≠ 3
braid[10]: b6n6{r4c8 r4c9} – b6n5{r4c9 r5c9} – c9n3{r5 r8} – r5n7{c8 c6} – r1c6{n7 n8} – c9n8{r5 r2} – r3n8{c8 c2} – r4c2{n8 n3} – c6n3{r8 r7} – c3n3{r8 .} ==> r4c8 ≠ 7
**braid[12]: c7n9{r3 r6} – b9n8{r9c7 r9c8} – r6c8{n9 n7} – r5c8{n8 n1} – r5c7{n8 n3} – b9n3{r9c7 r8c9} – r5n7{c8 c6} – r1c6{n7 n8} – r2c6{n8 n2} – c3n3{r8 r7} – c5n2{r1 r9} – r9n3{c7 .} ==> r3c7 ≠ 8**
whip[8]: r2c7{n7 n6} – r3c7{n6 n9} – r1n9{c8 c1} – c1n1{r1 r9} – c1n6{r9 r3} – r2c1{n6 n4} – r1c3{n4 n8} – r1c6{n8 .} ==> r2c6 ≠ 7
hidden-single-in-a-row ==> r2c7 = 7
whip[4]: r5n7{c8 c6} – r1c6{n7 n8} – r3n8{c4 c2} – b4n8{r6c2 .} ==> r5c8 ≠ 8
braid[9]: b2n9{r2c4 r3c4} – r3c7{n9 n6} – r2n9{c4 c1} – r2n6{c9 c3} – c3n2{r2 r8} – r2n4{c3 c9} – r8n4{c9 c4} – c4n6{r8 r9} – b7n6{r9c1 .} ==> r2c4 ≠ 2
braid[9]: b2n9{r2c4 r3c4} – r3c7{n9 n6} – r2n9{c4 c1} – r2n6{c9 c3} – c3n2{r2 r8} – r2c9{n8 n4} – r8n4{c9 c4} – c4n6{r8 r9} – b7n6{r9c1 .} ==> r2c4 ≠ 8
braid[11]: c4n2{r5 r8} – b6n1{r5c8 r4c8} – b6n6{r4c8 r4c9} – b6n5{r4c9 r5c9} – c9n3{r5 r8} – r8n4{c9 c3} – r8n6{c9 c7} – b9n1{r9c8 r9c7} – c1n1{r9 r1} – r3c7{n6 n9} – r1n9{c9 .} ==> r5c4 ≠ 1

whip[8]: r8n1{c4 c7} – r5n1{c7 c8} – r5n7{c8 c6} – c6n5{r5 r4} – r4c5{n5 n3} – b8n3{r9c5 r8c6} – c9n3{r8 r5} – b6n5{r5c9 .} ==> r7c6 ≠ 1

whip[8]: c1n1{r1 r9} – r7n1{c2 c8} – b9n9{r7c8 r7c9} – r1n9{c9 c8} – r3c7{n9 n6} – c1n6{r3 r2} – r1c3{n6 n8} – r1c9{n8 .} ==> r1c1 ≠ 4

whip[8]: c1n1{r1 r9} – r7n1{c2 c8} – b9n9{r7c8 r7c9} – r1n9{c9 c8} – r3c7{n9 n6} – r2n6{c9 c4} – r9c4{n6 n4} – c8n4{r9 .} ==> r1c1 ≠ 6

whip[10]: b6n6{r4c8 r4c9} – b6n5{r4c9 r5c9} – c9n3{r5 r8} – r8c7{n3 n1} – r9c7{n1 n8} – r5c7{n8 n3} – c3n3{r5 r7} – b7n6{r7c3 r8c3} – b8n6{r8c4 r7c5} – r1n6{c5 .} ==> r9c8 ≠ 6

braid[10]: r3c7{n6 n9} – r1n9{c8 c1} – c1n1{r1 r9} – r9c4{n1 n4} – r9c8{n4 n8} – r3n8{c8 c2} – c1n6{r9 r2} – r1c3{n8 n4} – c8n4{r9 r7} – b7n4{r9c2 .} ==> r3c4 ≠ 6

whip[4]: c4n6{r9 r2} – c1n6{r2 r3} – r3c7{n6 n9} – b2n9{r3c4 .} ==> r9c5 ≠ 6

whip[10]: b6n6{r4c8 r4c9} – b6n5{r4c9 r5c9} – c9n3{r5 r8} – r8c7{n3 n1} – b8n1{r8c6 r9c4} – r9n6{c4 c1} – r8n6{c3 c4} – r8n4{c4 c3} – c3n2{r8 r2} – r2n6{c3 .} ==> r7c8 ≠ 6

whip[11]: r3c7{n6 n9} – r1n9{c8 c1} – c1n1{r1 r9} – c2n1{r9 r1} – r1n2{c2 c5} – r2c6{n2 n8} – r2c9{n8 n4} – r1n4{c8 c3} – r8n4{c3 c4} – b8n2{r8c4 r8c6} – b8n1{r8c6 .} ==> r1c8 ≠ 6

whip[5]: c8n6{r4 r3} – r3c7{n6 n9} – r1n9{c8 c1} – c1n1{r1 r9} – r7n1{c2 .} ==> r4c8 ≠ 1

whip[1]: r4n1{c4 .} ==> r5c6 ≠ 1

whip[5]: r5n1{c7 c8} – r5n7{c8 c6} – r1c6{n7 n8} – c9n8{r1 r2} – c3n8{r2 .} ==> r5c7 ≠ 8

whip[4]: r5n1{c7 c8} – b6n7{r5c8 r6c8} – b6n9{r6c8 r6c7} – c7n8{r6 .} ==> r9c7 ≠ 1

whip[5]: r5c7{n1 n3} – b9n3{r9c7 r8c9} – c3n3{r8 r7} – c6n3{r7 r4} – c6n1{r4 .} ==> r8c7 ≠ 1

singles ==> r5c7 = 1, r5c8 = 7

whip[1]: r8n1{c4 .} ==> r9c4 ≠ 1

braid[7]: r8c7{n6 n3} – c3n2{r8 r2} – r2c6{n2 n8} – r8c9{n6 n4} – r2c9{n8 n6} – b9n6{r8c9 r9c7} – c4n6{r9 .} ==> r8c3 ≠ 6

whip[5]: b7n6{r7c3 r9c1} – r9c4{n6 n4} – r5n4{c4 c1} – c2n4{r6 r1} – c8n4{r1 .} ==> r7c3 ≠ 4

whip[5]: b9n9{r7c9 r7c8} – r7n1{c8 c2} – r7n4{c2 c5} – r9c4{n4 n6} – r8n6{c4 .} ==> r7c9 ≠ 6

whip[2]: r7n6{c3 c5} – c4n6{r9 .} ==> r2c3 ≠ 6

braid[6]: b7n6{r9c1 r7c3} – r8c7{n6 n3} – c3n3{r8 r5} – r6n3{c7 c5} – r9c4{n6 n4} – c5n4{r9 .} ==> r9c7 ≠ 6

whip[1]: b9n6{r8c7 .} ==> r8c4 ≠ 6

whip[8]: b4n7{r6c2 r4c2} – b4n8{r4c2 r5c3} – r5n4{c3 c4} – r9c4{n4 n6} – r7n6{c5 c3} – r1c3{n6 n4} – r2n4{c3 c9} – r8n4{c9 .} ==> r6c2 ≠ 4

braid[6]: b4n4{r5c1 r5c3} – b7n6{r9c1 r7c3} – c3n3{r7 r8} – b7n2{r8c3 r9c2} – r9c5{n4 n3} – b9n3{r9c7 .} ==> r9c1 ≠ 4

whip[8]: r7c3{n3 n6} – r9c1{n6 n1} – b9n1{r9c8 r7c8} – c2n1{r7 r1} – c2n4{r1 r9} – c2n2{r9 r3} – b3n2{r3c8 r1c8} – c8n4{r1 .} ==> r7c2 ≠ 3

whip[3]: r7c9{n4 n9} – r7c8{n9 n1} – r7c2{n1 .} ==> r7c5 ≠ 4

whip[6]: r8c7{n3 n6} – r3c7{n6 n9} – r1n9{c8 c1} – c1n1{r1 r9} – b7n6{r9c1 r7c3} – r7n3{c3 .} ==> r8c6 ≠ 3

whip[7]: r2c6{n8 n2} – c5n2{r3 r9} – c5n4{r9 r6} – b5n7{r6c5 r4c5} – r4c2{n7 n3} – r6c1{n3 n5} – r6c4{n5 .} ==> r4c6 ≠ 8

whip[6]: b4n8{r6c2 r5c3} – c6n8{r5 r2} – r2n2{c6 c3} – r3c2{n2 n3} – r6c2{n3 n7} – r4c2{n7 .} ==> r1c2 ≠ 8

whip[7]: b5n2{r5c4 r5c6} – r8c6{n2 n1} – b5n1{r4c6 r4c4} – b5n8{r4c4 r6c4} – r3c4{n8 n9} – c7n9{r3 r6} – r6c8{n9 .} ==> r5c4 ≠ 5

whip[8]: c4n1{r4 r8} – r8c6{n1 n2} – r2c6{n2 n8} – r3c4{n8 n9} – c7n9{r3 r6} – r6c8{n9 n8} – b5n8{r6c4 r5c4} – b5n2{r5c4 .} ==> r4c4 ≠ 5

whip[7]: c4n6{r9 r2} – b2n9{r2c4 r3c4} – c4n5{r3 r6} – b4n5{r6c1 r5c1} – r5n4{c1 c3} – c1n4{r6 r2} – r2n9{c1 .} ==> r9c4 ≠ 4
singles ==> r9c4 = 6, r2c4 = 9, r7c3 = 6
whip[1]: r7n3{c6 .} ==> r9c5 ≠ 3
whip[5]: c3n3{r5 r8} – r8c7{n3 n6} – r8c9{n6 n4} – c8n4{r9 r1} – r1c3{n4 .} ==> r5c3 ≠ 8
whip[1]: c3n8{r1 .} ==> r3c2 ≠ 8
whip[2]: b4n7{r4c2 r6c2} – c2n8{r6 .} ==> r4c2 ≠ 3
whip[2]: b4n7{r6c2 r4c2} – c2n8{r4 .} ==> r6c2 ≠ 3
whip[3]: b3n2{r1c8 r3c8} – r3n8{c8 c4} – r2c6{n8 .} ==> r1c5 ≠ 2
whip[4]: b8n4{r8c4 r9c5} – c8n4{r9 r1} – b3n2{r1c8 r3c8} – c5n2{r3 .} ==> r8c9 ≠ 4
whip[2]: r8c7{n3 n6} – r8c9{n6 .} ==> r9c7 ≠ 3
naked-single ==> r9c7 = 8
r9n3{c1 .} ==> r8c3 ≠ 3
hidden-single-in-a-column ==> r5c3 = 3
b4n4{r6c1 .} ==> r2c1 ≠ 4
naked-single ==> r2c1 = 6
whip[4]: c4n8{r6 r3} – c4n5{r3 r6} – r4n5{c5 c9} – r5c9{n5 .} ==> r5c6 ≠ 8
whip[1]: c6n8{r1 .} ==> r3c4 ≠ 8
singles to the end

### 5.10.3. An example of non-confluence for the $W_4$ whip resolution theory

As mentioned in the proof of the confluence property for the $B_n$ resolution theories (section 5.5), there is one step in this proof (step b) that would not work for the $W_n$ theories. But this did not prove that the $W_n$ theories do not have the confluence property. The puzzle in Figure 5.4 (Sudogen0_1M #279845) provides the missing proof, for the Sudoku CSP. n = 4 is the smallest n we could find with a counter-example to confluence.

| 9 | 8 | 1 | 7 |   | 3 | 2 | 5 |   |
|---|---|---|---|---|---|---|---|---|
| 7 | 5 | 2 |   |   | 1 | 9 |   |   |
| 3 | 6 | 4 |   | 9 |   |   |   | 8 |
|   | 1 | 7 | 3 |   |   |   | 9 | 2 |
|   | 4 | 3 |   |   | 9 |   | 6 |   |
|   | 9 |   |   |   | 7 |   |   |   |
| 4 |   |   | 1 |   |   |   | 2 | 9 |
|   | 2 | 9 |   |   | 8 |   |   |   |
|   |   |   |   | 9 |   | 5 |   |   |

| 9 | 8 | 1 | 7 | 6 | 3 | 2 | 5 | 4 |
|---|---|---|---|---|---|---|---|---|
| 7 | 5 | 2 | 4 | 8 | 1 | 9 | 3 | 6 |
| 3 | 6 | 4 | 5 | 9 | 2 | 1 | 7 | 8 |
| 8 | 1 | 7 | 3 | 5 | 6 | 4 | 9 | 2 |
| 2 | 4 | 3 | 8 | 1 | 9 | 7 | 6 | 5 |
| 6 | 9 | 5 | 2 | 4 | 7 | 8 | 1 | 3 |
| 4 | 7 | 8 | 1 | 3 | 5 | 6 | 2 | 9 |
| 5 | 2 | 9 | 6 | 7 | 8 | 3 | 4 | 1 |
| 1 | 3 | 6 | 9 | 2 | 4 | 5 | 8 | 7 |

**Figure 5.4.** *An example of non confluence of $W_4$: puzzle Sudogen0_1M #279845*

***** SudoRules 16.2 based on CSP-Rules 1.2, config: W *****
37 givens, 146 candidates, 792 csp-links and 792 links. Initial density = 1.97.
whip[1]: c7n6{r7 .} ==> r9c9 ≠ 6, r8c9 ≠ 6

whip[2]: c1n8{r6 r9} – c8n8{r9 .} ==> r6c3 ≠ 8
whip[1]: c3n8{r9 .} ==> r9c1 ≠ 8
whip[3]: c7n4{r6 r8} – c4n4{r8 r2} – b3n4{r2c9 .} ==> r6c9 ≠ 4
whip[3]: b7n5{r8c1 r7c3} – r7n8{c3 c7} – b9n6{r7c7 .} ==> r8c1 ≠ 6
whip[3]: b8n2{r9c5 r9c6} – r3c6{n2 n5} – r7c6{n5 .} ==> r9c5 ≠ 6
whip[3]: c6n4{r9 r4} – r4c7{n4 n8} – b9n8{r7c7 .} ==> r9c8 ≠ 4
whip[2]: r1n4{c5 c9} – b9n4{r9c9 .} ==> r8c5 ≠ 4

The resolution state $RS_1$ at this point is shown in Figure 5.5.

|   | c1 | c2 | c3 | c4 | c5 | c6 | c7 | c8 | c9 |   |
|---|----|----|----|----|----|----|----|----|----|---|
| r1 | 9 | 8 | 1 | 7 | n4 n6 | 3 | 2 | 5 | n4 n6 | r1 |
| r2 | 7 | 5 | 2 | n4 n6 n8 | n4 n6 n8 | 1 | 9 | n3 n4 | n3 n4 n6 | r2 |
| r3 | 3 | 6 | 4 | n2 n5 | 9 | n2 n5 | n1 n7 | n1 n7 | 8 | r3 |
| r4 | n5 n6 n8 | 1 | 7 | 3 | n4 n5 n6 n8 | n4 n5 n6 | n4 n8 | 9 | 2 | r4 |
| r5 | n2 n5 n8 | 4 | 3 | n2 n5 n8 | n1 n2 n5 n8 | 9 | n1 n7 n8 | 6 | n1 n5 n7 | r5 |
| r6 | n2 n5 n6 n8 | 9 | n5 n6 | n2 n4 n5 n6 n8 | n1 n2 n4 n5 n6 n8 | 7 | n1 n3 n4 n8 | n1 n3 n4 n8 | n1 n3 n5 | r6 |
| r7 | 4 | n3 n7 | n5 n6 n8 | 1 | n3 n5 n6 n7 | n5 n6 | n3 n6 n7 n8 | 2 | 9 | r7 |
| r8 | n1 n5 | 2 | 9 | n4 n5 n6 | n3 n5 n6 n7 | 8 | n1 n3 n4 n6 n7 | n1 n3 n4 n7 | n1 n3 n4 n7 | r8 |
| r9 | n1 n6 | n3 n7 | n6 n8 | 9 | n2 n3 n4 n7 | n2 n4 n6 | 5 | n1 n3 n7 n8 | n1 n3 n4 n7 | r9 |
|   | c1 | c2 | c3 | c4 | c5 | c6 | c7 | c8 | c9 |   |

*Figure 5.5. Resolution state $RS_1$ of puzzle Sudogen0_1M #279845*

After $RS_1$ has been reached, there are (at least) the following two resolution paths.

1) The first path starts with a general whip:

**whip[4]: c6n4{r4 r9} – c6n6{r9 r7} – r8c4{n6 n5} – c5n5{r8 .} ==> r4c6 ≠ 5**

It is worth analysing this whip by adding it a few details:

whip[4]: c6n4{r4 r9$_{(1)}$} – c6n6{r9 r7$_{(2)}$ r4*} – r8c4{n6 n5$_{(3)}$ n4#1} – c5n5{r8 . r4* r5* r6* r7#3} ==> r4c6≠5

The * sign corresponds to z-candidates, the # sign corresponds to t-candidates and the number following this # sign is the number of the right-linking candidate linked to this t-candidate (remember however that, by definition, these z- and t-candidates do not belong to the whip; we display them here for the only sake of illustrating how a whip deals with these additional candidates).

Notice that there is an alternative whip, for the same target, with the same first two cells and the last cell replaced by the slightly simpler: r3n5{c4 . c6*}. Using it instead would not change the sequel.

The end of this first resolution path has nothing noticeable:

whip[2]: b7n5{r7c3 r8c1} – r4n5{c1 .} ==> r7c5 ≠ 5
**whip[4]: r7c6{n5 n6} – r4c6{n6 n4} – r4c7{n4 n8} – r7n8{c7 .} ==> r7c3 ≠ 5**
singles ==> r8c1 = 5, r6c3 = 5, r5c9 = 5, r4c5 = 5, r3c4 = 5, r3c6 = 2, r9c5 = 2, r7c6 = 5, r5c7 = 7, r3c7 = 1, r3c8 = 7, r5c5 = 1, r9c1 = 1
whip[2]: b8n3{r7c5 r8c5} – b8n7{r8c5 .} ==> r7c5 ≠ 6
whip[2]: r7c2{n3 n7} – r7c5{n7 .} ==> r7c7 ≠ 3
whip[2]: b8n3{r8c5 r7c5} – b8n7{r7c5 .} ==> r8c5 ≠ 6
whip[2]: r9n4{c9 c6} – r4n4{c6 .} ==> r8c7 ≠ 4
whip[1]: c7n4{r4 .} ==> r6c8 ≠ 4
whip[3]: r9n4{c9 c6} – b8n6{r9c6 r8c4} – r8c7{n6 .} ==> r9c9 ≠ 3
whip[3]: r8c7{n3 n6} – r7c7{n6 n8} – r9c8{n8 .} ==> r8c9 ≠ 3, r8c8 ≠ 3
whip[3]: r6n2{c4 c1} – r6n6{c1 c5} – r4c6{n6 .} ==> r6c4 ≠ 4
whip[2]: c8n4{r2 r8} – c4n4{r8 .} ==> r2c9 ≠ 4, r2c5 ≠ 4
whip[2]: r1n4{c9 c5} – c4n4{r2 .} ==> r8c9 ≠ 4
**whip[4]: b9n6{r7c7 r8c7} – r8c4{n6 n4} – c6n4{r9 r4} – r4c7{n4 .} ==> r7c7 ≠ 8**
singles to the end

Now, if we activate braids and we re-start with our usual "simplest first" strategy, we get exactly the same path (there appears no non-whip braid). Thanks to the confluence property of $B_4$, we do not have to consider any other resolution path to claim that the correct B rating is B = 4. As W(P) ≤ B(P) for any P and we have found a resolution path for P with whips of lengths no more than 4, we can also claim that W(P) = 4.

2) Let us now consider what would have happened if we had followed an alternative resolution path. In state $RS_1$, before using the first whip[4] above, we could have chosen a whole sequence of simpler whips – "simpler" in the sense that they are special subtypes of whips, not in the sense of being shorter (these subtypes were introduced in *HLS*, but it is not necessary here to know their precise definitions, they are whips anyway, with the lengths indicated in square brackets):

***** SudoRules 13.7wter2 *****
;;; same path up to resolution state$RS_1$
xyzt-chain[4]: r7c6{n6 n5} – r3c6{n5 n2} – r9c6{n2 n4} – r8c4{n4 n6} ==> r8c5 ≠ 6, r7c5 ≠ 6

nrc-chain[4]:  b6n7{r5c7 r5c9} – b6n5{r5c9 r6c9} – c3n5{r6 r7} – r7n8{c3 c7} ==> r7c7 ≠ 7, r5c7 ≠ 8
naked-pairs-in-a-column c7{r3 r5}{n1 n7} ==> r8c7 ≠ 7, r8c7 ≠ 1, r6c7 ≠ 1
;;; Resolution state RS$_2$
**nrc-chain[4]: r9c3{n6 n8} – b9n8{r9c8 r7c7} – r4c7{n8 n4} – c6n4{r4 r9} ==> r9c6 ≠ 6**
;;; Resolution state RS$_3$
interaction row r9 with block b7 ==> r7c3 ≠ 6
**nrct-chain[5]:   c6n4{r4 r9} – c6n2{r9 r3} – r3n5{c6 c4} – r8c4{n5 n6} – r7c6{n6 n5} ==> r4c6≠5**
nrc-chain[2]: r4n5{c5 c1} – b7n5{r8c1 r7c3} ==> r7c5 ≠ 5
naked-pairs-in-a-row: r7{c2 c5}{n3 n7} ==> r7c7 ≠ 3
xy-chain[3]: r7c7{n6 n8} – r4c7{n8 n4} – r4c6{n4 n6} ==> r7c6 ≠ 6
singles to the end

Until we reach resolution state RS$_2$, the whip[4] of the first path is still available; but if we apply the nrc-chain[4] rule before this whip[4], it deletes the left-linking candidate n6r9c6 for its second CSP variable. Then, in the resulting state RS$_3$, there remains no whip[4]; the simplest whip available is a slightly longer nrct-chain[5]; it makes the same r4c6 ≠ 5 elimination.

Conclusion: if we considered only this second resolution path, we would find, erroneously, that the W rating of this puzzle is 5. This example is thus not only a clear case of non-confluence for whip theories, it is also a case in which this non-confluence leads to a bad evaluation of the W rating if we do not try all the paths. This is a very rare case.

Final remark: if we allow braids, even after the nrc-chain[4] is applied, there is a replacement braid for the missing whip[4] (and it is as provided in section 5.5.1 by the general proof of confluence for braid resolution theories):
**braid[4]:  c6n4{r4 r9} – c6n6{r4 r7} – r8c4{n6 n5 n4#1} – c5n5{r8 . r4* r5* r6* r7#3} ==> r4c6 ≠ 5**

The z-candidate n6r4c6 in cell 2 of the whip[4] is now used as a left-linking candidate in the braid, in which it is linked to the target.

### 5.10.4. A puzzle P with a whip of length 31 and B(P) = 19 [and gW(P) = 12]

What is the largest whip one can find? This is a very difficult question. The largest W rating we could obtain with random generators is 16 (and we could find only one puzzle with W=16 in more than 10,000,000). In Figure 5.5 of *CRT*, we gave an example of a puzzle (of unknown origin) with a whip of length 24. Since then, Mauricio, on the Player's Forum, has found one (Figure 5.6 below) with length 31. It does not prove that W(P) = 31, but after trying several resolution paths, we found none without a whip of length 31. Most interestingly, the B rating is B(P) = 19 only, suggesting that, in extremely rare cases, the gap between the W and B ratings, even when they are both finite, can be very large. Moreover, in chapter 7, it will be shown that the gW rating is only 12.

|   |   |   |   | 1 |   |   | 2 |   |
|---|---|---|---|---|---|---|---|---|
|   |   |   | 3 |   |   | 4 |   |   |
|   |   | 5 | 2 |   |   | 1 |   |   |
|   |   | 3 | 6 |   |   |   | 1 |   |
|   | 2 |   |   | 7 |   |   |   | 8 |
| 9 |   |   |   | 5 | 7 |   |   |   |
|   |   | 9 |   |   | 7 |   |   |   |
|   | 8 |   | 9 |   |   |   |   | 4 |
| 3 |   |   |   | 4 |   |   | 8 |   |

| 6 | 9 | 4 | 7 | 5 | 1 | 8 | 3 | 2 |
|---|---|---|---|---|---|---|---|---|
| 1 | 7 | 2 | 8 | 3 | 9 | 5 | 4 | 6 |
| 8 | 3 | 5 | 2 | 6 | 4 | 1 | 7 | 9 |
| 7 | 4 | 3 | 6 | 9 | 8 | 2 | 1 | 5 |
| 5 | 2 | 6 | 1 | 7 | 3 | 4 | 9 | 8 |
| 9 | 1 | 8 | 4 | 2 | 5 | 7 | 6 | 3 |
| 4 | 5 | 9 | 3 | 8 | 7 | 6 | 2 | 1 |
| 2 | 8 | 7 | 9 | 1 | 6 | 3 | 5 | 4 |
| 3 | 6 | 1 | 4 | 4 | 2 | 9 | 8 | 7 |

***Figure 5.6.*** *A puzzle P with W(P) = 31*

The path with whips provides a whip of length 31.

***** SudoRules 16.2 based on CSP-Rules 1.2, config: W *****
24 givens, 220 candidates, 1433 csp-links and 1433 links. Initial density = 1.49
whip[11]: c8n9{r1 r5} – r4c9{n9 n5} – r2n5{c9 c4} – b2n7{r2c4 r1c4} – b2n4{r1c4 r3c6} –
c6n9{r3 r4} – r5c6{n9 n3} – c4n3{r5 r7} – b8n8{r7c4 r7c5} – r4n8{c5 c1} – r3n8{c1 .} ==> r2c7 ≠ 9
whip[11]: r8n1{c1 c5} – r9c4{n1 n5} – c2n5{r9 r4} – b4n7{r4c2 r4c1} – b4n8{r4c1 r6c3} –
r6n1{c3 c4} – r6c5{n1 n2} – r4n2{c6 c7} – b6n4{r4c7 r5c7} – c4n4{r5 r1} – c3n4{r1 .} ==> r7c2 ≠ 1
whip[12]: b9n1{r7c9 r9c9} – r9c4{n1 n5} – c2n5{r9 r4} – r4c9{n5 n9} – b5n9{r4c6 r5c6} –
r2n9{c6 c2} – c2n1{r2 r6} – b5n1{r6c5 r5c4} – b5n3{r5c4 r6c4} – r6n4{c4 c3} – r5c3{n4 n6} –
r5c1{n6 .} ==> r7c9 ≠ 5
whip[12]: b9n9{r9c7 r9c9} – r4c9{n9 n5} – r2n5{c9 c4} – r9c4{n5 n1} – c5n1{r8 r6} – c2n1{r6 r2}
– r2n9{c2 c6} – b5n9{r5c6 r4c5} – b5n2{r4c5 r4c6} – c6n8{r4 r3} – b2n4{r3c6 r1c4} – b2n7{r1c4 .}
==> r9c7 ≠ 5
whip[14]: b3n8{r1c7 r2c7} – r2n5{c7 c4} – b2n7{r2c4 r1c4} – b2n4{r1c4 r3c6} – c6n8{r3 r4} –
c4n8{r6 r7} – b8n3{r7c4 r8c6} – r5c6{n3 n9} – r4c5{n9 n2} – r6c5{n2 n1} – c4n1{r6 r9} –
c2n1{r9 r2} – r2n9{c2 c9} – c8n9{r3 .} ==> r1c7 ≠ 5
whip[14]: b7n4{r7c1 r7c2} – c2n5{r7 r4} – b4n7{r4c2 r4c1} – b4n8{r4c1 r6c3} – r6n4{c3 c4} –
r4n4{c6 c7} – b6n2{r4c7 r6c8} – r6c5{n2 n1} – r5c4{n1 n3} – r5c6{n3 n9} – r4n9{c6 c9} –
r2n9{c9 c2} – c2n1{r2 r9} – r8n1{c1 .} ==> r7c1 ≠ 5
whip[17]: b2n4{r3c6 r1c4} – b2n7{r1c4 r2c4} – b2n5{r2c4 r1c5} – c5n9{r1 r4} – r4c9{n9 n5} –
b3n5{r2c9 r2c7} – r5n5{c7 c1} – r8n5{c1 c8} – b9n7{r8c8 r9c9} – c9n9{r9 r2} – b1n9{r2c2 r1c2} –
b1n3{r1c2 r3c2} – b1n4{r3c2 r3c1} – r3n8{c1 c5} – c6n8{r2 r4} – r4c1{n8 n7} – c2n7{r4 .} ==>
r3c6 ≠ 9
whip[17]: b4n8{r6c3 r4c1} – b4n7{r4c1 r4c2} – b4n5{r4c2 r5c1} – r5n1{c1 c4} – r9c4{n1 n5} –
b7n5{r9c2 r7c2} – c5n5{r7 r1} – c8n5{r1 r8} – b9n7{r8c8 r9c9} – r9n1{c9 c2} – c1n1{r8 r2} –
b1n2{r2c1 r2c3} – b1n8{r2c3 r1c3} – c3n4{r1 r5} – b6n4{r5c7 r4c7} – c7n5{r4 r2} – b3n8{r2c7 .}
==> r6c3 ≠ 1
**whip[31]: b3n8{r1c7 r2c7} – c1n8{r2 r4} – c6n8{r4 r3} – b2n4{r3c6 r1c4} – b2n7{r1c4 r2c4}
– b2n5{r2c4 r1c5} – b3n5{r1c8 r2c9} – r4c9{n5 n9} – r4c5{n9 n2} – r4c6{n2 n4} –**

**b5n9{r4c6 r5c6} – c6n3{r5 r8} – b8n2{r8c6 r9c6} – c6n6{r9 r2} – b2n9{r2c6 r3c5} – c8n9{r3 r1} – c7n9{r1 r9} – r4c7{n9 n5} – b4n5{r4c2 r5c1} – r8n5{c1 c8} – c8n7{r8 r3} – c9n7{r2 r9} – c3n7{r9 r8} – c3n2{r8 r2} – r2c1{n2 n1} – r8n1{c1 c5} – b8n6{r8c5 r7c5} – b9n6{r7c7 r8c7} – b3n6{r2c7 r3c9} – c1n6{r3 r1} – c1n7{r1 .} ==> r1c3 ≠ 8**

whip[6]: b4n8{r6c3 r4c1} – b1n8{r3c1 r2c3} – c6n8{r2 r3} – b2n4{r3c6 r1c4} – r6n4{c4 c2} – c3n4{r6 .} ==> r6c3 ≠ 6

whip[7]: r1n4{c3 c4} – r6n4{c4 c3} – c3n8{r6 r2} – b1n2{r2c3 r2c1} – b1n1{r2c1 r2c2} – b1n9{r2c2 r1c2} – b1n3{r1c2 .} ==> r3c2 ≠ 4

whip[6]: r3n4{c6 c1} – r3n8{c1 c5} – c6n8{r3 r4} – c6n4{r4 r5} – c3n4{r5 r6} – b4n8{r6c3 .} ==> r3c6 ≠ 6

whip[10]: b4n8{r6c3 r4c1} – r3n8{c1 c6} – r3n4{c6 c1} – b7n4{r7c1 r7c2} – r1n4{c2 c4} – b2n7{r1c4 r2c4} – c4n8{r2 r7} – c4n5{r7 r9} – c2n5{r9 r4} – b4n7{r4c2 .} ==> r6c5 ≠ 8

whip[6]: r6c5{n1 n2} – b6n2{r6c8 r4c7} – b6n4{r4c7 r5c7} – c4n4{r5 r1} – c3n4{r1 r6} – r6n8{c3 .} ==> r6c4 ≠ 1

whip[2]: r8n1{c3 c5} – r6n1{c5 .} ==> r9c2 ≠ 1

whip[6]: r4c9{n5 n9} – b5n9{r4c6 r5c6} – r2n9{c6 c2} – c2n1{r2 r6} – b5n1{r6c5 r5c4} – r9c4{n1 .} ==> r9c9 ≠ 5

whip[6]: r4c9{n5 n9} – b5n9{r4c6 r5c6} – r2n9{c6 c2} – c2n1{r2 r6} – r6c5{n1 n2} – b6n2{r6c8 .} ==> r4c7 ≠ 5

whip[6]: b8n8{r7c5 r7c4} – c4n1{r7 r5} – c4n3{r5 r6} – r6n8{c4 c3} – r6n4{c3 c2} – b4n1{r6c2 .} ==> r7c5 ≠ 1

whip[7]: b5n9{r4c6 r5c6} – r2n9{c6 c2} – c2n1{r2 r6} – b5n1{r6c5 r5c4} – b5n3{r5c4 r6c4} – r6n8{c4 c3} – r6n4{c3 .} ==> r4c9 ≠ 9

singles ==> r4c9 = 5, r5c1 = 5

biv-chain[2]: b4n1{r5c3 r6c2} – b4n6{r6c2 r5c3} ==> r5c3 ≠ 4

biv-chain[2]: r5n1{c3 c4} – c5n1{r6 r8} ==> r8c3 ≠ 1

whip[2]: b4n1{r6c2 r5c3} – b4n6{r5c3 .} ==> r6c2 ≠ 4

whip[2]: r6n8{c4 c3} – r6n4{c3 .} ==> r6c4 ≠ 3

whip[1]: r6n3{c9 .} ==> r5c8 ≠ 3, r5c7 ≠ 3

biv-chain[3]: b2n7{r1c4 r2c4} – r2n5{c4 c7} – c7n8{r2 r1} ==> r1c4 ≠ 8

whip[4]: r9n5{c2 c4} – r2n5{c4 c7} – r8n5{c7 c8} – b9n7{r8c8 .} ==> r9c2 ≠ 7

biv-chain[3]: r8n1{c1 c5} – r9c4{n1 n5} – r9c2{n5 n6} ==> r8c1 ≠ 6

whip[4]: c2n1{r2 r6} – b5n1{r6c5 r5c4} – r9c4{n1 n5} – r9c2{n5 .} ==> r2c2 ≠ 6

whip[4]: r9c4{n1 n5} – r9c2{n5 n6} – b4n6{r6c2 r5c3} – r5n1{c3 .} ==> r7c4 ≠ 1

biv-chain[3]: b9n1{r7c9 r9c9} – c4n1{r9 r5} – c4n3{r5 r7} ==> r7c9 ≠ 3

whip[4]: r7n1{c9 c1} – b7n4{r7c1 r7c2} – b7n5{r7c2 r9c2} – r9c4{n5 .} ==> r9c9 ≠ 1

hidden-single-in-a-block ==> r7c9 = 1

biv-chain[2]: r5n1{c3 c4} – r9n1{c4 c3} ==> r2c3 ≠ 1

whip[3]: b7n7{r9c3 r8c1} – c1n1{r8 r2} – b1n2{r2c1 .} ==> r2c3 ≠ 7

whip[4]: b4n6{r6c2 r5c3} – c3n1{r5 r9} – r9c4{n1 n5} – r9c2{n5 .} ==> r3c2 ≠ 6, r1c2 ≠ 6

whip[4]: b4n6{r6c2 r5c3} – c3n1{r5 r9} – r9c4{n1 n5} – b7n5{r9c2 .} ==> r7c2 ≠ 6

whip[4]: r6c4{n4 n8} – c3n8{r6 r2} – c6n8{r2 r3} – b2n4{r3c6 .} ==> r5c4 ≠ 4

biv-chain[2]: c3n4{r1 r6} – c4n4{r6 r1} ==> r1c2 ≠ 4, r1c1 ≠ 4

biv-chain[5]: r9c6{n2 n6} – c2n6{r9 r6} – r6n1{c2 c5} – r5c4{n1 n3} – b8n3{r7c4 r8c6} ==> r8c6 ≠ 2

biv-chain[5]: b7n1{r8c1 r9c3} – r9c4{n1 n5} – b7n5{r9c2 r7c2} – c2n4{r7 r4} – r4n7{c2 c1} ==> r8c1 ≠ 7

whip[1]: b7n7{r9c3 .} ==> r1c3 ≠ 7
whip[3]: r6n2{c8 c5} – c5n1{r6 r8} – r8c1{n1 .} ==> r8c8 ≠ 2
whip[5]: r4c2{n7 n4} – r6n4{c3 c4} – r1c4{n4 n5} – r9n5{c4 c2} – r7c2{n5 .} ==> r1c2 ≠ 7
whip[5]: b8n3{r8c6 r7c4} – r5c4{n3 n1} – r9c4{n1 n5} – r2n5{c4 c7} – r8n5{c7 .} ==> r8c8 ≠ 3
whip[5]: r5c8{n9 n6} – r6n6{c9 c2} – c2n1{r6 r2} – r2n9{c2 c9} – c8n9{r3 .} ==> r5c6 ≠ 9
whip[1]: r5n9{c8 .} ==> r4c7 ≠ 9
biv-chain[3]: b8n3{r7c4 r8c6} – r5c6{n3 n4} – r6c4{n4 n8} ==> r7c4 ≠ 8
hidden-single-in-a-block ==> r7c5 = 8
biv-chain[2]: b4n8{r4c1 r6c3} – c4n8{r6 r2} ==> r2c1 ≠ 8
biv-chain[2]: c3n8{r2 r6} – c4n8{r6 r2} ==> r2c7 ≠ 8
hidden-single-in-a-block ==> r1c7 = 8
whip[1]: c7n3{r8 .} ==> r7c8 ≠ 3
biv-chain[2]: c3n8{r2 r6} – c4n8{r6 r2} ==> r2c6 ≠ 8
biv-chain[2]: r3c5{n6 n9} – r2c6{n9 n6} ==> r1c5 ≠ 6
biv-chain[2]: r3n4{c1 c6} – r3n8{c6 c1} ==> r3c1 ≠ 7, r3c1 ≠ 6
whip[2]: r2n5{c7 c4} – b8n5{r9c4 .} ==> r8c7 ≠ 5
whip[2]: r2c6{n9 n6} – r3c5{n6 .} ==> r1c5 ≠ 9
singles ==> r1c5 = 5, r2c7 = 5, r8c8 = 5, r9c9 = 7, r8c3 = 7, r9c7 = 9, r5c8 = 9, r1c2 = 9, r3c2 = 3, r1c8 = 3, r6c9 = 3, r3c8 = 7
whip[1]: r1n6{c1 .} ==> r2c3 ≠ 6, r2c1 ≠ 6
whip[2]: c2n6{r9 r6} – c8n6{r6 .} ==> r7c1 ≠ 6 ; singles to the end

Radically different from the start, the path with braids shows that B(P) = 19.

***** SudoRules 16.2 based on CSP-Rules 1.2, config: B *****
24 givens, 220 candidates, 1433 csp-links and 1433 links. Initial density = 1.49
braid[8]: r9c4{n5 n1} – b9n1{r9c9 r7c9} – c5n1{r7 r6} – c2n1{r6 r2} – r4c9{n5 n9} – b5n9{r4c5 r5c6} – r2n9{c2 c7} – b9n9{r9c9 .} ==> r9c9 ≠ 5
braid[10]: c8n9{r1 r5} – r4c9{n9 n5} – b3n8{r1c7 r2c7} – r2n5{c7 c4} – b2n7{r2c4 r1c4} – b2n4{r1c4 r3c6} – c6n8{r2 r4} – c4n8{r1 r7} – r5c6{n4 n3} – b8n3{r8c6 .} ==> r1c7 ≠ 9
braid[10]: r8n1{c1 c5} – r9c4{n1 n5} – b7n4{r7c1 r7c2} – c2n5{r7 r4} – b4n7{r4c2 r4c1} – b4n8{r4c1 r6c3} – r6n4{c2 c4} – r4n4{c1 c7} – b6n2{r4c7 r6c8} – r6c5{n8 .} ==> r7c1 ≠ 1
whip[11]: c8n9{r1 r5} – r4c9{n9 n5} – r2n5{c9 c4} – b2n7{r2c4 r1c4} – b2n4{r1c4 r3c6} – c6n9{r3 r4} – r5c6{n9 n3} – c4n3{r5 r7} – b8n8{r7c4 r7c5} – r4n8{c5 c1} – r3n8{c1 .} ==> r2c7 ≠ 9
whip[11]: r8n1{c1 c5} – r9c4{n1 n5} – c2n5{r9 r4} – b4n7{r4c2 r4c1} – b4n8{r4c1 r6c3} – r6n1{c3 c4} – r6c5{n1 n2} – r4n2{c6 c7} – b6n4{r4c7 r5c7} – c4n4{r5 r1} – c3n4{r1 .} ==> r7c2 ≠ 1
braid[11]: b3n8{r1c7 r2c7} – c6n8{r2 r4} – b2n7{r1c4 r2c4} – r2n5{c4 c9} – r4c9{n5 n9} – r4c5{n8 n2} – b5n9{r4c5 r5c6} – r2n9{c6 c2} – r6c5{n2 n1} – c2n1{r2 r9} – r8n1{c5 .} ==> r1c4 ≠ 8
whip[11]: c6n3{r8 r5} – c4n3{r5 r7} – c7n3{r7 r1} – b3n8{r1c7 r2c7} – c4n8{r2 r6} – c6n8{r4 r3} – b2n4{r3c6 r1c4} – b2n7{r1c4 r2c4} – r2n5{c4 c9} – r4c9{n5 n9} – r5n9{c8 .} ==> r8c8 ≠ 3
braid[11]: b7n4{r7c1 r7c2} – r6n4{c2 c4} – b4n7{r4c1 r4c2} – c2n5{r4 r9} – r9c4{n5 n1} – b5n1{r5c4 r6c5} – r5c4{n1 n3} – r5c6{n3 n9} – c2n1{r6 r2} – r2n9{c2 c9} – c8n9{r5 .} ==> r4c1 ≠ 4
whip[11]: c7n2{r8 r4} – b5n2{r4c6 r6c5} – r7n2{c5 c1} – b7n4{r7c1 r7c2} – r4n4{c2 c6} – r6n4{c4 c3} – b4n8{r6c3 r4c1} – b4n7{r4c1 r4c2} – c2n5{r4 r9} – r9c4{n5 n1} – b5n1{r5c4 .} ==> r8c8 ≠ 2

braid[11]: r9c4{n5 n1} – c5n1{r8 r6} – c2n1{r6 r2} – b9n9{r9c7 r9c9} – r2n9{c9 c6} – b5n9{r5c6 r4c5} – b5n2{r6c5 r4c6} – r4n8{c6 c1} – r9n2{c7 c3} – b4n7{r4c1 r4c2} – r9n7{c9 .} ==> r9c7 ≠ 5

braid[11]: r4c9{n5 n9} – b5n9{r4c6 r5c6} – r2n9{c6 c2} – b9n1{r7c9 r9c9} – c2n1{r9 r6} – b5n1{r6c5 r5c4} – b5n3{r5c6 r6c4} – c4n4{r6 r1} – c9n3{r7 r3} – b2n7{r1c4 r2c4} – c9n7{r9 .} ==> r7c9 ≠ 5

braid[12]: b3n8{r1c7 r2c7} – r2n5{c7 c4} – b2n7{r2c4 r1c4} – b2n4{r1c4 r3c6} – c6n8{r3 r4} – r9c4{n5 n1} – b5n1{r5c4 r6c5} – b5n2{r4c6 r4c5} – b5n9{r4c5 r5c6} – c2n1{r6 r2} – r2n9{c2 c9} – c8n9{r5 .} ==> r1c7 ≠ 5

braid[12]: b7n4{r7c1 r7c2} – c2n5{r7 r4} – b4n7{r4c2 r4c1} – b4n8{r4c1 r6c3} – r6n4{c3 c4} – r4c9{n5 n9} – b5n9{r4c5 r5c6} – b5n3{r5c6 r5c4} – b5n1{r5c4 r6c5} – r2n9{c6 c2} – c2n1{r2 r9} – r8n1{c5 .} ==> r7c1 ≠ 5

whip[16]: b2n4{r3c6 r1c4} – b2n7{r1c4 r2c4} – b2n5{r2c4 r1c5} – c5n9{r1 r4} – r4c9{n9 n5} – b3n5{r2c9 r2c7} – r5n5{c7 c1} – r8n5{c1 c8} – b9n7{r8c8 r9c9} – c9n9{r9 r2} – b1n9{r2c2 r1c2} – b1n3{r1c2 r3c2} – c2n7{r3 r4} – r4c1{n7 n8} – r3n8{c1 c5} – c6n8{r3 .} ==> r3c6 ≠ 9

whip[16]: b4n8{r6c3 r4c1} – b4n7{r4c1 r4c2} – b4n5{r4c2 r5c1} – r5n1{c1 c4} – r9c4{n1 n5} – b7n5{r9c2 r7c2} – c5n5{r7 r1} – c8n5{r1 r8} – b9n7{r8c8 r9c9} – r9n1{c9 c2} – c1n1{r8 r2} – b1n2{r2c1 r2c3} – r2n7{c3 c4} – r1c4{n7 n4} – c3n4{r1 r5} – r6n4{c3 .} ==> r6c3 ≠ 1

**braid[19]: b1n3{r3c2 r1c2} – r1n4{c2 c4} – b2n7{r1c4 r2c4} – b2n5{r2c4 r1c5} – r1n9{c5 c8} – r3n9{c9 c5} – r6n4{c4 c3} – b4n8{r6c3 r4c1} – r3n8{c5 c6} – b2n6{r3c6 r2c6} – r4c5{n9 n2} – r6n2{c5 c8} – r9c6{n6 n2} – r8c6{n6 n3} – c1n4{r1 r7} – r7n2{c1 c7} – c7n3{r1 r5} – r5n4{c1 c6} – r5n9{c8 .} ==> r3c2 ≠ 4**

whip[6]: r3n4{c6 c1} – r3n8{c1 c5} – c6n8{r3 r4} – c6n4{r4 r5} – c3n4{r5 r6} – b4n8{r6c3 .} ==> r3c6 ≠ 6

braid[8]: b4n8{r6c3 r4c1} – r3n8{c1 c6} – r3n4{c6 c1} – b4n7{r4c1 r4c2} – b7n4{r7c1 r7c2} – c2n5{r7 r9} – r9c4{n5 n1} – c5n1{r8 .} ==> r6c5 ≠ 8

whip[6]: r6n8{c3 c4} – r6n4{c4 c2} – c3n4{r6 r1} – c3n8{r1 r2} – c6n8{r2 r3} – b2n4{r3c6 .} ==> r6c3 ≠ 6

whip[6]: r6c5{n1 n2} – b6n2{r6c8 r4c7} – b6n4{r4c7 r5c7} – c4n4{r5 r1} – c3n4{r1 r6} – r6n8{c3 .} ==> r6c4 ≠ 1

whip[2]: r8n1{c3 c5} – r6n1{c5 .} ==> r9c2 ≠ 1

whip[6]: r4c9{n5 n9} – b5n9{r4c6 r5c6} – r2n9{c6 c2} – c2n1{r2 r6} – r6c5{n1 n2} – b6n2{r6c8 .} ==> r4c7 ≠ 5

whip[6]: b8n8{r7c5 r7c4} – c4n1{r7 r5} – c4n3{r5 r6} – r6n8{c4 c3} – r6n4{c3 c2} – b4n1{r6c2 .} ==> r7c5 ≠ 1

whip[3]: b7n1{r9c3 r8c1} – r5n1{c1 c4} – b8n1{r9c4 .} ==> r2c3 ≠ 1

braid[6]: b5n9{r4c6 r5c6} – b9n9{r9c7 r9c9} – r2n9{c6 c2} – c2n1{r2 r6} – r6c5{n1 n2} – b6n2{r6c8 .} ==> r4c7 ≠ 9

whip[7]: b6n9{r5c8 r4c9} – r2n9{c9 c2} – c2n1{r2 r6} – b5n1{r6c5 r5c4} – b5n3{r5c4 r6c4} – r6n8{c4 c3} – r6n4{c3 .} ==> r5c6 ≠ 9

whip[1]: r5n9{c8 .} ==> r4c9 ≠ 9

singles ==> r4c9 = 5, r5c1 = 5

biv-chain[2]: b4n1{r5c3 r6c2} – b4n6{r6c2 r5c3} ==> r5c3 ≠ 4

biv-chain[2]: r5n1{c3 c4} – c5n1{r6 r8} ==> r8c3 ≠ 1

whip[2]: b4n1{r6c2 r5c3} – b4n6{r5c3 .} ==> r6c2 ≠ 4

whip[2]: r6n8{c4 c3} – r6n4{c3 .} ==> r6c4 ≠ 3

whip[1]: r6n3{c9 .} ==> r5c8 ≠ 3, r5c7 ≠ 3

biv-chain[3]: b8n3{r7c4 r8c6} – r5c6{n3 n4} – r6c4{n4 n8} ==> r7c4 ≠ 8
hidden-single-in-a-block ==> r7c5 = 8
biv-chain[2]: b4n8{r4c1 r6c3} – c4n8{r6 r2} ==> r2c1 ≠ 8
biv-chain[2]: r3n8{c1 c6} – r4n8{c6 c1} ==> r1c1 ≠ 8
biv-chain[2]: r3n4{c1 c6} – r3n8{c6 c1} ==> r3c1 ≠ 7, r3c1 ≠ 6
whip[2]: r2n5{c7 c4} – b8n5{r9c4 .} ==> r8c7 ≠ 5
whip[2]: r3n8{c6 c1} – r4n8{c1 .} ==> r2c6 ≠ 8
biv-chain[2]: r3c5{n6 n9} – r2c6{n9 n6} ==> r1c5 ≠ 6
whip[2]: r3c5{n9 n6} – r2c6{n6 .} ==> r1c5 ≠ 9
singles ==> r1c5 = 5, r2c7 = 5, r1c7 = 8, r8c8 = 5, r9c9 = 7, r7c9 = 1, r9c7 = 9, r5c8 = 9, r1c2 = 9, r3c2 = 3, r1c8 = 3, r6c9 = 3, r3c8 = 7
whip[1]: r1n6{c1 .} ==> r2c1 ≠ 6, r2c2 ≠ 6, r2c3 ≠ 6
whip[2]: c2n6{r9 r6} – c8n6{r6 .} ==> r7c1 ≠ 6
biv-chain[3]: c1n6{r1 r8} – c1n1{r8 r2} – r2c2{n1 n7} ==> r1c1 ≠ 7
biv-chain[3]: c1n6{r1 r8} – b7n1{r8c1 r9c3} – r5c3{n1 n6} ==> r1c3 ≠ 6
hidden-single-in-a-block ==> r1c1 = 6
whip[2]: r6n4{c4 c3} – r1n4{c3 .} ==> r5c4 ≠ 4
biv-chain[3]: r4c1{n7 n8} – r6c3{n8 n4} – r1c3{n4 n7} ==> r2c1 ≠ 7
biv-chain[3]: c2n7{r2 r4} – c1n7{r4 r8} – c1n1{r8 r2} ==> r2c2 ≠ 1
singles to the end

Considering such exceptional puzzles, it appears that the notion of simplicity of a resolution path can only be (very) relative.

### 5.10.5. A braid[3] that is not a whip[3]; also a proof that a puzzle has no solution

We shall use the puzzle in Figure 5.7 for two different purposes at the same time: giving an example of a braid[3] that is not a whip[3] and showing how our resolution rules can be used to prove that an instance has no solution (the steps of such a proof are exactly the same as those used to find a solution) .

|   | 3 |   |   | 6 |   |   |   |   |
|---|---|---|---|---|---|---|---|---|
|   | 5 | 1 |   |   |   |   |   |   |
| 6 |   |   | 2 | 3 | 4 |   |   |   |
|   | 7 |   |   |   |   | 5 |   |   |
|   |   | 9 |   |   |   |   |   | 7 |
|   | 6 | 4 |   | 3 |   | 8 |   |   |
|   | 4 |   |   |   |   |   | 9 | 1 |
|   |   | 2 |   |   | 8 | 3 |   |   |
|   |   |   |   |   |   |   |   |   |

*Figure 5.7. A puzzle P with a non-whip braid[3]*

***** SudoRules 16.2 based on CSP-Rules 1.2, config: B *****
22 givens, 242 candidates 1692 csp-links and 1692 links. Initial density = 1.45

whip[1]: r7n8{c1 .} ==> r9c3 ≠ 8, r9c2 ≠ 8, r9c1 ≠ 8
whip[2]: r3n9{c2 c9} – r6n9{c9 .} ==> r2c1 ≠ 9, r1c1 ≠ 9
whip[2]: b6n3{r4c9 r5c8} – b6n4{r5c8 .} ==> r4c9 ≠ 2
whip[2]: b6n4{r4c9 r5c8} – b6n3{r5c8 .} ==> r4c9 ≠ 6, r4c9 ≠ 9
whip[2]: b6n3{r5c8 r4c9} – b6n4{r4c9 .} ==> r5c8 ≠ 2, r5c8 ≠ 1
whip[2]: b6n4{r5c8 r4c9} – b6n3{r4c9 .} ==> r5c8 ≠ 6
whip[1]: b6n6{r4c7 .} ==> r9c7 ≠ 6, r7c7 ≠ 6, r2c7 ≠ 6
whip[2]: b3n3{r2c8 r2c9} – b3n6{r2c9 .} ==> r2c8 ≠ 2, r2c8 ≠ 7, r2c8 ≠ 8
whip[2]: b3n3{r2c9 r2c8} – b3n6{r2c8 .} ==> r2c9 ≠ 2
whip[2]: c2n2{r5 r1} – b3n2{r1c7 .} ==> r5c7 ≠ 2
whip[2]: b3n3{r2c9 r2c8} – b3n6{r2c8 .} ==> r2c9 ≠ 8, r2c9 ≠ 9
whip[2]: b6n2{r6c9 r4c7} – r2n2{c7 .} ==> r6c1 ≠ 2
whip[3]: b5n8{r4c4 r5c5} – r2n8{c5 c1} – b7n8{r7c1 .} ==> r4c3 ≠ 8
whip[3]: b4n8{r5c3 r4c1} – b7n8{r7c1 r7c3} – r2n8{c3 .} ==> r5c5 ≠ 8
whip[1]: b5n8{r4c4 .} ==> r4c1 ≠ 8
;;; Resolution state RS$_1$, displayed in Figure 5.8.

|    | c1 | c2 | c3 | c4 | c5 | c6 | c7 | c8 | c9 |    |
|----|----|----|----|----|----|----|----|----|----|----|
| r1 | n1 n2 n4 n7 n8 | n1 n2 n8 n9 | n3 | n4 n5 n7 n8 | n4 n5 n7 n8 n9 | n6 | n1 n2 n5 n7 n9 | n1 n2 n5 n7 n8 | n2 n5 n8 n9 | r1 |
| r2 | n2 n4 n7 n8 | n5 | n1 | n4 n7 n8 n9 | n4 n7 n9 | n2 n7 n9 | n3 n6 | n3 n6 |    | r2 |
| r3 | n6 | n1 n8 n9 | n1 n7 n8 n9 | n5 n7 n8 | n2 | n3 | n4 | n1 n7 n8 | n5 n8 n9 | r3 |
| r4 | n1 n2 n3 n9 | n7 | n1 n9 | n2 n4 n6 n8 | n1 n4 n6 n8 | n1 n2 n4 n9 | n1 n2 n6 n9 | n5 | n3 n4 | r4 |
| r5 | n1 n2 n3 n5 n8 | **n1** n2 n3 n8 | n1 n5 n8 | n9 | n1 n4 n6 n8 | n1 n2 n4 n5 | n1 n6 | n3 n4 | n7 | r5 |
| r6 | n1 n5 n9 | n6 | n4 | n2 n5 n7 | n3 | n1 n2 n5 n7 | n8 | n1 n2 | n2 n9 | r6 |
| r7 | n3 n5 n7 n8 | n4 | n5 n6 n7 n8 | n2 n3 n5 n6 n7 | n5 n6 n7 | n2 n5 n7 | n2 n5 n7 | n9 | n1 | r7 |
| r8 | n1 n5 n7 n9 | n1 n9 | n2 | n4 n5 n6 n7 | n1 n4 n5 n6 n7 n9 | n8 | n3 | n4 n6 n7 | n4 n5 n6 | r8 |
| r9 | n1 n3 n5 n7 | n1 n3 n9 | **n1** | n2 n3 n5 n6 n4 n5 n6 n7 | n1 n4 n5 n6 n7 n9 | n1 n2 n4 n5 n7 | n2 n5 n7 | n2 n4 n6 n7 n8 | n2 n4 n5 n6 n8 | r9 |
|    | c1 | c2 | c3 | c4 | c5 | c6 | c7 | c8 | c9 |    |

*Figure 5.8. Resolution state RS$_1$ for puzzle in Figure 5.7*

At this point, there is no whip[3] but we find two braids[3]:

**braid[3]: r8c2{n1 n9} – r4c3{n1 n9} – c1n9{r4 .} ==> r5c2 ≠ 1**

**braid[3]: r8c2{n1 n9} – r4c3{n1 n9} – c1n9{r9 .} ==> r9c3 ≠ 1**

Anticipating on the definitions in chapter 7 and as an illustration of theorem 7.6, these eliminations could also be done respectively by the following g-whips[3] :
g-whip[3]: r8c2{n1 n9} – c1n9{r8 r456} – r4c3{n9 .} ==> r5c2 ≠ 1
g-whip[3]: r8c2{n1 n9} – c1n9{r8 r456} – r4c3{n9 .} ==> r9c3 ≠ 1

Let us now see the rest of the proof (in resolution theory $B_7$) that this puzzle has no solution:
whip[6]: b6n2{r6c9 r4c7} – r2n2{c7 c1} – c2n2{r1 r5} – c2n3{r5 r9} – r7n3{c1 c4} – r7n2{c4 .} ==> r6c6 ≠ 2
whip[7]: c2n2{r1 r5} – c2n8{r5 r3} – r2c3{n8 n7} – r3c3{n7 n1} – r3n9{c3 c9} – c7n9{r2 r4} – r4c3{n9 .} ==> r1c2 ≠ 9
whip[7]: b8n3{r7c4 r9c4} – c2n3{r9 r5} – c1n3{r5 r7} – b7n8{r7c1 r7c3} – b4n8{r5c3 r5c1} – r5n2{c1 c6} – b8n2{r9c6 .} ==> r7c4 ≠ 5, r7c4 ≠ 6
whip[5]: b7n8{r7c1 r7c3} – r7n6{c3 c5} – c4n6{r9 r4} – b5n8{r4c4 r4c5} – r2n8{c5 .} ==> r1c1 ≠ 8, r5c1 ≠ 8
whip[3]: c1n8{r7 r2} – c2n8{r3 r5} – b4n3{r5c2 .} ==> r7c1 ≠ 3
hidden-single-in-a-row ==> r7c4 = 3
whip[3]: c2n2{r5 r1} – b3n2{r1c7 r2c7} – r7n2{c7 .} ==> r5c6 ≠ 2
whip[1]: r5n2{c1 .} ==> r4c1 ≠ 2
braid[6]: b7n8{r7c1 r7c3} – r7n6{c3 c5} – r2n2{c1 c7} – c4n6{r9 r4} – r7n2{c7 c6} – r4n2{c7 .} ==> r2c1 ≠ 8
hidden-single-in-a-column ==> r7c1 = 8
whip[6]: r4n8{c4 c5} – b5n6{r4c5 r5c5} – r5c7{n6 n1} – r5c6{n1 n5} – r5c3{n5 n8} – r2n8{c3 .} ==> r4c4 ≠ 4
whip[6]: r7n2{c7 c6} – r4n2{c6 c4} – b5n8{r4c4 r4c5} – r2n8{c5 c3} – b4n8{r5c3 r5c2} – c2n2{r5 .} ==> r1c7 ≠ 2
whip[7]: r2n8{c5 c3} – b4n8{r5c3 r5c2} – b4n2{r5c2 r5c1} – r2n2{c1 c7} – r7n2{c7 c6} – r4n2{c6 c4} – b5n8{r4c4 .} ==> r1c5 ≠ 8
whip[7]: r4n8{c4 c5} – r2n8{c5 c3} – b4n8{r5c3 r5c2} – c2n2{r5 r1} – r2n2{c1 c7} – r7n2{c7 c6} – r4n2{c6 .} ==> r4c4 ≠ 6
whip[1]: c4n6{r9 .} ==> r8c5 ≠ 6, r7c5 ≠ 6
hidden-single-in-a-row ==> r7c3 = 6
whip[1]: c4n6{r9 .} ==> r9c5 ≠ 6
whip[5]: r7n2{c6 c7} – r2n2{c7 c1} – b1n4{r2c1 r1c1} – c4n4{r1 r8} – b8n6{r8c4 .} ==> r9c4 ≠ 2
whip[1]: b8n2{r9c6 .} ==> r4c6 ≠ 2
whip[3]: r4n6{c7 c5} – b5n8{r4c5 r4c4} – r4n2{c4 .} ==> r4c7 ≠ 1, r4c7 ≠ 9
hidden-single-in-a-block ==> r6c9 = 9
whip[1]: r3n9{c2 .} ==> r2c3 ≠ 9
whip[3]: r4c3{n9 n1} – r6c1{n1 n5} – b7n5{r9c1 .} ==> r9c3 ≠ 9
whip[3]: r4n6{c5 c7} – r4n2{c7 c4} – b5n8{r4c4 .} ==> r4c5 ≠ 1, r4c5 ≠ 4
whip[4]: r2n2{c7 c1} – c2n2{r1 r5} – b4n8{r5c2 r5c3} – r2c3{n8 .} ==> r2c7 ≠ 7
whip[3]: c6n2{r7 r9} – c6n9{r9 r2} – r2c7{n9 .} ==> r7c7 ≠ 2
hidden-single-in-a-row ==> r7c6 = 2
whip[2]: c3n5{r5 r9} – c6n5{r9 .} ==> r5c5 ≠ 5
whip[3]: b6n1{r5c7 r6c8} – r6c1{n1 n5} – b5n5{r6c6 .} ==> r5c6 ≠ 1

whip[4]: c3n5{r9 r5} – b4n8{r5c3 r5c2} – c2n2{r5 r1} – c9n2{r1 .} ==> r9c9 ≠ 5
whip[4]: r4c3{n1 n9} – b1n9{r3c3 r3c2} – r3n1{c2 c8} – b6n1{r6c8 .} ==> r5c3 ≠ 1
whip[2]: c3n9{r3 r4} – c3n1{r4 .} ==> r3c3 ≠ 8, r3c3 ≠ 7
whip[4]: b3n1{r1c7 r3c8} – r6n1{c8 c6} – b5n7{r6c6 r6c4} – r3n7{c4 .} ==> r1c1 ≠ 1
whip[3]: r8c2{n1 n9} – b1n9{r3c2 r3c3} – b1n1{r3c3 .} ==> r9c2 ≠ 1
whip[3]: r8c2{n9 n1} – b1n1{r3c2 r3c3} – b1n9{r3c3 .} ==> r9c2 ≠ 9
naked-single ==> r9c2 = 3
whip[4]: b7n5{r9c3 r8c1} – r6c1{n5 n1} – r9n1{c1 c6} – r4n1{c6 .} ==> r9c5 ≠ 5
whip[4]: r3n7{c8 c4} – b5n7{r6c4 r6c6} – r6n1{c6 c1} – c3n1{r4 .} ==> r3c8 ≠ 1
whip[1]: b3n1{r1c7 .} ==> r1c2 ≠ 1
whip[2]: c2n9{r3 r8} – c2n1{r8 .} ==> r3c2 ≠ 8
whip[3]: r3c9{n8 n5} – r1c9{n5 n2} – r1c2{n2 .} ==> r1c8 ≠ 8
whip[4]: c6n9{r9 r2} – r2c7{n9 n2} – r9c7{n2 n5} – r9c3{n5 .} ==> r9c6 ≠ 7
whip[4]: r6c1{n1 n5} – r5c3{n5 n8} – r2c3{n8 n7} – c6n7{r2 .} ==> r6c6 ≠ 1
whip[2]: b4n1{r6c1 r4c3} – c6n1{r4 .} ==> r9c1 ≠ 1
whip[1]: r9n1{c6 .} ==> r8c5 ≠ 1
whip[3]: r1n9{c5 c7} – c7n1{r1 r5} – c5n1{r5 .} ==> r9c5 ≠ 9
whip[3]: r9n9{c1 c6} – c6n1{r9 r4} – r4c3{n1 .} ==> r4c1 ≠ 9
singles ==> r4c3 = 9, r3c3 = 1, r3c2 = 9, r8c2 = 1
whip[4]: c6n1{r4 r9} – b8n9{r9c6 r8c5} – r1n9{c5 c7} – c7n1{r1 .} ==> r5c5 ≠ 1
singles ==> r4c6 = 1, r4c1 = 3, r4c9 = 4, r5c8 = 3, r2c8 = 6, r2c9 = 3, r9c5 = 1
whip[3]: r9n6{c4 c9} – r8c9{n6 n5} – r3n5{c9 .} ==> r9c4 ≠ 5
whip[4]: b7n7{r9c3 r8c1} – r8n9{c1 c5} – r1n9{c5 c7} – b3n7{r1c7 .} ==> r9c8 ≠ 7
whip[4]: b7n5{r9c1 r9c3} – c3n7{r9 r2} – c6n7{r2 r6} – c6n5{r6 .} ==> r5c1 ≠ 5
whip[4]: c9n2{r1 r9} – c8n2{r9 r6} – r6n1{c8 c1} – r5c1{n1 .} ==> r1c1 ≠ 2
whip[4]: b7n7{r9c3 r8c1} – r8c8{n7 n4} – c4n4{r8 r1} – r1c1{n4 .} ==> r9c4 ≠ 7
whip[4]: b8n7{r7c5 r8c4} – b8n6{r8c4 r9c4} – c4n4{r9 r1} – r1c1{n4 .} ==> r1c5 ≠ 7
whip[4]: c4n4{r9 r1} – r1c1{n4 n7} – b3n7{r1c8 r3c8} – r8c8{n7 .} ==> r8c5 ≠ 4
whip[4]: r2n2{c7 c1} – r5c1{n2 n1} – r6n1{c1 c8} – b6n2{r6c8 .} ==> r9c7 ≠ 2
whip[2]: b9n8{r9c8 r9c9} – b9n2{r9c9 .} ==> r9c8 ≠ 4
hidden-single-in-a-block ==> r8c8 = 4
whip[1]: b9n7{r9c7 .} ==> r1c7 ≠ 7
whip[2]: b9n8{r9c9 r9c8} – b9n2{r9c8 .} ==> r9c9 ≠ 6
singles ==> r8c9 = 6, r9c4 = 6, r9c6 = 4, r5c6 = 5, r6c6 = 7, r2c6 = 9, r2c7 = 2, r4c7 = 6, r5c7 = 1, r6c8 = 2
NO SOLUTION: NO CANDIDATE FOR RC-CELL r6c4.

### 5.11. Whips in N-Queens and Latin Squares; definition of SudoQueens

In this final section, mainly about the N-Queens problem, we show that the rules introduced in this chapter work concretely for other CSPs than Sudoku or LatinSquare. We also show that N-Queens has whips of length 1 and how they look like. More examples will appear (with more detail) in chapters 14 to 16. Using the LatinSquare CSP, we also show that a CSP with no whips of length 1 can nevertheless have longer ones. Finally, we introduce the N-SudoQueens CSP.

*5.11.1. The N-Queens CSP*

Given an n×n chessboard, the n-Queens CSP consists of placing n queens on it in such a way that no two queens appear in the same row, column or diagonal.

Here again, as in the Sudoku case, we introduce redundant sets of CSP variables:
- for each r° in {r1, r2, …, rn}, CSP variable Xr° with values in {c1, c2, …, cn};
- for each c° in {c1, c2, …, cn}, CSP variable Xc° with values in {r1, r2, …, rn}.

We define CSP-Variable-Type as the sort with domain {r, c} and Constraint-Type as the super-sort of CSP-Variable-Type with domain {r, c, f, s} corresponding to the four types of constraints: along a row, a column, parallel to the first diagonal and parallel to the second diagonal. Notice that there are now other constraints (f and s) than those taken care of by the CSP variables (corresponding to the r and c in Constraint-Type). And there is no possibility of adding CSP variables for the constraints along these diagonals: although no two queens may appear in the same diagonal, there are diagonals with no queen (there are 2n-1 diagonals of each kind); if we tried to define them as CSP variables, some of them would have no value.

For each r° in {r1, r2, …, rn} and each c° in {c1, c2, …, cn}, we define label (r°, c°) or r°c° as corresponding to the two <variable, value> pairs <Xr°, c°> and <Xc°, r°> (which is equivalent to the implicit axiom: Xr° = c° ⇔ Xc°= r°). A label can be assimilated with a cell in the grid.

Easy details of the model (in particular the writing of the constraints along rows, columns and diagonals) are left as an exercise for the reader. Similarly, the explicit writing of the Basic Resolution Theory BRT(n-Queens) is considered as obvious. As for whips, they need no specific definition; they are part of our general theory.

In all the forthcoming figures for n-Queens, the * signs represent the given queens; the small ° signs represent the candidates eliminated by ECP at the start of the resolution process; the A, B, C, … letters represent the candidates eliminated by resolution rules after the first ECP, in this order; the + signs represent the queens placed by the Single rule (at any time in the resolution process).

Notice that all our solutions for n-Queens were obtained manually; therefore, the resolution path for some of them may not be the shortest possible and the resolution theory in which the solution is obtained may not be the weakest possible. For lack of a generator of minimal instances, all our examples were built manually and they remain elementary. Our only ambition with respect to the n-Queens CSP is to illustrate how our general concepts can be applied and how our patterns look in them; contrary to Sudoku, it is not to produce any classification results.

### 5.11.2. Simple whips of length 1 and 2 in 8-Queens

|    | c1 | c2 | c3 | c4 | c5 | c6 | c7 | c8 |
|----|----|----|----|----|----|----|----|----|
| r1 | ∘  | *  | ∘  | ∘  | ∘  | ∘  | ∘  | ∘  |
| r2 | ∘  |    | ∘  | ∘  | ∘  | ∘  | *  | ∘  |
| r3 | ∘  |    | ∘  | ∘  | *  | ∘  | ∘  | ∘  |
| r4 |    | ∘  |    | ∘  | ∘  | ∘  | ∘  | +  |
| r5 | +  | ∘  | ∘  | ∘  | ∘  | ∘  | ∘  |    |
| r6 | ∘  | ∘  | ∘  | +  | ∘  |    | ∘  | ∘  |
| r7 |    | ∘  | B  | C  | ∘  | +  | ∘  | ∘  |
| r8 | ∘  | ∘  | +  | A  | ∘  |    | ∘  |    |

*Figure 5.9.* An 8-Queens instance solved by whips

For the 8-Queens CSP, consider the instance described in Figure 5.9, with 3 queens already given (in positions r1c2, r2c7 and r3c5). After the first obvious ECP eliminations, the Single rule cannot be applied. But we have the following resolution path with whips of lengths 1 and 2.

***** Manual solution *****
whip[1]: r6{c4 .} ⇒ ¬r8c4 (A eliminated)
whip[2]: r6{c4 c6} – r8{c6 .} ⇒ ¬r7c3, ¬r7c4 (B and C eliminated)
single in c4: r6c4; single in c6: r7c6; single in r5: r5c1; single in r4: r4c8; single in r8: r8c3
Solution found in $W_2$.

Notice the first whip[1], in the grey cells, with an interaction of a column and a diagonal occurring in a row at a relatively small distance from the target; it proves that there are whips of length 1 in n-Queens and it shows how some of them can look.

### 5.11.3. Whips[1] in 10-Queens with long distance interactions

The instance of 10-Queens in Figure 5.10 shows that whip[1] interactions can happen on much longer distances than in the previous example. They can also happen at distance 0, i.e. in the row or column adjacent to the target (as in section 5.11.5 below), or at still much longer distances in n-Queens for very large n.

|    | c1 | c2 | c3 | c4 | c5 | c6 | c7 | c8 | c9 | c10 |
|----|----|----|----|----|----|----|----|----|----|-----|
| r1 | ○ | ○ | ○ | ○ | ○ | ○ | ✱ | ○ | ○ | ○ |
| r2 | ○ | ○ | ○ | ○ | ○ | ○ | ○ | ○ | ○ | ✱ |
| r3 |   | ○ | ○ | ○ | ○ | + | ○ | ○ | ○ | ○ |
| r4 | ○ | ○ | ✱ | ○ | ○ | ○ | ○ | ○ | ○ | ○ |
| r5 | + | ○ | ○ | ○ | ○ |   | ○ | ○ | ○ | ○ |
| r6 | ○ | ○ | ○ | ○ | ○ | ○ | ○ | ✱ | ○ | ○ |
| r7 | ○ |   | ○ | + | ○ | ○ | ○ | ○ | ○ | ○ |
| r8 | C | + | ○ | ○ |   | ○ | ○ | ○ | ○ | ○ |
| r9 | ○ | ○ | ○ | ○ | ○ | ○ | ○ | ○ | ✱ | ○ |
| r10| B | ○ | ○ | ○ | + | A | ○ | ○ | ○ | ○ |

***Figure 5.10.*** *A 10-Queens instance, with 3 whips[1] based on long distance interactions*

This puzzle has five queens already given (in r1c7, r2c10, r4c3, r6c8 and r9c9). Its first three whips[1] have interactions of a column and a diagonal in rows at long distances from their targets. After them, it can be solved by Singles.

***** Manual solution *****
whip[1]: r5{c6 .} ⇒ ¬r10c6 (A eliminated); whip in light grey cells with target A
whip[1]: r5{c6 .} ⇒ ¬r10c1 (B eliminated); "same" whip in light grey cells, but with target B
whip[1]: r3{c1 .} ⇒ ¬r8c1 (C eliminated); whip in dark grey cells with target C
single in r10: r10c5; single in r8: r8c2; single in r7: r7c4; single in r5: r5c1; single in r3: r3c6
Solution found in $W_1$.

### 5.11.4. Another kind of whip[1] in N-Queens

The instance of 9-Queens in Figure 5.11, with three queens already given (in r3c3, r6c2 and r9c7) has three whips[1] of another kind, relying on the interaction of three different constraints in a row or a column at a medium distance from the target. It can be solved in $W_4$.

***** Manual solution *****
whip[1]: r7{c6 .} ⇒ ¬r5c6 (A eliminated, whip on light grey cells)
whip[1]: r8{c5 .} ⇒ ¬r4c5 (B eliminated, whip on medium grey cells and r8c5)
whip[1]: c5{r2 .} ⇒ ¬r5c8 (C eliminated, whip on dark grey cells)

whip[1]: c6{r4 .} ⇒ ¬r4c9 (D eliminated)
whip[2]: r5{c9 c4} – r7{c4 .} ⇒ ¬r8c9 (E eliminated)
whip[3]: r5{c9 c4} – r2{c1 c5} – r8{c5 .} ⇒ ¬r1c9 (F eliminated)
whip[4]: r5{c9 c4} – r4{c8 c6} – r1{c6 c8} – r8{c1 .} ⇒ ¬r2c9 (G eliminated)
whip[4]: r5{c9 c4} – r1{c4 c6} – r4{c6 c8} – r7{c8 .} ⇒ ¬r2c9 (H eliminated)
single in c9: r5c9; single in c4: r1c4; single in r4: r4c6; single in r2: r2c1; single in r8: r8c5; single in r7: r7c8.
Solution found in $W_4$ or $gW_3$.
|    | c1 | c2 | c3 | c4 | c5 | c6 | c7 | c8 | c9 |
|----|----|----|----|----|----|----|----|----|----|
| r1 | ∘ | ∘ | ∘ | + | ∘ |   | ∘ |   | F |
| r2 | + | ∘ | ∘ | ∘ |   | ∘ | ∘ |   | H |
| r3 | ∘ | ∘ | ✱ | ∘ | ∘ | ∘ | ∘ | ∘ | ∘ |
| r4 |   | ∘ | ∘ | ∘ | B | + | ∘ |   | D |
| r5 | ∘ | ∘ | ∘ |   | ∘ | A | ∘ | C | + |
| r6 | ∘ | ✱ | ∘ | ∘ | ∘ | ∘ | ∘ | ∘ | ∘ |
| r7 | ∘ | ∘ | ∘ | G | ∘ |   | ∘ | + | ∘ |
| r8 |   | ∘ | ∘ | ∘ | + | ∘ | ∘ | ∘ | E |
| r9 | ∘ | ∘ | ∘ | ∘ | ∘ | ∘ | ✱ | ∘ | ∘ |

*Figure 5.11. A 9-Queens instance, with another kind of whip[1]*

### 5.11.5. An instance of 8-Queens with two solutions

Whips can also be used to produce a readable proof that an instance has two (or more) solutions. For the 8-Queens CSP, consider the instance displayed in Figure 5.12, with 3 queens already given (in positions r2c7, r3c5 and r4c8). Although it has the same solution as the example in section 5.11.2, we shall prove that it has two solutions.

***** Manual solution *****
;;; The first two whips[1] display an interaction of a row and a diagonal in a column at the shortest possible distance from the target:
whip[1]: c3{r7 .} ⇒ ¬r7c4 (A eliminated)
whip[1]: c3{r8 .} ⇒ ¬r8c2 (B eliminated)

|    | c1 | c2 | c3 | c4 | c5 | c6 | c7 | c8 |
|----|----|----|----|----|----|----|----|----|
| r1 |    |    | ∘  | E  | ∘  | ∘  | ∘  | ∘  |
| r2 | ∘  | ∘  | ∘  | ∘  | ∘  | ∘  | *  | ∘  |
| r3 | ∘  | ∘  | ∘  | ∘  | *  | ∘  | ∘  | ∘  |
| r4 | ∘  | ∘  | ∘  | ∘  | ∘  | ∘  | ∘  | *  |
| r5 |    |    | ∘  | ∘  | ∘  | C  | ∘  | ∘  |
| r6 | D  | ∘  | ∘  | +  | ∘  | ∘  | ∘  | ∘  |
| r7 | ∘  | ∘  |    | A  | ∘  | +  | ∘  | ∘  |
| r8 | ∘  | B  | +  | ∘  | ∘  |    | ∘  | ∘  |

*Figure 5.12. An instance of 8-Queens with two solutions, partially solved by whips*

;;; The third whip[1], in the grey cells, appearing after B has been eliminated, has an interaction of a column and a diagonal in a row at a longer distance from the target:
whip[1]: r8{c6 .} ⇒ ¬r5c6 (C eliminated)

;;; The fourth whip[1], appearing after C has been eliminated, has an interaction of a column and a diagonal in a row, again at the shortest possible distance from the target:
whip[1]: r5{c1 .} ⇒ ¬r6c1 (D eliminated)
whip[2]: c1{r1 r5} – c2{r5 .} ⇒ ¬r1c4 (E eliminated)
single in r6 ⇒ r6c4 ; single in c3 ⇒ r8c3 ; single in r7 ⇒ r7c6

At this point, the resolution path cannot go further because there appears to be two obvious solutions: r1c2+r5c1 (as in section 5.11.1) and r1c1+r5c2; but we have shown that whips can be used to lead from a situation where this was not obvious to one where it is.

### *5.11.6. An instance of 6-Queens with no solution*

As shown in section 5.10.5, whips or braids can also provide a readable proof that an instance has no solution. Of course, this is not specific to Sudoku but it is true for any CSP. And the proof that an instance has no solution can be as hard as finding a solution when there is one. It can also be very simple, as shown below.

Consider Figure 5.13, an instance of 6-Queens, with only two queens given in cells r4c5 and r5c2. Although these data show no direct contradiction with the constraints, a unique elimination by a whip[3] and two Singles are enough to make it obvious, without trying all the remaining possibilities, that there can be no solution.

|    | c1 | c2 | c3 | c4 | c5 | c6 |
|----|----|----|----|----|----|----|
| r1 |    | ∘  |    | +  | ∘  | ∘  |
| r2 |    | ∘  | ∘  |    | ∘  |    |
| r3 | A  | ∘  | +  | ∘  | ∘  | ∘  |
| r4 | ∘  | ∘  | ∘  | ∘  | ∗  | ∘  |
| r5 | ∘  | ∗  | ∘  | ∘  | ∘  | ∘  |
| r6 | ∘  | ∘  | ∘  |    | ∘  |    |

*Figure 5.13. An instance of 6-Queens with no solution; proven by a whip[3]*

***** Manual solution *****
whip[3]: r6{c4 c6} – r2{c6 c4} – r1{c4 .} ⇒ ¬r3c1 (A eliminated)
single in r3 ⇒ r3c3
single in r1 ⇒ r1c4
This puzzle has no solution: no value for Xr6

### *5.11.7. The absence of whip[1] does not preclude the existence of longer whips*

The non-existence of whips of length 1 in a CSP does not preclude the existence of longer whips. Figure 5.14 gives an example of a partial whip[3] in LatinSquare.

In this Figure, black horizontal lines represent CSP variables ($V_1$, $V_2$, $V_3$); they are supposed to have candidates only at their extremities ($L_k$ and $R_k$ candidates) or at their meeting points with arrows (z- and t- candidates). Dark grey vertical arrows represent links from Z to $L_1$ or from $R_k$ to $L_{k+1}$. Light grey arrows represent links to z- or t- candidates. Here, arrows represent only the flow of reasoning in the proof of the whip rule (by themselves, links are not orientated).

A particular interpretation of Figure 5.14 can be obtained by considering only labels (n, r, c) with a fixed Number n and by interpreting horizontal lines as rows and vertical lines as columns. Similarly, one can fix Row r or Column c. But these restricted visions of the symbolic representation, limited to rc-space (or cn-space, or rn-space), do not take into account the 3D symmetries of this CSP.

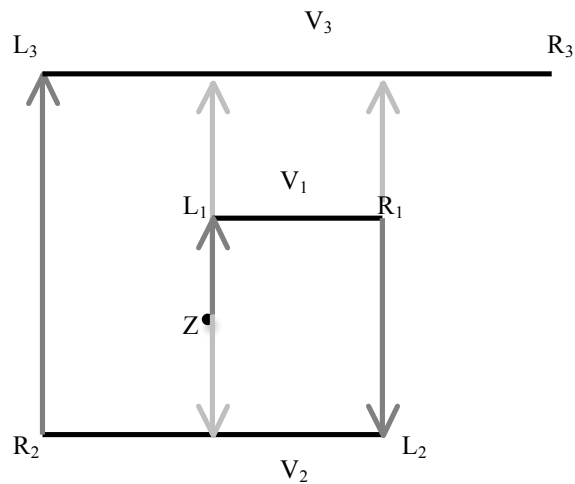

*Figure 5.14.* *A symbolic representation of a partial whip[3] in LatinSquare.*

Similar symbolic representations, for whips in a general CSP (Figure 11.1) and for generalised whips (Figures 9.1 and 11.2) can be seen in chapters 9 and 11.

### *5.11.8. Defining SudoQueens*

Given an integer n that is a square ($n = m^2$) and starting from the n-Queens CSP, one can define the n-SudoQueens CSP by the additional constraint that there should not be two queens in the same m×m block, where blocks are defined as in Sudoku.

In this new CSP, we can use the same two coordinate systems as in Sudoku, with the same relations between them. Because it implies that there must be one queen in each square, the new constraint can be taken care of by n new CSP-Variables $Xb1$, …, $Xbn$, all with domain {s1, …., sn} and/or by a new CSP-Variable-Type: b.

It is easy to check that n-SudoQueens has no instances for n=2 or n=4 (i.e. m=1 or m=2). But, as shown by the example in Figure 5.15, it has for n ≥ 9 (m ≥ 3).

In n-SudoQueens, one can find two types of whips[1]: the same as in n-Queens and the same as in Sudoku[n].

|    | c1 | c2 | c3 | c4 | c5 | c6 | c7 | c8 | c9 |
|----|----|----|----|----|----|----|----|----|----|
| r1 | ∗  | ∘  | ∘  | ∘  | ∘  | ∘  | ∘  | ∘  | ∘  |
| r2 | ∘  | ∘  | ∘  | ∘  | ∘  | ∘  | ∘  | ∗  | ∘  |
| r3 | ∘  | ∘  | ∘  | ∘  | ∗  | ∘  | ∘  | ∘  | ∘  |
| r4 | ∘  | ∘  | ∗  | ∘  | ∘  | ∘  | ∘  | ∘  | ∘  |
| r5 | ∘  | ∘  | ∘  | ∘  | ∘  | ∗  | ∘  | ∘  | ∘  |
| r6 | ∘  | ∘  | ∘  | ∘  | ∘  | ∘  | ∘  | ∘  | ∗  |
| r7 | ∘  | ∗  | ∘  | ∘  | ∘  | ∘  | ∘  | ∘  | ∘  |
| r8 | ∘  | ∘  | ∘  | ∗  | ∘  | ∘  | ∘  | ∘  | ∘  |
| r9 | ∘  | ∘  | ∘  | ∘  | ∘  | ∘  | ∗  | ∘  | ∘  |

*Figure 5.15. A complete grid for 9-SudoQueens*

# 6. Unbiased statistics and whip classification results

In the previous chapter, we gave a pure logic definition of the W and B ratings of an instance P, as the smallest n ($0 \leq n \leq \infty$) such that P can be solved by resolution theory $W_n$ [respectively $B_n$]. Because these theories involve longer and longer whips [resp. braids] as n increases, it is *a priori* meaningful for any CSP to chose W(P) [resp. B(P)] as a measure of complexity for P. In the Sudoku case, there are additional justifications, based on results[2] obtained with our SudoRules solver:

– W [resp. B] is strongly correlated with the logarithm of the number[3] of partial whips [resp. braids] one must check before finding the solution when the "simplest first" strategy is adopted[4];

– for W $\leq$ 9,[5] W [resp. B] is strongly correlated with SER, the Sudoku Explainer rating [Juillerat www]; this rating (version 1.2.1) is widely used in the Sudoku community in spite of its many shortcomings[6]; it often gives some rough idea of the difficulty of a puzzle for a human player (at least for SER $\leq$ 9.3);

– W is also well correlated with less popular ratings (see our website).

It should however be noted that a rating based on the hardest step (instead of e.g. the whole resolution path) can only be meaningful statistically. (This applies also to SER.) In particular, there remains much variance in the number of partial chains needed to solve Sudoku puzzles with W(P) = n, n fixed. Based on the thousands of resolution paths we observed in detail, one explanation is that a puzzle P with W(P) = n can be hard to solve with whips [or any other type of pattern: braids, g-whips, …] for two opposed reasons: either because it does not have enough smaller whips [patterns of this type] or because it has too many useless ones.

---

[2] Details and additional correlation results can be found on our website.

[3] Although this number is not completely independent of implementation (it depends in part on the resolution path chosen), it is statistically meaningful.

[4] In this situation, W is also strongly correlated with the logarithm of the resolution time, but this is mainly a consequence of the previous correlation (and computation times are too implementation-dependent to be good indicators).

[5] For larger values of W, the number of available instances in our unbiased samples is too small to compute meaningful correlations.

[6] SER is defined only by non-documented Java code, it is not invariant under logical symmetries and it is based neither on any general theory nor (for the most part of it) on any popular application-specific resolution techniques. Indeed, the main part of SER is based on the number of inference steps (which is implementation dependent) in a resolution procedure more or less equivalent to T&E(1) complemented by T&E(2) when T&E(1) is not enough; it is easy to see that this cannot be given by a purely logical definition (because a logical theory can put no limit on how many applications of its axioms may be used to prove a theorem). But it is free and it is the "less worse" of the currently available ratings.

The results[7] reported in this chapter required several months of (2.66 GHz) CPU time (for the generation of unbiased samples and for the computation of ratings). They will show that:

– building unbiased uncorrelated samples of minimal instances of a (fixed size) CSP and obtaining unbiased statistics can be very hard;

– (loopless) whips have a very strong resolution power, at least for Sudoku; the ten million puzzles we have produced using different kinds of random generators could all be solved by whips of relatively short length: 93.9% by whips of length no more than 4, 99.9% by whips of length no more than 7 and 99.99% by whips of length no more than 9 – see Table 6.4.

Only the main results of direct relevance to the topic of this book are provided here; many additional statistical results for Sudoku can be found on our website.

Although we can only present such results in the specific context of the Sudoku CSP, the sample generation methods described here (bottom-up, top-down and controlled-bias) could be extended to many CSPs. The specific P(n+1)/P(n) formula proven in section 6.2.2 for the controlled-bias generator will not hold in any CSP, but the same approach can in many cases help understand the existence of a very strong bias in the samples with respect to the number of clues (see the end of chapter 14 for an adaptation to the Futoshiki CSP). Probably, it can also help explain the well-known fact that, for many CSPs, it is very difficult to generate the hardest instances.

The number of clues may not be a criterion of much interest in itself, but the existence of such a strong bias in it suggests the possibility of a bias with respect to many other different classification criteria, even if they are weakly correlated with the number of clues: in the Sudoku case, preliminary analyses showed that the correlation coefficient between the W rating and the number of clues is only 0.12, but Tables 6.3 and 6.4 below show that the bias in the generators has nevertheless a very noticeable impact on the classification of instances according to the W rating.

Even in the very structured and apparently simple Sudoku domain, none of this was clear before the present analysis. In particular, as the results in *HLS* were based on a top-down generator, they were biased.

---

[7] We first published them on the late Sudoku Player's Forum (July to October 2009) and then in [Berthier 2009].

*Acknowledgements*: Thanks are due to "Eleven" for implementing the first modification (suexg-cb) of a well-known top-down generator (suexg, written in C) to make it compliant with the specification of controlled-bias defined below, and then several faster versions of it; this allowed to turn the whole idea into reality. Thanks to Paul Isaacson for adapting Brian Turner's fast solver so that it could be used instead of that of suexg. Thanks to Glenn Fowler (alias gsf) for providing an *a priori* unbiased source of complete grids: the full (compressed) collection of their equivalence classes together with a fast decompressor. Thanks also, for discussions and/or various contributions, to Allan Barker, Coloin, David P. Bird, Mike Metcalf, Red Ed (who was first to suggest the existence of a bias in the current generators). The informal collaboration that the controlled-bias idea sprouted on the late Sudoku Player's Forum was very productive: due to several independent optimisations, the last version of suexg-cb (which does not retain much of the original suexg code) is 200 times faster than the first.

All the generators mentioned below are available on our website.

### 6.1 Classical top-down and bottom-up generators

There is a very simple procedure for generating an unbiased sample of n uncorrelated minimal Sudoku puzzles:

```
1) set p = 0 and list = ();
2) if p = n then return list;
3) randomly choose a complete grid P;
4) for each cell in P, delete its value with probability 0.5, thus
obtaining a puzzle Q;
5) if Q is minimal then add Q to list, set p = p+1 and goto 2 else
goto 3.
```

Unfortunately, the probability of getting a valid puzzle this way is infinitesimal for each complete grid tried as a starting point (see last column of Table 6.2, which should be combined for each n with the probability of obtaining 81-n deletions). One has no choice but rely on more efficient generators. Before going further, let us introduce the two classical algorithms that have been widely used in the Sudoku community for generating minimal puzzles: bottom-up and top-down.

A standard bottom-up generator works as follows to produce n minimal puzzles:

```
1) set p = 0 and list = ();
2) if p = n then return list;
3a) set p = p+1 and start from an empty grid P;
3b) in P, randomly choose an undecided cell and a value for it, thus
getting a puzzle Q with one more clue than P;
3c) if Q is minimal, then add it to list and goto 2;
3d) if Q has several solutions, then set P = Q and goto 3b;
```

```
3e) if Q has no solution, then goto 3b (i.e. backtrack: forget Q and
try another cell from P).
```

A standard top-down generator works as follows to produce n minimal puzzles:

```
1) set p = 0 and list = ();
2) if p = n then return list;
3a) set p = p+1 and randomly choose a complete grid P;
3b) randomly choose one clue from P and delete it, thus obtaining a
puzzle Q;
3c) if Q still has only one solution but is not minimal, set P=Q and
goto 3b (for trying to delete one more clue);
3d) if Q is minimal, then add it to list and goto 2;
3e) otherwise, i.e. if Q has several solutions, then goto 3b (i.e.
reinsert the clue just deleted and try deleting another clue from P).
```

Notice that, in both cases, a minimal puzzle is produced from each complete random grid. Backtracking (i.e. clause 3e in both cases) makes any formal analysis of these algorithms very difficult. However, at first sight, it seems that it causes the generator to look for puzzles with fewer clues (this intuition will be confirmed in section 6.3). It may thus be suspected of introducing a strong, uncontrolled bias with respect to the number of clues, which, in turn, may induce a bias with respect to other properties of the collection of puzzles generated.

**6.2 A controlled-bias generator**

No unbiased generator of uncorrelated minimal puzzles is currently known and building such a generator with reasonable computation times seems out of reach. We therefore decided to proceed differently: taking the generators (more or less) as they are and applying corrections for the bias, if we can estimate it.

This idea was inspired by an article we read in a newspaper about what is done in digital cameras: instead of complex optimisations of the lenses to reduce typical anomalies (such as chromatic aberration, purple fringing, barrel or pincushion distortion…) – optimisations that lead to large and expensive lenses –, some camera makers now accept a small amount of these in the lenses and they take advantage of the huge computational power available in the processors to correct the result in real time with dedicated software before recording the photo.

The main question was then: can we determine the bias of the classical top-down or bottom-up generators? The answer was negative. But there appeared to be a medium way between "improving the lens to make it perfect" and "correcting its small defects by software": we devised a modification of the top-down generator that allows a precise mathematical computation of the bias.

*6.2.1. Definition of the controlled-bias generator*

Consider the following, modified top-down generator, the ***controlled-bias generator*** for producing n minimal uncorrelated puzzles:

```
1) set p = 0 and list = ();
2) if p = n then return list;
3a) randomly choose a complete grid P;
3b) randomly choose one clue from P and delete it, thus obtaining a
puzzle Q;
3c) if Q still has only one solution but is not minimal, set P=Q and
goto 3b (for trying to delete one more clue);
3d) if Q is minimal, then add it to list, set p = p+1 and goto 2;
3e) otherwise, i.e. if Q has several solutions, then goto 3a (i.e.
forget everything about P and restart with another complete grid).
```

The only difference with the top-down algorithm is in clause 3e: if a multi-solution puzzle is encountered, instead of backtracking to the previous state, the current complete grid is merely discarded and the search for a minimal puzzle is restarted with another complete grid.

Notice that, contrary to the standard bottom-up or top-down generators, which produce one minimal puzzle per complete grid, the controlled-bias generator will generally use several complete grids before it outputs a minimal puzzle. The efficiency question is: how many? Experimentations show that many complete grids (approximately 257,514 in the mean) are necessary before a minimal puzzle is reached. But this question is about the efficiency of the generator, it is not a conceptual problem.

The controlled-bias generator has the same output and will therefore produce minimal puzzles according to the same probability distribution as its following "virtual" counterpart:

```
1) set p = 0 and list = ();
2) if p = n then return list;
3a) randomly choose a complete grid P;
3b) if P has no more clue, then goto 2 else randomly choose one clue
from P and delete it, thus obtaining a puzzle Q;
3c) if Q is minimal, add Q to list, set P=Q, set p=p+1 and goto 3b;
3d) otherwise, set P=Q and goto 3b.
```

The only difference with the controlled-bias generator is that, once it has found a minimal or a multi-solution puzzle, instead of exiting, this virtual generator continues along a useless path until it reaches the empty grid.

But this virtual generator is interesting theoretically because it works similarly to the random uniform search defined in the next section and according to the same transition probabilities; and it outputs minimal puzzles according to the probability Pr on the set B of minimal puzzles defined below.

### 6.2.2. Analysis of the controlled-bias generator

We now build our formal probabilistic model of the controlled-bias generator.

Let us first introduce the notion of a *doubly indexed puzzle*. We consider only (single or multi solution) consistent puzzles P. The double index of a doubly indexed puzzle P has a clear intuitive meaning: the first index is one of its solution grids and the second index is a sequence (notice: not a set, but a sequence, i.e. an ordered set) of clue deletions leading from this complete grid to P. In a sense, the double index keeps track of the full generation process.

Given a doubly indexed puzzle Q, there is an underlying singly-indexed puzzle: the ordinary puzzle obtained by forgetting the second index of Q, i.e. by remembering the solution grid from which it came and by forgetting the order of the deletions leading from this solution to Q. Given a doubly indexed puzzle Q, there is also a non indexed puzzle, obtained by forgetting the two indices.

For a single solution doubly indexed puzzle, the first index is useless as it can be computed from the puzzle; in this case singly indexed and non-indexed are equivalent. This is true in particular for minimal puzzles. In terms of the generator, it could equivalently output minimal puzzles or couples (minimal-puzzle, solution).

Consider now the following layered structure (a forest, in the graph-theoretic sense, i.e. a set of disjoint trees, with branches pointing downwards), the nodes being (single or multi solution) doubly indexed puzzles:

– floor 81 : the N different complete solution grids (considered as puzzles), each indexed by itself and by the empty sequence; notice that all the puzzles at floor 81 have 81 clues;

– recursive step: given floor n+1, where each doubly indexed puzzle has n+1 clues and is indexed by a complete grid that solves it and by a sequence of length 81-(n+1), build floor n as follows:
each doubly indexed puzzle Q at floor n+1 sprouts n+1 branches; for each clue C in Q, there is a branch leading to a doubly indexed puzzle R at floor n: R is obtained from Q by removing clue C; its first index is identical to that of Q and its second index is the (81-n)-element sequence obtained by appending C to the end of the second index of Q; notice that all the doubly indexed puzzles at floor n have n clues and the length of their second index is equal to 1 + (81-(n+1)) = 81-n.

It is easy to see that, at floor n, each doubly indexed puzzle has an underlying singly indexed puzzle identical to that of (81 - n)! doubly indexed puzzles with the same first index (i.e. the same solution grid) at the same floor (including itself).

This is equivalent to saying that, at any floor n < 81, any singly indexed puzzle Q can be reached by exactly (81 - n)! different paths from the top (all of which start necessarily from the complete grid defined as the first index of Q). These paths are the (81 - n)! different ways of deleting one by one its missing 81-n clues from its solution grid.

Notice that this would not be true for non-indexed puzzles that have multiple solutions. This is where the first index is useful.

Let N be the number of complete grids (N is known to be close to $6.67 \times 10^{21}$, but this is pointless here). At each floor n, there are $N \times 81! / n!$ doubly indexed puzzles and $N \times 81! / (81-n)! / n!$ singly indexed puzzles. For each n, there is therefore a uniform probability $P(n) = 1/N \times 1/81! \times (81-n)! \times n!$ that a singly indexed puzzle Q at floor n is reached by a random (uniform) search starting from one of the complete grids. What is important here is the ratio: $P(n+1) / P(n) = (n + 1) / (81 - n)$, giving the relative probability of being reached by the generation process, for two singly indexed puzzles with respectively n+1 and n clues.

The above formula is valid globally if we start from all the complete grids, as above, but it is also valid for all the single solution puzzles if we start from a single complete grid (just forget N in the proof above). (Notice however that it is not valid if we start from a subgrid instead of a complete grid.)

Now, call B the set of (non indexed) minimal puzzles. On B, all the puzzles are minimal. Any puzzle strictly above B has redundant clues and a single solution. Notice that, for all the puzzles on B and above B, singly indexed and non-indexed puzzles are in one-to-one correspondence. Therefore, the relative probability of two minimal puzzles is given by the above formula.

On the set B of minimal puzzles, there is thus a probability Pr naturally induced by the different Pn's and it is the probability that a minimal puzzle Q is output by our controlled-bias generator. It depends only on the number of clues and it is defined by $Pr(Q) = P(n)$ if Q has n clues.

The most important here is that, by construction of Pr on B (a construction which models the workings of the virtual controlled bias generator), the fundamental relation: $Pr(n+1)/Pr(n) = (n+1)/(81-n)$ holds for any two minimal puzzles, with respectively n+1 and n clues.

For n < 41, this relation means that a minimal puzzle with n clues is more likely to be reached from the top than a minimal puzzle with n+1 clues. More precisely, we have: $Pr(40) = Pr(41)$, $Pr(39) = 42/40 \times Pr(40)$, $Pr(38) = 43/39 \times Pr(39)$. Repeated application of the formula gives $Pr(24) = 61.11 \times Pr(30)$: *a puzzle with 24 clues has about 61 times more chances of being output by the controlled-bias generator than a puzzle with 30 clues*. This is indeed a very strong bias.

A non-biased generator would give the same probability to all the minimal puzzles. The above analysis shows that *the controlled bias generator:*
*- is unbiased when restricted (by filtering its output) to n-clue puzzles, for any fixed n,*
*- is strongly biased towards puzzles with fewer clues,*
*- this bias is well known and given by Pr(n+1) / Pr(n) = (n + 1) / (81 – n),*
*- the puzzles produced are uncorrelated, provided that the complete grids are chosen in an uncorrelated way*.

*As we know precisely the bias with respect to uniformity, we can correct it easily by applying correction factors cf(n) to the probabilities on B. Only the relative values of the cf(n) is important: they satisfy cf(n+1) / cf(n) = (81-n)/(n+1).* Mathematically, after normalisation, cf is just the relative density of the uniform distribution on B with respect to the probability distribution Pr.

This analysis also shows that a classical top-down generator is still more strongly biased towards puzzles with fewer clues because, instead of discarding the current path when it meets a multi-solution puzzle, it backtracks to the previous floor and tries again to go deeper.

### 6.2.3. Computing unbiased means and standard deviations using a controlled-bias generator

In practice, how can one compute unbiased statistics of minimal puzzles based on a (large) sample produced by a controlled-bias generator? Consider any random variable X defined (at least) on the set of minimal puzzles. Define: on(n) = the number of n-clue puzzles in the sample, E(X, n) = the mean value of X for n-clue puzzles in the sample and $\sigma(X, n)$ = the standard deviation of X for n-clue puzzles in the sample.

The mean and standard-deviation of X on a sample are classically computed as:
mean(X) = $\sum_n [E(X, n) \times on(n)] / \sum_n on(n)$
$\sigma(X) = \sqrt{\{\sum_n [\sigma(X, n)^2 \times on(n)] / \sum_n [on(n)]\}}$.

The unbiased mean and standard deviation of X must then be estimated as (this is merely the mean and standard deviation for a weighted average):
*unbiased-mean(X) = $\sum_n [E(X, n) \times on(n) \times cf(n)] / \sum_n [on(n) \times cf(n)]$;*
*unbiased-$\sigma$(X) = $\sqrt{\{\sum_n [\sigma(X, n)^2 \times on(n) \times cf(n)] / \sum_n [on(n) \times cf(n)]\}}$*.

These formulæ show that the cf(n) sequence needs be defined only modulo a multiplicative factor. It is convenient to choose cf(26) = 1. This gives the following sequence of correction factors (in the range n = 19-31, which includes all the puzzles of all the samples we have obtained with all the random generators considered here):

[0.00134 0.00415 0.0120 0.0329 0.0843 0.204 0.464 1 2.037 3.929 7.180 12.445 20.474]

It may be shocking to consider that 30-clue puzzles in a sample must be given a weight 61 times greater than 24-clue puzzles, but it is a fact. As a result of this strong bias of the controlled-bias generator (strong but known and much smaller than the other generators), unbiased statistics for the mean number of clues of minimal puzzles (and any variable correlated with this number) must rely on extremely large samples with sufficiently many 29-clue and 30-clue puzzles.

### 6.3. The real distribution of clues and the number of minimal puzzles

The above formulæ show that the number-of-clue distribution of the controlled-bias generator is the key for computing unbiased statistics.

#### 6.3.1. The number-of-clue distribution as a function of the generator

| Generator →<br>sample size →<br>↓ #clues | bottom-up<br>1,000,000<br>% (sample) | top-down<br>1,000,000<br>% (sample) | ctr-bias<br>5,926,343<br>% (sample) | real<br>% (estimated) |
|---|---|---|---|---|
| 20 | 0.028 | 0.0044 | 0.0 | 0.0 |
| 21 | 0.856 | 0.24 | 0.0030 | 0.000034 |
| 22 | 8.24 | 3.45 | 0.11 | 0.0034 |
| 23 | 27.67 | 17.25 | 1.87 | 0.149 |
| 24 | 36.38 | 34.23 | 11.85 | 2.28 |
| 25 | 20.59 | 29.78 | 30.59 | 13.42 |
| 26 | 5.45 | 12.21 | 33.82 | 31.94 |
| 27 | 0.72 | 2.53 | 17.01 | 32.74 |
| 28 | 0.054 | 0.27 | 4.17 | 15.48 |
| 29 | 0.0024 | 0.017 | 0.52 | 3.56 |
| 30 | 0 | 0.001 | 0.035 | 0.41 |
| 31 | 0 | 0 | 0.0012 | 0.022 |
| **mean** | **23.87** | **24.38** | **25.667** | **26.577** |
| std-dev | 1.08 | 1.12 | 1.116 | 1.116 |

*Table 6.1: The experimental number-of-clue distribution (%) for the bottom-up, top-down and controlled-bias generators and the estimated real distribution.*

After applying the above formulæ to estimate the real number-of-clue distribution, Table 6.1 shows that the bias with respect to the number of clues is very strong in all the generators we have considered; moreover, *controlled-bias, top-*

162                    Pattern-Based Constraint Satisfaction and Logic Puzzles*down and bottom-up are increasingly biased towards puzzles with fewer clues.*
Graphically, the estimated number-of-clue distribution is very close to Gaussian.

Table 6.1 partially explains Tables 6.3 and 6.4 in section 6.4. More precisely, it explains why there can be a noticeable W rating bias in the samples produced by the bottom-up and top-down generators, in spite of the weak correlation coefficient between the number of clues and the W rating of a puzzle: the bias with respect to the number of clues is very strong in these generators.

### *6.3.2. Collateral result: the number of minimal puzzles*

The number of minimal Sudoku puzzles has been a longstanding open question. We can now provide precise estimates for the distribution of the mean number of n-clue minimal puzzles per complete grid (mean and standard deviation in the second and third columns of Table 6.2).

| number of clues | number of n-clue minimal puzzles per complete grid: mean | number of n-clue minimal puzzles per complete grid: relative error (~ 1 std dev) | mean number of tries |
|---|---|---|---|
| 20 | $6.152 \times 10^6$ | 70.7% | $7.6306 \times 10^{11}$ |
| 21 | $1.4654 \times 10^9$ | 7.81% | $9.3056 \times 10^9$ |
| 22 | $1.6208 \times 10^{12}$ | 1.23% | $2.2946 \times 10^8$ |
| 23 | $6.8827 \times 10^{12}$ | 0.30% | $1.3861 \times 10^7$ |
| 24 | $1.0637 \times 10^{14}$ | 0.12% | $2.1675 \times 10^6$ |
| 25 | $6.2495 \times 10^{14}$ | 0.074% | $8.4111 \times 10^5$ |
| 26 | $1.4855 \times 10^{15}$ | 0.071% | $7.6216 \times 10^5$ |
| 27 | $1.5228 \times 10^{15}$ | 0.10% | $1.5145 \times 10^6$ |
| 28 | $7.2063 \times 10^{14}$ | 0.20% | $6.1721 \times 10^6$ |
| 29 | $1.6751 \times 10^{14}$ | 0.56% | $4.8527 \times 10^7$ |
| 30 | $1.9277 \times 10^{13}$ | 2.2% | $7.3090 \times 10^8$ |
| 31 | $1.1240 \times 10^{12}$ | 11.6% | $2.0623 \times 10^{10}$ |
| 32 | $4.7465 \times 10^{10}$ | 70.7% | $7.6306 \times 10^{11}$ |
| **Total** | **$4.6655 \times 10^{15}$** | **0.065%** | |

*Table 6.2: Mean number of n-clue minimal puzzles per complete grid. Last column: inverse of the proportion of n-clue minimal puzzles among n-clue sub-grids*

Another number of interest (e.g. for the first naïve algorithm given in section 6.1) is the mean number of tries one must do to find an n-clue minimal puzzle by randomly deleting 81-n clues from a complete grid. It is the inverse of the proportion of n-clue minimal puzzles among n-clue sub-grids, given by the last column in Table 6.2.

One can also get:

– after multiplying the total mean by the number of complete grids (known to be 6,670,903,752,021,072,936,960 [Felgenhauer et al. 2005]), **the total number of minimal Sudoku puzzles: $3.1055 \times 10^{37}$**, *with 0.065% relative error;*

– after multiplying the total mean by the number of non isomorphic complete grids (known to be 5,472,730,538 [Russell et al. 2006]), **the total number of non isomorphic minimal Sudoku puzzles: $2.5477 \times 10^{25}$**, *also with 0.065% relative error.*

### 6.4. The W-rating distribution as a function of the generator

We can now apply the bias correction formulæ of section 6.2.3 to estimate the W rating distribution. Table 6.3 shows that the mean W rating of the minimal puzzles in a sample depends noticeably on the type of generator used to produce them and that all the generators give rise to mean complexity below the real values.

| Generator         | bottom-up | top-down | ctr-bias   | real |
|-------------------|-----------|----------|------------|------|
| sample size       | 10,000    | 50,000   | 5,926,343  |      |
| W rating : mean   | **1.80**  | **1.94** | **2.22**   | **2.45** |
| W rating : std dev| 1.24      | 1.29     | 1.35       | 1.39 |
| max W found in sample | 11    | 13       | 16         |      |

*Table 6.3: The W-rating means and standard deviations for bottom-up, top-down and controlled-bias generators, compared with the estimated real values.*

The mean W rating gives only a very pale idea of what really happens, because the first two levels, $W_0$ and $W_1$, concentrate a large part of the distribution, for any of the generators. With the full distributions, Table 6.4 provides more detail about the bias in the W rating for the three kinds of generators (with the same sample sizes as in Table 6.3). All these distributions have the same two modes as the real distribution, at levels $W_0$ and $W_3$. But, when one moves from bottom-up to top-down to controlled-bias to real, the mass of the distribution moves progressively to the right. This displacement towards higher complexity occurs mainly at the first W levels, after which it is only slight, but still visible.

More detailed analyses (available on our website), in particular with skewness and kurtosis, seem to show that there is a (non absolute) barrier of complexity, such that, when we consider n-clue puzzles and when the number n of clues increases:
- the n-clue mean W rating increases;
- the proportion of puzzles with W rating away from the n-clue mean increases;
but:
- the proportion of puzzles with W rating far below the n-clue mean increases;
- the proportion of puzzles with W rating far above the n-clue mean decreases.

Graphically, the W rating distribution of n-clue puzzles looks like a wave. When n increases, the wave moves to the right, with a longer tail on its left and a steeper front on its right. The same remarks apply if the W rating is replaced by the SER.

| Generator → <br> W-rating ↓ | bottom-up <br> % (sample) | top-down <br> % (sample) | ctr-bias <br> % (sample) | **real** <br> **% (estimated)** |
|---|---|---|---|---|
| 0 (first mode →) | 46.27 | 41.76 | 35.08 | **29.17** |
| 1 | 13.32 | 12.06 | 9.82 | **8.44** |
| 2 | 12.36 | 13.84 | 13.05 | **12.61** |
| 3 (second mode →) | 15.17 | 16.86 | 20.03 | **22.26** |
| 4 | 10.18 | 12.29 | 17.37 | **21.39** |
| 5 | 1.98 | 2.42 | 3.56 | **4.67** |
| 6 | 0.49 | 0.55 | 0.79 | **1.07** |
| 7 | 0.19 | 0.15 | 0.21 | **0.29** |
| 8 | 0.020 | 0.047 | 0.055 | **0.072** |
| 9 | 0.010 | 0.013 | 0.015 | **0.020** |
| 10 | 0* | $3.8 \, 10^{-3}$ | $4.4 \, 10^{-3}$ | **$5.5 \, 10^{-3}$** |
| 11 | 0.01* | $1.5 \, 10^{-3}$ | $1.2 \, 10^{-3}$ | **$1.5 \, 10^{-3}$** |
| 12-16 | 0* | $1.1 \, 10^{-3}$ | $4.3 \, 10^{-4}$ | **$5.4 \, 10^{-4}$** |

*Table 6.4: The W-rating distribution (in %) for bottom-up, top-down and controlled-bias generators, compared with the estimated real distribution. A * sign on a result means that the number of puzzles justifying it is too small to allow a precise value.*

### 6.5. Stability of the classification results

#### 6.5.1. Insensivity of the controlled-bias generator wrt the source of complete grids

There remains a final question: do the above results depend on the source of complete grids? Until now, we have done as if this was not a problem. Nevertheless, producing the unbiased and uncorrelated collections of complete grids, necessary in the first step of all the puzzle generators, is all but obvious. It is known that there are $6.67 \times 10^{21}$ complete grids; it is therefore impossible to have a generator scan them all. Up to isomorphisms, there are "only" $5.47 \times 10^9$ complete grids, but this remains a very large number and storing them in uncompressed format would require about half a terabyte.

In 2009, Glenn Fowler provided both a collection of all the (equivalence classes of) complete grids in a compressed format (only 6 gigabytes) and a real time decompressor. All the results reported above for the controlled bias generator were obtained with this *a priori* unbiased source of complete grids. (Notice that, due to the normalisation and compression of grids, it is unbiased only when one does full scans of its grids, whence the queer sizes of some of our samples of controlled-bias minimal puzzles).

Before this, all the generators we tried had a first phase consisting of creating a complete grid and this is where some type of bias could slip in at this level. Nevertheless, we tested several sources of complete grids based on very different generation principles and the classification results remained very stable.

This insensitivity of the controlled-bias generator to the source of complete grids can be understood intuitively: it deletes in the mean two thirds of the initial grid data and any structure that might be present in the complete grids and cause a bias is washed away by the deletion phase.

### *6.5.2. Insensivity of the classification results wrt the generators implementation*

As can be seen from additional results on our website, we have tested several independent implementations of the bottom-up and top-down generators, using in particular various pseudo-random number generators for the selection of clue deletions (or additions in the bottom-up case); they all lead to the same conclusions.

### **6.6. The W rating is a good approximation of the B rating**

The above statistical results are unchanged when the W rating is replaced by the B rating. Indeed, in 10,000 puzzles tested, only 20 (0.2%) have different W and B ratings. Moreover, ***in spite of non-confluence of the whip resolution theories, the maximum length of whips in a single resolution path using only loopless whips and obtained by the "simplest first" strategy (defined in section 5.5.2 for the B rating) is a good approximation of both the W and B ratings***.

# 7. g-labels, g-candidates, g-whips and g-braids

After introducing the purely structural notion of a "grouped-label" or "g-label", we give a new description of whips of length one. Having g-labels (or, equivalently, whips of length one) is an intrinsic property of a CSP with deep consequences for its resolution theories. When a CSP has g-labels, one can define two new families of resolution rules: g-whips and g-braids, extending the resolution power of whips and braids by allowing the presence of slightly more complex right-linking objects: g-candidates, i.e. groups of candidates related by pre-defined structural relationships, that act locally like the logical "or" of the candidates in the group.

**7.1. g-labels, g-links, g-candidates and whips[1]**

*7.1.1. g-labels and g-links*

*7.1.1.1. General definition of a grouped label (g-label) in a CSP*

Definition: in a CSP, a *potential-g-label* is a pair <V, g>, where V is a CSP variable and g is a set of labels for V, such that:

– the cardinality of g is greater than one, but g is not the full set of labels for V;

– there is at least one label l such that l is not a label for V and l is linked (possibly by different constraints) to all the labels in g.

Definition: a *g-label* is a potential g-label <V, g> that is "saturated" or "locally maximal" in the sense that, for any potential g-label <V, g'> with g' strictly larger than g, there is a label l that is not a label for V and that is linked to all the elements of g but not to all the elements of g'.

Miscellaneous remarks:

– when CSP variable V is clear, we often speak of g-label g, but one must be careful with this abuse of language; (see the Sudoku discussion in section 7.1.1.3);

– as a result of the first condition, a label is not a g-label and there are CSPs with no g-labels;

– one can introduce a new, auxiliary sort: g-Label, with a constant symbol for every g-label and with variable symbols g, g', $g_1$, $g_2$, …;

– the "saturation" or "local maximality" condition plays no role in all our theoretical analyses (in particular, it has no impact on the definition of a g-link); it is there mainly for efficiency reasons; it has the effect of minimising the number of g-labels one must consider when looking for chain patterns built on them; accepting non locally maximal g-labels would increase the computational complexity of the corresponding resolution rules without providing any more generality (as can easily be checked from the definitions of g-whips and g-braids below); for an example where this saturation condition appears as essential from a computational point of view and how it works in more complex cases than Sudoku, see section 15.5 on Kakuro;

– in LatinSquare, there are no g-labels; in Sudoku, all the elements of a g-label are linked to l by constraints of the same type; in N-Queens, there are g-labels but their different elements are always linked to l by two or three constraints of different types (see section 7.8.1); in Kakuro (section 15.5) there are two types of CSP-variables and two corresponding types of g-labels.

*7.1.1.2. g-links*

Definition: a g-label <V, g> and a label l are *g-linked* if l is not a label for V and l is linked to all the elements of g; and we define an auxiliary predicate *g-linked* with signature (g-Label, Label) by:
g-linked(<V, g>, l) ≡ ∀v ¬label(l, V, v) ∧ ∀l'∈g linked(l', l);

Definition: a g-label <V, g> and a label l are *compatible* if they are not g-linked.

Definition: a g-label <V, g> is compatible with a g-label <V', g'> if g contains some label l compatible with <V', g'> . Notice that this is a symmetric relation, in spite of the non symmetric definition (most of the time, we shall use this relation in its apparently non-symmetric form); it is equivalent to: there are some l ∈ g and some l' ∈ g' such that l and l' are not linked.

Definition: a label l [respectively a g-label <V, g>] is compatible with a set S of labels and g-labels if l [resp. <V, g>] is compatible with each element of S.

*7.1.1.3. Grouped labels (g-labels) in Sudoku*

As an example, let us analyse the situation in Sudoku. Informally, a g-label could be defined as the set of labels for a given Number "in" the intersection of a row and a block or "in" the intersection of a column and a block (these are the only possibilities). These intersections are known respectively as row-segments and column-segments (sometimes also as mini-rows and mini-columns).

Then, g-label (n°, r°, $c_{ijk}$) would be the mediator of a symmetric conjugacy relationship between the set of labels (n°, r°, $c°_1$) such that rc-cell (r°, $c°_1$) is in row r° but not in block b° and the set of labels for <variable, value> pairs <b°n°, $s°_2$>

such that rc-cell [b°, s°$_2$] is in block b° but not in row r°. Similarly, if (r$_{ijk}$, c°) = [b°, s$_{pqr}$], then g-label (n°, r$_{ijk}$, c°) would be the mediator of a conjugacy between the set of labels (n°, r°$_1$, c°) such that rc-cell (r°$_1$, c°) is in column c° but not in block b° and the set of labels for <variable, value> pairs <b°n°, s°$_2$> such that rc-cell [b°, s°$_2$] is in block b° but not in column c°.

"Conjugacy", in the above sentences, must be understood in the following sense. When two sets of labels are conjugated via a g-label as above, a proof that all the candidates from one set are impossible leads in an obvious way to a proof that all the candidates from the other set are also impossible. Thus, when one knows that, in row r° [resp. in column c°], number n° can only be in block b°, one can delete n° from all the rc-cells in block b° that are not in row r° [resp. not in column c°]. Conversely, when one knows that, in block b°, number n° can only be in row r° [resp. in column c°], one can delete n° from all the rc-cells in row r° [resp. in column c°] that are not in block b°. These rules are among the most basic ones in Sudoku; they are usually named row-block and column-block interactions (or "locked candidates"). In Sudoku, g-labels correspond to what is also sometimes called "hinges": they are hinges for the conjugacy. As shown in *HLS* (see also the end of section 7.1.2), these ***basic interactions are equivalent to whip[1]***.

Nevertheless, this kind of symmetric conjugacy between two CSP variables is specific to Sudoku. We have chosen to define the notion of a g-label in a much more general way, involving only one CSP variable, so that it can be applied when it is not the "intersection" of two CSP variables and there is no associated symmetric conjugacy relationship. In particular, g-labels in the N-Queens CSP (section 7.8.1) will not be defined by two CSP variables.

According to our formal definition, Sudoku has the following 972 "g-labels":

– for each Row r°, for each Number n°, three g-labels for CSP variable Xr°n°: <Xr°n°, r°n°c123>, <Xr°n°, r°n°c456> and <Xr°n°, r°n°c789>, where:
r°n°c123 is the set of three labels {(n°, r°, c1), (n°, r°, c2), (n°, r°, c3)};
r°n°c456 is the set of three labels {(n°, r°, c4), (n°, r°, c5), (n°, r°, c6)};
r°n°c789 is the set of three labels {(n°, r°, c7), (n°, r°, c8), (n°, r°, c9)};

– for each Column c°, for each Number n°, three g-labels for CSP variable Xc°n°: <Xc°n°, c°n°r123>, <Xc°n°, c°n°r456> and <Xc°n°, c°n°r789>, where:
c°n°r123 is the set of three labels {(n°, c°, r1), (n°, c°, r2), (n°, c°, r3)};
c°n°r456 is the set of three labels {(n°, c°, r4), (n°, c°, r5), (n°, c°, r6)};
c°n°r789 is the set of three labels {(n°, c°, r7), (n°, c°, r8), (n°, c°, r9)};

– for each Block b°, for each Number n°, three g-labels for CSP variable Xb°n°: <Xb°n°, b°n°s123>, <Xb°n°, c°n°s456> and <Xb°n°, c°n°s789>, where:
b°n°s123 is the set of three labels {(n°, b°, s1), (n°, b°, s2), (n°, b°, s3)};
b°n°s456 is the set of three labels {(n°, b°, s4), (n°, b°, s5), (n°, b°, s6)};
b°n°s789 is the set of three labels {(n°, b°, s7), (n°, b°, s8), (n°, b°, s9)};

– for each Block b°, for each Number n°, three g-labels for CSP variable Xb°n°:
<Xb°n°, b°n°s147>, <Xb°n°, c°n°s258> and <Xb°n°, c°n°s369>, where:
b°n°s147 is the set of three labels {(n°, b°, s1), (n°, b°, s4), (n°, b°, s7)};
b°n°s258 is the set of three labels {(n°, b°, s2), (n°, b°, s5), (n°, b°, s8)};
b°n°s369 is the set of three labels {(n°, b°, s3), (n°, b°, s6), (n°, b°, s9)}.

The two groups of g-labels for the Xbn CSP variables may seem redundant with respect to the first two groups: their sets of label triplets are the same as the sets of label triplets related to rows and columns. But they are not considered as g-labels for the same CSP variables. In Sudoku, this difference has always been in implicit existence with the classical distinction between the rules of interaction from blocks to rows (or columns) and rules of interaction from rows (or columns) to blocks, respectively called pointing and claiming (names that are now falling into oblivion).

Contrary to what we did for labels (considering them as equivalence classes of pre-labels), *we do not consider two g-labels as being essentially the same if they have the same sets of labels but different underlying CSP variables*. The reason for this will be clear after the SudoQueens example in section 7.8.3.

### 7.1.2. g-candidates and their correspondence with whips of length one

Definitions: we say that a g-label <V, g> for a CSP variable V is a *g-candidate* for V in a resolution state RS if there are at least two different labels $l_1$ and $l_2$ in g such that $l_1$ and $l_2$ are present as candidates in RS, i.e. RS $\models$ candidate($l_1$) and RS $\models$ candidate($l_2$). Thus, in the same spirit as in the definition of a g-label, we consider that an ordinary candidate is not a g-candidate. The above defined notion of "g-linked" can be extended straightforwardly from g-labels to g-candidates, by considering the complete g-labels underlying the g-candidates. Beware: it is not enough that all the actual candidates be linked; the underlying g-labels must be g-linked). As for "compatibility" between a candidate l and a g-candidate g, it is defined similarly, in terms of the underlying g-label of g, and there is the condition that g must contain at least two candidates compatible with l.

*g-labels act like the logical "or" of several candidates* (but not any combination of any candidates, only structurally fixed combinations for the same CSP variable, predefined by the set of g-labels): in any context in which the true value of V is one of those in the g-candidate, it is not necessary to know precisely which of them is true; one can always conclude that any candidate g-linked to this g-label must be false in this context.

It can also be noticed that g-labels could be used to define two kinds of extended elementary resolution rules (which could be called g-resolution rules, as they deal with g-labels, g-links, g-values and g-candidates in addition to labels, links, values and candidates): gS would assert a g-value predicate for a g-label <V, g> and gECP would eliminate any candidate g-linked to an asserted g-value <V, g>.

But the following remark will lead us further and will require no extension of the notion of a resolution theory. If <V, g> is a g-candidate for V, Z is a candidate g-linked to it and l = <V, x> is any candidate in g, then V{x .} is a whip[1] with target Z. Conversely, for any whip[1]: V{x .} with target Z, there must be at least another value x' for V such that <V, x'> is in g, is still a candidate and is linked to Z (otherwise, the whip would degenerate into a Single, a possibility we have excluded from the definition of a whip); if one defines g as the set of labels for V that are linked to Z, then <V, g> is a g-label for variable V and Z is g-linked to it.

### 7.2. g-bivalue chains, g-whips and g-braids

We now introduce extensions of bivalue chains, whips and braids by allowing the right-linking (but not the left-linking) objects to be either candidates or g-candidates.

Definition: in a resolution state RS, *a g-regular sequence of length n associated with a sequence $(V_1, … V_n)$ of CSP variables* is a sequence of length 2n [or 2n-1] ($L_1$, $R_1$, $L_2$, $R_2$, …. $L_n$, [$R_n$]), such that:
  – for 1≤k≤ n, $L_k$ is a candidate,
  – for 1≤k≤ n [or 1≤k<n], $R_k$ is a candidate or a g-candidate,
  – for each k, $L_k$ has a representative <$V_k$, $l_k$> with $V_k$ and $R_k$ is a candidate or a g-candidate <$V_k$, $r_k$> for $V_k$; this "strong continuity" or "strong g-continuity" (depending on what $R_k$ is) from $L_k$ to $R_k$ implies "continuity" or "g-continuity" (i.e. link or g-link) from $L_k$ to $R_k$.
The $L_k$ are called the *left-linking candidates* of the sequence and the $R_k$ the *right-linking candidates or g-candidates*.

Definition: A *g-regular chain* is a g-regular sequence that satisfies all the additional $R_{k-1}$ to $L_k$ g-continuity conditions: $L_k$ is linked or g-linked to $R_{k-1}$ for all k.

#### 7.2.1. Definition of g-bivalue chains

Definition: in any CSP and in any resolution state RS, given a candidate Z (which will be a target), a *g-bivalue-chain of length n* (n ≥ 1) is a *g-regular chain* ($L_1$, $R_1$, $L_2$, $R_2$, …. $L_n$, $R_n$) associated with a sequence ($V_1$, … $V_n$) of CSP variables, such that:
  – Z is neither equal to any candidate in {$L_1$, $R_1$, $L_2$, $R_2$, …. $L_n$, $R_n$}, nor a member of any g-candidate in this set, for any 1≤k<n;
  – Z is linked to $L_1$;

- $R_1$ is the only candidate or g-candidate for $V_1$ compatible with Z;
- for any $1 < k \leq n$, $R_k$ is the only candidate or g-candidate for $V_k$ compatible with $R_{k-1}$;
- Z is not a label for $V_m$;
- Z is linked or g-linked to $R_m$.

Notice that these conditions imply that Z cannot be a label for any of the CSP variables $V_k$.

***Theorem 7.1 (g-bivalue-chain rule for a general CSP): in any resolution state of any CSP, if Z is a target of a g-bivalue-chain, then it can be eliminated (formally, this rule concludes ¬candidate(Z)).***

Proof: the proof is short and obvious but it will be the basis for the proof of all our forthcoming generalised chain, whip and braid rules including g-labels.

If Z was True, then $L_1$ and all the other candidates for $V_1$ linked to Z would be eliminated by ECP; therefore $R_1$ would have to be or to contain the true value of $V_1$; but then $L_2$ and all the candidates for $V_2$ linked or g-linked to $R_1$ would be eliminated by ECP or $W_1$ and $R_2$ would have to be or to contain the true value of $V_2$....; finally $R_n$ would have to be or to contain the true value of $V_n$; which would contradict the hypothesis that Z was True. Therefore Z can only be False. qed.

Notation: a g-bivalue-chain of length n, together with a potential target elimination, is written symbolically as:
**g-biv-chain[n]: {$L_1$ $R_1$} – {$L_2$ $R_2$} – …… – {$L_n$ $R_n$} ⇒ ¬ candidate(Z)**,
where the curly brackets recall that the two candidates or g-candidates inside have representatives with the same CSP variable.

Re-writing the candidates or g-candidates as <variable, value> or <variable, g-value> pairs and "factoring" the CSP variables out of the pairs, a bivalue chain will also be written symbolically in either of the more explicit forms:
**g-biv-chain[n]: $V_1${$l_1$ $r_1$} – $V_2${$l_2$ $r_2$} – …… – $V_n${$l_n$ $r_n$} ⇒ ¬ candidate(Z)**, or:
**g-biv-chain[n]: $V_1${$l_1$ $r_1$} – $V_2${$l_2$ $r_2$} – …… – $V_n${$l_n$ $r_n$} ⇒ $V_Z \neq v_Z$**.

In spite of the apparently non reversible definition, one has:

***Theorem 7.2: a g-bivalue-chain is reversible***.

Proof: the main point of the proof is the construction of the reversed chain ($L'_1$, $R'_1$, $L'_2$, $R'_2$, …. $L'_n$, $R'_n$). It is based on the reversed sequence of CSP variables and defined as follows (for a similar theorem, see section 9.2.2):
- $L'_k = R_{n-k+1}$ if $R_{n-k+1}$ is a candidate; $L'_k$ = any element in $R_{n-k+1}$ if $R_{n-k+1}$ is a g-candidate; thus, $L'_k$ is always a candidate;

– R'$_k$ = L$_{n-k+1}$ plus all the candidates for V$_{n-k+1}$ that are linked to R$_{n-k}$; thus, R'$_k$ can be a candidate or a g-candidate.

### 7.2.2. Definition of g-whips

Definition: in a resolution state RS, given a candidate Z (which will be the target), a *g-whip* of length n (n ≥ 1) built on Z is a g-regular sequence (L$_1$, R$_1$, L$_2$, R$_2$, …. L$_n$) [notice that there is no R$_n$] associated with a sequence (V$_1$, … V$_n$) of CSP variables, such that:

– Z is neither equal to any candidate in {L$_1$, R$_1$, L$_2$, R$_2$, …. L$_n$} nor a member of any g-candidate in this set;

– L$_1$ is linked to Z;

– for each 1 < k ≤ n, L$_k$ is linked or g-linked to R$_{k-1}$; this is what we call g-continuity from R$_{k-1}$ to L$_k$;

– for any 1 ≤ k < n, R$_k$ is the only candidate or g-candidate for V$_k$ compatible with Z and with all the previous right-linking candidates and g-candidates (i.e. with Z and with all the R$_i$, 1 ≤ i < k);

– Z is not a label for V$_n$;

– V$_n$ has no candidate compatible with Z and with all the previous right-linking candidates and g-candidates (but V$_n$ has more than one candidate).

Notice that left-linking candidates are labels, as in the case of whips; they are not g-labels. Accepting g-labels instead of labels would lead to no added generality but it would entail unnecessary complications. This is the main reason for our restrictive definition of a g-label (i.e. a label is not a g-label).

Definition: as in the cases of bivalue-chains, whips and braids, in any of the above defined g-bivalue chains, g-whips or g-braids, a candidate other than L$_k$ for a CSP variable V$_k$ is called a t- [respectively a z-] candidate if it is incompatible with a previous right-linking candidate or g-candidate [resp. with the target]. And, here again, a candidate can be z- and t- at the same time and that the t- and z- candidates are not considered as being part of the pattern. Notice also that a right-linking g-candidate can contain z- and/or t-candidates, as long as it has more than one non-z and non-t candidate (otherwise, the only compatible candidate is considered as a mere right-linking candidate).

***Theorem 7.3 (g-whip rule for a general CSP): in any resolution state of any CSP, if Z is a target of a g-whip, then it can be eliminated (formally, this rule concludes ¬ candidate(Z)).***

Proof: the proof is a simple adaptation of that for g-bivalue-chains, adding the elimination of all the z-candidates by ECP and, at each step, the elimination of all the next t-candidates by ECP or W$_1$. The end is slightly different: the last condition on the g-whip entails that, if the target Z was True, there would be no possible value for the last variable $V_n$ (because it is not a CSP-Variable for Z).

### 7.2.3. Definition of g-braids

Definition: in a resolution state RS, given a candidate Z (which will be the target), a *g-braid* of length n (n ≥ 1) built on Z is a g-regular sequence ($L_1$, $R_1$, $L_2$, $R_2$, …. $L_n$) [notice that there is no $R_n$] associated with a sequence ($V_1$, … $V_n$) of CSP variables, such that:

– Z is neither equal to any candidate in {$L_1$, $R_1$, $L_2$, $R_2$, …. $L_n$} nor a member of any g-candidate in this set;

– $L_1$ is linked to Z;

– for any 1 < k ≤ n, $L_k$ is either linked to a previous right-linking candidate or to the target or g-linked to a previous right-linking g-candidate; this is the only (but major) structural difference with g-whips (for which the only linking possibility is $R_{k-1}$); the "g-continuity" condition of g-whips is not satisfied by g-braids;

– for any 1 ≤ k < n, $R_k$ is the only candidate or g-candidate for $V_k$ compatible with Z and with all the previous right-linking candidates and g-candidates (i.e. with Z and with all the $R_i$, 1 ≤ i < k);

– Z is not a label for $V_n$;

– $V_n$ has no candidate compatible with Z and with all the previous right-linking candidates and g-candidates (but $V_n$ has more than one candidate).

As in g-whips, left-linking candidates are labels, not g-labels. Here also, accepting g-labels instead of labels would lead to no added generality but it would entail unnecessary complications.

***Theorem 7.4 (g-braids rule for a general CSP): in any resolution state of any CSP, if Z is a target of a g-braid, then it can be eliminated (formally, this rule concludes ¬ candidate(Z)).***

Proof: obvious (almost the same as in the g-whips case).

### 7.2.4. Properties of g-whips and g-braids

g-whips and g-braids have properties very similar to those of whips and braids, namely: linearity, g-continuity (for g-whips), non anticipativeness, left-composability,… In the next sections, we shall see that g-braids also have the two strongest properties of braids: confluence and relationship with gT&E, i.e. T&E($W_1$).

**7.3. g-whip and g-braid resolution theories; the gW and gB ratings**

One can now define two new families of resolution theories and two new ratings, in a way that strictly parallels what was done for whips and braids in chapter 5. As was the case for the W and B ratings, the gW and gB ratings of an instance will be measures of the hardest step in its simplest resolution path with g-whips or g-braids; they will not take into account combinations of steps of the whole path.

*7.3.1. g-whip resolution theories in a general CSP; the gW rating*

Recall that BRT(CSP) is the Basic Resolution Theory of the CSP defined in section 4.3.

Definition: for any $n \geq 0$, let $gW_n$ be the following resolution theory:
– $gW_0$ = BRT(CSP) = $W_0$ = $B_0$,
– $gW_1$ = $gW_0 \cup$ {rules for g-whips of length 1} = $W_1$ (obviously),
– $gW_2$ = $gW_1 \cup$ {rules for g-whips of length 2},
– ....
– $gW_n$ = $gW_{n-1} \cup$ {rules for g-whips of length n},
– $gW_\infty = \cup_{n \geq 0} gW_n$.

Definition : the ***gW-rating*** of an instance P of the CSP, noted gW(P), is the smallest $n \leq \infty$ such that P can be solved within $gW_n$. An instance P has gW rating n if it can be solved using only g-whips of length no more than n but it cannot be solved using only g-whips of length strictly smaller than n. By convention, gW(P) = $\infty$ means that P cannot be solved by g-whips.

The gW rating has some good properties one can expect of a rating:

– it is defined in a purely logical way, independent of any implementation; the gW rating of an instance is an intrinsic property of this instance;

– in the Sudoku case, it is invariant under symmetry and supersymmetry; similar symmetry properties will be true for any CSP, if it has symmetries of any kind and they are properly formalised.

*7.3.2. g-braid resolution theories in a general CSP; the gB rating*

Definition: for any $n \geq 0$, let $gB_n$ be the following resolution theory:
– $gB_0$ = BRT(CSP) = $gW_0$ = $W_0$ = $B_0$,
– $gB_1$ = $gB_0 \cup$ {rules for g-braids of length 1} = $gW_1$ = $W_1$ = $B_1$,
– $gB_2$ = $gB_1 \cup$ {rules for g-braids of length 2},
– ....

- $gB_n = gB_{n-1} \cup \{\text{rules for g-braids of length n}\}$,
- $gB_\infty = \cup_{n \geq 0} gB_n$.

Definition : the ***gB-rating*** of an instance P of the CSP, noted gB(P), is the smallest $n \leq \infty$ such that P can be solved within $gB_n$. An instance P has gB rating n if it can be solved using only g-braids of length no more than n but it cannot be solved using only g-braids of length strictly smaller than n. By convention, gB(P) = $\infty$ means that P cannot be solved by g-braids.

The gB rating has all the good properties one can expect of a rating:

– it is defined in a purely logical way, independent of any implementation; the gB rating of an instance is an intrinsic property of this instance;

– as will be shown in the second next section, it is based on an increasing sequence of theories ($gB_n$) with the confluence property; this ensures *a priori* better computational properties ; in particular, one can define a "simplest first" resolution strategy able to provide the gB rating after following a single resolution path;

– in the Sudoku case, it is invariant under symmetry and supersymmetry ; similar properties will be true for any CSP with symmetries properly formalised.

### 7.4. Comparison of the ratings based on whips, braids, g-whips and g-braids

The first natural question is: how do these two new ratings differ from the W and B ratings associated with ordinary whips and braids? For any CSP, any instance P and any $1 \leq n \leq \infty$, it is obvious that $gB_n(P) \leq \{gW_n(P), B_n(P)\} \leq W_n(P)$, but the relationship between $gW_n(P)$ and $B_n(P)$ is not obvious at all.

Statistically, in Sudoku, there is surprisingly little difference between the four ratings for instances with finite W ratings. Based on 21,371 puzzles generated by the controlled bias generator, only 49 cases with gW(P) < W(P) were found. This is a proportion of 0.23%. In most of these cases, the difference was 1. In 3 cases, the difference was 2. In 1 case, the difference was 5 (see section 7.7.1).

In what follows, as there can be no confusion, we use the same symbol to name a resolution theory T and the set of instances of the CSP solvable in it, i.e. we use T to mean {P / P solvable in T}.

### 7.4.1. *In any CSP, $W_2 = B_2 \subseteq gW_2 = gB_2$*

Theorem 5.5 has shown that $W_2 = B_2$. The proof below will show that $gW_2 = gB_2$. As a result, one has $W_2 = B_2 \subseteq gW_2 = gB_2$.

This is the most one can hope in general: the inclusion $B_2 \subset gW_2$ is strict in Sudoku ($gW_2 \not\subset B_2$), as shown by the counter-example to equality in section 7.7.2. The example in section 7.7.3 will even show that ***$gW_2 \not\subset B_\infty$***.

***Theorem 7.5: In any CSP, any elimination done by a g-braid of length 2 can be done by a g-whip of same or shorter length; as a result, $gB_2 = gW_2$.***

Proof: Let $B = V_1\{l_1 \ r_1\} - V_2\{l_2 \ .\} \Rightarrow V_z \neq v_z$ be a g-braid[2] with target $Z = <V_Z, r_Z>$ in some resolution state RS. If variable $V_2$ has a candidate $<V_2, v'>$ (it may be $<V_2, l_2>$) such that $<V_2, v'>$ is linked or g-linked to $<V_1, r_1>$, then $V_1\{l_1 \ r_1\} - V_2\{v' \ .\} \Rightarrow V_z \neq v_z$ is a g-whip[2] with target Z. Otherwise, $<V_2, l_2>$ is linked to $<V_z, v_z>$ and $V_2\{l_2 \ .\} \Rightarrow V_z \neq v_z$ is a shorter g-whip[1] with target Z.

### 7.4.2. In any CSP, $gW_3 = gB_3$ and therefore $W_3 \subseteq B_3 \subseteq gW_3$

***Theorem 7.6: In any CSP, any elimination done by a g-braid of length 3 can be done by a whip or a g-whip of same or shorter length; as a result, $gB_3 = gW_3$ and $W_3 \subseteq B_3 \subseteq gW_3$.***

Proof: The proof is a little harder than that of $gB_2 = gW_2$. It involves three kinds of changes: 1) re-ordering the various cells; exchanging the roles of left-linking, right-linking and t- objects; exchanging candidates with g-candidates. It is a very good exercise on the manipulation of these notions.

Let $B = V_1\{l_1 \ r_1\} - V_2\{l_2 \ r_2\} - V_3\{l_3 \ .\} \Rightarrow V_z \neq v_z$ be a g-braid[3] in some resolution state RS. We can always suppose that is has been pruned of its useless branches, i.e. of any part $V_k\{l_k \ r_k\}$ such that no candidate for any posterior CSP variable is linked or g-linked to $r_k$. This entails in particular that CSP variable $V_3$ has a candidate linked or g-linked to $<V_2, r_2>$; by modifying B if necessary, we can always suppose it is $<V_3, l_3>$. Then all the other candidates for $V_3$ are linked or g-linked to (at least) one of $<V_Z, v_Z>$, $<V_1, r_1>$ or $<V_2, r_2>$. We now consider two subcases.

1) If CSP variable $V_2$ has at least one candidate linked or g-linked to $<V_1, r_1>$, we can always suppose it is $<V_2, l_2>$ (otherwise, we modify the $l_2$ of the original g-braid). Then B is a g-whip[3] built on target $<V_Z, v_Z>$.

2) Otherwise, all the candidates for $V_2$ other than $<V_2, r_2>$ are linked or g-linked to $<V_Z, v_Z>$; then "$V_2\{l_2 \ r_2\} - V_3\{l_3$ " is a possible beginning for a g-whip built on target $<V_Z, v_Z>$. Moreover, in this case, $V_3$ must have at least one candidate $<V_3, t_3>$ linked or g-linked to $<V_1, r_1>$ (otherwise $V_1\{l_1 \ r_1\}$ would be a useless branch of B and it would have been pruned). If $V_3$ has only one such $t_3$, let $gt_3$ be $t_3$; if $V_3$ has several such $t_3$, they can only belong to a same g-label, say $gt_3$, for $V_3$. Let $r'_1$ be $r_1$ if $r_1$ is a candidate and any candidate in $r_1$ if $r_1$ is a g-candidate. Then the following is a g-whip[3] built on target $<V_Z, v_Z>$: $V_2\{l_2 \ r_2\} - V_3\{l_3 \ gt_3\} - V_1\{r'_1 \ .\}$. qed.

Case 2 is where a tentative proof of $B_3 \subseteq W_3$ along similar lines would fail: in some subcases, we need a right-linking g-candidate gt3, even if B had only right-linking candidates. (Of course, this is not enough to prove that $B_3 \not\subseteq W_3$.)

### 7.4.3. General comparisons

Getting occasionally a lower rating is not the only advantage of having g-whips.

We already mentioned that the inclusion $B_2 \subset gW_2$ is strict in Sudoku (i.e. $gW_2 \not\subseteq B_2$). But we also have the much stronger (*a priori* unexpected) result that $gW_2$ cannot be reduced in general to whips or braids of any length, i.e. **$gW_2 \not\subseteq B_\infty$** (which obviously implies that $gW_\infty \not\subseteq B_\infty$). This will be shown by the example of section 7.7.3. Notice however that such instances will be very exceptional, at least for the Sudoku CSP, as "almost all" the randomly generated puzzles can be solved with whips (see chapter 6).

The simple counter-example in section 7.7.3 is related to the presence in the puzzle of a Sudoku specific pattern, a Swordfish (see chapter 8). The example in section 7.7.4 is much more complex but it shows that even when the Subset patterns are not involved, one can prove that $gW_\infty \not\subseteq B_\infty$ (indeed it shows that $gW_{18} \not\subseteq B_\infty$).

What about the converse? Is $B_\infty \subseteq gW_\infty$? With a puzzle that can be solved by braids (of maximal length 6) but not by g-whips, section 7.7.5 gives a negative answer: $B_\infty \not\subseteq gW_\infty$. What the smallest n such that $B_n \not\subseteq gW_n$ is remains an open question; we only know that $n \leq 6$.

Finally, none of $B_\infty$ and $gW_\infty$ is included in the other.

Now, as a g-whip is a particular case of a g-braid, one has $gW_n \subseteq gB_n$ for all n. But the converse is not true in general, except for $n = 0, 1, 2$ or 3. g-braids are a true generalisation of g-whips. Even in the Sudoku case (for which whips solve almost any puzzle), there seems to be (rare) examples of puzzles that can be solved with g-braids but (probably) not with g-whips: a probable one will appear in section 7.7.6.

## 7.5. The confluence property of the $gB_n$ resolution theories

### 7.5.1. The confluence property of g-braid resolution theories

*Theorem 7.7: each of the $gB_n$ resolution theories, $0 \leq n \leq \infty$, is stable for confluence; therefore it has the confluence property*.

Let n be fixed. We must show that, if an elimination of a candidate Z could have been done in a resolution state $RS_1$ by a g-braid B of length $m \leq n$ and with target Z, it will always still be possible, starting from any further state $RS_2$ obtained from $RS_1$ by consistency preserving assertions and eliminations, if we use a sequence of rules from $gB_n$. Let B be: $\{L_1\ R_1\} - \{L_2\ R_2\} - .... - \{L_p\ R_p\} - \{L_{p+1}\ R_{p+1}\} - ... - \{L_m\ .\}$, with target Z, where the $R_k$'s are candidates or g-candidates modulo Z and the previous $R_i$'s.

The proof follows that for braids in section 5.5, with a few additional subtleties. Consider first the state $RS_3$ obtained from $RS_2$ by applying repeatedly the rules in BRT until quiescence. As BRT has the confluence property, this state is uniquely defined. (Notice that we could legitimately apply rules from $W_1$ instead of only BRT, but this would not guarantee that they do all the eliminations needed in later steps of the following proof).

If, in $RS_3$, target Z has been eliminated, there remains nothing to prove. If target Z has been asserted, then the instance of the CSP is contradictory; if not yet detected in $RS_3$, this contradiction can be detected by CD in a state posterior to $RS_3$, reached by a series of applications of rules from $W_1$, following the g-braid structure of B.

Otherwise, we must consider all the elementary events related to B that can have happened between $RS_1$ and $RS_3$ as well as those we must provoke in posterior resolution states RS. For this, we start from B' = what remains of B in $RS_3$ and we let RS = $RS_3$. At this point, B' may not be a g-braid in RS. We progressively update RS and B' by repeating the following procedure, for p = 1 to p = m, until it produces a new (possibly shorter) g-braid B' with target Z in RS – a situation that is bound to happen. (This is a difference with the braids case: we have to consider a state RS posterior to $RS_3$). We return from this procedure as soon as B' is a g-braid in RS. All the references below are to the current RS and B'.

a) If, in RS, the left-linking or any t- or z- candidate of CSP variable $V_p$ has been asserted (as can be checked on what is done in the other steps, this can only have happened between $RS_1$ and $RS_3$), then all the candidates linked to it have been eliminated by ECP in $RS_3$, in particular: Z and/or the candidate(s) $R_k$ (k<p) to which it is linked and/or all the elements of the g-candidate(s) $R_k$ (k<p) to which it is g-linked; if Z is among them, there remains nothing to prove; otherwise, the procedure has already been successfully terminated by case f1 or f2$\alpha$ of the first such k.

b) If, in RS, left-linking candidate $L_p$ has been eliminated (but not asserted) (it can therefore no longer be used as a left-linking candidate in a g-braid) and if CSP variable $V_p$ still has a z- or a t- candidate $C_p$ (i.e. a candidate $C_p$ linked or g-linked to Z or to some previous $R_i$), then replace $L_p$ by $C_p$. Now, up to $C_p$, B' is a partial g-braid in RS with target Z. Notice that, even if $L_p$ was linked or g-linked to $R_{p-1}$ (e.g. if B was a g-whip) this may not be the case for $C_p$; therefore trying to prove a similar theorem for g-whips would fail here.

c) If, in RS, any t- or z- candidate of $V_p$ has been eliminated (but not asserted), this does not change the basic structure of B (at stage p). Continue with the same B'.

d) If, in RS, right-linking candidate $R_p$ or a candidate $R_p'$ in right-linking g-candidate $R_p$ has been asserted (p can therefore not be the last index of B'), $R_p$ can no longer be used as an element of a g-braid, because it is no longer a candidate or a g-candidate. Contrary to the proof for braids, and only because of this d case, we cannot be sure that this assertion occurred in $RS_3$; we must palliate this. First eliminate by ECP or $W_1$ any left-linking or t- candidate for any CSP variable of B' after p that is incompatible with $R_p$, i.e. linked or g-linked to it, if it is still present in RS. Now, considering the g-braid structure of B upwards from p, more eliminations and assertions can been done by rules from $W_1$. (Notice that we are not trying to do more eliminations or assertions than needed to get a g-braid in RS; in particular, we continue to consider $R_p$, not $R_p'$; in any case, it will be excised from B'; but, most of all, we do not have to find the shortest possible g-braid!)

Let q be the smallest number strictly greater than p such that, in RS, CSP variable $V_q$ still has a (left-linking, t- or z-) candidate $C_q$ that is not linked or g-linked to any of the $R_i$ for $p \leq i < q$ (by definition of a g-braid, $C_q$ is therefore linked or g-linked to Z or to some $R_i$ with $i < p$). Apply the following rules from $W_1$ (if they have not yet been applied between $RS_2$ and RS) for each of the CSP variables $V_u$ of B with index u increasing from p+1 to q-1 included:
- eliminate its left-linking candidate $L_u$ by ECP or $W_1$;
- at this stage, CSP variable $V_u$ had no left-linking, t- or z- candidate;
- if $R_u$ is a candidate, assert it by S and eliminate by ECP all the left-linking and t-candidates for CSP variables after u that are incompatible with $R_u$ in the current RS;
- if $R_u$ is a g-candidate, it cannot be asserted by S; eliminate by $W_1$ all the left-linking and t- candidates for CSP variables after u that are incompatible with $R_u$ in the current RS.

In the new RS thus obtained, excise from B' the part related to CSP variables p to q-1 (included) and, if $L_q$ has been eliminated in the passage from $RS_2$ to RS, replace it by $C_q$; for each integer $s \geq p$, decrease by q-p the index of CSP variable $V_s$ and of its candidates and g-candidates in the new B'. In RS, B' is now, up to p (the ex q), a partial g-braid in $gB_n$ with target Z.

e) If, in RS, left-linking candidate $L_p$ has been eliminated (but not asserted), and if CSP variable $V_p$ has no t- or z- candidate in RS (complementary to case b), then there are now two cases ($V_p$ must have at least one candidate).

e1) If $R_p$ is a candidate, then $V_p$ has only one possible value, namely $R_p$. If $R_p$ has not yet been asserted by S somewhere between $RS_2$ and RS, do it now; this case is now reducible to case d (because the assertion of $R_p$ also entails the elimination of $L_p$); go back to case d for the same value of p (this does not introduce an infinite loop!). Otherwise, go to the next p.

e2) If $R_p$ is a g-candidate, then $R_p$ cannot be asserted by S; use it, for any CSP variable after p, to eliminate by $W_1$ any of its t-candidates that is g-linked to $R_p$. Let q be the smallest number strictly greater than p such that, in RS, CSP variable $V_q$ still has a (left-linking, t- or z-) candidate $C_q$ that is not linked or g-linked to any of the $R_i$ for $p \leq i < q$. Replace RS by the state obtained after all the assertions and eliminations similar to those in case d above have been done. Then, in RS, excise the part of B' related to CSP variables p to q-1 (included), replace $L_q$ by $C_q$ (if $L_q$ has been eliminated in the passage from $RS_2$ to $RS_3$) and re-number the posterior elements of B', as in case d. In RS, B' is now, up to p (the ex q), a partial g-braid in $gB_n$ with target Z. If p is its last index, it is a g-braid; return it and stop.

f) Finally, consider eliminations occurring in a right-linking candidate or g-candidate $R_p$. This implies that p cannot be the last index of B'. There are two cases.

f1) If, in RS, right-linking candidate $R_p$ of B has been eliminated (but not asserted) or marked (by f2γ) in a previous step (i.e. it has become a t-candidate), then replace B' by its initial part: $\{L_1\ R_1\} - \{L_2\ R_2\} - \ldots - \{L_p\ .\}$. At this stage, B' is in RS a (possibly shorter) g-braid with target Z. Return B' and stop.

f2) If, in RS, a candidate in right-linking g-candidate $R_p$ has been eliminated (but not asserted) or marked in a previous step, then:

f2α) either there remains no unmarked candidate of $R_p$ in RS; then replace B' by its initial part: $\{L_1\ R_1\} - \{L_2\ R_2\} - \ldots - \{L_p\ .\}$; at this stage, B' is in RS a (possibly shorter) g-braid with target Z; return B' and stop;

f2β) or the remaining unmarked candidates of $R_p$ in RS still make a g-candidate and B' does not have to be changed;

f2γ) or there remains only one unmarked candidate $R_p$' of $R_p$; replace $R_p$ by $R_p$' in B'. We must also prepare the next steps by putting marks. Any t-candidate of B that was g-linked to $R_p$, if it is still present in RS, can still be considered as a t-candidate in B', where it is now linked to $R_p$' instead of being g-linked to $R_p$; this does not raise any problem. However, this substitution may entail that candidates that were not t-candidates in B become t-candidates in B'; if they are left-linking candidates of B', this is not a problem either; but if any of them is a right-linking candidate or an element of a right-linking g-candidate for B', then mark it so that the same procedure (i.e. f1 or f2) can be applied to it in a later step.

Notice that, as was the case for braids, this proof works only because the notions of being linked and g-linked do not depend on the resolution state.

### 7.5.2. g-braid resolution strategies consistent with the gB rating

As explained in section 4.5.3 and in exactly the same way as in the braids case, we can take advantage of the confluence property of g-braid resolution theories to define a "simplest firth" strategy that will always find the simplest (in terms of the length of the g-braids it will use) solution after following a single resolution path. As a result, it will also compute the gB rating of an instance. The following order satisfies this requirement:
ECP > S >
biv-chain[1] > whip[1] > g-whip[1] > braid[1] > g-braid[1] >
… > …
biv-chain[k] > whip[k] > g-whip[k] > braid[k] > g-braid[k] >
biv-chain[k+1] > whip[k+1] > g-whip[k+1] > braid[k+1] > g-braid[k+1] > …

Notice that bivalue-chains, whips, g-whips and braids being special cases of g-braids of same length, their explicit presence in the set of rules does not change the final result (z-chains and t-whips could also be added in the landscape). We put them here because when we look at a resolution path, it may be nicer to see simple patterns appear instead of more complex ones (g-braids). Also, it allows to see (in the Sudoku case) that, in practice, g-braids that are neither g-whips nor braids do not appear very often in the resolution paths.

Here, we have put g-whips before braids of same length, because they are structurally simpler and experiments confirm this complexity hierarchy (in terms of computation times and memory requirements). This choice has no impact on the gB rating.

As in the case of ordinary braids, the above ordering does not completely define a deterministic procedure: it does not set any precedence between different chains of same type and length. This could be done by using an ordering of the candidates instantiating them, based e.g. on their lexicographic order. But, here again, one can also decide that, for all practical purposes, which of these equally prioritised rule instantiations should be "fired" first should be chosen randomly (as in the default behaviour of CSP-Rules).

### 7.6. The "gT&E vs g-braids" theorem

In section 5.6.1, we defined the procedure T&E(T, Z, RS) for any candidate Z, any resolution state RS and any resolution theory T with the confluence property. In this section, we consider $T = W_1 = B_1$ and we set gT&E = T&E($W_1$). It is obvious that any elimination that can be done by a g-braid B can be done by gT&E, using a sequence of rules from $B_1 = W_1$, following the structure of B. The converse is more interesting:

***Theorem 7.8: for any instance of any CSP, any elimination that can be done by gT&E can be done by a g-braid. Any instance of a CSP that can be solved by gT&E can be solved by g-braids.***

Proof: Let RS be a resolution state and let Z be a candidate eliminated by gT&E(Z, RS) using some auxiliary resolution state RS'. Following the steps of resolution theory $B_1$ in RS', we progressively build a g-braid in RS with target Z. But we must do this in a little smarter way than in our proof for mere braids. First, remember that $B_1$ contains only four types of rules: ECP (which eliminates candidates), S (which asserts a value for a CSP variable), $W_1$ (whips of length 1, which eliminates candidates) and CD (which detects a contradiction on a CSP variable).

Consider the sequence $(P_1, P_2, …, P_k, …P_n)$ of rule applications in RS' based on rules from $W_1$ different from ECP and suppose that $P_n$ is the first occurrence of CD (there must be at least one occurrence of CD if Z is eliminated by gT&E). We first define the $R_k$ and $V_k$ sequences; starting from empty $R_k$ and $V_k$, for k = 1 to n-1:
- if $P_k$ is of type S, then it asserts a value $R_k$ for some CSP variable $V_k$; add $R_k$ and $V_k$ at the end of the appropriate sequences;
- if $P_k$ is of type whip[1]: $\{M_k .\} \Rightarrow \neg candidate(C_k)$ for some CSP variable $V_k$, then define $R_k$ as the g-candidate for $V_k$ that contains $M_k$ and is g-linked to $C_k$; (notice that $C_k$ will not necessarily be $L_{k+1}$); add $R_k$ and $V_k$ to the appropriate sequences.

We shall build a g-braid[n] in RS with target Z, with the $R_k$'s as its sequence of right-linking candidates or g-candidates and with the $V_k$'s as its sequence of first n-1 CSP variables. We only have to define properly the $L_k$'s. We do this successively for k = 1, …, k = n. As the proofs for k = 1 and for the passage from k to k+1 are almost identical, we skip the case k = 1. Suppose we have done it until k and consider CSP variable $V_{k+1}$.

Whatever rule $P_{k+1}$ is (S or whip[1]), the fact that it can be applied means that, apart from $R_{k+1}$ (if it is a candidate) or the labels contained in $R_{k+1}$ (if it is a g-candidate), all the other labels for CSP variable $V_{k+1}$ that were still candidates for $V_{k+1}$ in RS (and there must be at least one, say $L_{k+1}$) have been eliminated in RS' by the assertion of Z and the previous rule applications. But these previous eliminations can only result from being linked or g-linked to Z or to some $R_i$, i≤k. $\{L_{k+1} R_{k+1}\}$ is therefore a legitimate extension for our partial g-braid.

End of the procedure: at step n, a contradiction is obtained by CD for some variable $V_n$. It means that all the candidates for $V_n$ that were still candidates for $V_n$ in RS (and there must be at least one, say $L_n$) have been eliminated in RS' by the assertion of Z and the previous rule applications. But these previous eliminations can only result from being linked or g-linked to Z or to some $R_i$, i<n. $L_n$ is thus the last left-linking candidate of the g-braid we were looking for in RS.

Here again (as in the proof of confluence), this proof works only because the existence of a link or a g-link between two candidates does not depend on the resolution state. And, again, it is very unlikely that the gT&E procedure followed by the construction in this proof would produce the shortest available g-braid in RS.

**7.7. Exceptional examples**

This section provides the proofs by examples announced in section 7.4.

*7.7.1. A puzzle with W=B=7 and gW=2*

In section 7.4, we mentioned the rare case of a puzzle P with finite W rating but with very different W and gW ratings: gW(P) = W(P) - 5. One might think that this can happen only for hard puzzles, but the example in Figure 7.1 shows that it can also happen with relatively simple ones: here, we have gW(P) = 2 and W(P) = 7.

| 1 |   |   |   | 6 |   |   |   | 9 |
|---|---|---|---|---|---|---|---|---|
|   |   |   |   |   |   |   |   |   |
|   |   | 8 |   | 1 |   |   | 5 | 6 |
|   |   | 3 |   | 6 |   |   | 8 |   | 1 |
|   |   |   |   |   |   |   |   | 7 |
|   |   |   |   |   | 8 | 2 | 4 | 6 |
|   |   | 4 |   | 9 |   |   |   | 2 |
| 5 | 6 |   | 8 | 4 | 1 |   |   |   |
|   |   | 7 |   |   | 3 |   |   |   |

| 1 | 2 | 3 | 4 | 5 | 6 | 7 | 8 | 9 |
|---|---|---|---|---|---|---|---|---|
| 4 | 5 | 6 | 7 | 8 | 9 | 1 | 3 | 2 |
| 7 | 8 | 9 | 1 | 3 | 2 | 5 | 6 | 4 |
| 2 | 3 | 4 | 6 | 9 | 7 | 8 | 5 | 1 |
| 6 | 1 | 8 | 5 | 2 | 4 | 3 | 9 | 7 |
| 9 | 7 | 5 | 3 | 1 | 8 | 2 | 4 | 6 |
| 3 | 4 | 1 | 9 | 7 | 5 | 6 | 2 | 8 |
| 5 | 6 | 2 | 8 | 4 | 1 | 9 | 7 | 3 |
| 8 | 9 | 7 | 2 | 6 | 3 | 4 | 1 | 5 |

**Figure 7.1.** *Puzzle P (cb#41065 ) with W(P)=B(P)=7 and gW(P)=2*

1) If we accept g-whips, there is a very short resolution path:

```
***** SudoRules 16.2 based on CSP-Rules 1.2, config: gW *****
26 givens and 206 candidates, 1339 csp-links and 1339 links. Initial density = 1.59
singles ==> r8c9 = 3, r8c3 = 2
whip[1]: c9n5{r9 .} ==> r9c8 ≠ 5
whip[1]: r8n9{c7 .} ==> r9c8 ≠ 9, r9c7 ≠ 9,
whip[1]: r8n7{c8 .} ==> r7c7 ≠ 7
whip[1]: r6n3{c5 .} ==> r5c4 ≠ 3, r5c5 ≠ 3
;;; Resolution state RS₁
whip[2]: b4n6{r5c1 r5c3} – b4n8{r5c3 .} ==> r5c1 ≠ 4, r5c1 ≠ 2
whip[2]: b4n8{r5c1 r5c3} – b4n6{r5c3 .} ==> r5c1 ≠ 9
whip[2]: b4n6{r5c3 r5c1} – b4n8{r5c1 .} ==> r5c3 ≠ 5, r5c3 ≠ 4
whip[1] : r5n4{c6 .} ==> r4c6 ≠ 4
```

whip[2]: b4n6{r5c3 r5c1} – b4n8{r5c1 .} ==> r5c3 ≠ 1
whip[2]: b4n8{r5c3 r5c1} – b4n6{r5c1 .} ==> r5c3 ≠ 9
;;; Resolution state RS$_2$
**g-whip[2]: r3n7{c1 c456} – c4n7{r2 .} ==> r6c1 ≠ 7**
singles to the end

2) If we accept only whips, the resolution path is much longer:

***** SudoRules 16.2 based on CSP-Rules 1.2, config: W *****
;;; same path up to resolution stateRS$_2$
whip[3]: c4n7{r1 r6} – c2n7{r6 r2} – r3n7{c1 .} ==> r1c5 ≠ 7
whip[3]: c2n7{r1 r6} – c4n7{r6 r1} – r3n7{c6 .} ==> r2c1 ≠ 7
whip[3]: c4n7{r2 r6} – c2n7{r6 r1} – r3n7{c1 .} ==> r2c5 ≠ 7, r2c6 ≠ 7
whip[3]: r5n2{c6 c2} – r1n2{c2 c4} – b8n2{r9c4 .} ==> r4c5 ≠ 2
whip[3]: b4n7{r6c2 r4c1} – r3n7{c1 c6} – b8n7{r7c6 .} ==> r6c5 ≠ 7
whip[3]: r9c2{n1 n9} – r9c1{n9 n8} – r9c8{n8 .} ==> r9c7 ≠ 1
whip[3]: r9c8{n8 n1} – r9c2{n1 n9} – r9c1{n9 .} ==> r9c9 ≠ 8
whip[5]: r4n2{c1 c6} – r4n7{c6 c5} – b8n7{r7c5 r7c6} – r3n7{c6 c1} – r6c1{n7 .} ==> r4c1 ≠ 9
whip[5]: r4c8{n9 n5} – r4c5{n5 n7} – b8n7{r7c5 r7c6} – r3n7{c6 c1} – r6c1{n7 .} ==> r4c3 ≠ 9
whip[3]: r9c2{n9 n1} – c3n1{r7 r6} – c3n9{r6 .} ==> r2c2 ≠ 9
whip[6]: b3n1{r2c7 r2c8} – r9c8{n1 n8} – r9c1{n8 n9} – r6c1{n9 n7} – b1n7{r3c1 r1c2} – c4n7{r1 .} ==> r2c7 ≠ 7
**whip[7]: c3n6{r2 r5} – c1n6{r5 r2} – c1n4{r2 r4} – c1n2{r4 r3} – b3n2{r3c9 r2c9} – c9n8{r2 r7} – c3n8{r7 .} ==> r2c3 ≠ 4**
whip[7]: r3n3{c1 c5} – c3n3{r3 r7} – b7n1{r7c3 r9c2} – b7n9{r9c2 r9c1} – r6c1{n9 n7} – r3n7{c1 c6} – c4n7{r1 .} ==> r2c1 ≠ 3
whip[7]: r1n2{c5 c2} – c1n2{r2 r4} – c6n2{r4 r5} – r2n2{c6 c9} – r3c9{n2 n4} – c6n4{r3 r2} – c1n4{r2 .} ==> r3c5 ≠ 2
whip[7]: r4n2{c1 c6} – r4n7{c6 c5} – b8n7{r7c5 r7c6} – r3n7{c6 c1} – c1n2{r3 r2} – r1c2{n2 n5} – r2c2{n5 .} ==> r4c1 ≠ 4
hidden-single-in-a-block ==> r4c3 = 4
whip[4]: b5n4{r5c6 r5c4} – r1n4{c4 c7} – c7n7{r1 r8} – c7n9{r8 .} ==> r5c6 ≠ 9
whip[4]: r3c3{n9 n3} – r3c5{n3 n7} – c4n7{r1 r6} – r6c1{n7 .} ==> r3c1 ≠ 9
whip[5]: b3n7{r1c7 r2c8} – c8n1{r2 r9} – r7c7{n1 n6} – r9c7{n6 n4} – r1n4{c7 .} ==> r1c4 ≠ 7
;;; only now do we get the crucial elimination with a whip[2]:
**whip[2]: c4n7{r6 r2} – r3n7{c6 .} ==> r6c1 ≠ 7**
singles to the end

3) Interestingly (anticipating on chapter 8), this puzzle can also be solved with Subset rules (of size 3), but it gets a higher rating (S=3) than with g-whips (gW=2); i.e. g-whips are better than Subsets in this case.

***** SudoRules 16.2 based on CSP-Rules 1.2, config: gW+S *****
;;; same path up to resolution stateRS$_1$
hidden-pairs-in-a-row: r5{n6 n8}{c1 c3} ==> r5c3 ≠ 9, r5c3 ≠ 5, r5c3 ≠ 4, r5c3 ≠ 1, r5c1 ≠ 9, r5c1 ≠4
whip[1] : r5n4{c6 .} ==> r4c6 ≠ 4
hidden-pairs-in-a-row: r5{n6 n8}{c1 c3} ==> r5c1 ≠ 2

;;; same situation as $RS_2$ (all the whips[2] in the W or gW resolution paths are hidden pairs)
naked-triplets-in-a-row: r9{c1 c2 c8}{n8 n9 n1} ==> r9c9 ≠ 8, r9c7 ≠ 1
swordfish-in-rows: n7{r3 r4 r7}{c6 c1 c5} ==> r6c5 ≠ 7
;;; The crucial elimination is now obtained with a swordfish:
**swordfish-in-rows: n7{r3 r4 r7}{c6 c1 c5} ==> r6c1 ≠ 7**
singles to the end

### 7.7.2. $gW_2 \not\subset B_2$: a puzzle with W=3, B=3, gW=2, gB=2

Our second example (puzzle cb#1249 in Figure 7.2) proves that the obvious inclusion $B_2 \subset gW_2$ is not an equality in general ("obvious" because $B_2 = W_2$).

1) The resolution path with g-whips gives gW(P) = 2:

```
***** SudoRules 16.2 based on CSP-Rules 1.2, config: gW *****
27 givens, 200 candidates, 1254 csp-links and 1254 links. Initial density = 1.58
singles ==> r7c3 = 2, r1c2 = 2, r2c2 = 5, r2c4 = 7, r1c5 = 5, r2c6 = 9, r3c5 = 3, r4c7 = 9, r5c9 = 2
whip[1]: c9n1{r9 .} ==> r7c7 ≠ 1, r8c8 ≠ 1, r9c7 ≠ 1, r9c8 ≠ 1
whip[1]: c2n1{r9 .} ==> r7c1 ≠ 1, r8c1 ≠ 1, r9c1 ≠ 1
whip[1]: r4n6{c8 .} ==> r6c8 ≠ 6, r6c7 ≠ 6
whip[1]: r4n8{c8 .} ==> r5c8 ≠ 8, r5c7 ≠ 8
whip[2]: r4c6{n3 n5} – r4c4{n5 .} ==> r4c8 ≠ 3
whip[1]: r4n3{c4 .} ==> r5c6 ≠ 3
whip[2]: r4c6{n5 n3} – r4c4{n3 .} ==> r4c8 ≠ 5, r4c9 ≠ 5
singles ==> r6c8 = 5, r6c4 = 2, r9c5 = 2
;;; Resolution state RS₁
```
**g-whip[2]: c7n8{r1 r789} – r8n8{c9 .} ==> r1c1 ≠ 8**
singles to the end

|   | 3 | 4 |   | 6 |   |   |   | 9 |
|---|---|---|---|---|---|---|---|---|
|   |   |   | 8 |   |   |   | 2 | 3 |
| 7 |   |   | 1 |   | 2 | 5 |   |   |
| 2 | 7 | 1 |   | 4 |   |   |   |   |
| 5 |   |   | 6 |   |   |   |   |   |
|   | 3 |   |   |   | 8 |   |   | 7 |
|   |   |   |   |   |   | 9 |   |   |
|   | 4 | 5 | 9 |   |   | 2 |   |   |
|   |   | 7 |   |   | 4 |   |   |   |

| 1 | 2 | 3 | 4 | 5 | 6 | 7 | 8 | 9 |
|---|---|---|---|---|---|---|---|---|
| 4 | 5 | 6 | 7 | 8 | 9 | 1 | 2 | 3 |
| 7 | 8 | 9 | 1 | 3 | 2 | 5 | 4 | 6 |
| 2 | 7 | 1 | 3 | 4 | 5 | 9 | 6 | 8 |
| 5 | 8 | 4 | 6 | 9 | 7 | 3 | 1 | 2 |
| 6 | 3 | 9 | 2 | 1 | 8 | 4 | 5 | 7 |
| 3 | 6 | 2 | 5 | 7 | 1 | 8 | 9 | 4 |
| 8 | 4 | 5 | 9 | 6 | 3 | 2 | 7 | 1 |
| 9 | 1 | 7 | 8 | 2 | 4 | 6 | 3 | 5 |

***Figure 7.2.** A puzzle P (cb #1249 ) with gW(P)=2 and W(P)=B(P)=3*

2) The resolution path with whips gives W(P) = 3; the resolution path with braids is exactly the same, i.e. no non-whip braid appears in it, and B(P) = 3:

***** SudoRules 16.2 based on CSP-Rules 1.2, config: W *****
;;; same path up to resolution stateRS$_1$
whip[3]: b1n1{r1c1 r2c1} – c1n4{r2 r6} – r6c7{n4 .} ==> r1c7 ≠ 1
whip[3]: c7n6{r7 r2} – b3n1{r2c7 r1c8} – c8n7{r1 .} ==> r8c8 ≠ 6
whip[3]: c1n4{r2 r6} – r6c7{n4 n1} – r2n1{c7 .} ==> r2c1 ≠ 6
whip[3]: r2c3{n4 n6} – b4n6{r6c3 r6c1} – c1n4{r6 .} ==> r2c7 ≠ 4
whip[1]: r2n4{c1 .} ==> r3c3 ≠ 4
whip[3]: r2c7{n6 n1} – r6c7{n1 n4} – c8n4{r5 .} ==> r3c8 ≠ 6
whip[3]: b6n8{r4c9 r4c8} – r3c8{n8 n4} – c9n4{r3 .} ==> r7c9 ≠ 8
whip[3]: r2c3{n4 n6} – r2c7{n6 n1} – r6c7{n1 .} ==> r6c3 ≠ 4
whip[3]: b1n9{r3c2 r3c3} – b1n6{r3c3 r2c3} – r6c3{n6 .} ==> r3c2 ≠ 8
whip[3]: b1n8{r1c1 r3c3} – b1n9{r3c3 r3c2} – b7n9{r9c2 .} ==> r9c1 ≠ 8
whip[3]: c1n8{r8 r1} – c7n8{r1 r7} – r8n8{c9 .} ==> r9c2 ≠ 8
whip[3]: b7n8{r7c1 r7c2} – c7n8{r7 r9} – r8n8{c8 .} ==> r1c1 ≠ 8
singles to the end

### 7.7.3. $gW_2 \not\subset B_\infty$: a puzzle not solvable by braids of any length but solvable in $gW_2$

The example in Figure 7.3 (a puzzle from Mauricio's swordfish collection) allows to go much further: it proves that $gW_2 \not\subset B_\infty$ and therefore $gW_\infty \not\subset B_\infty$.

|   | 1 |   | 2 |   |   | 3 |   |   |
|---|---|---|---|---|---|---|---|---|
|   |   | 1 |   |   | 4 |   |   |   |
| 2 |   | 4 |   | 5 |   |   |   |   |
|   | 6 |   |   | 7 |   |   | 8 |   |
|   | 5 |   |   |   | 2 |   |   |   |
| 9 |   | 3 |   | 4 |   |   |   |   |
|   | 8 |   | 1 |   |   |   | 5 |   |
|   | 9 |   | 6 |   |   |   |   |   |
| 1 |   | 9 |   | 7 |   |   |   |   |

| 6 | 4 | 1 | 5 | 9 | 2 | 8 | 7 | 3 |
|---|---|---|---|---|---|---|---|---|
| 8 | 7 | 5 | 6 | 1 | 3 | 2 | 4 | 9 |
| 2 | 3 | 9 | 4 | 7 | 8 | 5 | 6 | 1 |
| 3 | 1 | 6 | 2 | 4 | 7 | 9 | 5 | 8 |
| 7 | 5 | 4 | 1 | 8 | 9 | 3 | 2 | 6 |
| 9 | 8 | 2 | 3 | 5 | 6 | 4 | 1 | 7 |
| 4 | 2 | 8 | 7 | 3 | 1 | 6 | 9 | 5 |
| 5 | 9 | 7 | 8 | 6 | 4 | 1 | 3 | 2 |
| 1 | 6 | 3 | 9 | 2 | 5 | 7 | 8 | 4 |

***Figure 7.3.*** *A puzzle P with B(P)=∞ but gW(P)=2*

Using the T&E procedure and the "T&E vs braids" theorem, it is easy to check that this puzzle is not solvable by braids, let alone by whips. But it is in gT&E and it can therefore be solved by g-braids. Let us try to do better and solve it by g-whips.

***** SudoRules 16.2 based on CSP-Rules 1.2, config: gW *****
24 givens, 214 candidates, 1289 csp-links and 1289 links. Initial density = 1.41
**g-whip[2]: c3n4{r5 r789} – r7n4{c2 .} ==> r5c5 ≠ 4**
**g-whip[2]: r1n9{c5 c789} – c9n9{r3 .} ==> r5c5 ≠ 9**
singles to the end

Anticipating on chapter 8, this puzzle can also be solved by Subsets of size 3, more precisely by Swordfish; actually, we find two Swordfish (for two different numbers) in the same three columns, a very exceptional situation. This puzzle will also count as a very rare example of a Swordfish not completely subsumed by whips.

***** SudoRules 16.2 based on CSP-Rules 1.2, config: B+S *****
24 givens, 214 candidates, 1289 csp-links and 1289 links. Initial density = 1.41
swordfish-in-columns n4{c3 c6 c9}{r9 r5 r8} ==> r9c5 ≠ 4, r9c2 ≠ 4, r8c1 ≠ 4, r5c5 ≠ 4, r5c1 ≠ 4
swordfish-in-columns n9{c3 c6 c9}{r3 r2 r5} ==> r5c7 ≠ 9, r5c5 ≠ 9  ; singles to the end

### 7.7.4. $gW_\infty \not\subset B_\infty$: a puzzle not solvable by braids of any length but solvable in $gW_{18}$

Even without invoking puzzles, as in section 7.7.3, involving the rare case of a Subset pattern that is not subsumed by whips or braids, there are examples that can be solved by g-whips but not by braids. Consider the puzzle (created by Arto Inkala) shown in Figure 7.4 (and known as "AI Broken Brick").

Using the T&E procedure and the "T&E vs braids" theorem, it is easy to check that this puzzle is not solvable by T&E and it has therefore no chance of being solvable by braids, let alone by whips. But it is solvable by gT&E and it can therefore be solved by g-braids. Let us try to do better and solve it by g-whips.

| 4 |   |   | 6 |   |   | 7 |   |   |
|---|---|---|---|---|---|---|---|---|
|   |   |   |   |   | 6 |   |   |   |
|   | 3 |   |   | 2 |   |   |   | 1 |
| 7 |   |   |   | 8 | 5 |   |   |   |
|   | 1 |   | 4 |   |   |   |   |   |
|   | 2 |   | 9 | 5 |   |   |   |   |
|   |   |   |   |   | 7 |   |   | 5 |
|   |   | 9 | 1 |   |   | 3 |   |   |
|   |   | 3 |   | 4 |   |   | 8 |   |

| 4 | 5 | 1 | 8 | 6 | 3 | 9 | 7 | 2 |
|---|---|---|---|---|---|---|---|---|
| 9 | 8 | 2 | 7 | 1 | 4 | 6 | 5 | 3 |
| 6 | 3 | 7 | 5 | 9 | 2 | 8 | 4 | 1 |
| 7 | 9 | 6 | 3 | 2 | 8 | 5 | 1 | 4 |
| 3 | 1 | 5 | 4 | 7 | 6 | 2 | 9 | 8 |
| 8 | 2 | 4 | 9 | 5 | 1 | 3 | 6 | 7 |
| 1 | 4 | 8 | 6 | 3 | 9 | 7 | 2 | 5 |
| 2 | 7 | 9 | 1 | 8 | 5 | 4 | 3 | 6 |
| 5 | 6 | 3 | 2 | 4 | 7 | 1 | 8 | 9 |

**Figure 7.4.** Puzzle "AI Broken Brick" with $B(P)=\infty$ and $gW(P)=18$

***** SudoRules 16.2 based on CSP-Rules 1.2, config: gW *****
23 givens, 219 candidates, 1366 csp-links and 1366 links. Initial density = 1.43
hidden-single-in-a-column ==> r2c6 = 4
whip[1] : c3n7{r3 .} ==> r2c2 ≠ 7
whip[4]: b4n5{r5c3 r5c1} – b4n9{r5c1 r4c2} – r2c2{n9 n8} – r1c2{n8 .} ==> r3c3 ≠ 5, r2c3 ≠ 5, r1c3 ≠ 5
hidden-single-in-a-column ==> r5c3 = 5
whip[9]: r6n7{c9 c6} – b5n1{r6c6 r4c5} – r4n3{c5 c4} – b5n6{r4c4 r5c6} – r8c6{n6 n5} – r9c6{n5 n9} – r7n9{c6 c8} – r5c8{n9 n2} – r4n2{c8 .} ==> r6c9 ≠ 3
whip[10]: b1n6{r3c1 r3c3} – r4c3{n6 n4} – r6c3{n4 n8} – r6c1{n8 n3} – b4n6{r6c1 r4c2} – c4n6{r4 r7} – b9n6{r7c8 r8c9} – c9n4{r8 r6} – r6c7{n4 n1} – r9n1{c7 .} ==> r9c1 ≠ 6

whip[13]: b1n6{r3c1 r3c3} – r4c3{n6 n4} – r6c3{n4 n8} – r6c1{n8 n3} – b4n6{r6c1 r4c2} – c4n6{r4 r9} – r8n6{c6 c9} – c9n4{r8 r6} – r6c7{n4 n1} – c6n1{r6 r1} – r1c3{n1 n2} – r7c3{n2 n1} – r9n1{c1 .} ==> r7c1 ≠ 6

whip[13]: c6n5{r8 r1} – r1n1{c6 c3} – r2n1{c1 c5} – b5n1{r4c5 r6c6} – c7n1{r6 r9} – r9c1{n1 n2} – b1n2{r2c1 r2c3} – b1n7{r2c3 r3c3} – b1n6{r3c3 r3c1} – r3n5{c1 c8} – r2c8{n5 n9} – c1n9{r2 r5} – c7n9{r5 .} ==> r9c4 ≠ 5

whip[1]: c4n5{r1 .} ==> r1c6 ≠ 5

whip[15]: r4c3{n6 n4} – r6c3{n4 n8} – r3c3{n8 n7} – b1n6{r3c3 r3c1} – b4n6{r5c1 r4c2} – c4n6{r4 r9} – r8n6{c6 c9} – c9n4{r8 r6} – r6n7{c9 c6} – r8c6{n7 n5} – r9c6{n5 n9} – r7n9{c6 c8} – b9n1{r7c8 r9c7} – r6n1{c7 c8} – r6n6{c8 .} ==> r7c3 ≠ 6

g-whip[15]: c7n1{r9 r6} – c6n1{r6 r1} – c5n1{r2 r4} – c8n1{r4 r7} – b9n9{r7c8 r9c9} – c6n9{r9 r7} – c6n3{r7 r456} – r4n3{c4 c9} – b3n3{r2c9 r1c7} – r1n9{c7 c2} – r4n9{c2 c8} – r2n9{c8 c5} – b2n3{r2c5 r2c4} – r2n7{c4 c3} – c3n1{r2 .} ==> r9c7 ≠ 2

**g-whip[18]: r4n1{c8 c5} – b2n1{r2c5 r1c6} – c6n9{r1 r789} – r7n9{c5 c6} – c6n3{r7 r456} – r4n3{c4 c9} – r4n2{c9 c4} – r5c5{n2 n7} – b6n7{r5c9 r6c9} – c9n4{r6 r8} – r8c7{n4 n2} – r8c5{n2 n8} – r3c5{n8 n9} – r2c5{n9 n3} – r1n3{c6 c7} – c7n9{r1 r9} – r9n1{c7 c1} – r9n2{c1 .} ==> r4c8 ≠ 9**

whip[5]: r4n9{c9 c2} – r1n9{c2 c6} – c6n1{r1 r6} – c7n1{r6 r9} – r9n9{c7 .} ==> r2c9 ≠ 9

g-whip[10]: c5n1{r2 r4} – c6n1{r6 r1} – b2n3{r1c6 r123c4} – r4n3{c4 c9} – r4n9{c9 c2} – r2c2{n9 n5} – r1c2{n5 n8} – r1n9{c2 c789} – r2c8{n9 n2} – r2c9{n2 .} ==> r2c5 ≠ 8

g-whip[14]: b9n1{r9c7 r7c8} – r4n1{c8 c5} – b2n1{r2c5 r1c6} – c6n9{r1 r7} – c6n3{r7 r456} – r4n3{c4 c9} – r4n9{c9 c2} – c2n4{r4 r789} – r7n4{c3 c2} – r7n6{c2 c4} – r4c4{n6 n2} – r9c4{n2 n7} – r9c6{n7 n5} – r8c6{n5 .} ==> r9c7 ≠ 9

naked-single ==> r9c7 = 1

whip[5]: c6n5{r8 r9} – r9n9{c6 c9} – r4n9{c9 c2} – r1c2{n9 n8} – r2c2{n8 .} ==> r8c2 ≠ 5

whip[7]: r4n9{c2 c9} – b9n9{r9c9 r7c8} – r2n9{c8 c5} – c5n1{r2 r4} – r4n3{c5 c4} – b2n3{r1c4 r1c6} – b2n1{r1c6 .} ==> r1c2 ≠ 9

whip[5]: r1c2{n8 n5} – r2c2{n5 n9} – b4n9{r4c2 r5c1} – b4n3{r5c1 r6c1} – b4n8{r6c1 .} ==> r1c3 ≠ 8, r2c3 ≠ 8, r3c3 ≠ 8

whip[8]: r1c2{n5 n8} – r2c2{n8 n9} – b4n9{r4c2 r5c1} – b4n3{r5c1 r6c1} – b4n8{r6c1 r6c3} – r6c7{n8 n4} – b3n4{r3c7 r3c8} – b3n5{r3c8 .} ==> r2c1 ≠ 5

whip[9]: r1c2{n8 n5} – r2c2{n5 n9} – b4n9{r4c2 r5c1} – b4n3{r5c1 r6c1} – b4n8{r6c1 r6c3} – r6c7{n8 n4} – r3c7{n4 n9} – c8n9{r3 r7} – c5n9{r7 .} ==> r3c1 ≠ 8

whip[10]: b7n5{r9c1 r9c2} – r1c2{n5 n8} – r2c2{n8 n9} – r4n9{c2 c9} – b9n9{r9c9 r7c8} – c5n9{r7 r3} – c6n9{r1 r9} – r9n7{c6 c4} – r3n7{c4 c3} – b1n6{r3c3 .} ==> r3c1 ≠ 5

whip[1]: c1n5{r9 .} ==> r9c2 ≠ 5

whip[3]: c2n6{r7 r4} – c2n9{r4 r2} – r3c1{n9 .} ==> r8c1 ≠ 6

whip[1]: b7n6{r9c2 .} ==> r4c2 ≠ 6

whip[5]: b1n8{r2c1 r1c2} – b3n8{r1c9 r3c7} – b3n4{r3c7 r3c8} – b3n5{r3c8 r2c8} – c2n5{r2 .} ==> r2c4 ≠ 8

whip[8]: r3c1{n9 n6} – r3c3{n6 n7} – r3c5{n7 n8} – b8n8{r8c5 r7c4} – c3n8{r7 r6} – r6c1{n8 n3} – r6c7{n3 n4} – b3n4{r3c7 .} ==> r3c8 ≠ 9

whip[10]: r1c2{n8 n5} – r2c2{n5 n9} – b4n9{r4c2 r5c1} – b4n3{r5c1 r6c1} – b4n8{r6c1 r6c3} – r6c7{n8 n4} – r8n4{c7 c9} – r8n6{c9 c6} – c4n6{r9 r4} – b4n6{r4c3 .} ==> r8c2 ≠ 8

whip[9]: r3c1{n9 n6} – r3c3{n6 n7} – r3c5{n7 n8} – r8n8{c5 c1} – r6c1{n8 n3} – r5c1{n3 n9} – b1n9{r2c1 r2c2} – c8n9{r2 r7} – c5n9{r7 .} ==> r3c7 ≠ 9

whip[8]: r8c7{n2 n4} – r3c7{n4 n8} – c5n8{r3 r7} – c3n8{r7 r6} – b6n8{r6c9 r5c9} – b6n7{r5c9 r6c9} – c9n4{r6 r4} – r6n4{c8 .} ==> r8c5 ≠ 2
whip[9]: r1c2{n8 n5} – r1c4{n5 n3} – r2n3{c5 c9} – r4n3{c9 c5} – c5n1{r4 r2} – r1c6{n1 n9} – r1c7{n9 n2} – r8c7{n2 n4} – r3c7{n4 .} ==> r1c9 ≠ 8
whip[9]: b3n3{r1c7 r2c9} – r4n3{c9 c5} – r4n1{c5 c8} – r6n1{c8 c6} – r1c6{n1 n9} – r1c9{n9 n2} – r4n2{c9 c4} – b8n2{r9c4 r7c5} – b8n9{r7c5 .} ==> r1c4 ≠ 3
whip[2]: r1c2{n8 n5} – r1c4{n5 .} ==> r1c7 ≠ 8
whip[4]: r1c4{n5 n8} – r3n8{c5 c7} – b3n4{r3c7 r3c8} – b3n5{r3c8 .} ==> r2c4 ≠ 5
whip[8]: c7n9{r1 r5} – c1n9{r5 r3} – b1n6{r3c1 r3c3} – r4c3{n6 n4} – r6c3{n4 n8} – c7n8{r6 r3} – b3n4{r3c7 r3c8} – b3n5{r3c8 .} ==> r2c8 ≠ 9
whip[1]: b3n9{r1c7 .} ==> r1c6 ≠ 9
whip[1]: c6n9{r9 .} ==> r7c5 ≠ 9
whip[4]: r1c3{n2 n1} – r2c3{n1 n7} – r2c4{n7 n3} – r1c6{n3 .} ==> r2c1 ≠ 2
whip[1]: c1n2{r9 .} ==> r7c3 ≠ 2
whip[5]: r6n1{c8 c6} – r1c6{n1 n3} – r2n3{c5 c9} – b3n8{r2c9 r3c7} – b3n4{r3c7 .} ==> r6c8 ≠ 4
whip[7]: r6c8{n6 n1} – c6n1{r6 r1} – c5n1{r2 r4} – r4n3{c5 c4} – c6n3{r6 r7} – r7n9{c6 c8} – b9n6{r7c8 .} ==> r4c9 ≠ 6
whip[8]: r8c7{n2 n4} – r8c9{n4 n6} – r8c2{n6 n7} – r9c2{n7 n6} – r9c4{n6 n7} – r2c4{n7 n3} – r2c9{n3 n8} – r3c7{n8 .} ==> r9c9 ≠ 2
whip[8]: c2n5{r2 r1} – b1n8{r1c2 r2c1} – c1n9{r2 r5} – c8n9{r5 r7} – r9n9{c9 c6} – r9n5{c6 c1} – r8c1{n5 n2} – b9n2{r8c7 .} ==> r2c2 ≠ 9
hidden-single-in-a-column ==> r4c2 = 9
whip[1]: c2n4{r8 .} ==> r7c3 ≠ 4
whip[2]: r1c2{n8 n5} – r2c2{n5 .} ==> r7c2 ≠ 8
whip[1]: c2n8{r1 .} ==> r2c1 ≠ 8
whip[5]: r8c5{n8 n7} – r3c5{n7 n9} – c1n9{r3 r2} – c1n1{r2 r7} – r7c3{n1 .} ==> r7c5 ≠ 8
whip[4]: b8n9{r7c6 r9c6} – r9n5{c6 c1} – r9n2{c1 c4} – r7c5{n2 .} ==> r7c6 ≠ 3
g-whip[2]: r4n3{c9 c456} – c6n3{r5 .} ==> r1c9 ≠ 3
whip[3]: b4n3{r5c1 r6c1} – c7n3{r6 r1} – c6n3{r1 .} ==> r5c5 ≠ 3
whip[3]: b4n3{r5c1 r6c1} – c7n3{r6 r1} – c6n3{r1 .} ==> r5c9 ≠ 3
whip[3]: c9n8{r6 r2} – b3n3{r2c9 r1c7} – c7n9{r1 .} ==> r5c7 ≠ 8
whip[3]: c4n8{r3 r7} – c3n8{r7 r6} – c7n8{r6 .} ==> r3c5 ≠ 8
hidden-single-in-a-column ==> r8c5 = 8
whip[2]: b2n5{r3c4 r1c4} – c4n8{r1 .} ==> r3c4 ≠ 7
whip[2]: r7n1{c1 c3} – r7n8{c3 .} ==> r7c1 ≠ 2
whip[3]: r9c2{n6 n7} – b8n7{r9c6 r8c6} – b8n5{r8c6 .} ==> r9c6 ≠ 6
whip[4]: r9c9{n9 n6} – r9c2{n6 n7} – r9c4{n7 n2} – r7n2{c5 .} ==> r7c8 ≠ 9
singles to the end

### 7.7.5. $B_\infty \not\subset gW_\infty$: a puzzle P with $gW(P) = \infty$ but $B(P) = 6$

With Figure 7.5, we now have the converse case of a puzzle P (of moderate difficulty) not solvable by g-whips but solvable by braids: B(P) = gB(P) = 6 but W(P) = gW(P) = ∞.

Not only is this puzzle not solvable by whips or g-whips, it allows no elimination at all by whips or g-whips at the start. Let us try with braids :

|   |   |   |   | 1 |   |   | 2 |   |
|---|---|---|---|---|---|---|---|---|
|   | 1 |   |   | 3 |   |   | 4 |   |
|   |   | 5 | 4 |   |   | 6 |   |   |
|   |   | 3 | 5 |   |   | 7 |   |   |
|   | 4 |   |   |   |   | 2 |   |   |
| 1 |   |   |   | 8 |   |   |   | 9 |
|   |   | 2 | 9 |   |   | 1 |   |   |
|   | 3 |   |   | 6 |   |   |   | 7 |
| 6 |   |   |   | 7 |   | 9 |   |   |

| 8 | 9 | 4 | 6 | 5 | 1 | 3 | 7 | 2 |
|---|---|---|---|---|---|---|---|---|
| 7 | 1 | 6 | 8 | 3 | 2 | 9 | 4 | 5 |
| 3 | 2 | 5 | 4 | 7 | 9 | 6 | 8 | 1 |
| 2 | 8 | 3 | 5 | 9 | 6 | 7 | 1 | 4 |
| 5 | 4 | 9 | 7 | 1 | 3 | 8 | 2 | 6 |
| 1 | 6 | 7 | 2 | 4 | 8 | 5 | 3 | 9 |
| 4 | 7 | 2 | 9 | 8 | 5 | 1 | 6 | 3 |
| 9 | 3 | 8 | 1 | 6 | 4 | 2 | 5 | 7 |
| 6 | 5 | 1 | 3 | 2 | 7 | 4 | 9 | 8 |

***Figure 7.5.*** *A puzzle P with B(P) = 6 but gW(P) = W(P) = ∞*

\*\*\*\*\* SudoRules 16.2 based on CSP-Rules 1.2, config: B \*\*\*\*\*
25 givens, 204 candidates, 1214 csp-links and 1214 links. Initial density = 1.47
braid[5]: b9n6{r7c9 r7c8} – r7n3{c8 c6} – r2c9{n8 n5} – c6n5{r2 r8} – r8c8{n8 .} ==> r7c9 ≠ 8
braid[5]: b9n2{r9c7 r8c7} – c7n4{r8 r6} – r8c8{n8 n5} – r6n5{c7 c2} – r9c2{n8 .} ==> r9c7 ≠ 8
braid[6]: r9c2{n5 n8} – r8c8{n5 n8} – r2c9{n5 n8} – c7n8{r1 r5} – c3n8{r2 r1} – c4n8{r9 .} ==> r9c9 ≠ 5
whip[6]: b4n5{r5c1 r6c2} – r9n5{c2 c5} – r1n5{c5 c8} – r2c9{n5 n8} – b6n8{r5c9 r4c8} – r8c8{n8 .} ==> r5c7 ≠ 5
braid[5]: r5c7{n8 n3} – r8c8{n8 n5} – c6n3{r5 r7} – r7c8{n8 n6} – r6c8{n6 .} ==> r4c8 ≠ 8
whip[3]: r5n5{c1 c9} – r2c9{n5 n8} – b6n8{r5c9 .} ==> r5c1 ≠ 8
braid[5]: r5c7{n8 n3} – r8c8{n8 n5} – c6n3{r5 r7} – r7c8{n8 n6} – r6c8{n6 .} ==> r8c7 ≠ 8
whip[6]: b2n5{r1c5 r2c6} – c9n5{r2 r5} – c1n5{r5 r8} – r9c2{n5 n8} – b8n8{r9c5 r8c4} – r8c8{n8 .} ==> r7c5 ≠ 5
braid[5]: r7c5{n8 n4} – r8c8{n8 n5} – r6n4{c5 c7} – r8c7{n5 n2} – r8c6{n5 .} ==> r7c8 ≠ 8
whip[3]: c5n5{r1 r9} – r9c2{n5 n8} – r7n8{c1 .} ==> r1c5 ≠ 8
whip[4]: r7c5{n8 n4} – r6n4{c5 c7} – b9n4{r9c7 r9c9} – b9n8{r9c9 .} ==> r8c4 ≠ 8
**braid[6]: b8n5{r8c6 r9c5} – r2c9{n5 n8} – r9c2{n5 n8} – r4n8{c9 c1} – b8n8{r9c5 r7c5} – r3n8{c9 .} ==> r2c6 ≠ 5**
hidden-single-in-a-block ==> r1c5 = 5
whip[2]: r9n5{c2 c7} – c8n5{r7 .} ==> r6c2 ≠ 5
hidden-single-in-a-block ==> r5c1 = 5
whip[6]: b8n3{r7c6 r9c4} – r6n3{c4 c7} – b6n4{r6c7 r4c9} – r9c9{n4 n8} – r8c8{n8 n5} – b6n5{r6c8 .} ==> r7c8 ≠ 3
whip[4]: r4c8{n1 n6} – r7c8{n6 n5} – b6n5{r6c8 r6c7} – b6n4{r6c7 .} ==> r4c9 ≠ 1
whip[5]: c8n3{r1 r6} – b6n5{r6c8 r6c7} – c7n3{r6 r9} – c7n4{r9 r8} – b9n2{r8c7 .} ==> r3c9 ≠ 3
whip[6]: r6n4{c7 c5} – c6n4{r4 r7} – b8n3{r7c6 r9c4} – r9c9{n3 n8} – r8c8{n8 n5} – b8n5{r8c6 .} ==> r8c7 ≠ 4
whip[4]: r8c4{n1 n2} – r8c7{n2 n5} – b8n5{r8c6 r7c6} – b8n3{r7c6 .} ==> r9c4 ≠ 1
whip[4]: b9n4{r9c9 r7c9} – r7n3{c9 c6} – b8n5{r7c6 r8c6} – b8n4{r8c6 .} ==> r9c3 ≠ 4

whip[6]: b8n5{r8c6 r7c6} – b8n3{r7c6 r9c4} – r9n2{c4 c7} – c7n4{r9 r6} – b6n5{r6c7 r6c8} – r6n3{c8 .} ==> r8c6 ≠ 2
whip[6]: c7n4{r9 r6} – b6n5{r6c7 r6c8} – r6n3{c8 c4} – b8n3{r9c4 r7c6} – b8n5{r7c6 r8c6} – r8c7{n5 .} ==> r9c7 ≠ 2
singles ==> r8c7 = 2, r8c4 = 1, r9c3 = 1
whip[5]: r2n7{c1 c4} – r5n7{c4 c5} – r3n7{c5 c8} – c8n1{r3 r4} – c5n1{r4 .} ==> r1c3 ≠ 7
whip[5]: r3c6{n2 n9} – r2c6{n9 n6} – r5c6{n6 n3} – b8n3{r7c6 r9c4} – b8n2{r9c4 .} ==> r3c5 ≠ 2
whip[5]: c2n9{r1 r4} – c5n9{r4 r5} – r5n1{c5 c9} – b3n1{r3c9 r3c8} – r3n3{c8 .} ==> r3c1 ≠ 9
whip[6]: b6n5{r6c8 r6c7} – r6n4{c7 c5} – r7c5{n4 n8} – r9c5{n8 n2} – r9c4{n2 n3} – r6n3{c4 .} ==> r6c8 ≠ 6
whip[6]: r3c9{n8 n1} – r5n1{c9 c5} – c5n9{r5 r4} – r5n9{c6 c3} – r5n7{c3 c4} – c5n7{r6 .} ==> r3c5 ≠ 8
whip[1]: c5n8{r9 .} ==> r9c4 ≠ 8
whip[4]: c4n8{r2 r1} – c7n8{r1 r5} – c3n8{r5 r8} – b9n8{r8c8 .} ==> r2c9 ≠ 8
naked-single ==> r2c9 = 5
whip[2]: c7n4{r6 r9} – c7n5{r9 .} ==> r6c7 ≠ 3
whip[2]: c7n4{r9 r6} – c7n5{r6 .} ==> r9c7 ≠ 3
whip[1]: b9n3{r9c9 .} ==> r5c9 ≠ 3
whip[4]: c7n4{r9 r6} – b6n5{r6c7 r6c8} – r6n3{c8 c4} – r9n3{c4 .} ==> r9c9 ≠ 4
whip[2]: r6n4{c5 c7} – c9n4{r4 .} ==> r7c5 ≠ 4
naked-single ==> r7c5 = 8
whip[2]: b6n4{r4c9 r6c7} – r9n4{c7 .} ==> r4c5 ≠ 4
whip[3]: r8c8{n5 n8} – r9n8{c9 c2} – b7n5{r9c2 .} ==> r7c8 ≠ 5
singles ==> r7c8 = 6, r4c8 = 1, r3c9 = 1, r5c5 = 1
whip[3]: b5n4{r4c6 r6c5} – r9c5{n4 n2} – r4c5{n2 .} ==> r4c6 ≠ 9
whip[3]: b9n4{r7c9 r9c7} – r9n5{c7 c2} – r7n5{c2 .} ==> r7c6 ≠ 4
whip[3]: r5c7{n3 n8} – c9n8{r5 r9} – r9n3{c9 .} ==> r5c4 ≠ 3
whip[3]: b4n7{r6c3 r5c3} – r5n9{c3 c6} – b5n3{r5c6 .} ==> r6c4 ≠ 7
whip[3]: r6n6{c2 c4} – b5n3{r6c4 r5c6} – r5n9{c6 .} ==> r5c3 ≠ 6
whip[4]: c3n4{r1 r8} – r8c6{n4 n5} – r8c8{n5 n8} – r3n8{c8 .} ==> r1c3 ≠ 8
whip[3]: b9n8{r9c9 r8c8} – c3n8{r8 r2} – r3n8{c2 .} ==> r5c9 ≠ 8
singles to the end

### 7.7.6. $gB_\infty \neq gW_\infty$: a puzzle solvable by g-braids but probably not by g-whips

   Finding a Sudoku puzzle solvable by g-braids but neither by braids nor by g-whips is very hard: one can rely neither on random generators (all the puzzles we produced with them – about ten millions – were solvable by whips) nor on Subset rules that would not be subsumed by g-whips but would be by g-braids (see chapter 8 for comments on this). The following (Figure 7.6) gives the only such puzzle (#77) in the Magictour-top1465 collection. Using the "gT&E vs g-braids" and "T&E vs braids" theorems, it is easy to show that it can be solved by g-braids but not by braids. And the following resolution path with g-whips shows that these are not enough either to make substantial advances in the solution.

| 7 |   |   |   |   | 4 |   |   |   |
|---|---|---|---|---|---|---|---|---|
|   | 2 |   |   | 7 |   |   | 8 |   |
|   |   | 3 |   |   | 8 |   |   | 9 |
|   |   |   | 5 |   |   | 3 |   |   |
|   | 6 |   |   | 2 |   |   | 9 |   |
|   |   | 1 |   |   | 7 |   |   | 6 |
|   |   |   | 3 |   |   | 9 |   |   |
|   | 3 |   |   | 4 |   |   | 6 |   |
|   |   | 9 |   |   | 1 |   |   | 5 |

| 7 | 9 | 8 | 6 | 3 | 5 | 4 | 2 | 1 |
|---|---|---|---|---|---|---|---|---|
| 1 | 2 | 6 | 9 | 7 | 4 | 5 | 8 | 3 |
| 4 | 5 | 3 | 2 | 1 | 8 | 6 | 7 | 9 |
| 9 | 7 | 2 | 5 | 8 | 6 | 3 | 1 | 4 |
| 5 | 6 | 4 | 1 | 2 | 3 | 8 | 9 | 7 |
| 3 | 8 | 1 | 4 | 9 | 7 | 2 | 5 | 6 |
| 6 | 1 | 7 | 3 | 5 | 2 | 9 | 4 | 8 |
| 8 | 3 | 5 | 7 | 4 | 9 | 1 | 6 | 2 |
| 2 | 4 | 9 | 8 | 6 | 1 | 7 | 3 | 5 |

**Figure 7.6.** *A puzzle (Magictour-top1465#77) solvable by g-braids but not by braids and probably not by g-whips*

***** SudoRules 16.2 based on CSP-Rules 1.2, config: gW *****
24 givens, 219 candidates, 1397 csp-links and 1397 links. Initial density = 1.46.
hidden-single-in-a-row ==> r9c8 = 3
whip[1]: r9n4{c2 .} ==> r7c1 ≠ 4, r7c2 ≠ 4, r7c3 ≠ 4
g-whip[8]: c4n6{r2 r9} – r1n6{c4 c3} – b1n8{r1c3 r1c2} – b1n9{r1c2 r2c1} – b1n1{r2c1 r3c123} – r3c5{n1 n5} – r7c5{n5 n8} – r9c5{n8 .} ==> r2c6 ≠ 6
whip[11]: c3n4{r4 r2} – c6n4{r2 r4} – b5n6{r4c6 r4c5} – r9c5{n6 n8} – r7c5{n8 n5} – r3c5{n5 n1} – c4n1{r3 r5} – b5n8{r5c4 r6c4} – b5n9{r6c4 r6c5} – r6c2{n9 n5} – r3c2{n5 .} ==> r5c1 ≠ 4
whip[12]: r9c5{n8 n6} – r7c5{n6 n5} – r3c5{n5 n1} – r4c5{n1 n9} – r6c4{n9 n4} – r5c6{n4 n3} – b4n3{r5c1 r6c1} – c1n9{r6 r2} – r2c4{n9 n6} – r1n6{c6 c3} – b1n8{r1c3 r1c2} – b1n1{r1c2 .} ==> r6c5 ≠ 8
g-whip[14]: r3n4{c4 c123} – c3n4{r2 r4} – c6n4{r4 r2} – r5c6{n4 n3} – b4n3{r5c1 r6c1} – r6n4{c1 c8} – r6n2{c8 c7} – b4n2{r6c1 r4c1} – r9n2{c1 c4} – c6n2{r8 r1} – b2n3{r1c6 r1c5} – r1c9{n3 n1} – r5n1{c9 c7} – b6n5{r5c7 .} ==> r5c4 ≠ 4
whip[15]: b3n6{r2c7 r3c7} – b3n7{r3c7 r3c8} – r3n2{c8 c4} – c4n4{r3 r6} – r5c6{n4 n3} – r6c5{n3 n9} – r4c6{n9 n6} – r1n6{c6 c3} – b1n8{r1c3 r1c2} – r6c2{n8 n5} – c8n5{r6 r1} – r1c6{n5 n9} – r1c4{n9 n1} – r5c4{n1 n8} – r5c1{n8 .} ==> r2c4 ≠ 6

After this point, there is no whip or g-whip of length less than 18. While trying g-whips[18], the number of partial g-whips to be analysed suddenly gets so large that SudoRules encounters memory overflow problems. Given the poor partial results above (only 8 eliminations after the HS(row)), it is unlikely that a g-whip solution can be obtained.

Exercise for the reader: write a better implementation of g-whips (less greedy for memory) and prove that there is indeed no g-whip solution.

### 7.7.7. A puzzle with all the W, B and gW ratings finite, but very different

In section 5.10.4, we mentioned a puzzle P (Figure 5.6) with W(P) = 31 and B(P) = 19. We shall now show that gW(P) = 12. This will show that, even when all the ratings are finite, they can, in extremely rare cases, be very different. The path with g-whips is radically different from the start from the paths with whips or braids. It can also be shown that gB(P) = 11.

Together with all the previous ones, this example shows that "obstructions" to the extension of partial whips into longer ones can sometimes be palliated by two very different mild forms of branching: as in braids or as in g-whips. Moreover, most of the time, the g-whip type is more powerful than the braid type, even though it does not subsume it.

***** SudoRules 16.2 based on CSP-Rules 1.2, config: gW *****
24 givens, 220 candidates, 1433 csp-links and 1433 links. Initial density = 1.49
g-whip[6]: b2n4{r3c6 r1c4} – c3n4{r1 r456} – r6n4{c2 c3} – b4n8{r6c3 r4c1} – r3n8{c1 c5} – c6n8{r3 .} ==> r3c6 ≠ 6, r3c6 ≠ 9
g-whip[6]: b4n8{r6c3 r4c1} – r3n8{c1 c456} – c6n8{r2 r3} – b2n4{r3c6 r1c4} – c3n4{r1 r5} – r6n4{c3 .} ==> r6c3 ≠ 6, r6c3 ≠ 1
whip[11]: c8n9{r1 r5} – r4c9{n9 n5} – r2n5{c9 c4} – b2n7{r2c4 r1c4} – b2n4{r1c4 r3c6} – r5c6{n4 n3} – c4n3{r5 r7} – b8n8{r7c4 r7c5} – b2n8{r3c5 r2c6} – r3n8{c6 c1} – r4n8{c1 .} ==> r2c7 ≠ 9
whip[11]: r8n1{c1 c5} – r9c4{n1 n5} – c2n5{r9 r4} – b4n7{r4c2 r4c1} – b4n8{r4c1 r6c3} – r6c5{n8 n2} – r4n2{c6 c7} – b6n4{r4c7 r5c7} – b4n4{r5c3 r6c2} – c3n4{r6 r1} – c4n4{r1 .} ==> r7c2 ≠ 1
**whip[12]: r6n4{c2 c4} – r4n4{c6 c7} – b6n2{r4c7 r6c8} – r6n3{c8 c9} – r6n6{c9 c2} – r6n1{c2 c5} – r5c4{n1 n3} – r5c6{n3 n9} – r4n9{c6 c9} – r2n9{c9 c2} – c2n1{r2 r9} – r8n1{c1 .} ==> r5c3 ≠ 4**
g-whip[5]: c3n4{r1 r6} – b4n8{r6c3 r4c1} – r3n8{c1 c456} – c6n8{r2 r3} – b2n4{r3c6 .} ==> r1c2 ≠ 4
g-whip[5]: c3n4{r1 r6} – b4n8{r6c3 r4c1} – r3n8{c1 c456} – c6n8{r2 r3} – b2n4{r3c6 .} ==> r1c1 ≠ 4
whip[10]: c3n4{r1 r6} – b4n8{r6c3 r4c1} – b1n8{r1c1 r2c3} – b1n2{r2c3 r2c1} – c1n7{r2 r8} – c2n7{r9 r4} – b4n5{r4c2 r5c1} – c1n1{r5 r7} – b9n1{r7c9 r9c9} – b9n7{r9c9 .} ==> r1c3 ≠ 7
whip[11]: c3n7{r9 r2} – b1n2{r2c3 r2c1} – b1n1{r2c1 r2c2} – c2n7{r2 r4} – r9n7{c2 c9} – b9n1{r9c9 r7c9} – c1n1{r7 r5} – b4n5{r5c1 r4c1} – r4c9{n5 n9} – r5n9{c8 c6} – r2n9{c6 .} ==> r8c1 ≠ 7
**g-whip[11]: b1n2{r2c1 r2c3} – c3n8{r2 r6} – c3n4{r6 r1} – r3n4{c2 c6} – r3n8{c6 c5} – b8n8{r7c5 r7c4} – b8n3{r7c4 r8c6} – r5c6{n3 n9} – c8n9{r5 r123} – r2n9{c9 c2} – b1n1{r2c2 .} ==> r2c1 ≠ 8**
**whip[12]: b9n1{r7c9 r9c9} – r9c4{n1 n5} – c2n5{r9 r4} – r4c9{n5 n9} – b5n9{r4c6 r5c6} – r2n9{c6 c2} – c2n1{r2 r6} – r5c3{n1 n6} – r5c1{n6 n4} – r5n1{c1 c4} – b5n3{r5c4 r6c4} – r6n4{c4 .} ==> r7c9 ≠ 5**
**whip[12]: r3c6{n8 n4} – b1n4{r3c2 r1c3} – r6c3{n4 n8} – c4n8{r6 r7} – b8n3{r7c4 r8c6} – r5c6{n3 n9} – r4c5{n9 n2} – r6c5{n2 n1} – b8n1{r8c5 r9c4} – c2n1{r9 r2} – r2n9{c2 c9} – c8n9{r3 .} ==> r3c5 ≠ 8**
whip[4]: c3n8{r1 r6} – c3n4{r6 r1} – r3n4{c2 c6} – r3n8{c6 .} ==> r1c1 ≠ 8
whip[10]: r3n8{c1 c6} – b2n4{r3c6 r1c4} – b1n4{r1c3 r3c2} – b1n3{r3c2 r1c2} – r1n7{c2 c8} – r2n7{c9 c4} – b2n5{r2c4 r1c5} – r1n9{c5 c7} – b9n9{r9c7 r9c9} – b9n7{r9c9 .} ==> r3c1 ≠ 7

whip[11]: r3n7{c9 c2} – b1n3{r3c2 r1c2} – b1n9{r1c2 r2c2} – r2n7{c2 c4} – c9n7{r2 r9} – b9n9{r9c9 r9c7} – r1n9{c7 c5} – b2n5{r1c5 r1c4} – r9c4{n5 n1} – b5n1{r5c4 r6c5} – c2n1{r6 .} ==> r1c8 ≠ 7

**whip[12]: b9n9{r9c7 r9c9} – r4c9{n9 n5} – r2n5{c9 c4} – r9c4{n5 n1} – c5n1{r8 r6} – c2n1{r6 r2} – r2n9{c2 c6} – r3c5{n9 n6} – r1c5{n6 n8} – c5n9{r1 r4} – b5n2{r4c5 r4c6} – c6n8{r4 .} ==> r9c7 ≠ 5**

**whip[12]: r6c3{n4 n8} – c1n8{r4 r3} – r3c6{n8 n4} – r4n4{c6 c7} – b6n2{r4c7 r6c8} – r6c5{n2 n1} – r5c4{n1 n3} – r5c6{n3 n9} – b6n9{r5c8 r4c9} – r2n9{c9 c2} – c2n1{r2 r9} – r8n1{c1 .} ==> r5c1 ≠ 4**

whip[4]: b4n5{r5c1 r4c2} – b4n7{r4c2 r4c1} – c1n8{r4 r3} – c1n4{r3 .} ==> r7c1 ≠ 5

g-whip[8]: r4c9{n5 n9} – b5n9{r4c6 r5c6} – r5n4{c6 c4} – b5n3{r5c4 r6c4} – b5n1{r6c4 r6c5} – r8n1{c5 c123} – c2n1{r9 r2} – r2n9{c2 .} ==> r5c7 ≠ 5

**whip[12]: r5n5{c1 c8} – r4c9{n5 n9} – r5n9{c8 c6} – r2n9{c6 c2} – c2n1{r2 r9} – r8n1{c1 c5} – c4n1{r9 r6} – b5n3{r6c4 r5c4} – c4n4{r5 r1} – c3n4{r1 r6} – r6c2{n4 n6} – r5c3{n6 .} ==> r5c1 ≠ 1**

biv-chain[3]: r5c1{n5 n6} – r1c1{n6 n7} – r4n7{c1 c2} ==> r4c2 ≠ 5

whip[1]: c2n5{r9 .} ==> r8c1 ≠ 5

whip[4]: b7n4{r7c1 r7c2} – b7n5{r7c2 r9c2} – r9c4{n5 n1} – b9n1{r9c9 .} ==> r7c1 ≠ 1

whip[6]: r6n2{c8 c5} – r7n2{c5 c1} – b7n4{r7c1 r7c2} – b7n5{r7c2 r9c2} – r9c4{n5 n1} – b5n1{r5c4 .} ==> r8c8 ≠ 2

whip[7]: c1n1{r8 r2} – c1n2{r2 r7} – b7n4{r7c1 r7c2} – b7n5{r7c2 r9c2} – c2n1{r9 r6} – b5n1{r6c5 r5c4} – r9c4{n1 .} ==> r8c1 ≠ 6

whip[9]: b7n5{r7c2 r9c2} – r9c4{n5 n1} – r5n1{c4 c3} – b4n6{r5c3 r5c1} – b4n5{r5c1 r4c1} – r4c9{n5 n9} – r5n9{c8 c6} – r2n9{c6 c2} – c2n1{r2 .} ==> r7c2 ≠ 6

**g-whip[11]: b7n4{r7c1 r7c2} – b7n5{r7c2 r9c2} – r9c4{n5 n1} – b5n1{r5c4 r6c5} – r6c2{n1 n6} – r6n4{c2 c4} – r5c4{n4 n3} – r5c6{n3 n9} – c8n9{r5 r123} – r2n9{c9 c2} – c2n1{r2 .} ==> r4c1 ≠ 4**

whip[5]: b2n7{r2c4 r1c4} – r1n4{c4 c3} – c1n4{r3 r7} – c1n2{r7 r8} – c1n1{r8 .} ==> r2c1 ≠ 7

whip[5]: r7c2{n5 n4} – c1n4{r7 r3} – b2n4{r3c6 r1c4} – b2n7{r1c4 r2c4} – b2n5{r2c4 .} ==> r7c5 ≠ 5

whip[6]: r7c2{n5 n4} – b4n4{r6c2 r6c3} – r1n4{c3 c4} – b2n7{r1c4 r2c4} – c4n5{r2 r9} – r8n5{c5 .} ==> r7c8 ≠ 5

whip[6]: b2n5{r1c5 r2c4} – r7n5{c4 c2} – r9n5{c2 c9} – r4n5{c9 c1} – c1n7{r4 r1} – b2n7{r1c4 .} ==> r1c7 ≠ 5

whip[6]: r7c2{n5 n4} – b4n4{r6c2 r6c3} – r1n4{c3 c4} – b2n7{r1c4 r2c4} – c4n5{r2 r9} – r8n5{c5 .} ==> r7c7 ≠ 5

whip[5]: b9n7{r9c9 r8c8} – b9n5{r8c8 r8c7} – c5n5{r8 r1} – r2n5{c4 c9} – r4c9{n5 .} ==> r9c9 ≠ 9

hidden-single-in-a-block ==> r9c7 = 9

whip[6]: c6n3{r8 r5} – r5n9{c6 c8} – r4c9{n9 n5} – c8n5{r5 r1} – b9n5{r8c8 r8c7} – c5n5{r8 .} ==> r8c8 ≠ 3

whip[6]: c1n1{r2 r8} – c1n2{r8 r7} – b9n2{r7c7 r8c7} – r8n3{c7 c6} – c6n6{r8 r9} – b8n2{r9c6 .} ==> r2c1 ≠ 6

whip[2]: r8c1{n2 n1} – r2c1{n1 .} ==> r7c1 ≠ 2

whip[7]: c5n5{r1 r8} – r9c4{n5 n1} – r5n1{c4 c3} – c2n1{r6 r2} – r2n9{c2 c9} – r4c9{n9 n5} – b9n5{r9c9 .} ==> r1c5 ≠ 9

biv-chain[2]: c5n9{r3 r4} – r5n9{c6 c8} ==> r3c8 ≠ 9

biv-chain[3]: c5n9{r3 r4} – b6n9{r4c9 r5c8} – r1n9{c8 c2} ==> r3c2 ≠ 9

whip[4]: c5n5{r1 r8} – c7n5{r8 r4} – r4c9{n5 n9} – c8n9{r5 .} ==> r1c8 ≠ 5
whip[1]: b3n5{r2c9 .} ==> r2c4 ≠ 5
whip[4]: r8n7{c3 c8} – c8n5{r8 r5} – r4n5{c7 c1} – b4n7{r4c1 .} ==> r9c2 ≠ 7
whip[1]: b7n7{r9c3 .} ==> r2c3 ≠ 7
whip[4]: r8n7{c3 c8} – c8n5{r8 r5} – r5c1{n5 n6} – r5c3{n6 .} ==> r8c3 ≠ 1
whip[4]: c8n7{r3 r8} – c8n5{r8 r5} – r4n5{c7 c1} – b4n7{r4c1 .} ==> r3c2 ≠ 7
whip[1]: r3n7{c9 .} ==> r2c9 ≠ 7
whip[4]: c5n5{r1 r8} – c8n5{r8 r5} – c7n5{r4 r2} – b3n8{r2c7 .} ==> r1c5 ≠ 8
whip[6]: r9c4{n5 n1} – r5n1{c4 c3} – c2n1{r6 r2} – b1n9{r2c2 r1c2} – c8n9{r1 r5} – r4c9{n9 .} ==> r9c9 ≠ 5
whip[1]: b9n5{r8c7 .} ==> r8c5 ≠ 5
hidden-single-in-a-column ==> r1c5 = 5
biv-chain[2]: b2n9{r2c6 r3c5} – b2n6{r3c5 r2c6} ==> r2c6 ≠ 8
whip[2]: c1n8{r4 r3} – c6n8{r3 .} ==> r4c5 ≠ 8
whip[3]: r9c6{n2 n6} – r8c5{n6 n1} – r8c1{n1 .} ==> r8c6 ≠ 2
whip[4]: b9n2{r7c7 r8c7} – r8c1{n2 n1} – c5n1{r8 r6} – c5n8{r6 .} ==> r7c5 ≠ 2
whip[1]: r7n2{c8 .} ==> r8c7 ≠ 2
whip[4]: r2n5{c7 c9} – r4c9{n5 n9} – r5n9{c8 c6} – r2c6{n9 .} ==> r2c7 ≠ 6
whip[4]: r2c6{n6 n9} – r2c9{n9 n5} – r4c9{n5 n9} – r5n9{c8 .} ==> r2c2 ≠ 6, r2c3 ≠ 6
biv-chain[5]: r6c9{n3 n6} – r2n6{c9 c6} – r3c5{n6 n9} – r4c5{n9 n2} – r6n2{c5 c8} ==> r6c8 ≠ 3
whip[5]: r1n9{c8 c2} – b1n3{r1c2 r3c2} – b1n6{r3c2 r3c1} – r3n8{c1 c6} – r3n4{c6 .} ==> r1c8 ≠ 6
biv-chain[6]: c9n1{r7 r9} – b9n7{r9c9 r8c8} – c8n5{r8 r5} – c1n5{r5 r4} – r4n8{c1 c6} – c5n8{r6 r7} ==> r7c5 ≠ 1
biv-chain[3]: r5n1{c3 c4} – c5n1{r6 r8} – c1n1{r8 r2} ==> r2c3 ≠ 1
whip[4]: r8n7{c3 c8} – r8n5{c8 c7} – r2c7{n5 n8} – r2c3{n8 .} ==> r8c3 ≠ 2
whip[2]: r8n1{c5 c1} – r8n2{c1 .} ==> r8c5 ≠ 6
whip[2]: r2n6{c9 c6} – b8n6{r9c6 .} ==> r7c9 ≠ 6
whip[3]: r9n7{c9 c3} – b7n1{r9c3 r8c1} – b7n2{r8c1 .} ==> r9c9 ≠ 1
hidden-single-in-a-block ==> r7c9 = 1
biv-chain[5]: b3n8{r1c7 r2c7} – r2n5{c7 c9} – r2n6{c9 c6} – c5n6{r3 r7} – r7n8{c5 c4} ==> r1c4 ≠ 8
whip[3]: r4c2{n7 n4} – c3n4{r6 r1} – r1c4{n4 .} ==> r1c2 ≠ 7
whip[4]: b6n4{r5c7 r4c7} – r4c2{n4 n7} – r2n7{c2 c4} – r1c4{n7 .} ==> r5c4 ≠ 4
whip[2]: c3n4{r6 r1} – c4n4{r1 .} ==> r6c2 ≠ 4
whip[2]: r5c3{n6 n1} – r6c2{n1 .} ==> r5c1 ≠ 6
singles ==> r5c1 = 5, r8c8 = 5, r9c9 = 7, r3c8 = 7, r8c3 = 7
whip[2]: r2n6{c9 c6} – r8n6{c6 .} ==> r1c7 ≠ 6
whip[1]: b3n6{r3c9 .} ==> r6c9 ≠ 6
singles ==> r6c9 = 3, r3c2 = 3
whip[1]: r1n6{c1 .} ==> r3c1 ≠ 6
biv-chain[3]: b4n6{r5c3 r6c2} – r1c2{n6 n9} – c8n9{r1 r5} ==> r5c8 ≠ 6
singles ==> r5c8 = 9, r4c9 = 5, r1c8 = 3, r1c7 = 8, r2c7 = 5, r1c2 = 9
whip[2]: c2n6{r9 r6} – c8n6{r6 .} ==> r7c1 ≠ 6
singles to the end

**7.8. g-labels and g-whips in N-Queens and in SudoQueens**

N-Queens provides an interesting example where g-labels are very different from those of Sudoku. See chapters 14 and 15 for still more different examples.

*7.8.1. g-labels in n-Queens*

We have seen in section 5.11 that, in the n-Queens CSP, one can identify a label with a cell in the grid. From the various examples of whip[1] we have already seen there, we can understand that the g-labels of n-Queens are:

– for variable Xr°:

  - all the symmetric sets of horizontal triplets of cells in row r° that are separated by k other cells, $0 \leq k \leq IP((n-3)/2)$, provided that: 1) either the second diagonal passing though the leftmost cell, the first diagonal passing through the rightmost cell and the column passing through the inner cell meet in a cell above r° and inside the grid; 2) or the first diagonal passing through the leftmost cell, the second diagonal passing through the rightmost cell and the column passing through the inner cell meet in a cell under r° and inside the grid. The labels l g-linked to such a g-label correspond to the meeting points; (there are at most 2 such labels, symmetric with respect to r°);

  - all the sets of horizontal pairs of cells in row r° that are separated by k other cells ($0 \leq k \leq n-2$), provided that the column passing through one cell and one of the two diagonals passing through the other cell meet in a cell inside the grid, and provided that they are not part of some of the previous g-labels (maximality condition). The labels l g-linked to such a g-label correspond to the meeting points; (depending on r°, k and n, there are at most 2 or 4 such labels, symmetric with respect to r° and the column containing $l_2$);

– for variable Xc°: similar g-labels obtained by 90° rotation.

Notice that, contrary to the Sudoku case, *any* label l g-linked to a g-label <V, g> for a CSP variable V must use at least two different types of constraints (row, column, first diagonal or second diagonal) for its links with the various elements of g and at least one of these constraints is not defined by a CSP variable.

A simple case of a g-whip[3] can already be seen in the example of Figure 5.11, section 5.11.4. The first whip[4] elimination there can be replaced by a g-whip[3]:

**g-whip[3]: r8{c5 c1} – r2{c1 c58} – r4{c8 .} ⇒ ¬ r2c9 (G eliminated)**

*7.8.2. A g-whip[3] example in 9-Queens*

Accepting the same solution grid as that in Figure 5.11, the puzzle in Figure 7.7 is based on the same first two givens, but a different third one (r3c3, r6c2 and r8c5).

***** Manual solution *****
whip[2]: c6{r1 r5} – c4{r5 .} ⇒ ¬r1c9 (A eliminated)
**g-whip[3]: r4{c8 c67} – r5{c6 c9} – r7{c9 .} ⇒ ¬ r1c8 (B eliminated)**
**g-whip[3]: r4{c8 c67} – r5{c6 c9} – r7{c9 .} ⇒ ¬ r9c8 (C eliminated)**
whip[3]: r9{c7 c1} – r2{c1 c7}- r4{c7 .} ⇒ ¬r7c9 (D eliminated)
single in r7: r7c8
whip[1]: r4{c7 .} ⇒ ¬r5c9 (E eliminated)
whip[2]: r4{c6 c7} – r2{c7 .} ⇒ ¬r9c1 (F eliminated)
single in r9: r9c7; single in c1: r2c1; single in r4: r4c6; single in r1: r1c4; single in r5: r5c9
Solution found in gW3.

|    | c1 | c2 | c3 | c4 | c5 | c6 | c7 | c8 | c9 |
|----|----|----|----|----|----|----|----|----|----|
| r1 | ∘  | ∘  | ∘  | +  | ∘  |    | ∘  | B  | A  |
| r2 | +  | ∘  | ∘  | ∘  | ∘  | ∘  |    |    |    |
| r3 | ∘  | ∘  | ✱  | ∘  | ∘  | ∘  | ∘  | ∘  | ∘  |
| r4 | ∘  | ∘  | ∘  | ∘  | ∘  | +  |    |    | ∘  |
| r5 | ∘  | ∘  | ∘  |    | ∘  |    | E  | ∘  | +  |
| r6 | ∘  | ✱  | ∘  | ∘  | ∘  | ∘  | ∘  | ∘  | ∘  |
| r7 | ∘  | ∘  | ∘  | ∘  | ∘  | ∘  | ∘  | +  | D  |
| r8 | ∘  | ∘  | ∘  | ∘  | ✱  | ∘  | ∘  | ∘  | ∘  |
| r9 | F  | ∘  | ∘  | ∘  | ∘  | ∘  | +  | C  | ∘  |

*Figure 7.7. g-whips in a 9-Queens instance*

### 7.8.3. g-labels in n-SudoQueens

n-SudoQueens was introduced in section 5.11.8. The g-labels of n-SudoQueens are both those of n-Sudoku (without their Number coordinate) and those of n-Queens. As a result, the set of labels of a g-label can be included in the set of labels of another g-label (for a different CSP variable). For instance, consider 9-SudoQueens and the following two g-labels:
- <Xb1, g1> associated with CSP variable Xb1: <Xb1, r3c123>,
- <Xr3, g2> associated with CSP variable Xr3: <Xr3, r3c12>.

Let l be a label with respective representatives (r, c) and [b, s] in the two coordinate systems. Then:

- l is g-linked to <Xb1, g1> if and only if b = b1;
- l is g-linked to <Xr3, g2> if and only if (r = r2 or r = r4) and (c = 1 or c = 2).

This example shows that, although the set of labels in g2 is included in the set of labels in g1, none of the sets of labels linked to them is included in the other. This justifies our definition of a g-label, in which the CSP variable is kept as an explicit component.

### 7.8.4. A g-whip[4] example in 9-SudoQueens

The puzzle in Figure 7.8 shows an example of a g-whip[4] in 9-SudoQueens.

|    | c1 | c2 | c3 | c4 | c5 | c6 | c7 | c8 | c9 |
|----|----|----|----|----|----|----|----|----|----|
| r1 | *  | ○  | ○  | ○  | ○  | ○  | ○  | ○  | ○  |
| r2 | ○  | ○  | ○  | ○  | ○  | ○  | ○  | *  | ○  |
| r3 | ○  | ○  | ○  | ○  | *  | ○  | ○  | ○  | ○  |
| r4 | ○  | -3 | -0 | ○  | ○  | ○  | +3 | ○  | -2 |
| r5 | ○  | -0 | ○  | A  | ○  | -0 | ○  | ○  | -0 |
| r6 | ○  | ○  | -0 | ○  | ○  | ○  | -2 | ○  | +2 |
| r7 | ○  | -0 | ○  | -0 | ○  | -0 | ○  | ○  | ○  |
| r8 | ○  | ○  |    | -0 | ○  | +1 | -0 | ○  | +2 |
| r9 | ○  |    | B  | -0 | ○  | +1 | -1 | ○  | ○  |

*Figure 7.8. A partial grid for 9-SudoQueens*

We shall also use this example to illustrate how one can find instances of a CSP manually. When we introduced n-SudoQueens in section 5.11.8, we did not know for sure whether this CSP was not too constrained to have instances, at least for small values of n. So we tried to find instances for increasing values of n. As mentioned in that section, there are no instances for n = 2 or n = 4. But we found the instance in Figure 5.15 for n = 9, by a heuristic technique of adding queens progressively in the cell that is linked to the fewest other cells, so that we destroy fewer possibilities for the next ones. We started by cells in the two main diagonals, as close as possible to a corner (lesser destruction). When we reached the situation in Figure 7.8 (three queens given, in cells r1c1, r2c8 and r3c5), block b5 had only two possibilities left; r5c4 is linked to 11 available cells and r5c6 to 12; so we chose to put a queen in r5c4; but we were unable to find a solution. We then tried to prove that r5c4 was impossible; this is how we found the following g-whip[4] and a first resolution path showing that there is a unique solution.

**g-whip[4]: c6{r7 r89} – b9{r9c7 r9c9} – b6{r4c9 r4c7} – r4{c2 .} ⇒ ¬ r5c4 (A eliminated)**

In this g-whip: r5c6 and r7c6 are z-candidates for Xr6 (1$^{st}$ cell); r8c7 is both a z- and a t-candidate for Xb9 (2$^{nd}$ cell); r5c6 is both a z- and a t- candidate for Xb6; r6c7, r4c9 and r6c9 and t-candidates for Xb6 (3$^{rd}$ cell); r4c3 and r5c2 are t-candidates for Xr4 (last cell).

In Figure 7.8, in addition to our previous conventions, the characters in bold in a cell mean the following:
"**-0**" : the z-candidates of this g-whip;
"**+n**" the right-linking candidate or g-candidate for the n-th CSP variable;
"**-n**" the candidates linked to the n-th previous right-linking pattern in the n-th cell; they can be left-linking or t-candidates for the (n+1)-th CSP variables.

[We keep this g-whip example here only for illustrative purposes. Later, we found a simpler pattern, a whip[3], for the same elimination:

***** Manual solution *****
**whip[3]: b4{r5c2 r4c2} – b9{r9c7 r8c9} – b6{r6c9 .} ⇒ ¬ r5c4 (A eliminated)**
;;; the sequel has nothing noticeable:
single in block b5: r5c6
whip[1]: c4{r9 .} ⇒ ¬r9c3 (B eliminated)
single in block b7: r7c2; single in column c3: r4c3; single in column c4: r8c4; single in row r6: r6c9;
single in row r9: r9c7
Solution found in gW$_4$ (The solution is given in Figure 5.15.) ]

**Part Three**

# BEYOND G-WHIPS AND G-BRAIDS

# 8. Subset rules in a general CSP

This chapter has the two complementary goals of defining elementary Subset rules in any CSP and of showing that whips, g-whips, braids and g-braids subsume "almost all" the instances of these rules. This is not to mean that such elementary Subset rules (that are globally much weaker than whips) should not be preferred to chain rules when they can be applied; on the contrary, they may provide a shorter or a better understandable solution. But, when merely added to them, they do not bring much more resolution power; things are different when they are combined, as they will be in chapter 9, with the general "zt-ing" technique of whips and braids. Preparing the introduction of such combinations is the third goal of this chapter.

For the Subsets of size greater than two, we pay particular attention to the definitions: we want them to be comprehensive enough to get the broadest coverage but restrictive enough to exclude degenerated cases: for us, two Singles do not make a Pair, a Pair and a Single do not make a Triplet, a Triplet and a Single do not make a Quad, two Pairs do not make a Quad, … This modelling choice is consistent with what has already been done in the Sudoku case in *HLS1*, but it is now also closely related to how these patterns can be assigned a well defined "size" and ranked with respect to the $W_n$, $B_n$, $gW_n$ and $gB_n$ hierarchies; this will be essential in chapter 9 when we take them as building blocks of "$S_p$-whips" and "$S_p$-braids".

In sections 8.2 to 8.4, we define an $S_p$-subset rule in the general CSP framework (for p = 2, p = 3 and p = 4 – corresponding respectively to Pairs, Triplets and Quads) and we illustrate it by the classical form it takes in Sudoku, depending on which families of CSP variables one considers. For Sudoku, we write the Subsets in rows and leave it to the reader to write the corresponding Subsets in columns and in blocks (e.g. using meta-theorems 4.1 and 4.3 on symmetry and analogy). We give both the English and the formal logic statements and we insist once more on the symmetry and super-symmetry relationships between Naked, Hidden and Super-Hidden Subsets of same size (see Figure 8.1). Subsets are the simplest example of how the general CSP framework unifies, in a still stronger way than the mere symmetry relationships already present in *HLS1*, patterns that would otherwise be considered as different: ***in the CSP framework, Naked, Hidden and Super-Hidden Subset rules are not only related by symmetry relationships (for Subsets of given size), they are the very same rule***. (Symmetry, super-symmetry and analogy of rules have already been illustrated in this book by whips and braids, but in a different, more powerful, way: they use only basic predicates having these properties.)

Though they were not formulated in CSP terms, all the classical Subset rules of sections 8.2 to 8.4 (except the Special Quads) were present in *HLS1*, in their Sudoku specific form. But our perspective here is different: we are less concerned with these patterns for themselves than with their relationship with whips and braids – whence the general subsumption theorems of section 8.6 and the choice of examples in section 8.7, mainly centred on showing rare cases not covered by subsumption.

## 8.1. Transversality, $S_p$-labels and $S_p$-links

In the same way as, in chapter 7, we had to introduce a distinction between g-labels (defined as maximal sets of labels) and g-candidates (that did not have to be maximal), we must now introduce a distinction between:

– $S_p$-labels, that can only refer to CSP variables and transversal sets of labels (which can be considered as a saturation or maximality condition for $S_p$-labels),

– and $S_p$-subsets, in which considerations about mandatory and optional candidates will appear.

### 8.1.1. Set of labels transversal to a set of CSP variables

Definition: for p>1, given a set of p different CSP variables $\{V_1, V_2, …, V_p\}$, we say that a non-empty set S of at most p different labels is *transversal* with respect to $\{V_1, V_2, …, V_p\}$ for constraint c if:

– none of these labels has a representative with two of these CSP variables;

– all these labels are pairwise linked by c;

– S is maximal, in the sense that no label pertaining to one of these CSP variables could be added to it without contradicting the first two conditions.

Remarks:

– the first condition will always be true for pairwise strongly disjoint CSP variables, i.e. CSP variables such that no two of them share a label; but we do not adopt this stronger condition on CSP variables; adopting it would not change the general theory (for Subsets in the present chapter and for Reversible-$S_p$-chains, $S_p$-whips and $S_p$-braids in chapter 9) and it would not restrict the applications to Sudoku; but it may restrict the applications to others CSPs; moreover, the corresponding definition for g-Subsets in chapter 10 would restrict the applications, even for Sudoku (see the example in section 10.3).

– the second condition could be generalised by allowing labels in the transversal set to be pairwise linked by different constraints. In LatinSquare or Sudoku, due to the theorems proven in chapter 11 of *HLS1*, such pairwise constraints can always be replaced by a global constraint as in the present definition; this is also obviously true in N-Queens. In case a CSP had a transversal set that could not be defined via a unique constraint, we think modelling choices should be investigated. Anyway, the apparently more general condition would not change the theory developed in this chapter and in chapter 9 (it is nowhere used in the proofs) – although it may have a noticeable negative impact on the complexity of any possible implementation.

Typical examples of transversal sets of labels occur when the CSP can be represented on a k dimensional grid and two candidates differing by only one coordinate are contradictory, as can be illustrated by the Sudoku or LatinSquare examples: given CSP variables Xrc1 and Xrc2, {<Xrc1, n°>, <Xrc2, n°>} is a transversal set of labels, for any fixed Number n°; given CSP variables Xrn1 and Xrn2, {<Xrn1, c°>, <Xrn2, c°>} is also a transversal set of labels for any fixed Column c°… But there is no reason to restrict the above definition to such cases of "geometrical transversality". In particular, a transversal set of labels does not have to be associated with a "transversal" CSP variable (in the sense that, e.g. in Sudoku, variable Xc°n° could be called transversal to variable Xr°n°): in N-Queens, given two CSP variables $Xr_1$ and $Xr_2$ corresponding to different rows, the set of intersections of any diagonal (which is not associated with any CSP variable) with these rows defines a transversal set of labels (see section 8.8.1 for an example).

### 8.1.2. $S_p$-labels and $S_p$-links

Definitions: for any integer p>1, an $S_p$-label is a couple of data: {CSPVars, TransvSets}, where CSPVars is a set of p different CSP variables and TransvSets is a set of p different transversal sets of labels for these variables (each one for a well defined constraint). An S-label is an $S_p$-label for some p >1.

Definition: a label l is $S_p$-linked or simply S-linked to an $S_p$-label S = {CSPVars, TransvSets} if there is some k, 1≤k≤p, such that l is linked by the constraint $c_k$ of $TransvSets_k$ to all the labels of $TransvSets_k$ (where $TransvSets_k$ is the k-th element of TransvSets). In these conditions, l is also called *a potential target of the $S_p$-label*.

Miscellaneous remarks:
– with this definition, a label and a g-label are not $S_p$-labels (due to the condition p>1); for labels, this is a mere matter of convention, but this choice is more convenient for the sequel;
– as a result of this condition, there may be CSPs with no $S_p$-labels for some p;
– different transversal sets in the $S_p$-label are not required to be disjoint;
– in a sense, an $S_p$-label specifies the maximal extent of a possible $S_p$-subset (as defined below), but it does not tackle non-degeneracy conditions.

Notation: in the forthcoming definition of Subsets, we shall need a means of specifying that, in some transversal sets, some labels must exist while others may exist or not. We shall write this as e.g. {<$V_1$, $v_1$>, <$V_2$, $v_2$>, …, (<$V_k$, $v_k$>), ….}.

This should be understood as follows: a label not surrounded with parentheses must exist; a pseudo-label surrounded with parentheses may exist or not; if it exists, then it is named $<V_k, v_k>$.

## 8.2. Pairs

### 8.2.1. Pairs in a general CSP

Definition: in any resolution state RS of any CSP, a *Pair* (or $S_2$-*subset*) is an $S_2$-label {CSPVars, TransvSets}, where:
– CSPVars = $\{V_1, V_2\}$,
– TransvSets is composed of the following transversal sets of labels:
 $\{<V_1, v_{11}>, <V_2, v_{21}>\}$ for constraint $c_1$,
 $\{<V_1, v_{12}>, <V_2, v_{22}>\}$ for constraint $c_2$,
such that:
– in RS, $V_1$ and $V_2$ are disjoint, i.e. they share no candidate;
– $<V_1, v_{11}> \neq <V_1, v_{12}>$ and $<V_2, v_{22}> \neq <V_2, v_{21}>$;
– in RS, $V_1$ has the two mandatory candidates $<V_1, v_{11}>$ and $<V_1, v_{12}>$ and no other candidate;
– in RS, $V_2$ has the two mandatory candidates $<V_2, v_{22}>$ and $<V_2, v_{21}>$ and no other candidate.

A *target of a Pair* is defined as a candidate $S_2$-linked to the underlying $S_2$-label.

**Theorem 8.1 ($S_2$ rule): in any CSP, a target of a Pair can be eliminated.**

Proof: as the two transversal sets play similar roles, we can suppose that Z is linked to both $<V_1, v_{11}>$ and $<V_2, v_{21}>$. If Z was True, these candidates would be eliminated by ECP. As $V_1$ and $V_2$ have only two candidates each, their other candidate ($<V_1, v_{12}>$, respectively $<V_2, v_{22}>$) would be asserted by S, which is contradictory, as they are linked. Notice that the proof works only because $V_1$ and $V_2$ share no candidate in RS (and therefore in no posterior resolution state).

The rest of this section shows how, choosing pairs of variables in different sub-families of CSP variables, the familiar Naked Pairs, Hidden Pairs and Super-Hidden Pairs (X-Wing) of Sudoku (or LatinSquare) appear as mere Pairs in the above defined sense.

### 8.2.2. Naked Pairs in Sudoku

For the definition of Naked Pairs, there can be no ambiguity and we adopt the standard formulation. Naked Pairs in a row, or NP(row), is the following rule:

if there is a row r and there are two different columns $c_1$ and $c_2$ and two different numbers $n_1$ and $n_2$, such that:
- the candidates for cell (r, $c_1$) are exactly the two numbers $n_1$ and $n_2$,
- the candidates for cell (r, $c_2$) are exactly the two numbers $n_1$ and $n_2$,
then eliminate the two numbers $n_1$ and $n_2$ from the candidates for any other rc-cell in row r in rc-space.

Validity is very easy to prove directly from this (almost) standard formulation of the problem: in row r, each of the two cells defined by columns $c_1$ and $c_2$ must get a value and only two values ($n_1$ and $n_2$) are available for them, which entails that, whatever distribution is made between them of these two values, none of these two values remains available for the other cells in the same row.

The logical formulation strictly parallels the English one (except that, as is often the case, something which is formulated in natural language as "if there exists a row …", which should apparently translate into an existential quantifier, must be written with a universal quantifier):

$\forall r \forall_{\neq}(c_1,c_2) \forall_{\neq}(n_1,n_2)$
    { candidate($n_1$, r, $c_1$) ∧ candidate($n_2$, r, $c_1$) ∧
    candidate($n_2$, r, $c_2$) ∧ candidate($n_1$, r, $c_2$) ∧
    $\forall c \in \{c_1, c_2\} \forall n \neq n_1, n_2$ ¬candidate(n, r, c)
⇒
    $\forall c \neq c_1, c_2 \ \forall n \in \{n_1, n_2\}$ ¬candidate(n, r, c) }.

Exercise: show that this is exactly what Pairs of the general definition give when applied to CSP variables $Xrc_1$ and $Xrc_2$, with transversal sets defined by CSP variables (considered as constraints) $Xrn_1$ and $Xrn_2$.

### *8.2.3. Hidden Pairs in Sudoku*

If we apply meta-theorem 4.2 to Naked Pairs in a row, permuting the words "number" and "column", we obtain the rule for Hidden Pairs in a row, or HP(row) (once transposed into rn-space, a Hidden Pairs in a row looks graphically like a Naked Pairs in a row would in rc-space):
if there is a row r and there are two different numbers $n_1$ and $n_2$ and two different columns $c_1$ and $c_2$, such that:
- the candidates (columns) of rn-cell (r, $n_1$) (in rn-space) are exactly $c_1$ and $c_2$,
- the candidates (columns) of rn-cell (r, $n_2$) (in rn-space) are exactly $c_1$ and $c_2$,
then eliminate the two columns $c_1$ and $c_2$ from the candidates for any other rn-cell (r, n) in row r in rn-space.

$\forall r \forall_{\neq}(n_1,n_2) \forall_{\neq}(c_1,c_2)$
    { candidate($n_1$, r, $c_1$) ∧ candidate($n_1$, r, $c_2$) ∧
    candidate($n_2$, r, $c_2$) ∧ candidate($n_2$, r, $c_1$) ∧

$\forall n \in \{n_1, n_2\} \forall c \neq c_1, c_2 \ \neg candidate(n, r, c)$
$\Rightarrow$
$\forall n \neq n_1, n_2 \forall c \in \{c_1, c_2\} \ \neg candidate(n, r, c) \}.$

Exercise: show that this is exactly what Pairs of the general definition give when applied to CSP variables $Xrn_1$ and $Xrn_2$, with transversal sets defined by CSP variables (considered as constraints) $Xrc_1$ and $Xrc_2$.

### 8.2.4. Super Hidden Pairs in Sudoku (X-Wing)

This is not yet the full story: one can iterate the application of meta-theorem 4.2 and a rule SHP(row) can be obtained from rule HP(row) by permuting the words "row" and "number". Let us first do this permutation formally, i.e. by applying the $S_{rn}$ transform to HP(row) = $S_{cn}$(NP(row)). We get the logical formulation for Super Hidden Pairs in rows, or SHP(row):

$\forall n \forall \neq (r_1, r_2) \forall \neq (c_1, c_2)$
  $\{ candidate(n, r_1, c_1) \wedge candidate(n, r_1, c_2) \wedge$
  $candidate(n, r_2, c_2) \wedge candidate(n, r_2, c_1) \wedge$
  $\forall r \in \{r_1, r_2\} \forall c \neq c_1, c_2 \ \neg candidate(n, r, c)$
$\Rightarrow$
  $\forall r \neq r_1, r_2 \forall c \in \{c_1, c_2\} \ \neg candidate(n, r, c) \}.$

Let us now try to understand the result, with a strict English transcription:
if there is a number n and there are two different rows $r_1$ and $r_2$ and two different columns $c_1$ and $c_2$ such that:
- the candidates (columns) of rn-cell ($r_1$, n) (in rn-space) are $c_1$ and $c_2$ and no other column,
- the candidates (columns) of rn-cell ($r_2$, n) (in rn-space) are $c_1$ and $c_2$ and no other column,
then eliminate the two columns $c_1$ and $c_2$ from the candidates (columns) for any other rn-cell (r, n) in column n in rn-space.

Exercise: show that this is exactly what Pairs of the general definition give when applied to CSP variables $Xr_1n$ and $Xr_2n$, with transversal sets defined by CSP variables (considered as constraints) $Xc_1n$ and $Xc_2n$.

As the meaning of this rule is not absolutely clear in rc-space, let us make it more explicit with a new equivalent formulation based on rc-space: if there is a number n and there are two different rows $r_1$ and $r_2$, such that, in these rows, n appears as a candidate in and only in columns $c_1$ and $c_2$, then, in any of these two columns, eliminate n from the candidates for any row other than $r_1$ and $r_2$. We find the usual formulation of X-Wing in rows. Finally, we have shown that the familiar X-Wing in rows is the super-hidden version of Naked Pairs in a row: SHP(row) ≡ $S_{rn}$(HP(row)) ≡ $S_{rn}$($S_{cn}$(NP(row))) = X-Wing(row).

### 8.3. Triplets

*8.3.1. Triplets in a general CSP*

There may be several formulations of Triplets. Here, we adopt one (cyclic form) that is neither too restrictive (the presence of some of the candidates potentially involved is not mandatory) nor too comprehensive (by making mandatory the presence of some of the candidates involved, it excludes degenerated cases). The justification was done in *HLS1* for Sudoku, but it is valid for the general CSP.

Definition: in any resolution state RS of any CSP, a *Triplet* (or $S_3$-subset) is an $S_3$-label {CSPVars, TransvSets}, where:
– CSPVars = {$V_1$, $V_2$, $V_3$},
– TransvSets is composed of the following transversal sets of labels:
  {<$V_1$, $v_{11}$>, (<$V_2$, $v_{21}$>), <$V_3$, $v_{31}$>} for constraint $c_1$,
  {<$V_1$, $v_{12}$>, <$V_2$, $v_{22}$>, (<$V_3$, $v_{32}$>)} for constraint $c_2$,
  {(<$V_1$, $v_{13}$>), <$V_2$, $v_{23}$>, <$V_3$, $v_{33}$>} for constraint $c_3$,
such that:
– in RS, $V_1$, $V_2$ and $V_3$ are pairwise disjoint, i.e. no two of these variables share a candidate;
– <$V_1$, $v_{11}$> ≠ <$V_1$, $v_{12}$>, <$V_2$, $v_{22}$> ≠ <$V_2$, $v_{23}$> and <$V_3$, $v_{33}$> ≠ <$V_3$, $v_{31}$>;
– in RS, $V_1$ has the two mandatory candidates <$V_1$, $v_{11}$> and <$V_1$, $v_{12}$>, one optional candidate <$V_1$, $v_{13}$> (supposing this label exists) and no other candidate;
– in RS, $V_2$ has the two mandatory candidates <$V_2$, $v_{22}$> and <$V_2$, $v_{23}$>, one optional candidate <$V_2$, $v_{21}$> (supposing this label exists) and no other candidate;
– in RS, $V_3$ has the two mandatory candidates <$V_3$, $v_{33}$> and <$V_3$, $v_{31}$>, one optional candidate <$V_3$, $v_{32}$> (supposing this label exists) and no other candidate.

A *target of a Triplet* is defined as a candidate $S_3$-linked to the underlying $S_3$-label.

***Theorem 8.2 ($S_3$ rule): in any CSP, a target of a Triplet can be eliminated.***

Proof: as the three transversal sets play similar roles, we can suppose that Z is linked to the first, i.e. to <$V_1$, $v_{11}$>, <$V_2$, $v_{21}$> (and <$V_3$, $v_{31}$> if it exists). If Z was True, these candidates (if they are present) would be eliminated by ECP. Each of $V_1$, $V_2$ and $V_3$ would have at most two candidates left. Any choice for $V_1$ would reduce to at most one the number of possibilities for each of $V_2$ and $V_3$ (due to the pairwise contradictions between members of each transversal set). Finally, the unique choice for $V_2$, if any, would in turn reduce to zero the number of possibilities for $V_3$.

The rest of this section shows how, choosing sets of three variables in different sub-families of CSP variables, the familiar Naked Triplets, Hidden Triplets and Super-Hidden Triplets (Swordfish) of Sudoku all appear as mere Triplets of the general CSP.

### *8.3.2. Naked Triplets in Sudoku*

There may be several definitions of Naked Triplets (see *HLS1* for a discussion). Here, we adopt the same as in *HLS1*, neither too restrictive nor too comprehensive (i.e. it does not allow degenerated cases). Naked Triplets in a row or NT(row):
if there is a row r and there are three different columns $c_1$, $c_2$ and $c_3$ and three different numbers $n_1$, $n_2$ and $n_3$, such that:
- cell $(r, c_1)$ has $n_1$ and $n_2$ among its candidates,
- cell $(r, c_2)$ has $n_2$ and $n_3$ among its candidates,
- cell $(r, c_3)$ has $n_3$ and $n_1$ among its candidates,
- none of the cells $(r, c_1)$, $(r, c_2)$ and $(r, c_3)$ has any candidate other than $n_1$, $n_2$ or $n_3$,
then eliminate the three numbers $n_1$, $n_2$ and $n_3$ from the candidates for any other cell in row r in rc-space.

$\forall r \forall_{\neq}(c_1,c_2,c_3) \forall_{\neq}(n_1,n_2,n_3)$
   { candidate$(n_1, r, c_1) \wedge$ candidate$(n_2, r, c_1) \wedge$
     candidate$(n_2, r, c_2) \wedge$ candidate$(n_3, r, c_2) \wedge$
     candidate$(n_3, r, c_3) \wedge$ candidate$(n_1, r, c_3) \wedge$
     $\forall c \in \{c_1, c_2, c_3\} \forall n \neq n_1, n_2, n_3 \, \neg$candidate$(n, r, c)$
  $\Rightarrow$
     $\forall c \neq c_1, c_2, c_3 \; \forall n \in \{n_1, n_2, n_3\} \, \neg$candidate$(n, r, c)$ }.

Exercise: show that this is exactly what Triplets of the general definition give when applied to CSP variables $Xrc_1$, $Xrc_2$ and $Xrc_3$, with transversal sets defined by CSP variables (considered as constraints) $Xrn_1$, $Xrn_2$ and $Xrn_3$.

### *8.3.3. Hidden Triplets in Sudoku*

If we apply meta-theorem 4.2 to Naked Triplets in a row, permuting the words "number" and "column", we obtain the rule for Hidden Triplets in a row, or HT(row):
if there is a row r, and there are three different numbers $n_1$, $n_2$ and $n_3$ and three different columns $c_1$, $c_2$ and $c_3$, such that:
- rn-cell $(r, n_1)$ (in in rn-space) has $c_1$ and $c_2$ among its candidates (columns),
- rn-cell $(r, n_2)$ (in in rn-space) has $c_2$ and $c_3$ among its candidates (columns),
- rn-cell $(r, n_3)$ (in in rn-space) has $c_3$ and $c_1$ among its candidates (columns),
- none of the rn-cells $(r, n_1)$, $(r, n_2)$ and $(r, n_3)$ (in in rn-space) has any remaining candidate (column) other than $c_1$, $c_2$ and $c_3$,
then eliminate the three columns $c_1$, $c_2$ and $c_3$ from the candidates for any other rn-cell $(r, n)$ in row r in rn-space.

$\forall r \forall_{\neq}(n_1,n_2,n_3)\forall_{\neq}(c_1,c_2,c_3)$
 { candidate($n_1$, r, $c_1$) $\wedge$ candidate($n_1$, r, $c_2$) $\wedge$
 candidate($n_2$, r, $c_2$) $\wedge$ candidate($n_2$, r, $c_3$) $\wedge$
 candidate($n_3$, r, $c_3$) $\wedge$ candidate($n_3$, r, $c_1$) $\wedge$
 $\forall n \in \{n_1, n_2, n_3\} \forall c \neq c_1,c_2,c_3 \neg$candidate(n, r, c)
$\Rightarrow$
 $\forall n \neq n_1,n_2,n_3 \forall c \in \{c_1, c_2, c_3\} \neg$candidate(n, r, c) }.

Exercise: show that this is exactly what Triplets of the general definition give when applied to CSP variables $Xrn_1$, $Xrn_2$ and $Xrn_3$, with transversal sets defined by CSP variables (considered as constraints) $Xrc_1$, $Xrc_2$ and $Xrc_3$.

### 8.3.4. Super Hidden Triplets in Sudoku (Swordfish)

As in the case of Pairs, one can iterate the application of meta-theorem 4.2 and a rule SHT(row) can be obtained from rule HT(row) by permuting the words "row" and "number". If we apply the $S_{rn}$ transform to HT(row) = $S_{cn}$(NT(row)), we get the logical formulation of Super Hidden Triplets in rows, or SHT(row):

$\forall n \forall_{\neq}(r_1,r_2,r_3)\forall_{\neq}(c_1,c_2,c_3)$
 { candidate(n, $r_1$, $c_1$) $\wedge$ candidate(n, $r_1$, $c_2$) $\wedge$
 candidate(n, $r_2$, $c_2$) $\wedge$ candidate(n, $r_2$, $c_3$) $\wedge$
 candidate(n, $r_3$, $c_3$) $\wedge$ candidate(n, $r_3$, $c_1$) $\wedge$
 $\forall r \in \{r_1, r_2, r_3\} \forall c \neq c_1,c_2,c_3 \neg$candidate(n, r, c)
$\Rightarrow$
 $\forall r \neq r_1,r_2,r_3 \forall c \in \{c_1, c_2, c_3\} \neg$candidate(n, r, c) }.

Let us now try to understand the result, first with a direct English transliteration: if there is a number n, and there are three different rows $r_1$, $r_2$ and $r_3$ and three different columns $c_1$, $c_2$ and $c_3$, such that:
- rn-cell ($r_1$, n) (in rn-space) has $c_1$ and $c_2$ among its candidates (columns),
- rn-cell ($r_2$, n) (in rn-space) has $c_2$ and $c_3$ among its candidates (columns),
- rn-cell ($r_3$, n) (in rn-space) has $c_3$ and $c_1$ among its candidates (columns),
- none of the rn-cells ($r_1$, n), ($r_2$, n) and ($r_3$, n) (in rn-space) has any candidate (column) other than $c_1$, $c_2$ and $c_3$,
then eliminate the three columns $c_1$, $c_2$ and $c_3$ from the candidates (columns) for any other rn-cell (r, n) in column n in rn-space in rn-space .

Exercise: show that this is exactly what Triplets of the general definition give when applied to CSP variables $Xr_1n$, $Xr_2n$ and $Xr_3n$, with transversal sets defined by CSP variables (considered as constraints) $Xc_1n$, $Xc_2n$ and $Xc_3n$.

As this is not yet very explicit, let us try to clarify it by expressing it in rc-space and by temporarily forgetting part of the conditions: if there is a number n and there are three different rows $r_1$, $r_2$ and $r_3$ and three different columns $c_1$, $c_2$ and $c_3$, such that for each of the three rows the instance of number n that must be somewhere in each of these rows can actually be only in either of the three columns, then in any of the three columns eliminate n from the candidates for any row different from the given three.

What we find is the usual formulation of the rule for Swordfish in rows. There remains one point: the part of the conditions we have temporarily discarded. It is precisely what prevents Swordfish in rows from reducing to X-Wing in rows.

## 8.4. Quads

### 8.4.1. Quads in a general CSP

Finding the proper formulation for Quads, guaranteeing that it covers no degenerated case, is less obvious than for Triplets. Indeed, the simplest way is to introduce two types of Quads: Cyclic and Special. (In order to avoid technicalities, we shall show that there can only be these two types for the Sudoku CSP, but the analysis can be transposed to the general framework.) We choose to write the Special Quad in such a way that it does not cover any case already covered by the Cyclic Quad. If we wanted to introduce larger Subsets, though one could always write a general formula expressing non-degeneracy (which would lead to computationally very inefficient implementations), it would get harder and harder to write an explicit (more efficient) list of non-degenerated subcases. [As we shall see soon, in the 9×9 Sudoku case, this would be useless.]

Definition: in any resolution state RS of any CSP, a *Cyclic Quad* (or *Cyclic $S_4$-subset*) is an $S_4$-label {CSPVars, TransvSets}, where:

– CSPVars = {$V_1$, $V_2$, $V_3$, $V_4$},

– TransvSets is composed of the following transversal sets of labels:

{<$V_1$, $v_{11}$>, (<$V_2$, $v_{21}$>), (<$V_3$, $v_{31}$>), <$V_4$, $v_{41}$>} for constraint $c_1$,
{<$V_1$, $v_{12}$>, <$V_2$, $v_{22}$>, (<$V_3$, $v_{32}$>), (<$V_4$, $v_{42}$>)} for constraint $c_2$,
{(<$V_1$, $v_{13}$>), <$V_2$, $v_{23}$>, <$V_3$, $v_{33}$>, (<$V_4$, $v_{43}$>)} for constraint $c_3$,
{(<$V_1$, $v_{14}$>), (<$V_2$, $v_{24}$>), <$V_3$, $v_{34}$>, <$V_4$, $v_{44}$>} for constraint $c_4$,

such that:

– in RS, $V_1$, $V_2$, $V_3$ and $V_4$ are pairwise disjoint, i.e. no two of these variables share a candidate;

– <$V_1$, $v_{11}$> ≠ <$V_1$, $v_{12}$>, <$V_2$, $v_{22}$> ≠ <$V_2$, $v_{23}$>, <$V_3$, $v_{33}$> ≠ <$V_3$, $v_{34}$> and <$V_4$, $v_{44}$> ≠ <$V_4$, $v_{41}$>;

– in RS, $V_1$ has the two mandatory candidates <$V_1$, $v_{11}$> and <$V_1$, $v_{12}$>, two optional candidates <$V_1$, $v_{13}$> and <$V_1$, $v_{14}$> (supposing any of these labels exists) and no other candidate,

– in RS, $V_2$ has the two mandatory candidates $<V_2, v_{22}>$ and $<V_2, v_{23}>$, two optional candidates $<V_2, v_{24}>$ and $<V_2, v_{21}>$ (supposing any of these labels exists) and no other candidate,

– in RS, $V_3$ has the two mandatory candidates $<V_3, v_{33}>$ and $<V_3, v_{34}>$, two optional candidates $<V_3, v_{31}>$ and $<V_3, v_{32}>$ (supposing any of these labels exists) and no other candidate,

– in RS, $V_4$ has the two mandatory candidates $<V_4, v_{44}>$ and $<V_4, v_{41}>$, two optional candidates $<V_4, v_{42}>$ and $<V_4, v_{43}>$ (supposing any of these labels exists) and no other candidate.

Definition: in any resolution state RS of any CSP, a *Special Quad* (or *Special $S_4$-subset*) is an $S_4$-label {CSPVars, TransvSets}, where:

– CSPVars = $\{V_1, V_2, V_3, V_4\}$,

– TransvSets is composed of the following transversal sets of labels:
$\{<V_1, v_{11}>, <V_2, v_{21}>, <V_3, v_{31}>, (<V_4, v_{41}>)\}$ for constraint $c_1$,
$\{<V_1, v_{12}>, (<V_2, v_{22}>), (<V_3, v_{32}>), <V_4, v_{42}>\}$ for constraint $c_2$,
$\{(<V_1, v_{13}>), <V_2, v_{23}>, (<V_3, v_{33}>), <V_4, v_{43}>\}$ for constraint $c_3$,
$\{(<V_1, v_{14}>), (<V_2, v_{24}>), <V_3, v_{34}>, <V_4, v_{44}>\}$ for constraint $c_4$,

such that:

– in RS, $V_1$, $V_2$, $V_3$ and $V_4$ are pairwise disjoint, i.e. no two of these variables share a candidate;

– $<V_1, v_{11}> \neq <V_1, v_{12}>$, $<V_2, v_{21}> \neq <V_2, v_{23}>$ and $<V_3, v_{31}> \neq <V_3, v_{34}>$; moreover $<V_4, v_{42}>$, $<V_4, v_{43}>$ and $<V_4, v_{44}>$ are pairwise different;

– in RS, $V_1$ has the two mandatory candidates $<V_1, v_{11}>$ and $<V_1, v_{12}>$ and no other candidate;

– in RS, $V_2$ has the two mandatory candidates $<V_2, v_{21}>$ and $<V_2, v_{23}>$ and no other candidate;

– in RS, $V_3$ has the two mandatory candidates $<V_3, v_{31}>$ and $<V_3, v_{34}>$ and no other candidate;

– in RS, $V_4$ has the three mandatory candidates $<V_4, v_{42}>$, $<V_4, v_{43}>$ and $<V_4, v_{44}>$ and no other candidate.

In both cases, a *target of a Quad* is defined as a candidate $S_4$-linked to the underlying $S_4$-label.

***Theorem 8.3 ($S_4$ rule): in any CSP, a target of a Quad can be eliminated.***

Proof for the cyclic case: as the four transversal sets play similar roles, we can suppose that Z is linked to all of $<V_1, v_{11}>$, $<V_2, v_{21}>$, $(<V_3, v_{31}>)$ and $(<V_4, v_{41}>)$. If Z was True, these candidates (if they are present) would be eliminated by ECP. Each of $V_1$, $V_2$, $V_3$ and $V_4$ would have at most three candidates left. Any choice for $V_1$ would reduce to at most two the number of possibilities for $V_2$, $V_3$ and $V_4$. Any further choice among the remaining candidates for $V_2$ would reduce to at most one the number of possibilities for $V_3$ and $V_4$. Finally the unique choice left for $V_3$, if any, would reduce to zero the number of possibilities for $V_4$.

Proof for the special case: there are four subcases (the last two of which are similar to the second):
- suppose Z is linked to all of $<V_1, v_{11}>$, $<V_2, v_{21}>$, $<V_3, v_{31}>$ (and $<V_4, v_{41}>$ if it exists). If Z was True, these candidates (if they are present) would be eliminated by ECP. Each of $V_1$, $V_2$, $V_3$, would have only one candidate left; choosing these as values would reduce to zero the number of possibilities for $V_4$.
- suppose Z is linked to all of $<V_1, v_{12}>$ (, $<V_2, v_{22}>$), ($<V_3, v_{32}>$) and $<V_4, v_{42}>$. If Z was True, $<V_1, v_{12}>$ (, $<V_2, v_{22}>$), ($<V_3, v_{32}>$) and $<V_4, v_{42}>$ would be eliminated by ECP; $<V_1, v_{11}>$ would then be asserted by S, which would eliminate $<V_2, v_{21}>$ and $<V_3, v_{31}>$. Then $<V_2, v_{23}>$ and $<V_3, v_{34}>$ would be asserted. This would leave no possibility for $V_4$.

The rest of this section shows how, choosing sets of four variables in different sub-families of CSP variables, the familiar Naked Quads, Hidden Quads and Super-Hidden Quads (Jellyfish) of Sudoku appear as mere Quads of the general CSP.

### *8.4.2. Naked Quads in Sudoku*

The good formulation for Naked Quads is a little harder to find than for Triplets.

Naked Quads in a row (first tentative formulation, sometimes called Strict Naked Quads or Complete Naked Quads): if there is a row and there are four numbers and four cells in this row whose remaining candidates are exactly these four numbers, then remove these four numbers from the candidates for the other cells in this row. But there is a major problem: it is unnecessarily restrictive and situations where it can be applied are extremely rare (actually, in 10,000,000 randomly generated minimal puzzles, we have found no example that would use this form of Quads if simpler rules, i.e. Subsets and whips of size strictly less than four, are allowed).

Naked Quads in a row (second tentative formulation, sometimes called Comprehensive Naked Quads): if there is a row and there are four numbers and four cells in this row such that all their candidates are among these four numbers, then remove these four numbers from the candidates for all the other cells in this row. But, again, it has a major problem: it includes Naked Triplets in a row, Naked Pairs in a row and even Naked Single in a row as special cases.

So, neither of the usual two formulations of the Naked Quads rule is correct according to our guiding principles. How then can one formulate it so that it is comprehensive but does not subsume any of the rules for Naked Subsets of smaller size? It is enough to make certain that the four cells have no candidate other than the four given numbers (say $n_1$, $n_2$, $n_3$ and $n_4$), that each of them has more than one candidate (it is not a Naked-Single), that no two of them have exactly the same two candidates (which would make a Naked Pairs in a row) and that no three of them form a Naked Triplets in a row. There are only two ways to satisfy these conditions.

The first, most general way is to impose candidates $n_1$ and $n_2$ for cell 1, candidates $n_2$ and $n_3$ for cell 2, candidates $n_3$ and $n_4$ for cell 3 and candidates $n_4$ and $n_1$ for cell 4. This is the "Cyclic Naked Quads". We get the final formulation of this first case, more complex than usual but with its full natural scope:

if there is a row r and there are four different columns $c_1$, $c_2$, $c_3$ and $c_4$, and four different numbers $n_1$, $n_2$, $n_3$ and $n_4$, such that:
- cell (r, $c_1$) has $n_1$ and $n_2$ among its candidates,
- cell (r, $c_2$) has $n_2$ and $n_3$ among its candidates,
- cell (r, $c_3$) has $n_3$ and $n_4$ among its candidates,
- cell (r, $c_4$) has $n_4$ and $n_1$ among its candidates,
- none of the cells (r, $c_1$), (r, $c_2$), (r, $c_3$) and (r, $c_4$) has any candidate other than $n_1$, $n_2$, $n_3$ or $n_4$,

then eliminate the four numbers $n_1$, $n_2$, $n_3$ and $n_4$ from the candidates for any other cell in row r in rc-space.

$\forall r \forall \neq (c_1,c_2,c_3,c_4) \forall \neq (n_1,n_2,n_3,n_4)$
  { candidate($n_1$, r, $c_1$) $\wedge$ candidate($n_2$, r, $c_1$) $\wedge$
    candidate($n_2$, r, $c_2$) $\wedge$ candidate($n_3$, r, $c_2$) $\wedge$
    candidate($n_3$, r, $c_3$) $\wedge$ candidate($n_4$, r, $c_3$) $\wedge$
    candidate($n_4$, r, $c_4$) $\wedge$ candidate($n_1$, r, $c_4$) $\wedge$
    $\forall c \in \{c_1, c_2, c_3, c_4\} \forall n \neq n_1,n_2,n_3,n_4$ $\neg$candidate(n, r, c)
  $\Rightarrow$
    $\forall c \neq c_1,c_2,c_3,c_4 \forall n \in \{n_1, n_2, n_3, n_4\}$ $\neg$candidate(n, r, c) }.

Exercise: show that this is exactly what Cyclic Quads of the general definition give when applied to CSP variables $Xrc_1$, $Xrc_2$, $Xrc_3$ and $Xrc_4$, with transversal sets defined by CSP variables (considered as constraints) $Xrn_1$, $Xrn_2$, $Xrn_3$ and $Xrn_4$.

The second way will be called Special Naked Quads in a row, a very rare pattern, with the following respective contents for its four cells: $\{n_1\ n_2\}$, $\{n_1\ n_3\}$, $\{n_1\ n_4\}$, $\{n_2\ n_3\ n_4\}$:

$\forall r \forall \neq (c_1,c_2,c_3,c_4) \forall \neq (n_1,n_2,n_3,n_4)$
  { candidate($n_1$, r, $c_1$) $\wedge$ candidate($n_2$, r, $c_1$) $\wedge$ $\forall n \neq n_1,n_2$ $\neg$candidate(n, r, $c_1$) $\wedge$
    candidate($n_1$, r, $c_2$) $\wedge$ candidate($n_3$, r, $c_2$) $\wedge$ $\forall n \neq n_1,n_3$ $\neg$candidate(n, r, $c_2$) $\wedge$
    candidate($n_1$, r, $c_3$) $\wedge$ candidate($n_4$, r, $c_3$) $\wedge$ $\forall n \neq n_1,n_4$ $\neg$candidate(n, r, $c_3$) $\wedge$
    candidate($n_2$, r, $c_4$) $\wedge$ candidate($n_3$, r, $c_4$) $\wedge$ candidate($n_4$, r, $c_4$)
       $\wedge$ $\forall n \neq n_2,n_3,n_4$ $\neg$candidate(n, r, $c_4$)
  $\Rightarrow$
    $\forall c \neq c_1,c_2,c_3,c_4 \forall n \in \{n_1, n_2, n_3, n_4\}$ $\neg$candidate(n, r, c) }.

Exercise: show that this is exactly what Special Quads of the general definition give when applied to CSP variables $Xrc_1$, $Xrc_2$, $Xrc_3$ and $Xrc_4$, with transversal sets defined by CSP variables (considered as constraints) $Xrn_1$, $Xrn_2$, $Xrn_3$ and $Xrn_4$.

Exercise: Transpose the above justification for the two definitions of Quads in Sudoku to the general CSP framework. (Show that there are no other possibilities than the Cyclic and Special Quads.)

### 8.4.3. Hidden Quads in Sudoku

The proper formulation of rules for Hidden Quads would not be obvious if we could not rely on super-symmetries and meta-theorem 4.2. But, if we apply meta-theorem 4.2 to Cyclic Naked Quads in a row and to Special Naked Quads in a row, permuting the words "number" and "column", we immediately obtain two rules, corresponding to what is known as "Hidden Quads in a row" in the Sudoku world:

Cyclic Hidden Quads in a row, or Cyclic HQ(row):
if there is a row r, and there are four different numbers $n_1$, $n_2$, $n_3$ and $n_4$ and four different columns $c_1$, $c_2$, $c_3$ and $c_4$, such that:
- rn-cell (r, $n_1$) (in rn-space) has $c_1$ and $c_2$ among its candidates (columns),
- rn-cell (r, $n_2$) (in in rn-space) has $c_2$ and $c_3$ among its candidates (columns),
- rn-cell (r, $n_3$) (in in rn-space) has $c_3$ and $c_4$ among its candidates (columns),
- rn-cell (r, $n_4$) (in in rn-space) has $c_4$ and $c_1$ among its candidates (columns),
- none of the rn-cells (r, $n_1$), (r, $n_2$), (r, $n_3$) and (r, $n_4$) (in in rn-space) has any remaining candidate (column) other than $c_1$, $c_2$, $c_3$ and $c_4$,
then eliminate the four columns $c_1$, $c_2$, $c_3$ and $c_4$ from the candidates for any other rn-cell (r, n) in row r in rn-space.

$\forall r \forall_{\neq}(n_1,n_2,n_3,n_4) \forall_{\neq}(c_1,c_2,c_3,c_4)$
  { candidate($n_1$, r, $c_1$) $\land$ candidate($n_1$, r, $c_2$) $\land$
    candidate($n_2$, r, $c_2$) $\land$ candidate($n_2$, r, $c_3$) $\land$
    candidate($n_3$, r, $c_3$) $\land$ candidate($n_3$, r, $c_4$) $\land$
    candidate($n_4$, r, $c_4$) $\land$ candidate($n_4$, r, $c_1$) $\land$
    $\forall n \in \{n_1, n_2, n_3, n_4\} \forall c \neq c_1,c_2,c_3,c_4$ $\neg$candidate(n, r, c)
  $\Rightarrow$
    $\forall n \neq n_1,n_2,n_3,n_4 \forall c \in \{c_1, c_2, c_3, c_4\}$ $\neg$candidate(n, r, c) }.

And Special Hidden Quads in a row, or Special HQ(row):

$\forall r \forall_{\neq}(n_1,n_2,n_3,n_4) \forall_{\neq}(c_1,c_2,c_3,c_4)$
  { candidate($n_1$, r, $c_1$) $\land$ candidate($n_1$, r, $c_2$) $\land$ $\forall c \neq c_1,c_2$ $\neg$candidate($n_1$, r, c) $\land$
    candidate($n_2$, r, $c_1$) $\land$ candidate($n_2$, r, $c_3$) $\land$ $\forall c \neq c_1,c_3$ $\neg$candidate($n_2$, r, c) $\land$
    candidate($n_3$, r, $c_1$) $\land$ candidate($n_3$, r, $c_4$) $\land \forall n \neq c_1,c_4$ $\neg$candidate($n_3$, r, c) $\land$
    candidate($n_4$, r, $c_2$) $\land$ candidate($n_4$, r, $c_3$) $\land$ candidate($n_4$, r, $c_4$) $\land$
      $\land$ $\forall c \neq c_2,c_3,c_4$ $\neg$candidate($n_4$, r, c)

$$\Rightarrow \forall n \neq n_1,n_2,n_3,n_4 \forall c \in \{c_1, c_2, c_3, c_4\} \neg candidate(n, r, c) \}.$$

Exercise: show that this is exactly what Cyclic and Special Quads of the general definition give when applied to CSP variables $Xrn_1$, $Xrn_2$, $Xrn_3$ and $Xrn_4$, with transversal sets defined by CSP variables (considered as constraints) $Xrc_1$, $Xrc_2$, $Xrc_3$ and $Xrc_4$.

### 8.4.4. Super Hidden Quads in Sudoku (Jellyfish)

Finally, there remains to consider a rule that should be called Cyclic Super Hidden Quads in rows, or SHQ(row), obtained from Cyclic Hidden Quads in a row by permuting the words "row" and "number", according to meta-theorem 4.2. Let us first do this formally, i.e. by applying the $S_{rn}$ transform to HQ(row) = $S_{cn}$(NQ(row)):

$\forall n \forall \neq (r_1,r_2,r_3,r_4) \forall \neq (c_1,c_2,c_3,c_4)$
  $\{ candidate(n, r_1, c_1) \land candidate(n, r_1, c_2) \land$
  $candidate(n, r_2, c_2) \land candidate(n, r_2, c_3) \land$
  $candidate(n, r_3, c_3) \land candidate(n, r_3, c_4) \land$
  $candidate(n, r_4, c_4) \land candidate(n, r_4, c_1) \land$
  $\forall r \in \{r_1, r_2, r_3, r_4\} \forall c \neq c_1,c_2,c_3,c_4 \neg candidate(n, r, c)$
$\Rightarrow$
  $\forall r \neq r_1,r_2,r_3,r_4 \forall c \in \{c_1, c_2, c_3, c_4\} \neg candidate(n, r, c) \}.$

Exercise: show that this is exactly what Cyclic Quads of the general definition give when applied to CSP variables $Xr_1n$, $Xr_2n$, $Xr_3n$ and $Xr_4n$, with transversal sets defined by CSP variables (considered as constraints) $Xc_1n$, $Xc_2n$, $Xc_3n$ and $Xc_4n$.

In the same way as in the Triplets case, we can clarify this rule by temporarily forgetting part of the conditions: if there is a number n and there are four different rows $r_1$, $r_2$, $r_3$ and $r_4$ and four different columns $c_1$, $c_2$, $c_3$ and $c_4$, such that for each of the four rows the instance of number n that must be somewhere in each of these rows can actually only be in either of the four columns, then in any of the four columns eliminate n from the candidates for any row different from the given four.

This is the usual formulation of the rule for Jellyfish in rows. The part we have temporarily discarded corresponds to the conditions we have added to Comprehensive Cyclic Naked Quads in a row; it is just what prevents Jellyfish in rows from reducing to X-Wing in rows or to Swordfish in rows. Finally, we have not only shown that the familiar Jellyfish in rows is the supersymmetric version of Cyclic Naked Quads in a row, but we have also found the proper way to write this rule according to our guiding principles, in as comprehensive a way as possible.

We leave it to the reader to write the rule for Special Super Hidden Quads or Special Jellyfish.

**8.5. Relations between Naked, Hidden and Super Hidden Subsets in Sudoku**

The so-called "fishy patterns" (X-Wing, Swordfish, Jellyfish, …) are very popular in the Sudoku micro-world, even the non-existent ones (such as Squirmbag, a would be Super Hidden Quintuplets in our vocabulary) and there are many very specific extensions of these patterns (such as "finned fish", "sashimi fish", … See also chapter 10 for another kind of extension).

As can be seen by looking at the logical formulæ in the previous sections, a graph similar to that in Figure 4.2 for Singles would not be enough to describe all the rules available for Subsets of size greater than one. Moreover, there is a major difference between Singles and larger Subsets: in the latter, there are different numbers of quantified variables of different sorts: Numbers, Rows and Columns. Building on these differences, the question now is, how far can one go in the iteration of theorem 4.2 and in the definition of Subset rules: Naked, Hidden, Super-Hidden, Super-Super-Hidden, …?

As for the Naked and Hidden Subsets, a well-known (and obvious) property of Subsets shows that we have found all of them: for any subset S of Numbers of size p ($1 \leq p < 9$), there is a complementary subset $S^c$ of size 9-p (with $1 \leq 9-p < 9$). And S forms a Naked Subset of size p on p cells in a row [respectively a column, a block], if and only if $S^c$ forms a Hidden Subset of size 9-p on the remaining 9-p cells in this row [resp. this column, this block]. As a result, no Naked or Hidden Subset rule for subsets of size greater than four is needed. For instance, Naked Quintuplets in a row is just Hidden Quads in the same row and Hidden Quintuplets in a row is just Naked Quads in the same row.

What was not known before *HLS1*, because super-symmetries had not been explicited, the mythical Super Hidden Quintuplets in a row (alias Squirmbag) is just Hidden Quads in a column (as shown by Figure 8.1 and the remarks above). This is a very interesting example of a named thing that had no independent existence.

Indeed, after the previous sections, several natural questions may arise, such as:

– what if, instead of applying symmetry $S_{cn}$ to NP(row), we apply symmetry $S_{rn}$?

– what if we formulate a rule analogous to X-Wing in rows but in rn-space – i.e. a rule that should be called Hidden X-Wing in rows or HXW(row) or HSHP(row)?

Do we get new unknown rules? The answer is no; the previous set of rules is strongly closed under symmetry and supersymmetry. More specifically, the full story is to be found in Figure 8.1. The first practical consequence of this is that it exempts us from looking for new types of Subset rules (but see chapter 10 for g-Subset rules). Checking the assertions of Figure 8.1 is an easy exercise about the $S_{rc}$, $S_{rn}$ and $S_{cn}$ transforms (one must just be very careful with the indices). As a detailed proof is available in *HLS*, we do not reproduce it here.

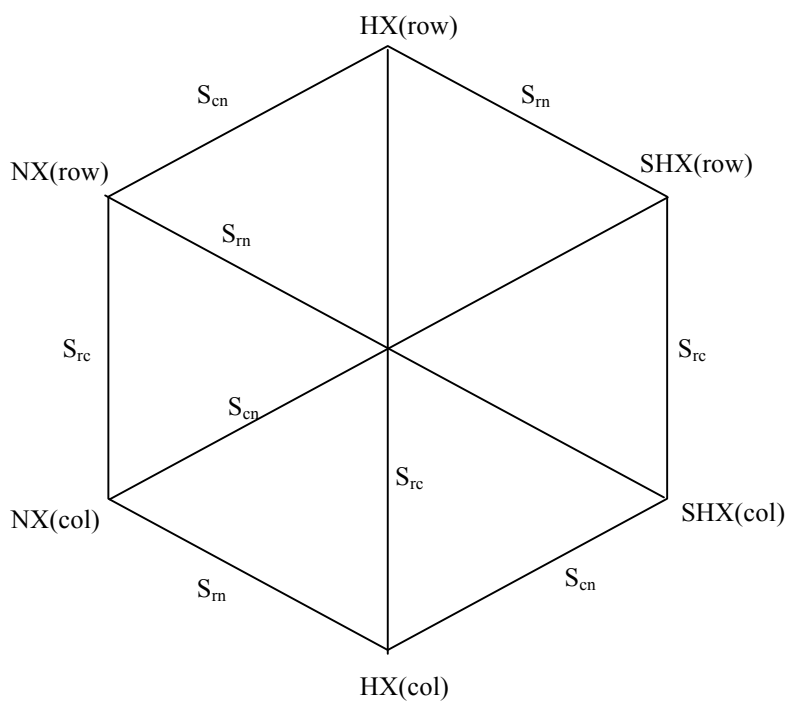

*Figure 8.1. Sudoku symmetries and supersymmetries (X = Pairs, Triplets or Quads – or Subsets of size ≤ IP(n/2) for Sudoku on n×n grids)*

[Historical note: after the first edition of *HLS*, we were informed that the idea of "another view of Fish" (i.e. of X-Wings, Swordfish and Jellyfish) had already been expressed by "Arcilla" on the late Sudoku Player's Forum, in the thread "a new (?) view of fish (naked or hidden)", November 3$^{rd}$, 2006. The same thread also shows that similar ideas had been mentioned even before, still in informal ways or in programmers jargon (e.g. "the same program can be used to find Naked Subsets and Fish"). All this was very smart, though it missed the mathematical notions of symmetry and supersymmetry and the closely related idea (first presented in *HLS1*) of introducing the four 2D (rc, rn, cn and bn) spaces and cells as first class concepts, with their associated representations in an Extended Sudoku Board. As a result, it did not develop into a global framework and it led neither to the meta-theorems of chapter 4, nor to the systematic relationships displayed in Figure 8.1 (some of which are not obvious at all), nor to the idea of hidden chains introduced in *HLS1*.]

**8.6. Subset resolution theories in a general CSP; confluence**

*8.6.1. Definition of the Subset resolution theories*

The principle of the definitions for Pairs, Triplets and Quads can easily be extended to larger Subsets, although, as we mentioned above, the conditions for non-degeneracy may be tedious to write explicitly. Given a non-degenerated Subset pattern, we define its size to be the number of CSP variables (or transversal sets) in its definition: Pairs have size 2, Triplets size 3, Quads size 4, … As should now be expected from what was done with our previous families of rules, we can define an increasing sequence of resolution theories.

Definition: In any CSP, the Subset resolution theories are defined as follows:
– $S_0 = BRT(CSP)$,
– $S_1 = W_1$,
– $S_2 = S_1 \cup$ {rules for non degenerated Pairs},
– $S_3 = S_2 \cup$ {rules for non degenerated Triplets},
– ….
– $S_{n+1} = S_n \cup$ {rules for non degenerated Subsets of size n+1},
– $S_\infty = \cup_{n \geq 0} S_n$.

Notice that, in this hierarchy of resolution theories, we put $W_1$ before Pairs; this is not only a matter of convention: as already noticed, whips of length 1 (when they exist) are the most basic pattern after Singles and it would not make much sense to define any resolution theory, apart from BRT(CSP), without them.

In 9×9 Sudoku or Latin Squares, $S_\infty = S_4$. More generally, in n×n Sudoku or Latin Squares, $S_\infty = S_p$, with p = IP(n/2) (where "IP" means the integer part).

***Theorem 8.4: in any CSP, each of the $S_n$ resolution theories is stable for confluence; therefore, it has the confluence property.***

Proof: let S be an $S_p$-subset (p≤n), for CSP variables $\{V_1, …, V_p\}$ and transversal sets (some of the labels below may be missing):
$\{<V_1, v_{11}>, <V_2, v_{21}>, …<V_p, v_{p1}>\}$
….
$\{<V_1, v_{1p}>, <V_2, v_{2p}>, …<V_p, v_{pp}>\}$

If Z is a target for S, it is linked to all the elements in some of these sets. There may happen two different events:
- if some optional candidate is eliminated from the transversal sets, what remains is still an $S_p$-subset and Z is still linked to it via the same transversal set;
- if a mandatory candidate is eliminated from a transversal set, either what remains is still an $S_p$-subset (due to the presence of the remaining optional candidates) or what remains can be split into two (or more) smaller Subsets or Singles and Z is still linked to one of them.
In any case, Z can still be eliminated by rules in $S_n$.

### *8.6.2. Complexity considerations (in Sudoku)*

When we increase the size p of Subsets (p goes up from 2 to 4 as we pass from Pairs to Quads, via Triplets), the number of possible cases in each row (forgetting the Special Quads) increases from $(9×8)^2 = 5184$ to $(9×8×7×6)^2 = 9,144,576$ (4 different columns and 4 different numbers). Multiplying this by 9 rows and by 8 patterns (3 Naked, 3 Hidden and 2 Super-Hidden), i.e. by 72, gives an idea of the increase in complexity (from 373,248 to 658,409,472). These figures can be significantly improved by ordering the columns and/or numbers (and this is essential for an effective implementation), but the relative order of magnitude remains the same. Programming Triplets and Quads as rules in a knowledge-based system is a very good exercise for AI students: they can see the importance of having a precise logical formulation before they start to code them in the specific formalism of their inference engine, they can be shown different techniques of rule optimisation and finally they can see at work Newell's famous distinction [Newell 1982] between the "knowledge level" (here a non-ambiguous English or MS-FOL formulation) and the "symbol level" (the rule in the syntax of the inference engine, where different logical conditions may have to be ordered, control facts may have to be added, different saliences, i.e. priorities of rules, may have to be introduced, ….).

### *8.6.3. Definition of the $W_n+S_n$, $gW_n+S_n$, $B_n+S_n$ and $gB_n+S_n$ theories and ratings*

Combining whips and Subsets, one can define the increasing sequence ($W_n+S_n$, $n≥0$) of resolution theories:
– $W_0+S_0 = BRT(CSP)$,
– $W_1+S_1 = W_1$,
– …
– $W_{n+1}+S_{n+1} = W_n+S_n \cup W_{n+1} \cup S_{n+1}$,
– …
– $W_\infty+S_\infty = \cup_{n≥0} W_n+S_n$.

Combining g-whips and Subsets, one can define in similar ways the increasing sequences ($gW_n+S_n$, $n≥0$), ($B_n+S_n$, $n≥0$) and ($gB_n+S_n$, $n≥0$) of resolution theories. And with each of these sequences, one can associate a rating. As a direct corollary to theorems 5.6 and 7.4 and to lemma 4.1, we get:

***Theorem 8.5: in any CSP, each of the $B_n+S_n$ and $gB_n+S_n$ resolution theories is stable for confluence; therefore, it has the confluence property.***

### 8.7. Whip subsumption results for Subset rules

After the previous definitions, this section describes the main relationships between Subsets and whips. In the Sudoku case, additional subsumption results can be found on our website for extended Subset patterns ("finned fish" and "sashimi fish"). In our opinion, this is the main section of this chapter; it establishes a strong link between the length of a whip or braid and the size of a Subset. For consistency reasons, patterns that can be seen either as whips [resp. g-whips, braids or g-braids] or as Subsets must be assigned the same W [resp. gW, B, gB] and S ratings. Moreover, the results proven here justify the *a priori* combinations (with the same n) of the $S_n$ and $W_n$, $gW_n$, $B_n$ or $gB_n$ theories used in the definitions in section 8.6.3.

#### *8.7.1. Subsumption and almost-subsumption theorems in a general CSP*

*8.7.1.1. Pairs*

**Theorem 8.6: $S_2 \subseteq W_2$ (whips of length 2 subsume all the Pairs).**

Proof: keeping the notations of theorem 8.1 and considering a target Z of the Pair that is linked to the first transversal set, the following whip eliminates Z:
whip[2]: $V_1\{v_{11}\ v_{12}\} - V_2\{v_{22}\ .\} \Rightarrow \neg$candidate(Z).

**The converse of the above theorem is false: $W_2 \not\subseteq S_2$**. For a deep understanding of whips, this is as interesting as the theorem itself. The Sudoku example in section 8.8.1 has W(P) = 2 but S(P) = 3. It also has three very instructive examples of whip[2] that cannot be considered as Pairs (or even as g-Pairs, see section 10.1.6.1.).

*8.7.1.2. Triplets*

**Theorem 8.7: $W_3$ subsumes "almost all" the Triplets.**

Proof: keeping the notations of theorem 8.2 and considering a target Z of the Triplet that is linked to the first transversal set (the three of them play similar roles), the following whip eliminates Z in any CSP:
whip[3]: $V_1\{v_{11}\ v_{12}\} - V_2\{v_{22}\ v_{23}\} - V_3\{v_{33}\ .\} \Rightarrow \neg$candidate(Z),
provided that $<V_1, v_{13}>$ is not a candidate for $V_1$.
The optional candidates of the Triplet appear in the whip as z- or t- candidates.

Considering that, in the above situation, the three CSP variables play symmetrical roles, there is only one case of a Triplet elimination that cannot be replaced by a whip[3] elimination. It occurs when the optional candidates for variables $V_1$, $V_2$ and $V_3$ in the transversal set to which the target is $S_3$-linked correspond to existing labels and are all effectively present in the resolution state.

This theorem is illustrated by the same Sudoku example as above (in section 8.8.1), whereas a Sudoku example of non-subsumption is given in section 8.8.2; it even shows that $\mathbf{S_3} \not\subset \mathbf{B_\infty}$.

Replacing whips by braids would not change the above results.

*8.7.1.3. Quads*

### Theorem 8.8: $W_4$ subsumes "almost all" the Cyclic Quads.

Keeping the notations of theorem 8.3, the following whip eliminates a target Z of the Cyclic Quad in any CSP:
whip[4]: $V_1\{v_{11}\ v_{12}\} - V_2\{v_{22}\ v_{23}\} - V_3\{v_{33}\ v_{34}\} - V_4\{v_{41}\ .\} \Rightarrow \neg candidate(Z)$,
provided that $<V_1, v_{13}>$ and $<V_1, v_{14}>$ (if they exist) are not candidates for $V_1$ and $<V_2, v_{23}>$ (if it exists) is not a candidate for $V_2$.
The optional candidates of the Quad appear in the whip as z- or t- candidates.

An exceptional example of non-subsumption for a Naked Quad elimination is given in section 8.8.3.

### Theorem 8.9: $B_4$ subsumes all the Special Quads.

Keeping the notations of theorem 8.3, let Z be a target of the Special Quad:
- if Z is linked to the first transversal set, the following braid eliminates Z:
braid[4]: $V_1\{v_{11}\ v_{12}\} - V_2\{v_{21}\ v_{23}\} - V_3\{v_{31}\ v_{34}\} - V_4\{v_{44}\ .\} \Rightarrow \neg candidate(Z)$,
in which the first three left-linking candidates are linked to Z;
- if Z is linked to another transversal set, say the second, the following whip eliminates Z:
whip[4]: $V_1\{v_{12}\ v_{11}\} - V_2\{v_{21}\ v_{23}\} - V_4\{v_{43}\ v_{44}\} - V_3\{v_{34}\ .\} \Rightarrow \neg candidate(Z)$,
in which candidate $<V_4, v_{42}>$ appears as a z candidate for the third CSP variable.

### 8.7.2. Statistical almost-subsumption results in Sudoku

The theorems in the previous subsection show that, for $n \leq 4$, "almost all" of the eliminations done by Subsets can be done by whips or braids. Can this "almost all", until now only specified by logical conditions, be given any numerical meaning? One has $W+S(P) \leq W(P)$ for any instance P and the question can be reformulated as: how frequently can the two ratings be different? Notice that this is not exactly an answer to our initial question, because equality of the ratings does not mean that the same eliminations were done; another resolution path may have been followed. Anyway, experiments with the first 10,000 random minimal puzzles in the Sudogen0 collection show that the W+S and the W ratings differ in only 8 cases: either non-subsumption cases are statistically very rare (as suggested by the above "almost subsumption" theorems) or they are well compensated by other eliminations.

*8.7.3. Comparison of the resolution power of whips and Subsets of same length*

Subsets are "almost" subsumed by whips of same length; but is there any reciprocal almost subsumption, so that both would have approximately the same resolution power? The answer is negative. The classification results in Table 8.1 show that, even with $W_1$ included in all the $S_n$ theories, Subsets have a very weak resolution power compared to whips. The W line comes from the "ctr-bias" column of Table 6.4; the S line is based on a series of 275,867 puzzles from the controlled-bias generator. Only the part of the Table in bold is meaningful for this comparison.

| rating → | 0 (BRT) | 1 ($S_1$=$W_1$) | 2 | 3 | 4 | 4<n<∞ |
|---|---|---|---|---|---|---|
| S | 35.08% | 9.82% | **5.44%** | **0.36%** | **0.011%** | **0%** |
| W | 35.08% | 9.82% | **13.05%** | **20.03%** | **17.37%** | **qsp 100%** |

*Table 8.1: non-cumulative S and W distributions for the controlled-bias generator*

One way of understanding these results is that the definition of Subsets is much more restrictive than the definition of whips of same size. In Subsets, transversal sets are defined by a single constraint. In whips, the fact of being linked to the target or to a given previous right-linking candidate plays a role very similar to each of these transversal sets. But being linked to a candidate is much less restrictive than being linked to it via a pre-assigned constraint. As shown by the almost subsumption results, the few Subset cases not covered by whips because of the restrictions on them related to sequentiality are too rarely met in practice to be able to compensate for this.

**8.8. Subsumption and non-subsumption examples from Sudoku**

This final section illustrates both subsumption and non-subsumption cases. It also shows concretely how Super Hidden Subsets can look like Naked ones in the appropriate 2D space.

*8.8.1. $W_2 \not\subset S_2$; also an example of Swordfish subsumption by a whip[3]*

Let us first prove that $W_2 \not\subset S_2$. For the puzzle P in Figure 8.2 (Royle17#18966), we shall show that $W(P) = 2$ and $S(P) = 3$.

After an initial sequence of 36 Hidden Singles, leading to the puzzle in the middle of Figure 8.2, we consider two resolution paths.

|   | 5 |   | 4 |   |   |   |   |   |
|---|---|---|---|---|---|---|---|---|
|   |   |   |   | 3 |   | 8 |   |   |
|   |   |   |   |   |   |   |   | 1 |
| 3 |   |   |   | 8 |   | 7 |   |   |
|   | 6 |   |   |   |   |   |   | 5 |
|   |   | 2 |   |   |   |   |   |   |
|   |   | 5 |   | 6 |   | 4 |   |   |
| 1 |   | 8 |   |   |   | 3 |   |   |
|   |   |   |   |   |   |   |   |   |

|   | 5 | 1 | 4 |   | 8 |   | 3 | 6 |
|---|---|---|---|---|---|---|---|---|
| 6 |   |   | 1 | 3 | 5 | 8 |   | 4 |
| 4 | 8 | 3 |   | 6 |   | 5 |   | 1 |
| 3 |   | 5 | 6 | 8 | 4 | 7 | 1 |   |
| 8 | 6 |   | 3 |   | 1 | 4 | 5 |   |
|   | 1 | 4 | 2 | 5 |   | 6 | 8 | 3 |
|   | 3 |   | 5 |   | 6 |   | 4 | 8 |
| 1 |   | 8 |   |   | 3 | 6 | 5 |   |
| 5 |   | 6 | 8 | 3 |   |   |   | 7 |

| 7 | 5 | 1 | 4 | 2 | 8 | 9 | 3 | 6 |
|---|---|---|---|---|---|---|---|---|
| 6 | 9 | 2 | 1 | 3 | 5 | 8 | 7 | 4 |
| 4 | 8 | 3 | 7 | 6 | 9 | 5 | 2 | 1 |
| 3 | 2 | 5 | 6 | 8 | 4 | 7 | 1 | 9 |
| 8 | 6 | 7 | 3 | 9 | 1 | 4 | 5 | 2 |
| 9 | 1 | 4 | 2 | 5 | 7 | 6 | 8 | 3 |
| 2 | 3 | 9 | 5 | 7 | 6 | 1 | 4 | 8 |
| 1 | 7 | 8 | 9 | 4 | 2 | 3 | 6 | 5 |
| 5 | 4 | 6 | 8 | 1 | 3 | 2 | 9 | 7 |

*Figure 8.2. Puzzle Royle17#18966: 1) original, 2) after initial Singles, 3) solution*

In the first path, using only the Subset theories, the simplest rule applicable is a Swordfish in columns (Figure 8.3); it allows four eliminations; after three have been done, Singles and ECP are enough to solve the puzzle, showing that S(P) = 3.

***** SudoRules 16.2 based on CSP-Rules 1.2, config: S *****
**swordfish-in-columns: n7{c2 c4 c8}{r2 r3 r8}** ==> r3c6 ≠ 7, r8c5 ≠ 7, r8c6 ≠ 7
 singles to the end

In the second path, using only the whip theories, the simplest applicable rules are ***three very instructive cases of whip[2] that cannot be considered as Pairs [or even as g-Pairs*** (see 10.1.6.1)], leading to eliminations unrelated to the above Swordfish; these are enough to solve the puzzle with Singles and ECP, showing that W(P) = 2.

***** SudoRules 16.2 based on CSP-Rules 1.2, config: W *****
**whip[2]: r1n7{c1 c5} – r5n7{c5 .} ==> r2c3 ≠ 7**
**whip[2]: r1n7{c5 c1} – r6n7{c1 .} ==> r3c6 ≠ 7**
**whip[2]: b4n7{r5c3 r6c1} – r1n7{c1 .} ==> r5c5 ≠ 7**
singles to the end

Now, forgetting the simple whip[2] eliminations, we can also use this example to show how a Swordfish looks like in the proper 2D space. Spotting this Swordfish in the standard representation (upper part of Figure 8.3) may be difficult because it seems to be very degenerated (three of the nine rc-cells on which it lies are even decided). However, in the cn-representation (lower part of Figure 8.3), it looks like a very incomplete Naked-Triplets, but still a non-degenerated one. Indeed, it is a hidden xy-chain[3] (defined in *HLS1* as a kind of bivalue-chain[3], but in rn- instead of rc- space, and therefore a whip[3]).

Exercise: based on the proof of theorem 8.7, write the four whips[3] allowing the eliminations of the four Swordfish targets.

|    | c1 | c2 | c3 | c4 | c5 | c6 | c7 | c8 | c9 |    |
|----|----|----|----|----|----|----|----|----|----|----|
| r1 | n2 n7 n9 | n5 | n1 | n4 | n2 n7 n9 | n8 | n2 n9 | n3 | n6 | r1 |
| r2 | n6 | n2 n7 n9 | n2 n7 n9 | n1 | n3 | n5 | n8 | n2 n7 n9 | n4 | r2 |
| r3 | n4 | n8 | n3 | n7 n9 | n6 | n2 n7 n9 | n5 | n2 n7 n9 | n1 | r3 |
| r4 | n3 | n2 n9 | n5 | n6 | n8 | n4 | n7 | n1 | n2 n9 | r4 |
| r5 | n8 | n6 | n2 n7 n9 | n3 | n7 n9 | n1 | n4 | n5 | n2 n9 | r5 |
| r6 | n7 n9 | n1 | n4 | n2 | n5 | n7 n9 | n6 | n8 | n3 | r6 |
| r7 | n2 n7 n9 | n3 | n2 n7 n9 | n5 | n1 n2 n7 n9 | n6 | n1 n2 n9 | n4 | n8 | r7 |
| r8 | n1 | n2 n4 n7 n9 | n8 | n7 n9 | n2 n4 n7 n9 | n2 n7 n9 | n3 | n6 | n5 | r8 |
| r9 | n5 | n2 n4 n9 | n6 | n8 | n1 n2 n4 n9 | n3 | n1 n2 n9 | n2 n9 | n7 | r9 |
|    | c1 | c2 | c3 | c4 | c5 | c6 | c7 | c8 | c9 |    |

|    | c1 | c2 | c3 | c4 | c5 | c6 | c7 | c8 | c9 |    |
|----|----|----|----|----|----|----|----|----|----|----|
| n1 | r8 | r6 | r1 | r2 | r7 r9 | r5 | r7 r9 | r4 | r3 | n1 |
| n2 | r1 r7 | r2 r4 r8 | r2 r5 r7 | r6 | r1 r7 r8 r9 | r3 r8 | r1 r7 r9 | r2 r3 r9 | r4 r5 | n2 |
| n3 | r4 | r7 | r3 | r5 | r2 | r9 | r8 | r1 | r6 | n3 |
| n4 | r3 | r8 r9 | r6 | r1 | r8 r9 | r4 | r5 | r7 | r2 | n4 |
| n5 | r9 | r1 | r4 | r7 | r6 | r2 | r3 | r5 | r8 | n5 |
| n6 | r2 | r5 | r9 | r4 | r3 | r7 | r6 | r8 | r1 | n6 |
| n7 | r1 r6 r7 | r2 r8 | r2 r5 r7 | r3 r8 | r1 r5 r7 r8 | r3 r6 r8 | r4 | r2 r3 | r9 | n7 |
| n8 | r5 | r3 | r8 | r9 | r4 | r1 | r2 | r6 | r7 | n8 |
| n9 | r1 r6 r7 | r2 r4 r8 r9 | r2 r5 r7 | r3 r8 | r1 r5 r7 r8 r9 | r3 r6 r8 | r1 r7 r9 | r2 r3 r9 | r4 r5 | n9 |
|    | c1 | c2 | c3 | c4 | c5 | c6 | c7 | c8 | c9 |    |

***Figure 8.3.*** *Puzzle Royle17#18966, seen in rc and cn spaces, after initial Singles have been applied. The four eliminations allowed by the Swordfish (in grey cells) are underlined.*

### 8.8.2. $S_3 \not\subset B_\infty$ : a Swordfish not subsumed by whips or braids

| | c1 | c2 | c3 | c4 | c5 | c6 | c7 | c8 | c9 | |
|---|---|---|---|---|---|---|---|---|---|---|
| r1 | n4 n5 n6 n7 n8 | n4 n6 n7 n8 | n1 | n5 n6 n7 n8 | n5 n7 n8 n9 | n2 | n6 n8 n9 | n6 n7 n8 n9 | n3 | r1 |
| r2 | n3 n5 n6 n7 n8 | n3 n6 n7 n8 | n3 n5 n7 n9 | n5 n6 n7 n8 | n1 | n3 n5 n6 n8 n9 | n2 n6 n8 n9 | n4 | n2 n6 n7 n9 | r2 |
| r3 | n2 | n3 n6 n7 n8 | n3 n7 n9 | n4 | n3 n7 n8 n9 | n3 n6 n8 n9 | n5 | n1 n6 n7 n8 n9 | n1 n6 n7 | r3 |
| r4 | n3 n4 | n1 n2 n3 n4 | n6 | n1 n2 n5 | n2 n4 n5 n9 | n7 | n1 n3 n9 | n1 n3 n5 n9 | n8 | r4 |
| r5 | n3 <u>n4</u> n7 n8 | n5 | n3 n4 n7 | n1 n6 n8 | <u>n4</u> <u>n9</u> | n4 n6 n8 n9 | n1 n3 n6 <u>n9</u> | n2 | n1 n6 n7 n9 | r5 |
| r6 | n9 | n1 n2 n7 n8 | n2 n7 | n3 | n2 n5 n8 | n5 n6 n8 | n4 | n1 n5 n6 n7 | n1 n6 n7 | r6 |
| r7 | n3 n4 n6 n7 | n2 n3 n4 n6 | n8 | n2 n7 | n2 n4 n7 | n1 | n2 n3 n6 n9 | n3 n6 n9 | n5 | r7 |
| r8 | n3 <u>n4</u> n5 n7 | n9 | n2 n3 n4 n5 n7 | n2 n5 n7 n8 | n6 | n3 n4 n5 n8 | n1 n2 n3 n8 | n1 n3 n8 | n1 n2 n4 | r8 |
| r9 | n1 | n2 n3 <u>n4</u> n6 | n2 n3 n4 n5 | n9 | n2 n3 <u>n4</u> n5 n8 | n3 n4 n5 n8 | n7 | n3 n6 n8 | n3 n2 n4 n6 | r9 |
| | c1 | c2 | c3 | c4 | c5 | c6 | c7 | c8 | c9 | |

| | c1 | c2 | c3 | c4 | c5 | c6 | c7 | c8 | c9 | |
|---|---|---|---|---|---|---|---|---|---|---|
| n1 | r9 | r4 r6 | r1 | r4 r5 | r2 | r7 | r4 r5 r8 | r3 r4 r6 r8 | r3 r5 r6 r8 | n1 |
| n2 | r3 | r4 r6 r7 r9 | r6 r7 r9 | r4 r7 r8 | r4 r6 r7 r9 | r1 | r2 r7 r8 | r5 | r2 r8 r9 | n2 |
| n3 | r2 r4 r5 r7 r8 | r2 r3 r4 r7 r9 | r2 r3 r5 r8 r9 | r6 | r3 r7 r9 | r2 r3 r8 r9 | r4 r5 r7 r8 | r4 r7 r8 r9 | r1 | n3 |
| n4 | r1 r4 <u>r5</u> r7 <u>r8</u> | r1 r4 r7 <u>r9</u> | | r5 r8 r9 | r3 | r4 <u>r5</u> r7 <u>r9</u> | r5 r8 r9 | r6 | r2 | r8 r9 | n4 |
| n5 | r1 r2 r8 | r5 | r2 r8 r9 | r1 r2 r4 r8 | r1 r4 r7 r9 | r2 r6 r8 r9 | r3 | r4 r6 | r7 | n5 |
| n6 | r1 r2 r7 | r1 r2 r3 r7 r9 | r4 | r1 r2 r4 | r8 | r2 r3 r5 r6 | r1 r2 r5 r7 | r1 r3 r6 r7 | r2 r3 r5 r6 r9 | n6 |
| n7 | r1 r2 r5 r7 r8 | r1 r2 r3 r7 | r2 r3 r5 r6 r8 | r1 r2 r7 r8 | r1 r3 r7 | r4 | r9 | r1 r3 r6 | r2 r3 r5 r6 | n7 |
| n8 | r1 r2 r5 | r1 r2 r3 r6 | r7 | r1 r2 r5 r8 | r1 r3 r5 r6 r9 | r2 r3 r5 r6 r8 r9 | r1 r2 r8 | r1 r3 r8 r9 | r4 | n8 |
| n9 | r6 | r8 | r2 r3 | r9 | r1 r3 r4 <u>r5</u> | r2 r3 r5 | r1 r2 <u>r4 r5</u> r7 | r1 r3 r4 r7 | r2 r3 r5 | n9 |
| | c1 | c2 | c3 | c4 | c5 | c6 | c7 | c8 | c9 | |

*Figure 8.4. Two Swordfish in columns at the same time, in rc and cn representations*

We have already met in section 7.7.3 (Figure 7.3, reproduced as Figure 8.5) the puzzle we shall now use to illustrate a case of non-subsumption of a Swordfish in columns by whips. We already know from section 7.7.3 that this puzzle cannot be solved by braids of any length, let alone by whips. However, it has a resolution path using only Swordfish (besides rules in BSRT), which proves that at least one of the Swordfish eliminations cannot be replaced by a whip or a braid elimination.

|   |   | 1 |   | 2 |   |   |   | 3 |
|---|---|---|---|---|---|---|---|---|
|   |   |   | 1 |   |   | 4 |   |   |
| 2 |   |   | 4 |   | 5 |   |   |   |
|   |   | 6 |   |   | 7 |   |   | 8 |
|   | 5 |   |   |   |   | 2 |   |   |
| 9 |   |   | 3 |   | 4 |   |   |   |
|   |   | 8 |   |   | 1 |   |   | 5 |
|   | 9 |   |   | 6 |   |   |   |   |
| 1 |   |   | 9 |   |   | 7 |   |   |

| 6 | 4 | 1 | 5 | 9 | 2 | 8 | 7 | 3 |
|---|---|---|---|---|---|---|---|---|
| 8 | 7 | 5 | 6 | 1 | 3 | 2 | 4 | 9 |
| 2 | 3 | 9 | 4 | 7 | 8 | 5 | 6 | 1 |
| 3 | 1 | 6 | 2 | 4 | 7 | 9 | 5 | 8 |
| 7 | 5 | 4 | 1 | 8 | 9 | 3 | 2 | 6 |
| 9 | 8 | 2 | 3 | 5 | 6 | 4 | 1 | 7 |
| 4 | 2 | 8 | 7 | 3 | 1 | 6 | 9 | 5 |
| 5 | 9 | 7 | 8 | 6 | 4 | 1 | 3 | 2 |
| 1 | 6 | 3 | 9 | 2 | 5 | 7 | 8 | 4 |

***Figure 8.5**. A puzzle P with W(P)=B(P)=∞ but S(P)=3*

***** SudoRules 16.2 based on CSP-Rules 1.2, config: S *****
24 givens, 214 candidates, 1289 csp-links and 1289 links. Initial density = 1.41
**swordfish-in-columns: n4{c3 c6 c9}{r5 r8 r9} ==> r9c5 ≠ 4, r9c2 ≠ 4, r8c1 ≠ 4, r5c5 ≠ 4, r5c1 ≠ 4**
**swordfish-in-columns: n9{c3 c6 c9}{r2 r3 r5} ==> r5c7 ≠ 9, r5c5 ≠ 9**
;;; this swordfish allows three more eliminations, but they are interrupted by singles
singles to the end

As for the advantages of considering the four 2D spaces, notice that in the upper part of Figure 8.4 (rc-space at the start of resolution), it is difficult to distinguish the two Swordfish, because they are in the same columns and they have three rc-cells in common. In the lower part (cn-space), it is obvious: they lie in different rows (for n).

Exercise: use theorem 8.7 and its proof to show exactly which eliminations done (or allowed) by the two Swordfish are subsumed by whips and which are not.

As previously shown in section 7.7.3, this puzzle can be solved by g-whips[2], but this is irrelevant to our present purposes, because these g-whips are unrelated to the two Swordfish.

### 8.8.3. A Jellyfish not subsumed by whips but solved by g-whips or (longer) braids

After theorem 8.8, whips subsume most cases of Cyclic Quads. But there are rare examples in which this is not the case, such as the puzzle in Figure 8.6 (#017#Mauricio-002#8#1). Not only is there a Quad elimination that cannot be done by whips or braids of length 4, but also there is no whip of length < 18 that could do it. We shall also use this puzzle to illustrate the fact that allowing/disallowing one more resolution rule can occasionally have dramatic effects on the classification of a puzzle, although the statistical effects seem to be minor.

|   |   |   |   |   |   |   |   |   |
|---|---|---|---|---|---|---|---|---|
|   |   |   |   |   |   |   |   |   |
|   |   |   |   | 1 |   |   |   |   |
|   | 1 | 2 | 3 |   | 4 | 5 | 6 |   |
|   |   |   |   |   |   |   |   |   |
|   | 2 | 3 |   |   |   | 7 | 8 |   |
|   | 4 | 7 |   | 6 |   | 1 | 2 |   |
|   |   |   |   |   |   |   |   |   |
|   | 3 | 1 | 8 |   | 7 | 6 | 4 |   |
|   | 5 | 8 |   |   |   | 2 | 3 |   |

| 4 | 9 | 5 | 7 | 8 | 6 | 3 | 1 | 2 |
|---|---|---|---|---|---|---|---|---|
| 3 | 7 | 6 | 5 | 1 | 2 | 8 | 9 | 4 |
| 8 | 1 | 2 | 3 | 9 | 4 | 5 | 6 | 7 |
| 1 | 8 | 9 | 2 | 7 | 3 | 4 | 5 | 6 |
| 6 | 2 | 3 | 4 | 5 | 1 | 7 | 8 | 9 |
| 5 | 4 | 7 | 9 | 6 | 8 | 1 | 2 | 3 |
| 2 | 6 | 4 | 1 | 3 | 5 | 9 | 7 | 8 |
| 9 | 3 | 1 | 8 | 2 | 7 | 6 | 4 | 5 |
| 9 | 5 | 8 | 6 | 4 | 9 | 2 | 3 | 1 |

**Figure 8.6.** *Puzzle P with W+S(P)=4, B(P) = 10, W(P) >18 and gW(P) = 4*

### 8.8.3.1. Solution with whips and subsets, W+S(P)=4

Let us first find a solution combining whips and Subsets:

***** SudoRules version 13.7wter2, config: W+S *****
nrc-chain[2]: c8n5{r4 r7} – r8n5{c9 c5} ==> r4c5 ≠ 5 (a special case of whip[2])
xyz-chain[3]: r6c4{n9 n5} – r5c5{n5 n4} – r9c5{n4 n9} ==> r4c5 ≠ 9 (a special case of whip[3])
**naked-quads-in-a-block: b5{r5c4 r5c5 r5c6 r6c4}{ n1 n4 n5 n9} ==> r4c4 ≠ 4, r4c5 ≠ 4**

;;; here , due to the simplest first strategy, the application of Naked Quad is "interrupted" by the availability of a simpler rule (this could be modified):
whip[1]: b4n4{r5c4 .} ==> r5c9 ≠ 4
;;; now the Quad continues:
**naked-quads-in-a-block: b5{r5c4 r5c5 r5c6 r6c4}{n1 n4 n5 n9} ==> r4c4 ≠ 1, r4c4 ≠ 5, r4c4 ≠ 9, r4c6 ≠ 5, r4c6 ≠ 9, r6c6 ≠ 5, r6c6 ≠ 9**
;;; Resolution state RS$_1$

The resolution state RS$_1$ reached at this point is displayed in Figure 8.7; here, we have artificially isolated the last elimination allowed by this Quad, for later reference, because the same resolution state will be reached by another resolution path using only braids.
;;; let us now continue past resolution state RS$_1$:
**naked-quads-in-a-block: b5{r5c4 r5c5 r5c6 r6c4}{n1 n4 n5 n9} ==> r4c6 ≠ 1**
hidden-single-in-row r4 ==> r4c1 = 1

;;; we now reach a resolution state RS$_2$ (Figure 8.8) in which there is a Jellyfish; notice that this Jellyfish was already present in resolution state RS$_1$.

|  | c1 | c2 | c3 | c4 | c5 | c6 | c7 | c8 | c9 |  |
|---|---|---|---|---|---|---|---|---|---|---|
| r1 | n3 n4 n5 n6 n7 n8 n9 | n6 n7 n8 n9 | n4 n5 n6 n9 | n2 n5 n6 n7 | n2 n5 n7 n8 n9 | n2 n5 n6 n8 n9 | n3 n4 n8 n9 | n1 n7 n9 | n1 n2 n3 n4 n7 n8 n9 | r1 |
| r2 | n3 n4 n5 n6 n7 n8 n9 | n6 n7 n8 n9 | n4 n5 n6 n9 | n2 n5 n6 n7 n9 | n1 | n2 n5 n6 n8 n9 | n4 n8 n9 | n7 | n2 n3 n4 n7 n8 n9 | r2 |
| r3 | n7 n8 n9 | n1 | n2 | n3 | n7 n8 n9 | n4 | n5 | n6 | n7 n8 n9 | r3 |
| r4 | n1 n5 n6 n8 n9 | n6 n8 n9 | n5 n6 n9 | n̶1̶ n2 n4 n5 n7 n̶9̶ | n2 n3 n4 n7 n8 | **n1** n2 n3 n̶5̶ n8 n̶9̶ | n3 n4 n9 | n5 n9 | n3 n4 n5 n6 n̶9̶ | r4 |
| r5 | n1 n5 n6 n9 | n2 | n3 | n1 n4 n5 n9 | n4 n5 n9 | n1 n5 n9 | n7 | n8 | (n4)n5n6 n9 | r5 |
| r6 | n5 n8 n9 | n4 | n7 | n5 n9 | n6 | n3 n̶5̶ n8 n̶9̶ | n1 | n2 | n3 n5 n9 | r6 |
| r7 | n2 n4 n6 n7 | n6 n7 n9 | n4 n9 | n1 n2 n4 n5 n6 n9 | n2 n3 n4 n5 n9 | n1 n2 n3 n5 n6 n9 | n8 n9 | n1 n5 n7 | n1 n5 n7 n8 n9 | r7 |
| r8 | n2 n9 | n3 | n1 | n8 | n2 n5 n9 | n7 | n6 | n4 | n5 n9 | r8 |
| r9 | n4 n6 n7 n9 | n5 | n8 | n1 n4 n6 n9 | n4 n9 | n1 n6 n9 | n2 | n3 | n1 n7 n9 | r9 |
|  | c1 | c2 | c3 | c4 | c5 | c6 | c7 | c8 | c9 |  |

***Figure 8.7.*** *resolution state RS₁: a Naked Quad in block b5 (in grey cells); the nine candidates eliminated by the Quad just before resolution state RS₁ is reached are barred; the candidate (n4r5c9) eliminated by the whip[1] is between parentheses; the next candidate (n1r4c6) the Quad could eliminate is underlined; it is the target of no whip or braid.*

;;; let us now continue past RS₂:
**jellyfish-in-columns: n9{c2 c3 c7 c8}{r1 r2 r4 r7} ==> r1c6 ≠ 9, r1c9 ≠ 9, r2c1 ≠ 9, r2c4 ≠ 9, r2c6 ≠ 9, r2c9 ≠ 9, r4c9 ≠ 9, r7c1 ≠ 9, r7c4 ≠ 9, r7c5 ≠ 9, r7c6 ≠ 9, r7c9 ≠ 9**
nrc-chain[3]: c6n9{r5 r9} – r9c5{n9 n4} – b5n4{r5c5 r5c4} ==> r5c4 ≠ 9 (a special kind of whip[3])
**jellyfish-in-columns: n9{c2 c3 c7 c8}{r1 r2 r4 r7} ==> r1c4 ≠ 9, r1c5 ≠ 9**
singles to the end

### 8.8.3.2. Using only braids, B(P)=10

Suppose we now want a pure braids solution and we do not allow Subset rules. Then we get B(P) = 10.

***** SudoRules 16.2 based on CSP-Rules 1.2, config: B *****
26 givens, 222 candidates, 1621 csp-links and 1621 links. Initial density = 1.65
whip[2]: c8n5{r4 r7} – r8n5{c9 .} ==> r4c5 ≠ 5
whip[3]: r6c4{n9 n5} – r5c5{n5 n4} – r9c5{n4 .} ==> r4c5 ≠ 9

|    | c1 | c2 | c3 | c4 | c5 | c6 | c7 | c8 | c9 |    |
|----|----|----|----|----|----|----|----|----|----|----|
| r1 | n3 n4 n5 n6 n7 n8 n9 | n6 n7 n8 n9 | n4 n5 n6 n9 | n2 n7 n9 | n2 n5 n6 n9 | n2 n5 n6 n8 n9 | n3 n4 n8 n9 | n1 n7 n9 | n1 n2 n3 n4 n7 n8 n9 | r1 |
| r2 | n3 n4 n5 n6 n7 n8 n9 | n6 n7 n8 n9 | n4 n5 n6 n9 | n2 n5 n6 n7 n9 | n1 | n2 n5 n6 n8 n9 | n3 n4 n8 n9 | n7 n9 | n2 n3 n4 n7 n8 n9 | r2 |
| r3 | n7 n8 n9 | n1 | n2 | n3 | n7 n8 n9 | n4 | n5 | n6 | n7 n8 n9 | r3 |
| r4 | n1 | n6 n8 n9 | n5 n6 n9 | n2 n7 | n2 n3 n7 n8 | n2 n3 n8 | n3 n4 n9 | n5 n9 | n3 n4 n5 n6 n9 | r4 |
| r5 | n5 n6 n9 | n2 | n3 | n1 n4 n5 n9 | n4 n5 n9 | n1 n5 n9 | n7 | n8 | n5 n6 n9 | r5 |
| r6 | n5 n8 n9 | n4 | n7 | n5 n9 | n6 | n3 n8 | n1 | n2 | n3 n5 n9 | r6 |
| r7 | n2 n4 n6 n7 | n6 n7 n9 | n4 n6 n9 | n1 n2 n4 n5 n6 n9 | n2 n3 n4 n5 n9 | n1 n2 n3 n5 n6 n9 | n8 n9 | n1 n5 n7 n9 | n1 n5 n7 n8 n9 | r7 |
| r8 | n2 n9 | n3 | n1 | n8 | n2 n5 n9 | n7 | n6 | n4 | n5 n9 | r8 |
| r9 | n4 n6 n7 n9 | n5 | n8 | n1 n4 n6 n9 | n4 n9 | n1 n6 n9 | n2 | n3 | n1 n7 n9 | r9 |
|    | c1 | c2 | c3 | c4 | c5 | c6 | c7 | c8 | c9 |    |

***Figure 8.8.*** *Resolution state RS$_2$: a Jellyfish not subsumed by whips or g-braids*

;;; the following whips[4] replace all but one of the eliminations allowed by the Naked Quad in the previous resolution path:

whip[4]: b5n7{r4c4 r4c5} – b5n2{r4c5 r4c6} – b5n3{r4c6 r6c6} – b5n8{r6c6 .} ==> r4c4 ≠ 1, r4c4 ≠ 4, r4c4 ≠ 5, r4c4 ≠ 9
whip[4]: b5n7{r4c5 r4c4} – b5n2{r4c4 r4c6} – b5n3{r4c6 r6c6} – b5n8{r6c6 .} ==> r4c5 ≠ 4
whip[1]: b4n4{r5c4 .} ==> r5c9 ≠ 4
whip[4]: r6c4{n5 n9} – r5c6{n9 n1} – r5c4{n1 n4} – r5c5{n4 .} ==> r4c6 ≠ 5
whip[4]: r6c4{n9 n5} – r5c6{n5 n1} – r5c4{n1 n4} – r5c5{n4 .} ==> r4c6 ≠ 9
whip[4]: r6c4{n5 n9} – r5c6{n9 n1} – r5c4{n1 n4} – r5c5{n4 .} ==> r6c6 ≠ 5
whip[4]: r6c4{n9 n5} – r5c6{n5 n1} – r5c4{n1 n4} – r5c5{n4 .} ==> r6c6 ≠ 9

Here, we have reached the same resolution state as RS$_1$. But now, candidate n1r4c6 (underlined in Figure 8.7), which could be eliminated by the Naked Quad in the previous resolution path, is the target of no whip or braid; it is *a rare case of a Quad elimination not subsumed by whips, braids, g-whips or g-braids*. As a consequence of this missing elimination, r4c1 = 1 cannot be asserted. Nevertheless, this does not prevent the Jellyfish from being present (it was already present in state RS$_1$). But, what is really exceptional here is that *none of the candidates that could be eliminated by the Jellyfish can be eliminated by a whip[4]*.

The resolution path with braids continues, much harder than with Subsets:

whip[5]: b4n1{r4c1 r5c1} – r5n6{c1 c9} – b6n5{r5c9 r6c9} – r6c4{n5 n9} – r5n9{c4 .} ==> r4c1 ≠ 5
whip[5]: b4n1{r4c1 r5c1} – r5n6{c1 c9} – b6n9{r5c9 r6c9} – r6c4{n9 n5} – r5n5{c4 .} ==> r4c1 ≠ 9
whip[5]: r4n1{c1 c6} – b5n8{r4c6 r6c6} – b5n3{r6c6 r4c5} – b5n2{r4c5 r4c4} – b5n7{r4c4 .} ==> r4c1 ≠ 8
whip[6]: r8c9{n9 n5} – c8n5{r7 r4} – c8n9{r4 r7} – c7n9{r7 r4} – c3n9{r4 r1} – c2n9{r2 .} ==> r2c9 ≠ 9
whip[6]: r8c9{n9 n5} – c8n5{r7 r4} – c8n9{r4 r7} – c7n9{r7 r4} – c3n9{r4 r2} – c2n9{r1 .} ==> r1c9 ≠ 9
whip[6]: r9c5{n9 n4} – r5c5{n4 n5} – r8n5{c5 c9} – b9n9{r8c9 r9c9} – b7n9{r9c1 r8c1} – r3n9{c1 .} ==> r7c5 ≠ 9
braid[6]: b5n5{r5c4 r6c4} – r8c9{n5 n9} – r6n9{c4 c1} – r3n9{c1 c5} – r5c5{n5 n4} – r9c5{n9 .} ==> r5c9 ≠ 5
whip[7]: r9c5{n4 n9} – r5c5{n9 n5} – r8n5{c5 c9} – r8n9{c9 c1} – r3n9{c1 c9} – r6n9{c9 c4} – r5n9{c4 .} ==> r7c5 ≠ 4
whip[7]: r9c5{n9 n4} – r5c5{n4 n5} – r8n5{c5 c9} – r8n9{c9 c1} – r3n9{c1 c9} – r6n9{c9 c4} – r5n9{c4 .} ==> r1c5 ≠ 9
braid[7]: r2c8{n7 n9} – r3c9{n9 n8} – c7n8{r2 r7} – c7n9{r7 r4} – r9n7{c9 c1} – r3c1{n7 n9} – b4n9{r6c1 .} ==> r1c9 ≠ 7
braid[7]: r2c8{n7 n9} – r3c9{n9 n8} – c7n8{r2 r7} – c7n9{r7 r4} – r9n7{c9 c1} – r3c1{n7 n9} – b4n9{r6c1 .} ==> r2c9 ≠ 7
braid[7]: r2c8{n7 n9} – r9n7{c1 c9} – r3c9{n7 n8} – c7n8{r1 r7} – c7n9{r1 r4} – r3c1{n7 n9} – b4n9{r6c1 .} ==> r2c1 ≠ 7
braid[7]: r6c4{n5 n9} – r8c9{n5 n9} – r5n9{c4 c1} – r3n9{c1 c5} – r8n5{c9 c5} – r5c5{n5 n4} – r9c5{n9 .} ==> r6c9 ≠ 5
whip[1]: b6n5{r4c8 .} ==> r4c3 ≠ 5
whip[1]: c3n5{r1 .} ==> r1c1 ≠ 5, r2c1 ≠ 5
whip[4]: c1n5{r5 r6} – r6c4{n5 n9} – r5n9{c4 c9} – r5n6{c9 .} ==> r5c1 ≠ 1
hidden-single-in-a-block ==> r4c1 = 1
**braid[10]: b4n8{r4c2 r6c1} – r6n5{c1 c4} – c5n3{r4 r7} – r6n9{c4 c9} – r8c9{n9 n5} – c5n5{r8 r1} – c5n2{r1 r8} – c5n7{r1 r3} – r3c1{n7 n9} – r8n9{c9 .} ==> r4c5 ≠ 8**
whip[1]: c5n8{r1 .}2 ==> r1c6 ≠ 8, r2c6 ≠ 8
whip[7]: b2n8{r1c5 r3c5} – c1n8{r3 r6} – r6n5{c1 c4} – r6n9{c4 c9} – r3n9{c9 c1} – r8n9{c1 c5} – r9n9{c6 .} ==> r1c2 ≠ 8
whip[8]: b2n8{r1c5 r3c5} – c5n7{r3 r4} – c5n3{r4 r7} – c5n2{r7 r8} – r8c1{n2 n9} – r3n9{c1 c9} – r6n9{c9 c4} – r5n9{c4 .} ==> r1c5 ≠ 5
braid[5]: r6n3{c9 c6} – c5n3{r4 r7} – r6c4{n9 n5} – r8c9{n9 n5} – c5n5{r8 .} ==> r6c9 ≠ 9
singles ==> r6c9 = 3, r6c6 = 8, r4c2 = 8
whip[2]: r6n9{c4 c1} – b7n9{r9c1 .} ==> r7c4 ≠ 9
whip[3]: r6n9{c4 c1} – r8n9{c1 c9} – r3n9{c9 .} ==> r5c5 ≠ 9
whip[3]: r9c5{n9 n4} – r5c5{n4 n5} – r6c4{n5 .} ==> r9c4 ≠ 9
whip[3]: r6n9{c1 c4} – r5n9{c4 c9} – b9n9{r9c9 .} ==> r7c1 ≠ 9
whip[4]: b4n9{r6c1 r4c3} – b6n9{r4c9 r5c9} – r8n9{c9 c5} – r3n9{c5 .} ==> r9c1 ≠ 9
whip[3]: b7n9{r7c2 r8c1} – b4n9{r5c1 r4c3} – b6n9{r4c9 .} ==> r7c9 ≠ 9
whip[4]: r9n9{c6 c9} – r8n9{c9 c1} – r5n9{c1 c4} – r6n9{c4 .} ==> r7c6 ≠ 9
whip[4]: b8n9{r9c6 r8c5} – r3n9{c5 c1} – r6n9{c1 c4} – r5n9{c4 .} ==> r9c9 ≠ 9

whip[1]: r9n9{c5 .} ==> r8c5 ≠ 9
whip[2]: r8n9{c9 c1} – b4n9{r5c1 .} ==> r4c9 ≠ 9
whip[3]: r6n9{c4 c1} – r3n9{c1 c9} – r8n9{c9 .} ==> r1c4 ≠ 9, r2c4 ≠ 9
whip[1]: c4n9{r6 .} ==> r5c6 ≠ 9
whip[3]: r6n9{c1 c4} – r5n9{c4 c9} – r8n9{c9 .} ==> r1c1 ≠ 9, r2c1 ≠ 9, r3c1 ≠ 9
whip[3]: r8n9{c9 c1} – r5n9{c1 c4} – r6n9{c4 .} ==> r3c9 ≠ 9
singles to the end

### 8.8.3.3. Using only whips, W(P)>18

Suppose now we wanted a solution with only whips. If a resolution path could be obtained with whips, some of them would have to be of length > 18, i.e. one has W(P) > 18. Actually, we did not try longer ones because of memory overflow problems and we did not insist because it did not seem interesting to go further.

### 8.8.3.4. Using g-whips, gW(P)=4

If we now use g-whips, we get gW(P) = 4, with a completely different resolution path (unrelated to the Quads in the first path):

***** SudoRules 16.2 based on CSP-Rules 1.2, config: gW *****
26 givens, 222 candidates and 1621 nrc-links
whip[2]: c8n5{r4 r7} – r8n5{c9 .} ==> r4c5 ≠ 5
whip[3]: r6c4{n9 n5} – r5c5{n5 n4} – r9c5{n4 .} ==> r4c5 ≠ 9

;;; after this point, the resolution path diverges completely with respect to the previous ones :
g-whip[3]: b6n9{r4c7 r456c9} – r3n9{c9 c5} – r8n9{c5 .} ==> r4c1 ≠ 9
g-whip[3]: b4n9{r4c3 r456c1} – r3n9{c1 c5} – r8n9{c5 .} ==> r4c9 ≠ 9
g-whip[3]: b7n9{r7c3 r789c1} – r3n9{c1 c9} – b9n9{r9c9 .} ==> r7c5 ≠ 9
g-whip[3]: b4n9{r6c1 r4c123} – b6n9{r4c7 r456c9} – b9n9{r9c9 .} ==> r7c1 ≠ 9
g-whip[3]: b7n9{r7c3 r789c1} – r5n9{c1 c456} – r6n9{c6 .} ==> r7c9 ≠ 9
whip[4]: b5n7{r4c4 r4c5} – b5n2{r4c5 r4c6} – b5n3{r4c6 r6c6} – b5n8{r6c6 .} ==> r4c4 ≠ 5, r4c4 ≠ 4, r4c4 ≠ 1, r4c4 ≠ 9
whip[4]: b5n7{r4c5 r4c4} – b5n2{r4c4 r4c6} – b5n3{r4c6 r6c6} – b5n8{r6c6 .} ==> r4c5 ≠ 4
whip[1] : r4n4{c9 .} ==> r5c9 ≠ 4
whip[4]: r6c4{n5 n9} – r5c6{n9 n1} – r5c4{n1 n4} – r5c5{n4 .} ==> r4c6 ≠ 5, r6c6 ≠ 5
whip[4]: r6c4{n9 n5} – r5c6{n5 n1} – r5c4{n1 n4} – r5c5{n4 .} ==> r4c6 ≠ 9, r6c6 ≠ 9
g-whip[3]: b9n9{r7c7 r789c9} – r6n9{c9 c1} – b7n9{r9c1 .} ==> r7c4 ≠ 9
**g-whip[4]: b4n9{r5c1 r4c123} – b6n9{r4c7 r456c9} – r8n9{c9 c5} – r9n9{c6 .} ==> r3c1 ≠ 9**
whip[3]: r3n9{c5 c9} – r8n9{c9 c1} – r6n9{c1 .} ==> r5c5 ≠ 9
whip[3]: r9c5{n9 n4} – r5c5{n4 n5} – r6c4{n5 .} ==> r9c4 ≠ 9
whip[4]: r9n7{c9 c1} – r3c1{n7 n8} – r3c9{n8 n9} – r2c8{n9 .} ==> r2c9 ≠ 7
whip[4]: r9n7{c9 c1} – r3c1{n7 n8} – r3c9{n8 n9} – r2c8{n9 .} ==> r1c9 ≠ 7
whip[4]: r2c8{n7 n9} – r3n9{c9 c5} – r3n7{c5 c9} – r9n7{c9 .} ==> r2c1 ≠ 7
whip[4]: r8c9{n5 n9} – r3n9{c9 c5} – r9c5{n9 n4} – r5c5{n4 .} ==> r8c5 ≠ 5
singles ==> r8c9 = 5, r4c8 = 5
whip[1]: c3n5{r1 .} ==> r1c1 ≠ 5, r2c1 ≠ 5

whip[3]: r8n9{c1 c5} – r9n9{c6 c9} – r3n9{c9 .} ==> r7c3 ≠ 9, r7c2 ≠ 9
whip[1]: b7n9{r9c1 .} ==> r1c1 ≠ 9, r2c1 ≠ 9, r5c1 ≠ 9, r6c1 ≠ 9
whip[1]: b4n9{r4c2 .} ==> r4c7 ≠ 9
whip[1]: b6n9{r6c9 .} ==> r9c9 ≠ 9
whip[1]: b9n9{r7c7 .} ==> r7c6 ≠ 9
whip[1]: b6n9{r6c9 .} ==> r1c9 ≠ 9, r2c9 ≠ 9, r3c9 ≠ 9
singles to the end

### 8.9. Subsets in N-Queens

Recalling that, in N-Queens, a label corresponds to a cell, we shall represent each transversal set in an $S_p$-subset pattern by p grey cells with the same shade of grey.

#### 8.9.1. A Pair in 7-Queens with a transversal set not associated with a CSP variable

The instance of 7-Queens in Figure 8.9, with two queens already placed in r2c1 and r6c4 has a Pair for CSP variables Xr4 and Xr7, with transversal sets {r4c5, r7c2} and {r4c7, r7c7}. These sets are defined as the intersections of the two rows with respectively a diagonal and a column. The first thus provides an example of a transversal set not defined via a "transversal" CSP variable.

|    | c1 | c2 | c3 | c4 | c5 | c6 | c7 |
|----|----|----|----|----|----|----|----|
| r1 | ∘  | ∘  |    | ∘  |    |    | B  |
| r2 | ∗  | ∘  | ∘  | ∘  | ∘  | ∘  | ∘  |
| r3 | ∘  | ∘  |    | ∘  |    | A  | ∘  |
| r4 | ∘  | ∘  | ∘  | ∘  | ░  | ∘  | ▓  |
| r5 | ∘  |    | ∘  | ∘  | ∘  |    | C  |
| r6 | ∘  | ∘  | ∘  | ∗  | ∘  | ∘  | ∘  |
| r7 | ∘  | ░  | ∘  | ∘  | ∘  | ∘  | ▓  |

*Figure 8.9. A 7-Queens instance, with a Pair*

***** Manual solution *****
whip[1]: r4{c5 .} ⇒ ¬r3c6 (A eliminated)
pair: {{Xr4, Xr7}, {{r4c5, r7c2}, {r4c7, r7c7}}} ⇒ ¬r1c7, ¬r5c7 (B and C eliminated)

Notice that A could have been eliminated by the Pair, because it is also linked to the first transversal set, but the whip[1] is applied before, because it is considered simpler. Both B and C are linked to the second transversal set.

Remember that the disjointness conditions of the definition bear on the candidates of the different CSP variables in the current resolution state and not on the transversal sets, let alone on the global transversal constraints (or transversal CSP variables) defining them, if any: here r2c7 is common to both constraints.

Finally, notice that, in conformance with the general theory, the Pair can be seen as a whip[2]:

whip[2]: ⇒ r4{c7 c5} – r7{c2 .} ⇒ ¬r1c7, ¬r5c7

### 8.9.2. A Pair in 10-Queens with transversal sets defined via transversal variables

|    | c1 | c2 | c3 | c4 | c5 | c6 | c7 | c8 | c9 | c10 |
|----|----|----|----|----|----|----|----|----|----|----|
| r1 | ∘  | ∘  | ∘  | ∘  | ∘  | ∘  | *  | ∘  | ∘  | ∘  |
| r2 | ∘  | ∘  | ∘  | ∘  | ∘  | ∘  | ∘  | ∘  | ∘  | *  |
| r3 |    | ∘  | ∘  | ∘  | ∘  | +  | ∘  | ∘  | ∘  | ∘  |
| r4 | ∘  | ∘  | *  | ∘  | ∘  | ∘  | ∘  | ∘  | ∘  | ∘  |
| r5 | +  | ∘  | ∘  | ∘  | ∘  |    | ∘  | ∘  | ∘  | ∘  |
| r6 | ∘  | ∘  | ∘  | ∘  | ∘  | ∘  | ∘  | *  | ∘  | ∘  |
| r7 | ∘  |    | ∘  | +  | ∘  | ∘  | ∘  | ∘  | ∘  | ∘  |
| r8 | B  | +  | ∘  | ∘  |    | ∘  | ∘  | ∘  | ∘  | ∘  |
| r9 | ∘  | ∘  | ∘  | ∘  | ∘  | ∘  | ∘  | ∘  | *  | ∘  |
| r10| C  | ∘  | ∘  | ∘  | +  | A  | ∘  | ∘  | ∘  | ∘  |

*Figure 8.10. A 10-Queens instance, with a Pair*

Consider again the 10-Queens instance in Figure 5.10 (section 5.11.2), reproduced below as Figure 8.10. Suppose we do not see the second and the third long distance interaction whips. We can still eliminate B and C, based on Pairs in rows (CSP variables Xr3, Xr5), in which the transversal sets correspond to the intersections with columns ("transversal CSP variables" Xc1, Xc6).

***** Manual solution *****
whip[1]: r3{c1 .} ⇒ ¬r10c6 (A eliminated)
pairs: {{Xr3, Xr5}, {c1{r3, r5}, c6{r3, r5}}} ⇒ ¬r8c1, ¬r10c1   (B, C eliminated)
single in r10: r10c5; single in r8: r8c2; single in r7: r7c4; single in r5: r5c1; single in r3: r3c6
Solution found in $W_2$.

### 8.9.3. Triplets in 9-Queens not subsumed by whip[3]

The instance of 9-Queens in Figure 8.11 has a complete Triplet (three candidates for the three CSP variables, i.e. all the optional candidates are present). The (unique) elimination (A) allowed by the Triplet cannot be replaced by a whip[3].

Here, the method is used to provide a simple proof that this instance has no solution.

***** Manual solution *****
triplets: {{Xr1, Xr3, Xr7}, {c1{r1, r3, r7}, c5{r1, r3, r7}, c7{r1, r3, r7}}} ⇒ ¬r6c1 (A eliminated)
whip[3]: r6{c2 c8} – r7{c7 c5} – r3{c5 .} ⇒ ¬r8c2 (B eliminated)
single in r8 ⇒ r8c8
whip[1]: c1{r7 .} ⇒ ¬r7c5 (C eliminated)
single in r7 ⇒ r7c1
This puzzle has no solution: no value for Xc2

|    | c1 | c2 | c3 | c4 | c5 | c6 | c7 | c8 | c9 |
|----|----|----|----|----|----|----|----|----|----|
| r1 |    | ○  | ○  | ○  |    | ○  |    | ○  | ○  |
| r2 | ○  | ○  | ✱  | ○  | ○  | ○  | ○  | ○  | ○  |
| r3 |    | ○  | ○  | ○  |    | ○  |    | ○  | ○  |
| r4 | ○  | ○  | ○  | ○  | ○  | ○  | ○  | ○  | ✱  |
| r5 | ○  | ○  | ○  | ✱  | ○  | ○  | ○  | ○  | ○  |
| r6 | A  |    | ○  | ○  | ○  | ○  | ○  |    | ○  |
| r7 | +  | ○  | ○  | ○  | C  | ○  |    | ○  | ○  |
| r8 | ○  | B  | ○  | ○  | ○  | ○  | ○  | +  | ○  |
| r9 | ○  | ○  | ○  | ○  | ○  | ✱  | ○  | ○  | ○  |

*Figure 8.11. A 9-Queens instance, with a complete Triplet*

# 09. Reversible-$S_p$-chains, $S_p$-whips and $S_p$-braids

In this chapter, we define more complex types of chains than the whips, g-whips and corresponding braids introduced until now[8]. At least for the Sudoku CSP, this entails that we are dealing with exceptional instances, either because they cannot be solved by the previous patterns or because the new ones give them a smaller rating.

The main idea is that there are patterns that can be considered as elementary or "atomic" and there are ways to combine them into more complex ones. Until now, typical "atomic" patterns have been single candidates in chapter 5 and g-candidates in chapter 7. And the typical way of combining them has been to assemble them into chains, whips, g-whips, braids and g-braids via what we shall now call the "zt-ing principle": in the context of these chains, i.e. "modulo the target (z) and the previous right-linking candidates (t)", they appear as single candidates or as g-candidates.

We shall now show that this principle can be extended to the $S_p$-subset patterns of chapter 8, more precisely: given any Subset resolution theory $S_p$ ($0 \leq p \leq \infty$) for any CSP, one can define $S_p$-whips and $S_p$-braids as generalised whips or braids that accept patterns from this family of rules (i.e. $S_{p'}$-subsets for any $p' \leq p$), in addition to candidates and g-candidates, for their right-linking elements – whereas their left-linking elements remain mere candidates, as in the case of whips and g-whips. In a sense, allowing the inclusion of such patterns introduces a restricted kind of look-ahead with respect to the original non-anticipating (no look-ahead) whips and g-whips, because each $S_{p'}$-subset is inserted into the chain as a whole and it increases its length by p' (its size) instead of 1; but this form of look-ahead is strictly controlled by the p parameter and by the very specific type of pattern the $S_{p'}$-subsets are.

If we consider that, in the context of a whip or a g-whip, the left-linking candidates have negative valence and the right-linking candidates or g-candidates have positive valence, then in the context of the new $S_p$-whips and $S_p$-braids, the right-linking $S_p$-subsets have positive valence, in the sense that, if the target was True in some resolution state RS, there would be some posterior resolution state in which they would appear as autonomous $S_p$-subsets.

---

[8] In the Sudoku context, we first introduced these extended whips and braids (with a different terminology) in the "Fully Supersymmetric Chains" thread of the late Sudoku Player's Forum (p. 14, October 17th, 2008).

In the next chapters, we shall see that one can go still further, but we think the intermediate step developed here is sufficiently interesting in its own. Moreover, it will be easier to justify certain choices we shall have to make later, after we have analysed the simpler case of $S_p$-whips (simpler mainly because, contrary to whips or braids, the $S_p$-subset patterns can be defined without any reference to their target).

Everything goes for $S_p$-whips as for g-whips (except that a few additional technicalities have to be faced). The main point to be noticed is that, when it comes to defining the concepts of $S_p$-links and $S_p$-compatibility, we always consider the $S_p$-labels underlying the $S_p$-subsets instead of the $S_p$-subsets themselves, in exactly the same way as we considered the full g-labels underlying the g-candidates when we defined g-links. The main reason for this choice is the same as that for g-links: we want all the notions related to linking and compatibility to be purely structural, i.e. we do not want them to depend on any particular resolution state; this will be essential for the confluence property of $S_p$-braid resolution theories (in section 9.4) and for the "T&E($S_p$) vs $S_p$-braids" theorem (in section 9.5). But there are also important computational benefits in doing so (such as the possibility of pre-computing all the $S_p$-labels and $S_p$-links – but we shall not dwell on implementation matters here).

## 9.1. $S_p$-links; $S_p$-subsets modulo other Subsets; $S_p$-regular sequences

### 9.1.1. $S_p$-links, $S_p$-compatibility

Definition: a label l is *compatible with an $S_p$-label* S if l is not $S_p$-linked to S (i.e. if, for each transversal set TS of S, there is at least one label l' in TS such that l is not linked to l').

Definition: a label l is *compatible* with a set R of labels, g-labels and S-labels if l is compatible with each element of R (in the senses of "compatible" already defined separately for labels, g-labels and $S_p$-labels).

Definitions: a label l is *$S_p$-linked to an $S_p$-subset* S if l is $S_p$-linked to the $S_p$-label underlying S; a label l is compatible with an $S_p$-subset if l is not $S_p$-linked to it; a label l is *compatible* with a set R of candidates, g-candidates and Subsets if l is compatible with each element of R (in the senses of "compatible" already defined separately for candidates, g-candidates and $S_p$-subsets).

Notice that, in conformance with what we mentioned in the introduction to this chapter, according to the definition of "$S_p$-linked to an $S_p$-subset", it is not enough for label l to be linked to all the actual candidates of one of its transversal sets: it must be linked to all the labels of one of its transversal sets.

### 9.1.2. $S_p$-subsets modulo a set of labels, g-labels and S-labels

All our forthcoming definitions (Reversible-$S_p$-chains, $S_p$-whips and $S_p$-braids) will be based on that of an $S_p$-subset modulo a set R of labels, g-labels and S-labels; in practice, R will be either the previous right-linking pattern or the set consisting of the target plus all the previous right-linking patterns (i.e. candidates, g-candidates and $S_k$-subsets).

Definition: in any resolution state of any CSP, given a set R of labels, g-labels and S-labels [or a set R of candidates, g-candidates and Subsets], a *Pair (or $S_2$-subset) modulo R* is an $S_2$-label {CSPVars, TransvSets}, where:
 – CSPVars = {$V_1$, $V_2$},
 – TransvSets is composed of the following transversal sets of labels:
    {<$V_1$, $v_{11}$>, <$V_2$, $v_{21}$>} for constraint $c_1$,
    {<$V_1$, $v_{12}$>, <$V_2$, $v_{22}$>} for constraint $c_2$,
such that:
 – in RS, $V_1$ and $V_2$ are disjoint, i.e. they share no candidate;
 – <$V_1$, $v_{11}$> ≠ <$V_1$, $v_{12}$> and <$V_2$, $v_{22}$> ≠ <$V_2$, $v_{21}$>;
 – in RS, $V_1$ has the two mandatory candidates <$V_1$, $v_{11}$> and <$V_1$, $v_{12}$> compatible with R and no other candidate compatible with R;
 – in RS, $V_2$ has the two mandatory candidates <$V_2$, $v_{21}$> and <$V_2$, $v_{22}$> compatible with R and no other candidate compatible with R.

Definition: in any resolution state of any CSP, given a set R of labels, g-labels and S-labels [or a set R of candidates, g-candidates and Subsets], a *Triplet (or $S_3$-subset) modulo R* is an $S_3$-label {CSPVars, TransvSets}, where:
 – CSPVars = {$V_1$, $V_2$, $V_3$},
 – TransvSets is composed of the following transversal sets of labels:
 – {<$V_1$, $v_{11}$>, (<$V_2$, $v_{21}$>), <$V_3$, $v_{31}$>} for constraint $c_1$,
 – {<$V_1$, $v_{12}$>, <$V_2$, $v_{22}$>, (<$V_3$, $v_{32}$>)} for constraint $c_2$,
 – {(<$V_1$, $v_{13}$>), <$V_2$, $v_{23}$>, <$V_3$, $v_{33}$>} for constraint $c_3$,
such that:
 – in RS, $V_1$, $V_2$ and $V_3$ are pairwise disjoint, i.e. no two of these variables share a candidate;
 – <$V_1$, $v_{11}$> ≠ <$V_1$, $v_{12}$>, <$V_2$, $v_{22}$> ≠ <$V_2$, $v_{23}$> and <$V_3$, $v_{33}$> ≠ <$V_3$, $v_{31}$>;
 – in RS, $V_1$ has the two mandatory candidates <$V_1$, $v_{11}$> and <$V_1$, $v_{12}$> compatible with R, one optional candidate <$V_1$, $v_{13}$> compatible with R (supposing this label exists) and no other candidate compatible with R;

– in RS, $V_2$ has the two mandatory candidates $<V_2, v_{22}>$ and $<V_2, v_{23}>$ compatible with R, one optional candidate $<V_2, v_{21}>$ compatible with R (supposing this label exists) and no other candidate compatible with R;

– in RS, $V_3$ has the two mandatory candidates $<V_3, v_{33}>$ and $<V_3, v_{31}>$ compatible with R, one optional candidate $<V_3, v_{32}>$ compatible with R (supposing this label exists) and no other candidate compatible with R.

We leave it to the reader to write the definitions of Subsets of larger sizes modulo R ($S_p$-subsets modulo R). The general idea is that, when one looks in RS at some $S_p$-label "modulo R", i.e. when all the candidates in RS incompatible with R are "forgotten", what remains in RS satisfies the conditions of a non degenerated Subset of size p based on this $S_p$-label.

Definition: in all the above cases, *a target of the $S_p$-subset modulo R* is defined as a target of the $S_p$-subset itself (i.e. as a candidate $S_p$-linked to its underlying $S_p$-label). The idea is that, in any context (e.g. in a chain) in which all the elements in R have positive valence, the $S_p$-subset itself will have positive valence and any of its targets will have negative valence.

### 9.1.3. $S_p$-regular sequences

As in the case of chains built on mere candidates, it is convenient to introduce an auxiliary notion before we define Reversible-$S_p$-chains, $S_p$-whips and $S_p$-braids.

Definition: let there be given an integer $1 \leq p \leq \infty$, an integer $m \geq 1$, a sequence $(q_1, ..., q_m)$ of integers, with $1 \leq q_k \leq p$ for all $1 \leq k \leq m$, and let $n = \sum_{1 \leq k \leq m} q_k$; let there also be given a sequence $(W_1, ..., W_m)$ of different sets of CSP variables of respective cardinalities $q_k$ and a sequence $(V_1, ..., V_m)$ of CSP variables such that $V_k \in W_k$ for all $1 \leq k \leq m$. We define *an $S_p$-regular sequence of length n associated with $(W_1, ... W_m)$ and $(V_1, ... V_m)$* to be a sequence of length 2m [or 2m-1] $(L_1, R_1, L_2, R_2, .... L_m, [R_m])$, such that:

– $q_m=1$ and $W_m = \{V_m\}$;

– for $1 \leq k \leq m$, $L_k$ is a candidate;

– for $1 \leq k \leq m$ [or $1 \leq k < m$], $R_k$ is a candidate or a g-candidate if $q_k=1$ and it is a (non degenerated) $Sq_k$-subset if $q_k>1$;

– for each $1 \leq k \leq m$ [or $1 \leq k < m$], one has *"strong continuity", "strong g-continuity" or "strong $Sq_k$-continuity" from $L_k$ to $R_k$*, namely:

  - if $R_k$ is a candidate ($q_k=1$ and $W_k=\{V_k\}$), $L_k$ and $R_k$ have a representative with $V_k$: $<V_k, l_k>$ and $<V_k, r_k>$,

  - if $R_k$ is a g-candidate ($q_k=1$ and $W_k=\{V_k\}$), $L_k$ is a candidate $<V_k, l_k>$ for $V_k$ and $R_k$ is a g-candidate $<V_k, r_k>$ for $V_k$ ($r_k$ being its set of values),

- if $R_k$ is an $Sq_k$-subset ($q_k>1$), then $W_k$ is its set of CSP variables and $L_k$ has a representative $<V_k, l_k>$ with $V_k$.

The $L_k$'s are called the *left-linking candidates* of the sequence and the $R_k$'s the *right-linking objects (or elements or patterns or Subsets)*.

Remarks:

– Notice the natural expression chosen for $L_k$ to $R_k$ continuity in case $R_k$ is a Subset.

– The definition of Subsets implies a disjointness condition on the sets of candidates for the CSP variables inside each $W_k$, but the present definition puts no *a priori* condition on the intersections of different $W_k$'s. In particular, $W_{k+i}$ may be a strict subset of $W_k$, if the right-linking elements in between give negative valence in $W_{k+i}$ to some candidates that had no individual valence assigned in $W_k$. This is not considered as an inner loop of the sequence.

Exercise: after reading all this chapter, comment on the condition $q_m=1$ and show that it entails no restriction in the sequel.

### 9.2. Reversible-$S_p$-chains

Reversible-$S_p$-chains are an extension of g-bivalue chains in which right-linking candidates may be replaced by g-candidates or $S_{p'}$-subsets (p'≤p). [One could imagine introducing an intermediate, restricted notion, in which g-candidates would not be not allowed; with the proper definition, extending that of bivalue chains, they would be reversible and give rise to resolution theories with the confluence property; but, for the same reasons as invoked in the definition of the Subset resolution theories, this would not make much sense in practice.]

#### 9.2.1. Definition of Reversible-$S_p$-chains

Definition: given an integer 1≤p≤∞ and a candidate Z (which will be a target), a *Reversible-$S_p$-chain* of length n (n ≥ 1) built on Z, noted $RS_pC[n]$, is an $S_p$-regular sequence ($L_1, R_1, L_2, R_2, …. L_m, R_m$) of length n associated with a sequence ($W_1, … W_m$) of sets of CSP variables and a sequence ($V_1, … V_m$) of CSP variables (with $V_k \in W_k$ for all 1≤k≤m and $W_m = \{V_m\}$), such that:

– Z is neither equal to any candidate in {$L_1, R_1, L_2, R_2, …. L_m, R_m$}, nor a member of any g-candidate in this set, nor equal to any label in the $Sq_k$-label of $R_k$ when $R_k$ is an $Sq_k$-subset, for any 1≤k<m;

– Z is linked to $L_1$;

– for each 1 < k ≤ m, $L_k$ is linked or g-linked or $Sq_{k-1}$-linked to $R_{k-1}$; this is the natural way of defining *"continuity" from $R_{k-1}$ to $L_k$*;

– $R_1$ is a candidate or a g-candidate or an $Sq_1$-subset modulo Z: $R_1$ is the only candidate or g-candidate or is the unique $Sq_1$-subset composed of all the candidates C for the CSP variables in $W_1$ such that C is compatible with Z;

– for any $1 < k \leq m$, $R_k$ is a candidate or a g-candidate or an $Sq_k$-subset modulo $R_{k-1}$: $R_k$ is the only candidate or g-candidate or (if $k \neq m$) is the unique $Sq_k$-subset composed of all the candidates C for the CSP variables in $W_k$ such that C is compatible with $R_{k-1}$;

– Z is not a label for $V_m$;

– Z is linked or g-linked to $R_m$.

***Theorem 9.1 (Reversible-$S_p$-chain rule for a general CSP): in any resolution state of any CSP, if Z is the target of a Reversible-$S_p$-chain, then it can be eliminated (formally, this rule concludes ¬candidate(Z)).***

Proof: if Z was True, then $L_1$ would be eliminated by ECP and $R_1$ would be asserted by S (if it is a candidate) or it would be a g-candidate or an $Sq_1$-subset; in any case, $L_2$ would be eliminated by ECP or $W_1$ or $Sq_1$. After iteration: $R_m$ would be asserted by S or it would be a g-candidate – which would contradict Z being True.

### 9.2.2. Reversibility of Reversible-$S_p$-chains in the general CSP

The following theorem justifies the name we have given these chains. Notice that it does in no way depend on the fact that the transversal sets defining the Subsets would be defined by "transversal" CSP variables.

***Theorem 9.2: a Reversible-$S_p$-chain is reversible.***

Proof: the main point of the proof is the construction of the reversed chain (a generalisation of the construction for g-bivalue-chains in section 7.2).

This construction can be followed in part using Figure 9.1. This Figure gives a symbolic representation of the end of a Reversible-$S_2$-chain and the start of the associated reversed chain. Horizontal solid lines represent CSP variables (both chains use the same global set of CSP variables); vertical dotted lines represent transversal sets: on horizontal lines, candidates can only exist at the intersections with dotted lines (here "horizontal" and "vertical" are in no way related to an underlying grid on which the CSP would have to be defined). Octagons are symbolic containers for the candidates in the right-linking $S_2$-subsets (solid lines for the initial chain, dotted lines for the reversed chain); they also show how CSP variables are grouped (differently) in each chain to define their respective Subsets.

Given a Reversible-$S_p$-chain ($L_1$, $R_1$, $L_2$, $R_2$, …. $L_m$, $R_m$) of length n built on Z and associated with the sequence ($W_1$, … $W_m$) of sets of CSP variables and the sequence ($V_1$, … $V_m$) of CSP variables, let us define a reversed $S_p$-chain of same length, with the same target Z and associated with a sequence $(W'_1, ..., W'_m)$ of sets of CSP variables and a sequence $(V'_1, ..., V'_m)$ of CSP variables that are closely related, but not identical, to the reversed sequences of $(W_1, ..., W_m)$ and $(V_1, ..., V_m)$ respectively, and with a sequence of sizes $(q'_1, ..., q'_m)$ such that its first m-1 elements are those of $(q_1, ..., q_{m-1})$ in reversed order and $q'_m=1$. Let $L'_1 = R_m$.

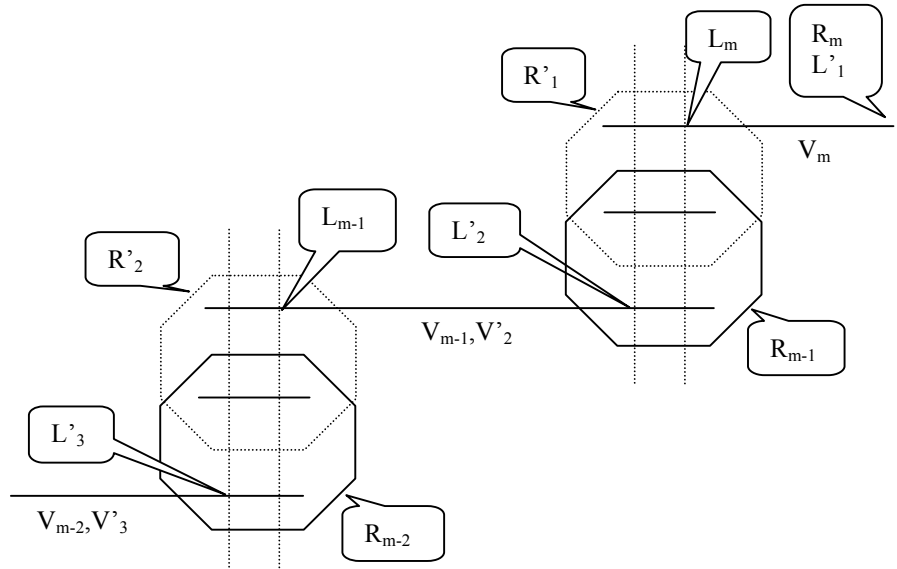

**Figure 9.1.** *A symbolic representation of the end of a Reversible-$S_2$-chain and the start of the associated reversed chain.*

We can now define $W'_1$, $V'_1$, $R'_1$ and $L'_2$, depending on what $R_{m-1}$ is:

– if $q_{m-1}=1$ and $R_{m-1}$ is a candidate or a g-candidate and it is linked or g-linked to only one candidate for $V_m$ (which implies that this candidate can only be $L_m$), then let $W'_1 = \{V_m\}$, $V'_1 = V_m$, $q'_1=1$ and $R'_1 = L_m$ ($R'_1$ is a candidate); let $L'_2 = R_{m-1}$ if $R_{m-1}$ is a candidate and $L'_2 =$ any candidate in $R_{m-1}$ if $R_{m-1}$ is a g-candidate;

– if $q_{m-1}=1$ and $R_{m-1}$ is a candidate or a g-candidate and it is linked or g-linked to several candidates for $V_m$ (which implies that these candidates can only be elements of a g-label for $V_m$, say g), let $W'_1 = \{V_m\}$, $V'_1 = V_m$, $q'_1=1$ and let $R'_1$ be the subset of g consisting of these candidates ($R'_1$ is a thus g-candidate); as before, let $L'_2 = R_{m-1}$ if $R_{m-1}$ is a candidate and $L'_2 =$ any candidate in $R_{m-1}$ if $R_{m-1}$ is a g-candidate;

– if $q_{m-1}>1$, then $R_{m-1}$ is an $Sq_{m-1}$-subset; let $W'_1 = W_{m-1} \cup \{V_m\} - \{V_{m-1}\}$; let $V'_1 = V_m$; and let $R'_1$ be the set of all the candidates for variables in $W'_1$. Because $R_m$ is the only candidate for $V_m$ modulo $R_{m-1}$, all the candidates for $V_m$ other than $L'_1 = R_m$ can only be in the transversal sets of $R_{m-1}$. Thus, forgetting $L'_1$, $R'_1$ together with the same transversal sets as $R_{m-1}$ is an $Sq_{m-1}$-subset and it has all the candidates for $V_{m-1}$ in $R_{m-1}$ as targets (and we take any of these as $L'_2$). As a result, all the other candidates for $V_{m-1}$ (i.e. all those that are compatible with $R'_1$) can only be in the transversal sets of $R_{m-2}$.

We are now in a situation in which $L'_2$ is defined and the above construction can be iterated, using $L'_2$ instead of $L'_1$, $R_{m-2}$ instead of $R_{m-1}$, $W_{m-1}$ instead of $W_m$ and $V_{m-1}$ instead of $V_m$ (once $L'_1$ was defined, the fact that $q_m=1$, i.e. that $R_m$ was a candidate or a g-candidate played no role in the above construction).

All this can be iterated until we can define the final $W'_m = \{V'_m\}$ with $V'_m = V_1$; $L_1$ or the g-candidate consisting of $L_1$ and the other candidates for $V_1$ linked to $Z$ can be taken as $R'_m$. qed.

Notice that, in this construction: even though $q_m=1$, one can have $q'_1 \neq 1$; and even if $q_1 \neq 1$, one always has $q'_m=1$, as in the definition of a Reversible-$S_p$-chain.

Exercise: check that this reversed chain does satisfy all the conditions in the definition of a Reversible-$S_p$-chain.

### 9.2.3. $RS_pC_n$ resolution theories and the $RS_pC$ ratings

As is now usual after introducing new rules, we can define a new increasing family of resolution theories. Here, we can do it for each p.

Definition: for each p, $1 \leq p \leq \infty$, one can define an increasing sequence ($RS_pC_n$, $n \geq 0$) of resolution theories:
 – $RS_pC_0 = BRT(CSP)$,
 – $RS_pC_1 = RS_pC_0 \cup$ {rules for Reversible-$S_p$-chains of length 1} = $W_1$,
 – $RS_pC_2 = RS_pC_1 \cup S_2$ (if $p \geq 2$) $\cup$ {rules for Reversible-$S_p$-chains of length 2},
 – ....
 – $RS_pC_n = RS_pC_{n-1} \cup S_n$ (if $p \geq n$) $\cup$ {rules for Reversible-$S_p$-chains of length n},
 – $RS_pC_\infty = \cup_{n \geq 0} RS_pC_n$.

For p=1, $S_1W_n = gW_n$. For p=∞, i.e. for Reversible-$S_p$-chains built on Subsets of *a priori* unrestricted size, we also write $RSC_n$ instead of $RS_\infty C_n$.

Definition: for any $1 \leq p \leq \infty$, the **$RS_pC$-rating** of an instance P, noted $RS_pC(P)$, is the smallest $n \leq \infty$ such that P can be solved within $RS_pC_n$, i.e. with Resersible $S_p$-Chains of total length not greater than n.

***Theorem 9.3:*** *all the $RS_pC_n$ resolution theories (for $1 \leq p \leq \infty$ and $n \geq 0$) are stable for confluence; therefore, they have the confluence property.*

Proof: we leave it as an exercise for the reader. (Using reversibility to propagate the consequences of value assertions and candidate deletions, it can be obtained via a drastic simplification of the proof for the $S_p$-braids case, theorem 9.9.)

### 9.2.4. Reversible-Subset-chains in Sudoku: grouped ALS chains and AICs

Non Sudoku experts can skip this sub-section or see the classical definitions of ALS chains (chains of Almost Locked Set) and AICs (Alternating Inference Chains / Nice Loops) in the over-abundant Sudoku literature, e.g. at www.sudopedia.org. Our main purpose here is to notice that the above Reversible-Subset-chains, defined for any CSP, correspond in Sudoku to these well-known patterns (though the above presentation provides a very unusual perspective of them).

In Sudoku, if one considers only the $X_{rc}$ CSP variables, Reversible-Subset-chains correspond to the classical grouped ALS-chains ("grouped" because we allow g-candidates as right-linking patterns). The only difference is, we never mention "Almost Locked Sets" (ALSs) or "Restricted Commons", we deal only with Subsets ("Locked Sets") modulo something.

If one uses all the $X_{rc}$, $X_{rn}$, $X_{cn}$ and $X_{bn}$ CSP variables, Reversible-Subset-chains correspond to the grouped AICs (Alternating Inference Chains).

[Historical note: what an AIC is has never been very clear in the Sudoku literature. (In what it differs from "Nice Loops", apart from being written in a different notation has never been very clear either; it seems to be more a matter of competition between different people than anything else). On the one hand, the definition of AICs is so vague that, transposed into our vocabulary, almost anything could be used as a right-linking pattern.

On the other hand, i.e. on the concrete side of things, the fact that "Fish" (our Super-Hidden Subsets) could be included in AICs has been mentioned only long after we introduced the more general $S_p$-whips and $S_p$-braids (in a different terminology); as the definition of the latter was fully supersymmetric and included all types of Subsets from the start, there was no need to make a special mention of Fish Subsets; in particular, all our classification results with Subsets in *HLS*, or those with $S_p$-braids mentioned in section 9.6 below, included Fish.

From an epistemological point of view, it is interesting to explore the reasons of this late recognition. In our opinion, there are four:
– the various notions involved lacked being formalised;
– in particular, there was an incomplete view of all the logical symmetries;

– the notions of an Almost Locked Set and of a Restricted Common, at the basis of ALS chains, are much more complicated than the notion of a Locked Set modulo something; they are difficult to deal with; in particular, their correct transposition to AICs, i.e. their extension to the rn, cn and bn spaces, seems to be difficult to do without having a complete logical formalisation; they also lead to the introduction of several levels of "almosting": AALSs, AAALSs (all of which are taken care of by the more general zt-ing principle);

– there was a strong insistence on chains having to be "reversible" (without any definition of this property); even for chains effectively reversible according to our definition, this blocked any view of them, such as the one exposed here, that would have allowed to shortcut the notion of a Restricted Common.]

## 9.3. $S_p$-whips and $S_p$-braids

$S_p$-whips and $S_p$-braids are an extension of g-whips and g-braids in which $S_{p'}$-subsets ($p' \leq p$) may appear as right-linking patterns. They can also be seen as extensions of the Reversible-$S_p$-chains: starting from the same $S_p$-subset bricks, the "almosting-principle" used to assemble Reversible-$S_p$-chains (a principle that only allows to "forget" candidates linked to the previous right-linking pattern) has to be replaced by the much more powerful "zt-ing principle" (a principle that allows to "forget" candidates linked to any of the previous right-linking patterns or to the target). In this replacement, reversibility is lost, but the most important property, non-anticipativeness, is preserved (with the above-mentioned remarks on the restricted form of look-ahead that corresponds to inner Subsets).

### 9.3.1. Definition of $S_p$-whips

Definition: given an integer $1 \leq p \leq \infty$ and a candidate Z (which will be the target), an $S_p$-whip of length n (n ≥ 1) built on Z is an $S_p$-regular sequence ($L_1$, $R_1$, $L_2$, $R_2$, …. $L_m$) [notice that there is no $R_m$] of length n, associated with a sequence ($W_1$, … $W_m$) of sets of CSP variables and a sequence ($V_1$, … $V_m$) of CSP variables (with $V_k \in W_k$ for all $1 \leq k < m$ and $W_m = \{V_m\}$), such that:

– Z is neither equal to any candidate in {$L_1$, $R_1$, $L_2$, $R_2$, …. $L_m$} nor a member of any g-candidate in this set nor equal to any element in the $Sq_k$-label of $R_k$ when $R_k$ is an $Sq_k$-subset, for any $1 \leq k \leq m$;

– $L_1$ is linked to Z;

– for each $1 < k \leq m$, $L_k$ is linked or g-linked or $Sq_{k-1}$-linked to $R_{k-1}$; this is the natural way of defining *"continuity" from $R_{k-1}$ to $L_k$*;

– for any $1 \leq k < m$, $R_k$ is a candidate or a g-candidate or an $Sq_k$-subset modulo Z and all the previous right-linking patterns: either $R_k$ is the only candidate or g-candidate compatible with Z and with all the $R_i$ with $1 \leq i < k$, or $R_k$ is the unique $Sq_k$-subset composed of all the candidates C for some of the CSP variables in $W_k$ such that C is compatible with Z and with all the $R_i$ with $1\leq i< k$;

– Z is not a label for $V_m$;

– $V_m$ has no candidate compatible with the target and with all the previous right-linking objects (but $V_m$ has more than one candidate).

***Theorem 9.4 ($S_p$-whip rule for a general CSP): in any resolution state of any CSP, if Z is a target of an $S_p$-whip, then it can be eliminated (formally, this rule concludes ¬ candidate(Z)).***

Proof: the proof is an easy adaptation of that for g-whips.

If Z was True, all the z-candidates would be eliminated by ECP and, iterating upwards from k=2: $R_{k-1}$ would be asserted by S or it would be a g-candidate or an $Sq_{k-1}$-subset; $R_{k-1}$ to $L_k$ continuity ensures that $L_k$ would be eliminated by ECP, $W_1$ or $Sq_{k-1}$; and the t- candidates would be eliminated by these rules. When m-1 is reached, $R_{m-1}$ would be asserted by S or it would be a whip[1] (a g-candidate) or a Subset with target $L_m$; finally, there would be no value left for $V_m$ (because Z itself is not a label for $V_m$).

### 9.3.2. Definition of $S_p$-braids

Definition: given an integer $1\leq p\leq\infty$ and a candidate Z (which will be the target), an *$S_p$-braid* of length n (n ≥ 1) built on Z is an $S_p$-regular sequence ($L_1, R_1, L_2, R_2,$ …. $L_m$) [notice that there is no $R_m$] of length n, associated with a sequence ($W_1, … W_m$) of sets of CSP variables and a sequence ($V_1, … V_m$) of CSP variables (with $V_k \in W_k$ for all $1\leq k<m$ and $W_m = \{V_m\}$), such that:

– Z is neither equal to any candidate in $\{L_1, R_1, L_2, R_2, …. L_m\}$ nor a member of any g-candidate in this set nor equal to any element in the $Sq_k$-label of $R_k$ when $R_k$ is an $Sq_k$-subset, for any $1\leq k\leq m$;

– $L_1$ is linked to Z;

– for each $1 < k \leq m$, $L_k$ is linked or g-linked or S-linked to Z or to some of the $R_i$, i<k; this is the only difference with $S_p$-whips;

– for any $1 \leq k < m$, $R_k$ is a candidate or a g-candidate or an $Sq_k$-subset modulo Z and all the previous right-linking patterns: either $R_k$ is the only candidate or g-candidate compatible with Z and with all the $R_i$ with $1\leq i< k$, or $R_k$ is the unique $Sq_k$-subset composed of all the candidates C for some of the CSP variables in $W_k$ such that C is compatible with Z and with all the $R_i$ with $1\leq i< k$;

– Z is not a label for $V_m$;

– $V_m$ has no candidate compatible with the target and with all the previous right-linking objects (but $V_m$ has more than one candidate).

***Theorem 9.5 ($S_p$-braid rule for a general CSP): in any resolution state of any CSP, if Z is a target of an $S_p$-braid, then it can be eliminated (formally, this rule concludes ¬ candidate(Z)).***

Proof: almost the same as in the $S_p$-whips case. The Z or $R_i$ (i<k) to $L_k$ condition replacing the $R_{k-1}$ to $L_k$ continuity condition allows the same intermediate conclusion for $L_k$.

Definition: in any of the above defined Reversible-$S_p$-chains, $S_p$-whips [and their obvious reversible $S_p$-z-whips specialisation, with no t- candidates (see section 9.3.4)] or $S_p$-braids, a candidate other than $L_k$ for any of the CSP variables ("global" variable $V_k$ or inner variables $V_{k,i}$ if $R_k$ is an inner Subset), is called a t- [respectively a z-] candidate if it is incompatible with a previous right-linking pattern [resp. with the target]. Notice that a candidate can be z- and t- at the same time and that the t- and z- candidates are not considered as being part of the pattern.

### 9.3.3. $S_p$-whip and $S_p$-braid resolution theories; $S_pW$ and $S_pB$ ratings

In exactly the same way as in the cases of whips, g-whips, braids and g-braids, one can now, for each p, define an increasing sequence of resolution theories. They now have two parameters, one (n) for the total length of the chain and one (p) for the maximum size of its inner Subsets. By convention, p=1 means no Subset, only candidates and g-candidates.

Definition: for each 1≤p≤∞, one can define an increasing sequence ($S_pW_n$, n ≥ 0) of resolution theories (similar definitions can be given for $S_p$-braids, merely by replacing everywhere "whip" by "braid" and "W" by "B"):
– $S_pW_0$ = BRT(CSP),
– $S_pW_1$ = $S_pW_0$ ∪ {rules for $S_p$-whips of length 1} = $W_1$,
– $S_pW_2$ = $S_pW_1$ ∪ $S_2$ (if p≥2) ∪ {rules for $S_p$-whips of length 2},
– ....
– $S_pW_n$ = $S_pW_{n-1}$ ∪ $S_n$ (if p≥n) ∪ {rules for $S_p$-whips of length n},
– $S_pW_∞$ = ∪$_{n≥0}$ $S_pW_n$.

For p=1, $S_1W_n$ = $gW_n$. For p=∞, i.e. for S-whips built on Subsets of *a priori* unrestricted size (but, in practice, p < n), we also write $SW_n$ instead of $S_∞W_n$.

Definition: for any 1≤p≤∞, the ***$S_pW$-rating*** of an instance P, noted $S_pW(P)$, is the smallest n ≤ ∞ such that P can be solved within $S_pW_n$, i.e. by $S_p$-whips of maximal total length n. By convention, $S_pW(P)$ = ∞ means that P cannot be solved by $S_p$-whips of any length; SW(P) = ∞ means that P cannot be solved by S-whips of any length including Subsets of any size.

Definition: similarly, for any 1≤p≤∞, the ***$S_pB$-rating*** of an instance P, noted $S_pB(P)$, is the smallest n ≤ ∞ such that P can be solved within $S_pB_n$. By convention, $S_pB(P) = ∞$ means that P cannot be solved by $S_p$-braids of any length.

For any 1≤p≤∞, the $S_pW$ and $S_pB$ ratings are defined in a purely logical way, independent of any implementation; the $S_pW$ and $S_pB$ ratings of an instance are intrinsic properties of this instance; moreover, as will be shown in the next section, for any fixed p (1≤p≤∞), the $S_pB$ rating is based on an increasing sequence of theories ($S_pB_n$, n≥0) with the confluence property and it can therefore be computed with a simplest first strategy based on the global length of the $S_p$-braids involved.

For any puzzle P, one has obviously $W(P) ≥ gW(P) = S_1W(P) ≥ S_2W(P) ≥ S_pW(P) ≥ S_{p+1}W(P) ≥ … ≥ S_∞W(P)$ and similar inequalities for the $S_pB(P)$.

Beware of not confusing the definitions in this section with those in section 8.6.3. In the latter case, whips and Subsets of same size are merely put together in the same set of rules; in the present section, whips and Subsets are fused into more complex structures. The respective notations can be remembered with the following mnemonic (and a similar one for braids): the "+" sign (and the repetition of size n) in W+S (and in $W_n+S_n$) indicate(s) the juxtaposition of two different things; the absence of a space between W and S in WS and in $WS_n$ indicates their fusion into new patterns.

Notice that consistent definitions of length for $S_p$-whips or $S_p$-braids and of the associated $S_pW$ and $S_pB$ ratings are highly constrained:

– by the fact that they are generalisations of the $RS_pC$ chains;

– by the subsumption theorems of section 8.7 and their obvious generalisations to Subsets of any size: in "many" cases of inclusion of such Subsets in an $S_p$-whip or $S_p$-braid, it will be possible to replace them by equivalent g-whips or g-braids and to transform the original $S_p$-whip or $S_p$-braid into an equivalent $W_p$-whip or $B_p$-braid (chapter 11). It seems natural to impose that, in such cases, the two visions of the "same" pattern lead to the same length (especially as length is taken as the measure of complexity of instances).

Finally, the confluence property of all the $S_pB_n$ resolution theories for each p, $1 ≤ p ≤ ∞$, (proven in section 9.4 below), allows to superimpose on $S_pB_n$ a "simplest first" strategy compatible with the $S_pB$ rating.

### 9.3.4. *$S_p$-z-whips and how they subsume $S_{p+1}$-subsets*

#### 9.3.4.1. *Definition of $S_p$-z-whips*

In simple terms, an $S_p$-z-whip is a particular kind of $S_p$-whip (as such, it allows the elimination of its target): it has no t-candidate (more precisely, no t-candidate that cannot also be considered as a z-candidate). It is easy to define the $S_p$-z-whip[n] resolution theories and to prove that they have the confluence property.

Definition: given an integer $1 \leq p \leq \infty$ and a candidate Z (which will be the target), an *$S_p$-z-whip* of length n (n ≥ 1) built on Z is an $S_p$-regular sequence ($L_1$, $R_1$, $L_2$, $R_2$, …. $L_m$) [notice that there is no $R_m$] of length n, associated with a sequence ($W_1$, … $W_m$) of sets of CSP variables and a sequence ($V_1$, … $V_m$) of CSP variables (with $V_k \in W_k$ for all $1 \leq k < m$ and $W_m = \{V_m\}$), such that:

– Z is neither equal to any candidate in $\{L_1, R_1, L_2, R_2, …. L_m\}$ nor a member of any g-candidate in this set nor equal to any element in the $Sq_k$-label of $R_k$ when $R_k$ is an $Sq_k$-subset, for any $1 \leq k \leq m$;

– $L_1$ is linked to Z;

– for each $1 < k \leq m$, $L_k$ is linked or g-linked or $Sq_{k-1}$-linked to $R_{k-1}$; this is the natural way of defining *"continuity" from $R_{k-1}$ to $L_k$*;

– for any $1 \leq k < m$, $R_k$ is a candidate or a g-candidate or an $Sq_k$-subset modulo Z: either $R_k$ is the only candidate or g-candidate compatible with Z, or $R_k$ is the unique $Sq_k$-subset composed of all the candidates C for some of the CSP variables in $W_k$ such that C is compatible with Z;

– Z is not a label for $V_m$;

– $V_m$ has no candidate compatible with Z (but $V_m$ has more than one candidate).

### 9.3.4.2. *Targets of $S_p$-subsets are targets of $S_{p-1}$-z-whips[p]*

**Theorem 9.6: a target of an $S_p$-subset is always also a target of an $S_{p-1}$-z-whip of length p**.

Proof: almost obvious. After renumbering the CSP variables, one can always suppose that Z is $S_p$-linked to transversal set $TS_1$ and that $V_1$ has a candidate $L_1$ = <$V_1$, $l_1$> to which Z is linked. Let $L_p$ = <$V_p$, $l_p$> be a candidate for $V_p$ not in $TS_1$ (there must be one if the $S_p$-subset is not degenerated). Let $R_2$ be the $S_{p-1}$-subset: $\{\{V_1,…, V_{p-1}\}, \{TS_2, …, TS_p\}\}$. Then Z is a normal target of the following whip:

$S_{p-1}$-z-whip[p]: $\{L_1\ R_2\} - V_p\{l_p\ .\} \Rightarrow \neg$candidate(Z).

### 9.3.5. *Type-2 targets of $S_p$-subsets*

It appears that an $S_p$-subset that has transversal sets with non-void intersections allows more eliminations than the "standard" ones defined in chapter 8. (This can happen only for p>2.)

Definition: a type-2 target of an $S_p$-subset is a candidate belonging to (at least) two of its transversal sets.

**Theorem 9.7: a type-2 target of an $S_p$-subset can be eliminated.**

Proof: suppose the type-2 target Z is a candidate for variable $V_1$ and it belongs to transversal sets $TS_1$ and $TS_2$. If Z was True, then all the other candidates in $TS_1$ or $TS_2$ or in a g-candidate in $TS_1$ or $TS_2$ would be eliminated by ECP. This would leave at most p-2 possibilities for the remaining p-1 CSP variables – which is contradictory, in exactly the same way as in the case of a normal target.

Notice however that this is a very unusual kind of elimination. Until now, for all the rules we have met, the target did not belong to the pattern. The following theorem shows that this "cannibalistic" abnormality can be palliated. It also justifies that we did not consider type-2 targets of $S_p$-subsets in chapter 8: these abnormal targets can always be eliminated by a simpler pattern. An illustration of this theorem will appear in section 10.3 for the more general case of $gS_p$-subsets.

***Theorem 9.8: A type-2 target of an $S_p$-subset is always the (normal) target of a shorter $S_{p-2}$-z-whip of length p-1.***

Proof: in a resolution state RS, let Z be a type-2 target of an $S_p$-subset with CSP variables $V_1$, … $V_p$ and transversal sets $TS_1$, … $TS_p$. One can always suppose that $V_1$ is the CSP variable for which Z is a candidate (there can be only one in RS) and that $TS_1$ and $TS_2$ are the two transversal sets to which Z belongs.

Firstly, each of the CSP variables $V_2$, $V_3$, … $V_p$ must have at least one candidate belonging neither to $TS_1$ nor to $TS_2$ (if it has several, choose one arbitrarily and name it $<V_2, c_2>$, … $<V_p, c_p>$, respectively). Otherwise, the initial $S_p$-subset would be degenerated; more precisely, Z could be eliminated by a whip[1] (or even by ECP after a Single) associated with (any of) the CSP variable(s) that has no such candidate.

Secondly, in $TS_1$ or $TS_2$, there must be at least one candidate for at least one of the CSP variables $V_2$, … $V_p$. Otherwise, the initial $S_p$-subset would be degenerated; more precisely, it would contain, among others, the $S_{p-2}$-subset $\{\{V_3, …, V_p\}, \{TS_3, …, TS_p\}\}$; this would allow to eliminate all the candidates for $V_1$ and $V_2$ that are not in $TS_1$ or $TS_2$; Z could then be eliminated by a whip[1] associated with $V_2$; and $V_1$ would have no candidate left. One can always suppose that there exists such a candidate $L_2$ for $V_2$, i.e. $L_2 = <V_2, l_2>$.

Modulo Z, we therefore have an $S_{p-2}$ subset $R_2$ with CSP variables $V_2$, … $V_{p-1}$ and transversal sets $TS_3$, … $TS_p$. Then, Z is a (normal) target of the following $S_{p-2}$-z-whip of length p-1:

$S_{p-2}$-z-whip[p-1]: $V_2\{l_2\ R_2\}$ – $V_p\{c_p\ .\} \Rightarrow \neg$candidate(Z).

### 9.3.6. Accepting type-2 targets of $S_p$-subsets in $S_pW$ and $S_pB$?

Theorem 9.8 alone does not guarantee that type-2 targets of $S_p$-subsets, if allowed to be used as left-linking candidates in the definitions of $S_p$-whips or $S_p$-braids, could not lead to (slightly) more general patterns than those in our current definitions. The following, if true, would provide a negative answer and it would complete the justification for not accepting type-2 targets in $S_p$-subsets: "for any $S_p$-whip or $S_p$-braid, according to an extended definition that would allow using type-2 targets of $S_{p'}$-subsets as left-linking candidates, there is an equivalent standard (i.e. satisfying the definitions of this chapter) $S_p$-whip or $S_p$-braid, respectively, of same or shorter length and with the same target". But, although we have no counter-example, this does not seem to be true in general.

In order to understand why it may not be true, consider the following tentative proof, using the notations of the definitions. If the situation occurs several times in the chain, the same kind of actions as defined below could be repeated.

If left-linking candidate $L_{k+1}$ is a type-2 target of Subset $S_k$, then consider CSP variable $V_{k+1}$ (notice that it cannot be the unique CSP variable of $S_k$ for which $L_{k+1}$ is a candidate) Z and all the previous right-linking objects:

– either it has another candidate, say $L'_k$, linked to all the candidates in some of the transversal sets of $S_k$ to which $L_{k+1}$ does not belong; then, one can replace $L_{k+1}$ with $L'_{k+1}$ in the original chain;

– or it has no such candidate but it has a candidate linked to Z or to a previous right-linking object; then, in the $S_p$-braid cases but *a priori* not in the $S_p$-whip case, one can replace $L_{k+1}$ with this candidate in the original chain;

– or it has no such candidate and no candidate linked to Z or to a previous right-linking object; then no possibility seems to be available.

We think that the cases in which this construction does not work and there are no alternative resolution paths are extremely rare and that allowing type-2 targets of $S_{p'}$-subsets inside $S_p$-whips or $S_p$-braids is not worth until experimental results show the contrary. Until then, we shall stick to our original definitions.

See Theorem 10.17 for complementary aspects of this question in case g-Subsets instead of Subsets are involved.

Remark: the main interest of the above tentative proof may be that it can be transposed to other situations; e.g. it can explain why, given an $S_{p'}$-subset S allowing the elimination of a target L that could be replaced by a whip elimination (according to the subsumption theorems), if the same S appears inside an $S_p$-whip or $S_p$-braid modulo the target and the previous right-linking patterns of this $S_p$-whip or $S_p$-braid and if L is used in this $S_p$-whip or $S_p$-braid as the next left-linking candidate, S can nevertheless not always be considered as an ordinary sub-whip of this global $S_p$- whip or $S_p$-braid. In particular, given that $W_2$ subsumes $S_2$, this gives an idea why $S_2W_n$ is not equal to $gW_n$ for n>2 (or why $S_2W_5 \not\subset gW_5$, as shown by the example in section 9.7.1; or why $S_2B$ is not equal to $gB$ in table 9.1).

### 9.4. The confluence property of the $S_pB_n$ resolution theories

***Theorem 9.9: each of the $S_pB_n$ resolution theories (for $1 \leq p \leq \infty$, $0 \leq n \leq \infty$) is stable for confluence; therefore, it has the confluence property.***

Proof: in order to keep the same notations as in the proof for the g-braid resolution theories (section 7.6), we prove the result for $S_rB_n$, r and n fixed. The proof follows the same general lines as that for g-braids in section 7.5. We keep the same numbering of the various cases to be considered. However, some new sub-cases appear and some cases have to be split into three, in order to take into account the different kinds of right-linking patterns. Marks now extend from case f to case d.

We must show that, if an elimination of a candidate Z could have been done in a resolution state $RS_1$ by an $S_r$-braid B of length n' $\leq$ n and with target Z, it will always still be possible, starting from any further state $RS_2$ obtained from $RS_1$ by consistency preserving assertions and eliminations, if we use a sequence of rules from $S_rB_n$. Let B be: $\{L_1\ R_1\} - \{L_2\ R_2\} - .... - \{L_p\ R_p\} - \{L_{p+1}\ R_{p+1}\} - ... - \{L_m\ .\}$, with target Z, where the $R_k$'s are now candidates or g-candidates or Subsets from $S_r$ modulo Z and the previous $R_i$'s.

Consider first the state $RS_3$ obtained from $RS_2$ by applying repeatedly the rules in $S_r$ until quiescence. As $S_r$ has the confluence property (by theorem 8.4), this state is uniquely defined.

If, in $RS_3$, target Z has been eliminated, there remains nothing to prove. If target Z has been asserted, then the instance of the CSP is contradictory; if not yet detected in $RS_3$, this contradiction can be detected by CD in a state posterior to $RS_3$, reached by a series of applications of rules from $S_r$, following the $S_r$-braid structure of B.

Otherwise, we must consider all the elementary events related to B that can have happened between $RS_1$ and $RS_3$ as well as those we must provoke in posterior resolution states RS. For this, we start from B' = what remains of B in $RS_3$ and we let RS = $RS_3$. At this point, B' may not be an $S_r$-braid in RS. We progressively update RS and B' by repeating the following procedure, for p = 1 to p = m, until it produces a new (possibly shorter) $S_r$-braid B' with target Z in resolution state RS – a situation that is bound to happen. (As in the g-braids case, and because we have included $W_1$ in $S_r$, we have to consider a state RS posterior to $RS_3$). Return from this procedure as soon as B' is a g-braid in RS. All the references below are to the current RS and B'.

a) If, in RS, any candidate that had negative valence in B – i.e. the left-linking candidate, or any t- or z- candidate, of CSP variable $V_p$, or any t- or z- candidate of $R_p$ in case $R_p$ is an inner Subset – has been asserted (this can only be between $RS_1$ and $RS_3$), then all the candidates linked to it have been eliminated by relevant rules from $S_r$ in $RS_3$, in particular: Z and/or all the candidate(s) $R_k$ (k<p) to which it is linked, and/or all the elements of the g-candidate(s) $R_k$ (k<p) to which it is g-linked, and/or all the candidates of the CSP variable in $W_k$ to which it belongs and/or all the candidates in the transversal set(s) of the $R_k$'s (k<p) to which it is S-linked (by the definition of an $S_r$-braid); if Z is among them, there remains nothing to prove; otherwise, the procedure has already either been successfully terminated by case f1 or f2α or dealt with by case d2 of the first previous such k.

b) If, in RS, left-linking candidate $L_p$ has been eliminated (but not asserted), it can no longer be used as a left-linking candidate in an $S_r$-braid. Suppose that either CSP variable $V_p$ still has a z- or a t- candidate $C_p$, or $R_p$ is an inner Subset and there is another CSP variable $V_p$' in its $W_p$ such that $V_p$' still has a z- or a t- candidate $C_p$; then replace $L_p$ by $C_p$ and (in the latter case) $V_p$ by $V_p$'. Now, up to $C_p$, B' is a partial $S_r$-braid in RS with target Z. Notice that, even if $L_p$ was linked or g-linked or $S_r$-linked to $R_{p-1}$ (e.g. if B was an $S_r$-whip) this may not be the case for $C_p$; therefore trying to prove a similar theorem for $S_r$-whips would fail here.

c) If, in RS, any t- or z- candidate of $V_p$ or of the inner Subset $S_p$ has been eliminated (but not asserted), this has not changed the basic structure of B (at stage p). Continue with the same B'.

d) Consider now assertions occurring in right-linking objects. There are two cases instead of one for g-braids.

d1) If, in RS, right-linking candidate $R_p$ or a candidate $R_p$' in right-linking g-candidate $R_p$ has been asserted (p can therefore not be the last index of B'), $R_p$ can no longer be used as an element of an $S_r$-braid, because it is no longer a candidate or a g-candidate. As in the proof for g-braids, and only because of this d1 case, we cannot be sure that this assertion occurred in $RS_3$. We must palliate this. First eliminate by ECP or $W_1$ any left-linking or t- candidate for any CSP variable of B' after p, including those in the inner Subsets, that is incompatible with $R_p$, i.e. linked or g-linked to it, if it is still present in RS. Now, considering the $S_r$-braid structure of B upwards from p, more eliminations and assertions can been done by rules from $S_r$. (Notice that, as in the g-braids case, we are not trying to do more eliminations or assertions than needed to get a g-braid in RS; in particular, we continue to consider $R_p$, not $R_p$'; in any case, it will be excised from B'; but, most of all, we do not have to find the shortest possible $S_r$-braid!)

Let q be the smallest number strictly greater than p such that CSP variable $V_q$ or some CSP variable $V_q$' in $W_q$ still has a left-linking, t- or z- candidate $C_q$ that is not linked, g-linked or S-linked to any of the $R_i$ for $p \leq i < q$ ($C_q$ is therefore linked, g-linked or S-linked to Z or to some $R_i$ with $i < p$). (For index q, there is thus a $V_q$' in $W_q$ and a candidate $C_q$ for $V_q$' such that $C_q$ is linked, g-linked or S-linked to Z or to some $R_i$ with $i < p$.)

Apply the following rules from $S_r$ (if they have not yet been applied between $RS_2$ and RS) for each of the CSP variables $V_u$ (and all the $V_{u,i}$ in $W_u$ if $R_u$ is an inner Subset) with index (or first index) u increasing from p+1 to q-1 included:
- eliminate its left-linking candidate ($L_u$) by ECP or $W_1$ or some $S_{r'}$ ($r' \leq r$);
- at this stage, CSP variable $V_u$ has no left-linking candidate and there remains no t- or z- candidate in $W_u$ if $R_u$ is an inner Subset;
- if $R_u$ is a candidate, assert it by S and eliminate by ECP all the candidates for CSP variables after u, including those in the inner Subsets, that are incompatible with $R_u$ in the current RS;
- if $R_u$ is a g-candidate, it cannot be asserted; eliminate by $W_1$ all the candidates for CSP variables after u, including those in the inner Subsets, that are incompatible with $R_u$ in the current RS;
- if $R_u$ is an $Sq_u$-subset, it cannot be asserted by $Sq_u$; eliminate by $Sq_u$ all the candidates for CSP variables after u, including those in the inner Subsets, that are incompatible with $R_u$ in the current RS.

In the new RS thus obtained, excise from B' the part related to CSP variables and inner Subsets p to q-1 (included); if $L_q$ has been eliminated in the passage from $RS_2$ to RS, replace it by $C_q$ (and, if necessary, replace $V_q$ by $V_q$'); for each integer $s \geq p$, decrease by q-p the index of CSP variable $V_s$, of its candidates and inner right-linking pattern (g-candidate or $S_{r'}$-subset) and of the set $W_s$, in the new B'. In RS, B' is now, up to p (the ex q), a partial $S_r$-braid in $S_rB_n$ with target Z.

d2) If, in RS, a candidate $C_p$ in a right-linking $Sq_p$-subset $R_p$ has been asserted or eliminated or marked in a previous step, $R_p$ can no longer be used as such as a right-linking Subset of an $S_r$-braid, because it may no longer be a (conditional) $Sq_p$-subset. Moreover, there may be several such candidates in $R_p$; consider them all at once. Notice that candidates can only have been asserted as values in the transition from $RS_1$ to $RS_3$ (the candidates asserted in case d1 are all excised from B') and that all the candidates for their CSP variables or in their transversal sets have also been eliminated in this transition. Delete from $R_p$ the CSP variables and the transversal sets corresponding to these asserted candidates (as we do not have type-2 targets, there is the same number of each). Call $R_p$' what remains of $R_p$ and replace $R_p$ by $R_p$' in B'. A few more questions must be dealt with:
- is there still a candidate for one of the CSP variables of $R_p$' that could play the role of a left-linking candidate for $R_p$'? If not, $R_p$' has already become an autonomous Subset in $RS_3$; excise it from B', together with a whole part of B' after it, along the same lines as in case d1;

- is $R_p$' still linked to the next part of B'? If not, excise it from B', together with a whole part of B' after it, as in the previous case;
- is $R_p$' degenerated (modulo Z and the previous $R_k$'s)? If so, this can easily be fixed by replacing $R_p$' with the corresponding Reversible-Subset-chain (modulo Z and the previous $R_k$'s);
- does $R_p$' or the Reversible-Subset-chain (modulo Z and the previous $R_k$'s) replacing it have more targets than $R_p$? If so, if any of these is a right-linking candidate or an element of a right-linking g-candidate or of an $S_r$-subset of B' for an index after p, then mark it so that the information can be used in cases d2, f1, f2 or f3 of later steps.

In RS, B' is now, up to p (the ex q), a partial $S_r$-braid in $S_rB_n$ with target Z.

e) If, in RS, a left-linking candidate $L_p$ has been eliminated (but not asserted) and CSP variable $V_p$ has no t- or z- candidate in $RS_2$ (complementary to case b), we now have to consider three cases instead of the two we had for g-braids.

e1) If $R_p$ is a candidate, then $V_p$ has only one possible value, namely $R_p$; if $R_p$ has not yet been asserted by S somewhere between $RS_2$ and RS, do it now; this case is now reducible to case d1 (because the assertion of $R_p$ also entails the elimination of $L_p$); go back to case d1 for the same value of p (in order to prevent an infinite loop, mark this case as already dealt with for the current step).

e2) If $R_p$ is a g-candidate, then $R_p$ cannot be asserted by S; however, it can still be used, for any CSP variable after p, to eliminate by $W_1$ any of its t-candidates that is g-linked to $R_p$. Let q be the smallest number strictly greater than p such that, in RS, CSP variable $V_q$ still has a left-linking, t- or z- candidate $C_q$ that is not linked or g-linked or S-linked to any of the $R_i$ for $p \leq i < q$. Replace RS by the state obtained after all the assertions and eliminations similar to those in case d1 above have been done. Then, in RS, excise the part of B' related to CSP variables p to q-1 (included), replace $L_q$ by $C_q$ (if $L_q$ has been eliminated in the passage from $RS_2$ to RS) and re-number the posterior elements of B', as in case d1. In RS, B' is now, up to p (the ex q), a partial $S_r$-braid in $S_rB_n$ with target Z.

e3) If $R_p$ is an $Sq_p$-subset, then $R_p$ is no longer linked via $L_p$ to a previous right-linking element of the braid. If none of the CSP variables $V_p$' in $W_p$ has a z- or t-candidate $C_p$ that can be linked, g-linked or S-linked to Z or to a previous $R_i$, (situation complementary to case b), it means that the elimination of $L_p$ has turned $R_p$ into an unconditional $Sq_p$-subset. Let q be the smallest number strictly greater than p such that, in RS, CSP variable $V_q$ has a left-linking, t- or z- candidate $C_q$ that is not linked or g-linked or S-linked to any of the $R_i$ for $p \leq i < q$. Replace RS by the state obtained after all the assertions and eliminations similar to those in case d1 above have been done. Then, in RS, excise the part of B' related to CSP variables p to q-1 (included), replace $L_q$ by $C_q$ (if $L_q$ has been eliminated in the passage from RS$_2$ to RS) and re-number the posterior elements of B', as in case d1. In RS, B' is now, up to p (the ex q), a partial $S_r$-braid in $S_rB_n$ with target Z.

　f) Finally, consider eliminations occurring in a right-linking object $R_p$. This implies that p cannot be the last index of B'. There are three cases.

　f1) If, in RS, right-linking candidate $R_p$ of B has been eliminated (but not asserted) or marked, then replace B' by its initial part:
{L$_1$ R$_1$} − {L$_2$ R$_2$} − …. − {L$_p$ .}. At this stage, B' is in RS a (possibly shorter) $S_r$-braid with target Z. Return B' and stop.

　f2) If, in RS, a candidate in right-linking g-candidate $R_p$ has been eliminated (but not asserted) or marked, then:

　f2α) either there remains no unmarked candidate of $R_p$ in RS; then replace B' by its initial part: {L$_1$ R$_1$} − {L$_2$ R$_2$} − …. − {L$_p$ .}; at this stage, B' is in RS a (possibly shorter) $S_r$-braid with target Z; return B' and stop;

　f2β) or the remaining unmarked candidates of $R_p$ in RS still make a g-candidate and B' does not have to be changed;

　f2γ) or there remains only one unmarked candidate $C_p$ of $R_p$; replace $R_p$ by $C_p$ in B'. We must also prepare the next steps by putting marks. Any t-candidate of B that was g-linked to $R_p$, if it is still present in RS, can still be considered as a t-candidate in B', where it is now linked to $C_p$ instead of g-linked to $R_p$; this does not raise any problem. However, this substitution may entail that candidates that were not t-candidates in B become t-candidates in B'; if they are left-linking candidates of B', this is not a problem either; but if any of them is a right-linking candidate or an element of a right-linking g-candidate or of an $S_{r'}$-subset of B', then mark it so that the same procedure (i.e. f1, f2 or f3) can be applied to it in a later step.

　f3) If, in RS, a candidate $C_p$ in right-linking $Sq_p$-subset $R_p$ has been eliminated (but not asserted) or marked, this has been dealt with in case d2.

　Notice that, as was the case for braids and g-braids, this proof works only because the notions of being linked, g-linked or S-linked do not depend on the resolution state.

## 9.5. The "T&E($S_p$) vs $S_p$-braids" theorem, 1≤p≤∞

Any resolution theory T stable for confluence has the confluence property and the procedure T&E(T) can therefore be defined (see section 5.6.1). Taking T = $S_p$, it is obvious that any elimination done by an $S_p$-braid can be done by T&E($S_p$). As was the case for braids and for g-braids, the converse is true:

***Theorem 9.10: for any $1 \leq p \leq \infty$, any elimination done by $T\&E(S_p)$ can be done by an $S_p$-braid.***

The proof is very similar to the g-braids case.

Proof: Let RS be a resolution state and let Z be a candidate eliminated by $T\&E(S_p, Z, RS)$ using some auxiliary resolution state RS'. Following the steps of resolution theory $S_p$ in RS', we progressively build an $S_p$-braid in RS with target Z. First, remember that $S_p$ contains only five types of rules: ECP (which eliminates candidates), $W_1$ (whips of length 1, which eliminates candidates), $S_{p'}$ (which eliminates targets of $S_{p'}$-subsets, $p' \leq p$), S (which asserts a value for a CSP variable) and CD (which detects a contradiction on a CSP variable).

Consider the sequence $(P_1, P_2, \ldots, P_k, \ldots P_m)$ of rule applications in RS' based on rules from $S_p$ different from ECP and suppose that $P_m$ is the first occurrence of CD (there must be at least one occurrence of CD if Z is eliminated by $T\&E(S_p, Z, RS)$). We first define the $R_k$, $V_k$, $W_k$ and $q_k$ sequences, for $k < m$:
- if $P_k$ is of type S, then it asserts a value $R_k$ for some CSP variable $V_k$; let $W_k = \{V_k\}$ and $q_k=1$;
- if $P_k$ is of type whip[1]: $\{M_k \;.\} \Rightarrow \neg \text{candidate}(C_k)$ for some CSP variable $V_k$, then define $R_k$ as the g-candidate of $V_k$ that contains $M_k$ and is g-linked to $C_k$; (notice that $C_k$ will not necessarily be $L_{k+1}$); let $W_k = \{V_k\}$ and $q_k=1$;
- if $P_k$ is of type $S_{p'}$, then define $R_k$ as the non degenerated $S_{p'}$-subset used by the condition part of $P_k$, as it appears at the time when $P_k$ is applied; let $W_k$ be the set of CSP variables of $R_k$ and $q_k=p'$; in this case, $V_k$ will be defined later.

We shall build an $S_p$-braid[n] in RS with target Z, with the $R_k$'s as its sequence of right-linking candidates or g-candidates or $Sq_k$-subsets, with the $W_k$'s as its sequence of sets of CSP variables, with the $q_k$'s as its sequence of sizes and with $n = \sum_{1 \leq k \leq m} q_k$ (setting $q_m = 1$). We only have to define properly the $L_k$'s, $q_k$'s and $V_k$'s with $V_k \in W_k$. We do this by recursion, successively for $k = 1$ to $k = m$. As the proofs for $k = 1$ and for the passage from k to k+1 are almost identical, we skip the case $k = 1$. Suppose we have done it until k and consider the set $W_{k+1}$ of CSP variables.

Whatever rule $P_{k+1}$ is (S or whip[1] or $S_{p'}$), the fact that it can be applied means that, apart from $R_{k+1}$ (if it is a candidate) or the candidates contained in $R_{k+1}$ (if it is a g-candidate or an $S_{p'}$-subset), all the other candidates for all the CSP variables in $W_{k+1}$ that were still present in RS (and there must be at least one, say $L_{k+1}$, for some CSP variable $V_{k+1} \in W_{k+1}$) have been eliminated in RS' by the assertion of Z and the previous rule applications. But these previous eliminations can only result from being linked or g-linked or S-linked to Z or to some $R_i$, $i \leq k$. The data $L_{k+1}$, $R_{k+1}$ and $V_{k+1} \in W_{k+1}$ therefore define a legitimate extension for our partial $S_p$-braid.

End of the procedure: at step m, a contradiction is obtained by CD for a CSP variable $V_m$. It means that all the candidates for $V_m$ that were still candidates for $V_m$ in RS (and there must be at least one, say $L_m$) have been eliminated in RS' by the assertion of Z and the previous rule applications. But these previous eliminations can only result from being linked or g-linked or S-linked to Z or to some $R_i$, i<m. $L_m$ is thus the last left-linking candidate of the $S_p$-braid we were looking for in RS and we can take $W_m=\{V_m\}$. qed.

Remarks:

– here again, this proof works only because the existence of a link, g-link or $S_p$-link between a candidate and a pattern does not depend on the resolution state;

– as in the previous cases of braids and g-braids, it is very unlikely that following the T&E($S_p$) procedure to produce an $S_p$-braid, as in the construction in this proof, would provide the shortest available one in resolution state RS (and this intuition is confirmed by experience).

### 9.6. The scope of $S_p$-braids (in Sudoku)

The "T&E($S_p$) vs $S_p$-braids" theorem can be used to estimate with simple calculations the scope of $S_p$-braids (which is also an upper boundary for the scope of $S_p$-whips) for any p, without having to find effectively the resolution paths.

Several times in this book, we mentioned that all the Sudoku puzzles in a set of random collections of about 10,000,000 minimal puzzles generated according to different methods (several independent implementations of the bottom-up, top-down and controlled-bias algorithms) could be solved by T&E or (equivalently) by braids (indeed, we also mentioned that they can all be solved by whips).

As a result, Sudoku puzzles that are not in the scope of braids are extremely rare and if one wants to compare the scopes of several types of more complex $S_p$-braids, one can only do this on a collection of exceptionally hard puzzles. The natural choice for this is Glenn Fowler's (alias gsf) highly non random, manually selected collection of 8152 puzzles [gsf's www]; it contains puzzles of varying levels of difficulty, with a very strong bias for the hardest ones (and a tendency for repetition of puzzles with similar patterns of givens, i.e. obtained by variations from a previous pattern); although this is no longer true, its top level part has long been considered as containing the hardest known puzzles.

In Table 9.1[9], the first column defines the sets of puzzles under consideration: gsf's list is decomposed into slices of 500 puzzles each (but the last ones), starting from the top (i.e. from the hardest in his classification). The next columns show how many puzzles of each slice can be solved using the $S_p$-braids mentioned in the first row (each column includes the results of the previous columns).

---

[9] We first published these results on the late Sudoku Player's Forum, "Abominable T&E and Lovely Braids" thread, p. 3, October 2008.

| Resolution theory → ↓ slice of puzzles | $B_\infty$ | $gB_\infty$ | $S_2B_\infty$ | $S_3B_\infty$ | $S_4B_\infty$ | $S_{4Fin}B_\infty$ | $+x_2y_2$ |
|---|---|---|---|---|---|---|---|
| 1-500 | 0 | 187 | 336 | 414 | 443 | 466 | 489 |
| 500-1000 | 0 | 178 | 335 | 415 | 460 | 480 | 497 |
| 1001-1500 | 0 | 163 | 382 | 451 | 486 | 494 | 500 |
| 1501-2000 | 0 | 168 | 397 | 476 | 490 | 496 | 499 |
| 2001-2500 | 0 | 135 | 367 | 434 | 474 | 489 | 497 |
| 2501-3000 | 0 | 116 | 334 | 443 | 479 | 493 | 499 |
| 3001-3500 | 1 | 120 | 335 | 424 | 473 | 486 | 498 |
| 3501-4000 | 0 | 113 | 325 | 426 | 472 | 493 | 499 |
| 4001-4500 | 1 | 104 | 298 | 395 | 448 | 471 | 497 |
| 4501-5000 | 0 | 231 | 399 | 450 | 482 | 494 | 499 |
| 5001-5500 | 47 | 348 | 487 | 500 | | | |
| 5501-6000 | 434 | 490 | 500 | | | | |
| 6001-6500 | 487 | 500 | | | | | |
| 6501-7000 | 494 | 500 | | | | | |
| 7001-8152 | 1152 | | | | | | |
| Total solved | 2616 | 4505 | 6647 | 7480 | 7859 | 8014 | 8126 |
| Total unsolved | 5536 | 3647 | 1505 | 672 | 293 | 136 | 26 |

**Table 9.1.** *Cumulated number of puzzles solved by $S_p$-braids with $p' \leq p$, for each slice of 500 puzzles in gsf's list. Missing cells in a row are intended to make it easier to see when a slice is completely solved.*

This table shows that almost all the hardest puzzles, known at the time when gsf's list was published, can be solved with braids (of unspecified total length) built on (Naked, Hidden and Super-Hidden) Subsets, on Finned Fish (a variant of Fish with additional candidates linked to the target, i.e. a z-Fish) and on x2y2-belts[10].

---

[10] $x_2y_2$ belts are our formal interpretation of a pattern known as a "hidden-pairs loop" or "sk-loop". This extremely symmetric pattern originated in the famous EasterMonster puzzle created by "jpf" and was discovered by Steven Kurzhals [see chapter 13 for details]. EasterMonster has long been considered as the hardest known puzzle and it has given rise to many variants, in the hope of finding still harder ones; as a result, it is over-represented in gsf's list.

However, "Eleven" recently reported [Eleven www] that he generated more than fifteen million "potentially hardest" minimal puzzles (including 90% of the known puzzles with SER > 11) that, in the vocabulary of the present book, cannot be solved by T&E. He used a kind of genetic programming algorithm (an innovative idea in the search of hard Sudoku puzzles), starting with a random collection of minimal puzzles as the seed and mutating them by withdrawal and addition of clues. Later, using T&E($S_4$) together with other filters (whose precise description would be irrelevant here) for cropping the current population, and taking the widely used SER rating[11] of puzzles as the selection function, he published a sub-collection of 26,370 minimal puzzles [Eleven 2011] that cannot be solved by T&E($S_4$). The question of a resolution theory T such that T&E(T), or associated T-braids, could solve all the known Sudoku puzzles is thus more open (although these are extremely exceptional instances, in proportion: even fifteen millions in approximately $2.5 \times 10^{25}$ non essentially equivalent minimal puzzles, as estimated in section 6.3.2, would not be much). The new non-T&E($S_4$) sub-collection makes it very unlikely that such a "universal" T-braids resolution theory could be based mainly on S-braids or braids with variants of Subsets as inner patterns. This point will be re-examined in section 11.3 after we have introduced the more powerful notion of a B-braid.

### 9.7. Examples

As examples of Reversible-Subset-chains abound in the Sudoku web forums, we shall not give any here. They can be found under the names of Alternating Inference Chains (AICs) or Nice Loops, as explained in section 9.2.4. Still more examples of a very special case, ALS chains (chains of Almost Locked Sets), can also be found; they are AICs (in the broad sense we have given them here) restricted to rc-space.

#### 9.7.1. $S_2W_5 \not\subset gW_5$: an $S_2$-whip[5] not subsumed by a g-whip[5] (+ $S_2W_5 \not\subset gW_{13}$)

| 7 | | 8 | | | 3 | | | |
|---|---|---|---|---|---|---|---|---|
| | | | 2 | 1 | | | | |
| 5 | | | | | | | | |
| | 4 | | | | | 8 | 2 | 6 |
| 3 | | | | 8 | | | | |
| | | | 1 | | | | 9 | 3 |
| | 9 | | 6 | | | | | 4 |
| | | | | 7 | 5 | | | |
| | | | | | | | | |

| 7 | 2 | 8 | 9 | 4 | 6 | 3 | 1 | 5 |
|---|---|---|---|---|---|---|---|---|
| 9 | 3 | 4 | 2 | 5 | 1 | 6 | 7 | 8 |
| 5 | 1 | 6 | 7 | 3 | 8 | 2 | 4 | 9 |
| 1 | 4 | 7 | 5 | 9 | 3 | 8 | 2 | 6 |
| 3 | 6 | 9 | 4 | 8 | 2 | 1 | 5 | 7 |
| 8 | 5 | 2 | 1 | 6 | 7 | 4 | 9 | 3 |
| 2 | 9 | 3 | 6 | 1 | 5 | 7 | 8 | 4 |
| 4 | 8 | 1 | 3 | 7 | 9 | 5 | 6 | 2 |
| 6 | 7 | 5 | 8 | 2 | 4 | 9 | 3 | 1 |

***Figure 9.2.*** *A puzzle P with W(P) = 13*

---

[11] See the first note of chapter 6.

The puzzle P in Figure 9.2 provides an example of an $S_2$-whip[5] that is not equivalent to a whip or a g-whip of same length. This puzzle has moderate complexity (though it is on the high side of the fuzzy boundary of puzzles solvable by humans): $W(P) = gW(P) = 13$.

|     | c1 | c2 | c3 | c4 | c5 | c6 | c7 | c8 | c9 |     |
| --- | --- | --- | --- | --- | --- | --- | --- | --- | --- | --- |
| r1 | 7 | n1 n2 n6 | 8 | n4 n5 n9 | n4 n5 n6 n9 | n4 n5 n6 n9 | 3 | n1 n4 n5 n6 | n1 n2 n5 n9 | r1 |
| r2 | n4 n6 n9 | n3 n6 | n3 n4 n6 | 2 | n3 n4 n5 n6 n9 | 1 | n4 n6 n7 **n9** | n4 n5 n6 n7 n8 | n5 n7 n8 | r2 |
| r3 | 5 | n1 n2 n3 n6 | n1 n2 n3 n4 n6 n9 | n7 n8 | n3 n4 n6 n9 | n7 n8 | n1 n2 n4 n6 | n1 n4 n6 | n1 n2 n9 | r3 |
| r4 | n1 n9 | 4 | n1 n5 n7 n9 | n3 n5 n7 n9 | n5 n9 | n3 n5 n7 n9 | 8 | 2 | 6 | r4 |
| r5 | 3 | n2 n6 n7 | n2 n6 n7 n9 | n4 n7 n9 | 8 | n2 n4 n6 n9 | n1 n4 n7 | n1 n4 n5 n7 | n1 n5 n7 | r5 |
| r6 | n2 n6 n8 | n2 n5 n6 n7 n8 | n2 n5 n6 n7 | 1 | n2 n4 n5 n6 n7 | n2 n4 n5 n6 n7 | n4 n7 | 9 | 3 | r6 |
| r7 | n1 n2 n8 | 9 | n1 n2 n3 n5 n7 | 6 | n2 n3 n5 n8 | n1 n2 n5 | n1 n2 n7 | n1 n3 n7 n8 | 4 | r7 |
| r8 | n1 n2 n4 n6 n8 | n1 n2 n3 n6 n8 | n1 n2 n3 n4 n6 | n3 n4 n8 n9 | 7 | n2 n3 n4 n8 n9 | 5 | n1 n3 n6 n8 | n1 n2 n8 n9 | r8 |
| r9 | n1 n2 n4 n6 n8 | n1 n3 n5 n6 n7 n8 | n1 n2 n3 n4 n5 n6 n7 | n3 n4 n5 n8 n9 | n1 n2 n4 n5 n9 | n2 n3 n4 n5 n8 n9 | n1 n2 n6 n7 n9 | n1 n3 n6 n7 n8 | n1 n2 n7 n8 n9 | r9 |
|     | c1 | c2 | c3 | c4 | c5 | c6 | c7 | c8 | c9 |     |

**Figure 9.3.** Resolution state $RS_1$ of puzzle P in Figure 9.2

The first (easy) steps of the resolution paths with whips or g-whips are identical.

***** SudoRules 16.2 based on CSP-Rules 1.2, config: W *****
20 givens, 267 candidates, 2000 csp-links and 2000 links. Initial density = 1.41
whip[1]: r4n1{c1 .} ==> r5c3 ≠ 1, r5c2 ≠ 1
whip[1]: r2n8{c8 .} ==> r3c9 ≠ 8, r3c8 ≠ 8
whip[1]: r2n7{c7 .} ==> r3c9 ≠ 7, r3c8 ≠ 7, r3c7 ≠ 7
whip[1]: b6n5{r5c9 .} ==> r5c2 ≠ 5, r5c3 ≠ 5, r5c4 ≠ 5, r5c6 ≠ 5
whip[2]: b2n7{r3c4 r3c6} – b2n8{r3c6 .} ==> r3c4 ≠ 4, r3c4 ≠ 3
whip[2]: b2n8{r3c4 r3c6} – b2n7{r3c6 .} ==> r3c4 ≠ 9
whip[2]: b2n7{r3c6 r3c4} – b2n8{r3c4 .} ==> r3c6 ≠ 6, r3c6 ≠ 4, r3c6 ≠ 3
whip[1]: b2n3{r2c5 .} ==> r9c5 ≠ 3, r7c5 ≠ 3, r4c5 ≠ 3
whip[2]: b2n8{r3c6 r3c4} – b2n7{r3c4 .} ==> r3c6 ≠ 9
whip[3]: c1n2{r9 r6} – b4n8{r6c1 r6c2} – c2n5{r6 .} ==> r9c2 ≠ 2

whip[4]: b3n8{r2c9 r2c8} – r2n5{c8 c5} – r4c5{n5 n9} – c1n9{r4 .} ==> r2c9 ≠ 9
whip[5]: r4n7{c6 c3} – b4n1{r4c3 r4c1} – b4n9{r4c1 r5c3} – r5n6{c3 c2} – r5n2{c2 .} ==> r5c6 ≠ 7

The resolution state $RS_1$ reached at this point is displayed in Figure 9.3.

After $RS_1$, both resolution paths with whips or g-whips continue with a whip[6] and a whip[8]:

whip[6]: c2n7{r5 r9} – c9n7{r9 r2} – b3n8{r2c9 r2c8} – r2n5{c8 c5} – r4c5{n5 n9} – r5n9{c6 .} ==> r5c3 ≠ 7
whip[8]: c1n2{r7 r6} – b4n8{r6c1 r6c2} – c2n5{r6 r9} – c2n7{r9 r5} – b6n7{r5c7 r6c7} – r7c7{n7 n1} – b8n1{r7c5 r9c5} – c5n2{r9 .} ==> r7c3 ≠ 2

After these two whips, the two resolution paths diverge (one has either a whip[12] or a g-whip[8]), but they finally both give a rating of 13: W(P)=gW(P)=13. As they have nothing noticeable, we skip them.

What is interesting in the context of this chapter is that, in state $RS_1$, there appears a shorter pattern than those provided by whips or g-whips, an $S_2$-whip[5] (Although, in row r2, n7 appears in column c7, i.e. outside the two cells of the hidden pair, n7r2c7 is a z-candidate and can be "forgotten": we have a hidden pair modulo the target.):

**$S_2$-whip[5]: c1n9{r2 r4} – r4c5{n9 n5} – r2{c5n5 HP:[c8 c9][n7 n8]} –r2n7{c8 .} => r2c7 ≠ 9**

Indeed, the two full resolution paths with whips and g-whips show more: as the elimination r2c7≠9 appears only after a whip[13], it shows that the above $S_2$-whip[5] cannot be replaced by a g-whip, even longer, with length less than 13.

For a more complex example of an $S_2$-braid (one of total length 14), see Figure 13.6 and section 13.5.1.

# 10. g-Subsets, Reversible-$gS_p$-chains, $gS_p$-whips and $gS_p$-braids

This chapter extends the definitions and results of chapters 8 and 9 by allowing the basic elements of Subsets (the "intersections" between the CSP variables and the transversal sets) to be g-candidates instead of candidates. While $gS_p$-subsets are an extension of $S_p$-subsets in which g-transversal sets of candidates or g-candidates replace transversal sets of candidates, $gS_p$-whips (respectively $gS_p$-braids) are an extension of $S_p$-whips (resp. $S_p$-braids) in which $gS_p$-subsets replace $S_p$-subsets. The situation is similar to that in chapter 7, when we extended all the definitions and results from whips (resp. braids) to g-whips (resp. g-braids). For this reason and the following additional ones, we shall give precise definitions and theorems (at the risk of some apparent redundancy) but we shall be rather sketchy for their proofs:

  – the first two parts of this chapter strictly parallel chapters 8 and 9 respectively;

  – it seems that exploiting all the possibilities of these new g-Subsets is rather difficult in practice; in Sudoku, g-Subsets appear either as "Franken Fish" or as "Mutant Fish" (see section 10.1.6); as far as we know, these exotic Fish patterns have never before been considered as g-Subsets, i.e. as the "grouped" version of Subsets (for the reason that Subsets themselves have never been considered in the full generality allowed by the CSP point of view developed in chapter 8);

  – our personal opinion is that, most of the time, it is often easier to find and understand a solution with whips or g-whips, when it exists (see the subsumption results in section 10.1.5), than with such patterns; but we acknowledge that some Sudoku Fishermen may have a different opinion; the main advantage of a g-Subset is that, like a Subset, it often allows several eliminations at once (see section 10.3);

  – although this is not a problem for their general theory, finding explicitly all the non degenerated subcases of $gS_p$-subsets is very difficult for p>3;

  – as for the $gS_p$-whips and $gS_p$-braids obtained by allowing these new g-Subsets as right-linking objects, even if their theories can easily be developed in a strict parallel to those of $S_p$-whips and $S_p$-braids (as shown in section 10.2), they seem to be rather complex structures; in Sudoku, they include the "Fishy Cycles" (which are already "almost" subsumed by the simpler $S_p$-whips and $S_p$-braids).

For a better understanding of the concepts involved, it may be a good idea to read the detailed example in section 10.3 in parallel with the first two sections.

**10.1. g-Subsets**

*10.1.1. g-transversality, gS$_p$-labels and gS$_p$-links*

In the same way as, in chapters 7 and 8, we had to introduce a distinction between g-labels or S$_p$-labels (defined as maximal sets of labels) and g-candidates or S$_p$-subsets (that did not have to be maximal), we must now introduce a distinction between gS$_p$-labels that can only refer to CSP variables and to g-transversal sets of labels and g-labels (which can be considered as a kind of saturation or maximality condition on gS$_p$-labels), and gS$_p$-subsets in which considerations about mandatory and optional candidates or g-candidates will appear.

*10.1.1.1. Set of labels and g-labels g-transversal to a set of disjoint CSP variables*

Definition: for p>1, given a set of p different CSP variables $\{V_1, V_2, …, V_p\}$, we say that a set S of at most p different labels and g-labels is *g-transversal* with respect to $\{V_1, V_2, …, V_p\}$ for constraint c if:

– 1) none of the labels in S or contained in a g-label in S has a representative for two of these CSP variables;

– 2) all the labels in S or contained in a g-label in S are pairwise linked by some constraint;

– 3) all the labels in S are pairwise linked by constraint c;

– 4) each g-label in S contains a "distinguished" label that is linked by constraint c to all the labels in S and to all the other distinguished labels of all the g-labels of S;

– 5) S is maximal, in the sense that no label or g-label pertaining to one of these CSP variables could be added to it without contradicting the first two conditions.

Remarks:

– as in the definition of transversal sets, the first condition will always be true for pairwise strongly disjoint CSP variables, but, for the same reasons as before, we do not take this as a necessary condition;

– conditions 2, 3 and 4 together express that constraint c plays a role for the whole transversal set; forgetting the idea of such a global constraint and adopting only condition 2 would not change the general theory developed in this chapter (but the totality of the second remark in section 8.1.1 after the definition of a transversal set applies here also);

– conversely, one could imagine replacing conditions 2, 3 and 4 by the stronger one: all the labels in S or contained in a g-label in S are pairwise linked by constraint c; in Sudoku, it is obvious that this would not change anything (because all the g-labels involve blocks); but for the general CSP, this may be too restrictive.

*10.1.1.2. gS$_p$-labels and gS$_p$-links*

Definitions: for any integer p>1, a *gS$_p$-label* is a couple of data: {CSPVars, TransvSets}, where CSPVars is a set of p different CSP variables and TransvSets is a set of p different g-transversal sets of labels and g-labels for these variables (each one for a well defined constraint). A gS-label is a gS$_p$-label for some p > 1.

Definition: a label l is *gS$_p$-linked* or simply *gS-linked* to a gS$_p$-label S = {CSPVars, TransvSets} if there is some k with $1 \leq k \leq p$ and such that:

– l is linked or g-linked to all the labels and g-labels in the k-th element TransvSets$_k$ of TransvSets,

– l is linked by the constraint c$_k$ of TransvSets$_k$ to all the labels and all the distinguished labels contained in all the g-labels in TransvSets$_k$.

In these conditions, l is also called *a potential target of the gS$_p$-label*.

Definition: Two sets of labels or g-labels are said to be "strongly transitively disjoint" if no label appearing in one of them (even as an element of a g-label) can appear in the other (even as an element of a g-label). This is much stronger than saying that these two sets are disjoint (in the usual set-theoretic sense of "disjoint"); if these sets are only disjoint, two g-labels, one in each of the sets, can be different but share a label (in Sudoku, <Xr1n1, r1n1c123> and <Xc1n1, c1n1r123> share the label with representative <r1c1, n1>); or a label in one set can be contained in a g-label in the other (e.g. <r1n1, c1> in <Xc1n1, c1n1r123>).

Definition: In a resolution state RS, two sets of candidates or g-candidates are said to be "transitively disjoint" if no candidate effectively present in one of them (even as an element of a g-candidate) is effectively present in the other (even as a candidate in a g-candidate); again this is much stronger than saying that these two sets are disjoint. However, this is weaker than saying that the sets obtained by considering the labels and g-labels underlying the candidates and g-candidates in these two sets are strongly transitively disjoint.

Miscellaneous remarks about gS$_p$-labels:

– with the above definition of a gS$_p$-label, a label and a g-label are not gS$_p$-labels (due to the condition p > 1); for labels, this is a mere matter of convention, but this choice is more convenient for the sequel;

– different transversal sets in a gS$_p$-label are not required to be pairwise transitively disjoint, let alone pairwise disjoint; such conditions will appear only in the definitions of g-Subsets and only with respect to candidates (not labels);

– a gS$_p$-label corresponds to the maximal extent of a possible gS$_p$-subset (as defined below), but it does not tackle non-degeneracy conditions.

Notation: in the definition of g-Subsets, as in the case of Subsets, we shall need a means of specifying that, in some g-transversal sets, some labels or g-labels must exist while others may exist or not. We shall write this as e.g. {<$V_1$, $v_1$>, <$V_2$, $v_2$>, …, (<$V_k$, $v_k$>), ….}. This should be understood as follows: a label or g-label not surrounded with parentheses must exist; a "label" or "g-label" surrounded with parentheses, like (<$V_k$, $v_k$>), may exist or not; if it exists, then it is named <$V_k$, $v_k$>.

### 10.1.2. g-Pairs

Definition: in any resolution state RS of any CSP, a *g-Pair* (or *$gS_2$-subset*) is a $gS_2$-label {CSPVars, TransvSets}, where:

– CSPVars = {$V_1$, $V_2$},

– TransvSets is composed of the following g-transversal sets of labels and g-labels:

   {<$V_1$, $v_{11}$>, <$V_2$, $v_{21}$>} for constraint $c_1$,
   {<$V_1$, $v_{12}$>, <$V_2$, $v_{22}$>} for constraint $c_2$,

such that:

– in RS, $V_1$ and $V_2$ are disjoint, i.e. they share no candidate;

– in RS, {<$V_1$, $v_{11}$>} and {<$V_1$, $v_{12}$>} are transitively disjoint; {<$V_2$, $v_{22}$>} and {<$V_2$, $v_{21}$>} are transitively disjoint;

– in RS, $V_1$ has the two mandatory candidates or g-candidates <$V_1$, $v_{11}$> and <$V_1$, $v_{12}$> and no other candidate or g-candidate;

– in RS, $V_2$ has the two mandatory candidates or g-candidates <$V_2$, $v_{21}$> and <$V_2$, $v_{22}$> and no other candidate or g-candidate.

A *target of a g-Pair* is a candidate $gS_2$-linked to the underlying $gS_2$-label.

**Theorem 10.1 ($gS_2$ rule): in any CSP, a target of a g-Pair can be eliminated.**

Proof: as the two g-transversal sets play similar roles, we can suppose that Z is linked or g-linked to <$V_1$, $v_{11}$> and <$V_2$, $v_{21}$>. If Z was True, these candidates or all the candidates these g-candidates contain would be eliminated by ECP. As $V_1$ and $V_2$ have only two candidates or g-candidates each, their other candidate or g-candidate (<$V_1$, $v_{12}$>, respectively <$V_2$, $v_{22}$>) would be or would contain their real value, which is impossible, as both are linked or g-linked. Here again, the proof works only because $V_1$ and $V_2$ share no candidate in RS (and in no posterior resolution state).

### 10.1.3. g-Triplets

There may be several formulations of g-Triplets. Here again, as in the case of ordinary Triplets, we adopt one that is neither too restrictive (the presence of some of the candidates or g-candidates potentially involved is not mandatory) nor too comprehensive (i.e., by making mandatory the presence of some of the candidates or g-candidates involved, it does not allow degenerated cases).

Definition: in any resolution state RS of any CSP, a *g-Triplet* (or *gS$_3$-subset*) is a gS$_3$-label {CSPVars, TransvSets}, where:

– CSPVars = {$V_1$, $V_2$, $V_3$},

– TransvSets is composed of the following g-transversal sets of labels and g-labels:

– {<$V_1$, $v_{11}$>, (<$V_2$, $v_{21}$>), <$V_3$, $v_{31}$>} for constraint $c_1$,

– {<$V_1$, $v_{12}$>, <$V_2$, $v_{22}$>, (<$V_3$, $v_{32}$>)} for constraint $c_2$,

– {(<$V_1$, $v_{13}$>), <$V_2$, $v_{23}$>, <$V_3$, $v_{33}$>} for constraint $c_3$,

such that:

– in RS, $V_1$, $V_2$ and $V_3$ are pairwise disjoint;

– in RS, {<$V_1$, $v_{11}$>} and {<$V_1$, $v_{12}$>} are transitively disjoint; {<$V_2$, $v_{22}$>} and {<$V_2$, $v_{23}$>} are transitively disjoint; {<$V_3$, $v_{33}$>} and {<$V_3$, $v_{31}$>} are transitively disjoint;

– in RS, $V_1$ has the two mandatory candidates or g-candidates <$V_1$, $v_{11}$> and <$V_1$, $v_{12}$>, one optional candidate or g-candidate <$V_1$, $v_{13}$> (supposing this label or g-label exists) and no other candidate or g-candidate;

– in RS, $V_2$ has the two mandatory candidates or g-candidates <$V_2$, $v_{22}$> and <$V_2$, $v_{23}$>, one optional candidate or g-candidate <$V_2$, $v_{21}$> (supposing this label or g-label exists) and no other candidate or g-candidate;

– in RS, $V_3$ has the two mandatory candidates or g-candidates <$V_3$, $v_{33}$> and <$V_3$, $v_{31}$>, one optional candidate or g-candidate <$V_3$, $v_{32}$> (supposing this label or g-label exists) and no other candidate or g-candidate.

A *target of a g-Triplet* is defined as a candidate gS$_3$-linked to the underlying gS$_3$-label.

***Theorem 10.2 (gS$_3$ rule): in any CSP, a target of a g-Triplet can be eliminated.***

Proof: as the three g-transversal sets play similar roles, we can suppose that Z is gS$_3$-linked to the first, i.e. linked or g-linked to <$V_1$, $v_{11}$>, <$V_2$, $v_{21}$> and <$V_3$, $v_{31}$> if it exists. If Z was True, these candidates or all the candidates these g-candidates contain (if they are present) would be eliminated by ECP. Each of $V_1$, $V_2$ and $V_3$ would have at most two candidates or g-candidates left. Any choice for $V_1$ would reduce to at most one the number of possibilities (in terms of candidates and g-candidates) for each of $V_2$ and $V_3$ (due to the pairwise contradictions between members of each g-transversal sets). Finally, the unique choice for $V_2$ (still in terms of candidates and g-candidates), if any, would in turn reduce to zero the number of possibilities for $V_3$.

### *10.1.4. g-Quads*

Finding the proper formulation for g-Quads, guaranteeing that it covers no degenerated case, is less obvious than for g-Triplets. Borrowing to the Quads case, we consider two types of g-Quads: Cyclic and Special, and we choose to write the Special g-Quad in such a way that it does not cover any case already covered by the Cyclic g-Quad.

Definition: in any resolution state RS of any CSP, a *Cyclic g-Quad* (or *Cyclic $gS_4$-subset*) is a $gS_4$-label {CSPVars, TransvSets}, where:

– CSPVars = {$V_1$, $V_2$, $V_3$, $V_4$},

– TransvSets is composed of the following g-transversal sets of labels and g-labels:

– {<$V_1$, $v_{11}$>, (<$V_2$, $v_{21}$>), (<$V_3$, $v_{31}$>), <$V_4$, $v_{41}$>} for constraint $c_1$,

– {<$V_1$, $v_{12}$>, <$V_2$, $v_{22}$>, (<$V_3$, $v_{32}$>), (<$V_4$, $v_{42}$>)} for constraint $c_2$,

– {(<$V_1$, $v_{13}$>), <$V_2$, $v_{23}$>, <$V_3$, $v_{33}$>, (<$V_4$, $v_{43}$>)} for constraint $c_3$,

– {(<$V_1$, $v_{14}$>), (<$V_2$, $v_{24}$>), <$V_3$, $v_{34}$>, <$V_4$, $v_{44}$>} for constraint $c_4$,

such that:

– in RS, $V_1$, $V_2$, $V_3$ and $V_4$ are pairwise disjoint, i.e. no two of these variables share a candidate;

– in RS, {<$V_1$, $v_{11}$>} and {<$V_1$, $v_{12}$>} are transitively disjoint; {<$V_2$, $v_{22}$>} and {<$V_2$, $v_{23}$>} are transitively disjoint; {<$V_3$, $v_{33}$>} and {<$V_3$, $v_{34}$>} are transitively disjoint; {<$V_4$, $v_{44}$>} and {<$V_4$, $v_{41}$>} are transitively disjoint;

– in RS, $V_1$ has the two mandatory candidates or g-candidates <$V_1$, $v_{11}$> and <$V_1$, $v_{12}$>, two optional candidates or g-candidates <$V_1$, $v_{13}$> and <$V_1$, $v_{14}$> (supposing any of these labels exists) and no other candidate or g-candidate;

– in RS, $V_2$ has the two mandatory candidates or g-candidates <$V_2$, $v_{22}$> and <$V_2$, $v_{23}$>, two optional candidates or g-candidates <$V_2$, $v_{24}$> and <$V_2$, $v_{21}$> (supposing any of these labels exists) and no other candidate or g-candidate;

– in RS, $V_3$ has the two mandatory candidates or g-candidates <$V_3$, $v_{33}$> and <$V_3$, $v_{34}$>, two optional candidates or g-candidates <$V_3$, $v_{31}$> and <$V_3$, $v_{32}$> (supposing any of these labels exists) and no other candidate or g-candidate;

– in RS, $V_4$ has the two mandatory candidates or g-candidates <$V_4$, $v_{44}$> and <$V_4$, $v_{41}$>, two optional candidates or g-candidates <$V_4$, $v_{42}$> and <$V_4$, $v_{43}$> (supposing any of these labels exists) and no other candidate or g-candidate.

Definition: in any resolution state RS of any CSP, a *Special g-Quad* (or *Special $gS_4$-subset*) is a $gS_4$-label {CSPVars, TransvSets}, where:

– CSPVars = {$V_1$, $V_2$, $V_3$, $V_4$},

– TransvSets is composed of the following transversal sets of labels and g-labels:

$\{<V_1, v_{11}>, <V_2, v_{21}>, <V_3, v_{31}>, (<V_4, v_{41}>\})$ for constraint $c_1$,
$\{<V_1, v_{12}>, (<V_2, v_{22}>), (<V_3, v_{32}>), <V_4, v_{42}>\}$ for constraint $c_2$,
$\{(<V_1, v_{13}>), <V_2, v_{23}>, (<V_3, v_{33}>), <V_4, v_{43}>\}$ for constraint $c_3$,
$\{(<V_1, v_{14}>), (<V_2, v_{24}>), <V_3, v_{34}>, <V_4, v_{44}>\}$ for constraint $c_4$,

such that:

– in RS, $V_1$, $V_2$, $V_3$ and $V_4$ are pairwise disjoint, i.e. no two of these variables share a candidate;

– in RS, $<V_1, v_{11}>$ and $<V_1, v_{12}>$ are transitively disjoint; $<V_2, v_{21}>$ and $<V_2, v_{23}>$ are transitively disjoint; $<V_3, v_{31}>$ and $<V_3, v_{34}>$ are transitively disjoint; moreover, $<V_4, v_{42}>$, $<V_4, v_{43}>$ and $<V_4, v_{44}>$ are pairwise transitively disjoint;

– in RS, $V_1$ has the two mandatory candidates or g-candidates $<V_1, v_{11}>$ and $<V_1, v_{12}>$ and no other candidate or g-candidate;

– in RS, $V_2$ has the two mandatory candidates or g-candidates $<V_2, v_{21}>$ and $<V_2, v_{23}>$ and $<V_1, v_{12}>$ and no other candidate or g-candidate;

– in RS, $V_3$ has the two mandatory candidates or g-candidates $<V_3, v_{31}>$ and $<V_3, v_{34}>$ and $<V_1, v_{12}>$ and no other candidate or g-candidate;

– in RS, $V_4$ has the three mandatory candidates or g-candidates $<V_4, v_{42}>$, $<V_4, v_{43}>$ and $<V_4, v_{44}>$ and no other candidate or g-candidate.

In both cases, a *target of a g-Quad* is defined as a candidate $gS_4$-linked to the underlying $gS_4$-label.

***Theorem 10.3 ($gS_4$ rule): in any CSP, a target of a Cyclic or Special g-Quad can be eliminated.***

Proof for the cyclic case: as the four g-transversal sets play similar roles, we can suppose that Z is linked or g-linked to $<V_1, v_{11}>$, $<V_2, v_{21}>$, $<V_3, v_{31}>$ if it exists and $<V_4, v_{41}>$ if it exists. If Z was True, these candidates or all the candidates these g-candidates contain (if they are present) would be eliminated by ECP. Each of $V_1$, $V_2$, $V_3$ and $V_4$ would have at most three candidates or g-candidates left. Any choice for $V_1$ would reduce to at most two the number of possibilities (in terms of candidates and g-candidates) for $V_2$, $V_3$ and $V_4$. Any further choice among the remaining candidates or g-candidates for $V_2$ would reduce to at most one the number of possibilities (still in terms of candidates and g-candidates) for $V_3$ and $V_4$. Finally, the unique choice left for $V_3$ (still in terms of candidates and g-candidates), if any, would reduce to zero the number of possibilities for $V_4$.

Proof for the special case: there are four subcases (the last two of which are similar to the second):
- suppose Z is linked or g-linked to both $<V_1, v_{11}>$, $<V_2, v_{21}>$, $<V_3, v_{31}>$ and $<V_4, v_{41}>$ if it exists. If Z was True, these candidates or all the candidates these g-candidates contain (if they are present) would be eliminated by ECP. Each of $V_1$, $V_2$, $V_3$, would have only one candidate or g-candidate left; choosing these candidates or any candidate in these g-candidates as their respective values would reduce to zero the number of possibilities for $V_4$.

- suppose Z is linked to both $<V_1, v_{12}>$, $<V_2, v_{22}>$ if it exists, $<V_3, v_{32}>$ and $<V_4, v_{42}>$ if it exists. If Z was True, $<V_1, v_{12}>$ and $<V_4, v_{42}>$ or all the candidates they contain would be eliminated by ECP; $<V_1, v_{11}>$ would then be True, which would eliminate $<V_2, v_{21}>$ and $<V_3, v_{31}>$ or all the candidates they contain. Then $<V_2, v_{23}>$ and $<V_3, v_{34}>$ would be True. This would leave no possibility for $V_4$.

If we wanted to introduce larger g-Subsets, it would get harder and harder to write separate formulæ guaranteeing non-degeneracy of each subcase. We leave this as a (difficult) exercise for the reader. Contrary to Subsets, in the 9×9 Sudoku case, there can be g-Subsets of size larger than 4: see the example of a Franken Squirmbag (size 5) in section 10.3.

### 10.1.5. $gS_p$-subset theories and confluence

All of section 8.6 can be transposed and extended from $S_p$-subsets to $gS_p$-subsets: definition of the $gS_p$ resolution theories, proof of their stability for confluence, definition of the $gW_p+gS_p$ and $gB_p+gS_p$ resolution theories (in which g-whips[p] or g-braids[p] are added to $gS_p$-subsets) and proof of their stability for confluence.

### 10.1.6. Subsumption results for g-Subsets

*10.1.6.1. g-Pairs*

**Theorem 10.4: $gS_2 \subseteq gW_2$ (g-whips of length 2 subsume all the g-Pairs).**

Proof: keeping the notations of theorem 10.1 and considering a target Z of the g-Pair that is $gS_2$-linked to the first g-transversal set, i.e. linked or g-linked to both $<V_1, v_{11}>$ and $<V_2, v_{21}>$, the following g-whip[2] eliminates Z:
g-whip[2]: $V_1\{v^*_{11}\ v_{12}\} - V_2\{v^*_{22}\ .\} \Rightarrow \neg candidate(Z)$,
where $v^*_{11}$ [respectively $v^*_{22}$] is $v_{11}$ [resp. $v_{22}$] if $<V_1, v_{11}>$ [resp. $<V_2, v_{22}>$] is a candidate and it is any element chosen in $v_{11}$ [resp. $v_{22}$] if $<V_1, v_{11}>$ [resp. $<V_2, v_{22}>$] is a g-candidate. In case $<V_1, v_{11}>$ is a g-candidate, the candidates in $<V_1, v_{11}>$ other than $<V_1, v^*_{11}>$ are z-candidates in the whip[2]; in case $<V_2, v_{22}>$ is a g-candidate, the candidates in $<V_2, v_{22}>$ other than $<V_2, v^*_{22}>$ are t-candidates in the whip[2].

**The converse of the above theorem is false: $gW_2 \not\subseteq gS_2$; indeed, $W_2 \not\subseteq gS_2$.** For a deep understanding of both whips and g-Subsets, this is as interesting as the theorem itself. Using the example in section 8.8.1, we have concluded in section 8.7.1 that $W_2 \not\subseteq S_2$. But the same example can now be used to show that $W_2 \not\subseteq gS_2$.

The three whips[2] defined in section 8.8.1 can be considered no more as g-Pairs than as Pairs.

It is nevertheless instructive to understand how a tentative proof of the inclusion gW$_2 \subseteq$ gS$_2$ would fail. We would have to proceed as follows. Let
g-whip[2]: V$_1$\{v$_{11}$ v$_{12}$\} − V$_2$\{v$_{22}$ .\} $\Rightarrow$ ¬candidate(Z), be a g-whip[2] with target Z, associated with CSP variables (V$_1$, V$_2$). Consider all the possible candidates or g-candidates for each of these variables.

For V$_1$, they can only be:

− <V$_1$, v'$_{11}$> = the candidate or g-candidate consisting of <V$_1$, v$_{11}$> and all the candidates for V$_1$ linked to Z;

− and <V$_1$, v'$_{12}$> = <V$_1$, v$_{12}$> (a candidate or a g-candidate, with no element linked to Z).

For V$_2$, they can only be:

− <V$_2$, v'$_{22}$> = the candidate or g-candidate consisting of <V$_2$, v$_{22}$> and all the candidates for V$_2$ linked to <V$_1$, v$_{12}$>;

− and <V$_2$, v'$_{21}$> = the candidate or g-candidate consisting of all the candidates for V$_2$ linked to Z but not to <V$_1$, v$_{12}$>.

We have thus built a g-transversal set \{<V$_1$, v'$_{12}$>, <V$_2$, v'$_{22}$>\}, but \{<V$_1$, v'$_{11}$>, <V$_2$, v'$_{21}$>\} may not be a g-transversal set: the target Z is linked to these two candidates or g-candidates that may not be linked together by any constraint. This is exactly the situation with the three whips[2] in the example of section 8.8.1.

*10.1.6.2. g-Triplets*

### Theorem 10.5: gW$_3$ subsumes "almost all" the g-Triplets.

Proof: keeping the notations of theorem 10.3 and considering a target Z of the g-Triplet that is gS-linked to the first g-transversal set (the three of them play similar roles), the following g-whip eliminates Z in any CSP:
g-whip[3]: V$_1$\{v*$_{11}$ v$_{12}$\} − V$_2$\{v*$_{22}$ v$_{23}$\} − V$_3$\{v*$_{33}$ .\} $\Rightarrow$ ¬candidate(Z),
provided that <V$_1$, v$_{13}$> is not a candidate or a g-candidate for V$_1$.
Here, v*$_{11}$ [respectively v*$_{22}$, v*$_{33}$] is v$_{11}$ [resp. v$_{22}$, v$_{33}$] if <V$_1$, v$_{11}$> [resp. <V$_2$, v$_{22}$>, <V$_3$, v$_{33}$>] is a candidate and it is any element chosen in v$_{11}$ [resp. v$_{22}$, v$_{33}$] if <V$_1$, v$_{11}$> [resp. <V$_2$, v$_{22}$>, <V$_3$, v$_{33}$>] is a g-candidate.
The optional candidates and the elements of the optional g-candidates of the g-Triplet appear in the g-whip as z- or t- candidates.

Considering that, in the above situation, the three CSP variables play symmetrical roles, there is only one case of a g-Triplet elimination that cannot be replaced by a g-whip[3] elimination. It occurs when the optional candidates or g-candidates for variables V$_1$, V$_2$ and V$_3$ in the g-transversal set to which the target is gS-linked correspond to existing labels or g-labels and are all effectively present in the resolution state.

*10.1.6.3. g-Quads*

### Theorem 10.6: $gW_4$ subsumes "almost all" the Cyclic g-Quads.

Keeping the notations of theorem 10.5, the following g-whip eliminates a target Z of the Cyclic g-Quad in any CSP:
g-whip[4]:  $V_1\{v^*_{11}\ v_{12}\}$  –  $V_2\{v^*_{22}\ v_{23}\}$  –  $V_3\{v^*_{33}\ v_{34}\}$  –  $V_4\{v^*_{41}\ .\}$  ⇒ ¬candidate(Z),
provided that $<V_1, v_{13}>$ and $<V_1, v_{14}>$ are not candidates for $V_1$ and $<V_2, v_{23}>$ is not a candidate for $V_2$,
with the $v^*_{xy}$ defined as before.
The optional candidates and the elements of the optional g-candidates of the g-Quad appear in the g-whip as z- or t- candidates.

### Theorem 10.7: $gB_4$ subsumes all the Special g-Quads.

Keeping the notations of theorem 10.5, let Z be a target of the Special g-Quad:
- if Z is $gS_4$-linked to the first g-transversal set, the following g-braid eliminates Z:
g-braid[4]: $V_1\{v^*_{11}\ v_{12}\}$  –  $V_2\{v^*_{21}\ v_{23}\}$  –  $V_3\{v^*_{31}\ v_{34}\}$  –  $V_4\{v^*_{44}\ .\}$  ⇒ ¬candidate(Z),
in which the first three left-linking candidates are linked to Z;
- if Z is $gS_4$-linked to another g-transversal set, say the second, the following g-whip eliminates Z:
g-whip[4]: $V_1\{v^*_{12}\ v_{11}\}$  –  $V_2\{v^*_{21}\ v_{23}\}$  –  $V_4\{v^*_{43}\ v_{44}\}$  –  $V_3\{v^*_{34}\ .\}$  ⇒ ¬candidate(Z),
in which candidate $<V_4, v_{42}>$ appears as a z-candidate for the third CSP variable.

*10.1.7. g-Subsets in Sudoku*

Although the concept of a g-Subset has never been considered as such in Sudoku, the point we want to make here is that g-Subsets have been in existence for a very long time, under other names: they appear as the "Franken Fish" and "Mutant Fish" patterns.

The difference between the two kinds depends on the specific geometry of Sudoku and is of little interest for the general theory developed here. Let us therefore mention it quickly, transposed into the vocabulary of this book. For a given number n°, a standard Fish in rows [respectively in columns] (of size p) uses only p different Xrn° [resp. Xcn°] CSP variables and p different transversal sets defined by Xcn° [resp. Xrn°] constraints. A Franken Fish in rows (of size p) is defined as an extended Fish (Super-Hidden Subset) pattern of size p in which either some of the p CSP variables are of type Xbn° instead of Xrn° [resp. Xcn°] *or* some of the p transversal sets are defined by Xbn° constraints instead of Xcn° [resp. Xrn°] constraints. In a Mutant Fish, rows, columns and blocks may all appear in both CSP variables (i.e. these may be Xr°c°, Xrn°, Xcn° and Xbn°) and in constraints defining the transversal sets, which makes them much more complex than Franken Fish.

For more details and for examples, see sudopedia.org. For a (maybe not exhaustive) review of the various possibilities, we direct the reader to the specialised forums, where he will find that there is a handful of people who consecrate their time to studying and naming them (together with their "finned", "sushi", "sashimi" and other extensions). There is also a recent free java Sudoku solver, specialised in Fish: Hodoku – as far as we know, the only solver implementing (almost) all the known possibilities. See also the detailed example in section 10.3 below.

## 10.2. Reversible-gS$_p$-chains, gS$_p$-whips and gS$_p$-braids

When we try to apply the zt-ing principle to g-Subsets, everything goes for gS$_p$-whips and gS$_p$-braids as for S$_p$-whips and S$_p$-braids. Here again, when it comes to defining the concepts of gS$_p$-links and gS$_p$-compatibility, we always consider the gS$_p$-labels underlying the gS$_p$-subsets instead of the gS$_p$-subsets themselves, in exactly the same way as we considered the full S$_p$-labels underlying the S$_p$-subsets when we defined S$_p$-links. The main reason for this choice is the same as in the S$_p$-links case: we want all the notions related to linking and compatibility to be purely structural (see chapter 9 for more detail).

### 10.2.1. gS$_p$-links; gS$_p$-subsets modulo other g-Subsets; gS$_p$-regular sequences

#### 10.2.1.1. gS$_p$-links, gS$_p$-compatibility

Definition: a label l is *compatible with a gS$_p$-label* S if l is not gS$_p$-linked to S (i.e. if, for each g-transversal set TS of S, there is at least one label or g-label l' in TS such that l is not linked or g-linked to l').

Definition: a label l is *compatible* with a set R of labels, g-labels, S-labels and gS-labels if l is compatible with each element of R (in the senses of "compatible" already defined separately for labels, g-labels, S$_p$-labels and gS$_p$-labels).

Definitions: a label l is *gS$_p$-linked to a gS$_p$-subset* S if l is gS$_p$-linked to the gS$_p$-label underlying S; a label l is compatible with a gS$_p$-subset if l is not gS$_p$-linked to it; a label l is *compatible* with a set R of candidates, g-candidates, Subsets and g-Subsets if l is compatible with each element of R (in the senses of "compatible" already defined separately for candidates, g-candidates, S$_p$-subsets and gS$_p$-subsets).

Notice that, in conformance with what we mentioned at the beginning of section 10.2, according to the definition of "gS$_p$-linked to a gS$_p$-subset", it is not enough for label l to be linked or g-linked to all the actual candidates and g-candidates of one of its transversal sets: it must be linked or g-linked to all the labels and g-labels of one of its transversal sets.

*10.2.1.2. $gS_p$-subsets modulo a set of labels, g-labels, S-labels and gS-labels*

All our forthcoming definitions (Reversible-$gS_p$-chains, $gS_p$-whips and $gS_p$-braids) will be based on that of a $gS_p$-subset modulo a set R of labels, g-labels, S-labels and gS-labels; in practice, R will be either the previous right-linking pattern or the set consisting of the target plus all the previous right-linking patterns (i.e. candidates, g-candidates, $S_k$-subsets and $gS_k$-subsets).

Definition: in any resolution state of any CSP, given a set R of labels, g-labels, S-labels and gS-labels [or a set R of candidates, g-candidates, Subsets and g-Subsets], a *g-Pair (or $gS_2$-subset) modulo R* is a $gS_2$-label {CSPVars, TransvSets}, where:

– CSPVars = {$V_1$, $V_2$},

– TransvSets is composed of the following transversal sets of labels and g-labels:

– {<$V_1$, $v_{11}$>, <$V_2$, $v_{21}$>} for constraint $c_1$,

– {<$V_1$, $v_{12}$>, <$V_2$, $v_{22}$>} for constraint $c_2$,

such that:

– in RS, $V_1$ and $V_2$ are disjoint, i.e. they share no candidate;

– in RS, {<$V_1$, $v_{11}$>} and {<$V_1$, $v_{12}$>} are transitively disjoint; in RS, {<$V_2$, $v_{22}$>} and {<$V_2$, $v_{21}$>} are transitively disjoint;

– in RS, $V_1$ has the two mandatory candidates or g-candidates <$V_1$, $v_{11}$> and <$V_1$, $v_{12}$> compatible with R and no other candidate or g-candidate compatible with R;

– in RS, $V_2$ has the two mandatory candidates or g-candidates <$V_2$, $v_{21}$> and <$V_2$, $v_{22}$> compatible with R and no other candidate or g-candidate compatible with R.

Definition: in any resolution state of any CSP, given a set R of labels, g-labels, S-labels and gS-labels [or a set R of candidates, g-candidates, Subsets and g-Subsets], a *g-Triplet (or $gS_3$-subset) modulo R* is a $gS_3$-label {CSPVars, TransvSets}, where:

– CSPVars = {$V_1$, $V_2$, $V_3$},

– TransvSets is composed of the following transversal sets of labels and g-labels:

– {<$V_1$, $v_{11}$>, (<$V_2$, $v_{21}$>), <$V_3$, $v_{31}$>} for constraint $c_1$,

– {<$V_1$, $v_{12}$>, <$V_2$, $v_{22}$>, (<$V_3$, $v_{32}$>)} for constraint $c_2$,

– {(<V$_1$, v$_{13}$>), <V$_2$, v$_{23}$>, <V$_3$, v$_{33}$>} for constraint c$_3$,

such that:

– in RS, V$_1$, V$_2$ and V$_3$ are pairwise disjoint;

– in RS, {<V$_1$, v$_{11}$>} and {<V$_1$, v$_{12}$>} are transitively disjoint; {<V$_2$, v$_{22}$>} and {<V$_2$, v$_{23}$>} are transitively disjoint; {<V$_3$, v$_{33}$>} and {<V$_3$, v$_{31}$>} are transitively disjoint;

– in RS, V$_1$ has the two mandatory candidates or g-candidates <V$_1$, v$_{11}$> and <V$_1$, v$_{12}$> compatible with R, one optional candidate or g-candidate <V$_1$, v$_{13}$> compatible with R (supposing this label or g-label exists), and no other candidate or g-candidate compatible with R;

– in RS, V$_2$ has the two mandatory candidates or g-candidates <V$_2$, v$_{22}$> and <V$_2$, v$_{23}$> compatible with R, one optional candidate or g-candidate <V$_2$, v$_{21}$> compatible with R (supposing this label or g-label exists), and no other candidate or g-candidate compatible with R;

– in RS, V$_3$ has the two mandatory candidates or g-candidates <V$_3$, v$_{33}$> and <V$_3$, v$_{31}$> compatible with R, one optional candidate or g-candidate <V$_3$, v$_{32}$> compatible with R (supposing this label or g-label exists), and no other candidate or g-candidate compatible with R.

We leave it to the reader to write the definitions of g-Subsets of larger sizes modulo R (gS$_p$-subsets modulo R). The general idea is that, when one looks in RS at some gS$_p$-label "modulo R", i.e. when all the candidates and g-candidates in RS incompatible with R are "forgotten", what remains in RS satisfies the conditions of a non degenerated g-Subset of size p based on this gS$_p$-label.

Definition: in all the above cases, *a target of the gS$_p$-subset modulo R* is defined as a target of the gS$_p$-subset itself (i.e. as a candidate gS$_p$-linked to its underlying gS$_p$-label).

The idea is that, in any context (e.g. in a chain) in which the elements in R have positive valence, the gS$_p$-subset itself will have positive valence and any of its targets will have negative valence.

### 10.2.1.3. gS$_p$-regular sequences

As in the previous chapter, it is convenient to introduce an auxiliary notion before we define Reversible-gS$_p$-chains, gS$_p$-whips and gS$_p$-braids.

Definition: let there be given an integer $1 \leq p \leq \infty$, an integer $m \geq 1$, a sequence ($q_1$, …, $q_m$) of integers, with $1 \leq q_k \leq p$ for all $1 \leq k \leq m$, and let $n = \sum_{1 \leq k \leq m} q_k$; let there also be given a sequence ($W_1$, …, $W_m$) of different sets of CSP variables of respective cardinalities $q_k$ and a sequence ($V_1$, …, $V_m$) of CSP variables such that $V_k \in W_k$ for all $1 \leq k \leq m$. We define *a gS$_p$-regular sequence of length n associated with ($W_1$, …* $W_m$) and ($V_1, ... V_m$) to be a sequence of length 2m [or 2m-1] ($L_1$, $R_1$, $L_2$, $R_2$, .... $L_m$, [$R_m$]), such that:

– $q_m=1$ and $W_m = \{V_m\}$,

– for 1≤k≤ m, $L_k$ is a candidate;

– for 1≤k≤ m [or 1≤k<m], $R_k$ is a candidate or a g-candidate if $q_k=1$ and it is a (non degenerated) $Sq_k$-subset or $gSq_k$-subset if $q_k>1$;

– for each 1≤k≤ m [or 1≤k<m], one has *"strong continuity", "strong g-continuity", "strong $Sq_k$-continuity" or "strong $gSq_k$-continuity"* from $L_k$ to $R_k$:

   - if $R_k$ is a candidate ($q_k=1$ and $W_k=\{V_k\}$), $L_k$ and $R_k$ have a representative with $V_k$: <$V_k$, $l_k$> and <$V_k$, $r_k$>,

   - if $R_k$ is a g-candidate ($q_k=1$ and $W_k=\{V_k\}$), $L_k$ has a representative <$V_k$, $l_k$> with $V_k$ and $R_k$ is a g-candidate <$V_k$, $r_k$> for Vk ($r_k$ being its set of values),

   - if $R_k$ is an $Sq_k$-subset or a $gSq_k$-subset ($q_k>1$), then $W_k$ is its set of CSP variables and $L_k$ has a representative with $V_k$.

The $L_k$ are called the *left-linking candidates* of the sequence and the $R_k$ the *right-linking objects (or elements or patterns or g-Subsets)*. Notice that the natural expression of $L_k$ to $R_k$ continuity in case $R_k$ is a g-Subset is the same as if it is a Subset.

Notice also that the definition of a g-Subset implies a disjointness condition on the sets of candidates for the CSP variables inside each $W_k$, but for a $gS_p$-regular sequence there is no condition on the intersections of different $W_k$'s. In particular, $W_{k+i}$ may be a strict subset of $W_k$, if the right-linking elements in between give negative valence in $W_{k+i}$ to some candidates or g-candidates that had no individual valence assigned in $W_k$. This is not considered as an inner loop of the sequence.

### 10.2.2. Reversible-$gS_p$-chains

Reversible-$gS_p$-chains are an extension of Reversible-$S_p$-chains in which right-linking $S_{p'}$-subsets may be replaced by $gS_{p'}$-subsets (p'≤p).

#### 10.2.2.1. Definition of Reversible-$gS_p$-chains

Definition: given an integer 1≤p≤∞ and a candidate Z (which will be a target), a *Reversible-$gS_p$-chain* of length n (n ≥ 1) is a $gS_p$-regular sequence ($L_1$, $R_1$, $L_2$, $R_2$, .... $L_m$, $R_m$) of length n associated with a sequence ($W_1$, … $W_m$) of sets of CSP variables and a sequence ($V_1$, … $V_m$) of CSP variables (with $V_k \in W_k$ for all 1≤k<m), such that:

– Z is neither equal to any candidate in {$L_1$, $R_1$, $L_2$, $R_2$, …. $L_m$, $R_m$} nor a member of any g-candidate in this set nor equal to any label in the $Sq_k$-label or $gSq_k$-label of $R_k$ when $R_k$ is an $Sq_k$-subset or a $gSq_k$-subset, for any 1≤k<m;

– Z is linked to L$_1$;

– for each $1 < k \leq m$, L$_k$ is linked or g-linked or Sq$_{k-1}$-linked or gSq$_{k-1}$-linked to R$_{k-1}$; this is the natural way of defining *"continuity" from R$_{k-1}$ to L$_k$*;

– R$_1$ is a candidate or a g-candidate or an Sq$_1$-subset or a gSq$_1$-subset modulo Z: R$_1$ is the only candidate or g-candidate or is the unique Sq$_1$-subset or is the unique gSq$_1$-subset composed of all the candidates C for the CSP variables in W$_1$ such that C is compatible with Z;

– for any $1 < k \leq m$, R$_k$ is a candidate or a g-candidate or an Sq$_k$-subset or a gSq$_k$-subset modulo R$_{k-1}$: R$_k$ is the only candidate or g-candidate or (if k≠m) is the unique Sq$_k$-subset or (if k≠m) is the unique gSq$_k$-subset composed of all the candidates C for the CSP variables in W$_k$ such that C is compatible with R$_{k-1}$;

– Z is not a label for V$_m$;

– Z is linked to L$_1$ and to R$_m$.

***Theorem 10.8 (Reversible-gS$_p$-chain rule for a general CSP): in any resolution state of any CSP, if Z is a target of a Reversible-gS$_p$-chain, then it can be eliminated (formally, this rule concludes ¬ candidate(Z)).***

Proof: if Z was True, then L$_1$ would be eliminated by ECP and R$_1$ would be asserted by S (if it is a candidate) or it would be a g-candidate or an Sq$_1$-subset or a gSq$_1$-subset; in any case, L$_2$ would be eliminated by ECP or W$_1$ or Sq$_1$ or gSq$_1$. By induction, we arrive at: R$_m$ would be asserted by S or it would be a g-candidate – which would contradict Z being True.

*10.2.2.2. Reversibility of Reversible-gS$_p$-chains in the general CSP*

The following theorem justifies the name we have given these chains.

***Theorem 10.9: a Reversible-gS$_p$-chain is reversible***.

Proof: the main point of the proof is the construction of the reversed chain. As it is a simple transposition of the proof for Reversible-S$_p$-chains in section 9.2.2, we leave it as an exercise for the reader. Figure 9.1 can still be used as a partial visual support for the proof, but now the intersections between horizontal lines (CSP variables) and vertical lines (g-transversal sets) must be interpreted as candidates or g-candidates for these CSP variables instead of only candidates.

*10.2.3. gS$_p$-whips and gS$_p$-braids*

gS$_p$-whips and gS$_p$-braids are an extension of g-whips and g-braids in which S$_{p'}$-subsets and gS$_{p'}$-subsets (p'≤p) may appear as right-linking patterns. They can also be seen as extensions of the Reversible-gS$_p$-chains by application of the zt-ing instead of the almost-ing principle.

*10.2.3.1. Definition of gS$_p$-whips*

Definition: given an integer $1 \leq p \leq \infty$ and a candidate Z (which will be the target), a *gS$_p$-whip* of length n (n ≥ 1) built on Z is a gS$_p$-regular sequence (L$_1$, R$_1$, L$_2$, R$_2$, …. L$_m$) [notice that there is no R$_m$] of length n, associated with a sequence (W$_1$, … W$_m$) of sets of CSP variables and a sequence (V$_1$, … V$_m$) of CSP variables (with V$_k \in$ W$_k$ for all 1≤k<m and W$_m$ = {V$_m$}), such that:

– Z is neither equal to any candidate in {L$_1$, R$_1$, L$_2$, R$_2$, …. L$_m$} nor a member of any g-candidate in this set nor equal to any label in the Sq$_k$-label or gSq$_k$-label of R$_k$ when R$_k$ is an Sq$_k$-subset or a gSq$_k$-subset, for any 1≤k<m;

– L$_1$ is linked to Z;

– for each 1 < k ≤ m, L$_k$ is linked or g-linked or Sq$_{k-1}$-linked or gSq$_{k-1}$-linked to R$_{k-1}$; this is a form of "continuity" from R$_{k-1}$ to L$_k$;

– for any 1 ≤ k < m, R$_k$ is a candidate or a g-candidate or an Sq$_k$-subset or a gSq$_k$-subset modulo Z and all the previous right-linking patterns: either R$_k$ is the only candidate or g-candidate compatible with Z and with all the R$_i$ with 1≤ i< k, or R$_k$ is the unique Sq$_k$-subset or gSq$_k$-subset composed of all the candidates C for some of the CSP variables in W$_k$ such that C is compatible with Z and with all the R$_i$ with 1≤ i< k;

– Z is not a label for V$_m$;

– V$_m$ has no candidate compatible with the target and with all the previous right-linking objects (but V$_m$ has more than one candidate).

***Theorem 10.10 (gS$_p$-whip rule for a general CSP): in any resolution state of any CSP, if Z is a target of a gS$_p$-whip, then it can be eliminated (formally, this rule concludes ¬candidate(Z)).***

Proof: the proof is an easy adaptation of that for the S$_p$-whips. Supposing Z was True and iterating upwards: R$_{k-1}$ would be asserted by S or it would be a g-candidate or an Sq$_{k-1}$-subset or a gSq$_{k-1}$-subset; due to R$_{k-1}$ to L$_k$ continuity, L$_k$ would be eliminated by rule ECP, W$_1$, Sq$_{k-1}$ or gSq$_{k-1}$; as usual the z- and t- candidates would be progressively eliminated. When m-1 is reached, R$_{m-1}$ would have positive valence and there would be no possible value left for V$_m$ (because Z itself is not a label for V$_m$).

*10.2.3.2. Definition of gS$_p$-braids*

Definition: given an integer $1 \leq p \leq \infty$ and a candidate Z (which will be the target), a *gS$_p$-braid* of length n (n ≥ 1) built on Z is a gS$_p$-regular sequence (L$_1$, R$_1$, L$_2$, R$_2$, …. L$_m$) [notice that there is no R$_m$] of length n, associated with a sequence (W$_1$, … W$_m$) of sets of CSP variables and a sequence (V$_1$, … V$_m$) of CSP variables (with V$_k \in$ W$_k$ for all 1≤k<m and W$_m$ = {V$_m$}), such that:

  – Z is neither equal to any candidate in {$L_1$, $R_1$, $L_2$, $R_2$, …. $L_m$} nor a member of any g-candidate in this set nor equal to any label in the $Sq_k$-label or $gSq_k$-label of $R_k$ when $R_k$ is an $Sq_k$-subset or a $gSq_k$-subset, for any $1 \leq k < m$;

  – $L_1$ is linked to Z;

  – for each $1 < k \leq m$, $L_k$ is linked or g-linked or S-linked or gS-linked to Z or to some of the $R_i$, $i<k$; this is the only difference with $gS_p$-whips;

  – for any $1 \leq k < m$, $R_k$ is a candidate or a g-candidate or an $Sq_k$-subset or a $gSq_k$-subset modulo Z and all the previous right-linking patterns: either $R_k$ is the only candidate or g-candidate compatible with Z and with all the $R_i$ with $1 \leq i < k$, or $R_k$ is the unique $Sq_k$-subset or $gSq_k$-subset composed of all the candidates C for some of the CSP variables in $W_k$ such that C is compatible with Z and with all the $R_i$ with $1 \leq i < k$;

  – Z is not a label for $V_m$;

  – $V_m$ has no candidate compatible with the target and with all the previous right-linking objects (but $V_m$ has more than one candidate).

**Theorem 10.11 ($gS_p$-braid rule for a general CSP): in any resolution state of any CSP, if Z is a target of a $gS_p$-braid, then it can be eliminated (formally, this rule concludes ¬candidate(Z)).**

Proof: almost the same as the proof for $gS_p$-whips. The condition replacing $R_{k-1}$ to $L_k$ continuity still allows the elimination of $L_k$ by ECP.

*10.2.3.3. $gS_p$-whip and $gS_p$-braid resolution theories; $gS_pW$ and $gS_pB$ ratings*

In the same way as for the $S_p$-whips or $S_p$-braids cases, one can define increasing sequences of resolution theories, with two parameters, one (n) for the total length of the chain and one (p) for the maximum size of inner Subsets or g-Subsets. By convention, p=1 means no Subset or g-Subset, only candidates and g-candidates.

Definition: for each p, $1 \leq p \leq \infty$, define the increasing sequence ($gS_pW_n$, $n \geq 0$) of resolution theories as follows (similar definitions can be given for $gS_p$-braids, by replacing everywhere "$gS_p$-whip" by "$gS_p$-braid" and "$gS_pW$" by "$gS_pB$"):

  – $gS_pW_0$ = BRT(CSP),

  – $gS_pW_1 = gS_pW_0 \cup$ {rules for $gS_p$-whips of length 1} = $W_1$,

  – $gS_pW_2 = gS_pW_1 \cup gS_2$ (if $p \geq 2$) $\cup$ {rules for $gS_p$-whips of length 2},

  – ....

  – $gS_pW_n = gS_pW_{n-1} \cup gS_n$ (if $p \geq n$) $\cup$ {rules for $gS_p$-whips of length n},

  – $gS_pW_\infty = \cup_{n \geq 0} gS_pW_n$.

For p=∞, i.e. for gS-whips built on g-Subsets of *a priori* unrestricted size, we also write $gSW_n$ instead of $gS_\infty W_n$.

Definitions: for any $1 \leq p \leq \infty$, the ***gS$_p$W-rating*** of an instance P, noted gS$_p$W(P), is the smallest $n \leq \infty$ such that P can be solved within gS$_p$W$_n$. Similarly, for any $1 \leq p \leq \infty$, the ***gS$_p$B-rating*** of an instance P, noted gS$_p$B(P), is the smallest $n \leq \infty$ such that P can be solved within gS$_p$B$_n$.

Obviously, setting p=1, gS$_1$W$_n$ = gW$_n$ and gS$_1$W(P) = gW(P) for any instance. Similarly, gS$_1$B$_n$ = gB$_n$ and gS$_1$B(P) = gB(P) for any instance

For any $1 \leq p \leq \infty$, the gS$_p$W and gS$_p$B ratings are defined in a purely logical way, independent of any implementation; they are intrinsic properties of each instance; moreover, for any fixed p ($1 \leq p \leq \infty$), the gS$_p$B rating is based on an increasing sequence of theories (gS$_p$B$_n$, $n \geq 0$) with the confluence property (theorem 10.12).

For any puzzle P, one has obviously gW(P) = gS$_1$W(P) $\geq$ gS$_2$W(P) $\geq$ gS$_p$W(P) $\geq$ gS$_{p+1}$W(P) $\geq$ … $\geq$ gS$_\infty$W(P) and similar inequalities for the gS$_p$B(P).

*10.2.3.4. The confluence property of all the gS$_p$B$_n$ resolution theories*

***Theorem 10.12: in any CSP, each of the gS$_p$B$_n$ resolution theories ($1 \leq p \leq \infty$, $0 \leq n \leq \infty$) is stable for confluence; therefore, it has the confluence property.***

Proof: as it is a simple adaptation of the proof for the S$_p$B$_n$ resolution theories, we leave it as an exercise for the reader. We could even allow type-2 targets.

As usual, the confluence property of all the gS$_p$B$_n$ resolution theories for each p, $1 \leq p \leq \infty$, allows to superimpose on gS$_p$B$_n$ a "simplest first" strategy compatible with the gS$_p$B rating.

*10.2.3.5. The "T&E(gS$_p$) vs gS$_p$-braids" theorem, $1 \leq p \leq \infty$*

Any resolution theory T stable for confluence has the confluence property and the procedure T&E(T) can therefore be defined (see section 5.6.1). Taking T = gS$_p$, it is obvious that any elimination done by a gS$_p$-braid can be done by T&E(gS$_p$). As was the case for braids, for g-braids and for S$_p$-braids, the converse is true:

***Theorem 10.13: for any $1 \leq p \leq \infty$, any elimination done by T&E(gS$_p$) can be done by a gS$_p$-braid.***

As the proof closely follows that for S$_p$-braids, we leave it to the reader.

*10.2.4. gS$_p$-z-whips and their relationships with gS$_p$-subsets*

All the definitions and results in sections 9.3.4 to 9.3.6 can be extended with only slight changes. Informally speaking, a gS$_p$-z-whip[n] can be defined as a gS$_p$-whip[n] with no t-candidate that is not also a z-candidate. And gS$_p$-z-whip resolution theories can be defined and shown to have the confluence property.

***Theorem 10.14: a target Z of a $gS_p$-subset is always a target of a $gS_{p-1}$-z-whip of length p***.

Proof: almost obvious. One can always suppose that Z is $gS_p$-linked to $TS_1$ and that $V_1$ has a candidate or a g-candidate to which Z is linked or g-linked. Let $L_1$ = <$V_1$, $l_1$> be this candidate or any candidate in this g-candidate. Let $L_p$ = <$V_p$, $l_p$> be a candidate for $V_p$ not in $TS_1$ (there must be one if the $gS_p$-subset is not degenerated). Let $R_2$ be the $gS_{p-1}$-subset: {{$V_1$, …, $V_{p-1}$}, {$TS_2$, …, $TS_p$}}. Then the desired chain is $RgS_{p-1}C[p]$: {$L_1$ $R_2$} – $V_p${$l_p$ .}. qed.

A $gS_p$-subset that has g-transversal sets with "transitively non-void" intersections allows more eliminations than the "standard" ones defined in section 10.1. (This can happen only for p>2.)

Definition: a type-2 target of a $gS_p$-subset is a candidate belonging, either as an element or as a member of a g-label, to (at least) two of its g-transversal sets.

***Theorem 10.15: a type-2 target of a $gS_p$-subset can be eliminated.***

Proof: suppose a type-2 target Z is a candidate for variable $V_1$ and belongs to g-transversal sets $TS_1$ and $TS_2$. If Z was True, then all the candidates in $TS_1$ or $TS_2$ or in a g-label in $TS_1$ or $TS_2$ would be eliminated by ECP. This would leave only p-2 possibilities (in terms of candidates or g-candidates) for the remaining p-1 CSP variables – which is contradictory, in the same way as if it was for a normal target.

As in the case of $S_p$-subsets, this is a very unusual kind of elimination. The following theorem shows that, here also, this abnormality can be palliated. It also justifies that we did not consider type-2 targets of $gS_p$-subsets in section 10.1: these abnormal targets can always be eliminated by a simpler pattern. An illustration of the following theorem will appear in section 10.3 (for a "Franken Squirmbag").

***Theorem 10.16: A type-2 target of a $gS_p$-subset is always the (normal) target of a shorter $gS_{p-2}$-z-whip of length p-1.***

Proof: in a resolution state RS, let Z be a type-2 target of a $gS_p$-subset with CSP variables $V_1$, … $V_p$ and g-transversal sets $TS_1$, … $TS_p$. One can always suppose that $V_1$ is the CSP variable for which Z is a candidate (there can be only one in RS) and that $TS_1$ and $TS_2$ are the two g-transversal sets to which Z belongs.

Firstly, each of the CSP variables $V_2$, $V_3$, … $V_p$ must have at least one candidate or g-candidate of the $gS_p$-subset that is not in $TS_1$ or $TS_2$ (if it has several, choose one arbitrarily and name it <$V_2$, $c_2$>, … <$V_p$, $c_p$>, respectively). Otherwise, the $gS_p$-subset would be degenerated; more precisely, Z could be eliminated by a whip[1] (or even by ECP after a Single) associated with (any of) the CSP variable(s) that has no such candidate.

Secondly, in $TS_1$ or $TS_2$, there must be at least one candidate for at least one of the CSP variables $V_2, \ldots V_p$. Otherwise, the initial $gS_p$-subset would be degenerated; more precisely, it would contain, among others, the $gS_{p-2}$-subset $\{\{V_3, \ldots, V_p\}, \{TS_3, \ldots, TS_p\}\}$; this would allow to eliminate all the candidates for $V_1$ and $V_2$ that are not in $TS_1$ or $TS_2$; Z could then be eliminated by a whip[1] associated with $V_2$; and $V_1$ would have no candidate left. One can always suppose that there exists such a candidate $L_2$ for $V_2$, i.e. $L_2 = <V_2, l_2>$.

Modulo Z, we therefore have a $gS_{p-2}$ subset $R_2$ with CSP variables $V_2, \ldots V_{p-1}$ and g-transversal sets $TS_3, \ldots TS_p$. Let $c_p{*}$ be $c_p$ if it is a candidate or any element in $c_p$ if it is a g-candidate. Then, Z is a (normal) target of the following $gS_{p-2}$-z-whip of length p-1: $gS_{p-2}$-z-whip[p-1]: $V_2\{l_2\ R_2\} - V_p\{c_p\ .\} \Rightarrow \neg candidate(Z)$. qed.

Theorem 10.16 allows to replace any elimination of a candidate Z as a type-2 target for a $gS_p$-subset by the elimination of Z as a normal target for a $gS_{p-2}$-z-whip[p-1]. But, as was the case for $S_p$-subsets in section 9.3.6 and for similar reasons, this is not enough to guarantee that type-2 targets of $gS_p$-subsets, if allowed to be used as left-linking candidates in the definitions of $gS_p$-whips or $gS_p$-braids, could not lead to (slightly) more general patterns than those in our current definitions, due to the (probably rare) cases similar to those evoked in section 9.3.6. However, in the present case, one can prove the following:

***Theorem 10.17: for any $1 \leq p \leq \infty$, for any n>2, if a $RgS_pC[n]$ (respectively a $gS_pW[n]$, a $gS_pB[n]$) has a left-linking candidate $L_k$ that is a type-2 target of an inner $gS_p$-subset, then it can be seen as a normal (i.e. with no inner type-2 targets) $RgS_qC[n]$ (resp. $gS_qW[n]$, $gS_qB[n]$), for some q>p, i.e. with larger inner g-Subsets.***

Proof: almost obvious. Every time a left-linking candidate appears as a type-2 target, it suffices to merge its g-Subset with the next pattern in the sequence. Notice that this would not work for a "non-g" version of this theorem, because, even in this case, the next pattern could be a g-candidate.

Unfortunately, g-Subsets obtained by this (rather artificial) method tend to be very close to degeneracy.

## 10.3. A detailed example

We shall use the puzzle in Figure 10.1 (taken from the examples that go with the Hodoku solver [Hodoku www]) for several purposes:

– it will provide an example of a $gS_5$-subset and illustrate that, in conformance with our definition, the g-transversal sets do not have to meet all its CSP variables;

– it will illustrate the application of theorem 10.16 to the type-2 targets of a $gS_5$-subset;

– it will provide an example of a Reversible-gS$_2$-chain;

– it will illustrate alternative solutions using either gS$_5$-subsets and Reversible Chains or g-whips[5].

|   |   |   | 6 | 1 |   | 4 | 7 |   |
|---|---|---|---|---|---|---|---|---|
|   |   |   |   | 4 | 8 |   | 5 | 2 |
|   |   |   |   |   |   |   |   |   |
|   | 1 | 5 |   |   | 2 |   |   | 6 |
|   | 7 | 4 |   | 5 |   |   | 1 | 8 |
|   |   |   |   |   |   |   |   |   |
|   | 2 | 8 |   | 7 | 5 |   |   | 4 |
|   | 5 | 6 |   | 3 | 4 |   | 2 | 1 |

| 8 | 4 | 2 | 5 | 9 | 7 | 1 | 6 | 3 |
|---|---|---|---|---|---|---|---|---|
| 5 | 9 | 3 | 2 | 6 | 1 | 8 | 4 | 7 |
| 7 | 6 | 1 | 3 | 4 | 8 | 9 | 5 | 2 |
| 6 | 8 | 9 | 4 | 1 | 3 | 2 | 7 | 5 |
| 3 | 1 | 5 | 7 | 8 | 2 | 4 | 9 | 6 |
| 2 | 7 | 4 | 6 | 5 | 9 | 3 | 1 | 8 |
| 4 | 3 | 7 | 1 | 2 | 6 | 5 | 8 | 9 |
| 1 | 2 | 8 | 9 | 7 | 5 | 6 | 3 | 4 |
| 9 | 5 | 6 | 8 | 3 | 4 | 7 | 2 | 1 |

***Figure 10.1.** A puzzle P with W(P) = 4, gW(P) = 5, gSW(P) = 5*

### 10.3.1. Solution using only gS$_p$-subsets and Reversible-gS$_p$-chains

Let us first see what is obtained if we use the Hodoku software mentioned in section 10.1.7, when only basic rules plus xy-chains, Subsets, Finned-Fish, Franken Fish, Mutant Fish and Kraken Fish (a kind of Fish Chains to be discussed below) are activated. We keep Hodoku's self-explaining notation. In the first three patterns, the Finned Swordfish, the various "f" indicate the fins. "Finned Swordfish" is a classical variant of Swordfish with additional candidates linked to the target; in our view, it is merely a "z-Swordfish" (or z-SHT); the eliminations allowed here by the three instances of this pattern can also be done by g-whips[3] (see section 10.3.2).

***** Hodoku 2.0.1 *****
Finned Swordfish: 3 c239 r147 fr2c2 fr2c3 fr3c2 fr3c3 => r1c1≠3
Finned Swordfish: 9 r239 c147 fr2c2 fr2c3 fr3c2 fr3c3 => r1c1≠9
Finned Swordfish: 9 c569 r147 fr5c5 fr6c6 => r4c4≠9
;;; Resolution state RS$_1$

Now, Hodoku reaches a resolution state RS$_1$ (displayed in Figure 10.2) with a "Franken Squirmbag in columns" for Number 9: in the five Columns c2, c3, c5, c6 and c9 (in light grey), Number 9 appears only in Rows r1, r4 and r7 and in Blocks b1 and b5 (in dark grey).

**Franken Squirmbag: 9 c23569 r147b15 => r1c23478,r2347c1,r4c5678,r567c4,r7c78≠9**

In the approach of this chapter, this is a gS$_5$-subset: the five CSP variables are $X_{c2n9}$, $X_{c3n9}$, $X_{c5n9}$, $X_{c6n9}$ and $X_{c9n9}$ (symbolised by light grey columns); the five g-transversal sets are defined by CSP variables (considered as constraints) $X_{r1n9}$, $X_{r4n9}$, $X_{r7n9}$, $X_{b1n9}$ and $X_{b5n9}$ (symbolised by three dark grey rows and two dark grey blocks). The targets of the $gS_5$-subset are all the candidates (the fourteen ones in bold underlined characters in Figure 10.1) $S_5$-linked to one of the transversal sets, i.e. all the Numbers 9 in r1, r4, r7, b1 or b5 but not in any of c2, c3, c5, c6 or c9.

|    | c1 | c2 | c3 | c4 | c5 | c6 | c7 | c8 | c9 |    |
|----|----|----|----|----|----|----|----|----|----|----|
| r1 | n1 n2 n4 n5 n6 n7 n8 | n3 n4 n6 n8 **n9** | n1 n2 n3 n7 **n9** | n2 n3 n5 n7 **n9** | n2 n9 | n3 n7 n9 | n1 n3 n6 n8 **n9** | n3 n6 n8 **n9** | n3 n9 | r1 |
| r2 | n2 n3 n5 n8 **n9** | n3 n8 n9 | n2 n3 n9 | n2 n3 n5 n9 | n6 | n1 | n3 n8 n9 | n4 | n7 | r2 |
| r3 | n1 n3 n6 n7 **n9** | n3 n6 n9 | n1 n3 n7 n9 | n3 n7 n9 | n4 | n8 | n1 n3 n6 n9 | n5 | n2 | r3 |
| r4 | n2 n3 n6 n8 **n9** | n3 n6 n8 n9 | n2 n3 n9 | n1 n3 n4 n6 n7 n8 | n1 n8 n9 | n3 n7 n9 | n2 n3 n7 **n9** | n3 n7 **n9** | n3 n5 n9 | r4 |
| r5 | n3 n8 n9 | n1 | n5 | n3 n4 n7 n8 **n9** | n8 n9 | n2 | n3 n4 n7 n9 | n3 n7 n9 | n6 | r5 |
| r6 | n2 n3 n6 n9 | n7 | n4 | n3 n6 **n9** | n5 | n3 n6 n9 | n2 n3 n9 | n1 | n8 | r6 |
| r7 | n1 n3 n4 n7 **n9** | n3 n4 n9 | n1 n3 n7 n9 | n1 n2 n8 **n9** | n1 n2 n8 n9 | n6 n9 | n3 n5 n6 n7 n8 **n9** | n3 n6 n7 n8 **n9** | n3 n5 n9 | r7 |
| r8 | n1 n3 n9 | n2 | n8 | n1 n6 n9 | n7 | n5 | n3 n6 n9 | n3 n6 n9 | n4 | r8 |
| r9 | n7 n9 | n5 | n6 | n8 n9 | n3 | n4 | n7 n8 n9 | n2 | n1 | r9 |
|    | c1 | c2 | c3 | c4 | c5 | c6 | c7 | c8 | c9 |    |

***Figure 10.2.*** *Resolution state $RS_1$ of P, in which there appears a Franken Squirmbag*

At first sight, this Franken Squirmbag leads to a very impressive result: eighteen eliminations done by a single pattern. Notice that, contrary to whips that generally eliminate only one candidate at a time (though associated whips obtained by permutations can often eliminate more candidates), a Subset often eliminates several candidates; but eighteen is really exceptional.

However, a closer look shows some difference in the eliminations done by Hodoku for its Franken Squirmbag and those done by our $gS_5$-subset: in addition to the fourteen candidates of the latter, the former eliminates the following four candidates (in underlined but not bold characters): n9r1c2, n9r1c3, n9r4c5 and n9r4c6. These are examples of the type-2 targets evoked in section 10.2.2.3. We shall take advantage of them to illustrate how, according to theorem 10.8, they could be eliminated by shorter gS-z-whips with smaller inner g-Subsets. Here, we have p=5, so we find $gS_3$-z-whip[4]:

gS$_3$-z-whip[4]: n9{r1c5 gS$_3$:{c5 c6 c9}{r4 r7 b5}} – c3n9{r4 .} ==> r1c2 ≠ 9
gS$_3$-z-whip[4]: n9{r1c5 gS$_3$:{c5 c6 c9}{r4 r7 b5}} – c2n9{r4 .} ==> r1c3 ≠ 9
gS$_3$-z-whip[4]: n9{r4c5 gS$_3$:{c2 c3 c9}{r1 r7 b1}} – c6n9{r1 .} ==> r4c5 ≠ 9
gS$_3$-z-whip[4]: n9{r4c5 gS$_3$:{c2 c3 c9}{r1 r7 b1}} – c6n9{r1 .} ==> r4c6 ≠ 9

The end of the Hodoku resolution path has nothing noticeable:

XYZ-Wing: 7/6/3 in r4c68,r6c4 => r4c4≠3 (a particular kind of z-whip[2])
XYZ-Wing: 9/3/6 in r6c46,r7c6 => r4c6≠6 (a particular kind of z-whip[2])
Naked Pair: 3,7 in r4c68 => r4c12379≠3, r4c47≠7
Locked Candidates Type 1 (Pointing): 3 in b4 => r2378c1≠3 (i.e. whip[1])
Locked Candidates Type 1 (Pointing): 3 in b7 => r7c789≠3 (i.e. whip[1])
Hidden Single: r1c9=3
Locked Candidates Type 1 (Pointing): 3 in b2 => r56c4≠3 (i.e. whip[1])
Singles: r6c4=6, r7c6=6
Locked Candidates Type 1 (Pointing): 9 in b3 => r5689c7≠9 (i.e. whip[1])
Locked Candidates Type 2 (Claiming): 9 in r1 => r23c4≠9 (i.e. whip[1])
Locked Candidates Type 2 (Claiming): 9 in c4 => r7c5≠9 (i.e. whip[1])
Naked Pair: 7,8 in r7c8,r9c7 => r7c7≠7, r7c7≠8
Singles: r7c7=5, r7c9=9, r4c9=5, r5c8=9, r5c5=8, r5c1=3, r4c5=1, r4c4=4, r5c4=7, r5c7=4, r3c4=3, r4c6=3, r6c6=9, r1c6=7, r6c1=2, r6c7=3, r4c3=9, r4c7=2, r4c8=7, r7c5=2, r1c5=9, r7c8=8, r1c8=6, r8c8=3, r7c4=1, r8c4=9, r9c4=8, r8c1=1, r8c7=6, r9c7=7, r9c1=9
Naked Triple: 4,5,8 in r1c12,r2c1 => r2c2≠8
XY-Chain: 5 5- r1c4 -2- r1c3 -1- r3c3 -7- r3c1 -6- r4c1 -8- r2c1 -5 => r1c1,r2c4 ≠ 5
Singles to the end

### 10.3.2. Solution using only g-whips

The resolution path with whips has nothing noticeable; it gives W(P) = 9. We shall skip it. But the path with g-whips gives gW(P) = 5. The "SQ" comment at the end of a line indicates that the elimination is one available with the Franken Squirmbag; "SQ2" indicates that it is a type-2 target.

As can be seen, most of the Squirmbag eliminations can be done by shorter g-whips or even shorter whips. The "<<<<" comment indicates a whip[4] elimination not available to the Franken Squirmbag but that can be done before it.

***** SudoRules 16.2 based on CSP-Rules 1.2, config: gW *****
28 givens, 201 candidates, 1425 csp-links and 1425 links. Initial density = 1.77
g-whip[3]: b2n9{r1c5 r123c4} – r9n9{c4 c7} – b3n9{r1c7 .} ==> r1c1 ≠ 9
g-whip[3]: b4n9{r4c3 r456c1} – r9n9{c1 c7} – r8n9{c7 .} ==> r4c4 ≠ 9
g-whip[3]: b4n3{r5c1 r4c123} – c9n3{r4 r7} – r8n3{c8 .} ==> r1c1 ≠ 3
;;; Resolution state RS$_1$
g-whip[3]: b8n9{r7c6 r789c4} – r2n9{c4 c123} – r3n9{c3 .} ==> r7c7 ≠ 9 ; SQ
**whip[4]: b2n3{r3c4 r1c6} – c9n3{r1 r7} – b7n3{r7c3 r8c1} – b4n3{r4c1 .} ==> r4c4 ≠ 3** ; <<<<
g-whip[4]: b4n9{r6c1 r4c123} – c9n9{r4 r1} – c6n9{r1 r6} – c5n9{r4 .} ==> r7c1 ≠ 9 ; SQ
g-whip[4]: b7n9{r8c1 r7c123} – c9n9{r7 r1} – c6n9{r1 r6} – c5n9{r4 .} ==> r4c1 ≠ 9 ; SQ
g-whip[4]: b2n9{r1c5 r123c4} – b8n9{r8c4 r7c456} – c2n9{r7 r4} – c9n9{r4 .} ==> r1c3 ≠ 9 ; SQ2

g-whip[4]: b2n9{r1c5 r123c4} – b8n9{r8c4 r7c456} – c3n9{r7 r4} – c9n9{r4 .} ==> r1c2 ≠ 9 ; SQ2
g-whip[4]: b4n9{r4c3 r456c1} – b7n9{r8c1 r7c123} – c6n9{r7 r1} – c9n9{r1 .} ==> r4c5 ≠ 9 ; SQ2
g-whip[4]: b4n9{r4c3 r456c1} – b7n9{r8c1 r7c123} – c5n9{r7 r1} – c9n9{r1 .} ==> r4c6 ≠ 9 ; SQ2
g-whip[4]: b3n9{r3c7 r1c789} – b2n9{r1c5 r123c4} – r9n9{c4 c1} – b4n9{r5c1 .} ==> r4c7 ≠ 9 ; SQ
g-whip[4]: b2n9{r3c4 r1c456} – b3n9{r1c8 r123c7} – r6n9{c7 c1} – r9n9{c1 .} ==> r5c4 ≠ 9 ; SQ
g-whip[4]: b7n9{r7c3 r789c1} – b4n9{r5c1 r4c123} – c9n9{r4 r1} – b2n9{r1c4 .} ==> r7c4 ≠ 9 ; SQ
g-whip[4]: b8n9{r8c4 r7c456} – c9n9{r7 r4} – c3n9{r4 r123} – c2n9{r3 .} ==> r1c4 ≠ 9 ; SQ
g-whip[5]: b2n9{r1c6 r123c4} – r9n9{c4 c1} – r8n9{c1 c8} – c9n9{r7 r4} – b4n9{r4c2 .} ==> r1c7 ≠ 9 ; SQ
g-whip[5]: b4n9{r4c3 r456c1} – b7n9{r8c1 r7c123} – c9n9{r7 r1} – c6n9{r1 r6} – c5n9{r5 .} ==> r4c8 ≠ 9 ; SQ
whip[3]: r5n4{c7 c4} – r5n7{c4 c8} – r4c8{n7 .} ==> r5c7 ≠ 3
whip[4]: r4c8{n3 n7} – r5n7{c8 c4} – r5n3{c4 c1} – b7n3{r8c1 .} ==> r7c8 ≠ 3
whip[5]: b5n4{r4c4 r5c4} – b5n7{r5c4 r4c6} – r4c8{n7 n3} – r5n3{c8 c1} – b4n8{r5c1 .} ==> r4c4 ≠ 8
whip[5]: c6n7{r1 r4} – r4c8{n7 n3} – c9n3{r4 r7} – r8n3{c8 c1} – b4n3{r4c1 .} ==> r1c6 ≠ 3
whip[1]: c6n3{r6 .} ==> r6c4 ≠ 3, r5c4 ≠ 3
whip[4]: r6c4{n9 n6} – b8n6{r8c4 r7c6} – c6n9{r7 r1} – b3n9{r1c9 .} ==> r6c7 ≠ 9
whip[5]: r4c5{n1 n8} – r5c5{n8 n9} – b6n9{r5c7 r4c9} – b6n5{r4c9 r4c7} – r4n4{c7 .} ==> r4c4 ≠ 1
hidden-single-in-a-block ==> r4c5 = 1
whip[1]: r4n8{c1 .} ==> r5c1 ≠ 8
whip[4]: b7n3{r7c3 r8c1} – r5c1{n3 n9} – b6n9{r5c8 r4c9} – r1c9{n9 .} ==> r7c9 ≠ 3
whip[5]: b1n4{r1c2 r1c1} – b1n5{r1c1 r2c1} – c1n8{r2 r4} – c1n2{r4 r6} – b4n6{r6c1 .} ==> r1c2 ≠ 6
whip[5]: b1n5{r2c1 r1c1} – c1n8{r1 r4} – c1n2{r4 r6} – r6c7{n2 n3} – r5n3{c8 .} ==> r2c1 ≠ 3
whip[5]: b6n4{r5c7 r4c7} – c7n5{r4 r7} – r7c9{n5 n9} – c8n9{r8 r1} – c5n9{r1 .} ==> r5c7 ≠ 9
g-whip[3]: b7n9{r7c3 r789c1} – b4n9{r5c1 r4c123} – b6n9{r4c9 .} ==> r7c8 ≠ 9
whip[5]: b6n9{r4c9 r5c8} – c5n9{r5 r7} – c6n9{r7 r6} – b5n3{r6c6 r4c6} – c9n3{r4 .} ==> r1c9 ≠ 9
naked-single ==> r1c9 = 3
whip[3]: c9n9{r7 r4} – r5n9{c8 c1} – b7n9{r9c1 .} ==> r7c5 ≠ 9
whip[3]: r6c4{n9 n6} – b8n6{r8c4 r7c6} – b8n9{r7c6 .} ==> r3c4 ≠ 9, r2c4 ≠ 9
whip[1]: b2n9{r1c5 .} ==> r1c8 ≠ 9
whip[1]: b3n9{r3c7 .} ==> r9c7 ≠ 9, r8c7 ≠ 9
whip[3]: r9n9{c4 c1} – r5n9{c1 c8} – r8n9{c8 .} ==> r6c4 ≠ 9 ; SQ
singles ==> r6c4 = 6, r7c6 = 6
whip[2]: r9c7{n7 n8} – r7c8{n8 .} ==> r7c7 ≠ 7
whip[2]: r9c7{n8 n7} – r7c8{n7 .} ==> r7c7 ≠ 8
whip[2]: r4c8{n3 n7} – r4c6{n7 .} ==> r4c1 ≠ 3, r4c2 ≠ 3, r4c3 ≠ 3
whip[1]: b4n3{r6c1 .} ==> r3c1 ≠ 3, r7c1 ≠ 3, r8c1 ≠ 3
whip[1]: r8n3{c8 .} ==> r7c7 ≠ 3
singles ==> r7c7 = 5, r7c9 = 9, r4c9 = 5, r5c8 = 9, r5c1 = 3, r5c5 = 8, r7c5 = 2, r1c5 = 9, r1c6 = 7, r3c4 = 3, r4c6 = 3, r6c6 = 9, r6c1 = 2, r4c3 = 9, r6c7 = 3, r8c7 = 6, r8c8 = 3, r4c8 = 7, r5c7 = 4, r4c7 = 2, r5c4 = 7, r7c8 = 8, r9c7 = 7, r9c1 = 9, r8c1 = 1, r8c4 = 9, r9c4 = 8, r1c8 = 6, r7c4 = 1, r4c4 = 4
whip[3]: r2c1{n8 n5} – r1c1{n5 n4} – r1c2{n4 .} ==> r2c2 ≠ 8
whip[4]: r2c7{n8 n9} – r2c2{n9 n3} – r7c2{n3 n4} – r1c2{n4 .} ==> r2c1 ≠ 8
singles to the end

# 11. $W_p$-whips, $B_p$-braids and the T&E(2) instances

In chapters 7, 9 and 10, we have extended the possibilities for right-linking elements of whips and braids from candidates to respectively g-candidates, Subsets and g-Subsets – whereas we always kept left-linking elements restricted to mere candidates. In the present chapter, we shall show that whips and braids themselves can be used as right-linking patterns. For each $1 \leq p \leq \infty$, we shall define two increasing sequences of resolution theories ($W_pW_n$ and $B_pB_n$, $0 \leq n \leq \infty$) and we shall associate with them two new ratings, $W_pW$ and $B_pB$.

We shall prove two main results for $B_p$-braids, similar to those proven for all our previous generalised braid theories: the confluence property of all the $B_pB_n$ resolution theories (providing the $B_pB$ ratings with all the good properties of the previous similar ratings) and a "T&E($B_p$) vs $B_p$-braids" theorem.

We shall also prove that there is a close relationship, given by the "T&E(2) vs B-braids" theorem, between B-braids and an iterated (depth 2) Trial-and-Error procedure. As very fast programs can easily be written for T&E(2), this theorem provides an easy way of checking if an instance of a CSP can be solved by B-braids, without actually finding explicitly its B-braids resolution path and its BB rating. A practical consequence for Sudoku is that, as all the known minimal puzzles can be solved by T&E(2), they all have a finite BB rating.

## 11.1. $W_p$-labels and $B_p$-labels; $W_p$-whips and $B_p$-braids

### 11.1.1. $W_p$-labels and $B_p$-labels; $W_p$-links and $B_p$-links

When one wants to allow a pattern P as a right-linking object of a whip or a braid, the first step is to explicit the P-label underlying its definition, independently of any resolution state. The following definition of a W-label extracts from the definition of a whip its structural part: only the part, but all the part, that does not depend on the resolution state, i.e. that can be expressed with labels and links, without referring to actual candidates.

Definition: for any $n \geq 1$, a $W_n$-label is a structured list $(Z, (V_1, L_1, R_1), …, (V_{n-1}, L_{n-1}, R_{n-1}), (V_n, L_n))$, such that:
– for any $1 \leq k \leq n$, $V_k$ is a CSP variable;

– Z, all the $L_k$'s and all the $R_k$'s are labels;

– in the sequence of labels $(L_1, R_1, …, L_{n-1}, R_{n-1}, L_n)$, any two consecutive elements are different;

– Z does not belong to $\{L_1, R_1, L_2, R_2, …. L_n\}$;

– $L_1$ is linked to Z;

– right-to-left continuity: for any $1<k≤n$, $L_k$ is linked to $R_{k-1}$;

– strong left-to-right continuity: for any $1≤k<n$, $L_k$ and $R_k$ are labels for $V_k$;

– $L_n$ is a label for $V_n$;

– Z is not a label for $V_n$.

Definition: a $B_n$-label is a structured sequence as above, with the right-to-left continuity condition replaced by:

– for any $1<k≤n$, $L_k$ is linked to Z or to a previous $R_i$.

Definitions: a label l is $W_n$-linked [respectively $B_n$-linked] to a $W_n$-label [resp. a $B_n$-label] $(Z, (V_1, L_1, R_1), …, (V_{n-1}, L_{n-1}, R_{n-1}), (V_n, L_n))$ if l is equal to Z. The index n in "$W_n$-linked" or "$B_n$-linked" may be dropped, as there can be no ambiguity. A label l is compatible with the above $W_n$-label [resp. $B_n$-label] if it is not $W_n$-linked [resp. $B_n$-linked] to it.

One can now give an alternative equivalent definition of a whip [or a braid], in which the structural and non-structural conditions are completely separated:

Definition: in a resolution state RS, given a candidate Z (which will be the target), a whip [respectively a braid] of length n ($n ≥ 1$) built on Z is a $W_n$-label [resp. a $B_n$-label] $(Z, (V_1, L_1, R_1), …, (V_{n-1}, L_{n-1}, R_{n-1}), (V_n, L_n))$, such that:

– all the $L_k$'s and all the $R_k$'s are candidates (not only labels);

– for any $1 ≤ k < n$, $R_k$ is the only candidate for $V_k$ compatible with Z and with all the previous right-linking candidates (i.e. with Z and with all the $R_i$, $1 ≤ i < k$);

– $V_n$ has no candidate compatible with Z and with all the previous right-linking candidates (but $V_n$ has more than one candidate – a non-degeneracy condition).

## 11.1.2. Equivalence of whips or braids

Until now, we have been very strict on the targets of whips [or braids]: a whip [or a braid] has only one target, specified in its definition. But, sometimes there is another whip [braid] with an underlying $W_n$-label [$B_n$-label] strongly equivalent (definition below) to that of the first whip [braid] and allowing to eliminate its own target. This entailed no problem until now, because the second whip [braid] could be written after the first and it did not change the W [B] rating of an instance. But if a whip [braid] is to be inserted into another one as a right-linking pattern, then it should not be counted several times if it serves to justify several t-candidates. The following definitions palliate this problem. Notice that, in the manual editing of all our previous resolution paths, we have implicitly used them, every time two eliminations appeared in the same line.

Definition (structural): two $W_n$-labels [$B_n$-labels] are strongly equivalent if they differ only by their targets. This is obviously an equivalence relation.

Definition (non-structural): in a resolution state RS, two whips [braids] of length n are strongly equivalent if their underlying $W_n$-labels [$B_n$-labels] are strongly equivalent. This is an equivalence relation.

Remarks about strongly equivalent whips [resp. braids]:

– the definition entails that the whips [resp. braids] have the same t-candidates;

– it also supposes that, for each CSP variable of the common $W_n$- [resp. $B_n$-] label, every candidate that is not a left-linking, t- or right- linking candidate must be a z-candidate for both whips [braids] simultaneously (i.e. it is linked to their two different targets);

– due to the second remark, there is no simple way of replacing this definition by a purely structural one; but if it is satisfied in a resolution state RS, then it will be satisfied in any posterior state in which both whips [braids] are still defined; we say that it is *persistent*, which, for some purposes, is almost as good as being structural;

– having no z-candidates, as in t-whips [t-braids], gives rise to strongly equivalent whips [braids]; but this is not a necessary condition;

– a $W_n$-label [$B_n$-label] can be interpreted as a potential whip [braid], waiting for the elimination of some candidates from its CSP variables before it becomes an actual one.

Definition (non-structural): an *extended target of a whip W in a resolution state RS* is a target of any whip strongly equivalent to W in RS.

Remarks:

– there is an obvious correspondence between a $W_1$-label and the set consisting of a g-label and its targets (seen as whip[1] targets); and a label l is g-linked to a g-label g if and only if l is an extended target of the $W_1$-label corresponding to g;

– if Z' is an extended target of a whip W in a resolution state RS, then it remains one in any posterior resolution state in which W (or a strongly equivalent whip) is still present and Z' is still a candidate; being an extended target is a persistent property;

– however, a candidate Z' that is not an extended target for W in RS may become one in a posterior resolution state RS', in case all the z-candidates of W in RS that are not t-candidates of W and that are not linked to Z', have been eliminated along the path between RS and RS'.

Definition (non-structural): a candidate C is compatible with a whip W in a resolution state RS if it is not an extended target of W in RS. Transposing the above remarks: if C is incompatible with a whip W in RS, then it remains incompatible with W in any posterior resolution state in which W is still present and C is still a candidate; but if C is compatible with W in RS, it may become incompatible with W in a posterior resolution state. It is therefore necessary to be always clear about the resolution state under consideration. Said otherwise, incompatibility with a whip is persistent, compatibility is not.

### 11.1.3. Definition of $W_p$-whips, $W_p$-braids and $B_p$-braids

Special care must be taken with the definition of whips accepting inner whips as right-linking patterns:

– global variables of the global whip and inner variables of each of its inner sub-whips must not be confused;

– similarly, global and inner left-to-right linking conditions must not be confused;

– for a proper definition of the global size, the conditions must be written in a form that does not allow degeneracy of the inner whips; fortunately, this is much easier to do than for inner Subsets: one only has to make sure that contradictions in the inner whips can only occur on their last CSP variables (i.e. not before);

– it must not be forgotten that, as is always the case for all the inner patterns of generalised whips, inner whips will appear as "reversed" whips (modulo the target and the previous right-linking objects), in the sense that their targets will have to appear as the next left-linking candidate.

Definition: in any resolution state RS of any CSP, for any $n \geq 1$ and $1 \leq p < n$, a $W_p$-*whip[n]* is a structured list $(Z, (V_1, L_1, R_1, q_1), \ldots, (V_{m-1}, L_{m-1}, R_{m-1}, q_{m-1}), (V_m, L_m, q_m))$, with $m \leq n$, that satisfies the following structural and non-structural conditions:

structural conditions (that could be considered as defining a "$W_p$-regular sequence of length n"):

– all the $q_k$'s are integers; $1 \leq q_k \leq p$ for all $1 \leq k \leq m$, $q_m = 1$ and $n = \sum_{1 \leq k \leq m} q_k$;

– for any $1 \leq k \leq m$, $V_k$ is a CSP variable;

– for each $1 \leq k \leq m$, $L_k$ is a label for $V_k$;

– for each $1 \leq k < m$, $R_k$ is a label or a $W_1$-label if $q_k = 1$ and it is a $Wq_k$-label if $q_k > 1$;

– $L_1$ is linked to Z;

– right-to-left continuity: for any $1 \leq k \leq m$, $L_k$ is linked or $W_{k-1}$-linked to $R_{k-1}$;

– for any $1 \leq k < m$, the following "strong continuity or strong W-continuity from $L_k$ to $R_k$", implying *"continuity or W-continuity from $L_k$ to $R_k$"*, is satisfied:

   - if $q_k=1$ and $R_k$ is a label, then $R_k$ (as well as $L_k$) is a label for $V_k$;

   - if $q_k\geq 1$ and $R_k$ is a $Wq_k$-label, then $V_k$ is one of its CSP variables (it does not have to be the last one – see the comments after Figure 11.4);

  – Z is not a label for $V_m$;

non-structural conditions:

  – Z and all the $L_k$'s are candidates (not only labels);

  – for any $1\leq k<m$: if $q_k=1$ and $R_k$ is a label, then, $R_k$ is the only candidate for $V_k$ compatible *in RS* with Z and with all the previous right-linking patterns $R_i$; if $q_k\geq 1$ and $R_k$ is a $Wq_k$-label $(Z_k, (V_{k,1}, L_{k,1}, R_{k,1}), …, (V_{k,i}, L_{k,i}, R_{k,i}), …, (V_{k,qk-1}, L_{k,qk-1}, R_{k,qk-1}), (V_{k,qk}, L_{k,qk}))$, then:

   - for each $i<q_k$: $L_{k,i}$ and $R_{k,i}$ are candidates (not only labels) for CSP variable $V_{k,i}$ of $R_k$;

   - for each $i<q_k$: $R_{k,i}$ is the only candidate for $V_{k,i}$ compatible *in RS* with Z, with the previous right-linking patterns $R_i$ ($i<k$) of the global $W_p$-whip[n] being defined, with the previous right-linking candidates $R_{k,i'}$ ($i'<i$) inside $R_k$, *and with $Z_k$*;

   - $L_{k,qk}$ is a candidate for $V_{k,qk}$ (not only a label); $V_{k,qk}$ has no candidate compatible *in RS* with Z, with the previous right-linking patterns $R_i$ ($i<k$) of the global $W_p$-whip[n] being defined, with the previous right-linking candidates $R_{k,i}$ ($i<q_k$) inside $R_k$ *and with $Z_k$* (but $V_{k,qk}$ has more than one candidate compatible *in RS* with Z and with the previous right-linking objects $R_i$ ($i<k$) of the global $W_p$-whip – this is the non-degeneracy condition of the inner $R_k$ whip);

  – $V_m$ has no candidate compatible *in RS* with the target and with all the previous right-linking objects of the global $W_p$-whip (but $V_m$ has more than one candidate – the usual non-degeneracy condition of the global $W_p$-whip).

  Remark: for all n, a $W_1$-whip[n] is the same thing as a g-whip[n].

  Definition: for any $n\geq 1$ and $1\leq p<n$, a *$W_p$-braid[n]* is a structured list as above, with the structural right-to-left continuity condition of a $W_p$-whip[n] replaced by:

  – for any $1\leq k\leq m$, $L_k$ is linked or W-linked to Z or to a previous $R_i$.

  Definition: in the previous definition, if the inner $Wq_k$-labels are replaced by $Bq_k$-labels, one obtains $B_p$-braids[n].

  Definitions: in any of the above defined $W_p$-whips or $B_p$-braids, a candidate other than $L_k$ for any of the "global" CSP variables $V_k$ is called a global t- [respectively global z-] candidate if it is incompatible with a previous right-linking pattern [resp. with the target Z]; a candidate for a "local" or "inner" CSP variable $V_{k,i}$ of an inner braid $R_k$ is called a local (or inner) t- [respectively local (or inner) z-] candidate if it is incompatible with a previous local right-linking candidate $R_{k,j}$, $j<i$ [resp. with the local target $Z_k$ of $R_k$].

Notice that a candidate can be at the same time global and local, z- and t-. Notice also that, as in all our previous definitions, the (global or local) z- and t- candidates are not considered as being part of the $W_p$-whip or $B_p$-braid patterns.

Remarks:

– in the above definitions, as in any of the previously defined types of generalised whips or braids, left-linking elements of the global $W_p$-whip [$B_p$-braid] are mere candidates (and not more general patterns);

– as shown by the fact that inner whips or braids are "reversed" (see Figure 11.2 or the proof of the $W_p$-whip elimination theorem), the acceptance of whips[p] or braids[p] as right-linking patterns amounts to accepting some form of look-ahead of size p (a form different, globally less restricted than that accepted in $S_p$-whips or $S_p$-braids);

– in the same way as all the types of braids we have met before, $B_p$-braids[n] are interesting for the confluence property of the $B_pB_n$ theories and for the "T&E($B_p$) vs $B_pB$" theorem (see proofs below); and $W_p$-whips are interesting as a simpler (and hopefully good) approximation of $B_p$-braids;

– one could also define $B_p$-whips[n]; *a priori*, there does not seem to be any good reason for imposing an "outer" continuity condition if the inner bricks do not enjoy their own inner continuity, but it may be useful as an approximation tool.

### *11.1.4. Graphico-symbolic representations*

The symbolic representations in Figures 11.1 and 11.2 may help understand how a partial $W_2$-whip[3] differs from an ordinary partial whip[3]. In these Figures:

– black horizontal lines represent CSP variables; they are supposed to have candidates only at their extremities or at their meeting points with arrows;

– dark grey straight oblique arrows represent links from Z to $L_1$ or from $R_k$ to $L_{k+1}$ and also, in the second Figure, inner links from $R_{i,k}$ to $L_{i,k+1}$;

– light grey arrows represent links to z- or t- candidates in the global whip and (in the second Figure) in an inner whip (the straight ones represent links to candidates in the same g-label as the next left-linking candidate);

– the straight double-sided dark grey arrow in the second Figure represents the double role of $L_2$ as a target of the inner whip (descending arrow) and as the next left-linking candidate (ascending arrow);

– the orientations of arrows represent the way links are used in the proof of the whip or $W_2$-whip rule; by themselves, links are not orientated; but these orientations also illustrate the idea that inner whips correspond to some form of look-ahead.

11. $W_p$-whips, $B_p$-braids and the T&E(2) instances       295

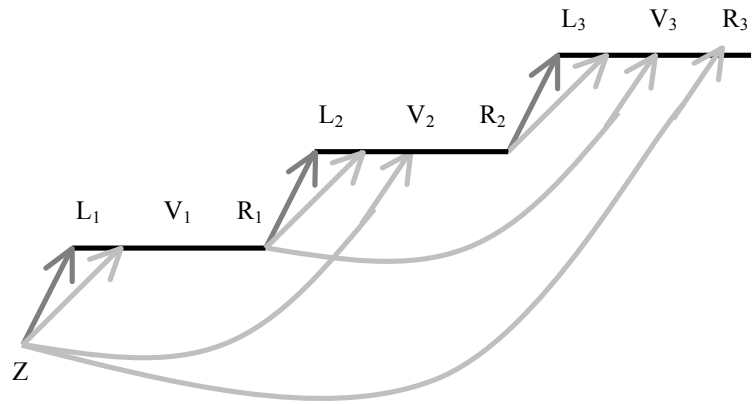

*Figure 11.1. A graphico-symbolic representation of a partial whip[3]*

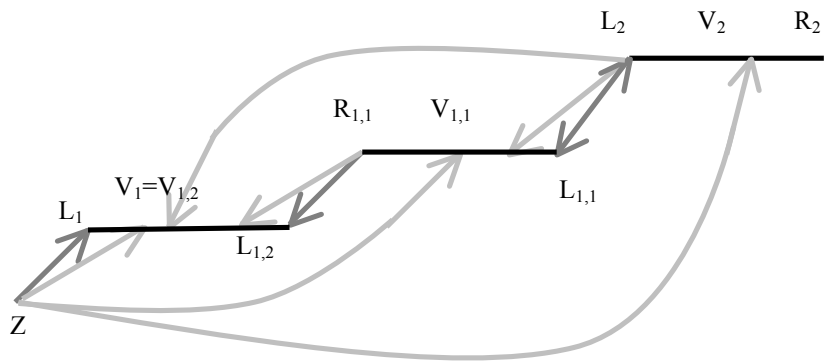

*Figure 11.2. A graphico-symbolic representation of a partial $W_2$-whip[3]. One can see an inner whip[2] modulo Z, with target $L_2$: ($L_2$, ($V_{1,1}$, $L_{1,1}$, $R_{1,1}$) ($V_{1,2}$, $L_{1,2}$)).*

*11.1.5. Elimination theorems*

***Theorem 11.1 ($W_p$-whip elimination theorem): given a $W_p$-whip, one can eliminate its target.***

Proof: obvious. The main point was having the correct definitions. If Z was True, then $L_1$ and all the candidates linked to Z (the global z-candidates) would be eliminated by ECP; if $R_1$ is a label, then it would be asserted by S; if $R_1$ is a $Wq_1$-label, then, after these first series of eliminations, it would be a whip[$q_1$] with target $L_2$ and $L_2$ would be eliminated by rule $Wq_1$. We can iterate until we reach $L_m$ would be eliminated by ECP or by rule $Wq_{m-1}$. (As usual, the global t-candidates would be progressively eliminated by ECP or some $Wq_k$). The last condition implies that $V_m$ would have no possible value.

***Theorem 11.2 ($B_p$-braid elimination theorem): given a $B_p$-braid, one can eliminate its target.***

Proof: almost the same as the proof for $W_p$-whips (with any reference to $Wq_k$ replaced by one to $Bq_k$), the main difference being the condition replacing right-to-left continuity, which still implies that $L_k$ would be eliminated by ECP.

*11.1.6. $W_p$-whip and $B_p$-braid resolution theories; the $W_pW$ and $B_pB$ ratings*

For each integer p with $1 \leq p \leq \infty$, one can define an increasing sequence ($W_pW_n$, n≥0) of resolution theories based on $W_p$-whips:
- $W_pW_0$ = BRT(CSP),
- $W_pW_1 = W_pW_0 \cup$ {rules for $W_p$-whips of length 1} = $W_1$,
- …
- $W_pW_n = W_pW_{n-1} \cup$ {rules for $W_p$-whips of length n},
- $W_pW_\infty = \cup_{n \geq 0} W_pW_n$.

One has obvious similar definitions for ($B_pB_n$, n≥0).

And, for each $1 \leq p \leq \infty$, one can also define in the usual way the $W_pW$ [respectively $B_pB$] rating associated with the increasing sequence ($W_pW_n$, n≥0) [resp. $B_pB_n$, n≥0] of resolution theories. It is obvious that, for any instance Q, $W_pW(Q)$ [resp. $B_pB(Q)$], considered as a function of p, is non-increasing.

One can also define the WW and BB ratings as being equal to $W_\infty W$ and $B_\infty B$, respectively, when no restriction is put *a priori* on the lengths of the inner whips [resp. braids] (of course, in each W-whip [resp. B-braid], they can only be smaller than its global length).

Remarks:

– it was important to properly define the length of a W$_p$-whip [or B$_p$-braid] in a way that takes into account the lengths of all its elements, because some W$_p$-whips [or B$_p$-braids] may be equivalent to (g)S$_p$-whips [or (g)S$_p$-braids]; for consistency of the ratings, they must be given the same size, whichever way they are considered;

– with the confluence property of all the B$_p$B$_n$ resolution theories (see section 11.2), the B$_p$B ratings have the same good properties as those mentioned for previous generalised braid theories; however, non-anticipativeness is no longer true; it is replaced by a restricted form of look-ahead, controlled by the maximum size p of the inner braids;

– as an obvious corollary to theorem 11.5 below, the BB rating is finite for any instance of a CSP that can be solved by T&E(2). ***In Sudoku, this entails that all the known minimal puzzles have a finite BB rating – a rating that is obviously invariant under symmetry and supersymmetry***.

### 11.1.7. gS$_p$W$_n$+W$_p$W$_n$ and gS$_p$W$_n$+B$_p$B$_n$ theories; associated ratings

Allowing gS$_p$-subsets or whips[p] as right-linking patterns in different whips, one can hope to get still more powerful resolution theories. For each $1 \leq p \leq \infty$, one can define an increasing sequence gS$_p$W$_n$+W$_p$W$_n$, $0 \leq n \leq \infty$, of resolution theories:
– gS$_p$W$_0$+W$_p$W$_0$ = BRT(CSP),
– gS$_p$W$_1$+W$_p$W$_1$ = gS$_p$W$_0$+W$_p$W$_0$ ∪ gS$_p$W$_1$ ∪ W$_p$W$_1$ = W$_1$,
– …
– gS$_p$W$_n$+W$_p$W$_n$ = gS$_p$W$_{n-1}$+W$_p$W$_{n-1}$ ∪ gS$_p$W$_n$ ∪ W$_p$W$_n$,
– …
– gS$_p$W$_\infty$+W$_p$W$_\infty$ = ∪$_{n \geq 0}$ gS$_p$W$_n$+W$_p$W$_n$.

One can introduce obvious similar definitions for gS$_p$B$_n$ +B$_p$B$_n$, $0 \leq n \leq \infty$.

And, for each $1 \leq p \leq \infty$, one can define in the usual way the gS$_p$W+W$_p$W [respectively gS$_p$B+W$_p$B] rating associated with the increasing sequence gS$_p$W$_n$+W$_p$W$_n$, n≥0, [resp. gS$_p$B$_n$+B$_p$B$_n$, n≥0] of resolution theories.

One can also define the gSW+WW and gSB+BB ratings in the usual way.

It is a straightforward corollary to lemma 4.1 and theorems 10.15 and 11.3 (below) that all the gS$_p$B$_n$+B$_p$B$_n$ resolution theories are stable for confluence and have the confluence property. A "simplest first" strategy can therefore be defined. Or rather several "simplest first" strategies: the question is, for each n, do we give precedence to gS$_p$-braids[n] or to B$_p$-braids[n]? These definitions leave us the freedom of choosing priorities between Subsets and whips. Moreover, the (probably limited) increased resolution power of these combined theories (with respect to the B$_p$-braids) is probably not worth its cost in terms of computational complexity.

### 11.1.8. $(gS_p+W_p)$-whip and $(gS_p+B_p)$-braid theories; associated ratings

Going one step further, one can allow both $gS_p$-subsets and whips[p] as right-linking patterns in the same whips, in the hope of getting the most powerful theories. For each $1 \leq p \leq \infty$, one can define an increasing sequence $(gS_p+W_p)W_n$, $0 \leq n \leq \infty$, of resolution theories:

- $(gS_p+W_p)W_0 = BRT(CSP)$,
- $(gS_p+W_p)W_1 = W_1$,
- …
- $(gS_p+W_p)W_n = (gS_p+W_p)W_{n-1} \cup$ {rules for whips of total length n, with inner $gS_p$-subsets and $W_p$-whips},
- …
- $(gS_p+W_p)W_\infty = \cup_{n \geq 0} (gS_p+W_p)W_n$.

One can introduce obvious similar definitions for $(gS_p+B_p)B_n$, $0 \leq n \leq \infty$.

And, for each $1 \leq p \leq \infty$, one can define in the usual way the $(gS_p+W_p)W$ [respectively $(gS_p+B_p)B$] rating associated with the increasing sequence $(gS_p+W_p)W_n$, $n \geq 0$, [resp. $(gS_p+W_p)B_n$, $n \geq 0$] of resolution theories.

One can also define the $(gS+W)W$ and $(gS+B)B$ ratings in the usual way.

Contrary to the previous case, the confluence property of the $(gS_p+B_p)B_n$ resolution theories must now be proven directly; this can be done by combining the proofs for the $gS_pB_n$ and the $B_pB_n$ theories (we leave it as an exercise for the reader). A "simplest first" strategy can therefore be defined, or rather several "simplest first" strategies, each providing all the $(gS_p+B_p)B$ ratings with good properties. But, as in the previous case, the (probably limited) increased resolution power is probably not worth the computational cost of so complex braids.

### 11.1.9. More graphico-symbolic representations

#### 11.1.9.1. Similarities between Subsets and whips

As suggested by the proof of confluence in the next section, there is a remarkable and deep similarity between Subsets and whips/braids of same size p. The definitions of both concepts involve p different CSP variables and p sets of candidates for these variables:

- for $S_p$-subsets: p transversal sets of candidates, defined by p fixed constraints;
- for whips/braids[p]: p sets consisting of candidates linked (by any constraint) to the target or to one of the previous right-linking candidates (a total of p also!).

These similarities can be represented symbolically in Figure 11.3 (for p = 4).

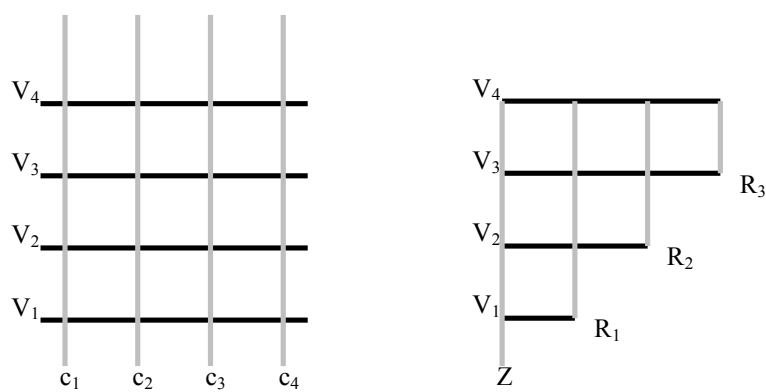

*Figure 11.3. A symbolic representation of the similarities between a Subset and a whip*

Horizontal black lines represent CSP variables in $\{V_1, V_2, V_3, V_4\}$. In a Subset (leftmost part of the Figure), each vertical grey line represents a fixed constraint in $\{c_1, c_2, c_3, c_4\}$. In a whip or a braid (rightmost part of the Figure), each of these lines represents the existence of a link (along *any* constraint) with the target or with a determined element in the sequence of p-1 right-linking candidates. In horizontal lines, candidates may exist only at the intersections with vertical lines; in the whip/braid case, an intersection may represent several candidates (in the same g-label for the corresponding CSP variable). In spite of their deep conceptual differences, the ideas represented by "vertical lines" can be used in much the same ways in several proofs, such as the confluence property and the "T&E($B_p$) vs $B_p$-braids" theorem.

For whips, the rightmost part of this Figure is an alternative view to that of Figure 11.1. The latter stressed the various links the target or a right-linking candidate can have with z- and/or t- candidates for various posterior CSP variables. The present view abstracts from these differences, considering that only the existence of a link is important. We insist that, contrary to the Subsets case in the leftmost part of the Figure and contrary to what these vertical lines may intuitively suggest, candidates in a vertical line do not have to be pairwise linked (and, in general, they are not).

*11.1.9.2. Another graphical representation of W-whips and B-braids*

Based on the similarities between Subsets and whips or braids and on Figure 11.3, another type of graphical representation for a W-whip can be given in Figure 11.4, maybe more readable than that in Figure 11.2. Here, the conventions are the same as in Figure 11.3: a horizontal black line represents a CSP variable, a vertical grey arrow represents the existence of a link (along *any* constraint) with the candidate (target or right-linking) at the origin of the arrow. It is supposed that, in the current resolution state, candidates for a CSP variable are present only at its intersection with some vertical arrow (and an intersection may represent several candidates in the same g-label for this CSP variable).

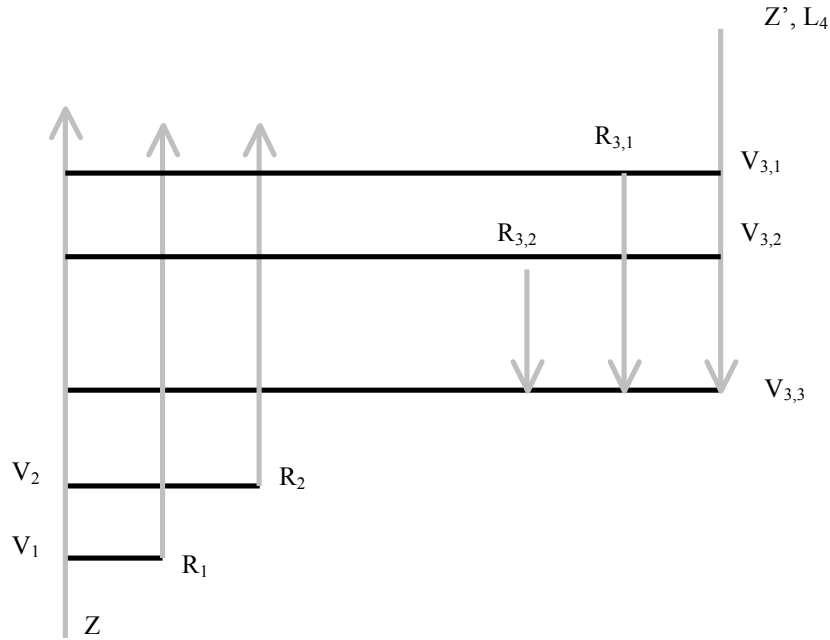

*Figure 11.4. A symbolic representation of a partial $W_3$-whip[5]*

Read from left to right, this example starts with a standard partial whip[2] with target Z, CSP variables $V_1$, $V_2$ and right-linking candidates $R_1$, $R_2$. Then, there appears an inner whip[3] with target Z'=$L_4$ (which will be the left-linking candidate for the next part of a larger global W-whip), inner CSP variables $V_{3,1}$, $V_{3,2}$ and $V_{3,3}$, and inner right-linking candidates $R_{3,1}$ and $R_{3,2}$. Here, CSP variable $V_{3,3}$ has no candidate compatible with Z, $R_1$, $R_2$, Z'=$L_4$, $R_{3,1}$ and $R_{3,2}$.

Notice that, if CSP variable $V_{3,3}$ had no candidate linked to Z, $R_1$ or $R_2$, the final contradiction in the inner whip would still occur in $V_{3,3}$, but $V_3$ could not be taken to be $V_{3,3}$. (This illustrates why, in our definition in section 11.1.3, $V_k$ does not have to be the last element of $W_k$).

## 11.2. The confluence property of the B$_p$B$_n$ resolution theories

We now prove the main property of B$_p$-braid resolution theories.

***Theorem 11.3: each of the B$_p$B$_n$ resolution theories ($1 \leq p \leq \infty$, $0 \leq n \leq \infty$) is stable for confluence; therefore, it has the confluence property.***

Proof: in order to keep the same notations as in the proofs for the g-braids (section 7.5) and the S$_r$-braids (section 9.4), we prove the result for B$_r$B$_n$, r and n fixed.

We must show that, if an elimination of a candidate Z could have been done in a resolution state RS$_1$ by a B$_r$-braid B of length n' $\leq$ n and with target Z, it will always still be possible, starting from any further state RS$_2$ obtained from RS$_1$ by consistency preserving assertions and eliminations, if we use a sequence of rules from B$_r$B$_n$. Let B be: $\{L_1\ R_1\} - \{L_2\ R_2\} - .... - \{L_p\ R_p\} - \{L_{p+1}\ R_{p+1}\} - ... - \{L_m\ .\}$, with target Z, where the R$_k$'s are candidates or braids in B$_r$ modulo Z and the previous R$_i$'s. For inner braids, we use the notations in the definition (section 11.1).

The proof follows the same general lines as that for g-braids and S$_r$-braids. Indeed, it is remarkably close to that for S$_r$-braids, with transversal sets replaced by the sets of candidates linked to some right-linking object (see the similarities in Figure 11.3 and discussion in section 11.2.2). For technical reasons, we keep a separate case for inner braids of length 1, i.e. g-whips.

Consider first the state RS$_3$ obtained from RS$_2$ by applying repeatedly the rules in BRT until quiescence. As BRT has the confluence property by theorem 5.6, this state is uniquely defined. (Notice that, thanks to theorem 5.6 and the inclusion B$_n$ ⊂ B$_r$B$_n$, we could use B$_n$ instead of BRT, but, apart from dispensing us of introducing marks, it does not seem to make the proof simpler.)

If, in RS$_3$, target Z has been eliminated, the proof is finished. If target Z has been asserted, then the instance of the CSP is contradictory; if not yet detected in RS$_3$, this contradiction can be detected by CD in a state posterior to RS$_3$, reached by a series of applications of rules from B$_r$, following the B$_r$-braid structure of B.

Otherwise, we must consider all the elementary events related to B that can have happened between RS$_1$ and RS$_3$, as well as those we must provoke in posterior resolution states RS. For this, we start from B' = what remains of B in RS$_3$ and we let RS = RS$_3$. At this point, B' may not be an S$_r$-braid in RS. We progressively update RS and B' by repeating the following procedure, for p = 1 to p = m, until it produces a new (possibly shorter) B$_r$-braid B' with target Z in resolution state RS – a situation that is bound to happen. We return from this procedure as soon as B' is a B$_r$-braid in RS. All the references below are to the current RS and B'.

a) If, in RS, any candidate that had negative valence in B – i.e. the left-linking candidate, or any t- or z- candidate, of CSP variable $V_p$, or any global or local t- or z- candidate of $R_p$ in case $R_p$ is an inner braid – has been asserted (this can only be between $RS_1$ and $RS_3$), then all the candidates linked to it have been eliminated by relevant rules from BRT in $RS_3$, in particular: Z and/or all the candidate(s) $R_k$ (k<p) to which it is linked, and/or all the elements of the g-candidate(s) $R_k$ (k<p) to which it is g-linked, and/or all the inner candidates to which it is linked of the inner $R_k$ braids (k<p) to which it is B-linked (by the definition of a $B_r$-braid); if Z is among them, there remains nothing to prove; otherwise, the procedure has already either been successfully terminated by case f1 or f2α and/or dealt with by case d2 of the previous such k's for which $R_k$ is an inner braid of length $q_k \geq 2$.

b) If, in $RS_3$, left-linking candidate $L_p$ has been eliminated (but not asserted), it can no longer be used as a left-linking candidate in a $B_r$-braid. Suppose that either CSP variable $V_p$ still has a z- or a t- candidate $C_p$, or that $R_p$ is an inner braid of length $q_p \geq 2$ and there is another CSP variable $V_p'$ in its $W_p$ sequence of CSP variables such that $V_p'$ still has a z- or a t- candidate $C_p$; then, in B', replace $L_p$ by $C_p$ and (in the latter case) $V_p$ by $V_p'$. Now, up to $C_p$, B' is a partial $B_r$-braid in RS with target Z. Notice that, even if $L_p$ was linked or g-linked or B-linked to $R_{p-1}$ (e.g. if B was a $B_r$-whip) this may not be the case for $C_p$; therefore trying to prove along the same lines a similar theorem for $B_r$-whips would fail here.

c) If, in RS, any t- or z- candidate of $V_p$ or of the inner braid $R_p$ (if $R_p$ is an inner braid) has been eliminated (but not asserted), this has not changed the basic structure of B (at stage p). Continue with the same B'.

d) Consider now assertions occurring in right-linking objects of the global $B_r$-braid. There are two cases instead of one for g-braids: assertions occurring in a right-linking candidate or g-candidate (case d1) and assertions occurring anywhere in an inner braid $R_p$ of length $q_p \geq 2$ (case d2).

d1) If, in RS, right-linking candidate $R_p$ or a candidate $R_p'$ in right-linking g-candidate $R_p$ has been asserted (p can therefore not be the last index of B'), $R_p$ can no longer be used as an element of a $B_r$-braid, because it is no longer a candidate or a g-candidate. As in the proof for $S_r$-braids, and only because of this d1 case, we cannot be sure that this assertion occurred in $RS_3$. We must palliate this. First eliminate by ECP or $W_1$ any left-linking or t- candidate for any CSP variable of B' after p, including those in the inner braids, that is incompatible with $R_p$, i.e. linked or g-linked to it, if it is still present in RS. Now, considering the $B_r$-braid structure of B upwards from p, more eliminations and assertions can be done by rules from $B_r$. (Notice that, as in the $S_r$-braids case, we are not trying to do more eliminations or assertions than needed to get a $B_r$-braid in RS; in particular, we continue to consider $R_p$, not $R_p$'; in any case, it will be excised from B'; but, most of all, we do not have to find the shortest possible $B_r$-braid!)

Let q be the smallest number strictly greater than p such that CSP variable $V_q$ or some CSP variable $V_q$' in $W_q$ still has a global left-linking, t- or z- candidate $C_q$ that is not linked, g-linked or B-linked to any of the $R_i$ for $p \leq i < q$. (For index q, there is thus a $V_q$' in $W_q$ and a candidate $C_q$ for $V_q$' such that $C_q$ is linked, g-linked or B-linked to Z or to some $R_i$ with i < p.)

Apply the following rules from $B_r$ (if they have not yet been applied between $RS_2$ and RS) for each of the CSP variables $V_u$ (and all the $V_{u,i}$ in $W_u$ if $R_u$ is an inner braid) with index (or first index) u increasing from p+1 to q-1 included:
- eliminate by ECP or $W_1$ or some $B_{r'}$ (r'≤r) any candidate for any CSP variable in $W_u$ that is incompatible with $R_{u-1}$;
- at this stage, CSP variable $V_u$ has no left-linking, z- or t- candidate and there remains no global t- or z- candidate in $W_u$ if $R_u$ is an inner braid;
- if $R_u$ is a candidate, assert it by S and eliminate by ECP all the candidates for CSP variables after u, including those in the inner braids, that are incompatible with $R_u$ in the current RS;
- if $R_u$ is a g-candidate, it cannot be asserted; eliminate by $W_1$ all the candidates for CSP variables after u, including those in the inner braids, that are incompatible with $R_u$ in the current RS;
- if $R_u$ is an inner braid in $Bq_u$, it cannot be asserted by $Bq_u$; eliminate by $Bq_u$ all the candidates for CSP variables after u, including those in the inner braids, that are incompatible with $R_u$ in the current RS (this includes the target of $R_u$).

In the new RS thus obtained, excise from B' the part related to CSP variables and inner braids p to q-1 (included); if $L_q$ has been eliminated in the passage from $RS_2$ to RS, replace it by $C_q$ (and, if necessary, replace $V_q$ by $V_q$'); for each integer $s \geq p$, decrease by q-p the index of CSP variable $V_s$, of its candidates and inner right-linking pattern (g-candidate or braid) and of the set $W_s$, in the new B'. In RS, B' is now, up to p (the ex q), a partial $B_r$-braid in $B_rB_n$ with target Z.

d2) If, in RS, a candidate $C_p$ in a right-linking braid $R_p$ with $q_p \geq 2$ has been asserted or eliminated or marked in a previous step, $R_p$ can no longer be used as such as a right-linking inner braid of a $B_r$-braid, because it may no longer be an inner braid. Moreover, there may be several such candidates in $R_p$; consider them all at once. Notice that candidates can only have been asserted as values in the transition from $RS_1$ to $RS_3$ (the candidates asserted in case d1 are all excised from B') and that all the candidates for their CSP variables and all the (global or local) t-candidates they justified in B have also been eliminated in this transition.

Delete from $R_p$ the CSP variables and the local t-candidates corresponding to these asserted candidates. Call $R_p$' what remains of $R_p$ and replace $R_p$ by $R_p$' in B'. A few more questions must be dealt with:

- is there still a candidate for one of the CSP variables of $R_p$' that could play the role of a left-linking candidate for $R_p$'? If not, $R_p$' has already become an autonomous braid in $RS_3$; excise it from B', together with a whole part of B' after it, along the same lines as in case d1;

- is $R_p$' still linked to the next part of B'? If not, excise it from B', together with a whole part of B' after it, as in the previous case;

- $R_p$' may be degenerated (modulo Z and the previous $R_k$'s); this can easily be fixed by replacing $R_p$' with a sequence of right-linking candidates and/or smaller inner braids (modulo Z and the previous $R_k$'s);

- $R_p$' or the sequence of right-linking candidates and/or smaller inner braids replacing it may have more targets than $R_p$; if any of these is a right-linking candidate or an element of a right-linking g-candidate or of an inner $B_{r'}$-braid of B' for some index after p, then mark it so that the information can be used in cases d2, f1, f2 or f3 of later steps.

In RS, B' is now, up to p (the ex q), a partial $B_r$-braid in $B_rB_n$ with target Z.

e) If, in RS, a left-linking candidate $L_p$ has been eliminated (but not asserted) and CSP variable $V_p$ has no t- or z- candidate in $RS_2$ (complementary to case b), we now have to consider three cases instead of the two we had for g-braids.

e1) If $R_p$ is a candidate, then $V_p$ has only one possible value, namely $R_p$; if $R_p$ has not yet been asserted by S somewhere between $RS_2$ and RS, do it now; this case is now reducible to case d1 (because the assertion of $R_p$ also entails the elimination of $L_p$); go back to case d1 for the same value of p (in order to prevent an infinite loop, mark this case as already dealt with for the current step).

e2) If $R_p$ is a g-candidate, then $R_p$ cannot be asserted by S; however, it can still be used, for any CSP variable after p, to eliminate by $W_1$ any of its t-candidates that is g-linked to $R_p$. Let q be the smallest number strictly greater than p such that, in RS, CSP variable $V_q$ still has a global left-linking, t- or z- candidate $C_q$ that is not linked or g-linked or B-linked to any of the $R_i$ for $p \leq i < q$. Replace RS by the state obtained after all the assertions and eliminations similar to those in case d1 above have been done. Then, in RS, excise the part of B' related to CSP variables p to q-1 (included), replace $L_q$ by $C_q$ (if $L_q$ has been eliminated in the passage from $RS_2$ to RS) and re-number the posterior elements of B', as in case d1. In RS, B' is now, up to p (the ex q), a partial $B_r$-braid in $B_rB_n$ with target Z.

e3) If $R_p$ is an inner braid, then $R_p$ is no longer linked via $L_p$ to a previous right-linking element of the braid. If none of the CSP variables $V_p$' in $W_p$ has a z- or t-candidate $C_p$ that can be linked, g-linked or B-linked to Z or to a previous $R_i$, (situation complementary to case b), it means that the elimination of $L_p$ has turned $R_p$ into an unconditional braid. Let q be the smallest number strictly greater than p such that, in RS, CSP variable $V_q$ has a global left-linking, t- or z- candidate $C_q$ that is not linked or g-linked or B-linked to any of the R$_i$ for p ≤ i < q. Replace RS by the state obtained after all the assertions and eliminations similar to those in case d1 above have been done. Then, in RS, excise the part of B' related to CSP variables p to q-1 (included), replace L$_q$ by C$_q$ (if L$_q$ has been eliminated in the passage from RS$_2$ to RS) and re-number the posterior elements of B', as in case d1. In RS, B' is now, up to p (the ex q), a partial B$_r$-braid in B$_r$B$_n$ with target Z.

f) Finally, consider eliminations occurring in a right-linking object R$_p$. This implies that p cannot be the last index of B'. There are three cases.

f1) If, in RS, right-linking candidate R$_p$ of B has been eliminated (but not asserted) or marked, then replace B' by its initial part:
{L$_1$ R$_1$} – {L$_2$ R$_2$} – …. – {L$_p$ .}. At this stage, B' is in RS a (possibly shorter) B$_r$-braid with target Z. Return B' and stop.

f2) If, in RS, a candidate in right-linking g-candidate R$_p$ has been eliminated (but not asserted) or marked, then:

f2α) either there remains no unmarked candidate of R$_p$ in RS; then replace B' by its initial part: {L$_1$ R$_1$} – {L$_2$ R$_2$} – …. – {L$_p$ .}; at this stage, B' is in RS a (possibly shorter) B$_r$-braid with target Z; return B' and stop;

f2β) or the remaining unmarked candidates of R$_p$ in RS still make a g-candidate and B' does not have to be changed;

f2γ) or there remains only one unmarked candidate C$_p$ of R$_p$; replace R$_p$ by C$_p$ in B'. We must also prepare the next steps by putting marks. Any t-candidate of B that was g-linked to R$_p$, if it is still present in RS, can still be considered as a t-candidate in B', where it is now linked to C$_p$ instead of g-linked to R$_p$; this does not raise any problem. However, this substitution may entail that candidates that were not t-candidates in B become t-candidates in B'; if they are left-linking candidates of B', this is not a problem either; but if any of them is a right-linking candidate or an element of a right-linking g-candidate or of an inner braid of B', then mark it so that the same procedure (i.e. f1, f2 or f3) can be applied to it in a later step.

f3) If, in RS, a candidate C$_p$ in right-linking braid R$_p$ of length q$_p$ ≥ 2 has been eliminated (but not asserted) or marked, this has been dealt with in case d2.

Notice that this proof works only because the notions of being linked and g-linked do not depend on the resolution state (they are structural) and the notion of being B-linked is persistent.

**11.3. The "T&E($B_p$) vs $B_p$-braids" and "T&E(2) vs B-braids" theorems**

For $B_p$-braids, for any $p \geq 1$, we are now prepared to expect some extension of the "T&E vs braids" theorem, a "T&E($B_p$) vs $B_p$-braids"; it will be theorem 11.4. But, the really new result (with respect to our above-mentioned expectations) is, if p is infinite, there will also appear a new kind of extension, the "T&E(2) vs B-braids" theorem (theorem 11.5), associated with the iteration of T&E at depth 2.

*11.3.1. The "T&E($B_p$) vs $B_p$-braids" theorem*

As the T&E(T, Z, RS) procedure can been defined for any resolution theory T with the confluence property (see section 5.6.1), T&E($B_p$, Z, RS) can be defined for every p. It is obvious that an elimination done by a $B_p$-braid can be done by T&E($B_p$). The converse is true:

***Theorem 11.4: for any $p \geq 1$, any elimination done by T&E($B_p$) can be done by a $B_p$-braid. As a result, any puzzle solvable by T&E($B_p$) can be solved by $B_p$-braids.***

Proof: it is an easy adaptation of that for g-braids (which are the case p=1 of $B_p$-braids). As the above proof of confluence, it is also remarkably close to the proof for $S_p$-braids, with transversal sets replaced by the sets of candidates linked to some right-linking object (see the similarities in Figure 11.3).

Let RS be a resolution state and let Z be a candidate eliminated by T&E($B_p$, Z, RS), using some auxiliary resolution state RS'. Following the successive applications of rules from resolution theory $B_p$ in RS', we progressively build a $B_p$-braid in RS with target Z. First, remember that $B_p$ contains only four types of rules: ECP (which eliminates candidates), $B_{p'}$ (which eliminates targets of $B_{p'}$-braids, $p' \leq p$), S (which asserts a value for a CSP variable) and CD (which detects a contradiction on a CSP variable).

Consider the sequence $(P_1, P_2, \ldots, P_k, \ldots P_m)$ of rule applications in RS' based on rules from $B_p$ different from ECP and suppose that $P_m$ is the first occurrence of CD (there must be at least one occurrence of CD if Z is eliminated by T&E($B_p$, Z, RS)). We first define the $R_k$, $V_k$, $W_k$ and $q_k$ sequences, for k < m:
- if $P_k$ is of type S, then it asserts a value $R_k$ for some CSP variable $V_k$; let $W_k = \{V_k\}$ and $q_k=1$;
- if $P_k$ is of type $B_{p'}$, then define $R_k$ as the non degenerated $B_{p'}$-braid used by the condition part of $P_k$, as it appears at the time when $P_k$ is applied; let $W_k$ be the sequence of CSP variables of $R_k$ and $q_k=p'$; in this case, $V_k$ will be defined later.

We shall build a $B_p$-braid[n] in RS with target Z, with the $R_k$'s as its sequence of right-linking candidates or $Bq_k$-braids, with the $W_k$'s as its sequence of sequences of CSP variables, with the $q_k$'s as its sequence of sizes and with n= $\sum_{1 \leq k \leq m} q_k$ (setting $q_m=1$). We only have to define properly the $L_k$'s, $q_k$'s and $V_k$'s with $V_k \in W_k$. We do this by recursion, successively for k = 1 to k = m. As the proofs for k = 1 and for the passage from k to k+1 are almost identical, we skip the case k = 1. Suppose we have done it until k and consider the set $W_{k+1}$ of CSP variables.

Whatever rule $P_{k+1}$ is (S or $Bq_{k+1}$), the fact that it can be applied means that, apart from $R_{k+1}$ (if it is a candidate) or the labels contained in $R_{k+1}$ (if it is an $Sq_{k+1}$-braid), all the other labels for all the CSP variables in $W_{k+1}$ that were still candidates in RS (and there must be at least one, say $L_{k+1}$, for some CSP variable $V_{k+1}$ of $W_{k+1}$) have been eliminated in RS' by the assertion of Z and the previous rule applications. But these previous eliminations can only result from being linked or B-linked to Z or to some $R_i$, i≤k. The data $L_{k+1}$, $R_{k+1}$ and $V_{k+1} \in W_{k+1}$ therefore define a legitimate extension for our partial $B_p$-braid.

End of the procedure: at step m, a contradiction is obtained by CD for a CSP variable $V_m$. It means that all the candidates for $V_m$ that were still candidates for $V_m$ in RS (and there must be at least one, say $L_m$) have been eliminated in RS' by the assertion of Z and the previous rule applications. But these previous eliminations can only result from being linked or B-linked to Z or to some $R_i$, i<m. $L_m$ is thus the last left-linking candidate of the $B_p$-braid we were looking for in RS and we can take $W_m=\{V_m\}$. qed.

Here again (as in the proof of confluence), this proof works only because the notions of being linked and g-linked are structural and the notion of being B-linked is persistent. It is also again very unlikely that following the T&E($B_p$) procedure to produce a $B_p$-braid, as in the construction in this proof, would produce the shortest available one in resolution state RS.

### 11.3.2. Definition of the T&E(T, P, n) procedure

In section 5.6.1, we defined the procedures T&E(T, Z, RS) and T&E(T, RS) for any resolution theory T with the confluence property, any candidate Z and any resolution state RS. We can now define the iterated versions of these procedures.

Definition: given a resolution theory T with the confluence property, a resolution state RS and an integer n, the two procedures *Trial-and-Error based on T at depth n for Z in RS* and *Trial-and-Error based on T at depth n in RS* [respectively *T&E(T, Z, RS, n)* and *T&E(T, RS, n)*] are defined by mutual recursion as follows:

T&E(T, Z, RS, 1) = T&E(T, Z, RS) and T&E(T, RS, 1) = T&E(T, RS), where the right-hand sides have been defined in section 5.6.1.

For n>1, T&E(T, Z, RS, n) is defined as follows:
- make a copy $RS_1$ of RS; in $RS_1$, delete Z as a candidate and assert it as a value;
- apply T&E(T, $RS_1$, n-1);

- if $RS_1$ has become a contradictory state (detected by CD), then delete Z from RS (*sic*: RS, not $RS_1$); otherwise, do nothing (in particular if a solution is obtained in $RS_1$, merely forget it);
- return the (possibly) modified RS state.

For n>1, T&E(T, RS, n) is defined as follows:
a) in RS, apply the rules in T until quiescence; if the resulting RS is a solution or a contradictory state, then return it and stop;
b) mark all the candidates remaining in RS as "not-tried";
c) choose some "not-tried" candidate Z, un-mark it and apply T&E(T, Z, RS, n);
d) if Z has been eliminated from RS by this procedure,
    then goto a
    else if there remains at least one "not-tried" candidate in RS
        then goto c else return RS and stop.

Notice that every time a candidate is eliminated by step d of T&E(T, RS, n), all the other candidates (remaining after step a) are re-marked as "not-tried" by step b. Thus, the same candidate can be tried several times in different resolution states. Even with T having the confluence property, this is necessary to guarantee that the result does not depend on the order used to try the candidates (in step c).

Definition: given a resolution theory T with the confluence property and an instance P with initial resolution state $RS_P$, we define *T&E(T, P, n)* as T&E(T, $RS_P$, n).

Definition: for an instance P, *the T&E-depth of P, d(P)*, is the smallest n≥0 such that P can be solved by T&E(n), with the convention that T&E(0) = BRT(CSP).

### 11.3.3. The "T&E(2) vs B-braids" theorem

In the previous definition, taking T = BRT(CSP) and n = 2, and forgetting as usual the reference to BRT(CSP), we get procedures T&E(Z, RS, 2), T&E(RS, 2) and T&E(P, 2). We write T&E(2) when P is clear. It is obvious that an elimination done by a $B_p$-braid of any length can be done by T&E(2). The converse is more interesting:

***Theorem 11.5**: any elimination done by T&E(2) can be done by a $B_p$-braid[n] for some p and some n. As a result, any instance solvable by T&E(2) can be solved by B-braids.*

The proof is a mere iteration of the previous proof. Let RS be a resolution state and let Z be a candidate eliminated by T&E(Z, RS, 2), using some auxiliary resolution state RS'. Following the successive events in RS', we progressively build a $B_p$-braid in RS with target Z. First, notice that there are only four types of such events: three are applications of rules from BRT [ECP (which eliminates candidates), S (which asserts a value for a CSP variable) and CD (which detects a contradiction on a CSP variable)] and the fourth is a call to some T&E($Z_{k+1}$, $RS_k$), where $RS_k$ is the resolution state reached after the k-th event.

Consider the sequence ($P_1$, $P_2$, …, $P_k$, …$P_m$) of such events in RS', forgetting those associated with rule ECP, and suppose that $P_m$ is the first occurrence of CD (there must be at least one occurrence of CD if Z is eliminated by T&E(Z, RS, 2)). We first replace the $P_k$ sequence by a sequence of rule applications:
- if $P_k$ is of type S, then we extend the B-braid under construction exactly as in the T&E(1) case;
- if $P_k$ is a call to T&E($Z_{k+1}$, $RS_k$), then, applying theorem 5.7, we replace it in $RS_k$ by a braid[$q_k$] with target $Z_{k+1}$ for some $q_k≥1$. There remains only to notice that such a braid in $RS_k$ is the same thing as a B$q_k$-braid with target $Z_{k+1}$ in RS, modulo Z and the previous right-linking candidates of the global B-braid under construction.

The rest of the proof is as in theorem 11.4. We skip it.

### 11.3.4. Application of the "T&E(2) vs B-braids" theorem to Sudoku

As the T&E(n) procedure is easy to code in efficient ways, it is also easy to check that ***all the known minimal 9×9 Sudoku puzzles can be solved by T&E(2)***; therefore ***they can all be solved by B-braids and they all have a finite BB rating***. This includes the hardest ones recently generated by "Eleven", as introduced in section 9.6; as we had previously checked that all the published "hardest" puzzles (and conjectured that all the puzzles) could be solved by T&E(2), after he announced his results, we asked him if it was true of his puzzles; Eleven kindly checked this with his program and provided a positive answer; later, when the sublist of his 26,370 hardest became available, we also checked them positively for this property with our independent program. For details on their $B_pB$ classification with respect to parameter p, see section 11.4.2.

In terms of the T&E-depth d(P), this means that there are only 3 possibilities for any puzzle P:

– d(P) = 0 ⇔ no T&E is necessary ⇔ P is solvable by BSRT;

– d(P) = 1 ⇔ only one T&E hypothesis needs be considered at a time ⇔ P is solvable by braids;

– d(P) = 2 ⇔ only two or fewer T&E hypotheses need be considered at a time ⇔ P is solvable by T&E(B) ⇔ P is solvable by B-braids.

Moreover, as there is only a finite (although huge) number of minimal puzzles, it entails that ***there is some p (possibly large) such all the known minimal 9×9 Sudoku puzzles have a finite $B_pB$ rating***.

Two questions remain open: whether all the minimal puzzles (not only all the known ones) can be solved by T&E(2) [we have strong reasons to believe that this is true – *our T&E(2) conjecture*] and what the value of the smallest such p is [we have strong reasons to believe that it is 7 – *our $B_7B$ conjecture*, see section 11.4].

Knowing for certain the smallest p would be interesting, because it would define the maximum look-ahead necessary when one tries to find a solution by structured search with only one hypothesis at a time and with no guessing. Whatever its actual value, it is also clear that such a p would provide a universal rating for Sudoku, in the restricted sense that it would ensure a finite rating to every puzzle (which the BB rating already does, but without a predefined finite value of p).

However, these universal ratings (BB or this $B_pB$) cannot be considered as universal in the non-technical sense that they would be associated with *the* "simplest" solution. As we have seen, although all the whip, braid and generalised whip or braid ratings we have introduced are largely mutually compatible (only rarely do they give different ratings to a puzzle), the cases where they differ also prove that it is not possible to have a single formal definition of simplicity.

## 11.4. The scope of $B_p$-braids in Sudoku

As already mentioned many times in this book, 9×9 Sudoku puzzles that cannot be solved by braids (or whips) are very rare (in percentage; less than one in ten millions). The only available sources of such puzzles are biased, for various non mutually exclusive reasons: they may have been created with particular patterns of givens (e.g. various kinds of symmetries or quasi-symmetries in the given cells, as in the 16×16 and 25×25 examples of section 11.5 below) and/or by algorithms biased by construction for the particular purpose of finding hard instances.

This section can therefore have no more statistical pretension than section 9.6. Instead, we shall review collections of extreme puzzles from different sources. The main result here is that B-braids, i.e. braids accepting inner braids as their right-linking elements, allow to solve all the known (standard, i.e. minimal 9×9) Sudoku puzzles, giving strong credit to our old conjecture that all the puzzles (not only the known ones) can be solved by T&E(2); this is a noticeable difference with S-braids, i.e. braids with inner Subsets. Moreover, it will provide very good reasons for making the stronger *conjecture that all the minimal 9×9 Sudoku puzzles can be solved by $B_p$-braids with p ≤ 7*.

It may be useful to notice that the results reported in this section required innumerable months of handcrafting and CPU time: depending on the source, more or less of each (but in any case not ours) for puzzle creation; mainly CPU (ours) for ratings.

### 11.4.1. Comparison of scope for S$_p$-braids and B$_p$-braids (gsf's collection)

Table 11.1 is the analogue for B$_p$-braids of Table 9.1 (for S$_p$-braids); it is relative to gsf's collection mentioned in section 9.6, with the same slices of 500 puzzles. As the last puzzles can all be solved by g-braids, we have restricted the list to the first 6,000. For easier comparison of the resolution powers of the two series of patterns, small figures recall the values obtained in Table 9.1 (for p ≤ 4). This table shows that *all the puzzles in gsf's list can be solved by B$_p$-braids with p ≤ 6 – and all but 4 (belonging to the first series of 500) can be solved by B$_p$-braids with p ≤ 5*.

Considering the next sub-sections, Table 11.1 also shows that the top-level of this list can no longer be considered as containing the hardest known puzzles. But we keep it here, for two reasons: it has long been *the* reference and it is still interesting for comparing the resolution power of S$_p$-braids and B$_p$-braids.

| Resolution theory → <br> ↓ slice of puzzles | gB$_*$ | B$_2$B$_*$ | B$_3$B$_*$ | B$_4$B$_*$ | B$_5$B$_*$ | B$_6$B$_*$ |
|---|---|---|---|---|---|---|
| 1-500 | 187 | 369$_{336}$ | 457$_{414}$ | 482$_{443}$ | 496 | 500 |
| 500-1000 | 178 | 364$_{335}$ | 462$_{415}$ | 496$_{460}$ | 500 | |
| 1001-1500 | 163 | 421$_{382}$ | 492$_{451}$ | 500$_{486}$ | | |
| 1501-2000 | 168 | 437$_{397}$ | 499$_{476}$ | 500$_{490}$ | | |
| 2001-2500 | 135 | 412$_{367}$ | 498$_{434}$ | 500$_{474}$ | | |
| 2501-3000 | 116 | 386$_{334}$ | 495$_{443}$ | 500$_{479}$ | | |
| 3001-3500 | 120 | 389$_{335}$ | 496$_{424}$ | 500$_{473}$ | | |
| 3501-4000 | 113 | 372$_{325}$ | 493$_{426}$ | 500$_{472}$ | | |
| 4001-4500 | 104 | 345$_{298}$ | 475$_{395}$ | 499$_{448}$ | 500 | |
| 4501-5000 | 231 | 433$_{399}$ | 493$_{450}$ | 500$_{482}$ | | |
| 5001-5500 | 348 | 495$_{487}$ | 500$_{500}$ | | | |
| 5501-6000 | 490 | 500$_{500}$ | | | | |
| Total solved /6000 | 2353 | 4923 $_{4495}$ | 5860 $_{5328}$ | 5977 $_{5707}$ | 5996 | 6000 |
| Total unsolved /6000 | 3647 | 1077 $_{1505}$ | 140 $_{672}$ | 23 $_{293}$ | 4 | 0 |

**Table 11.1.** *Cumulated number of puzzles solved by B$_p$-braids, p'≤p, for each slice of 500 puzzles in gsf's list. The second column here (for gB$_*$) corresponds to the third in Table 9.1.*

### 11.4.2. Eleven's collection of puzzles solvable by B$_p$-braids but not by S$_p$-braids

Let us now turn to the collection recently generated by "Eleven", already introduced at the end of section 9.6. Eleven has made public [Eleven 2011] a list of 26,370 puzzles that, by construction, cannot be solved by S$_4$-braids (and therefore not by any S-braids), which recommends them for consideration among the hardest. Needless to say, this list has been a great leap forward into the realm of the hardest puzzles. We have already stated in section 11.3.4 that they can all be solved by T&E(2) and therefore by B-braids; let us now be more precise about the maximum value of p for which $B_p$-braids are enough.

As this collection has been generated with the explicit purpose of maximising the SER, we have organised the distribution table (Table 11.2) by slices of constant SER (notwithstanding all the possible criticisms about SER as a measure of complexity). Contrary to the above presentation of gsf's list, slices have variable size. In a row, empty cells on the left or right mean that all the puzzles in the slice can be solved by $B_{p'}$-braids for some of the p' in the other cells. For the slices with SER < 11.3, we restricted our analysis to the first hundred puzzles in each of them.

Table 11.2 shows that *all the puzzles in Eleven's "no $S_p$-braids" collection can be solved by $B_p$-braids with p ≤ 7; moreover, only two of them cannot be solved by $B_p$-braids with p ≤ 6* and only 36 cannot be solved by $B_p$-braids with p ≤ 5. It also shows that there is some vague correlation but no systematic relationship between the SER and the minimum p of a puzzle P.

| Resolution theory → | | | | | | | |
|---|---|---|---|---|---|---|---|
| # of puzzles in slice ↓ | puzzles tried in this slice | SER | $B_2B_*$ | $B_3B_*$ | $B_4B_*$ | $B_5B_*$ | $B_6B_*$ | $B_7B_*$ |
| 4 | 1-4 (all) | 11.9 | | | | $3_{75}$ | 0 | **1**$_{25}$ |
| 20 | 5-24 (all) | 11.8 | | | $1_5$ | $8_{40}$ | $10_{50}$ | **1**$_5$ |
| 34 | 25-58 (all) | 11.7 | | | $4_{12}$ | $20_{59}$ | $10_{29}$ | |
| 48 | 59-106 (all) | 11.6 | | | $17_{36}$ | $27_{56}$ | $4_8$ | |
| 109 | 107-215 (all) | 11.5 | | $9_8$ | $56_{51}$ | $43_{40}$ | $1_1$ | |
| 263 | 216-478 (all) | 11.4 | | $35_{13}$ | $131_{50}$ | $88_{34}$ | $9_3$ | |
| 1207 | 479-578 | 11.3 | | 24 | 64 | 12 | | |
| 1689 | 1686-1785 | 11.2 | 3 | 45 | 44 | 8 | | |
| 2656 | 3375-3474 | 11.1 | 9 | 60 | 30 | 1 | | |
| 1818 | 6031-6130 | 11.0 | 23 | 70 | 7 | | | |
| 2427 | 7849-7948 | 10.9 | 70 | 23 | 7 | | | |
| 2931 | 10276-11275 | 10.8 | 22 | 49 | 28 | 1 | | |
| 4606 | 13207-13306 | 10.7 | 32 | 54 | 14 | | | |
| 8558 | 17813-17912 | 10.6 | 50 | 46 | 4 | | | |

*Table 11.2. For each slice of puzzles of given SER (SER version 1.2.1) in Eleven's collection, non-cumulated number (and percentage, in small digits) of puzzles solved by $B_p$-braids.*

[Additional comments on Table 11.2, for Sudoku experts:

– each slice with fixed SER has been ordered by Eleven according to secondary and ternary criteria, respectively EP and ED, the Sudoku Explainer rating of the hardest elimination step before the first assertion step (resp. of the first elimination step); these ratings do not seem to have any impact on the B$_?$B classification results;

– the discontinuity in behaviour between SER 10.9 and SER 10.8 is inherent in the definitions of these SER values; in between them there are two discontinuities, one in the types of "contradiction / forcing chains" it is based upon (this can easily be noticed in their names, even though these types are defined only by their Java code) and (an anomalous) one in the number of "nodes" used by these "chains": "`10.8: Dynamic + Forcing Chains` (*289-384 nodes*) `CRCD Forcing Chains`"; `10.9: Dynamic + Multiple Forcing Chains` (*73-96 nodes*) `CRCD Forcing Chains`; this discontinuity is only one of the many inconsistencies of the SER.]

Table 11.3 (which does not claim for exhaustivity beyond the sub-slices used in Table 11.2) displays the most remarkable puzzles in Eleven's collection, according to the following criteria: they have *either* extreme SER (4 puzzles with SER=11.9 and 20 with SER=11.8) *or* extreme p (2 puzzles with p=7 and 34 with p=6). The three occurrences (in bold), with values 4 or 7 for p, are unexpected. Considering the values of SER between 11.8 and 11.4, it is also unexpected that only one percent of the puzzles with SER=11.5 have p≥6; but this may be due to some bias in Eleven's collection and/or to some other obscure anomaly of the SER.

| puzzle | # in Eleven's list | SER | p |
| --- | --- | --- | --- |
| ..3....8..5.1....66....74....8.9..4.7....5...1.6..8.....9...2.....2....8..2...3.4 | 1 | 11.9 | 5 |
| .2.4...8.....8...68....71..2..5...9..95.......4..3..........1..7..28....4.....6.3.. | 2 | 11.9 | 5 |
| ..3....8..5...2.17..........5.8..6.9.12....8....3...6.9....5..4....7.....1.6.2 | 3 | 11.9 | **7** |
| ..3..6.8....1..2......7...4..9..8.6..3..4...1.7.2.....3.....5.....5..6..98.....5. | 4 | 11.9 | 5 |
| 1.......9..67...2..8....4.......75.3....5....2....6.3......9....8..6....4....1...25...6. | 5 | 11.8 | 6 |
| 1...6.8....7..1.........5.6..9.4......7.2...3.8....76..3.....1..5.4.9........2.7.... | 6 | 11.8 | 5 |
| ...4...89..7..92......3...526....1......19......7....1..5...9...4...6..29......8....3 | 7 | 11.8 | **4** |
| .........94....92.....7..45...1.3.....7.6..9..8....7..2.3.7...8.....6.1....9.....5.2. | 8 | 11.8 | 5 |
| ...4...8...7..92......3...526....1......19......7....1..5.......4..1.8...3..6..29.. | 9 | 11.8 | 5 |
| .2.4..7....6......17...3.....5....6..4.2..9........5..8...1..8....9...7.......92.3.. | 10 | 11.8 | 5 |
| 1.......9.5....2....87....4.2...3........48.5....8.6....7....6..4.5...........1....9.3.. | 11 | 11.8 | 6 |
| ..3.5.7....4.....9....6..2........5...8.3.9.....6.8............8.1.....75.....4.2.....3.5.8 | 12 | 11.8 | 6 |
| ..34....8........1.37..........2.....9........5..8....6....7.4...51....8.7....5....9....62.5. | 13 | 11.8 | 6 |
| 1......8......92....6.3....52.....8....5.7....6.5....4..47............91..3...6....7 | 14 | 11.8 | 5 |
| ......7.9...1....3..8....74......9.....8.5.....5...75.6......2....2.6.......13......94....8.... | 15 | 11.8 | 6 |

| Puzzle | # | SER | p |
|---|---|---|---|
| ..34......5..89...78...2....2.....5..7...6..41....9.....5........6.8...9...2..1...3.. | 16 | 11.8 | 6 |
| .....6..94...8.2.....7...1.2.9....8....4.3.9....6.....5.3.8.4........5......7...1... | 17 | 11.8 | 6 |
| .....6.8....1...2...9..7...5..5.4...734....8...97.........9...6..7...3....4....2...1. | 18 | 11.8 | 5 |
| ..3..6.8....1......9..7....4..8..6..3....4....2.....5.1..2.9...37........94...5.. | 19 | 11.8 | 5 |
| ..34.........8...668..7.1.....5......9..1.6.......2..4...5.....28.....96.97.....1. | 20 | 11.8 | 6 |
| .2....78.4......6.9..7..1.....5.....3......1.......9.12..7..1.8...5....4.....67.3... | 21 | 11.8 | 5 |
| .2....67..4....8......9........3....7.5.8....4..1.3....2....9...5....6.1...3....2..6.7 | 22 | 11.8 | **7** |
| .2.4...7....5....9.3.6.....7......5..8.9.7...2......4.6..3........1..85......1....1...9.3 | 23 | 11.8 | 6 |
| 1....6......8.2....9.7....5.7.3.....5.......16.....4....73..59......48.....2......3....... | 24 | 11.8 | 6 |
| ..34...8...........7...25..2..........49...1.9......6.7......5..6..9.1..3...8.34.... | 26 | 11.7 | 6 |
| .2....6.......1....3...9.7....5...5.....78.3......1.8.....4.5......4.9.8.....6.....2.....9........7 | 27 | 11.7 | 6 |
| .234.........8....36......4........5...6..19....3......7.8....19......2.5........7.4.3....9. | 29 | 11.7 | 6 |
| .2...67........8......91......23......7...7......34......1......8.....9....5...2....4......6....342. | 30 | 11.7 | 6 |
| ....5.78......1..9.......7...1..9....1...3..6.2.......4.9.3.....7....3.48........6.....2.....5. | 34 | 11.7 | 6 |
| ......67...5.1...3......3....4...8..4.......3..52.......9...1...2.7......6....3.9....1.....5......8 | 45 | 11.7 | 6 |
| ...4......9.5....9.3..........1.52..8........6.4......1...53....42....7.8.....67......7......3 | 46 | 11.7 | 6 |
| 1......89....8..2....82..5......93.......5......4....7..6.3...6......1......4.....7........28.....5. | 53 | 11.7 | 6 |
| 1.......7.9.57......3....8.7......2....4........68......38......5.......1.....4.......9.....2......3..56. | 54 | 11.7 | 6 |
| 1.......7.9.57......3....8.7......2....4........68......38......5.......1..2......9......4......3..56. | 55 | 11.7 | 6 |
| .2....67..4....8....2...9......5......8.....4.5..........3.1....2.....89......6.3..1....7......73.. | 85 | 11.6 | 6 |
| ..3......94....8.2....6.7....1.2.....9.....8...4......3........1.5.3.8.4.........6.......7......5... | 99 | 11.6 | 6 |
| ..3......9....1....63........75........196......4.........7......5........6....21..92......3..8.........4... | 100 | 11.6 | 6 |
| 1.......7.9.57......3....8.7......2....4........68......38......5.......1..2......9......4......3..56. | 103 | 11.6 | 6 |
| .....5.7..4......9........83.....1..8..........12...6..9....7........6......18.3..8......2......54... | 176 | 11.5 | 6 |
| 1....6..8...5.....9......837...........3.....4..42......8..6......1....2..4..8.......7.....39.........5. | 287 | 11.4 | 6 |
| 1....5......7..9......8.3...4......5.1....6..6.8......4.........7......3..2.........4.......2....3..89.......2 | 289 | 11.4 | 6 |
| ......5..8....71..2......2...1..4....4........7.......6..196......3..3......5......5......9........426...... | 335 | 11.4 | 6 |
| ...34...7....5.....9..........3....6...........8...4..7...2..91.................6.2.....23......4..8....2..5... | 342 | 11.4 | 6 |
| .2..5.........71.9....6......2........2..........8....4......91....9...3..4......76......5.........13.............47. | 349 | 11.4 | 6 |
| 1....67.....5................927......87.....4.3.........1....57........3..........9........8..6.....75...........4...2. | 357 | 11.4 | 6 |
| 12.4.........4...1......6.....8.3.........5.....9.7.6.......2.74.................9.3............35..........7....12 | 365 | 11.4 | 6 |
| ...45...8........92.......7....452.......3......8........586....7.3............6.7......4.9.......1..... | 391 | 11.4 | 6 |
| .2.4..............6........7......35......8......63...91.7.9................2......8.......1.35......75.9. | 441 | 11.4 | 6 |

*Table 11.3. Puzzles from Eleven's collection with extreme SER or p*

### 11.4.3. Other extreme puzzles are also solvable by $B_P$-braids with $p \leq 7$

Finally, as much effort has been invested over the years by many people in the search for extreme puzzles (according to various criteria, but mostly SER), let us consider a few famous ones proposed as such by different creators. Our source here is the meta-collection compiled by "Champagne" [Penet 2012], based on previously known lists and on his own creations. However, as we have already considered Eleven's puzzles separately in the previous sub-section (mainly because he provided some description of his generation process and he used the same one uniformly to produce his whole collection), we have extracted them from the results in the Tables below (this will also make further comparisons easier). For convenience, let us call "Champagne-minus-Eleven" the resulting collection.

Tables 11.4 and 11.5 are the respective analogues of Tables 11.2 and 11.3 for this complementary collection (but now limited to its puzzles with SER $\geq$ 11.6, because it does not lead to anything new). Only one puzzle requires $B_7$-braids.

| Resolution theory → | | | | | | | |
|---|---|---|---|---|---|---|---|
| # of puzzles in slice | SER | $B_2B_*$ | $B_3B_*$ | $B_4B_*$ | $B_5B_*$ | $B_6B_*$ | $B_7B_*$ |
| 3 | 1-3 | 11.9 | | | | $2_{66}$ | $1_{33}$ | $0_0$ |
| 16 | 4-19 | 11.8 | | | $2_{12}$ | $7_{44}$ | $6_{38}$ | **$1_6$** |
| 22 | 20-41 | 11.7 | | | $4_{18}$ | $10_{46}$ | $8_{36}$ | |
| 49 | 42-90 | 11.6 | | $5_{10}$ | $21_{43}$ | $13_{27}$ | $10_{20}$ | |

*Table 11.4. For each slice of puzzles of given SER in Champagne-minus-Eleven's meta-collection, non-cumulated number (and percentage, in small digits) of puzzles solved by $B_p$-braids.*

| puzzle | creator | name in list | SER | p |
|---|---|---|---|---|
| 98.7.....7....6...6.5.....4...5.3...79..5.....2...1..85..9......1....4.....3.2. | GPenet | **Champagne_dry** | 11.9 | 6 |
| 98.7.....6.....87...7....5.4...3.5......65....9......2...1..86....5.....1.3.........4..2 | GPenet | kz0_11523 | 11.9 | 5 |
| ........39.....1...5...3.5.8....8.9...6.7....2....1...4........9.8...5....2....6...4..7..... | Tarek | **Golden Nugget** | 11.9 | 5 |
| 98.76....54..........7..59..4....75......3.....2......1...6.9...87....4....1.....2....3 | GPenet | kz1a_15497 | 11.8 | 5 |
| .2...67...4...8.......93........9...57...1....7....2.....61.3....4....6....8........6...5.2. | Tarek | tarx0075 | 11.8 | 5 |
| 2......6.5...8...1....4......9....7.3.1........82........7.5.3.....9.....4.....8...1..5.6........2 | Tarek | tarek-ultra-0203 | 11.8 | 5 |
| 98.7.....6.....5.....4..93...5.........6...7...8.......9...24.....1........4.....9.1.........1.32 | GPenet | H1 | 11.8 | 5 |
| 1........2...94....5....6.....7......89...4....3.6........8.4.......2.....1...7.........6...5.8...3. | gsf | 2007-05-24-003 | 11.8 | 5 |
| 6.........2.9.4....5......1.....7......5...84.........2........3.5.4.2........6....3....9.8....7......1 | Coloin | coloin-04-10_13 | 11.8 | 6 |
| 1........2.9.4.......5....6.....7......5.3.4........6..........58.4......2......6....3....9.8.7..........1 | Coloin | coloin-04-10_14 | 11.8 | 5 |
| 1.......2.3.4......5.....6.....7......5.8.4.........29........3.....9......7......1.9....8.4.2........6.. | Hp54 | Hp54_4 | 11.8 | 6 |
| 1........2.3.4.....5.....6......7......5.9.4.........23........8.....9......2.....6....9...8.4.7...........1 | Hp54 | Hp54_1 | 11.8 | 6 |
| 98.76....75.....9......6.........8.....4....3...2........1..95...8......86....5.......3......4....1.2. | GPenet | KZ1C_23862 | 11.8 | 5 |
| ...3.8.......7...2.........6....9.1...........3........596..9......54.1....45....8......3..........27...... | Tarek | 071223170000 | 11.8 | 6 |
| 1........6.5.7........8......3.......4.........5.8.9........3........8.92........6........3.....7....5.2........4......1 | Coloin | H1 | 11.8 | 6 |
| 5.........9.2.1.....7......8....3......4.6...........5........2.7.1.....3......8......6.....4.2.9.........5 | Metcalf | no name | 11.8 | **7** |
| 98.7......6..89.....5..4...7...3.9....6...7...2...51.6..8.3.......1.4.........2 | GPenet | H3 | 11.8 | 6 |
| 98.7......6..89.....5..4...7...3.9....6...7...2...41.6..8.3.......1...5.........2. | GPenet | H2 | 11.8 | 5 |
| 98.7.......7..6..........57..4...3..2...1...6...3...9..8..2.......4..3.......1..86..5... | GPenet | cy4_9253 | 11.8 | 5 |
| 1........9.4...3.8.......2....6...7..58.........2.......7.4.5.....6.....2...3.8...7.9..........1 | Tarek | tarek-2803 | 11.7 | 6 |
| 1........2.9.4.....5.....6.....7...5.3.4.......96..........8.4...2.....6.....3...9.8.7........1 | Coloin | Coloin_04_10 | 11.7 | 6 |
| 1........2.3.4.....5.....6.....7...5.8.3..........7........95..8.7......6.....9...8.3......2......1 | jpf | jpf-04-08 | 11.7 | 6 |
| ..34..7........9..2......1..5.2...........38...6...6.43...........2...9........5..1.6..8..3... | Tarek | pearly6000-4268 | 11.7 | 6 |
| ..34..7........9..2......1..5.27..........38.....6.....43...........2..9........5..1.6..8...3... | Tarek | pearly6000-3802 | 11.7 | 6 |
| 1........2..34.....5....6......7.....85..9......3..6........8..9........2.....1...7.......6..9..8..3. | jpf | jpf-04/14/84 | 11.7 | 6 |
| 98.7......7..6..8......5..4...37........6.........2....31....3...98......1.....2......5..4 | GPenet | H8 | 11.7 | 6 |
| ..1...5..2..4....6..3....7........6.28..........9...2........4..65.....1...9..8....4......7......3 | Coloin | H2 | 11.7 | 6 |
| 1........2.3.4.....5.....6.....7...5.8.3..........74......9..8.7......6......9...8.3......2......1 | jpf | jpf-04/14/02 | 11.6 | 6 |
| ..34....8..5....1...7.......6..1.......5...8.9...2.6........7...294..........3....4.......8..5... | Tarek | pearly6000-4143 | 11.6 | 6 |
| 3........8.7..5.....1.....6.....4...9.2.1..........4........97..2..4........3.....5...2..7.....8.......6 | Tarek | tarek-ultra-0313 | 11.6 | 6 |
| 1........2.9.4.....5.....6.....7...5.9.3..........7........85...4.7.......6.....3...9.8......2......1 | jpf | **Easter Monster** | 11.6 | 6 |
| .......35......2.6...3.5..8......5.9...6.7.....9...1...4.........6.8....9...2.1......4.....7... | Coloin | H4 | 11.6 | 6 |
| ..345..........9......2.34....1........7..4.2.8..9........6....28..5....6........9...7......1 | Tarek | pearly6000-3238 | 11.6 | 6 |
| 987........65..........49...8.....5...8..7........3...4......2....1.6.7...5........4......3......1.2. | GPenet | H10 | 11.6 | 6 |
| 5..........9.2.1........7.......8.....3........4........2..........5.........7.6.1.......3........8........6......4.2.9..........5 | StrmCkr | StrmCkr_103 | 11.6 | 6 |
| ........8......6.....12.........2....6.5...15...9...8.........3........4...7........3........8.......21.......6.7...4........ | Coloin | H5 | 11.6 | 6 |
| 98.7......6..89.....5..4...7...3.9....6...7...2...41.6..8.3.......1.5...........2 | GPenet | H15 | 11.6 | 6 |

*Table 11.5. Puzzles in Champagne-minus-Eleven's meta-collection with extreme SER or p (names of famous puzzles appear in bold; the only puzzle in $B_7B$ was not famous before).*

**11.5. Existence and classification of instances beyond T&E(2)**

*11.5.1. Existence of instances beyond T&E(2)*

Anticipating on a question that might naturally arise now, let us notice that our T&E(2) conjecture for the standard 9×9 Sudoku CSP cannot be extended to larger Sudoku grids, let alone to any CSP. "Blue", a participant of the Sudoku Programmer's Forum, reported[12] that, using his generator – *a priori* biased towards easier instances, because of the top-down kind (see chapter 6):

– 46% of his randomly generated 16×16 minimal puzzles required T&E(2), although he could not find one requiring T&E(3) in 1.9 million random tries;

– 90% of his randomly generated 25×25 minimal puzzles required (at least) T&E(3).

---

[12] http://www.setbb.com/sudoku/viewtopic.php?t=2117&start=135&mforum=sudoku

```
. 4 . 9 . 2 . . . . . 6 . D . A     . . . 4 . E . 2 . 3 . F . . . 9
B . A . D . . . 4 . . . E . F .     . . G . 4 . 7 . 6 . 8 . . . F .
. 3 . 5 . . . 1 . 7 . . . 4 . C     . 5 . A . C . G . 7 . . . E . .
6 . 1 . . . 8 . F . 9 . . . 5 .     2 . 1 . A . D . 9 . . . C . . .
. C . . . 6 . 8 . 1 . F . . . 5     . F . B . 9 . 1 . . . 6 . . . G
4 . . . B . C . 3 . 7 . 2 . . .     G . A . 8 . B . . . E . . . 6 .
. . . 3 . A . 7 . E . 4 . 1 . .     . 7 . 3 . D . . . 8 . . . F . 5
. . 8 . F . 4 . C . 5 . 7 . B .     6 . C . 5 . . . 2 . . . 4 . 1 .
. . . G . C . 6 . 9 . B . 2 . .     . A . C . . . B . . . 3 . 7 . 2
9 . . . 1 . D . 2 . E . G . . .     4 . D . . . 1 . . . A . G . 3 .
. B . . . E . 3 . A . 5 . . . 7     . 1 . . . 7 . . . C . 8 . 4 . B
C . 4 . . . 5 . 7 . 1 . . . 6 .     3 . . . E . . . 4 . 6 . 9 . 5 .
. D . 8 . . . A . 3 . . . B . G     . . . D . . . 3 . 4 . 2 . 6 . E
5 . F . 6 . . . 9 . . . 1 . 8 .     . . 9 . . . G . C . 1 . 7 . D .
. 9 . 2 . 8 . . . . . 7 . 6 . E     . 2 . . . 5 . D . F . 9 . G . .
1 . 6 . C . E . . . D . 9 . 7 .     7 . . . F . C . 5 . 3 . 8 . . .
```

**Figure 11.5.** *Two 16×16 puzzles with T&E-depth ≥ 3 (hexadecimal notation)*

```
. . J E F 4 . . 3 . . A D . . . . 7 . G . . C . .
. . I 9 P . . 1 . . 4 6 . . J . . . 3 . 8 . . K
C M 2 K . . E 6 . . O . . 3 . B . . . . . G . . 5 .
8 L 1 . . D I . G K . . B . E . 4 . . . . . F . 9
3 B . . . F . N J . . . . 5 . D . . M . . 2 . O L
1 . . B N . 7 G . . 5 . . . 6 . . D . . . . A J P
. . 8 O . 5 D . . J . I . . . . E . . 2 . L . 9 4
. J K . D A . . C . N . H . . 8 . . P . 5 3 . . 2
A 4 . 6 7 . . 3 . H . C . . K . . . . I B M O . .
. . 5 P . . F . K . 1 . . 2 G . . . 9 6 8 D N . .
. . 6 . . K . L . 4 . . 9 . . E . . D H I . . M F
M . . . I . 8 . 9 . . D . . P . 2 4 . . . . L 3 .
N . . . . 3 . B . . F . . 7 . M 8 P I . . K . . G
. . A . . . G . . D . . 4 . 5 N K 9 . . . C . B .
. 3 . G . . . . P . . 2 . I 8 7 6 . . C . . 1 . .
9 . . . 2 . . H . . 3 . 5 O N P . . K 4 . F . . .
. 6 . C . . . . . B . G F 4 2 . . 1 A . P 9 . . E
D . F . . P . . . . . J 6 L . . 9 G . B O . . K .
. 5 . . E . . 8 . . D B K . . I 3 . 6 . . . H . 7
. . . 1 . . N . F G . 7 . . 9 L . E 8 . . B . 2 .
. . . . . M . E D 2 . . . J A . B 3 . . L . 4 . 6
. I . . L . 4 . H F . . . 8 . 1 O . . J . E . N .
5 . . F . 6 L K B . . E . . O . . . 2 . 9 . 8 . .
. . E . B . 3 A . . 9 H . . . . . F . L . G . 1 O
. 8 . 4 M O . . . I . . 3 F . . A . . . 2 . 5 P .
```

**Figure 11.6.** *A 25×25 puzzle with T&E-depth ≥ 4*

Blue also found a few minimal puzzles with symmetries in the pattern of given cells; it is known that such symmetries often lead to harder puzzles in the mean (although the hardest ones do not necessarily have any symmetries at all). T&E(d) computation times grow so fast with d that he computed only a lower bound for d, but this is enough for our present purposes. He posted (in the same thread):

– fifteen 16×16 minimal puzzles requiring at least T&E(3) (two of them appear in Figure 11.5, in hexadecimal notation),

– two 25×25 minimal puzzles requiring at least T&E(4) (one of them is given in Figure 11.6).

What this suggests is that, as grid size n increases, the depth of T&E required by the worst cases will also increase unboundedly (and this would probably remain true of mean case analysis). At what speed it increases remains an open (and apparently very difficult) question.

Even though, as remarked in the Introduction, any finite Constraint Satisfaction Problem can be reformulated as a CSP with only binary constraints, one must sometimes consider "implied" or "derived" constraints that are not binary. Indeed, what this section has shown goes much further: there are naturally binary CSPs (such as large size Sudoku) with minimal instances beyond T&E(2) and this corresponds to the necessity of tackling contradictions among more than two labels. How this can be done will be the topic of the next chapter.

### 11.5.2. Classification of instances beyond T&E(2)

In any CSP, instances P can be classified according to the minimum depth of T&E, $d(P)$, necessary to solve them – whatever the maximum value of $d(P)$ may be in this particular CSP. Moreover, as T&E(d) is equivalent to T&E(B, d-1), the various instances P within each of the layers thus defined can be further sub-classified, inside their layer $d(P)$, according to the smallest p such that they can be solved by T&E($B_p$, $d(P)$-1). This is what we did with the 9×9 Sudoku CSP: for the $d(P)=1$ case in chapter 6 and for the $d(P)=2$ case in section 11.4.

This is a reasonable classification, because the deepest level of T&E is also that which entails the highest computational cost (roughly speaking, mean computation times are close to exponential in d for a fixed size CSP). In line with our previous remarks on the multiplicity of ratings, it seems to us that, for instances in and beyond T&E(2), this is much more informative than what a single rating (some counterpart of the BB or $B_7B$ ratings in case d=2) can provide.

At first sight, as T&E(d) is a procedure, the above classification may seem extra-logical; but it could be shown that the "T&E vs braids" and "T&E(2) vs B-braids" theorems can be generalised, so that being solvable by T&E(d) is equivalent to being solvable by some resolution theory. However, the patterns necessary to do this would be so complex (with several levels of inner braids) that this theoretical result would probably be of no practical use.

*11.5.3. Depth of T&E versus backdoor-size*

Finally, when speaking of T&E(d), the notion of a backdoor inevitably comes to mind and a question immediately arises: is there a relationship between the backdoor-size of an instance and the depth of T&E necessary to solve it?

Definition: given an instance P of a CSP, *the backdoor-size of P, b(P)*, is the smallest integer n≥0 such that there exists a set B of n labels (a backdoor set) that, when added to the givens of P, allows to solve P within BRT(CSP). (As the CSP is finite, there is always such an n.)

[More generally, one can also define the backdoor size b(T, P) of P for any resolution theory T as the smallest n such that there exists a set B of n labels (the T-backdoor set) that, when added to the givens of P, allows to solve P in T. And one can ask about the relationship between b(T, P) and d(T, P), where d(T, P) is defined similarly to d(P). For simplicity, we shall consider here only T = BRT(CSP), but more on this topic can be found on our website. The notion of a strong T backdoor is also introduced there, with an application to the famous EasterMonster puzzle.]

Now, given an instance P of the CSP, one can associate with it two intrinsic constants: its backdoor size b(P) and its T&E-depth d(p). And our initial question gets formalised as: is there a relationship between d(p) and b(p)? The answer is not obvious because the backdoor-size b(P) is based on guessing b values (it is therefore largely incompatible with our approach and with the usual requirements of Sudoku players), whereas the T&E-depth d(P) is based on proving that some sets of d (or fewer) hypotheses are contradictory. In particular, none of the relations d ≤ b or b ≤ d or of their negations is obvious in the general CSP.

Let us therefore consider the Sudoku example again. It has long been believed that all the puzzles P had backdoor size b(P) ≤ 2, but Easter Monster was the first example with b(P) = 3. There are now strong reasons to conjecture that b(P) ≤ 3 for any 9×9 puzzle (and this is true for all the known ones).

Consider first the question "d(P) ≤ b(P)?". If a puzzle can be solved by T&E at depth d, it does not mean that one can choose d fixed hypotheses to generate all the auxiliary grids necessary to the T&E procedure: indeed, this procedure may make hypotheses on any sets of d candidates. But we currently have no explicit counter-example to d(P) ≤ b(P). Notice that, if we consider gsf's lists related to backdoors [gsf www], either his "FN-1" list of 1,183 puzzles P with b(P) = 1 or his "FN-2" list of 28,948 puzzles P with b(P) = 2, all of them can be solved by ordinary T&E, i.e. they have d(P) = 1, thus satisfying d(P) ≤ b(P).

As for the question "b(P) ≤ d(P)?", it is easy to find counter-examples. If we consider again gsf's FN-2 list of 28,948 puzzles P with b(P) = 2, all of them can be solved by ordinary T&E, i.e. they satisfy 1 = d(P) < b(P) = 2.

One can even find counter-examples to the question "b(P) ≤ d(P) + 1?". If we consider gsf's list [gsf www] of 14 puzzles P with b(P) = 3, reproduced below:

```
#1    1.......2.9.4...5...6...7...5.9.3........7.......85..4.7.....6...3...9.8...2.....1 ; Easter-Monster; SER= 11.6
#2    9.......5.4.3.....6....2....1...8.74........2........8.6.7.1.....9...3...7.4.....5.....2 ; tarek-ultra-.3..; SER= 11.3
#3    7........4.2.6....1....5....8....3.91........5........2.3.9.8......7....6....9.2....4......5 ; tarek-ultra-.3.1; SER= 11.3
#4    1......89......91.2......4......76.....3....4.....9....2....5...4.7.......5.....8.1..6.3..... ; tarek-4/.8; SER= 11.5
#5    1........2..34...5...6....7.....89..4....3.6.....9.4......2.....1..7........6...5.8..3. ; jpf-.4/14/.8; SER= 11.2
#6    5.......3.2.6.....1....8....9....4.7.1.......3........42...7.9......5....1...7.2....3........8 ; tarek-ultra-.3.2; SER= 11.2
#7    1.......2..34.....5...6......7......5....4....3.1......894.....2......1...7..........6....5.9..3. ; jpf-.4-1.; SER= 11.2
#8    1........6.2.5.......4......3......7.....4.85.........1.........24.8...7....3....5...9.2.6........1 ; coloin; SER= 11.3
#9    1........6.2.5.......4......3......7.....4.89........2.4.........15.8...7....3....5...9.2.6........1 ; coloin-.5/11/.1; SER= 11.4
#10   ..1...2...3......4.5....3....6....1.7......4......8....9.2....3.........8.6...5...3.....2....7.. ; ocean-2..7-.5-29-1; SER= 9.4
#11   3.......2........54.........6.....1.2...3..............648.............9.7....5........1.8.5...6..... ; gfroyle-2..7-.5-3.-4; SER= 3.6
#12   .8..9....3.......6....3....4......1....5..2.....9.....7....8...65.......1......2.7...........4..... ; gfroyle-2..7-.5-3.-3; SER= 4.2
#13   .....2.58..4.3........1...........6.....715.......2.......4......2.....59............3.67......... ; gfroyle-2..7-.5-3.-2; SER= 5.7
#14   ...5.....13.8........4........3........61.....9........8........5.........6.7...2..1.......3...........49. ; gfroyle-2..7-.5-3.-1; SER= 6.6
```

then four of them (numbers 10, 11, 12 and 14) can be solved by ordinary T&E(1), i.e. they satisfy 1 = d(P) < b(P) - 1 = 2; [the remaining ten can be solved by T&E(2), i.e. they satisfy d(P) = 2 and therefore d(P) = b(P) - 1].

The last four puzzles in this small collection are interesting because they show that a large backdoor size can also be found in easy instances (i.e. with small SER) and therefore backdoor size cannot have much to do with the difficulty of solving.

As a result, it does not seem that the notion of backdoor size (intrinsically based on guessing) can shed much light on classifications of puzzles, like those based on the resolution rules defined in this book, that reject *a priori* any form of guessing. This conclusion is strengthened if we consider larger size grids: on the Sudoku Programmer's Forum[13], Tarek has proposed a 16×16 puzzle P (Figure 11.7, in hexadecimal notation) with backdoor size b(P) = 5. P can be solved by whips of length 9 (which is relatively easy for a 16×16 puzzle). It entails that b(P) = 5 but d(P) = 1.

We shall now give the full resolution path of this puzzle (as compacted as possible) for the main purpose of suggesting a reason why 16×16 Sudoku has never (and in our opinion will never) become popular: even for relatively easy puzzles, with nothing special, there are always lots of tedious eliminations. It is essential to understand that the length of the present path can in no way be compared to those we gave in chapters 5 or 7 for whips or g-whips: those examples had very long paths (and, for most of them, very long whips) because they were quite exceptional in some respect; here the length is typical of all but the easiest 16×16 puzzles. This also shows how our theoretical interpretations of the general requirements of simplicity or explainability can be challenged by more pratical concerns of boredom for instances of CSPs that are not really designed for human solving. It seems that this concern will appear every time a CSP has many candidates (see also the Numbrix® and Hidato® examples in chapter 16).

---

[13] http://www.setbb.com/sudoku/viewtopic.php?t=2117&start=154&mforum=sudoku

```
. . . . C . . . . B . 1 8 . 6 .
1 5 . E . . 7 4 D . . . C . B .
. . B . . 9 . . . . 5 . . . . .
3 A . . . . G F 8 6 . . . . 4 E
9 F . . . . . 2 A E . . . . G .
. . 7 . . E . . . . 9 . . 3 . .
E . . G . . 1 B . 4 . . 7 . 8 .
. C . 8 6 . D . . . . F 5 . . .
. . . 6 E . . . . F . 9 1 . 2 .
G . . 3 . . 6 . . A . . D . . C
. . 5 . . 2 . . . . 6 . . 4 . .
A . . . . . . 5 7 8 . . . . . G
. 8 . . . . . . B 9 . . . . 5 .
. . 1 . . 7 . . . . E . . B . .
. E . B 8 . 2 . . C . . 4 . A 9
. 7 . D . . 9 . . 3 . A 6 . C .
```

**Figure 11.7.** *A 16×16 puzzle (from Tarek) with backdoor size = 5 and T&E-depth = 1*

***** SudoRules 16.2 based on CSP-Rules 1.2, config: W *****
91 givens, 871 candidates, 8498 csp-links and 8498 links. Initial density = 0.56
singles ==> r5c13 = 11, r12c14 = 6, r1c9 = 9
whip[1]: c7n11{r12 .} ==> r9c6 ≠ 11, r10c5 ≠ 11, r10c6 ≠ 11, r11c5 ≠ 11, r12c5 ≠ 11, r12c6 ≠ 11
whip[1]: r1n14{c7 .} ==> r3c8 ≠ 14, r3c7 ≠ 14
whip[2]: r2n6{c3 c6} – r2n8{c6 .} ==> r2c3 ≠ 2
whip[2]: r2n8{c3 c6} – r2n6{c6 .} ==> r2c3 ≠ 9
singles ==> r2c14 = 9, r4c13 = 2, r8c15 = 9, r7c5 = 9, r8c14 = 14, r13c13 = 14, r16c8 = 14, r1c7 = 14
whip[2]: c7n5{r6 r14} – c10n5{r14 .} ==> r6c5 ≠ 5
whip[2]: r4c11{n12 n7} – r4c12{n7 .} ==> r3c9 ≠ 12, r3c12 ≠ 12
whip[1]: b3n12{r4c12 .} ==> r4c4 ≠ 12, r4c3 ≠ 12
whip[2]: r4c11{n7 n12} – r4c12{n12 .} ==> r4c14 ≠ 7, r1c11 ≠ 7, r3c10 ≠ 7
singles ==> r8c10 = 7, r5c5 = 7
whip[2]: c10n1{r11 r6} – b5n1{r6c2 .} ==> r11c4 ≠ 1
whip[2]: r3c10{n16 n2} – r2c10{n2 .} ==> r14c10 ≠ 16, r11c10 ≠ 16, r6c10 ≠ 16
whip[1]: c10n16{r2 .} ==> r1c11 ≠ 16, r2c11 ≠ 16, r2c12 ≠ 16, r3c9 ≠ 16, r3c12 ≠ 16
whip[2]: r3c10{n2 n16} – r2c10{n16 .} ==> r14c10 ≠ 2, r6c10 ≠ 2
whip[1]: c10n2{r2 .} ==> r1c11 ≠ 2, r2c11 ≠ 2, r2c12 ≠ 2
whip[1]: r1n2{c1 .} ==> r3c1 ≠ 2, r3c2 ≠ 2, r3c4 ≠ 2
naked-single ==> r2c12 = 3
whip[1]: c10n2{r2 .} ==> r3c9 ≠ 2, r3c12 ≠ 2
whip[2]: r2c11{n15 n10} – r2c16{n10 .} ==> r2c3 ≠ 15
whip[2]: r2c11{n10 n15} – r2c16{n15 .} ==> r2c5 ≠ 10
singles ==> r2c5 = 2, r2c10 = 16, r3c10 = 2

whip[2]: r2c11{n10 n15} – r2c16{n15 .} ==> r2c6 ≠ 10
whip[2]: r4c11{n7 n12} – r4c12{n12 .} ==> r3c12 ≠ 7
whip[1]: b3n7{r4c12 .} ==> r4c4 ≠ 7
singles ==> r4c4 = 9,> r4c3 = 13, r14c2 = 9, r7c2 = 3
whip[1]: b5n13{r6c1 .} ==> r6c16 ≠ 13, r6c15 ≠ 13, r6c12 ≠ 13, r6c10 ≠ 13
whip[1]: b13n16{r16c3 .} ==> r1c3 ≠ 16
whip[2]: b9n9{r12c3 r10c3} – b9n14{r10c3 .} ==> r12c3 ≠ 15, r12c3 ≠ 12, r12c3 ≠ 2, r12c3 ≠ 4
whip[2]: b9n9{r10c3 r12c3} – b9n14{r12c3 .} ==> r10c3 ≠ 15, r10c3 ≠ 2, r10c3 ≠ 4, r10c3 ≠ 8
whip[2]: c4n5{r5 r14} – c10n5{r14 .} ==> r6c1 ≠ 5
whip[1]: c1n5{r16 .} ==> r14c4 ≠ 5
whip[3]: c7n5{r6 r14} – c10n5{r14 r6} – r7n5{c12 .} ==> r5c6 ≠ 5
whip[3]: r7n5{c12 c6} – c7n5{r6 r14} – c10n5{r14 .} ==> r5c12 ≠ 5, r6c9 ≠ 5, r6c12 ≠ 5
whip[3]: r6c10{n1 n5} – b5n5{r6c4 r5c4} – b5n1{r5c4 .} ==> r6c9 ≠ 1, r6c15 ≠ 1
singles ==> r6c15 = 15, r6c13 = 10, r7c6 = 15, r7c3 = 10, r3c15 = 1, r4c14 = 5, r9c16 = 5,
r11c16 = 11, r9c14 = 10, r10c14 = 8, r10c5 = 15, r1c6 = 5, r14c15 = 13, r14c10 = 5, r6c10 = 1,
r11c10 = 13, r11c2 = 1, r5c4 = 1, r8c16 = 1, r6c4 = 5, r5c7 = 5, r15c1 = 5, r16c5 = 5, r16c6 = 11,
r4c6 = 1, r10c6 = 4, r4c5 = 11
whip[1]: b6n4{r6c7 .} ==> r6c16 ≠ 4
singles ==> r5c16 = 4, r5c3 = 6, r2c3 = 8, r2c6 = 6, r3c2 = 6, r1c2 = 16
whips[1]: b6n4{r6c7 .} ==> r6c1 ≠ 4, r6c2 ≠ 4;   c2n4{r12 .} ==> r9c3 ≠ 4
naked-single ==> r9c3 = 12
whips[1]: c2n4{r12 .} ==> r9c1 ≠ 4, r12c4 ≠ 4;   c15n3{r11 .} ==> r12c13 ≠ 3, r11c13 ≠ 3
whip[1]: b12n15{r12c13 .} ==> r14c13 ≠ 15, r3c13 ≠ 15
whip[2]: b16n3{r14c16 r13c16} – r1n3{c16 .} ==> r14c8 ≠ 3
whip[4]: r10n7{c15 c8} – b10n9{r10c8 r11c8} – r11c13{n9 n15} – r11c4{n15 .} ==> r11c15 ≠ 7
hidden-single-in-a-block ==> r10c15 = 7
whip[4]: r12c15{n3 n14} – b9n14{r12c3 r10c3} – r10n9{c3 c8} – b10n1{r10c8 .} ==> r12c5 ≠ 3
whip[5]: r12n1{c11 c5} – r10c8{n1 n9} – r11n9{c8 c13} – b12n15{r11c13 r12c13} – r12c4{n15 .}
==> r12c11 ≠ 2
whip[5]: c6n8{r9 r5} – r6n8{c8 c12} – b7n11{r6c12 r8c11} – c1n11{r8 r6} – c1n13{r6 .} ==>
r9c1 ≠ 8
hidden-single-in-a-block ==> r11c1 = 8
whip[1]: b9n15{r12c4 .} ==> r14c4 ≠ 15, r13c4 ≠ 15, r1c4 ≠ 15, r3c4 ≠ 15
whip[3]: r11n7{c8 c4} – r11n15{c4 c13} – r11n9{c13 .} ==> r11c8 ≠ 16, r11c8 ≠ 12, r11c8 ≠ 10,
r11c8 ≠ 3
whip[6]: b11n5{r10c12  r10c9} – r10n14{c9  c3} – r12c3{n14 n9} – r12c13{n9 n15} – r12c4{n15 n2} – r10c2{n2 .} ==> r10c12 ≠ 11
whip[2]: b7n11{r6c12 r8c11} – r10n11{c11 .} ==> r6c2 ≠ 11
whip[1]: c2n11{r12 .} ==> r9c1 ≠ 11
whip[4]: b10n11{r9c7 r12c7} – c12n11{r12 r6} – r6n8{c12 c8} – b2n8{r3c8 .} ==> r9c7 ≠ 8
whip[4]: c6n8{r9 r5} – r6n8{c8 c12} – r6n11{c12 c1} – c1n13{r6 .} ==> r9c6 ≠ 13
whip[5]: r9n7{c8 c1} – c1n13{r9 r6} – r6n11{c1 c12} – r6n8{c12 c7} – b2n8{r3c7 .} ==> r9c8 ≠ 8
hidden-single-in-a-block ==> r9c6 = 8
whip[1]: b6n8{r6c8 .} ==> r6c12 ≠ 8
whip[4]: c8n12{r14 r6} – r6n8{c8 c7} – c7n4{r6 r13} – b14n15{r13c7 .} ==> r14c7 ≠ 12
whip[4]: c8n12{r14 r6} – r6n8{c8 c7} – c7n4{r6 r14} – b14n15{r14c7 .} ==> r13c7 ≠ 12

whip[7]: b10n16{r11c5 r9c8} – b6n16{r6c8 r8c6} – b6n10{r8c6 r8c8} – b6n3{r8c8 r5c6} – r15c6{n3 n13} – b10n13{r12c6 r12c5} – c5n1{r12 .} ==> r13c5 ≠ 16

whip[8]: r6n11{c12 c1} – c1n13{r6 r9} – r9n7{c1 c8} – b10n16{r9c8 r11c5} – r11c12{n16 n14} – r10n14{c12 c3} – r10n9{c3 c8} – r11c8{n9 .} ==> r6c12 ≠ 12

whip[8]: r10c2{n11 n2} – b11n2{r10c9 r12c12} – r12c4{n2 n15} – b12n15{r12c13 r11c13} – r11n9{c13 c8} – r10c8{n9 n1} – b11n1{r10c9 r12c11} – r12n4{c11 .} ==> r12c2 ≠ 11

**whip[9]: b8n2{r7c14 r6c16} – r6c2{n2 n13} – b9n13{r12c2 r9c1} – r9n7{c1 c8} – r11c8{n7 n9} – r10c8{n9 n1} – r10c11{n1 n11} – b7n11{r8c11 r6c12} – r6c1{n11 .} ==> r7c11 ≠ 2**

whip[3]: b8n13{r7c16 r5c14} – b8n12{r5c14 r7c14} – r7c11{n12 .} ==> r7c12 ≠ 13

whip[3]: r7c11{n12 n13} – b8n13{r7c16 r5c14} – b8n12{r5c14 .} ==> r7c12 ≠ 12, r7c9 ≠ 12

**whip[9]: r6n11{c12 c1} – c1n13{r6 r9} – b9n7{r9c1 r11c4} – b9n15{r11c4 r12c4} – r12n2{c4 c2} – r12n4{c2 c11} – r12n1{c11 c5} – r10c8{n1 n9} – r11c8{n9 .} ==> r12c12 ≠ 11**

singles ==> r6c12 = 11, r8c1 = 11, r8c3 = 4

whips[1]: r8n2{c11 .} ==> r7c12 ≠ 2, r7c9 ≠ 2;   r7n2{c16 .} ==> r6c16 ≠ 2

naked-single ==> r6c16 = 6

whip[1]: r8n2{c11 .} ==> r6c9 ≠ 2

whip[7]: r12n11{c11 c7} – r9c7{n11 n3} – b11n3{r9c11 r11c9} – c9n12{r11 r6} – b6n12{r6c7 r5c6} – r12c6{n12 n13} – r12c5{n13 .} ==> r12c11 ≠ 1

singles ==> r12c5 = 1, r10c8 = 9, r11c8 = 7, r11c4 = 15, r12c4 = 2, r10c2 = 11, r11c13 = 9, r12c13 = 15, r10c3 = 14, r12c3 = 9, r6c2 = 2, r6c1 = 13, r9c1 = 7

whip[5]: c6n16{r15 r8} – c6n10{r8 r13} – b14n12{r13c6 r14c8} – b14n6{r14c8 r15c8} – c8n1{r15 .} ==> r13c8 ≠ 16

whip[5]: c6n16{r15 r8} – c6n10{r8 r13} – b14n12{r13c6 r13c8} – b14n6{r13c8 r15c8} – c8n1{r15 .} ==> r14c8 ≠ 16

whip[5]: c8n1{r15 r13} – b14n6{r13c8 r14c8} – b14n12{r14c8 r13c6} – c6n16{r13 r8} – c6n10{r8 .} ==> r15c8 ≠ 16

whip[5]: c8n16{r6 r9} – b10n13{r9c8 r12c6} – r15c6{n13 n3} – b6n3{r5c6 r8c8} – b6n10{r8c8 .} ==> r8c6 ≠ 16

whip[1]: c6n16{r15 .} ==> r14c5 ≠ 16

whip[4]: b10n10{r11c7 r11c5} – c5n16{r11 r6} – b6n4{r6c5 r6c7} – c7n8{r6 .} ==> r3c7 ≠ 10

whip[5]: c9n3{r11 r8} – b7n2{r8c9 r8c11} – b7n16{r8c11 r6c9} – b11n16{r11c9 r11c12} – c5n16{r11 .} ==> r9c11 ≠ 3

whip[5]: r6c9{n12 n16} – c5n16{r6 r11} – r11c12{n16 n14} – r12c12{n14 n4} – r3c12{n4 .} ==> r5c12 ≠ 12, r11c9 ≠ 12

singles ==> r6c9 = 12, r7c11 = 13, r5c12 = 8, r5c11 = 3, r5c6 = 12, r5c14 = 13, r7c16 = 2, r7c14 = 12, r16c11 = 8, r16c16 = 15, r2c16 = 10, r2c11 = 15, r1c11 = 10, r14c16 = 8

whip[1]: b3n4{r3c12 .} ==> r3c4 ≠ 4, r3c1 ≠ 4

whip[1]: b7n16{r8c11 .} ==> r8c8 ≠ 16

whip[2]: r1n3{c16 c8} – r1n13{c8 .} ==> r1c16 ≠ 7

whip[3]: c8n1{r13 r15} – b14n6{r15c8 r14c8} – b14n12{r14c8 .} ==> r13c8 ≠ 13, r13c8 ≠ 10, r13c8 ≠ 3

whip[3]: c8n1{r15 r13} – b14n6{r13c8 r14c8} – b14n12{r14c8 .} ==> r15c8 ≠ 13, r15c8 ≠ 3

whip[3]: b14n12{r14c8 r13c8} – b14n1{r13c8 r15c8} – b14n6{r15c8 .} ==> r14c8 ≠ 10

whip[3]: r12c6{n13 n3} – b6n3{r8c6 r8c8} – r1c8{n3 .} ==> r9c8 ≠ 13

singles ==> r12c6 = 13, r12c2 = 4, r9c2 = 13, r13c5 = 13, r15c12 = 13

whip[3]: r15c6{n16 n3} – r14n3{c5 c13} – r14n16{c13 .} ==> r15c11 ≠ 16, r15c9 ≠ 16

whip[4]: r1n7{c14 c4} – b1n4{r1c4 r1c1} – r16c1{n4 n2} – b16n2{r16c14 .} ==> r13c14 ≠ 7
whip[4]: b2n10{r3c5 r3c8} – b2n8{r3c8 r3c7} – r6c7{n8 n4} – b14n4{r14c7 .} ==> r14c5 ≠ 10
whip[3]: b14n4{r14c7 r13c7} – b14n15{r13c7 r14c7} – r14n10{c7 .} ==> r14c4 ≠ 4
whip[4]: r6n4{c7 c5} – r14c5{n4 n3} – c13n3{r14 r3} – r3c7{n3 .} ==> r6c7 ≠ 8
singles ==> r6c7 = 4, r6c5 = 16, r6c8 = 8, r3c7 = 8, r9c8 = 16, r14c5 = 4
whip[5]: r14n16{c12 c13} – c14n16{r16 r3} – c14n15{r3 r1} – r1c3{n15 n2} – r16c3{n2 .} ==> r16c9 ≠ 16
whip[5]: c11n16{r13 r8} – r8c9{n16 n2} – r16c9{n2 n1} – b11n1{r10c9 r10c11} – c11n2{r10 .} ==> r13c11 ≠ 4
singles ==> r9c11 = 4, r9c9 = 3, r9c7 = 11, r12c11 = 11, r4c11 = 12, r4c12 = 7
whip[3]: r14n16{c12 c13} – b16n3{r14c13 r13c16} – r13n7{c16 .} ==> r13c11 ≠ 16
singles ==> r8c11 = 16, r8c9 = 2
whip[3]: b16n2{r13c14 r16c14} – r16c1{n2 n4} – b15n4{r16c9 .} ==> r13c12 ≠ 2
whip[4]: b16n2{r13c14 r16c14} – r16n1{c14 c9} – b11n1{r10c9 r10c11} – c11n2{r10 .} ==> r13c3 ≠ 2, r13c1 ≠ 2
whip[4]: r15c11{n1 n7} – r13c11{n7 n2} – b16n2{r13c14 r16c14} – r16n1{c14 .} ==> r15c9 ≠ 1
whip[4]: b15n4{r13c12 r16c9} – c9n1{r16 r10} – b11n5{r10c9 r10c12} – r7c12{n5 .} ==> r13c12 ≠ 6
whip[3]: c12n2{r14 r10} – c12n5{r10 r7} – c12n6{r7 .} ==> r14c12 ≠ 16
whip[3]: r15n15{c3 c9} – r15n6{c9 c8} – r13n6{c8 .} ==> r13c1 ≠ 15
whip[4]: r15n15{c9 c3} – b13n3{r15c3 r13c3} – b16n3{r13c16 r14c13 r14n16{c13 .} ==> r14c9 ≠ 15
singles to the end

# 12. Patterns of proof and associated classifications

Until now, our approach has been based on the resolution paradigm introduced in section 1.2 and formalised in chapter 4. The fundamental confluence property of the various T-braid theories and the "T&E(T) vs T-braids" theorems together show that this paradigm applies well, *in theory*, to instances in T&E(1) or T&E(2): in each case, a "simplest first" strategy and a universal rating [respectively B or BB] can be defined. Moreover, the many concrete examples provided in this book show that, for instances in T&E(1) or gT&E(1), it also applies well *in practice* to CSP's as varied as Sudoku, N-Queens, Futoshiki, Kakuro, Numbrix® and Hidato® (for the latter, see chapters 14 to 16). For small values of p, this practical aspect could be extended to instances in T&E($S_p$).

Now, for the general instances in T&E(2) and beyond gT&E(1), even though the paradigm still applies in theory, resolution paths for most of them require B-braids (even S-braids are generally not enough). But the structure of B-braids is rather complex, as it relies on "contextual" indirect contradictions between two candidates. Considering our general readability requirement, one may legitimately wonder whether a solution based on such patterns can satisfy it. The first two sections below will introduce apparently simpler patterns; they will finally be shown to have the same resolution power, but they correspond to very different views of resolution.

Beyond the technicalities associated with these new extended whip and braid patterns, our main point here is thus of a more epistemological nature. Re-assessing our initial resolution paradigm, it revolves around the notion of a *pattern of proof*.

Because the fuzzy borderline of complexity that may compel us to switch from the extreme requirement of finding the "simplest" solution to the relaxed one of finding a "readable" solution seems to cut through the T&E(2) land, this chapter starts by exploring various ways of dealing with such instances. Even the possibility of defining the notion of a "simplest solution" becomes much more questionable in T&E(2). Moreover, beyond gT&E(1), the simplicity and the understandability requirements may be at strong variance; we thus discuss various options for their interpretation. In particular, we show that "equivalent" structures (e.g. B*-braids vs B-braids) can correspond to viewpoints that lead to very different classifications. Finally, we show that a pure logic approach along the same lines as above, is still possible in theory, even for instances beyond T&E(2), although it may require some extensions to our initial resolution paradigm and the computational complexity may be much higher, depending on which *patterns of proof* one is willing to accept.

### 12.1. Bi-whips, bi-braids, confluence and bi-T&E

From a purely abstract logical point of view, given a single-solution instance of a CSP, if a candidate for some CSP variable is not its final value, then it is contradictory with anything, including itself. As shown by the various resolution theories we had to define, this does not imply that it can be proven to be contradictory either "easily" or in an "understandable" way or in a "constructive" way or in a "pattern-based" way. When one deals with T&E(2) instances, one has to consider contradictions arising from pairs of candidates in addition to contradictions arising from a single candidate; but the previous remarks also apply to pairs of candidates; as a result, when saying that two candidates are incompatible, one should always specify how this incompatibility is supposed to be proven.

In order to catch by constructive logical patterns some types of indirect pairwise contradictions between candidates, we shall first introduce the concepts of a bi-whip and a bi-braid. For each integer n≥1, the ***bi-whip[n] and bi-braid[n] incompatibility relations between candidates $Z_1$ and $Z_2$ will be two different constructive restricted forms of the abstract logical $nand_2$ predicate defined by:***
$$nand_2(Z_1, Z_2) \equiv \neg[candidate(Z_1) \wedge candidate(Z_2)].$$

As should now be expected from all the generalised whips and braids we have met in the previous chapters, we shall show that bi-braids have a smooth theory (one can prove a form of stability for confluence and a "bi-braid vs bi-T&E" theorem) and bi-whips are a structurally nicer and easier to compute (hopefully good) logical approximation of bi-braids. In the next section, we shall see how each of these patterns can be used to define new extended whip or braid patterns (namely W*-whips and B*-braids) that have a significantly simpler structure than W-whips and B-braids, although they are also based on indirect pairwise contradictions.

#### 12.1.1. Definition of bi-whips and bi-braids

Apart from being based on two candidates instead of one, bi-whips and bi-braids are very much like whips and braids, respectively. But, instead of leading to eliminations, they prove contradictions between pairs of candidates.

Definition: given a resolution state RS and two different candidates $Z_1$ and $Z_2$ in RS that are not linked, for any n≥1, a bi-whip[n] built on $Z_1$ and $Z_2$ is a structured list ($\{Z_1, Z_2\}$, $(V_1, L_1, R_1)$, …, $(V_{n-1}, L_{n-1}, R_{n-1})$, $(V_n, L_n)$), such that:
– for any 1≤k≤n, $V_k$ is a CSP variable;
– $Z_1$, $Z_2$, all the $L_k$'s and all the $R_k$'s are candidates in RS;

– in the sequence ($L_1$, $R_1$, …, $L_{n-1}$, $R_{n-1}$, $L_n$), any two consecutive elements are different;

– $Z_1$ and $Z_2$ do not belong to $\{L_1, R_1, L_2, R_2, …. L_n\}$;

– $L_1$ is linked to $Z_1$ or $Z_2$;

– right-to-left continuity: for any $1<k\leq n$, $L_k$ is linked to $R_{k-1}$;

– strong left-to-right continuity: for any $1\leq k<n$, $L_k$ and $R_k$ are candidates for $V_k$;

– $L_n$ is a candidate for $V_n$;

– at least one of $Z_1$ and $Z_2$ is not a label for $V_n$;

– for any $1\leq k<n$, $R_k$ is the only candidate for $V_k$ compatible with $Z_1$, $Z_2$ and all the previous $R_i$ ($i<k$);

– $V_n$ has no candidate compatible with $Z_1$, $Z_2$ and all the previous $R_i$ ($i<n$); (but $V_n$ has more than one candidate – the usual non-degeneracy condition).

Definition: a bi-braid[n] built on $Z_1$ and $Z_2$ is a structured list as above, with the right-to-left continuity condition replaced by:

– for any $1<k\leq n$, $L_k$ is linked to $Z_1$ or $Z_2$ or a previous $R_i$.

Definitions: given a resolution state RS and $n \geq 1$, two different candidates $Z_1$ and $Z_2$ in RS that are not linked are said *bi-whip[n]* (respectively *bi-braid[n]*) *incompatible or contradictory in RS* if there is in RS some bi-whip[n] (resp. some bi-braid[n]) built on $Z_1$ and $Z_2$. $Z_1$ and $Z_2$ are said *bi-whip* (respectively *bi-braid*) *incompatible or contradictory in RS* if they are bi-whip[n] (resp. bi-braid[n]) incompatible in RS for some $n \geq 1$. We also say that they are bi-whip (resp. bi-braid) contradictory.

Remarks:

– in order to avoid confusion with B-braids, the "bi" in "bi-whip" and "bi-braid" should be pronounced [ai] as in "bye bye";

– in a bi-whip[n] or a bi-braid[n], n is called the length; notice that, according to the above definitions, as was the case with those for whips or braids and all their generalisations, no initial structured strict sublist ($\{Z_1, Z_2\}$, ($V_1, L_1, R_1$), …, ($V_{n-1}, L_{k-1}, R_{k-1}$), ($V_k, L_k$)) of ($\{Z_1, Z_2\}$, ($V_1, L_1, R_1$), …, ($V_{n-1}, L_{n-1}, R_{n-1}$), ($V_n, L_n$)), with $k < n$, is a bi-whip[n] or a bi-braid[n] based on $Z_1$ and $Z_2$ in RS; this is our usual non-degeneracy condition for whip-like or braid-like structures;

– we use the same terminology of z- and t- candidates as for whips; here, a z-candidate is a candidate linked to (at least) one of $Z_1$ and $Z_2$;

– in any CSP, for any bi-braid[2] built on $Z_1$ and $Z_2$ there is a bi-whip[2] built on $Z_1$ and $Z_2$ (the proof is similar to that for ordinary braids – see theorem 5.5);

– if, in a resolution state RS, W is a partial whip [respectively a partial braid] based on Z and C is a left-linking or a t-candidate of W, then Z and C are obviously bi-whip [resp. bi-braid] incompatible in RS, unless C is linked to Z: the final CSP variable through which the bi-whip [resp. bi-braid] contradiction is made explicit is the first $V_k$ such that C is linked to $R_k$;

– if the present definitions were extended to the case $Z_1 = Z_2$, a bi-whip [resp. a bi-braid] built on $Z_1$ and $Z_2$ would merely be a whip [resp. a braid] of same length built on $Z_1$; as usual in our approach, we exclude this case because it is degenerated.

### 12.1.2. Bi-whips[1] in Sudoku

In Sudoku, a typical bi-whip[1] contradiction occurs between $Z_1$ and $Z_2$ when, in a row r, number n appears as a candidate in only two blocks $b_1$ and $b_2$ and when $Z_1$ [respectively $Z_2$] is a candidate for number n in an rc-cell situated in $b_1$ [resp. $b_2$] but not in r (as in the leftmost part of Figure 12.1); as usual, the role of rows and blocks can be permuted (as in the rightmost part of Figure 12.1); and rows can be replaced by columns.

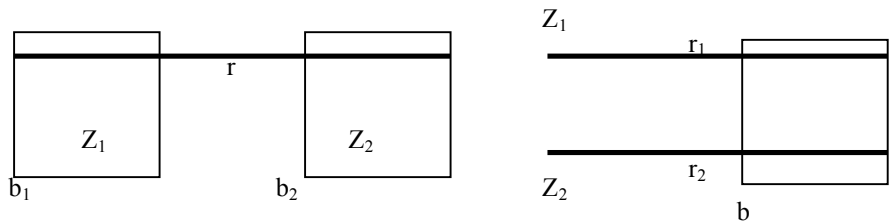

*Figure 12.1. Two typical types of bi-whips[1] in Sudoku*

But there are other possibilities: when, in a row r, number n is a candidate in only two columns $c_1$ and $c_2$ and $Z_1$ and $Z_2$ are candidates for n, respectively in $c_1$ and $c_2$, anywhere outside r; or when a 2D cell is bivalue and each of $Z_1$ and $Z_2$ is linked to a different candidate for this cell. See section 12.3.5 for more examples.

### 12.1.3. Definition of the bi-braid logical theories

Let us now define an increasing sequence of logical theories for bi-braids. Notice that we speak of logical theories, not of resolution theories; only later (in section 12.2) will bi-braids be used in resolution theories.

We first define, for each n ≥ 1, an auxiliary predicate:
bi-braid[n]($z_1$, $z_2$, $l_1$, $r_1$, $V_1$, $l_2$, $r_2$, $V_2$, …, $l_{n-1}$, $r_{n-1}$, $V_{n-1}$, $l_n$, $V_n$), with signature (label, label, [label, label, CSP-variable]$^{n-1}$, label, CSP-Variable). Formally, it is the description of a bi-braid[n] in terms of candidates and links, in the style of what was written for whips in section 5.2.2.

We also introduce a new constant "bi-braid" in the domain of Constraint or Constraint-Type (depending on which modelling choice has been made for the CSP under consideration). We correlatively allow predicate "linked-by" to accept this constant as its last argument. Predicate "linked" is still obtained from "linked-by" by existentially quantifying its last argument (of type Constraint or Constraint-Type). Notice that, contrary to all the links considered until now, links of type bi-braid are not structural, they may be dynamically created, but this will not be a problem in the sequel, because they are persistent.

We can now define BRT*(CSP) as the usual BRT(CSP), but with the domain of its Constraint or Constraint-Type extended as described above; as a result, its ECP rule is naturally extended into ECP* so as to take care of the corresponding new constraints; compared with BRT, it can be described as containing the additional rule:

**ECP-bi-braid:** $\forall_{\neq} l_1, l_2 \{[\text{value}(l_1) \wedge \text{linked}(l_1, l_2, \text{bi-braid})] \Rightarrow \neg \text{candidate}(l_2)\}$.

Definition: a *bi-braid[n] contradiction rule* is a formula in the "condition => action" form, where "condition" is a bi-braid[n] for two different candidates $Z_1$ and $Z_2$ and "action" is the assertion of ground atomic formula "linked-by($Z_1$, $Z_2$, bi-braid)".

Definition: for any $n \geq 0$, let $biB_n$ be the following logical theory:
- $biB_0$ = BRT*(CSP);
- $biB_1 = biB_0 \cup \{$ bi-braid[1] contradiction rules $\}$,
- $biB_2 = biB_1 \cup \{$ bi-braid[2] contradiction rules $\}$,
- ....
- $biB_n = biB_{n-1} \cup \{$ bi-braid[n] contradiction rules $\}$,
- $biB_\infty = \cup_{n \geq 0} biB_n$.

Notice that, in and of itself, none of the $biB_n$ theories allows any elimination that could not be done in the original BRT(CSP).

### 12.1.4. Stability for confluence of the bi-braid logical theories

We have not defined the $biB_n$ as resolution theories (as such they would be no more than BRT(CSP)), but the definition (in section 4.5) of stability for confluence can be extended to them as follows.

Definition: a logical bi-braid theory T is *stable for confluence* if, for any instance P of the CSP, for any resolution state $RS_1$ of P and for any rule R in T applicable in state $RS_1$ for asserting a bi-braid contradiction for two different candidates $Z_1$ and $Z_2$ – in the form of a "linked-by($Z_1$, $Z_2$, bi-braid)" ground atomic formula – , if any set Y of consistency preserving assertions and/or eliminations is done before R is applied, leading to a resolution state $RS_2$, and if it destroys the pattern of R (R can therefore no longer be applied for asserting a bi-braid link between $Z_1$ and $Z_2$), then there always exists a sequence of rules in T that will allow the assertion of a bi-braid link between $Z_1$ and $Z_2$ (possibly based on a shorter bi-braid).

The following theorem is mainly a preamble to the proof of the confluence property of the $B*_pB_m$ resolution theories defined in section 12.2.

***Theorem 12.1: for any $0 \leq n \leq \infty$, bi-braid theory $biB_n$ is stable for confluence.***

Proof: the following proof is a straightforward adaptation of that of theorem 5.6, with only very slight changes.

Let $n<\infty$ be fixed (the case $n=\infty$ is an obvious corollary to all the cases $n<\infty$). We shall show that, if there is a bi-braid B of length $m \leq n$ built on $Z_1$ and $Z_2$ in some resolution state $RS_1$, then, for any further resolution state $RS_2$ obtained from $RS_1$ by consistency preserving assertions and eliminations, in the resolution state $RS_3$ obtained from $RS_2$ by applying all the rules in BRT until quiescence, if both $Z_1$ and $Z_2$ are still candidates in $RS_3$, there will always be in $RS_3$ a bi-braid of length $m' \leq m$ built on $Z_1$ and $Z_2$.

Let B be: $\{L_1\ R_1\} - \{L_2\ R_2\} - \ldots - \{L_p\ R_p\} - \{L_{p+1}\ R_{p+1}\} - \ldots - \{L_m\ .\}$.

First notice that, as BRT has the confluence property (theorem 4.1), state $RS_3$ is uniquely defined, independently of the way we apply the rules in BRT.

If any of $Z_1$ or $Z_2$ has been eliminated or asserted in $RS_3$, there remains nothing to prove. Otherwise, we must consider all the elementary events related to B that can have happened between $RS_1$ and $RS_3$ (all the possibilities are marked by the same letter as in the proof of theorem 5.6). For this, we start from B' = what remains of B in $RS_3$. At this point, B' may not be a bi-braid in $RS_3$. We repeat the following procedure, for $p = 1$ to $p = m$, producing in the end a new (possibly shorter) bi-braid B' built on $Z_1$ and $Z_2$ in $RS_3$. All the references below are to the current B'.

a) If, in $RS_3$, the left-linking or any t- or z- candidate of CSP variable $V_p$ has been asserted, then $Z_1$ or $Z_2$ and/or the previous $R_k$('s) to which $L_p$ is linked must have been eliminated by ECP in the passage from $RS_2$ to $RS_3$ (if it was not yet eliminated in $RS_2$); if $Z_1$ or $Z_2$ is among these eliminations, there remains nothing to prove; otherwise, the procedure has already been successfully terminated by case f of the first such k.

b) If, in $RS_3$, left-linking candidate $L_p$ has been eliminated (but not asserted) (it can therefore no longer be used as a left-linking candidate in a bi-braid) and if CSP variable $V_p$ still has a z- or a t- candidate $C_p$, then replace $L_p$ by $C_p$; now, up to $C_p$, B' is a partial bi-braid built on $Z_1$ and $Z_2$ in $RS_3$. Notice that, even if $L_p$ was linked to $R_{p-1}$ (as it would if B was a bi-whip), this may not be the case for $C_p$; therefore trying to prove a similar theorem for bi-whips would fail here, as in the whips case.

c) If, in $RS_3$, any t- or z- candidate of $V_p$ has been eliminated (but not asserted), this has not changed the basic structure of B (at stage p). Continue with the same B'.

d) If, in $RS_3$, right-linking candidate $R_p$ has been asserted (p can therefore not be the last index of B'), it can no longer be used as an element of a bi-braid, because it is no longer a candidate. Notice that all the left-linking and t- candidates for CSP variables of B after p that were incompatible in B with $R_p$, i.e. linked to it, if still present in $RS_2$, must have been eliminated by ECP somewhere between $RS_2$ and $RS_3$. But, considering the bi-braid structure of B upwards from p, more eliminations and assertions must have been done by rules from BRT between $RS_2$ and $RS_3$.

Let q be the smallest number strictly greater than p such that, in $RS_3$, CSP variable $V_q$ still has a (left-linking, t- or z-) candidate $C_q$ that is not linked to any of the $R_i$ for $p \leq i < q$ (by definition of a bi-braid, $C_q$ is therefore linked to $Z_1$ or to $Z_2$ or to some $R_i$ with $i < p$). Between $RS_2$ and $RS_3$, the following rules from BRT must have been applied for each of the CSP variables $V_u$ of B with index u increasing from p+1 to q-1 included: eliminate its left-linking candidate ($L_u$) by ECP, assert its right-linking candidate ($R_u$) by S, eliminate by ECP all the left-linking and t- candidates for CSP variables after u that were incompatible in B with the newly asserted candidate ($R_u$).

In $RS_3$, excise from B' the part related to CSP variables p to q-1 (included) and (if $L_q$ has been eliminated in the passage from $RS_1$ to $RS_3$) replace $L_q$ by $C_q$; for each integer $s \geq p$, decrease by q-p the index of CSP variable $V_s$ and of its candidates in B'; in $RS_3$, B' is now, up to p (the ex q), a partial bi-braid in $B_n$ built on $Z_1$ and $Z_2$.

e) If, in $RS_3$, left-linking candidate $L_p$ has been eliminated (but not asserted) and if CSP variable $V_p$ has no t- or z- candidate in $RS_3$ (complementary to case b), then $V_p$ has only one possible value in $RS_3$, namely $R_p$; $R_p$ must therefore have been asserted by S somewhere between $RS_1$ and $RS_3$; this case has therefore been dealt with by case d (because the assertion of $R_p$ also entails the elimination of $L_p$).

f) If, in $RS_3$, right-linking candidate $R_p$ of B has been eliminated (but not asserted), in which case p cannot be the last index of B', then replace B' by its initial part: $\{L_1 R_1\} - \{L_2 R_2\} - \ldots - \{L_p .\}$. At this stage, B' is in $RS_3$ a shorter bi-braid built on $Z_1$ and $Z_2$. Return B' and stop.

Notice that, as was the case for ordinary braids, for the bi-braid thus obtained, its sequence of CSP variables is a sub-sequence W' of those of B, its right-linking candidates are those of B belonging to the sub-sequence W', its left-linking candidates are those of B belonging to the sub-sequence W', each of them possibly replaced by a t-candidate of B for the same CSP variable.

### 12.1.5. Definition of the bi-T&E($Z_1$, $Z_2$, RS) procedure

Definition: given a resolution state RS and two different non-linked candidates $Z_1$ and $Z_2$ in RS, *bi-T&E($Z_1$, $Z_2$, RS) or bi-Trial-and-Error($Z_1$, $Z_2$, RS)* is the following procedure:
- make a copy RS' of RS; in RS', delete $Z_1$ and $Z_2$ as candidates and assert them as values;
- in RS', apply repeatedly all the rules in BRT(CSP) until quiescence;
- if RS' has become a contradictory state (detected by axiom CD), then assert linked-by($Z_1$, $Z_2$, bi-braid) in RS (*sic*: in RS, not in RS').

Remarks:

– this definition is meaningful only because BRT(CSP) has the confluence property for any CSP: otherwise, the result of "applying repeatedly in RS' all the rules in BRT until quiescence" may not be uniquely defined;

– if we extended this definition to the degenerated case $Z_1 = Z_2$, bi-T&E($Z_1$, $Z_1$, RS) would assert "linked-by($Z_1$, $Z_1$, bi-braid)" in RS if and only if T&E($Z_1$, RS) eliminates $Z_1$ from RS;

– in case $Z_1 \neq Z_2$, but $Z_1$ would be eliminated by T&E($Z_1$, RS) or $Z_2$ by T&E($Z_2$, RS), then bi-T&E($Z_1$, $Z_2$, RS) would assert linked-by($Z_1$, $Z_2$, bi-braid); in the sequel, we shall avoid such situations by systematically applying T&E before bi-T&E.

### 12.1.6. The bi-T&E($Z_1$, $Z_2$, RS) procedure vs bi-braid incompatibilities

It is obvious that, for any bi-braid incompatibility obtained via a bi-braid B built on two different candidates $Z_1$ and $Z_2$ in some resolution state RS, procedure bi-T&E($Z_1$, $Z_2$, RS) will assert "linked-by($Z_1$, $Z_2$, bi-braid)" in RS; this can easily be seen by applying in RS' a sequence of rules from BRT following the bi-braid structure of B. The converse is more interesting.

***Theorem 12.2 ("bi-T&E vs bi-braid"): for any instance of any CSP, for any resolution state RS and for any pair of different non-linked candidates $Z_1$ and $Z_2$ in RS, if bi-T&E($Z_1$, $Z_2$, RS) asserts "linked-by($Z_1$, $Z_2$, bi-braid)" in RS, then there is in RS a bi-braid (of undefined length) built on $Z_1$ and $Z_2$.***

Proof: it is a straightforward adaptation of the corresponding proof for braids (theorem 5.7).

Let RS' be the auxiliary resolution state used by bi-T&E($Z_1$, $Z_2$, RS). Following the steps of BRT in RS', we progressively define a bi-braid in RS built on $Z_1$ and $Z_2$. First, remember that BRT contains three types of rules: ECP (which eliminates candidates), S (which asserts a value for a CSP variable) and CD (which detects a contradiction on a CSP variable).

Consider the first step of BRT in RS' that is an application of rule S, asserting some label $R_1$ as a value. As $R_1$ was not a value in RS, there must have been in RS' some elimination of a candidate, say $L_1$, for a CSP variable $V_1$ of which $R_1$ is a candidate, and the elimination of $L_1$ (which made the assertion of $R_1$ by S possible in RS') can only have been made possible in RS' by the assertion of $Z_1$ and $Z_2$. But if $L_1$ has been eliminated in RS', it can only be by ECP and because it is linked to $Z_1$ or $Z_2$. Then {$L_1$ $R_1$} is the first pair of candidates of our bi-braid in RS and $V_1$ is its first CSP variable. (Notice that there may be other z-candidates for $V_1$, but this is pointless, we can choose any of them as $L_1$ and consider the remaining ones as z-candidates).

The sequel is done by recursion. Suppose we have built a bi-braid in RS corresponding to the part of the BRT resolution in RS' up to its k-th assertion step. Let $R_{k+1}$ be the next candidate asserted by BRT in RS'. As $R_{k+1}$ was not a value in RS, there must have been in RS' some elimination of a candidate, say $L_{k+1}$, for a CSP variable $V_{k+1}$ of which $R_{k+1}$ is a candidate, and the elimination of $L_{k+1}$ (which made the assertion of $R_{k+1}$ possible in RS') can only have been made possible in RS' by the assertion of $Z_1$ and/or $Z_2$ and/or of some of the previous $R_i$. But if $L_{k+1}$ has been eliminated in RS', it can only be by ECP and because it is linked to $Z_1$ or $Z_2$ or to some of the previous $R_i$, say C. Then our partial bi-braid in RS can be extended to a longer one, with {$L_{k+1}$ $R_{k+1}$} added to its candidates, $L_{k+1}$ linked to C, and $V_{k+1}$ added to its sequence of CSP variables.

End of the procedure: a contradiction is supposed to be obtained by BRT in RS'. As, in BRT, only ECP can eliminate a candidate, a contradiction is obtained if a value asserted in RS', i.e. if $Z_1$ or $Z_2$ or one of the $R_i$, i<n, eliminates in RS' (via ECP) a candidate, say $L_n$, that was the last one for a corresponding variable $V_n$ and that is linked to $Z_1$ or $Z_2$ or one of the $R_i$, i<n. $L_n$ and $V_n$ are thus the last left-linking candidate and CSP variable of the bi-braid we were looking for in RS.

Nothing can guarantee that both $Z_1$ and $Z_2$ have effectively been used in this construction, but this is not a problem: if one of them has not, it only means that we are in the special case mentioned in the second remark at the end of section 12.1.5.

## 12.2. $W^*_p$-whips and $B^*_p$-braids

We now introduce $W^*_p$-whips and $B^*_p$-braids. Basically, they are like ordinary whips or braids in which indirect, but non-contextual (contrary to W-whips and B-braids), bi-whip or bi-braid contradictions between candidates are allowed wherever direct links can appear in whips or braids. We shall show that every B*-braid can be considered as a B-braid and conversely (surprisingly) – if we forget any notion of length. We shall also define associated resolution theories and show that they have the confluence property, thereby allowing to define a B*B rating. It is interesting to notice that different views of "equivalent" structures (B-braids vs B*-braids) lead to very different ratings and classifications.

### 12.2.1. Definition of $W^*_p$-whips and $B^*_p$-braids

Definition: given a resolution state RS of any CSP, an integer p with $1 \leq p \leq \infty$, an integer $m \geq 1$ and a candidate Z in RS, a $W^*_p$-whip[m] built on Z is a structured list (Z, ($V_1$, $L_1$, $R_1$), …, ($V_{m-1}$, $L_{m-1}$, $R_{m-1}$), ($V_m$, $L_m$)) that satisfies the following conditions:

– for any $1 \leq k \leq m$, $V_k$ is a CSP variable;

– Z, all the $L_k$'s and all the $R_k$'s are candidates;

– in the sequence of labels ($L_1$, $R_1$, …, $L_{m-1}$, $R_{m-1}$, $L_m$), any two consecutive elements are different;

– Z does not belong to $\{L_1, R_1, L_2, R_2, …. L_m\}$;

– either $L_1$ is linked to Z or $L_1$ and Z are bi-whip[p'] incompatible *in RS* for some $p' \leq p$;

– extended right-to-left continuity: for any $1 < k \leq m$, either $L_k$ is linked to $R_{k-1}$ or $L_k$ and $R_{k-1}$ are bi-whip[p'] incompatible *in RS* for some $p' \leq p$;

– strong left-to-right continuity: for any $1 \leq k < m$, $L_k$ and $R_k$ are candidates for $V_k$;

– Z is not a label for $V_m$;

– for any $1 \leq k < m$: $R_k$ is the only candidate for $V_k$ that is compatible and not bi-whip[p'] incompatible *in RS* for any $p' \leq p$ with Z and with all the previous right-linking candidates $R_i$;

– $V_m$ has no candidate compatible and not bi-whip[p'] incompatible *in RS* for any $p' \leq p$ with Z and with all the previous right-linking candidates (but $V_m$ has more than one candidate – our usual non-degeneracy condition of the global structure being defined).

Definition: a *B\*-braid[m]* is a structured list as above, with "bi-whip [in]compatible" replaced everywhere by "bi-braid [in]compatible" and with the extended right-to-left continuity condition replaced by:

– for any $1 \leq k \leq m$, $L_k$ is linked to $Z_1$ or to $Z_2$ or to a previous $R_i$ or $L_k$ is bi-braid[p'] incompatible *in RS* for some $p' \leq p$ with $Z_1$ or with $Z_2$ or with a previous $R_i$.

In both cases, Z is called the target and m the pseudo-length ("pseudo" because it does not take into account the lengths of the inner bi-whip/bi-braid contradictions). If $p = \infty$, we discard as usual the p index and we write W*-whip and B*-braid.

The case m = 1 is worth some comment. Recalling our definition of forcing-whips in section 5.9, condition patterns of rules in $W^*_pW_1$ (respectively $B^*_pB_1$) could also be named *forcing-bi-whips* (resp. *forcing-bi-braids*), because the target Z of a W*$_p$-whip[1] (resp. a B*$_p$-braid[1] ) is bi-whip (resp. bi-braid) incompatible with all the candidates of CSP variable $V_1$: using the symmetry of bi-whips (resp. bi-braids) with respect to their $\{Z_1, Z_2\}$ pair and looking backwards towards Z from the first CSP variable $V_1$, all the candidates for $V_1$ contradict Z.

***Theorem 12.3 (W\*-whip and B\*-braid elimination theorem): given a W\*-whip [respectively a B\*-braid] built on candidate Z, its target Z can be eliminated.***

Proof: obvious.

Figure 12.2 is a graphico-symbolic representation of a W*-whip[4] built on Z. Vertical lines represent the sequence of its CSP variables, from left to right; all the other lines represent direct links or bi-whip contradictions (undifferentiated); curved ones represent distant contradictions (corresponding to global z- and t- candidates); a candidate for a CSP variable can only exist at an endpoint of another line.

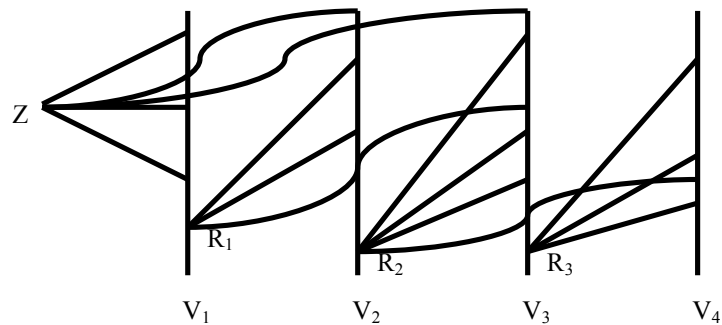

***Figure 12.2.*** *A graphico-symbolic representation of a W\*-whip[4] built on Z*

By comparing the definitions or the graphico-symbolic representations, it appears that B*$_p$-braids are a particular case of B$_p$-braids and W*-whips are a particular case of W*-whips (they are even the same pattern, length notwithstanding, in the m=0 case, if we extend the above definition to this case).

The main difference between a B$_p$-braid and a B*$_p$-braid (still apart from not taking into account the lengths of the inner structures) is that *the inner bi-braids of a B\*$_p$-braid rely on no global z- or t- candidates, contrary to the inner braids used in a B$_p$-braid. Said otherwise, the indirect contradictions between two candidates appearing in a B-braid may be contextual (they may depend on Z and the previous right-linking candidates), while they may not in a B\*-braid*. It makes the structure of a B*-braid apparently much simpler. However, there is a surprising "equivalence".

***Theorem 12.4 (pseudo-equivalence of B\*-braids and B-braids):*** *in any resolution state RS of any CSP, for any candidate Z in RS, if there is in RS a B\*-braid based on Z, then there is also in RS a B-braid based on Z. Conversely, if there is in RS a B-braid[m] based on Z, then there is in RS a B\*$_{m-1}$-braid[1], i.e. a forcing bi-braid[m-1], based on Z.*

Proof: the first part is a result of the preceding remarks. As for the converse, consider a B-braid[m] B with target Z. It is easy to see that, for any global left-linking or t- candidate C of B, the part of B before C is a bi-braid of length less than m based on Z and C. Taking all such bi-braids for all the candidates of the last CSP variable of B, we get the desired B\*-braid[1] (or forcing bi-braid).

Remarks:

– although the first part of the theorem remains true when we replace B\*-braids by W\*-whips and B-braids by W-whips, the converse does not;

– in our view, in spite of this theorem, B\*-braids and B-braids cannot be considered as the "same" pattern: the notion of length [or pseudo-length] that we have always introduced at the same time as each of our patterns is an integral and an essential part of their definition;

– as all the (known) Sudoku puzzles are in T&E(2) or less, they can all be solved by forcing bi-braids; but this is not a very interesting result, as it gives no indication on the maximum length of the necessary bi-braids;

– the smallest p such that an instance Q is in B\*$_p$B may be much larger than the smallest p such that Q is in B$_p$B;

– in the Sudoku community, vague notions of a "contradiction chain" and a "nested contradiction chain" have been in existence for a long time (with vaguely contradictory variants of each). As far as any precise interpretation can be given (based e.g. on the outputs of Sudoku Explainer), a "contradiction chain" is a sequence of applications of Single and ECP rules, i.e. of rules from BRT, based on the assumption of some candidate (if Z is True, then …); said otherwise, it is a proof in BRT. It has also been argued that whips and braids are "the same thing as" contradiction chains; this claim is based on too vague definitions of contradiction chains to be refuted, but such a chain is much more closer to our T&E procedure (or to some version of it that partly controls the total number of inferences) than to a whip or a braid. In the same view, W-whips, B-braids, W\*-whips and B\*-braids would all be "the same thing as" nested contradiction chains, although the latter are much closer to our T&E(2) procedure. Notwithstanding all this, our main point here is, the "same-thing" view completely forgets the notion of length inherently associated with each of our patterns, the confluence property of the braid resolution theories that can only be defined with this specific notion of length, the "simplest-first" strategy it justifies and the associated ratings. The "same-thing" view could as well be applied to the various types of "contradiction chains" on which Sudoku

Explainer is based (and to its associated measure of complexity, basically the number of inference steps, of which we have already mentioned that it cannot be defined by any logical theory – see note 5 of chapter 6); accordingly, the "same" pattern would thus be assigned three different ratings.

### 12.2.2. The $W^*_p W_m$ and $B^*_p B_m$ resolution theories; confluence property of $B^*_p B_m$

Definition: given an integer p with $1 \leq p \leq \infty$, we define in the now usual way the following increasing sequence of resolution theories:
 – $W^*_p W_0$ = BRT(CSP),
 – $W^*_p W_1 = W^*_p W_0 \cup W_1 \cup$ {rules for $W^*_p$-whips of pseudo-length 1},
 – …
 – $W^*_p W_m = W^*_p W_{m-1} \cup W_m \cup$ {rules for $W^*_p$-whips of pseudo-length m},
 – $W^*_p W_\infty = \cup_{m \geq 0} W^*_p W_m$.

One has obvious similar definitions for ($B^*_p B_m$, m≥0).

$W^*_p W_m$ [respectively $B^*_p B_m$] is based on W*-whips [resp. B*-braids] of maximum pseudo-length m with inner bi-whips [resp. bi-braids] of maximum length p. As for all our previous patterns, we also define $W^* W_m$ [resp. $B^* B_n$] as $W^*_\infty W_m$ [resp. $B^*_\infty B_m$], in which inner bi-whips [resp. bi-braids] may have unrestricted length.

Similarly to the whip, g-whip, S-whip, W-whip cases, or to the corresponding braid cases, one can associate a $W^*_p W$ [resp. a $B^*_p B$] rating with these families of resolution theories. Using theorem 12.1, it is easy to prove the following confluence property, so that the $B^*_p B$ rating has good computational properties. However, it should be stressed that these ratings neglect all the inner bi-whips [resp. bi-braids] and their meaning is therefore something queer, like "rating modulo the inner bi-whips [resp. bi-braids]". In particular, the $B^*_p B$ rating of any instance in T&E(1) is always 0.

***Theorem 12.5: for any p and n with $1 \leq p, m \leq \infty$, resolution theory $B^*_p B_m$ has the confluence property.***

Proof: as it is an easy combination of the proofs of theorems 5.6 and 12.1, we leave it as an exercise for the reader.

### 12.2.3. The T&E* and T&E*$_p$ procedures

Remember from section 12.1.3 the definitions of the extended basic resolution theory *BRT\** (the analogue of BRT, but in the language with an additional constant "bi-braid" in Constraint or Constraint-Type) and of the logical theories *$biB_p$* for p ≥ 1; remember also that, after theorem 12.1, all of these theories are stable for confluence (an essential property for the following definition to be meaningful).

Definition: for any p with 1≤p≤∞, given a resolution state RS of a CSP, in the above-defined extended language, *procedure T&E*$^{*2}_p$*(RS)* is defined as follows:
- loop until a solution or a contradiction is found or until quiescence:
   - in RS, apply repeatedly the rules of $B_p$ (braids of length ≤ p) until quiescence; if a solution is found, return it and stop;
   - loop until a solution or a contradiction is found or until quiescence:
      . in RS, apply the rules of bi$B_p$ to all the candidate pairs; (this step can only add links for the "bi-braid" constraint);
      . in RS, apply repeatedly the rules of BRT* (now use the bi-braid links produced in the previous step); if a solution is found, return it and stop;
   - end loop;
- end loop.

In case p = ∞, this is equivalent to T&E*$^2$(RS):
- loop until a solution or a contradiction is found or until quiescence:
   - set RS = T&E(RS); if a solution or a contradiction is found, return it and stop;
   - loop until a solution or a contradiction is found or until quiescence:
      . set RS = bi-T&E(RS); (this step can only add links for the "bi-braid" constraint);
      . set RS = T&E(BRT*); (now use the bi-braid links produced in the previous step);
   - end loop;
- end loop.

***Theorem 12.6 ("T&E*$^2_p$ vs B*$_p$-braids"): for any CSP and any p with 1≤p≤∞, if, in some resolution state RS in the extended language of BRT*, procedure T&E*$^2_p$(RS) deletes a candidate Z, then there exists in RS either a braid[p] with target Z or a B*$_p$-braid with target Z using only inner bi-braids no longer than p. For any p with 1≤p≤ ∞, an instance P of a CSP can be solved by T&E*$^2_p$ if and only if it can be solved by B*$_p$-braids.***

Proof: obvious, along the same lines as for all the previous similar theorems.

As all the previous similar theorems, this provides an easy means of checking whether an instance is in B*$_p$B.

In case p = ∞ and as a corollary to theorems 11.5 [T&E(2) vs B-braid], 12.4 [pseudo-equivalence of B*-braids and B-braids] and 12.6, one has the following theorem. (Exercise: write a direct proof.)

***Theorem 12.7: T&E*$^2$ ≡ T&E(2) – only in the limited sense that these two very different procedures always produce the same result.***

## 12.3. Patterns of proof and associated classifications

Given an instance in T&E(2), we have seen that it can be solved in very different ways with generic generalised chain patterns. It is one of the purposes of this section to review the various possibilities in the general case and to show how they work in practice. But, for the sake of providing this analysis with a broader perspective, we shall first reassess the general readability requirements one can reasonably put on the solution of instances beyond T&E(1) or gT&E(1); this will lead to the informal notion of a pattern of proof.

### *12.3.1. Rating versus classification*

As shown by the multitude of possible ratings introduced in this book – all logically grounded –, rating or classification of instances cannot be primary goals in themselves, even in a context restricted by the sole purpose of pattern-based solving; they have to be justified by higher level requirements such as simplicity or readability of a possible solution. Conversely, such vague requirements can only be made precise if there is some objective way of measuring how they are satisfied. So, rating and classification principles on the one hand and requirements analysis on the other, are intimately related. As a preamble to the sequel, let us clarify the difference we have implicitly made since chapter 11 between classification and rating:

– in a rating system (e.g. in any of the W, gW, B, gB, SB, BB or $B_pB$ systems), every instance is assigned a unique, possibly infinite value; it is supposed to represent its complexity (with respect to the corresponding family of resolution rules); any two instances can thus be directly compared;

– in a classification system, e.g. in the very broad T&E(?) or in $B_?B$ or in a similar $B^*_?B$, one has several levels and sublevels of complexity (think of the classification of electrons in an atom, with the various s, p, d, f … layers and sub-layers) and one does not necessarily try to compare directly instances from different levels (even if the $B_pB$ and the $B_qB$ ratings of an instance are indeed comparable, the smallest value of p such that $B_pB$ is finite is often more interesting than the $B_pB$ rating itself).

In this book, broad classifications have always been easier to compute than ratings; this may not be a general *a priori* difference between classification and rating, but a result of less stringent requirements; ratings have always supposed that we could exhibit a resolution path, whereas classifications allowed the use of the "T&E(T) vs T-braids" theorem; this is a clear advantage of classifications. As shown in chapter 13, a broad classification system such as the $B_?B$ can be enough (with no explicit computation of any rating) when one wants to analyse the worth of introducing a new resolution rule.

As for the relationship between the two approaches, given a rating system, a one-level classification system can obviously be obtained from it, based on the different values of the rating. But, conversely, given a classification system, there may be no unique rating compatible with it. However, in both cases, our minimal *a priori* requirements for a classification or a rating system are the same:

– it should be purely logical, i.e. it should be defined by an increasing sequence of CSP resolution theories (preferably with the confluence property, for better computational properties, although this is not necessary in theory);

– it should be invariant under all the logical symmetries of the CSP (which should be a consequence of the previous condition if predicate "linked" is correctly defined); by "logical symmetries", we mean more than just the obvious geometric ones (e.g., in Sudoku, we include in them the analogies between rows and blocks).

As for ratings, considering all those that have been defined in this book, one could add an *a posteriori* requirement: any rating should somehow be based on the number of CSP variables necessary to formulate the patterns in the resolution rules; as a result, the most natural ratings satisfying these conditions are those introduced until now (there is no reason to use any function of this number rather than the number itself).

### 12.3.2. The first two generic ways of solving a T&E(2) instance

Given an instance Q in T&E(2) [and not in gT&E(1)], the concrete question is, how can one present its solution in a readable form, as simply as possible? After chapter 11, there seemed to be only two theoretical options for simplicity, corresponding respectively to the rating and classification views: either one wants to find a resolution path leading to the smallest BB rating of Q or one wants a $B_pB$ solution of Q with the smallest p (and then, optionally, with the shortest possible $B_p$-braids, or $W_p$-whips if any, for this fixed value of p).

The first case corresponds to the idea that we want the hardest elimination to be globally as simple as possible (including in it all that is necessary to prove it); it is conceptually clear although it seems computationally intractable for instances beyond gT&E(1) [or T&E($S_p$) for small values of p].

In the second case, the first step is to find the smallest p; this can easily be done by applying the T&E($B_p$) procedure to Q for increasing values of p until it is able to solve Q. Having found the smallest p, we know that Q has a $B_p$-braids solution. If p = 1, this means a g-whip solution. For higher values of p, one can (at least in theory) look for the "simplest" $B_p$-braids solution (i.e. the solution with the smallest $B_pB$ rating) by applying the simplest-first strategy within resolution theory $B_pB$. How this works in practice has been illustrated in section 11.5 for p = 2.

In each of these approaches, the limits of readability are reached. Contrary to the puzzles in T&E(1) or gT&E(1), that have readable solutions with whips or g-whips (sometimes, they also require braids and g-braids), we consider that the $B_pB$ classification for $p > 2$ is more interesting from an abstract classification point of view (as in section 11.4) than for actually producing the simplest resolution path in $B_pB$ (in the sense of the $B_pB$ rating). In any event, it is likely that, for such instances, the readability requirement could hardly be met by any approach, unless the meanings of "readable" or "simplest" are significantly extended beyond those we have assigned them until now. It is our next goal to explore how this could be done.

### 12.3.3. The need to reassess our requirements for instances beyond gT&E(1)

It should first be recalled that, when it faces really new problems, any scientific discipline seldom progresses by keeping untouched the current formulation of its most general principles. Such principles can also be considered as requirements of "ultimate" understandability, because everything else (relevant to this discipline) should be explained by them. In parallel with a constant tendency to increased rigour or formalisation, there is in science a tendency to generalisation (and experience shows that, most of the time, these tendencies go hand in hand). Thus, the principle of mass conservation had to be extended to a principe of mass-energy conservation, Galilean symmetry to Poincaré symmetry and so on. Similarly, in evolution theory, the natural selection paradigm had to be extended from Darwin's initial view of "best" fitness in some niche to a more opportunistic view of fitness.

Many people with little scientific practice believe that science has general *a priori Laws*; but this is a very wrong view of science; from the outside, the "laws" of a scientific discipline may seem to be *a priori*, perhaps because they generally evolve only on long time scales, but they are indeed the result of its historical development. They are the super structure built to present all of its results in a unified framework.

There does not seem to be any general way of specifying how the general laws of a scientific discipline should be defined or interpreted or modified when necessary. However, they should be absolute and universal; although they are subject to (rare) change, the very notion of a general law would be meaningless without these two properties: they are absolute and universal in the current state of development of the discipline. But they should first of all also satisfy a complementary commonsense "principle of reasonableness" (or principle of submission to reality). In a not so different context, this has best been expressed in everyday terms by the King in "The Little Prince" (we added the italics):

   – Sire… over what do you rule?
   – Over everything, said the king, with magnificent simplicity.
   – Over everything?

> The king made a gesture, which took in his planet, the other planets, and all the stars.
> – Over all that? asked the little prince.
> – Over all that, the king answered.
> For his **rule** was not only **absolute**: it was also **universal**.
> – And the stars obey you?
> – Certainly they do, the king said. They obey **instantly**. I do not permit insubordination.
> Such power was a thing for the little prince to marvel at. […] he plucked up his courage to ask the king a favor:
> – I should like to see a sunset… Do me that kindness… Order the sun to set…
> – If I ordered a general to fly from one flower to another like a butterfly, or to write a tragic drama, or to change himself into a sea bird, and if the general did not carry out the order that he had received, which one of us would be in the wrong?, the king demanded. The general, or myself?
> – You, said the little prince firmly.
> – Exactly. One must require from each one the duty which each one can perform, the king went on. Accepted authority rests first of all on reason. If you ordered your people to go and throw themselves into the sea, they would rise up in revolution. ***I have the right to require obedience because my orders are reasonable***.

Although in a much less grandiose context, instances of a CSP beyond T&E(1) or gT&E(1) challenge the universality of our initial views of simplicity and rating. It is then time for a little more thinking about our requirements and our interpretations of them. Instead of sticking too strictly to these views, let us analyse, based on the concrete results of the previous sections, what could be considered as "reasonable" alternative interpretations of the vague requirements of simplicity, understandability, explanainability or readability for the resolution paths of such extreme instances.

### 12.3.4. The B*B and B*$_p$B approaches

For instances in T&E(2), the present chapter has introduced two new alternative ways of presenting a solution: compute either the bi-braid contradictions or the bi-braid[p'] contradictions for all the p' with $1 \leq p' \leq p$, and use them in addition to the direct contradiction links to find either the shortest B*-braids or the shortest B*$_p$-braids – shortest with respect to the pseudo-lengths of these B*-braids or B*$_p$-braids, i.e. when these indirect bi-braid contradiction links are considered as being no more complex than the direct ones.

Logically speaking, this amounts to considering as lemmas all the necessary bi-braid contradictions (resp. bi-braid[p'] contradictions for any p'≤p) and not counting their complexity in the proofs of the eliminations. From a technical point of view, as these contradictions are not structural, they may have to be continuously updated, with potentially new instances created every time a candidate is deleted during the resolution process. But this is not an unknown situation in mathematics: they can be considered as lemmas, intertwined with theorems (here the eliminations).

From a rating point of view, this approach consists of hiding part of the complexity of each elimination step (possibly the main part, as p is taken larger and larger). It does not allow to define ratings compatible with the previously defined ones, because it reduces to the same complexity (zero) all the inner bi-braid or bi-braid[p'] pairwise contradictions. It is equivalent to allowing a hidden level of T&E (possibly restricted by p), namely bi-T&E.

In the examples in the next two subsections, especially the first, if one considers without caution the resolution paths thus obtained, these instances may seem to be easy or relatively easy; but this kind of presentation obviously hides the main part of complexity.

One possibility for less cheating with complexity is to use these quasi resolution paths only as guides to a complete solution and to justify each of the B*-braids or $B^*_p$-braids they contain by the precise bi-braids on which it relies: once $Z_1$ and $Z_2$ are given and it is known that they are bi-braid incompatible (e.g. because we have first found it by applying the fast bi-T&E procedure), it is not very difficult to find the shortest bi-braid that can actually prove this in the current resolution state.

If eliminations are considered as theorems and bi-braid contradictions as lemmas, this approach provides the shortest theorems, at the cost of lemmas of uncontrolled complexity or of complexity bounded by p: even if, afterwards, we find the shortest bi-braid contradictions for each $(Z_1, Z_2)$ pair used in the path, there is no reason why the B*-braids or $B^*_{p'}$-braids in the path would globally use the shortest possible inner bi-braids; on the contrary, complexity evacuated from the B*-braids has to be compensated by complexity in the inner bi-braids.

What the above two approaches suggest, together with the two (BB and $B_pB$) of chapter 11, is that, for such T&E(2) instances, our initial requirement of simplicity cannot be understood in the simple terms of a unique rating or classification system. This raises more difficult questions that are relevant to automatic theorem proving in general: what does one really want in terms of readability or understandability of the proofs, e.g. what type of global *pattern of proof* does one allow these proofs to respect? Here, B*-braids or $B^*_p$-braids and the corresponding whip versions, can be considered as general patterns of proof, the inner bi-whips and/or bi-braids being there to fill in the details. We shall soon see that there are many alternative possibilities, but let us first illustrate these two B*B and $B^*_pB$ approaches.

### *12.3.5. Two examples of the B*B approcah*

For instances in T&E(2), possibly after applying more elementary rules (such as braids or g-braids), one can look for a solution with B*-braids of shortest pseudo-lengths, based on all the possible inner bi-braid contradictions (with no *a priori* limit on their lengths). If one is not interested in the exact bi-braids used (in the "details" of the proof), one can even start by computing the bi-braid contradictions via the much faster bi-T&E procedure.

Whether the bi-braids are continuously updated or updated when needed is an option that cannot change the final result; if they are updated at least when no more B*-braid[1] is available, theorem 12.4 guarantees that a solution in terms of forcing bi-braids will be found. However, if they are not updated at all, a B*-braid solution may not be found: for instance, it will not for the first puzzle in Eleven's collection.

*12.3.5.1 EasterMonster*

Consider first the famous EasterMonster puzzle (see Figure 13.1 in the next chapter). EasterMonster is in $B_6B$, but it has a very specific pattern allowing a series of thirteen eliminations, after which it is in $B_2B$ – see definition and discussion of this pattern (a "belt of crosses") in chapter 13.

```
***** SudoRules 16.2  based on CSP-Rules 1.2, config: B*B  *****
21 givens, 239 candidates, 1546 csp-links and 1546 links. Initial density = 1.36
belt[4] made of crosses:
   cross in block b1 with center r2c2
      horizontal outer candidates: 3 8; vertical outer candidates: 4 8; inner candidates: 2 7
   cross in block b3 with center r2c8
      horizontal outer candidates: 3 8; vertical outer candidates: 3 9; inner candidates: 1 6
   cross in block b9 with center r8c8
      horizontal outer candidates: 4 5; vertical outer candidates: 3 9; inner candidates: 2 7
   cross in block b7 with center r8c2
      horizontal outer candidates: 4 5; vertical outer candidates: 4 8;  inner candidates: 1 6
==>
   eliminations in rows: r2c5 ≠ 3, r2c5 ≠ 8, r2c6 ≠ 8, r8c4 ≠ 5, r8c5 ≠ 4
   eliminations in columns: r5c2 ≠ 4, r5c2 ≠ 8, r5c8 ≠ 3, r5c8 ≠ 9
   eliminations in blocks: r1c3 ≠ 7, r3c1 ≠ 2, r7c3 ≠ 1, r9c1 ≠ 6
```

After these eliminations, there is no available g-braid but the last part of theorem 12.4 guarantees that there is a solution with B*-braids[1] (or forcing bi-braids), using inner bi-braids of unrestricted length. At this point, we computed the set of 6,166 direct + bi-braid contradictions (via the bi-T&E procedure, using the "bi-braid vs bi-T&E" theorem, which is enough when we do not want to take the lengths of the inner bi-braids into account); this set does not even need to be updated before the solution is obtained by a relatively short sequence of B*-braids[1]. Remember however what we said above: giving this resolution path without providing any details about the inner bi-braids of unrestricted length is obviously cheating with complexity.

```
b*-braid[1]: r6c3{n9 .} ==> r9c1 ≠ 4
b*-braid[1]: b7n4{r9c2 .} ==> r8c9 ≠ 5
b*-braid[1]: b7n8{r9c2 .} ==> r8c9 ≠ 4
singles ==> r8c9 = 7, r4c8 = 7
```

b*-braid[1]: r9c8{n9 .} ==> r8c7 ≠ 2
hidden-single-in-a-block ==> r7c8 = 2
b*-braid[1]: r8c7{n5 .} ==> r8c5 ≠ 6
b*-braid[1]: r9c1{n9 .} ==> r8c4 ≠ 6
hidden-single-in-a-row ==> r8c1 = 6
b*-braid[1]: r8c4{n2 .} ==> r9c6 ≠ 8
b*-braid[1]: r8c4{n2 .} ==> r9c6 ≠ 5
b*-braid[1]: r8c4{n2 .} ==> r9c4 ≠ 5
b*-braid[1]: r8c4{n2 .} ==> r9c4 ≠ 3
b*-braid[1]: b9n9{r9c8 .} ==> r8c4 ≠ 1
singles ==> r8c4 = 2, r8c5 = 1, r7c2 = 1
b*-braid[1]: r5c4{n6 .} ==> r9c6 ≠ 7
hidden-single-in-a-block ==> r9c4 = 7
b*-braid[1]: r9c6{n6 .} ==> r9c7 ≠ 4
b*-braid[1]: b8n6{r9c6 .} ==> r9c5 ≠ 8
b*-braid[1]: b8n6{r9c6 .} ==> r9c5 ≠ 4
b*-braid[1]: c4n1{r5 .} ==> r8c3 ≠ 4
singles ==> r8c3 = 5, r8c7 = 4, r3c9 = 4, r3c1 = 5, r9c7 = 5, r5c9 = 5
b*-braid[1]: r1c7{n9 .} ==> r9c2 ≠ 8
singles to the end

### 12.3.5.2 Eleven#3 (a puzzle in $B_7B$)

In order to show that the B*B approach does not exclude the hardest instances (hardest according to the $B_?B$ classification), consider now the first of Eleven's puzzles in $B_7B$ (#3 in his list, Figure 12.3), one of the three hardest known ones in this classification, as mentioned in section 11.4.2.

|   | 3 |   |   |   | 8 |   |   |   |
|---|---|---|---|---|---|---|---|---|
|   | 5 |   |   |   | 2 |   | 1 |   |
| 7 |   |   |   |   |   |   |   |   |
|   |   | 5 |   | 8 |   |   |   | 6 |
|   | 9 | 1 | 2 |   |   |   |   |   |
| 8 |   |   |   | 3 |   |   |   |   |
|   | 6 | 9 |   |   |   |   |   | 5 |
|   |   | 4 |   |   |   | 7 |   |   |
|   |   |   | 1 |   | 6 |   | 2 |   |

*Figure 12.3. Puzzle Eleven#3*

The same technique as for EasterMonster works, although it now requires B*-braids of greater pseudo-lengths, up to six, if we never update the initial bi-braid contradictions. There are 4,518 initial direct links plus bi-braid contradictions. In the forthcoming resolution path, only B*-braids are active. The following can be considered as the general lines of a proof of the solution based on only bi-braid contradictions available in the initial resolution state (and not giving the details is again cheating with complexity).

***** SudoRules 16.2   based on CSP-Rules 1.2, config: B*B   *****
22 givens, 238 candidates, 1609 csp-links and 1609 links. Initial density = 1.43
b*-braid[1]: r9n8{c4 .} ==> r6c8 ≠ 4
b*-braid[1]: r5n4{c9 .} ==> r6c3 ≠ 7
b*-braid[1]: r9n8{c4 .} ==> r4c8 ≠ 4
b*-braid[1]: r9n9{c8 .} ==> r4c8 ≠ 3
b*-braid[1]: b9n9{r9c8 .} ==> r3c6 ≠ 6
b*-braid[1]: r2c8{n9 .} ==> r3c6 ≠ 5
b*-braid[1]: b5n7{r6c5 .} ==> r3c6 ≠ 4
b*-braid[1]: b9n9{r9c8 .} ==> r3c5 ≠ 9
b*-braid[1]: b9n9{r9c8 .} ==> r2c5 ≠ 9
b*-braid[1]: b9n9{r9c8 .} ==> r1c6 ≠ 7
b*-braid[1]: b5n7{r6c5 .} ==> r1c6 ≠ 4
b*-braid[1]: b9n9{r9c8 .} ==> r1c5 ≠ 9
b*-braid[1]: b9n9{r9c8 .} ==> r1c5 ≠ 7
b*-braid[1]: b9n9{r9c8 .} ==> r1c4 ≠ 7
b*-braid[2]: c6n4{r2 r5} – b8n7{r9c4 .} ==> r3c3 ≠ 6
b*-braid[2]: b1n9{r1c1 r2c1} – c3n7{r9 .} ==> r5c7 ≠ 5
b*-braid[2]: r7c3{n1 n7} – r9n7{c3 .} ==> r8c1 ≠ 3
b*-braid[3]: c1n9{r1 r8} – r4c8{n9 n2} – r6c8{n2 .} ==> r3c8 ≠ 4
b*-braid[4]: r2n4{c4 c1} – c1n9{r2 r8} – r4c8{n9 n2} – r6c8{n2 .} ==> r3c8 ≠ 3
b*-braid[4]: c3n1{r3 r4} – c2n1{r4 r1} – c2n4{r1 r6} – b4n7{r6c2 .} ==> r4c2 ≠ 2
b*-braid[4]: c2n3{r4 r8} – b7n8{r7c3 r9c3} – r8n1{c1 c7} – r6c7{n1 .} ==> r4c2 ≠ 1
b*-braid[4]: r5c3{n5 n6} – r2c6{n4 n9} – r1c6{n9 n1} – r3c6{n1 .} ==> r2c4 ≠ 6
b*-braid[5]: r2c8{n3 n6} – r2c1{n6 n4} – c8n4{r2 r5} – r7n3{c5 c1} – r4c1{n3 .} ==> r5c1 ≠ 5
b*-braid[1]: c1n5{r9 .} ==> r9c3 ≠ 5
b*-braid[3]: b1n6{r2c1 r1c1} – r5c1{n6 n3} – b6n3{r5c9 .} ==> r6c7 ≠ 5
b*-braid[1]: b6n5{r6c8 .} ==> r3c8 ≠ 5
b*-braid[4]: c8n5{r5 r6} – r5c1{n3 n6} – b1n6{r1c1 r2c3} – c6n4{r9 .} ==> r5c8 ≠ 4
b*-braid[4]: r6n6{c4 c3} – r5n5{c8 c3} – c8n4{r2 r9} – b8n4{r9c6 .} ==> r2c6 ≠ 7
b*-braid[4]: r3c8{n9 n6} – r2c8{n6 n4} – r1n9{c1 c6} – r2c6{n9 .} ==> r3c7 ≠ 9
b*-braid[4]: r2n7{c5 c4} – r2n4{c4 c1} – r5n4{c1 c7} – c5n3{r8 .} ==> r7c3 ≠ 8
b*-braid[5]: r2n4{c4 c1} – r5c1{n4 n3} – r4c1{n2 n1} – b1n9{r3c3 r2c3} – r2c6{n9 .} ==> r2c5 ≠ 6
b*-braid[5]: r3n9{c6 c3} – r9c3{n9 n7} – b4n1{r4c1 r4c3} – r7c3{n1 n2} – c1n2{r8 .} ==> r1c6 ≠ 2
b*-braid[5]: c4n8{r2 r3} – c3n6{r2 r5} – r5c1{n6 n3} – c2n3{r4 r8} – b8n7{r9c4 .} ==> r5c9 ≠ 4
b*-braid[5]: r5c1{n3 n6} – b1n6{r1c1 r2c3} – r2c6{n4 n9} – r2c8{n9 n3} – r5c8{n3 .} ==> r6c3 ≠ 1
b*-braid[5]: r5c1{n3 n6} – b1n6{r1c1 r2c3} – r2c6{n4 n9} – r2c8{n9 n3} – r5c8{n3 .} ==> r6c3 ≠ 2
**b*-braid[6]: c1n6{r1 r2} – c1n9{r2 r8} – r7c5{n4 n8} – b7n8{r8c2 r9c2} – r9n3{c2 c4} – r8n3{c5 .} ==> r2c8 ≠ 4**
b*-braid[1]: c8n4{r9 .} ==> r7c7 ≠ 4
b*-braid[5]: r7n4{c5 c8} – r6n1{c7 c8} – c3n2{r4 r7} – c6n2{r7 r8} – c6n6{r8 .} ==> r1c6 ≠ 5
b*-braid[1]: c6n5{r9 .} ==> r8c5 ≠ 5
**b*-braid[6]: b9n3{r7c7 r9c8} – b7n8{r8c2 r9c2} – r5n7{c3 c6} – r5n4{c6 c1} – r2n4{c1 c6} – b8n4{r9c6 .} ==> r8c1 ≠ 2**

**b\*-braid[6]: c3n6{r2 r6} – c8n5{r5 r6} – r7n4{c8 c5} – r7n3{c5 c1} – c1n2{r7 r1} – r8c2{n1 .}**
**==> r2c6 ≠ 4**

From this point on, the solution can be found by similar B\*-braids of maximum pseudo-length 3. But it can also be found by g-braids or even by g-whips of maximum length 8; in the context of this chapter, this can be considered as easy and we shall skip the end of the path.

### 12.3.6. An example of the B\*$_p$B approach, p fixed

Instead of looking for a solution with B\*-braids with the shortest pseudo-lengths, with no restriction on the lengths of the inner bi-braids, as in the previous examples, one can restrict these to some length p (possibly after choosing the smallest possible p) and look for a solution with B\*$_p$-braids with the shortest pseudo-lengths.

|    | c1 | c2 | c3 | c4 | c5 | c6 | c7 | c8 | c9 |    |
|----|----|----|----|----|----|----|----|----|----|----|
| r1 | n1 | n2 n3 n4 n5 n6 n8 | n3 n4 n6 n7 n8 | n2 n4 n5 n6 n9 | n2 n3 n4 n5 n9 | n3 n5 n6 n9 | n2 n7 n8 n9 | n3 n4 n8 n9 | n2 n3 n4 n9 | r1 |
| r2 | n2 n3 n4 n5 n7 | n2 n3 n4 n5 | n3 n4 n7 | n1 | n8 | n3 n5 n9 | n2 n7 n9 | n3 n4 n9 | n6 | r2 |
| r3 | n2 n3 n4 n6 n8 | n2 n3 n4 n6 n8 | n9 | n2 n4 n6 | n2 n3 n4 | n7 | n1 n2 n8 | n5 | n1 n2 n3 n4 | r3 |
| r4 | n2 n4 n7 | n1 n2 n4 | n1 n4 n7 | n2 n5 n7 n9 | n6 | n1 n5 n9 | n3 | n1 n4 n9 | n8 | r4 |
| r5 | n2 n3 n6 n8 | n1 n2 n3 n6 n8 n9 | n5 | n2 n8 n9 | n1 n2 n9 | n4 | n1 n2 n6 n9 | n7 | n1 n2 n9 | r5 |
| r6 | n2 n4 n6 n7 n8 | n1 n2 n4 n6 n8 n9 | n1 n4 n7 n8 | n3 | n1 n2 n7 n9 | n1 n8 n9 | n5 | n1 n4 n6 n9 | n1 n2 n4 n9 | r6 |
| r7 | n3 n4 n5 n6 n8 | n1 n3 n4 n5 n6 n8 n9 | n1 n4 n6 n8 | n4 n6 n8 n9 | n1 n3 n4 n7 n9 | n2 | n1 n6 n8 n9 | n1 n3 n6 n8 n9 | n7 | r7 |
| r8 | n3 n6 n8 | n1 n3 n6 n8 | n2 | n6 n7 n8 n9 | n1 n3 n7 n9 | n1 n3 n6 n8 | n4 | n1 n3 n6 n8 n9 | n5 | r8 |
| r9 | n9 | n7 | n1 n3 n4 n6 n8 | n4 n5 n6 n8 | n1 n3 n4 n5 | n1 n3 n5 n6 n8 | n1 n6 n8 | n2 | n1 n3 | r9 |
|    | c1 | c2 | c3 | c4 | c5 | c6 | c7 | c8 | c9 |    |

***Figure 12.4.** Resolution state RS$_1$ of puzzle (eleven #26370) in T&E(2)*

The last puzzle (#26370) in Eleven's collection, obtainable from Figure 12.4 by deleting n5r6c7 as a given, provides an example of the B\*$_p$B approach for p = 2. It can easily be checked that it is in T&E(B$_2$) and therefore in B$_2$B. We shall now prove that it is also in B*$_2$B and even in W*$_2$W, mainly for the sake of showing how a solution based on B*$_2$-braids or W*$_2$-whips can look like.

***** SudoRules 16.2 based on CSP-Rules 1.2, config: W*$_2$W *****
22 givens, 245 candidates, 1708 csp-links and 1708 links. Initial density = 1.43
hidden-single-in-a-column ==> r6c7 = 5
whip[1]: r9n5{c6 .} ==> r7c4 ≠ 5, r7c5 ≠ 5

;;; resolution state RS$_1$, displayed in Figure 12.4.

After these obvious steps, there is no whip or g-whip and we enter the realm of bi-braids and B*$_2$-braids (actually the simpler realm of bi-whips and W*$_2$-whips). There are now three W*$_2$-whip eliminations; each of the last two is made possible by the previous one (in the present case, the new bi-whips that may appear after each of these steps play no role in the last two eliminations):

**w*2-whip[8]: c3n8{r7 r9} – r8n8{c1 c8} – c6n8{r8 r6} – r5n6{c1 c7} – r9c7{n6 n1} – r5n1{c7 c5} – r7c5{n1 n3} – c3n3{r7 .} ==> r7c4 ≠ 8**
**w*2-whip[4]: c3n3{r7 r9} – c8n3{r1 r8} – r7c4{n9 n6} – c3n6{r7 .} ==> r7c5 ≠ 3**
**w*2-whip[4]: c3n6{r7 r9} – r8n6{c1 c8} – r7c5{n9 n1} – c8n1{r7 .} ==> r7c4 ≠ 6**

After this point, the resolution path is entirely in B$_1$B = gB and it can even be expressed by short g-whips of maximum length 5; as it has nothing noticeable and it can be considered as easy in the context of this chapter, we skip it.

The above W*$_2$-whips can be considered as defining the (in the present case, very simple) general *pattern of proof* or the main lines of a forthcoming full proof. Let us now fill in the details of this proof. For this purpose, for each of the CSP variables involved, we first mark its right-linking candidate with an integer between brackets, its standard z- and t- candidates with our usual * and # symbols, and its remaining candidates with the capital letters associated with the following bi-whips, if they must be justified either as left-linking candidates or as additional ones that are bi-whip incompatible with the target or with a previous right-linking one:

w*2-whip[8]: c3n8{r7 r9$_{(1)}$ r1$_M$ r6$_A$} – r8n8{c1 c8$_{(2)}$ c2$_{\#1}$ c4$_*$ c6$_*$} – c6n8{r8 r6$_{(3)}$ r9$_*$} – r5n6{c1$_N$ c7$_{(4)}$ c2$_P$} – r9c7{n6 n1$_{(5)}$ n8$_{\#1}$} – r5n1{c7 c5$_{(6)}$ c2$_Q$ c9$_B$} – r7c5{n1 n3$_{(7)}$ n4$_C$ n9$_D$} – c3n3{r7 . r1$_R$ r2$_S$ r9} ==> r7c4 ≠ 8
w*2-whip[4]:   c3n3{r7 r9$_{(1)}$ r1$_R$ r2$_S$} – c8n3{r1$_E$ r8$_{(2)}$ r2$_F$ r7$_*$} – r7c4{n9$_G$ n6$_{(3)}$ n4$_H$} – c3n6{r7 . r1$_J$ r6$_T$ r9} ==> r7c5 ≠ 3 (made possible by the elimination of n8r7c4)
w*2-whip[4]: c3n6{r7 r9$_{(1)}$ r1$_J$ r6$_T$} – r8n6{c1 c8$_{(2)}$ c2$_{\#1}$ c4$_*$ c6$_*$} – r7c5{n9$_K$ n4$_L$ } – c8n1{r7 . r4$_U$ r6$_V$ r8$_{\#2}$} ==> r7c4 ≠ 6 (made possible by the elimination of n3r7c5)

The following twenty bi-whips, necessary to justify these three W*$_2$-whips, are only a small subset of the 677 bi-whips[1] and 300 bi-whips[2] available in RS$_1$ – to be compared also with the 1,374 direct links between the candidates remaining in RS$_1$. We mark them with the letters corresponding to those used for indexing the W*$_2$-whips. Taken individually, none of them is very complex, but their number shows that the solution is far from being as simple as would falsely be suggested if we gave no more detail than the above three $W^*_2$-whips in bold.

A: bi-whip[1]: c6n8{r9 .} ==> bi-whip-contrad(n8r6c3 n8r7c4)
B: bi-whip[1]: c8n1{r8 .} ==> bi-whip-contrad(n1r5c9 n1r9c7)
C: bi-whip[1]: r9n4{c5 .} ==> bi-whip-contrad(n4r7c5 n8r9c3)
D: bi-whip[1]: b9n9{r8c8 .} ==> bi-whip-contrad(n8r8c8 n9r7c5)
E: bi-whip[1]: b9n3{r9c9 .} ==> bi-whip-contrad(n3r1c8 n3r9c3)
F: bi-whip[1]: b9n3{r9c9 .} ==> bi-whip-contrad(n3r2c8 n3r9c3)
G: bi-whip[1]: b9n9{r8c8 .} ==> bi-whip-contrad(n3r8c8 n9r7c4)
H: bi-whip[1]: r9n4{c5 .} ==> bi-whip-contrad(n3r9c3 n4r7c4)
J: bi-whip[1]: c6n6{r9 .} ==> bi-whip-contrad(n6r1c3 n6r7c4)
K: bi-whip[1]: b9n9{r8c8 .} ==> bi-whip-contrad(n6r8c8 n9r7c5)
L: bi-whip[1]: r9n4{c5 .} ==> bi-whip-contrad(n4r7c5 n6r9c3)
M: bi-whip[2]: b3n8{r1c7 r3c7} – r9n8{c7 .} ==> bi-whip-contrad(n8r1c3 n8r7c4)
N: bi-whip[2]: b4n3{r5c1 r5c2} – r5n8{c2 .} ==> bi-whip-contrad(n6r5c1 n8r7c4)
P: bi-whip[2]: b4n3{r5c2 r5c1} – r5n8{c1 .} ==> bi-whip-contrad(n6r5c2 n8r7c4)
Q: bi-whip[2]: b4n3{r5c2 r5c1} – b4n8{r5c1 .} ==> bi-whip-contrad(n1r5c2 n8r6c6)
R: bi-whip[2]: r3n3{c1 c9} – r9n3{c9 .} ==> bi-whip-contrad(n3r1c3 n3r7c5)
S: bi-whip[2]: r3n3{c1 c9} – r9n3{c9 .} ==> bi-whip-contrad(n3r2c3 n3r7c5)
T: bi-whip[2]: r5n6{c1 c7} – r9n6{c7 .} ==> bi-whip-contrad(n6r6c3 n6r7c4)
U: bi-whip[2]: r5n1{c5 c2} – r8n1{c2 .} ==> bi-whip-contrad(n1r4c8 n1r7c5)
V: bi-whip[2]: r5n1{c5 c2} – r8n1{c2 .} ==> bi-whip-contrad(n1r6c8 n1r7c5)

Exercise: re-write the above $W^*_2$-whips as $W_2$-braids and compute their lengths as $W_2$-braids (notice that these may not be the smallest possible $W_2$-braids).

In this example – relatively simple for a puzzle in T&E(2) –, three $W^*_2$-whip eliminations are enough to bring the puzzle to a much easier situation (in gW). But, this was the last puzzle in Eleven's list [one of the easiest in T&E(2) and not in gT&E(1)] and, for harder puzzles, many more eliminations based on much longer bi-braid contradictions will generally be necessary. Moreover, there is no guarantee that a puzzle in $B_pB$ has a solution in $B^*_pB$.

### *12.3.7. Theorems and lemmas of equal complexity: the [B*B] classification*

There appears to be a compromise between the above two options:

– minimising the complexity of the theorems (eliminations), at the cost of lemmas of unrestricted complexity, as in the B*B approach illustrated in section 12.3.5,

– minimising the complexity of the lemmas (bi-braid contradictions) they rely on, at the cost of theorems of unrestricted complexity, as in the $B^*_?B$ approach illustrated in section 12.3.6 (in which the smallest p such that there is a solution in $B^*_pB$ is first looked for).

Indeed, as each $B^*_p B_m$ has the confluence property, one can vary p and m arbitrarily and many compromises are possible. In particular, one can require that theorems and lemmas have the same maximum complexity. This amounts to setting p = m. Considering then the increasing sequence ($B^*_p B_p$, p≥0) of resolution theories, one can define the [B*B] rating in the usual way.

Definition: given an instance Q of a CSP, its *[B*B] rating* is the smallest p such that Q can be solved in $B^*_p B_p$, i.e. by B*-braids of maximum pseudo-length p relying on inner bi-braids of maximum length p. Having an infinite [B*B] rating means that Q cannot be solved by B*-braids, i.e. that it is not in T&E(2).

As this rating is mainly intended for instances in T&E(2) beyond gT&E (although it could in theory apply to any instance), we prefer considering it as a sub-classification of instances in T&E(2).

Such a compromise may be justified in logical puzzles or from an abstract logical point of view. However, if we tried to extend it to automated theorem proving in general, it would be, from a mathematical point of view, at variance with the usual implicit and non formalisable requirement that lemmas should be "meaningful"; whether lemmas are harder to prove than theorems is irrelevant. This leaves aside the question of deciding in general what should be called a lemma and what a theorem: the difference cannot be formalised in logic; it is based on meaning, on the possibility of a non-contextual (or, at least, not too much contextual) formulation and on more or less arbitrary choices; however, in the context of this chapter, considering binary contradictions as lemmas and eliminations as theorems and considering both as "meaningful" sounds quite natural, so that this general question can be skipped.

Moreover, although accepting theorems and lemmas of equal complexity may seem to be a rational choice, it does not take into account considerations about the complexity of choosing which lemmas to use. If there are n candidates, then there are n(n-1)/2 candidate pairs. As a result, the number of potential bi-braid contradictions is much larger than the number of potential braids of same length. So that it may seem better to define a rating based on the ($B^*_p B_{p(p-1)/2}$, p≥0) or even a ($B^*_p B_{f(p)}$, p≥0) sequence of resolution theories, where f is a function increasing (much) faster than p. As the possibilities for such f functions are almost unlimited, and there does not seem to be any really good choice, we shall not dwell on them.

### *12.3.8. Different patterns of proof involved in the above approaches*

Perhaps the simplest way of analysing the differences between the above approaches consists of exhibiting their respective *global* patterns of proof. Recalling the remarks in section 5.7.7 about the no OR-branching in any of the patterns introduced in this book, they can be considered to refer to the *local* patterns of proof used in each elimination step; we are now dealing with the pattern of the whole proof. In the following patterns of proof, we use a standard notation for patterns in general, where the vertical bar "|" means "or", "*" means zero or more occurrences, and (…)$_{p = 1, …}$ means a sequence indexed by values of p increasing from 1 to infinity (i.e. to an *a priori* unbounded finite value).

For easier comparison with the harder theories under discussion here, let us first mention the global pattern of a proof in the "elementary" $B_p$ or $gB_p$ (including B = $B_∞$ or gB = $gB_∞$) resolution theories:

[$E_p$ | A]*, with:
    $E_p$ = candidate elimination in $B_p$ or $gB_p$
    A = assertion by Single

We can now write the patterns of proof underlying the various approaches analysed in this chapter:

– for the $B_pB$ (including BB = $B_∞B$) approach, p fixed:

[$E_p$ | A]*, with:
    $E_p$ = candidate elimination in $B_pB$
    A = assertion by Single

– for the $B^*_pB$ and $B^*B$ (= $B^*_∞B$) approach, p fixed:

[[$E_p$ | A]*$L_p$*]*, with:
    $E_p$ = candidate elimination in $B^*_pB$
    A = assertion by Single
    $L_p$ = assertion of a bi-braid contradiction in $biB_p$

– for the $[B^*B]_?$ approach:

([$E_p$ | A]*$L_p$*)$_{p = 1, …}$, with:
    $E_p$ = candidate elimination in $[B^*B]_p$
    A = assertion by Single
    $L_p$ = assertion of a bi-braid contradiction in $biB_p$

In the $B_pB$ case (p = 2, … ∞), this apparently simple description must however be completed by recalling that each proof of a candidate-elimination theorem in $B_pB$ relies on a much more complex structure than in the other cases: as $B_p$-braids include inner braids that may depend on the target and on previous right-linking candidates, these should be considered as "contextual lemmas", i.e. lemmas whose scopes are restricted to very particular situations. In all the other cases, each theorem or lemma is valid in the current resolution state with no further restriction. $E_p$, a candidate elimination in $B_pB$, with its contextual lemmas (= inner braids) made explicit, obeys the following pattern, in which "||" means that both actions should be done in parallel and freely intertwined. As before, this should be considered as a general pattern of proof, not as a procedure.

```
start proof (find a partial-braid[1])
Loop until a full B-braid is found
   continue main proof (extend the current partial B-braid) ||
      find a contextual elimination (inner braid)
end loop
```

### 12.4. d-whips, d-braids, $W^{*d}$-whips and $B^{*d}$-braids

Most of what has been done in the previous sections can be further generalised so as to take into account the indirect contradictions between more than two candidates that inevitably appear for instances in T&E(3) and beyond. As the various possible requirements on solutions and on how to mix d-contradictions with various values of d are still more numerous but they also are straightforward extensions of the above, we shall give precise definitions but we shall leave the theorems and their proofs as exercises.

Indeed, more than all the technical possibilities suggested below, what is remarkable here, as mentioned in the Introduction, is that the existence of instances requiring T&E(d) with $d \geq 3$, together with the equivalence of T&E(d) with $B^d$-braid contradictions, shows that, in order to get a constructive solution, it is sometimes necessary to consider derived constraints among more than two labels, even though the given CSP was initially supposed to have only binary constraints. The gap between the what and the how is still more impressive than suggested by the unary and binary derived constraints we had to introduce with resolution rules for whips, braids, W-whips, B-braids, W*-whips and B*-braids.

#### 12.4.1. d-whips and d-braids

Definition: given $d \geq 1$ and given d different candidates $Z_1, Z_2, \ldots, Z_d$ in a resolution state RS, with no two of them linked, for any $n \geq 1$, a d-whip[n] built on $Z_1, Z_2, \ldots, Z_d$ is a structured list $(\{Z_1, Z_2, \ldots, Z_d\}, (V_1, L_1, R_1), \ldots, (V_{n-1}, L_{n-1}, R_{n-1}), (V_n, L_n))$, such that:
   – for any $1 \leq k \leq n$, $V_k$ is a CSP variable;
   – $Z_1, Z_2, \ldots, Z_d$, all the $L_k$'s and all the $R_k$'s are candidates in RS;
   – in the sequence $(L_1, R_1, \ldots, L_{n-1}, R_{n-1}, L_n)$, any two consecutive elements are different;
   – none of $Z_1, Z_2, \ldots$ and $Z_d$ belongs to $\{L_1, R_1, L_2, R_2, \ldots, L_n\}$;
   – $L_1$ is linked to $Z_1, Z_2, \ldots$ or $Z_d$;

– right-to-left continuity: for any $1<k\leq n$, $L_k$ is linked to $R_{k-1}$;
– strong left-to-right continuity: for any $1\leq k<n$, $L_k$ and $R_k$ are candidates for $V_k$;
– $L_n$ is a candidate for $V_n$;
– at least one of $Z_1, Z_2, \ldots$ and $Z_d$ is not a label for $V_n$;
– for any $1\leq k<n$: $R_k$ is the only candidate for $V_k$ compatible with $Z_1, Z_2, \ldots, Z_d$ and all the previous $R_i$ ($i<k$);
– $V_n$ has no candidate compatible with $Z_1, Z_2, \ldots, Z_d$ and all the previous $R_i$ ($i<n$); (but $V_n$ has more than one candidate).

Remark: 2-whips[n] are the same thing as the bi-whips[n] defined in section 11.4.1.

Definition: given $d\geq 1$ and d candidates $Z_1, Z_2, \ldots, Z_d$ in a resolution state RS, with no two of them linked, for any $n\geq 1$, a d-braid[n] built on $Z_1, Z_2, \ldots, Z_d$ is a structured list as above, with the right-to-left continuity condition replaced by:
– for any $1<k\leq n$, $L_k$ is linked to $Z_1, Z_2, \ldots$, or $Z_d$ or a previous $R_i$.

Definitions: given a resolution state RS, d different candidates $Z_1, Z_2, \ldots$ and $Z_d$ in RS, such that no two of them are linked, are said *d-whip[n]* (respectively *d-braid[n]*) *incompatible or contradictory in RS* if there exists in RS some d-whip[n] (resp. some d-braid[n]) built on $Z_1, Z_2, \ldots$ and $Z_d$. $Z_1, Z_2, \ldots$ and $Z_d$ are said *d-whip* (respectively *d-braid*) *incompatible or contradictory in RS* if there is some n such that, in RS, they are d-whip[n] (resp. d-braid[n]) incompatible.

Now defining the $\text{nand}_d(Z_1, Z_2, \ldots, Z_d)$ predicate as
$\text{nand}_d(Z_1, Z_2, \ldots, Z_d) \equiv \neg[\text{candidate}(Z_1) \wedge \text{candidate}(Z_2) \wedge \ldots \wedge \text{candidate}(Z_d)]$,
it is obvious that **all the d-whip[n] and d-braid[n] contradiction relations between d candidates are constructive restricted forms of this pure logic $\text{nand}_d$ predicate.** Moreover, all these relations are symmetric in all their arguments.

Exercise: define d-braid[m] logical theories and prove their stability for confluence; define a procedure d-T&E($Z_1, Z_2, \ldots, Z_d$) and prove its equivalence with the existence of a d-braid of unrestricted length built on $Z_1, Z_2, \ldots, Z_d$.

### 12.4.2. $W^{*d}$-whips and $B^{*d}$-braids

These d-whips and d-braids can now be used in a way very close to the way bi-whips and bi-braids have been used in section 12.2.

Definition: given a resolution state RS of any CSP and a candidate Z in RS, a $B^{*d-1}$-*braid* based on Z is a structured list $(Z, (V_1, L_1, R_1), \ldots, (V_{m-1}, L_{m-1}, R_{m-1}), (V_m, L_m))$ that satisfies the following conditions:
– for any $1\leq k\leq m$, $V_k$ is a CSP variable;

– Z, all the $L_k$'s and all the $R_k$'s are candidates;

– in the sequence of labels $(L_1, R_1, …, L_{m-1}, R_{m-1}, L_m)$, any two consecutive elements are different;

– Z does not belong to $\{L_1, R_1, L_2, R_2, …. L_m\}$;

– $L_1$ is linked to Z;

– for any $1<k\leq m$, there is some d', $0\leq d'<d$, such that $L_k$ is d'-braid incompatible *in RS* with a subset of d' elements taken from $\{Z\} \cup \{R_j, j<k\}$;

– strong left-to-right continuity: for any $1\leq k<m$, $L_k$ and $R_k$ are candidates for $V_k$;

– Z is not a label for $V_m$;

– for any $1\leq k<m$: $R_k$ is the only candidate for $V_k$ that is not d'-braid incompatible *in RS* with a subset of d' elements taken from $\{Z\} \cup \{R_j, j<k\}$ for some d', $0\leq d'<d$;

– $V_m$ has no candidate that is not d'-braid incompatible *in RS* with a subset of d' elements taken from $\{Z\} \cup \{R_j, j<k\}$ for some d', $0\leq d'<d$; (but $V_m$ has more than one candidate – the usual non-degenaracy condition).

We can now define the following increasing sequence of resolution theories:

– $B^{*0}$ = BRT(CSP);

– $B^{*1} = B_\infty = \cup_{n\geq 0} B_n$, the now familiar braids resolution theory; …

– $B^{*d} = B^{*d-1}$ ∪ rules for $B^{*d-1}$-braids.

Exercise: prove that all these theories have the confluence property, define the appropriate T&E$^{*d}$ procedure and prove an equivalence theorem.

Notice that the passage from $B^{*d-1}$ to $B^{*d}$ could be replaced by the addition of a single formula with precondition the existence of a d-braid and with conclusion the assertion of a nand$_d$ about d candidates (this is not a standard resolution rule, but it is still a logic formula with no disjunction):

$B^{*d} = B^{*d-1} \cup \forall l_1 \forall l_2 … \forall l_d [\text{Bi-T\&E-contrad}_d(l_1, l_2, …, l_d) \Rightarrow \text{nand}_d(l_1, l_2, …, l_d)]$

such nand$_d$ would then be used inside $B^{*d}$ by the standard laws of constructive logic.

Notice also that, if the above definitions took care of the lengths of the various d'-braids used in the d'-braid contradiction relations, the total length of a $B^{*d}$-braid could be defined; and a "universal" rating B*∗B could also be defined. However, the computational complexity of $B^{*d}$-braids may make them computationally intractable. Alternatively, for an instance in T&E(d), one could define, upwards from deeper to shallower layers, a sequence $(p_d, …, p_2, p_1)$ of the minimal sizes necessary for each of the sets of d'-braid contradictions, assuming at each level all the deeper Bi-T&E-contrad$_{d''}$ contradictions obtainable with the previous maximal allowed $p_{d''}$ lengths.

**Part Four**

# MATTERS OF MODELLING

# 13. Application-specific rules (the sk-loop in Sudoku)

As a counterpoint to the first chapter about modelling a Constraint Satisfaction Problem, the present one will tackle the problem of modelling resolution rules. Until now, we have been little concerned with modelling questions relative to rules (exceptions are the discussion about how to define g-labels or to express the non-degeneracy conditions of Subsets): all our rules were progressively derived from the two most basic types of rules for the Sudoku CSP, namely xy-chains (i.e. bivalue-chains restricted to rc-space) and Subset rules. In the process, our guiding principles have been theoretical: they rested on the analysis of how to transpose them to the general CSP, how to prove them and how to generalise these proofs further and further (of course, these are also modelling principles). In this respect, the approach followed in this book is very close to that we first applied in *HLS*.

However, over the years, participants of Sudoku forums have kept following a very different, example-based approach. Various types of rules have been proposed, in application-specific forms and usually in very informal presentations. Here, we shall examine a single example of such a tentative rule for the following purposes:

– we shall illustrate how putting a few examples together and saying they have the same pattern is a good start for defining a new resolution rule for a given CSP but it is very far from enough for doing this in a non-ambiguous way; at some point, a theoretical analysis is needed; in other terms, the example-based and the theory-based approaches are more complementary than opposite;

– we shall also show how new kinds of rules can be formalised, starting from examples; this will raise the delicate question of boundary cases;

– finally, we shall show how our general $B_?B$ classification allows to measure the impact of application-specific rules on hard instances.

It should be stressed that our purposes are only illustrative of what can be done when a new rule is suspected to have been discovered; this chapter is in no case intended as a review of the "exotic" patterns that may have appeared in forums.

We shall start from the famous EasterMonster Sudoku puzzle (created by jpf). It has long been considered as the hardest puzzle and the first pattern-based elimination of candidates for it was obtained by Steven Kurzhals with a rule he introduced in several Sudoku forums (in a rather sketchy way). Since then, this (now classical) pattern has been known under several names: hidden-pairs loop, sk-loop…

### 13.1. The EasterMonster family of puzzles and the sk-loop

Consider Figure 13.1. Informally, given the content of the four grey cells in each of the blocks at the four corners of the grid, the sk-loop rule says that the following thirteen candidates (crossed in the Figure) can be eliminated:
- in row r2 outside blocks b1 and b3, numbers n3 and n8: n3r2c5, n8r2c5, n8r2c6,
- in row r8 outside blocks b7 and b9, numbers n4 and n5: n4r8c5, n5r8c4,
- in column c2 outside blocks b1 and b7, numbers n4 and n8: n4r5c2, n8r5c2,
- in column c8 outside blocks b3 and b9, numbers n3 and n9: n3r5c8, n9r5c8,
- in block b1 outside the four grey cells, numbers n2 and n7: n2r3c1, n7r1c3,
- in block b3 outside the four grey cells, numbers n1 and n6: nothing,
- in block b7 outside the four grey cells, numbers n1 and n6: n1r7c3, n6r9c1,
- in block b9 outside the four grey cells, numbers n2 and n7: nothing.

From Table 11.5, we know that EasterMonter is in $B_6B$; after these eliminations, it can be solved in $B_2B$. Being in $B_2B$ is far from being easy, but this is clearly much better than being in $B_6B$. So, undoubtedly, the sk-loop is worth some consideration.

*Figure 13.1. EasterMonster (outer candidates in bold), from $B_6B$ to $B_2B$*

Soon after this pattern was discovered, many variants of EasterMonster were found; they all displayed the same set of four blocks forming a rectangle, each with a "cross" of four cells, each of these cells having three numbers, more or less as in EasterMonster. It should be noted that the clues in the central block have no influence on the contents of the 16 cells of the sk-loop, so that many variants can easily be obtained by merely changing them; the only condition is keeping the puzzle minimal (or at least ensuring it has a unique solution).

|   |   |   |   |   |   |   |   |   |
|---|---|---|---|---|---|---|---|---|
| x |   |   |   |   |   |   |   | x |
|   | x | x |   |   |   | x |   |   |
|   |   | x |   |   | x |   |   |   |
|   | x | x |   |   |   |   |   |   |
|   |   |   |   | x |   |   |   |   |
|   |   |   |   |   | x | x | x |   |
|   | x |   |   |   | x |   |   |   |
|   | x |   |   |   | x | x |   |   |
| x |   |   |   |   |   |   |   | x |

*Figure 13.2.* The pattern of given cells in Metcalf's puzzle

|    | c1 | c2 | c3 | c4 | c5 | c6 | c7 | c8 | c9 |    |
|----|----|----|----|----|----|----|----|----|----|----|
| r1 | n5 | **n1** n3<br>**n7** | n1<br>n4 n6<br>n7 | n3<br>n4<br>n7 n8 | n2 n3<br>n6<br>n7 n8 | n2 n3<br>n6<br>n8 | n1 n2<br>n4 n6<br>n8 | **n4** **n6**<br>n8 | n9 | r1 |
| r2 | n3<br>**n4** **n6** | n2 | **n4** **n6**<br>n9 | n1 | n3<br>~~n4~~ ~~n6~~<br>n7 n8 n9 | n3<br>n5 ~~n6~~<br>n8 n9 | **n4** n5 **n6** | n7 | **n4** **n6**<br>n8 | r2 |
| r3 | n1<br>n4 n6<br>n7 | **n1**<br>**n7** n9 | n8 | n4 n5<br>n7 n9 | n2<br>n4 n6<br>n7 n9 | n2<br>n5 n6<br>n9 | n3 | **n4** n5 **n6** | n1 n2<br>n4 n6 | r3 |
| r4 | n1 n2 n3<br>n7 n8 | n4 | n1 n2<br>n5<br>n7 n9 | n6 | n1 n3<br>n8 n9 | n1 n3<br>n8 n9 | n2<br>n5<br>n7 n9 | n3<br>n5<br>n8 n9 | n2 n3<br>n7 n8 | r4 |
| r5 | n1 n2 n3<br>n6<br>n7 n8 | ~~n1~~ n3<br>~~n7~~ n8 n9 | n1 n2<br>n6<br>n7 n9 | n3<br>n4<br>n8 n9 | n5 | n1 n3<br>n8 n9 | n2<br>n4 n6<br>n7 n9 | n3<br>~~n4~~ ~~n6~~<br>n8 n9 | n2 n3<br>n4 n6<br>n7 n8 | r5 |
| r6 | n3<br>n6<br>n8 | n3<br>n5<br>n8 n9 | n5 n6<br>n9 | n2 | n3<br>n4<br>n8 n9 | n7 | n4 n5 n6 | n1 | n3<br>n4 n6<br>n8 | r6 |
| r7 | n1 n2<br>n4<br>n7 | **n1**<br>n5<br>**n7** | n3 | n5<br>n7 n9 | n1 n2<br>n6<br>n7 n9 | n1 n2<br>n5 n6<br>n9 | n8 | **n4** **n6**<br>n9 | n1<br>n4 n6<br>n7 | r7 |
| r8 | **n1**<br>**n7** n8 | n6 | **n1**<br>n5<br>**n7** | n3<br>n5<br>~~n7~~ n8 n9 | ~~n1~~ n3<br>~~n7~~ n8 n9 | n4 | **n1**<br>**n7** n9 | n2 | **n1** n3<br>**n7** | r8 |
| r9 | n9 | **n1**<br>**n7** n8 | n1 n2<br>n4<br>n7 | n3<br>n7 n8 | n1 n2 n3<br>n6<br>n7 n8 | n1 n2 n3<br>n6<br>n7 | **n1**<br>n4<br>**n7** | n3<br>**n6** | n5 | r9 |
|    | c1 | c2 | c3 | c4 | c5 | c6 | c7 | c8 | c9 |    |

*Figure 13.3.* Metcalf's puzzle (outer candidates in bold), from $B_7B$ to $B_4B$

Even before the sk-loop was found in it, EasterMonster was famous not only for its high SER (11.6, the highest known at that time) but also for the quasi-symmetries in its pattern of clues. However, there is another minimal puzzle of still higher SER, with both an sk-loop and a more beautifully symmetric pattern of clues (shown in Figure 13.2). It is Metcalf's puzzle (with its initial resolution state given in Figure 13.3). We have already met it in Table 11.5 as one of the only three known (as of this writing) puzzles in $B_7B$; it also has the highest known SER (11.8) for a puzzle with an sk-loop. It should be noted that, contrary to EasterMonster, there are two blocks in which the pairs of numbers in bold appear in the four grey cells.

The eliminations allowed by the sk-loop are: r5c2≠7, r5c2≠1, r2c6≠6, r2c5≠6, r2c5≠4, r5c8≠6, r5c8≠4, r8c5≠7, r8c5≠1, r8c4≠7. After this, Metcalf's puzzle can be solved in $B_4B$. Again, this is not easy, but still easier than $B_7B$. (Later, we shall see an sk-loop puzzle for which the eliminations do not change its $B_3B$ classification).

**13.2. How to define a resolution rule from a set of examples**

Several descriptions and several proofs of an sk-loop rule have been proposed soon after it was introduced, but as we were no more satisfied by them than by the overall description of the pattern, we have tried to write our own formalisation. However, it is not obvious to define a resolution rule corresponding to a given example or set of examples. We shall ask a few questions that must be answered during this process. This will lead us to provide not one but two non *a priori* equivalent formal interpretations of the sk-loop. (There is a positive aspect of using computers in Sudoku solving: when programming a rule, one has to answer each of these questions. Conversely, the logical conditions that will be defined below for our formal interpretations can be understood as specifications for an implementation.)

The first question is about transforming the constants appearing in one or several examples into variables. It may seem easy to replace specific numbers, rows, columns and blocks by generic variables, but the relations they should have may be ambiguous; the question is, which of the relations present in the examples (e.g. the equality of two candidate-Numbers) are actually meaningful and which are purely coincidental? In our opinion, this can only be settled while trying to prove the rule.

A second group of two non-obvious questions is: which candidates present in the example(s) can be made optional and which additional candidates can be allowed as optional candidates in each of the cells? In chapter 8, we have already seen examples of how to deal with this when we defined the non-degenracy conditions for Subset rules. Even with this elementary case, it was far from obvious for Subsets of size greater than three. The two interpretations of sk-loops introduced in the next subsections answer these two questions. In the EasterMonster example and in all the variants that first appeared, each of the sixteen cells concerned by the sk-loop had exactly three candidates. With the following definitions, they may have 2, 3 or 4. Therefore, each of these interpretations is already a non obvious generalisation of the first known examples. The puzzle in Figure 13.4 shows that this is indeed useful.

Third question: what conditions do ensure that the pattern will not degenerate into something simpler? Anticipating on the definition of a belt of crosses in the next subsection, if all the sets of inner candidates in the 2k crosses are the same and all the sets of horizontal + vertical candidates are the same, then the whole belt degenerates into a set of k Naked-Quads in rows, k Naked-Quads in columns and 2k Naked-Quads in blocks (whether this can really happen is another question).

Fourth question: can the overall structure be generalised? Can it be included in a whole family of patterns (e.g. in the sense that xy-chains or whips of different lengths form a family)? The following two interpretations in terms of belts of crosses or of x2y2-chains both allow *a priori* larger patterns, each with more different physical shapes than the "standard" one. None has been found until now for the 9×9 puzzle (there does not seem to be room enough for them), but there is no obvious reason for not finding any in larger grids.

Fifth question, intimately related to the previous one: what are the building blocks of the overall structure? The following two interpretations rely on different building blocks (crosses *versus* x2y2-segments). The second is easily seen to be *a priori* more "atomic" and more general than the first; it also provides a better understanding of how the rule can be proven. Nevertheless, none of the known examples satisfies the second but not the first.

Sixth question: where should the pattern be classified in a complexity hierarchy? Notice that this is not an abstract question that could be independent of the tentative formulation of the rule. In order to state the rule precisely, such classification decisions have to be made (be it implicitly). For instance, the non-degeneracy condition we formulated for Quads in chapter 8 supposes that Quads are more complex than Triplets and Pairs (in this case, it is not really open to discussion, but it is nonetheless necessary for making the non-degeneracy condition meaningful). In the rating or classification approach of this book, as the sk-loop involves sixteen rc-cells, i.e. sixteen CSP variables, it should be ranked somewhere close to $W_{16}$, $B_{16}$, $gW_{16}$, $gB_{16}$, $SB_{16}$ or $BB_{16}$.

### 13.3. First interpretation of an sk-loop: crosses and belts of crosses

Let us now introduce our first interpretation (the most straightforward one) of the sk-loop by defining its building blocks ("crosses") and the ways (via crosses "aligned" along "spines") they can be combined into a full pattern ("a belt of crosses") allowing eliminations. Our definitions try to be as general as possible, considering the way we shall prove the rule; they go *a priori* much beyond the mere EasterMonster case (and they are meaningful for any grid size).

Definition: a *cross* is defined by the following two sets of data and conditions:

1) a pattern of cells:

– a block b;

– a row r and a column c that both intersect b; the intersection of r and c will be called the "center" of the cross;

– two different cells, each in both row r and block b, and none equal to the center of the cross; they will be called the horizontal ends of the cross;

– two different cells, each in both column c and block b, and none equal to the center of the cross; they will be called the vertical ends of the cross.

The "center" of a cross is a conceptual center, it does not have to be the physical center of block b. However, by a proper puzzle isomorphism, any "cross" can be made to look like a physical cross (whence the name we have chosen for them); in the examples below, they will appear directly as physical crosses in EasterMonster (Figure 13.1), in Metcalf's puzzle (Figure 13.3) and in Tarek's puzzle (Figure 13.4), and only indirectly (i.e. after an iso) in Ronk's puzzle (Figure 13.6). Notice that the above conditions imply that the four "ends of the cross" are different cells (they will be drawn in light grey in the forthcoming Figures).

2) a pattern of candidates in the four ends of the cross:

– two different "horizontal outer" candidate-Numbers;

– two different "vertical outer" candidate-Numbers; (each of them may be equal to an horizontal outer one);

– two different "inner" candidate-Numbers, each different from any of the (horizontal and vertical) outer candidate-Numbers;

– none of the four ends is decided;

– each of the two horizontal ends of the cross contains only inner and horizontal outer candidate-Numbers; each of the inner and horizontal outer candidate-Numbers appears in at least one of the two horizontal ends of the cross;

– each of the two vertical ends of the cross contains only inner and vertical outer candidate-Numbers; each of the inner and vertical outer candidate-Numbers appears in at least one of the two horizontal ends of the cross.

Forgetting the condition on the four undecided ends would lead to invalid eliminations; it is not a consequence of all the other assumptions (not even a practical one), as shown by the following six puzzles from Eleven's collection:

......7..4..18...6.....2.1..4..9...3..9..15.....7...2..6.......83.5.......4.3...8  ER/EP/ED=10.8/10.8/9.9 #10526
12.........6.8...2.8...3...2...65..8...9...4...7...5.......4.9....5..3...6..7...1  ER/EP/ED=10.7/10.7/9.4 #13852

```
..34......5...92..6...7.....1......8.....1.259..5..1...9...58....4.6.....6.3...7.   ER/EP/ED=10.6/1.2/1.2 #23210
1....6.8....7....2.8.3........9....7..2.3....5..1.4....4..56...9......36......1.   ER/EP/ED=10.6/1.2/1.2 #24974
1..45..8..........6.....75..2.4........1.2...9...9.3.........6...7..18...2..8...53..   ER/EP/ED=10.6/1.2/1.2 #26051
12..5...9...7......7.....5..2.1.4.9...9......4.4.6...2.3.2.1.........6..92.....81..   ER/EP/ED=10.6/1.2/1.2 #26342
```

Notice that in the set of 1,662 known (as of this writing) puzzles with an sk-loop, we have found none in which one of the four ends of a cross did not contain any of the outer candidate-Numbers and we have found only one (Tarek's puzzle, displayed in Figure 13.4) in which one of the four ends of a cross did not contain any of the inner candidate-Numbers. Because of this example (and only because the same elimination proof is valid for it), the definition we give here for the pattern of candidates in a cross is slightly more general than that available on our website.

**Figure 13.4.** Tarek's puzzle (#071223170000, outer candidates in bold), from $B_6B$ to $B_3B$

Definitions: two crosses are *row-aligned* [respectively *column-aligned*] if they are in different blocks, they are centred in the same row [resp. column] and they have the same set of horizontal [resp. vertical] outer candidate-Numbers.

Definition: a *spine* for a belt of even length 2k is defined by a sequence of 2k cells in different blocks such that, when repeating the first at the end of the list, two consecutive cells are alternatively in the same row and in the same column (these cells define the global structure of the belts to be defined below).

Definition: a *belt of crosses* of even length 2k is defined by the following data and conditions:
– a spine for a belt of length 2k;
– for each of the cells in the spine, a cross centred on this cell;
– when repeating the first cross at the end of the list, consecutive crosses are alternatively row-aligned and column-aligned;
– the 2k sets of inner candidate-Numbers of the 2k crosses are not all equal; the sets of horizontal + vertical candidates-Numbers of the 2k crosses are not all equal (non-degeneracy condition).

In the EasterMonster, Metcalf's and Tarek's examples, the spine is (r2c2, r2c8, r8c8, r8c2); it forms a square. As of now, no minimal 9×9 puzzle has been found with a belt of crosses based on a spine longer than 4: it is easy to find such spines, but when it comes to placing the clues in such a way that the proper candidate patterns appear in the crosses, it seems there is not enough room on the grid for such structures. On the other hand, there does not seem to be any reason for not finding any in larger grids.

***Theorem 13.1: Given a belt of crosses, one can eliminate:***
*- in each of its blocks: any inner candidate-Number of this block that is not in any end of the cross in this block,*
*- in each "central" row of consecutive row-aligned crosses: any horizontal outer candidate-Number (they are the same for the two crosses) that is not in any end of the crosses in the two blocks,*
*- in each "central" column of consecutive column-aligned crosses: any vertical outer candidate-Number (they are the same for the two crosses) that is not in any end of the crosses in the two blocks.*

Proof: Let us call $C_1$, $C_2$, … $C_{2k}$ the 2k crosses and $b_1$, $b_2$, … $b_{2k}$ their blocks. Suppose we start with two row-aligned crosses. We shall only prove that the inner candidate-Numbers in block $b_1$ outside the ends of $C_1$ can be eliminated; the proofs for the other parts of the theorem are similar.

If none of the horizontal ends of $C_1$ contains any of the inner candidate-Numbers of $C_1$, then one has successively:
- the horizontal ends of $C_1$ together contain both of the horizontal outer candidate-Numbers of $C_1$;
- the horizontal ends of $C_2$ (which are row-aligned with those of $C_1$) contain none of the horizontal outer candidate-Numbers of $C_2$ (which are the same as those of $C_1$);
- the horizontal ends of $C_2$ contain both of the inner candidate-Numbers of $C_2$ (only two candidate-Numbers for two cells);

- the vertical ends of $C_2$ contain none of the inner candidate-Numbers of $C_2$ (block constraint);
- the vertical ends of $C_2$ together contain both of the vertical outer candidate-Numbers of $C_2$ (only two candidate-Numbers for two cells);
…
- the horizontal ends of $C_{2k}$ contain both of the inner candidate-Numbers of $C_{2k}$;
- the vertical ends of $C_{2k}$ contain none of the inner candidate-Numbers of $C_{2k}$;
- the vertical ends of $C_{2k}$ together contain both of the vertical outer candidate-Numbers of $C_{2k}$;
- the vertical ends of $C_1$ contain none of the vertical outer candidate-Numbers of $C_1$;
- the vertical ends of $C_1$ together contain both of the inner candidate-Numbers of $C_1$.

Similarly, if none of the vertical ends of $C_1$ contains any of the inner candidate-Numbers of $C_1$, then the horizontal ends of $C_1$ together contain both of the inner candidate-Numbers of $C_1$. Combining these two results, if there is a branch of $C_1$ whose two ends contain none of the inner candidate-Numbers of $C_1$, then these two candidate-Numbers must be in the ends of the other branch. Finally, using contraposition, if there is a branch of $C_1$ that contains one and only one of the inner candidate-Number of $C_1$, then the other inner candidate-Number of $C_1$ must be in the other branch.

In any case, given the whole belt of crosses, each of the two inner candidate-Numbers must be in one of the four ends of $C_1$. Whence the eliminations of the inner candidate-Numbers in $b_1$ outside the four ends of $C_1$. Qed.

Notice that, if the inner candidates appeared somewhere in the row of a cross in a block, outside the ends of this cross, they could also be eliminated (this is probably pointless in 9×9 puzzles because in all the known examples the center of the block is occupied by a clue, but it might happen in larger grids).

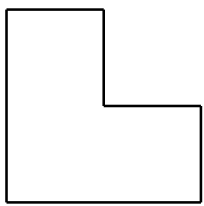

*Figure 13.5. A possible spine for a belt of length 6.*

Remarks on the shape of the spine: with belts of length 4, the spine can only be a rectangle; moreover, as floors and towers can be permuted, there is essentially one spine. However, with belts of length 6, the spine can (must) have different shapes (but always with spine ends in different blocks), e.g. as shown in Figure 13.5.

Finally, in anticipation of section 13.5, notice that a belt of crosses of length 4 can always be seen as a g-Subset of size sixteen, with its sixteen CSP variables defined by the sixteen rc-cells forming the ends of the four crosses and its sixteen transversal sets defined by four rn constraints (the two rows of the centers combined with the two horizontal Number-candidates), four cn constraints (the two columns of the centers combined with the two horizontal Number-candidates) and eight bn constraints (the four blocks of the centers, each combined with its two inner Number-candidates). It is easy to check that this view allows the same eliminations as the belt-of-crosses view, but it is not consistent with the original view of the pattern as a loop (i.e. some kind of closed chain).

### 13.4. Second interpretation of an sk-loop: x2y2-chains

The above interpretation of the sk-loop is the simplest one from the player's point of view, because crosses can easily be seen inside a block; but it hides the symmetrical roles played by blocks and rows or columns in our proof. Whence our second interpretation based on the remark that the proof in the previous section illustrates the x2y2-transfer principle to be enunciated below. Our building blocks will now be "x2y2-segments".

Definitions (classical): a *rowblock* is the intersection of a row and a block; a *colblock* is the intersection of a column and a block; a *segment* is either a rowblock or a colblock.

Definition: an *x2y2-segment* of type row [respectively col] is defined by the following data and conditions:

– a rowblock [resp a colblock] with two distinguished non-decided cells called its ends (as before, they do not have to be the physical ends, they are the conceptual ends);

– two different "left-linking" candidate-Numbers;

– two different "right-linking" candidate-Numbers, each different from any of the left-linking candidate-Numbers;

– each of the two ends of the x2y2-segment contains only left-linking and right-linking candidate-Numbers;

– each of the left-linking and each of the right-linking candidate-Numbers appears in at least one of the two ends of the x2y2-segment.

Definition: two x2y2-segments are *chainable* if:

– the set of right-linking candidate-Numbers of the first is equal to the set of left-linking candidate-Numbers of the second;

– they have no end in common;

– they satisfy either of the following conditions:
- they are both of type row, they lie in the same row but in different blocks,
- they are both of type column, they lie in the same column but in different blocks,
- the first is of type row, the second of type col and they lie in the same block,
- the first is of type col, the second of type row and they lie in the same block,
- they are both of type row, they lie in the same block but in different rows,
- they are both of type col, they lie in the same block but in different columns.

Remarks:

– two chainable x2y2-segments always share a single unit: respectively row and column for the first two cases listed above, and block for the remaining four cases;

– the last two cases introduce completely new possibilities that were not available in the "belt of crosses" view;

– given an x2y2-segment, its reverse can be defined as the x2y2-segment based on the same rowblock [or colblock] and ends, but with the roles of left-linking and right-linking candidate-Numbers interchanged;

– if two x2y2-segments are chainable, then the reversed segments taken in reversed order are chainable.

The basic (and obvious) property of an x2y2-segment is that if none of its left-linking candidate-Numbers is true in any of its two cells, then each of its right-linking candidate-Numbers must be true in one of its two cells.

This readily extends to a basic property of chainable x2y2-segments, where it constitutes what we call *the x2y2-transfer principle (a natural generalisation of the classical xy-transfer principle for bivalue chains)*: if none of the left-linking candidate-Numbers of the first x2y2-segment is true in any of its two cells, then each of the right-linking candidate-Numbers of the second x2y2-segment must be true in one of its two cells. This can in turn be extended to any sequence of chainable x2y2-segments as described below:

| xy-transfer principle | x2y2-transfer principle |
|---|---|
| if not x1 then y1       | if neither of x1 and x'1 then both of y1 and y'1 |
|   then not x2 |   then neither of x2 and x'2 |
|     then y2 |     then both of y2 and y'2 |
|       then not x3 |       then neither of x3 and x'3 |
|         then y3 ... |         then both of y3 and y'3 .... |

Definition: an *x2y2-belt* of length n is a sequence of n different x2y2-segments, $S_1, S_2, ..., S_n$, all different, such that (setting $S_{n+1} = S_1$) for each j in {1, 2, ..., n}, $S_k$ and $S_{k+1}$ are chainable.

Remarks:

– notice the circularity condition (as for belts of crosses);

– the length n does not have to be even;

– if one defines the reversed belt as being the sequence of the reversed segments taken in the reversed order, then it is an x2y2-belt (this is needed in the proof of the following theorem).

Definition: Given an x2y2-belt, the targets of one of its couples of consecutive (therefore chainable) x2y2-segments (still setting $S_{n+1} = S_1$) are the candidate-Numbers equal to the right-linking candidates of the first segment (or the left-linking candidates of the second) and belonging to their unique common unit but to none of their ends.

Definition: The targets of an x2y2-belt are the targets of any of its couples of consecutive (therefore chainable) x2y2-segments (still setting $S_{n+1} = S_1$).

***Theorem 13.2: Given an x2y2-belt, any of its targets can be eliminated.***

The proof is essentially the same as that we gave above for belts of crosses. It is based on iterating the x2y2-transfer principle in both directions and concluding in the same way as in that proof.

Remarks on the shape of the spine:

– given an x2y2-belt, we can define its spine more or less as previously (details are left to the reader);

– with x2y2-belts of length 4, the spine can only be a rectangle, but it can now be "flat", e.g. with four rowblocks in two different rows and two different blocks;

– with belts of length 6, the spine can have new different shapes, e.g. as in Figure 13.5, but with the horizontal or the vertical branch, or both, flattened as in the previous case;

– a question remains open: can one build a (9×9 or larger) puzzle with an x2y2-belt with any of these spines? As of today, no example has been found, but there does not seem to be any reason why this would not be possible for larger grids.

**13.5. Should the above definitions be generalised further?**

Consider the example in Figure 13.6. In blocks b5, b6 and b9, the conditions for a belt of crosses seem to be satisfied with the possible outer candidates in these blocks written in bold (this is the only possibility for a belt). But there is a problem for extending this to block b8: there is only one horizontal outer candidate-Number (n8), only one vertical outer candidate-Number (n3), and there are too many inner candidate-Numbers (either n4 in r7c5 and r8c4 or n9 in r7c6 and r9c4).

Supposing none of n1 and n2 was in r4c5 or r4c6, we can proceed as in section 13.4 until we show that n5 and n8 must be in r7c8 and r7c9. But afterwards, the chain-like reasoning used in the standard case does not allow to conclude than none of n3 and n5 can be in any of r5c4 and r6c4. The x2y2 chain of reasoning is broken; it can only be patched with a piece of reasoning very specific to this situation.

|    | c1 | c2 | c3 | c4 | c5 | c6 | c7 | c8 | c9 |    |
|----|----|----|----|----|----|----|----|----|----|----|
| r1 | n9 | n8 | n1 n2 n3 | n7 | n1 n2 n3 n4 n6 | n2 n3 n5 n6 | n1 n2 n4 n5 n6 | n1 n3 n4 n5 n6 | n2 n3 n5 n6 | r1 |
| r2 | n7 | n1 n2 | n6 | n1 n2 ~~n3~~ n4 n5 n9 | n1 n2 n3 n4 | n2 n3 n5 n9 | n8 | n1 n3 n4 n5 | n2 n3 n5 n9 | r2 |
| r3 | n1 n2 n3 | n5 | n4 | n1 n2 ~~n3~~ n9 | n1 n2 n3 n6 n8 | n2 n3 n6 n8 n9 | n1 n2 ~~n6~~ n9 | n1 n3 ~~n6~~ n7 | n2 n3 n6 n7 n9 | r3 |
| r4 | n6 | n1 n2 n4 ~~n7~~ n9 | n1 n2 n5 ~~n7~~ n9 | n8 | n1 n2 **n7** | n2 **n5** **n7** | n3 | n4 **n5** **n7** | **n5** ~~n7~~ n9 | r4 |
| r5 | n1 n3 n4 n5 n8 | n1 n4 n7 | n1 n3 n5 n7 n8 | n1 **n3** **n5** | n9 | n3 ~~n5~~ n6 n7 | n4 **n5 n6** | n2 | ~~n5~~ n6 n7 n8 | r5 |
| r6 | n2 n3 n5 n8 | n2 n7 | n2 n3 n5 n7 n8 n9 | n2 **n3** **n5** | ~~n2~~ n3 n6 n7 | n4 | **n5 n6** n9 | ~~n5~~ n7 n8 | n1 | r6 |
| r7 | n1 n2 n4 n5 ~~n8~~ | n3 | n1 n2 n5 ~~n8~~ n9 | n6 | n2 **n8** | n2 **n8** n9 | n7 | n1 **n5** **n8** | n2 **n5** **n8** | r7 |
| r8 | n1 n2 n4 n8 | n1 n2 n4 n6 n7 | n1 n2 n7 n8 | n2 **n3** n4 | n5 | ~~n2~~ n3 n7 n8 | n1 n2 **n6** | n9 | ~~n2~~ n3 n6 n8 | r8 |
| r9 | n2 n5 n8 | n2 n6 n7 | n2 n5 n7 n8 n9 | n2 **n3** n9 | ~~n2~~ n3 n7 n8 | n1 | n2 **n5 n6** n8 | n3 ~~n5~~ n6 | n4 | r9 |
|    | c1 | c2 | c3 | c4 | c5 | c6 | c7 | c8 | c9 |    |

**Figure 13.6.** *Ronk's puzzle (outer candidates in bold), from $B_3B$ to $B_3B$*

An alternative approach is to give a definition in terms of g-Subsets, with:

– sixteen CSP variables associated with the sixteen rc-cells at the ends of the crosses and pseudo-crosses (in light grey): Xr4c5, Xr4c6, Xr5c4, Xr6c4; Xr4c8, Xr4c9, Xr5c7, Xr6c7; Xr7c5, Xr7c6, Xr8c4, Xr9c4; Xr7c8, Xr7c9, Xr8c7, Xr9c7;

– and sixteen g-transversal sets defined by the following sixteen g-transversal constraints: two of type rn (r4n7, r7n8), two of type cn (c4n3, c7n6) and twelve of type bn (b5n1, b5n2, b5n5, b6n4, b6n5, b6n9, b8n2, b8n4, b8n9, b9n1, b9n2, b9n5);

one can then eliminate the following sixteen candidates: r4c2≠7, r4c3≠7, r5c6≠5, r5c9≠5, r7c1≠8, r7c3≠8, r8c6≠2, r8c9≠2, r2c4≠3, r3c4≠3, r6c5≠2, r9c5≠2, r1c7≠6, r3c7≠6, r6c8≠5, r9c8≠5 (the number of eliminations is also sixteen by mere chance).

Four of these eliminations would not be justified by a belt of crosses: r5c6≠5, r5c9≠5, r6c8≠5 and r9c8≠5. On the other hand, the following eight eliminations that would have been allowed by a belt of crosses are not justified by the present g-Subset[16]: r4c3≠5, r7c1≠5, r7c3≠5, r2c4≠ 5, r1c7≠5, r8c1≠5, r8c3≠5, r8c3≠2.

The question now is: should the informal sk-loop rule be extended beyond its initial scope and beyond our interpretations in terms of belts of crosses or x2y2-chains, so that it applies to this puzzle? The only obvious way this could be done is by redefining it as a particular kind of g-Subset[16] based on sixteen CSP variables associated with cells forming a pattern of crosses on the grid, as described above; it does not seem that any general condition on the g-transversal sets could be added to that defined by the general concept of a g-Subset[16]. As shown by the statistics in Tables 8.1 and 11.1, Subsets are very inefficient compared to braids of same size; so, one should find very good reasons (such as the frequency of occurrence of this generalised pattern – but it seems to be very rare) before swapping to such a new definition. In any case, if such an extension was adopted, the name sk-loop should certainly have to be changed (at least to "generalised sk-loop" or "mutant sk-loop"), as the initial loop idea that led to its definition would be completely lost. Our purpose here is not to provide a final answer, but only to illustrate the kind of questions that arise when trying to formalise new resolution rules.

### 13.5.1. Another $S_2$-braid example

Incidentally, as mentioned in section 9.7.1, this puzzle provides nice examples of $S_2$-braids. From the initial state in Figure 13.6, no elimination can be done by a braid or a g-braid. The first patterns we find are sixteen $S_2$-braids, corresponding to the eliminations allowed by the g-Subset[16] and then nothing more can be done with $S_2$-braids (the puzzle being in $B_3B$, the next elimination must be an $S_3$-braid).

$S_2$-braid[14]: b9n3{r9c8 r8c9} – b9n6{r8c9 r789c7} – b9n8{r8c9 r7c789} – {n8r7c5 NP: b8{r7c5 r8c4}{n2 n4}} – r7c6{n2 n9} – r9c4{n9 n3} – {n3r5c4 HP: b5{r5c6 r6c5}{n3 n6}} – b5n7{r5c6 r4c456} – r4c8{n7 n4} – r5c7{n4 n5} – r4c9{n5 n9} – r6c7{n9 .} ==> r9c8 ≠ 5

Let us now provide some explanatory detail, after introducing a few markers:

– a number between brackets after each right-linking object in the braid;

– independent numberings of these candidates or patterns for different branches of the braid, each of them starting after the number from which it branches out; here, there is only one main branch (1, 2, ... 11) plus a small secondary branch (2'); this $S_2$-braid is almost an $S_2$-whip;

– explicit addition of z- and t- candidates, with symbol "*" for a z-candidate and with symbol "#n" for a t-candidate (with n = the number of the previous right-linking candidate or pattern to which it is linked); remember however that these candidates are not part of the braid.

$S_2$-braid[14]: b9n3{r9c8 r8c9$_{(1)}$} – b9n6{r8c9 r789c7$_{(2')}$ r9c8*} – b9n8{r8c9 r7c789$_{(2)}$ r9c8*} – {n8r7c5 NP: b8{r7c5 r8c4}{n2 n4}$_{(3)}$ n3r8c4$_{\#1}$} – r7c6{n2 n9$_{(4)}$ n8$_{\#2}$} – r9c4{n9 n3$_{(5)}$ n2$_{\#3}$} – {n3r5c4 HP: b5{r5c6 r6c5}{n3 n6}$_{(6)}$ n3r6c4$_{\#5}$} – b5n7{r5c6 r4c456$_{(7)}$ r6c5$_{\#6}$} – r4c8{n7 n4$_{(8)}$ n5*} – r5c7{n4 n5$_{(9)}$ n6$_{\#2'}$} – r4c9{n5 n9$_{(10)}$ n7$_{\#7}$} – r6c7{n9 . n5$_{\#9}$ n6#2'} ==> r9c8 ≠ 5

Remember that, by definition, $S_p$-braids may include g-candidates as right-linking objects. They appear here in cells (2'), (2) and (7).

It can be checked that all the other eliminations allowed by the g-Subset[16] can be done by similar $S_2$-braids.

### 13.6. Measuring the impact of an application-specific rule

Three complementary aspects of the impact of a specific rule can be considered:

– How often does it appear in instances of the CSP? The answer obviously depends on how we classify the specific rule with respect to the general purpose ones, but all the known cases of Sudoku-specific rules appear very rarely if we put them after whips or braids (g-whips, g-braids) of same size. This can be verified for sk-loops (which rarely interact with generic rules) and even for the nonspecific rules for Subsets (where it is a consequence of the general subsumption theorems).

– For instances in T&E(1) or gT&E(1), how much can it modify their W, B, gW or gB ratings? We have already seen examples in Sudoku of how allowing Subsets can (very rarely) either slightly decrease the W rating of a puzzle or make it solvable by whips when it was not without the Subset rules.

– For instances in T&E(2), the previous question becomes: how much can it change their $B_?B$ classification? In this section, we shall concentrate on this type of impact and we shall consider the sk-loop example (in its belt-of-crosses formalisation).

As shown by the Subset examples in chapter 8, there does not seem to be much correlation between the intrinsic complexity of a new rule and its possible impact on rating. The reason is that there seems to be no limit to the possible modifications the elimination of a single candidate or a few ones can entail. The same goes for the impact on classification: in this chapter, we have seen that, after eliminating all their sk-loop targets, Metcalf's puzzle moves from $B_7B$ down to $B_4B$, EasterMonster from $B_6B$ to $B_2B$, Tarek's puzzle from $B_6B$ to $B_3B$, while Ronk's puzzle remains in $B_3B$; the example in Figure 13.7 (SER 10.5, obtained by adding a diagonal clue to Metcalf's puzzle, thus making it non-minimal, but looking like a snowflake) moves from $B_3B$ to gB. As a result, the impact of an application-specific rule can only be evaluated either statistically (when unbiased collections of instances are available) or on individual instances. For Subsets, we could give statistical evaluations; but, as of this writing, for the sk-loop, we can only analyse its individual impact.

*Figure 13.7. Snowflake, a non-minimal variant of Metcalf's puzzle, from $B_3B$ to gB*

### 13.7. Can an (apparently) application-specific rule be made general?

In the first parts of this book, we have shown that many rules that originated in Sudoku could be re-written in such a way that they are meaningful for any CSP:
– all our (standard and generalised) whip and braid rules,
– the classical Subset rules (and the finned fish, which are mere S-whips),
– the classical Franken and Mutant Fish (which now appear as g-Subset rules).

Can an application-specific resolution rule be made general? The sk-loop example and the discussion in section 13.5 together indicate that if such is the case, it may have to be at the cost of loosing much of the general idea at the origin of the rule (in the present case, even if the basic geometrical structure is kept unchanged).

# 14. Transitive constraints and Futoshiki

Futoshiki (literally "inequality" in Japanese), another logic puzzle, appeared a few years ago and is becoming relatively popular in Japan (but still much less than Sudoku). It is interesting in the context of this book for the following reasons:

– contrary to our main Sudoku example, it has more constraints (inequality constraints between the contents of adjacent cells) than the "strong" ones due to its CSP variables;

– in its "pure" form, it has no clue other than such inequality constraints (i.e. it has no predefined value for any of the cells);

– contrary to the Sudoku constraints, the inequality constraints are asymmetric;

– although the set of CSP variables is fixed as in LatinSquare or Sudoku, the set of the inequality constraints depends on the set of given inequalities;

– Futoshiki has g-labels that, given an instance, do not depend on its resolution state, in conformance with our general definition (g-labels are structural); but, unlike Sudoku, they depend on the instance under consideration (unless one wants to introduce a whole set of universal but useless g-labels);

– contrary to Sudoku, g-labels involve sets of values instead of sets of cells.

In spite of all these noticeable differences, we shall show that our approach is quite relevant to it, even for very hard instances.

## 14.1. Introducing Futoshiki and modelling it as a CSP

### 14.1.1. Definition of Futoshiki

Like an n×n Sudoku, an n×n Futoshiki is a special kind of n×n Latin Square. An n×n Futoshiki puzzle requires the placement of numbers from 1 to n in the cells of an n×n square grid in such a way that each of these numbers appears only once [and therefore exactly once] in each row and in each column. Unlike Sudoku, grid size n does not have to be the square of some integer m ($n = m^2$) and there are no m×m or any other block constraints (in this respect, it is much closer to LatinSquare than to Sudoku).

However, in any instance of Futoshiki, there are specific inequality constraints between elements in rows and columns, as in the example of Figure 14.1. A strict inequality sign between two contiguous cells in a row [respectively a column] means that the values in these two cells must be related by this inequality. These signs can appear graphically in four different shapes (<, >, ∧, ∨), but it should be stressed that they all have the same "strictly less than" meaning, they define only one new type of constraint; their appearance is only used to state graphically in which order the two cells are involved in the inequality. [14]

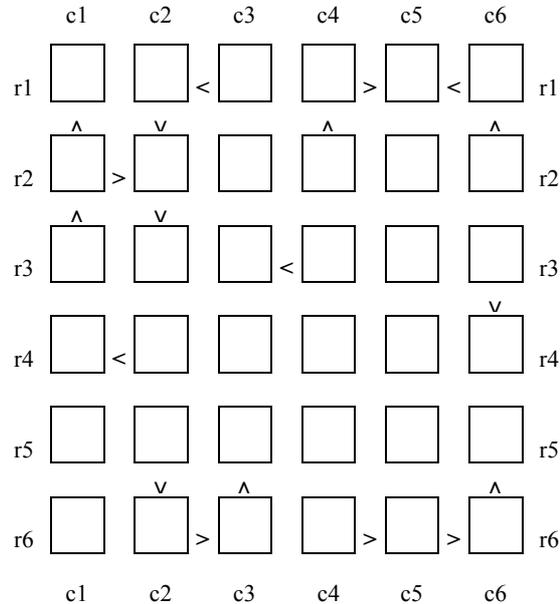

*Figure 14.1.* A 6×6 Futoshiki puzzle (clues of #M5121, from atksolutions.com)

As in Sudoku or in any logic puzzle with a reasonable definition, a "well-formed" puzzle is supposed to have a unique solution. Sometimes, clues are given in some of the cells (with the obvious meaning that they should be their final values, as in Sudoku); but this is not compulsory: inequality constraints can be enough to ensure uniqueness of a solution, as in Figure 14.1.

---

[14] There is a variant of Futoshiki (also unnamed, as far as we know) in which inequality signs are supposed to relate any two cells in different contiguous sectors in the same row [or column] – where a sector is defined as a contiguous set of cells delimited by two such signs. As it does not call for a radical change to the analyses of this chapter, we shall not consider it here.

We shall call "pure Futoshiki" a puzzle with no other clue than inequalities, although we are not aware of any name being currently used to make such a distinction. In the following, all our examples will be pure Futoshikis, but this does not change anything to their discussion. Our personal preference for pure Futoshiki is related to its pure geometric aspect (and also to the fact that, in the context of this book, this makes it look more complementary to our main Sudoku example); however, this will remain abstract, as we shall not investigate the kinds of geometric properties of the set of inequality signs that might have implications on the solution (and which type of implications, if any).

(In the "impure" case, i.e. if both digits and inequalities can be given, from a theoretical or CSP point of view, especially if one wants to define minimal instances, both predefined values and inequality constraints should be put on the same footing and considered as clues. Without this precision, there might be an ambiguity on the interpretation of "minimal": should one consider minimality with respect to a fixed set of inequalities or should one consider both types of clues as one set – each choice raises a few questions of its own.)

Futoshiki has obvious symmetries, some from LatinSquare (row-column symmetry, reflection) and some related to inequalities. If P is an n×n Futoshiki puzzle [or complete grid] and if P' is obtained from P by reversing all the inequality signs and replacing every Number k by n-k+1, then P' is an n×n Futoshiki puzzle [or complete grid]. But rows [or columns] can obviously not be permuted.

### *14.1.2. The sorts, CSP variables, labels and constraint types of Futoshiki*

Futoshiki has Number, Row and Column sorts similar to those of Sudoku, but with ranges corresponding to the grid size. There is a predicate "<" with signature (Number, Number), with the axiom of transitivity and with axiom n1 < n2 < n3 < …

The "natural" CSP variables of k×k Futoshiki are the $k^2$ $X_{rc}$ variables, with r in {r1, …, rk} and c in {c1, …, ck}: in the original formulation, one value in {1, …, k} must be found for each of them; in the formalisation, one value of sort Number must be found. However, as in Sudoku, one can define the rn and cn representations and corresponding $X_{rn}$ and $X_{cn}$ CSP variables, bringing the total number of CSP variables to $3 \times k^2$. Notice that there are no "block-number", i.e. no $X_{bn}$, CSP variables. Accordingly, one can define an extended Futoshiki board, with rc, rn and cn cells representing the $X_{rc}$, $X_{rn}$ and $X_{cn}$ variables, respectively.

Labels are defined as (n, r, c) triplets (also notated nrc), as in Sudoku or LatinSquare. There are thus $k^3$ labels. Label nrc or (n, r, c) is the equivalence class of the three pre-labels: <Xrc, n>, <Xrn, c>, <Xcn, r>. For details, see chapter 2, the only difference being the absence of pre-labels corresponding to Xbn CSP variables.

There are four Constraint-Types: rc, rn, cn, < (not to be confused with predicate "<" on Numbers). In a graphical representation, the < constraint may appear in four different shapes: (<, >, ∧, ∨) but this is one and only one Constraint-Type.

Constraints of type rc, rn, cn between labels are exactly as in Sudoku. As for the inequality Constraint-Type, it may seem to introduce essentially asymmetric relations and one may wonder how it can be modelled as a set of symmetric links between labels, as required by our CSP modelling approach. But this is straightforward:

– for each row $r°$, for each pair of cells $r°c_1°$ and $r°c_2°$ related by the inequality sign < in $r°$ (in any of the two shapes it can take in a row: <, >), the inequality constraint between these two cells is completely taken into account by the set of ground atomic formulæ "linked-by($n_1°r°c_1°$, $n_2°r°c_2°$, <)" for all the Numbers $n_1°$ and $n_2°$ such that $n_1° \geq n_2°$.

– for each column $c°$, for each pair of cells $r_1°c°$ and $r_2°c°$ related by the inequality sign < in $c°$ (in any of the two shapes it can take in a column: ∧, ∨), the inequality constraint between these two cells is completely taken into account by the set of ground atomic formulæ "linked-by($n_1°r_1°c°$, $n_2°r_2°c°$, <)" for all the Numbers $n_1°$ and $n_2°$ such that $n_1° \geq n_2°$.

Futoshiki has the same (Naked, Hidden and Super-Hidden) Subsets and Subset rules as LatinSquare (said otherwise, it has the same Subset rules as Sudoku, except those based on blocks). Futoshiki has whips of length 1 (as shown by the forthcoming example) and therefore it has g-labels (see section 14.5 for details).

Finally, there is nothing special to say about its Basic Resolution Theory, except that, in its "pure" form (i.e. with no predefined values), contrary to Sudoku, its elementary constraint propagation rules (ECP), which take into account not only the rc, rn and cn constraints, but also the < constraint, can eliminate no candidate at the start; all the $k^3$ labels will therefore appear as candidates in the initial resolution state $RS_P$ of any puzzle P. As a result, no minimal "pure" Futoshiki puzzle can be solved in BRT(Futoshiki); but this in itself entails no other significant difference.

## 14.2. Ascending chains and whips

Futoshiki has a very simple and well known rule that does not seem to have any standard name; we shall call it the ascending-chain rule; it is usually considered as a rule of its own and, as far as we know, it has never before been noticed that it corresponds to the interaction rules of Sudoku and that it can be simulated by a mere repetition of the whip[1] rule.

*14.2.1. The weak and strong forms of the ascending chain rule*

Definition: in n×n Futoshiki, an *ascending chain of length k* (1 < k < n) is a sequence of k+1 cells, each adjacent in its row or column to the next one and related to it by the "<" sign.

Notice that these cells may all be in the same row [such as (r3c3, r3c4) in Figure 14.1], or in the same column [such as (r1c1, r2c2, r2c3)], but they may also be spread on several rows and columns [such as (r3c2, r2c2, r1c2, r1c3) or (r5c2, r6c2, r6c3, r5c3)]. The definition of length (k if the chain lies on k+1 cells) will be justified by theorem 14.2.

***The ascending chain rule (weak version)***: in n×n Futoshiki, if $(C_0, C_1, …, C_k)$ is an ascending chain of length k, then, for any i with $0 \leq i \leq k$, k candidate-Numbers can be deleted from $C_i$, namely:
- the i candidate-Numbers j with $1 \leq j \leq i$;
- the k-i candidate-Numbers j with $n-(k-i)+1 \leq j \leq n$.

There are several obvious consequences. In a well-formed n×n Futoshiki puzzle:
- there can be no ascending chain of length n or greater;
- an ascending chain of length n-1 completely determines the values of all its cells.

Indeed, this is the (more or less) standard formulation of the rule, but a stronger version is often needed in practice.

***The ascending chain rule (strong version)***: in n×n Futoshiki, if $(C_0, C_1, …, C_k)$ is an ascending chain of length k and if, in the current resolution state, $m_0$ is the smallest candidate for $C_0$ and $M_k$ is the largest candidate for $C_k$, then:
- for any i with $0 < i \leq k$, all the candidate-Numbers j with $1 \leq j \leq m_0+i-1$ can be deleted from $C_i$;
- for any i with $0 \leq i < k$, all the candidate-Numbers j with $M_k-(k-i)+1 \leq j \leq n$ can be deleted from $C_i$.

The proof of both versions is straightforward, either directly (by counting the number of smaller / greater values there must be in the other cells of the chain) or as a corollary to theorem 14.1 below.

Exercise (easy): write the proof of the strong version.

Besides taking into consideration the minimum or maximum values still present in the endpoints in the current resolution state, the strong version of the ascending chain rule differs from the weak one by one more point. The latter could be restricted with no damage to maximal ascending chains, but the former gets its full strength only if it can be applied to non-maximal ones. Consider for instance the chain in Figure 14.2 (the cells $C_0, C_1, …, C_5$ do not have to be in the same row or column, they only have to be related by <). Each cell is displayed with the candidates remaing after the weak rule has been applied. Suppose now that some other rule application deletes n3 in $C_2$. Then, by considering the sub-chain $C_3, \ldots, C_5$, the strong rule can delete n4 in $C_3$, n5 in $C_4$ and n6 in $C_5$. Similarly, if n6 was deleted from $C_2$ by another rule, then the strong rule could delete n5 from $C_2$ and n4 from $C_1$. Of course, these eliminations could also be done by whips[1] (using theorem 14.1), but this example shows that only the strongest form of the ascending chain rule captures its full power.

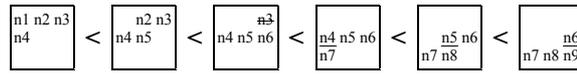

**Figure 14.2.** *A symbolic representation of the strong ascending chain rule.*

### 14.2.2. Ascending chains vs whips[1]

**Theorem 14.1:** *Any elimination done by the ascending chain rule (weak or strong version) can be obtained by a sequence of applications of the whip[1] rule.*

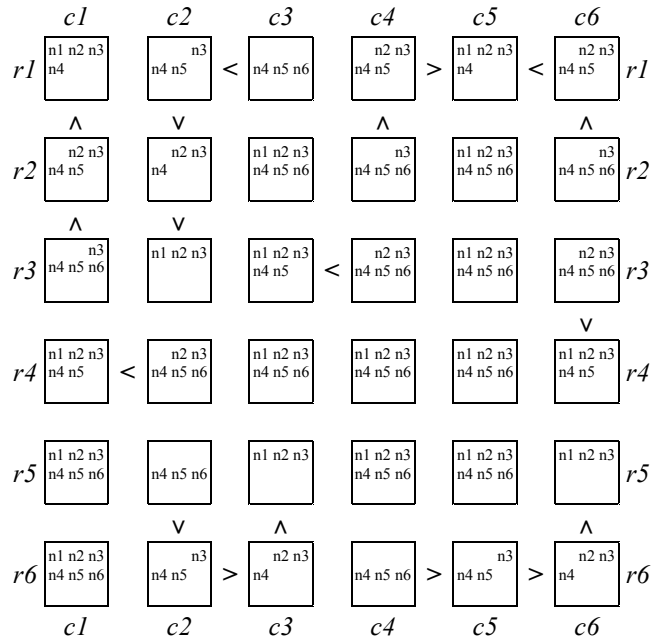

**Figure 14.3.** *Resolution state $RS_1$ of the puzzle in Figure 14.1*

Conversely, it is not true that any whip[1] comes in this way from an ascending chain. Without entering into details, think of a puzzle with constraint r1c4 < r1c5 and a resolution state with the following kind of whip[1]: r1n3{c4 .} ==> r1c5 ≠ 2.

Proof for the weak version: for the first series of eliminations described in the definition of the ascending-chain rule, proceed upwards from number 1 to number k and for each of these numbers backwards from cell $C_k$ to cell $C_1$; for the second series of eliminations, proceed downwards from number n down to number n-(k-i)+1 and for each of these numbers forwards from cell $C_0$ to cell $C_{k-1}$.

The proof of theorem 14.1 will be better understood after reading the following example, in which we show how the whip[1] rule applies to the easy puzzle in Figure 14.1 to make a lot of ascending-chain eliminations (we choose the whips[1] in the order mentioned in the proof).

***** Manual solution *****
;;; concentrating first on the upper-left corner
whip[1]: r1c6{n1 .} ==> r2c6 ≠ 1; whip[1]: r1c5{n1 .} ==> r1c6 ≠ 1; whip[1]: r1c6{n2 .} ==> r2c6 ≠ 2
whip[1]: r1c6{n6 .} ==> r1c5 ≠ 6; whip[1]: r2c6{n6 .} ==> r1c6 ≠ 6; whip[1]: r1c6{n5 .} ==> r1c5 ≠ 5
whip[1]: r1c4{n1 .} ==> r2c4 ≠ 1; whip[1]: r1c5{n1 .} ==> r1c4 ≠ 1; whip[1]: r1c4{n2 .} ==> r2c2 ≠ 2
whip[1]: r2c4{n6 .} ==> r1c4 ≠ 6
... lots of similar eliminations related to the remaining ascending chains

Figure 14.3 shows the state $RS_1$ reached after all these rules have been applied. Starting from resolution state $RS_1$, we now have the following resolution path. Notice that, if $RS_1$ was not merely taken as our starting state, some of the following Single rules could be applied earlier in the path. As usual, we do not write the ECP rule firings, but they are applied whenever possible, immediately after the Singles. They include not only constraint propagation according to the rc, rn and cn constraints, but also according to the inequality constraint, in conformance with the general definition of BRT(CSP) in section 4.3.

*****    FutoRules 1.2  based on CSP-Rules 1.2, config: W   *****
singles: r6c1 = 1, r1c5 = 1, r1c3 = 6, r3c2 = 1, r2c3 = 1, r6c4 = 6
whip[1]: r1c1{n2 .} ==> r2c1 ≠ 2; whip[1]: r2c1{n3 .} ==> r3c1 ≠ 3; whip[1]: r3c2{n5 .} ==> r3c4 ≠ 5
whip[1]: r5c3{n2 .} ==> r6c3 ≠ 2; whip[1]: r6c3{n3 .} ==> r6c2 ≠ 3; whip[1]: r6c2{n4 .} ==> r5c2 ≠ 4
whip[1]: r3c4{n5 .} ==> r3c3 ≠ 5
singles: r4c3 = 5, r6c6 = 2, r5c6 = 1, r4c4 = 1, r2c2 = 2
whip[1]: r1c6{n3 .} ==> r2c6 ≠ 3; whip[1]: r4c6{n3 .} ==> r3c6 ≠ 3; whip[1]: r4c1{n1 .} ==> r4c2 ≠ 1
whip[1] r2c4{n5 .} ==> r1c4 ≠ 5

;;; Resolution state $RS_2$, displayed in Figure 14.4. After $RS_2$ is reached, the simplest rules are short whips[2].

whip[2]: c2n3{r1 r4} – c6n3{r4 .} ==> r1c1 ≠ 3, r1c4 ≠ 3 (this is also an XWing)
whip[2]: r1n5{c2 c6} – r1n3{c6 .} ==> r1c2 ≠ 4
whip[2]: r1c1{n4 n2} – r1c4{n2 .} ==> r1c6 ≠ 4 (this is also a Naked Pair)

whip[2]: c6n5{r2 r3} – c6n6{r3 .} ==> r2c6 ≠ 4
whip[2]: c1n5{r3 r5} – c1n6{r5 .} ==> r3c1 ≠ 4
whip[2]: r3n3{c3 c5} – r3n2{c5 .} ==> r3c3 ≠ 4
singles to the end

Exercise: check these whips on Figure 14.4.

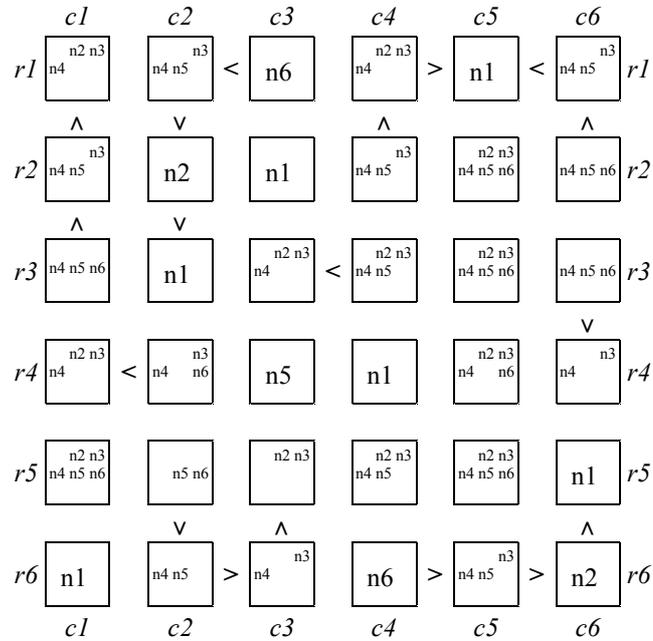

**Figure 14.4.** *Resolution state $RS_2$ of the puzzle in Figure 14.1*

### 14.2.3. Remarks on the rating of ascending chains

Depending on how we consider the ascending-chain rule, we may be tempted to assign it different ratings. If we decompose it as above – as a sequence of whips[1] – the W rating of this part of the resolution path is 1; otherwise, if we consider it as an independent rule, it seems we should assign it a rating equal to the length of the chain. This reflects an unavoidable difference in viewpoints:

– either one prefers "atomic" rules (here whips[1]) to which more complex ones can be reduced and one has to apply them multiple times (this is the approach followed in this book);

– or one prefers to define more complex rules (here the ascending-chain rule) each application of which leads to a large sets of eliminations.

The good solution in our view is that one can use the ascending chain rule in its original form (which is much easier to apply systematically), but remember that it is equivalent to a sequence of whips[1] and therefore grant it rating 1, independent of length. This is an interesting example of rule reduction, because the real underlying complexity (supposing our view of rating is still based on the hardest step) is drastically less than might appear from a quick look at the usual formulation of the rule. This is also more consistent with our intuition of simplicity.

Notice also that, in a pure Futoshiki puzzle, there always are initial ascending-chain eliminations and that these could be considered as obvious domain restrictions; one could decide to systematically choose as initial resolution state for a puzzle P the $RS_1$ (obtained immediately after all these restrictions) instead of the usual $RS_P$ of the general theory (consisting of all the candidates in all the undecided cells).

As a result of the ascending chain rule, many extreme values (1 and n and those close to them) will often be eliminated before the medium ones. This introduces an interesting asymmetry between extreme and medium values and it suggests the heuristics of trying to place or eliminate first the extreme values. However, as any heuristics, its efficiency should be tested by statistical studies, for which this chapter can have no pretension: there is no available generator of Futoshiki puzzles, *a fortiori* no controlled-bias one. Notice that, in "pure" n×n Futoshiki, as long as only this rule (and the hill and valley rules) is applied, the set of candidates for any cell can have no "hole": it can only be a full sub-interval $[k_1, …, k_2]$ of $[1, …, n]$.

**14.3. Hills, valleys and S-whips**

One can obtain more eliminations by combining two different ascending chains that both live completely in a single row or column, provided that they form a "valley" or a "hill" in this row or column; these eliminations can only be done at the top of the hill or at the bottom of the valley. It seems these classical rules have no standard name, but "hill" and "valley" sound appropriate.

*14.3.1. The hill rule and the valley rule*

Definitions: a *hill* is a pair of ascending chains $(C_0, C_1, …, C_k)$ and $(C'_0, C'_1, …, C'_{k'})$ of lengths k and k', all completely in the same row [or column], such that $C_k = C'_{k'}$ and $(C_0, C_1, …, C_{k-1})$ and $(C'_0, C'_1, …, C'_{k'-1})$ are disjoint. A *valley* of length l is a pair of ascending chains $(C_0, C_1, …, C_k)$ and $(C'_0, C'_1, …, C'_{k'})$ of lengths k and k', all completely in the same row [or column], such that $C'_0 = C_0$ and $(C_1, C_2, …, C_k)$ and $(C'_1, C'_2, …, C'_{k'})$ are disjoint. The length of the hill or valley is defined as $l = k + k'$.

***The hill rule (weak form)***: in n×n Futoshiki, if $(C_0, C_1, …, C_k)$ and $(C'_0, C'_1, …, C'_{k'})$ form a hill, then one can eliminate from $C_k$ the $k+k'$ candidate-Numbers between 1 and $k+k'$ included.

***The valley rule (weak form)***: in n×n Futoshiki, if $(C_0, C_1, …, C_k)$ and $(C_0, C'_1, …, C'_{k'})$ form a valley, then one can eliminate from $C_0$ the $k+k'$ candidate-Numbers between $n-(k+k')+1$ and $n$ included.

Proof: by counting the number of cells that must have a smaller value than $C_k$ [respectively a larger value than $C_0$].

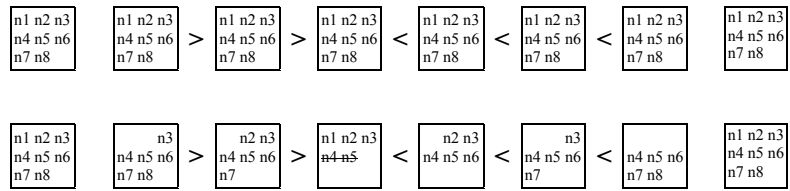

***Figure 14.5.*** *Illustration of the valley rule in an 8×8 Futoshiki, n = 8, k = 2, k'=3, k+k' = 5, n-(k+k')+1 = 4; candidates 8 to 4 can be eliminated from the fourth cell. First line, before any elimination. Second line, after the whip[1] (or ascending chain) eliminations of section 14.2.1. In the second line, the two crossed candidates are eliminated by the valley rule.*

***The hill rule (strong form)***: in n×n Futoshiki, if $(C_0, C_1, …, C_k)$ and $(C_0, C'_1, …, C'_{k'})$ form a hill and if, in the current resolution state, m is the smallest candidate-Number still present in $C_0$ or $C'_0$, then one can eliminate from $C_k$ the $m+k+k'$ candidate-Numbers between 1 and $m+k+k'-1$ included.

***The valley rule (strong form)***: in n×n Futoshiki, if $(C_0, C_1, …, C_k)$ and $(C_0, C'_1, …, C'_{k'})$ form a valley and if, in the current resolution state, M is the largest candidate-Number still present in $C_k$ or $C'_{k'}$, then one can eliminate from $C_0$ the $M+k+k'$ candidate-Numbers between $n-M-(k+k')+1$ and $n$ included.

The proofs of the strong forms are similar to those of their weak forms. Moreover, the remarks we made about the two forms of ascending chains can be transposed in an obvious way to hills and valleys.

*14.3.2. Hills, valleys and S-whips*

***Theorem 14.2:*** *Any elimination done by the hill or the valley rule using ascending chains of lengths k and k' can be obtained by the application of* *whips[1] and/or S-whips of total length no more than k+k' with a single inner Subset of size no more than k+k'-1. If k = 1 (or k' = 1), it can be obtained by a whip[k+k'].*

Proof: it is enough to prove the hill or valley eliminations that cannot be done by the simpler ascending chain rule. The case k = 1 or k' = 1 is obvious. For clarity, we shall prove the general case (k ≠ 1 and k' ≠ 1) only in the example of Figure 14.4, i.e. the eliminations rc4 ≠ 5 and rc4 ≠ 4:

whip[4]: rc5{n5 n6} – rc6{n6 n7} – rc7{n7 n8} – rc3{n8 .} ==> rc4 ≠ 5
$S_4$-whip[5]: r{c5n4 $S_4${c5 c6 c7 c3}{n5 n6 n7 n8}} – rc2{n8 .} ==> rc4 ≠ 4

Notice that, contrary to the ascending chain rule, the hill or valley rules cannot in general be reduced to combinations of elementary rules. Therefore, their place in the complexity hierarchy should be defined by their length. However, as our purposes in this chapter are only illustrative, we put them all just after whip[1], i.e. just after (both weak and strong) ascending chains.

This theorem justifies our definition of the "length" of a hill or valley: whether we consider it as such or as an S-whip, we get the same length. In Sudoku, an S-whip is a relatively complex pattern. It is interesting that Futoshiki provides very natural and easy to find instances of it (but there may be more complex ones, similar to those in Sudoku).

### 14.4. A detailed example using the hill rule, the valley rule and Subsets

Let us now illustrate this rule with the (relatively hard) pure 7×7 Futoshiki puzzle defined by its <, >, ∧ and ∨ inequality symbols in Figure 14.6. Contrary to the example in section 14.2, we now use explicit ascending chain rules, both their weak version, notated e.g. asc[3]: r3c7<r2c7<r1c7<r1c6), and their strong version, notated e.g. str-asc[2]: r6c3<r6c2<r5c2. We write their apparent length, but both are fundamentally mere whips[1]. The reason for making a distinction between weak and strong cases is only contingent: the weak version can only be used for initialisation purposes leading to state $RS_1$ while the strong one can only be used later; this is easily done in FutoRules by assigning them different priorities.

In the (easy but tedious) part of the resolution path leading to $RS_1$, the initial ascending chains are applied in a random order, independent of their apparent lengths; the application of Singles has been suspended during this first phase.

The path was generated by FutoRules, our Futoshiki solver based on our general CSP-Rules solver. What we needed to add to CSP-Rules is input-output functions (most of which are the same as in SudoRules) as well as functions for defining the inequality constraints from the data given in the form described in the next paragraph. In order to identify ascending chains, hills and valleys, even in cases where they are subsumed by whips, we also added specific rules for them.[15]

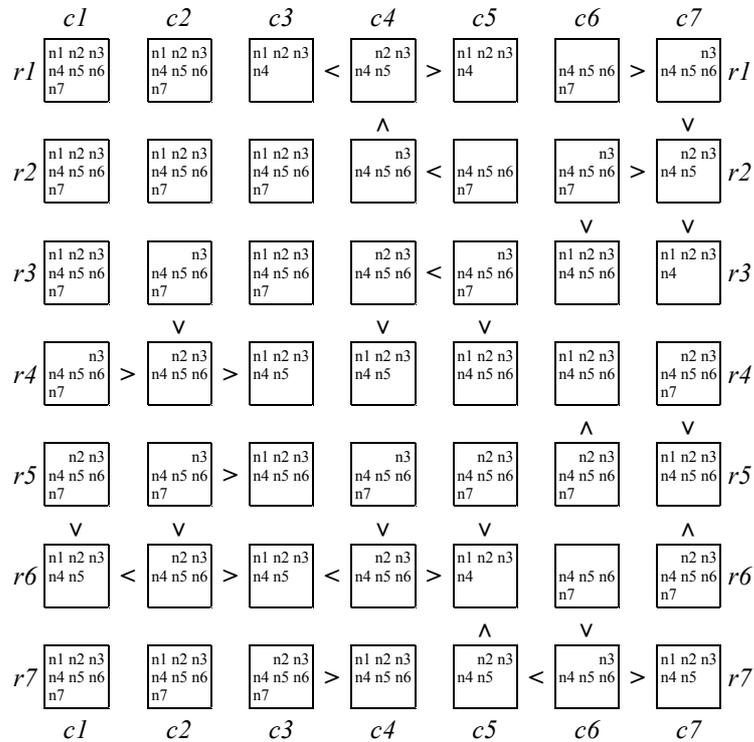

**Figure 14.6.** State $RS_1$ of a 7×7 Futoshiki puzzle (clues of #H662, from atksolutions.com)

   The second, third and fourth lines in the output display the compact representation we use and we recommend for any type (pure or not) of n×n Futoshiki (this is how we feed FutoRules with puzzles). It is made of three series of symbols. The first sequence is for the n×n clues in the cells, exactly as in Sudoku. The second and the third represent inequalities. The second is the sequence of (n-1)×n inequality signs present in the n successive rows, from top to bottom. The third is the sequence of the same number (n-1)×n of inequality signs present in the n successive colums, from left to right. The symbols "." and "-" are placeholders, meaning respectively the absence of any digit or inequality sign. In case one only wants to deal with "pure" Futoshiki, the first sequence can be discarded.

---

[15] To give a rough idea of what is needed for using CSP-Rules to solve another CSP, the total time it took us was two days, including extensive testing. Of course, as grid geometry is the same, forgetting blocks, we could re-use much of what we had done for SudoRules; for very different types of CSPs, in particular with a different geometry, it may take longer. In SudoRules or FutoRules, the specific part (of source code) is about 3% of the generic CSP-Rules part.

\*\*\*\*\*  FutoRules 1.2 based on CSP-Rules 1.2, config: W+S   \*\*\*\*\*
.................................................
--<>->---<.>---<-->>-----.>----<><>---->-<>
---->---.>.>--------<.>-.>---->.><-.>-<.>>>.><-
0 givens, 343 candidates, 3087 csp-links and 3759 links. Initial density = 1.60
asc[1]: r5c7<r6c7 ==> r6c7 ≠ 1, r5c7 ≠ 7
asc[1]: r5c7<r4c7 ==> r4c7 ≠ 1
asc[3]: r3c7<r2c7<r1c7<r1c6 ==> r3c7 ≠ 7, r3c7 ≠ 6, r3c7 ≠ 5, r2c7 ≠ 7, r2c7 ≠ 6, r2c7 ≠ 1, r1c7 ≠ 7, r1c7 ≠ 2, r1c7 ≠ 1, r1c6 ≠ 3, r1c6 ≠ 2, r1c6 ≠ 1
asc[2]: r3c7<r2c7<r2c6 ==> r2c6 ≠ 2, r2c6 ≠ 1
asc[1]: r4c6<r5c6 ==> r5c6 ≠ 1, r4c6 ≠ 7; asc[1]: r3c6<r2c6 ==> r3c6 ≠ 7
asc[3]: r6c5<r7c5<r7c6<r6c6 ==> r7c6 ≠ 7, r7c6 ≠ 2, r7c6 ≠ 1, r7c5 ≠ 7, r7c5 ≠ 6, r7c5 ≠ 1, r6c6 ≠ 3, r6c6 ≠ 2, r6c6 ≠ 1, r6c5 ≠ 7, r6c5 ≠ 6, r6c5 ≠ 5
asc[1]: r6c5<r5c5 ==> r5c5 ≠ 1; asc[1]: r4c5<r3c5 ==> r4c5 ≠ 7, r3c5 ≠ 1
asc[2]: r4c4<r3c4<r3c5 ==> r4c4 ≠ 7, r4c4 ≠ 6, r3c5 ≠ 2, r3c4 ≠ 7, r3c4 ≠ 1
asc[1]: r6c1<r5c1 ==> r6c1 ≠ 7, r5c1 ≠ 1; asc[2]: r7c7<r7c6<r6c6 ==> r7c7 ≠ 7, r7c7 ≠ 6
asc[1]: r7c4<r7c3 ==> r7c4 ≠ 7, r7c3 ≠ 1
asc[2]: r6c5<r6c4<r5c4 ==> r6c4 ≠ 7, r6c4 ≠ 1, r5c4 ≠ 2, r5c4 ≠ 1
asc[2]: r6c3<r6c4<r5c4 ==> r6c3 ≠ 7, r6c3 ≠ 6
asc[2]: r6c3<r6c2<r5c2 ==> r6c2 ≠ 7, r6c2 ≠ 1, r5c2 ≠ 2, r5c2 ≠ 1
asc[2]: r6c1<r6c2<r5c2 ==> r6c1 ≠ 6
asc[1]: r5c3<r5c2 ==> r5c3 ≠ 7
asc[2]: r4c3<r4c2<r3c2 ==> r4c3 ≠ 7, r4c3 ≠ 6, r4c2 ≠ 7, r4c2 ≠ 1, r3c2 ≠ 2, r3c2 ≠ 1
asc[2]: r4c3<r4c2<r4c1 ==> r4c1 ≠ 2, r4c1 ≠ 1
asc[3]: r1c5<r1c4<r2c4<r2c5 ==> r2c5 ≠ 3, r2c5 ≠ 2, r2c5 ≠ 1, r2c4 ≠ 7, r2c4 ≠ 2, r2c4 ≠ 1, r1c5 ≠ 7, r1c5 ≠ 6, r1c5 ≠ 5, r1c4 ≠ 7, r1c4 ≠ 6, r1c4 ≠ 1
asc[3]: r1c3<r1c4<r2c4<r2c5 ==> r1c3 ≠ 7, r1c3 ≠ 6, r1c3 ≠ 5

;;; Resolution state RS$_1$, displayed in Figure 14.6. Starting from RS$_1$, after a Single and a few ascending chains (strong form, just enabled by the Single), we find our first hills and valleys.

hidden-single-in-a-column: r5c4 = 7
str-asc[1]: r5c3<r5c2 ==> r5c3 ≠ 6; str-asc[1]: r6c2<r5c2 ==> r6c2 ≠ 6
str-asc[1]: r6c1<r6c2 ==> r6c1 ≠ 5
str-asc[2]: r6c3<r6c2<r5c2 ==> r6c3 ≠ 5

valley[2]: r4c7>r5c7<r6c7 ==> r5c7 ≠ 6
hill[2]: r6c3<r6c4>r6c5 ==> r6c4 ≠ 2
hill[2]: r6c1<r6c2>r6c3 ==> r6c2 ≠ 2
str-asc[1]: r6c2<r5c2 ==> r5c2 ≠ 3
hill[2]: r1c3<r1c4>r1c5 ==> r1c4 ≠ 2

The rest of the resolution path has nothing noticeable; from the outside, it looks exactly like one for Sudoku. However, it is worth checking the t-candidates in the various whips, because many of them rely on the inequality constraints. As for the presence of Subsets, it is here for illustration purposes only: all these instances are subsumed by whips.

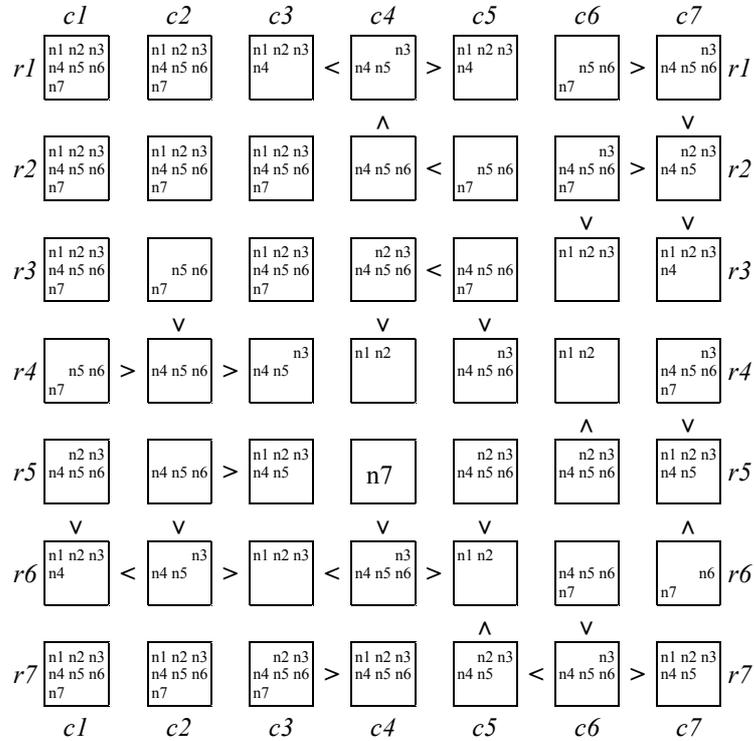

**Figure 14.7.** State $RS_2$ of the 7×7 Futoshiki puzzle in Figure 14.5.

str-asc[1]: r1c4<r2c4 ==> r2c4 ≠ 3; str-asc[1]: r2c4<r2c5 ==> r2c5 ≠ 4
whip[2]: c4n2{r4 r7} – c4n1{r7 .} ==> r4c4 ≠ 3, r4c4 ≠ 4, r4c4 ≠ 5
whip[2]: r4c3{n2 n1} – r4c4{n1 .} ==> r4c2 ≠ 2
str-asc[1]: r4c2<r4c1 ==> r4c1 ≠ 3; str-asc[1]: r4c2<r3c2 ==> r3c2 ≠ 3
whip[2]: c6n2{r4 r3} – c6n1{r3 .} ==> r4c6 ≠ 3, r4c6 ≠ 4, r4c6 ≠ 5
naked-pairs-in-a-row r4{c4 c6}{n1 n2} ==> r4c5 ≠ 1, r4c5 ≠ 2, r4c7 ≠ 2
str-asc[1]: r4c5<r3c5 ==> r3c5 ≠ 3
naked-pairs-in-a-row r4{c4 c6}{n1 n2} ==> r4c3 ≠ 1, r4c3 ≠ 2
str-asc[1]: r4c3<r4c2 ==> r4c2 ≠ 3; str-asc[1]: r4c2<r3c2 ==> r3c2 ≠ 4
str-asc[1]: r4c2<r4c1 ==> r4c1 ≠ 4

whip[2]: c5n1{r6 r1} – c5n2{r1 .} ==> r6c5 ≠ 4, r6c5 ≠ 3
whip[4]: c7n7{r6 r4} – c7n6{r4 r1} – r1c6{n6 n7} – r6n7{c6 .} ==> r6c7 ≠ 5, r6c7 ≠ 4, r6c7 ≠ 3, r6c7 ≠ 2
whip[3]: r6n3{c2 c1} – r6n1{c1 c5} – r6n2{c5 .} ==> r6c3 ≠ 4
whip[6]: c6n1{r3 r4} – c6n2{r4 r5} – c6n3{r5 r7} – r7c5{n3 n2} – r7c7{n2 n1} – c4n1{r7 .} ==> r3c6 ≠ 4, r3c6 ≠ 5, r3c6 ≠ 6
whip[10]: r1c7{n4 n3} – r1c4{n3 n5} – r2c4{n5 n6} – r2c5{n6 n7} – c6n7{r2 r6} – r6n5{c6 c2} – r5c2{n5 n6} – r3c2{n6 n7} – r1n7{c2 c1} – r1n6{c1 .} ==> r1c6 ≠ 4

;;; Resolution state RS$_2$, displayed in Figure 14.7. After RS$_2$ is reached, the longest whip in the path appears.

**whip[12]: r6n7{c6 c7} – r6n6{c7 c4} – r6n5{c4 c2} – r5c2{n5 n6} – r3c2{n6 n7} – c5n7{r3 r2} – c3n7{r2 r7} – r7n6{c3 c1} – r7c6{n6 n3} – r7n5{c6 c4} – r7n4{c4 c2} – r4c2{n4 .} ==> r6c6 ≠ 4**
whip[4]: r6n4{c1 c4} – r6n5{c4 c6} – r6n7{c6 c7} – r6n6{c7 .} ==> r6c2 ≠ 3
str-asc[1]: r6c2<r5c2 ==> r5c2 ≠ 4
whip[2]: r4c2{n6 n4} – r6c2{n4 .} ==> r3c2 ≠ 5
whip[2]: r3c5{n6 n7} – r3c2{n7 .} ==> r3c4 ≠ 6
naked-triplets-in-a-column c2{r4 r5 r6}{n4 n5 n6} ==> r3c2 ≠ 6, r7c2 ≠ 6, r7c2 ≠ 5, r7c2 ≠ 4
singles : r3c2 = 7, r2c5 = 7, r7c3 = 7
str-asc[1]: r4c5<r3c5 ==> r4c5 ≠ 6
whip[2]: r4n7{c1 c7} – r4n6{c7 .} ==> r4c1 ≠ 5
naked-triplets-in-a-column c2{r4 r5 r6}{n4 n5 n6} ==> r1c2 ≠ 6, r2c2 ≠ 4, r2c2 ≠ 5, r2c2 ≠ 6
whip[2]: r1n7{c6 c1} – r1n6{c1 .} ==> r1c6 ≠ 5
whip[2]: r6c6{n6 n7} – r1c6{n7 .} ==> r7c6 ≠ 6
str-asc[1]: r7c7<r7c6 ==> r7c7 ≠ 5; str-asc[2]: r6c5<r7c5<r7c6 ==> r7c5 ≠ 5
whip[2]: c5n6{r3 r5} – c5n5{r5 .} ==> r3c5 ≠ 4
naked-triplets-in-a-column c2{r4 r5 r6}{n4 n6 n5} ==> r1c2 ≠ 4, r1c2 ≠ 5
whip[3]: r7c5{n3 n2} – r7c7{n2 n1} – r7c2{n1 .} ==> r7c6 ≠ 3
whip[5]: r2c6{n3 n6} – r2c4{n6 n4} – r1c4{n5 n3} – r1n4{c5 c1} – r1n5{c1 .} ==> r2c7 ≠ 5
str-asc[1]: r3c7<r2c7 ==> r3c7 ≠ 4
whip[3]: c7n7{r4 r6} – c7n6{r6 r1} – c7n5{r1 .} ==> r4c7 ≠ 4, r4c7 ≠ 3
whip[2]: r4n4{c2 c5} – r4n3{c5 .} ==> r4c3 ≠ 5
whip[4]: c3n6{r2 r3} – c5n6{r3 r5} – r5c2{n6 n5} – c3n5{r5 .} ==> r2c3 ≠ 1, r2c3 ≠ 2, r2c3 ≠ 3, r2c3 ≠ 4

;;; Resolution state RS$_3$

whip[6]: r2n1{c1 c2} – r2n2{c2 c7} – r3c7{n3 n1} – c6n1{r3 r4} – c4n1{r4 r7} – r7n6{c4 .} ==> r2c1 ≠ 6
whip[6]: c3n6{r3 r2} – c3n5{r2 r5} – r5c2{n5 n6} – c5n6{r5 r3} – c5n5{r3 r4} – r4n3{c5 .} ==> r3c3 ≠ 3
whip[7]: r2c4{n5 n6} – r7n6{c4 c1} – r7n5{c1 c6} – r6n5{c6 c2} – r5c2{n5 n6} – c5n6{r5 r3} – c3n6{r3 .} ==> r1c4 ≠ 5
str-asc[1]: r1c3<r1c4 ==> r1c3 ≠ 4
str-asc[1]: r1c5<r1c4 ==> r1c5 ≠ 4
naked-triplets-in-a-row r1{c2 c3 c5}{n1 n2 n3} ==> r1c4 ≠ 3, r1c7 ≠ 3

naked-single: r1c4 = 4
naked-pairs-in-a-row r2{c3 c4}{n5 n6} ==> r2c6 ≠ 5, r2c6 ≠ 6
str-asc[2]: r3c7<r2c7<r2c6 ==> r3c7 ≠ 3, r2c7 ≠ 4
naked-pairs-in-a-row r2{c3 c4}{n5 n6} ==> r2c1 ≠ 5
naked-triplets-in-a-column c7{r1 r4 r6}{n6 n5 n7} ==> r5c7 ≠ 5
naked-triplets-in-a-row r1{c2 c3 c5}{n1 n2 n3} ==> r1c1 ≠ 1, r1c1 ≠ 2, r1c1 ≠ 3
whip[3]: r3c5{n5 n6} – c3n6{r3 r2} – r2n5{c3 .} ==> r3c4 ≠ 5
naked-triplets-in-a-row r3{c4 c6 c7}{n2 n3 n1} ==> r3c1 ≠ 1, r3c1 ≠ 2, r3c1 ≠ 3, r3c3 ≠ 1, r3c3 ≠ 2
whip[3]: c4n1{r7 r4} – c6n1{r4 r3} – r3n3{c6 .} ==> r7c4 ≠ 3
whip[3]: r4n2{c4 c6} – c6n1{r4 r3} – r3n3{c6 .} ==> r3c4 ≠ 2
naked-single: r3c4 = 3
naked-pairs-in-a-column c4{r2 r6}{n5 n6} ==> r7c4 ≠ 6
singles: r7c1 = 6, r4c1 = 7, r1c1 = 5
str-asc[1]: r6c1<r5c1 ==> r6c1 ≠ 4
naked-single : r3c1 = 4
str-asc[1]: r6c1<r5c1 ==> r6c1 ≠ 3
singles: r1c7 = 6, r4c7 = 5, r6c7 = 7, r1c6 = 7, r4c2 = 6
str-asc[2]: r6c3<r6c2<r5c2 ==> r6c2 ≠ 5
str-asc[1]: r5c3<r5c2 ==> r5c3 ≠ 5
singles: r6c2 = 4, r5c2 = 5, r3c5 = 5, r3c3 = 6, r2c3 = 5, r2c4 = 6, r6c4 = 5, r6c6 = 6, r5c5 = 6, r7c6 = 5, r6c3 = 3, r4c3 = 4, r4c5 = 3, r1c2 = 3, r7c7 = 3, r2c7 = 2, r3c7 = 1, r5c7 = 4, r3c6 = 2, r4c6 = 1, r4c4 = 2, r7c4 = 1, r7c2 = 2, r7c5 = 4, r5c6 = 3
str-asc[1]: r6c1<r5c1 ==> r6c1 ≠ 2
nine naked-singles to the end: r6c1 = 1, r2c1 = 3, r6c5 = 2, r1c5 = 1, r1c3 = 2, r5c3 = 1, r2c6 = 4, r5c1 = 2, r2c2 = 1

This puzzle also provides an example of braids in Futoshiki. Indeed, if we activate braids, we get the same resolution path upto state $RS_2$. But, the elimination done by the whip[12] coming immediately after $RS_2$ in the path with whips can now be done by a braid[10], which leads to a B rating of 10 for this puzzle. We did not investigate whether the whip[12] for this elimination is due to some non-confluence phenomenon in this puzzle or if its W rating is effectively 12.

**braid[10]: r6n7{c6 c7} – r6n6{c7 c4} – r6n5{c4 c2} – r5c2{n5 n6} – r4c2{n6 n4} – r7c6{n4 n3} – r3c2{n5 n7} – r7c5{n3 n2} – r7c2{n2 n1} – r7c7{n5 .} ==> r6c6 ≠ 4**

The next steps of the path are the same as in the above resolution path without braids, upto state $RS_3$. We do not repeat them here. After $RS_3$, there is a braid[5] eliminating a candidate that was not eliminated by whips. After it, the two paths diverge, even though they share many patterns (such as pairs and triplets).

braid[5]: r3c5{n6 n5} – c3n6{r3 r2} – c3n5{r2 r5} – r5c2{n5 n6} – c5n6{r5 .} ==> r3c1 ≠ 6
braid[5]: r3n6{c3 c5} – r4n3{c3 c5} – c5n5{r4 r5} – c3n5{r5 r2} – c3n6{r3 .} ==> r3c3 ≠ 3
whip[6]: r2n1{c1 c2} – r2n2{c2 c7} – r3c7{n3 n1} – c6n1{r3 r4} – c4n1{r4 r7} – r7n6{c4 .} ==> r2c1 ≠ 6
braid[6]: c5n6{r5 r3} – r5c2{n6 n5} – c5n5{r5 r4} – r6c2{n5 n4} – r4n4{c5 c3} – r4n3{c5 .} ==> r5c6 ≠ 6

braid[6]: c5n6{r5 r3} – r5c2{n6 n5} – c5n5{r5 r4} – r4c1{n6 n7} – c2n6{r5 r4} – r4c7{n7 .} ==> r5c1 ≠ 6

whip[7]: r2c4{n5 n6} – r7n6{c4 c1} – r7n5{c1 c6} – r6n5{c6 c2} – r5c2{n5 n6} – c5n6{r5 r3} – c3n6{r3 .} ==> r1c4 ≠ 5

str-asc[1]: r1c3<r1c4 ==> r1c3 ≠ 4; str-asc[1]: r1c5<r1c4 ==> r1c5 ≠ 4

naked-triplets-in-a-row r1{c2 c3 c5}{n1 n2 n3} ==> r1c4 ≠ 3

naked-single: r1c4 = 4

naked-pairs-in-a-row r2{c3 c4}{n5 n6} ==> r2c6 ≠ 6, r2c6 ≠ 5

str-asc[2]: r3c7<r2c7<r2c6 ==> r3c7 ≠ 3, r2c7 ≠ 4

naked-pairs-in-a-row r2{c3 c4}{n5 n6} ==> r2c1 ≠ 5

hidden-pairs-in-a-column c6{n6 n7}{r1 r6} ==> r6c6 ≠ 5

naked-pairs-in-a-row r6{c6 c7}{n6 n7} ==> r6c4 ≠ 6

naked-triplets-in-a-column c7{r1 r4 r6}{n6 n5 n7} ==> r5c7 ≠ 5

naked-triplets-in-a-row r1{c2 c3 c5}{n1 n2 n3} ==> r1c1 ≠ 3, r1c1 ≠ 2, r1c1 ≠ 1

whip[3]: r3c5{n5 n6} – c3n6{r3 r2} – r2n5{c3 .} ==> r3c4 ≠ 5

naked-triplets-in-a-row r3{c4 c6 c7}{n2 n3 n1} ==> r3c3 ≠ 2, r3c3 ≠ 1, r3c1 ≠ 3, r3c1 ≠ 2, r3c1 ≠ 1

whip[2]: r5c1{n2 n5} – r3c1{n5 .} ==> r6c1 ≠ 4

singles: r6c2 = 4, r6c4 = 5, r2c4 = 6, r2c3 = 5, r7c1 = 6, r4c1 = 7, r1c1 = 5, r3c1 = 4, r3c3 = 6, r3c5 = 5, r1c7 = 6, r4c7 = 5, r4c2 = 6, r5c2 = 5, r6c7 = 7, r6c6 = 6, r1c6 = 7, r7c6 = 5, r5c5 = 6, r2c6 = 4

str-asc[1]: r6c1<r5c1 ==> r6c1 ≠ 3

singles ==> r6c3 = 3, r4c3 = 4, r4c5 = 3, r1c2 = 3, r5c7 = 4, r7c5 = 4, r5c3 = 1, r1c3 = 2, r1c5 = 1, r6c5 = 2, r6c1 = 1, r2c2 = 1, r7c2 = 2

whip[3]: c6n1{r3 r4} – c4n1{r4 r7} – c4n3{r7 .} ==> r3c ≠ 3

eleven singles to the end: r5c6 = 3, r5c1 = 2, r2c1 = 3, r2c7 = 2, r3c7 = 1, r7c7 = 3, r7c4 = 1, r4c4 = 2, r3c4 = 3, r4c6 = 1, r3c6 = 2

## 14.5. g-labels, g-whips and g-braids in Futoshiki

In n×n Futoshiki, let us define the following sets of Numbers:
$k^+ = \{k, k+1, …, n\}$ for any $k < n$ and $k^- = \{1, 2 …, k\}$ for any $k > 1$.

We can now define the g-labels of Futoshiki. For any $X_{rc}$ CSP variable:

– there is a g-label $< X_{rc}, k^+ >$, or $k^+rc$ for short, provided that cell rc is adjacent in a row [respectively in a column] to at least one cell r'c' such that there is a < [resp. a ∧] inequality sign between rc and r'c'. It is easy to see that label k'r'c' for this adjacent cell is g-linked to g-label $k^+rc$ according to the general definition in chapter 7 if and only if k' ≤ k.

– there is a g-label $< X_{rc}, k^- >$, or $k^-rc$ for short, provided that cell rc is adjacent in a row [respectively in a column] to at least one cell r'c' such that there is a > [resp. a ∨] inequality sign between rc and r'c'. It is easy to see that label k'r'c' for this adjacent cell is g-linked to g-label $k^-rc$ according to the general definition in chapter 7 if and only if k' ≥ k.

Remark: the set of g-labels is fixed in the sense that it does not vary during the resolution process, i.e. it does not depend on the resolution state of a given instance, but, contrary to Sudoku, it is different for each instance. Alternatively, one could move the < condition between cells from the definition of g-labels to the definition of predicate g-linked. This would introduce lots of useless g-labels, but it would not change anything in theory or in practice. [From a programming point of view, it may be easier to have a set of g-labels independent of the instance; but one can also have a universal set of "potential" g-labels and a subset of real g-labels for each instance.]

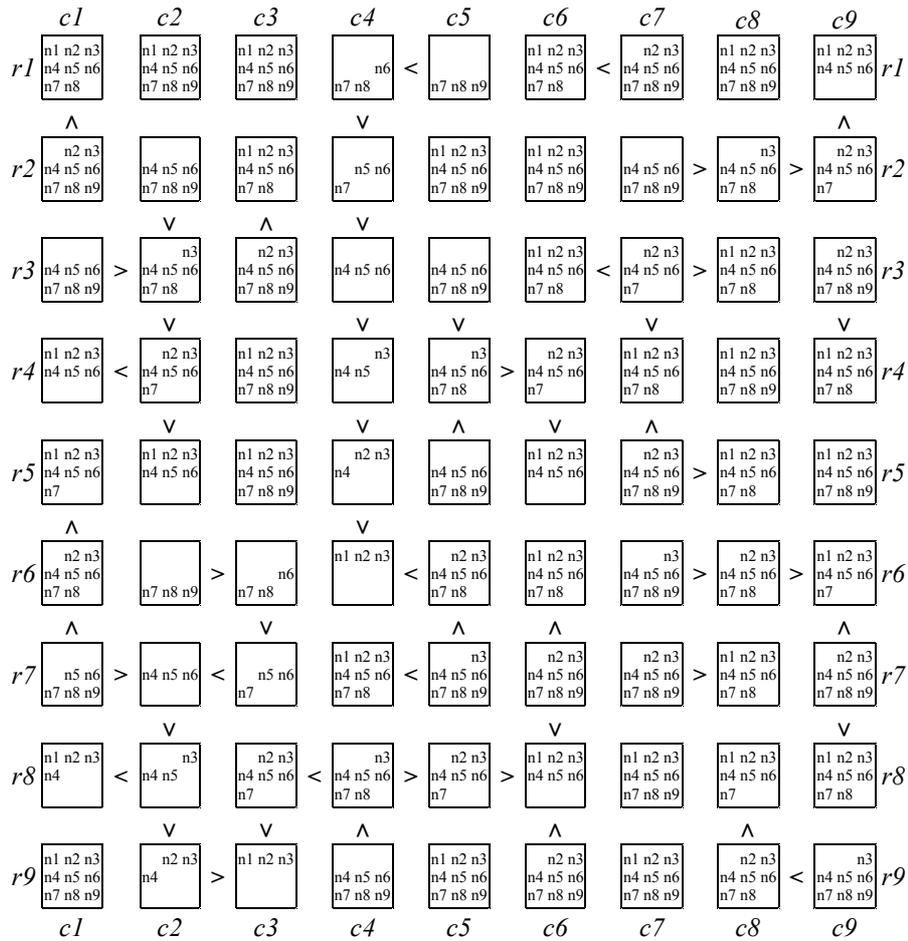

**Figure 14.8.** State $RS_1$ of a 9×9 Futoshiki puzzle (clues of H1117 from atksolutions.com)

Let us now see how g-labels appear in practice in g-whips and g-braids. These are the means by which ascending (or "descending") chains can both be made "contextual" (depending on the target and the previous right-linking candidates) and be included as parts of whip-like or braid-like patterns.

Consider the puzzle defined by the set of inequality signs in Figure 14.8. This hard puzzle cannot be solved by whips, braids or g-whips, even taken together; it requires g-braids. We conjecture that, as in Sudoku, this is an exceptional instance, but there is no available large collection of Futoshiki puzzles that would allow to test this. Figure 14.8 displays the state $RS_1$ reached after all the obvious weak ascending chain eliminations have been done.

After $RS_1$, one has, as is usual, a series of singles, strong ascending chains, hills and valleys, due to the interactions between different chains.

```
*****  FutoRules 1.2 based on CSP-Rules 1.2 , config: gB+S  *****
....................................................................
---<-<--------->>>----<>-<--->---------->-->-<-->>><-<->-<-<>>---->-----<
<---<<--->>>-->>-<--->->>>>>>--<-><-<----->-<><-->><-----------<<->--<>-
0 givens, 729 candidates, 8748 csp-links and 10728 links. Initial density = 1.01
… Starting from resolution state RS₁
singles ==> r9c4 = 9, r8c7 = 9, r6c2 = 9
str-asc[1]: r1c6<r1c7 ==> r1c6 ≠ 8; str-asc[1]: r3c6<r3c7 ==> r3c6 ≠ 8
str-asc[1]: r3c8<r3c7 ==> r3c8 ≠ 8; str-asc[1]: r3c2<r2c2 ==> r3c2 ≠ 8
str-asc[2]: r4c1<r4c2<r3c2 ==> r4c2 ≠ 7; str-asc[2]: r4c1<r4c2<r3c2 ==> r4c1 ≠ 6
str-asc[1]: r5c8<r5c7 ==> r5c8 ≠ 8 ; str-asc[1]: r6c8<r6c7 ==> r6c8 ≠ 8
str-asc[1]: r6c9<r6c8 ==> r6c9 ≠ 7; str-asc[1]: r7c8<r7c7 ==> r7c8 ≠ 8
str-asc[2]: r5c2<r4c2<r3c2 ==> r5c2 ≠ 6; str-asc[1]: r4c7<r3c7 ==> r4c7 ≠ 8
str-asc[1]: r9c8<r9c9 ==> r9c8 ≠ 8; str-asc[1]: r8c8<r9c8 ==> r8c8 ≠ 7
str-asc[1]: r2c8<r2c7 ==> r2c8 ≠ 8; str-asc[2]: r1c9<r2c9<r2c8 ==> r2c9 ≠ 7
str-asc[2]: r1c9<r2c9<r2c8 ==> r1c9 ≠ 6
hill[2]: r6c9<r7c9>r8c9 ==> r7c9 ≠ 2
hill[2]: r6c6<r7c6>r8c6 ==> r7c6 ≠ 2
valley[2]: r3c5>r4c5<r5c5 ==> r4c5 ≠ 8
str-asc[2]: r5c6<r4c6<r4c5 ==> r5c6 ≠ 6, r4c6 ≠ 7
hill[3]: r8c3<r8c4<r8c5>r8c6 ==> r8c4 ≠ 3
hill[2]: r3c6<r3c7>r3c8 ==> r3c7 ≠ 2
str-valley[2]: r3c7>r4c7<r5c7 ==> r4c7 ≠ 7
```

;;; Resolution state $RS_2$, displayed in Figure 14.9. Notice that, until now, there is no "hole" in any of the sets of candidates for a cell.

Following $RS_2$, there appears a series of strong ascending chains, subsets, whips and braids, in the same vein as in the previous example.

```
hidden-pairs-in-a-column c8{n8 n9}{r1 r4} ==> r4c8 and r1c8 ≠ 7, 6, 5, 4, 3, 2, 1
whip[2]: r1c5{n8 n9} – r1c8{n9 .} ==> r1c4 ≠ 8
```

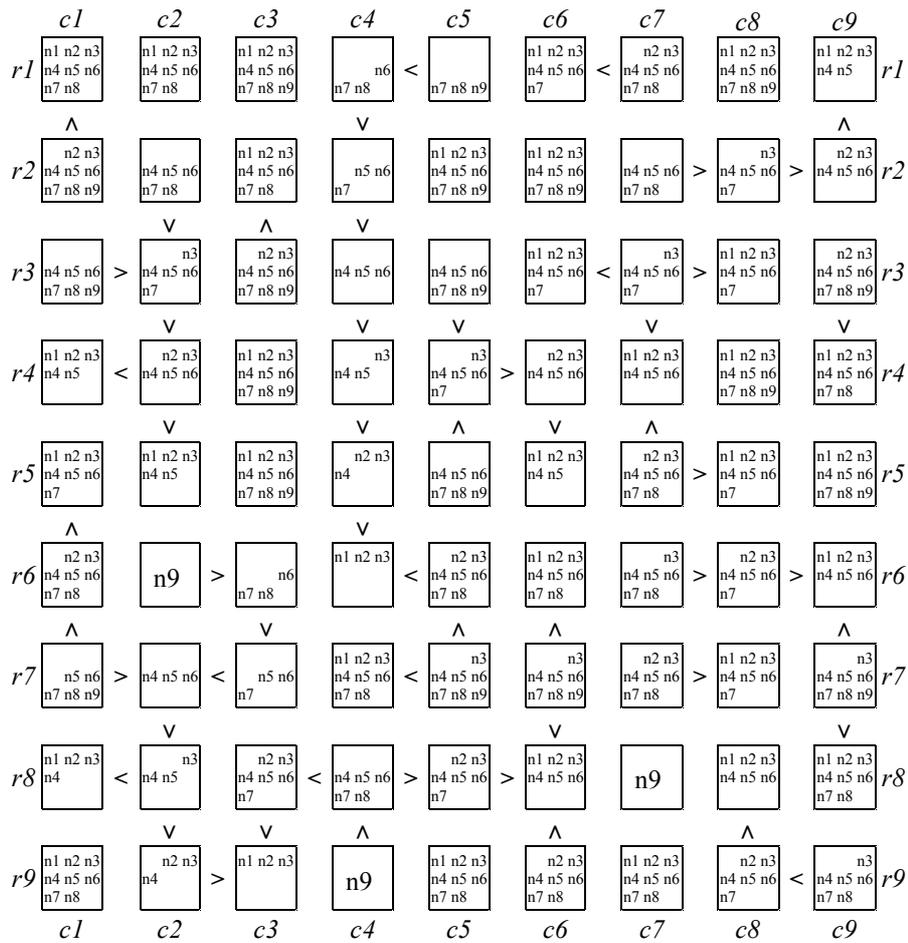

**Figure 14.9.** State $RS_2$ of the 9×9 Futoshiki puzzle of Figure 14.8.

str-asc[5]: r6c4<r5c4<r4c4<r3c4<r2c4<r1c4 ==> r6c4 ≠ 3, r5c4 ≠ 4, r4c4 ≠ 5, r3c4 ≠ 6, r2c4 ≠ 7
whip[2]: r6c4{n2 n1} – r6c9{n1 .} ==> r6c8 ≠ 2
str-asc[1]: r6c8<r6c7 ==> r6c7 ≠ 3
whip[2]: c2n7{r2 r1} – c2n8{r1 .} ==> r2c2 ≠ 6, 5, 4
whip[2]: r2c2{n7 n8} – r2c7{n8 .} ==> r2c8 ≠ 7
str-asc[2]: r1c9<r2c9<r2c8 ==> r2c9 ≠ 6, r1c9 ≠ 5
whip[2]: r2c4{n5 n6} – r2c8{n6 .} ==> r2c9 ≠ 5
str-asc[1]: r1c9<r2c9 ==> r1c9 ≠ 4
whip[2]: r7n2{c8 c4} – r7n1{c4 .} ==> r7c8 ≠ 3, 4, 5, 6, 7
whip[2]: r7c8{n2 n1} – r8c8{n1 .} ==> r9c8 ≠ 2

str-asc[1]: r9c8<r9c9 ==> r9c9 ≠ 3
whip[2]: r8n7{c4 c9} – r8n8{c9 .} ==> r8c4 ≠ 6, 5, 4
whip[3]: c4n8{r8 r7} – r7c5{n8 n9} – r7c9{n9 .} ==> r8c9 ≠ 8
hidden-single-in-a-row ==> r8c4 = 8
whip[3]: r1c7{n7 n8} – r1c5{n8 n9} – r1c8{n9 .} ==> r1c6 ≠ 7
whip[3]: r6c8{n6 n7} – r6c7{n7 n8} – r6c3{n8 .} ==> r6c9 ≠ 6

;;; this is now the first place a "hole" is introduced in a cell:
whip[4]: r5c4{n3 n2} – r5c8{n2 n1} – r7n1{c8 c4} – r6c4{n1 .} ==> r5c7 ≠ 3
whip[4]: r1c4{n6 n7} – r1c7{n7 n8} – r1c5{n8 n9} – r1c8{n9 .} ==> r1c6 ≠ 6
whip[4]: c2n8{r2 r1} – c8n8{r1 r4} – r4n9{c8 c3} – r3c3{n9 .} ==> r2c3 ≠ 8

;;; this is the second place a "hole" is introduced in a cell (exercise: find the next ones)
whip[5]: r3n1{c6 c8} – r7n1{c8 c4} – c4n5{r7 r2} – c4n6{r2 r1} – c4n7{r1 .} ==> r3c6 ≠ 5
braid[6]: c2n8{r1 r2} – r2c1{n8 n9} – r1c8{n8 n9} – r1c5{n8 n7} – c2n7{r1 r3} – r3c1{n9 .} ==> r1c1 ≠ 8
**whip[7]: r3c4{n4 n5} – r2c4{n5 n6} – r1c4{n6 n7} – c2n7{r1 r2} – c2n8{r2 r1} – r1c5{n8 n9} – r1c8{n9 .} ==> r3c1 ≠ 4, r3c2 ≠ 4 ;;; third hole (in r3c2)**
**whip[7]: r4c4{n4 n3} – r4c6{n3 n2} – r5c6{n2 n1} – r3n1{c6 c8} – r7n1{c8 c4} – r6c4{n1 n2} – r5c4{n2 .} ==> r4c5 ≠ 4**
whip[3]: r3c4{n5 n4} – r4c4{n4 n3} – r4c5{n3 .} ==> r3c5 ≠ 5
**whip[7]: r3n1{c6 c8} – r7n1{c8 c4} – c4n7{r7 r1} – c2n7{r1 r2} – c2n8{r2 r1} – r1c5{n8 n9} – r1c8{n9 .} ==> r3c6 ≠ 7**
whip[3]: c6n9{r7 r2} – c6n8{r2 r9} – c6n7{r9 .} ==> r7c6 ≠ 6, 5, 4, 3
**braid[7]: r4n9{c3 c8} – r4n8{c8 c9} – r4n7{c9 c5} – r1c8{n9 n8} – r1c5{n7 n9} – r3c5{n4 n8} – r5c5{n9 .} ==> r4c3 ≠ 6, 5, 4, 3, 2, 1**
**braid[7]: r7n1{c4 c8} – r7n2{c8 c7} – r7n3{c7 c9} – c4n1{r7 r6} – r6c9{n5 n2} – r1c9{n2 n1} – r8c9{n7 .} ==> r7c4 ≠ 7**
hidden-single-in-a-column ==> r1c4 = 7
naked-pairs-in-a-row r1{c5 c8}{n8 n9} ==> r1c7 ≠ 8, r1c3 ≠ 9, r1c3 ≠ 8, r1c2 ≠ 8
singles ==> r2c2 = 8, r3c2 = 7
whip[2]: r7c1{n8 n9} – r3c1{n9 .} ==> r6c1 ≠ 8
str-asc[1]: r5c1<r6c1 ==> r5c1 ≠ 7
whip[2]: r7c5{n8 n9} – r1c5{n9 .} ==> r6c5 ≠ 8

;;; Resolution state RS₃, displayed in Figure 14.10. This is where this example becomes really interesting, because the first g-whips and g-braids appear now (g-labels appear in cells r3c5 and r3c9).

**g-whip[6]: r3c1{n8 n9} – r3c9{n9 n7˙} – r4n8{c9 c8} – r4n9{c8 c3} – r4n7{c3 c5} – r3c5{n4 .} ==> r3c3 ≠ 8**
**g-whip[5]: r3c3{n2 n9} – r3c1{n9 n8} – r3c5{n8 n7˙} – r4n7{c5 c9} – r3c9{n2 .} ==> r2c3 ≠ 7**
**g-braid[6]: r3c1{n9 n8} – r3c9{n8 n7˙} – r4n9{c3 c8} – r4n8{c9 c3} – r4n7{c9 c5} – r3c5{n9 .} ==> r3c3 ≠ 9**
str-asc[1]: r2c3<r3c3 ==> r2c3 ≠ 6
**g-braid[6]: r3c1{n8 n9} – r3c5{n9 n7˙} – r3c9{n9 n7˙} – r4n7{c9 c3} – r4n8{c9 c8} – r4n9{c8 .} ==> r3c7 ≠ 8**

Exercise: check all the z- and t-candidates of these g-whips and g-braids; also check their right-to-left links.

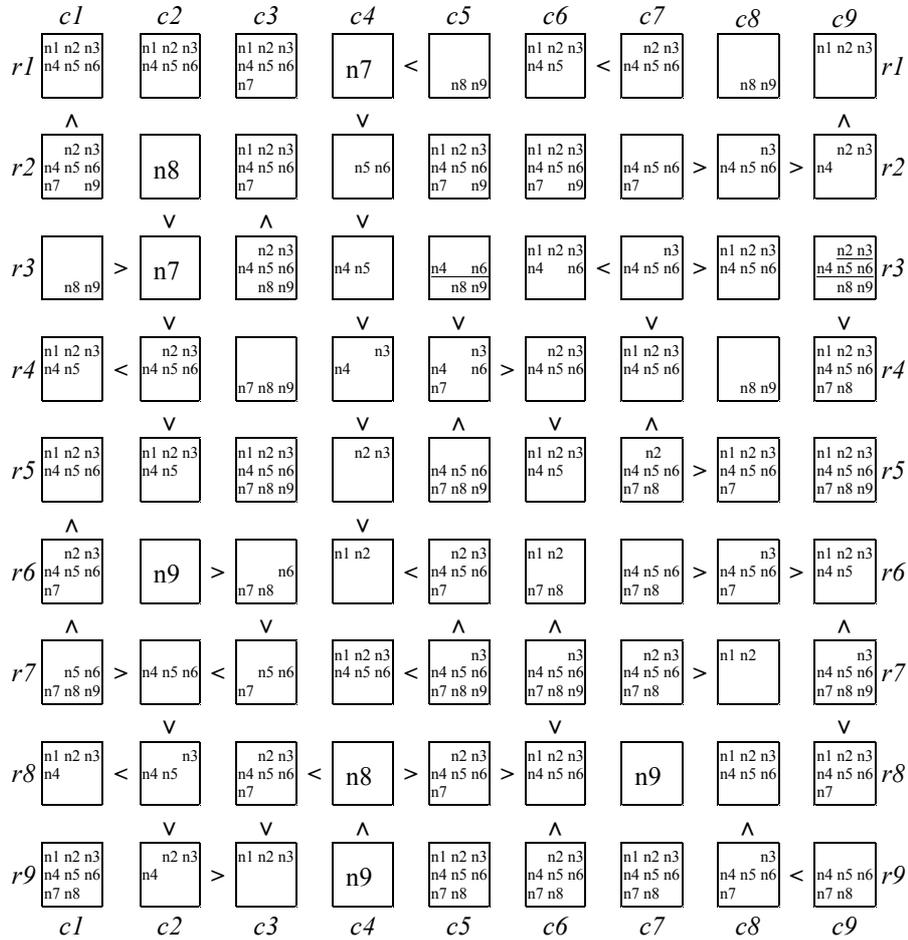

**Figure 14.10.** *State $RS_3$ of the 9×9 Futoshiki puzzle of Figure 14.8. The g-candidates $\underline{n7\ r3c5}$ and $\underline{n7\ r3c9}$ used in the subsequent g-whips and g-braids are underlined*

str-asc[1]: r4c7<r3c7 ==> r4c7 ≠ 6; str-asc[1]: r3c8<r3c7 ==> r3c8 ≠ 6
str-asc[1]: r3c6<r3c7 ==> r3c6 ≠ 6
braid[7]: r7n1{c4 c8} – r7n2{c8 c7} – r7n3{c7 c9} – c4n1{r7 r6} – r6c9{n5 n2} – r1c9{n2 n1} – r8c9{n7 .} ==> r7c4 ≠ 6
hidden-single-in-a-column ==> r2c4 = 6

**whip[7]:** c4n5{r3 r7} – r7n1{c4 c8} – r7n2{c8 c7} – r7n3{c7 c9} – c9n9{r7 r5} – c9n6{r5 r9} – c9n8{r9 .} ==> r3c9 ≠ 5
**braid[7]:** r7n1{c4 c8} – r7n2{c8 c7} – r7n3{c7 c9} – c4n1{r7 r6} – r6c9{n5 n2} – r1c9{n2 n1} – r8c9{n7 .} ==> r7c4 ≠ 5
hidden-single-in-a-column ==> r3c4 = 5
**whip[7]:** c4n4{r4 r7} – r7n1{c4 c8} – r7n2{c8 c7} – r7n3{c7 c9} – r2c9{n3 n2} – r1c9{n3 n1} – r8c9{n1 .} ==> r4c9 ≠ 4
**braid[7]:** r7n1{c4 c8} – r7n2{c8 c7} – r7n3{c7 c9} – c4n1{r7 r6} – r6c9{n5 n2} – r1c9{n2 n1} – r8c9{n7 .} ==> r7c4 ≠ 4
hidden-single-in-a-column ==> r4c4 = 4
**g-braid[7]:** r3c1{n9 n8} – r3c9{n8 n7⁻} – r1n9{c5 c8} – r4n9{c8 c3} – r4n7{c9 c5} – r1c5{n9 n8} – r5c5{n9 .} ==> r3c5 ≠ 9
whip[3]: r3c5{n4 n8} – r5c5{n8 n9} – r1c5{n9 .} ==> r4c5 ≠ 7
str-asc[2]: r5c6<r4c6<r4c5 ==> r5c6 ≠ 5, r4c6 ≠ 6
hidden-triplets-in-a-row r4{n7 n8 n9}{c3 c9 c8} ==> r4c9 ≠ 6, 5, 3, 2, 1
str-asc[1]: r4c9<r3c9 ==> r3c9 ≠ 6, 4, 3, 2
naked-pairs-in-a-row r3{c1 c9}{n8 n9} ==> r3c5 ≠ 8
str-asc[1]: r4c5<r3c5 ==> r4c5 ≠ 6
hidden-single-in-a-row ==> r4c2 = 6
str-asc[2]: r8c1<r8c2<r7c2 ==> r8c2 ≠ 5, r8c1 ≠ 4; str-asc[2]: r9c3<r9c2<r8c2 ==> r9c3 ≠ 3, 4
str-asc[2]: r5c6<r4c6<r4c5 ==> r5c6 ≠ 4, 5; str-asc[1]: r5c6<r4c6 ==> r5c6 ≠ 3
whip[2]: r5c8{n2 n1} – r5c6{n1 .} ==> r5c7 ≠ 2
whip[2]: r8c6{n2 n1} – r5c6{n1 .} ==> r9c6 ≠ 2
whip[2]: r2c3{n2 n1} – r9c3{n1 .} ==> r3c3 ≠ 2
hidden-pairs-in-a-row r3{n1 n2}{c6 c8} ==> r3c8 ≠ 4, 3
naked-pairs-in-a-column c8{r3 r7}{n1 n2} ==> r8c8 ≠ 2, 1
str-asc[1]: r8c8<r9c8 ==> r9c8 ≠ 3; str-asc[1]: r9c8<r9c9 ==> r9c9 ≠ 4
naked-pairs-in-a-column c8{r3 r7}{n1 n2} ==> r5c8 ≠ 2, 1
hidden-pairs-in-a-row r3{n1 n2}{c6 c8} ==> r3c6 ≠ 4, 3
naked-pairs-in-a-column c6{r3 r5}{n1 n2} ==> r8c6 ≠ 2, 1
str-asc[1]: r8c6<r8c5 ==> r8c5 ≠ 3, 2; str-asc[1]: r8c6<r9c6 ==> r9c6 ≠ 3
naked-pairs-in-a-column c6{r3 r5}{n1 n2} ==> r6c6 ≠ 2, 1, r4c6 ≠ 2
singles: r4c6 = 3, r4c5 = 5, r3c5 = 6
str-asc[1]: r2c3<r3c3 ==> r2c3 ≠ 5, 4; str-asc[1]: r8c6<r9c6 ==> r9c6 ≠ 4
str-asc[1]: r8c6<r8c5 ==> r8c5 ≠ 4
naked-single ==> r8c5 = 7
naked-pairs-in-a-column c5{r1 r5}{n8 n9} ==> r9c5 ≠ 8, r7c5 ≠ 9, 8
str-asc[1]: r6c5<r7c5 ==> r6c5 ≠ 4
naked-pairs-in-a-column c5{r1 r5}{n8 n9} ==> r2c5 ≠ 9
naked-pairs-in-a-column c6{r3 r5}{n1 n2} ==> r2c6 ≠ 2, 1, r1c6 ≠ 2, 1
str-asc[1]: r1c6<r1c7 ==> r1c7 ≠ 4, 3, 2
whip[2]: r5c1{n2 n1} – r4c1{n1 .} ==> r6c1 ≠ 2
whip[2]: r1c1{n2 n1} – r4c1{n1 .} ==> r2c1 ≠ 2
whip[2]: r1c6{n5 n4} – r8c6{n4 .} ==> r9c6 ≠ 5
swordfish-in-columns n9{c3 c5 c8}{r4 r5 r1} ==> r5c9 ≠ 9
whip[3]: r9c3{n2 n1} – c7n1{r9 r4} – r4n2{c7 .} ==> r9c1 ≠ 2
whip[3]: r6c5{n3 n2} – r6c9{n2 n1} – r6c4{n1 .} ==> r6c8 ≠ 3

str-asc[1]: r6c8<r6c7 ==> r6c7 ≠ 4
whip[4]: c4n3{r5 r7} – r7n1{c4 c8} – r3n1{c8 c6} – c6n2{r3 .} ==> r5c4 ≠ 2
naked-single ==> r5c4 = 3
str-asc[1]: r5c8<r5c7 ==> r5c7 ≠ 4
naked-pairs-in-a-row r7{c4 c8}{n1 n2} ==> r7c7 ≠ 2
hidden-pairs-in-a-column c7{n1 n2}{r4 r9} ==> r9c7 ≠ 8, 7, 6, 5, 4, 3
naked-pairs-in-a-row r9{c3 c7}{n1 n2} ==> r9c5 ≠ 2, 1
singles: r2c5 = 1, r6c5 = 2, r6c4 = 1, r7c4 = 2, r7c8 = 1, r3c8 = 2, r3c6 = 1, r5c6 = 2
str-asc[1]: r6c9<r7c9 ==> r7c9 ≠ 3
naked-pairs-in-a-row r9{c3 c7}{n1 n2} ==> r9c2 ≠ 2
singles: r9c2 = 3, r8c2 = 4, r7c2 = 5, r5c2 = 1, r1c2 = 2, r9c5 = 4, r7c5 = 3, r3c7 = 3, r3c3 = 4
str-asc[1]: r9c8<r9c9 ==> r9c9 ≠ 5; str-asc[1]: r5c1<r6c1 ==> r6c1 ≠ 4, 3
singles: r6c9 = 3, r1c9 = 1, r8c1 = 1, r4c1 = 2, r4c7 = 1, r9c7 = 2, r9c3 = 1
str-asc[1]: r1c1<r2c1 ==> r2c1 ≠ 3
singles: r1c1 = 3, r1c6 = 4, r6c8 = 4, r5c7 ≠ 5, r6c3 ≠ 6
hidden-pairs-in-a-column c3{n2 n3}{r2 r8} ==> r8c3 ≠ 6, 5
whip[2]: r2c8{n5 n3} – r8c8{n3 .} ==> r9c8 ≠ 5
hidden-single-in-a-row ==> r9c1 = 5
str-asc[1]: r6c1<r7c1 ==> r7c1 ≠ 6; str-asc[1]: r9c8<r9c9 ==> r9c9 ≠ 6
naked-pairs-in-a-column c9{r4 r9}{n7 n8} ==> r7c9 ≠ 8, 7, r5c9 ≠ 8, 7, r3c9 ≠ 8
singles: r3c9 = 9, r3c1 = 8
str-asc[1]: r8c9<r7c9 ==> r8c9 ≠ 6
x-wing-in-rows n6{r8 r9}{c6 c8} ==> r6c6 ≠ 6, r5c8 ≠ 6
whip[2]: r7c9{n6 n4} – c9n5{r8 .} ==> r5c9 ≠ 6
singles and a whip[2] (r2n3{c8 c3} – r2n2{c3 .} ==> r2c9 ≠ 4) to the end

## 14.6. Modelling transitive constraints

Let us now discuss our modelling of Futoshiki and see how it can be generalised to transitive constraints in any CSP.

Definition: a constraint c is *transitive* if, whenever one has linked-by($l_1$, $l_2$, c) and linked-by($l_2$, $l_3$, c) for labels $l_1$, $l_2$ and $l_3$, then one also has linked-by($l_1$, $l_3$, c).

An ascending chain has been given the same rating as a whip[1], independently of its length, but, in the current approach, if it appears as a part of a whip or a braid, it still contributes to the length of the whip by its real length. (This did not appear in the example of section 14.5, because all the g-whips and g-braids included only inequality sub-chains of length one.) This may seem inconsistent. However, there is a very simple way out of this dilemma: instead of modelling the inequality constraints by defining direct contradiction links only between candidates in adjacent cells related by an inequality sign, one can define contradiction links between candidates in any two cells belonging to an ascending chain.

Thus, in an n×n Futoshiki, if $C_0, C_1, ..., C_k$, is an ascending chain and $n_i$ is any Number, $n_iC_0$ would not only be linked by < to $n1C_1, n2C_1, ..., n_iC_1$, but also to $n1C_2, n2C_2, ..., n_{i+1}C_2$, to $n1C_3, n2C_3, ..., n_{i+2}C_3$ and so on.

As a result of using these new direct links, the whole notion of an ascending chain could disappear from the resolution paths (but not the notions of a hill and a valley). What is used here is only the transitivity property of the < constraint; the underlying order does not even have to be total. Obviously, this technique can be applied to any transitive constraint in any CSP and it may seem to be an appropriate general way of dealing with the propagation of such constraints.

However, which of the above two representations one should choose for a transitive constraint, with the consequence of modifying in possibly radical ways the rating of all the chain patterns relying partly on such constraints, is ultimately a modelling decision. In the Futoshiki CSP, the decision should take into account which kinds of readers or players are aimed at: keeping in mind our requirement that each step in the resolution path should be understandable, it would certainly be a very bad idea for beginners; but for advanced players, it may be compulsory in order to avoid the boredom of displaying so many obvious steps.

Notice that, even if these additional links are adopted as primary constraints, it does not entail that a g-whip or g-braid will never have to consider several parts of an ascending chain: it may need to justify t-candidates in its subsequent parts by the explicit presence of an intermediate right-linking candidate.

### 14.7. Hints for further studies on Futoshiki

As an abstract CSP, pure n×n Futoshiki could become an interesting topic for a detailed case study in the same vein as what we have done for Sudoku, with two additional possibilities: 1) as grid size n can take any value (it does not have to be a square $m^2$), it should be easier to analyse how its statistical properties vary with it, in particular what the ratios of minimal instances having various T&E depths are; 2) for any fixed size n, the "geometry" of constraints and therefore their initial density and tightness can be varied with much more freedom. (See section 17.2.2 for a definition and a discussion of these two notions.) We shall leave all this to motivated readers, but let us make a few remarks on the generation of minimal puzzles.

Given a complete n×n Latin Square LS (all the cells filled with values), it can be completed further into a complete "impure" Futoshiki grid by adding the correct inequality sign between any pair of cells adjacent in a row or a column; there are N = 2n(n-1) such signs. Now forgetting all the values in the cells, one gets a complete "pure" Futoshiki grid FP. FP is guaranteed by construction to have a Futoshiki solution, but it is not guaranteed to have a unique one. It would be nice to have some theorem like: "an n×n pure Futoshiki puzzle in which all the inequalities between adjacent cells are specified has a unique solution". But we have not been able to find a simple proof of this. Indeed, we did not try hard, because we can merely discard such an instance if it does not have a unique solution.

In any case, any minimal pure Futoshiki puzzle can be obtained from a complete LatinSquare by applying this process followed by a top-down algorithm similar to that described in chapter 6.

Given such a top-down generator, it would be easy to adapt it as in chapter 6 to make it controlled-bias (but there is currently no available source code). A formula similar to that in chapter 6 can be proven; in n×n Futoshiki, the number of inequality signs in a complete grid is $N = 2n(n-1)$ and N plays the role of the number of cells in Sudoku. If k is the number of remaining clues, one has the "controlled-bias" formula: ***P(k+1) / P(k) = (k+1) / (N-k)***, which allows to compute unbiased statistics from those obtained with collection provided by the controlled-bias generator.

### *14.7.1. Combining the Sudoku and Futoshiki constraints: Sudoshiki*

We think Futoshiki, as a game, will never become as popular as Sudoku:

– an inequality constraint is too weak; it entails too few consequences when a candidate is asserted (contrary to a Sudoku constraint of bn type), unless it is included in a long ascending chain. The maximal length of ascending chains is n-1 (in which case all the cells in the chain are completely solved). If the length is close to this value, the chain will "most of the time" make parts of the puzzle close to trivial. As a result, there cannot be many long chains in a non easy puzzle and having to use repeatedly the inequality constraint (even if written in the extended form introduced in section 14.6) for many short ones is quite tedious.

– g-labels also are too weak; their action is too local (only between cells connected by an inequality). g-labels in Sudoku or N-Queens are more exciting.

– besides ascending chains, hills and valleys, there does not seem to be many possibilities of finding Futoshiki-specific resolution rules.

In this perspective, another game we think worth exploring could be called "***Sudoshiki***" in fake-japanese: restrict grid size in the same way as in Sudoku ($n=m^2$) and combine the constraints of Sudoku and Futoshiki, i.e. add to Futoshiki the block constraints. This should palliate the above-mentioned weakness of the inequality constraints. Sudoshiki has all the g-labels of Sudoku plus those of Futoshiki. Probably, for better complementarity with Sudoku, the most interesting form would be "pure" Sudoshiki, in which clues can only be inequalities. Pure Sudoshiki has the same controlled-bias formula as Futoshiki, which opens the door to statistical analyses.

# 15. Non-binary arithmetic constraints and Kakuro

The logico-arithmetic game of Kakuro (abbreviation of japanese "kasan kurosu", best translated as "cross sums", by analogy with crosswords) is often presented as the numerical, "cross-cultural" analogue of crosswords. Obviously, this can only apply to the structure of the grid, not to the game itself: deprived of any linguistic or cultural aspect similar to wordplay and knowledge about vocabulary, it may look to crosswords addicts as a very poor analogue. Nevertheless, this is irrelevant to our purposes. In the context of the present book, Kakuro is indeed worth some consideration, for the following two main reasons:

– unlike all our previous examples, in its natural formulation, it has *non-binary* arithmetic constraints; in the first page of the Introduction we only alluded to the possibility of reducing such constraints to binary ones by introducing new CSP variables; we shall now show how this general idea can be made to work in practice; notice that this must be done as far as possible in such a way that the additional CSP variables do not have too large domains – i.e. in an application-specific way;

– it has g-labels that are more complex than in our previous examples and that require some theoretical analysis in order to provide them with a simplified representation; above all, these g-labels illustrate the importance of the "saturation" condition introduced in the definition of chapter 7 with respect to efficiency.

There are also more technical reasons:

– in addition to its set of "natural" ones, Kakuro has additional CSP variables that depend on the instance under consideration; (in all our previous examples, the CSP variables were not concerned by such dependency, even if the other constraints were); these variables are intrinsically related to the non-binary constraints;

– the links between the labels for the "natural" CSP variables and for the additional ones may seem to be non-symmetric (they are based on set-theoretic membership), but this will allow to illustrate the difference between the abstract relation "linked" (which must be symmetric) and the semantic relations on which it may be based; (in Futoshiki, the initial "<" relation between two cells was also non-symmetric but it was replaced in a rather obvious way by an equivalent set of symmetric non-equality links between labels for these cells);

– given an instance, g-labels do not depend on its resolution state, in conformance with our general definition (g-labels are structural); but they depend on the instance under consideration;

– it has Naked Subsets, but it does not have systematically corresponding Hidden ones; and it has no Super-Hidden ones if we strictly apply the general definitions of chapter 8, although more complex similar patterns could be defined.

Notice that there is a straightforward translation of a Kakuro puzzle into a linear programming problem, a kind of problem for which there are very efficient (widely and freely available) programs – much more efficient in this case than any general CSP solving program. As mentioned in the general CSP case, if solving efficiency was our only requirement, all of this chapter would be totally irrelevant.

**15.1. Introducing Kakuro**

*15.1.1. Definition of Kakuro*

Kakuro is played on a k×k square grid (with arbitrary k), with two types of cells, called "black" and "white". As in crosswords, black cells are used both as separators and as clue holders (in crosswords, they hold references to the clues rather than the clues themselves, but this is irrelevant). The upper row and the leftmost column contain only black cells.

Figure 15.1 shows the (standard) graphical representation of a Kakuro puzzle that will be used in this book. "Black" cells are in light grey; they are either empty or separated into two parts by a descending diagonal. A horizontal [respectively a vertical] clue, if any, occupies the upper rightmost [resp. lower leftmost] half of the cell. As in all our previous examples, the white cells can be pre-filled with small digits representing their possible values, i.e. with candidate-Numbers. Why we have underlined some of the clues will be explained in section 15.1.3.

In the following definitions, although "block" is often used instead, we adopt the word "sector", in order to avoid confusion with blocks in Sudoku: sectors in Kakuro cannot be used in the same ways as blocks in Sudoku and they do not have the same relationship with g-labels and whips[1].

Definitions: a *horizontal sector* [respectively a *vertical sector*] is a maximal set of contiguous white cells in the same row [resp. column].

A horizontal [resp. vertical] sector is thus always delimited by two black cells or by one black cell and the right end of a row [resp. the bottom of a column]; "contiguous" means that there is no black cell between any two of its white cells.

Definition: The black cell horizontally [resp. vertically] just before the first cell of a sector is called the horizontal [resp. vertical] *controller of this sector;* we also say it is the controller of each cell in the sector.

As far as we know, this notion of a controller has not been made explicit before, but we find it very convenient for many of our forthcoming definitions. Obviously, each white cell belongs to one and only one horizontal [resp. vertical] sector and it has one and only one horizontal [resp. vertical] controller, whether or not this controller contains a clue (as defined below) for it.

|  𝒦  | 11 | 45 | 11 | 7 |  |  |  | 10 | 13 |
|---|---|---|---|---|---|---|---|---|---|
| 12<br>16 |  |  |  |  |  | 4 | 15<br>45 |  |  |
| 35 |  |  |  |  | 11<br>19 |  |  |  |  |
| 11 |  |  |  | 7 | 11<br>34 |  |  | 7 | 15 |
| 38 |  |  |  |  |  | 9 |  |  |  |
|  | 4 | 12<br>13 |  |  |  |  | 18<br>4 |  |  |
| 15 |  |  |  | 24 |  |  |  | 25 | 7 |
| 8 |  |  |  | 24<br>25 |  |  |  |  |  |
|  | 5 | 7<br>13 |  |  |  | 16 | 21<br>7 |  |  |
| 15 |  |  |  |  | 17 |  |  |  |  |
| 11 |  |  |  |  | 29 |  |  |  |  |

**Figure 15.1.** *An 11×11 Kakuro puzzle (clues of #M72601, from atksolutions.com)*

In a Kakuro puzzle, all the white cells are initially empty and the goal is to find for each of them a value in the set of digits {1, ..., 9} (independent of grid size) such that these values satisfy the following two types of constraints:

– the constraints of "mutual exclusion" in each sector: in any (horizontal or vertical) sector, the same digit may not appear twice; but, contrary to all our previous examples, there is no such constraint globally in each row or column (in any case, it would be impossible to satisfy it for grids of size larger than nine – unless the set of digits is extended beyond 9);

– the sum constraints defined by the clues in the black cells, as follows.

A black cell C may contain zero, one or two types of clues:

– a *horizontal clue* S is an integer in the uppermost right corner of C meaning that the sum of the digits in the horizontal sector it controls must be equal to S;

– a *vertical clue* S is an integer in the lowermost left corner of C meaning that the sum of the digits in the vertical sector it controls must be equal to S.

As a result of the above definitions, the maximal size of any sector is nine. If there is a horizontal [resp. vertical] clue S in a black cell, we also say that this clue controls the horizontal [resp. vertical] sector controlled by the black cell. If p is the size of the sector, we call (S, p) the parameters of the sector or of the clue. We say that a digit is (S, p)-compatible if there exists at least one combination of p digits with sum S.

### 15.1.2. Miscellaneous remarks on Kakuro

As in all our previous examples, a well-formed Kakuro puzzle is supposed to have one and only one solution. This is "guaranteed" by most of the websites proposing Kakuro puzzles and all the examples we shall deal with do satisfy it (but no minimality condition is ever evoked).

Even if the global grid is square, the "real" one, i.e. the set of white cells, can have any shape one may want: it suffices to put enough black cells in the rightmost columns and/or in the lower rows. In particular, rectangular grids will often appear. The grid does not have to be simply connected (i.e. it may have an ulimited number of holes, made of isolated or grouped black cells). However, it must be *connected* (if black cells are considered as the ocean, it may have neither separate "continents" nor "islands" in the holes), otherwise it would be equivalent to several independent grids (see section 15.7 for a more formal definition).

A horizontal or vertical clue cannot be greater than 45. In case it is 45 (which implies that it controls 9 cells), considering the general constraint that all the candidate-Numbers in the sector must be different, it does not convey any content beyond the boundary information. Some websites adopt the convention of discarding it, but a few things will be easier to formulate if, on the contrary, we make it compulsory.

There is often a convention that no clue bears on only one white cell; we adopt it for definiteness, but this does not have much impact on our forthcoming analyses. In any case, this situation would fall under those examined in section 15.7.

There is also sometimes an implicit convention that every sector has an explicit clue; the reason is that sectors with no clue allow many more possibilities for their cells (with no sum restriction in a sector of size p, any of the 9!(9-p)!/p!

combinations of p different values is allowed), which makes the puzzle much more difficult to solve. As it has no impact on our theoretical analyses, we do not adopt it.

There is a total of 120 different legitimate clues in a puzzle, i.e. of (S, p) compatible pairs. As these are easily computable or available on many websites, we do not list them here. Following the vocabulary used on some websites, we shall also speak of an (S, p) clue as an "S-in-p".

### 15.1.3. "Magic" sectors

*Combinations* of digits that can appear in a sector will play a major role in the sequel. For certain clues, depending on the (S, p) pair of the sector they control, there is only one possible combination of Numbers fulfilling the sum constraint (notwithstanding all the possible permutations of these Numbers within the controlled cells). There are thirty four such "magic" cases (that can easily be computed or found on several Kakuro websites). By abuse of language, when the context is clear, we shall speak of "magic" sectors and "magic" sums, but what's "magic" is only the (S, p) pair.

| Sector size | Sum | Combination | Sector size | Sum | Combination |
|---|---|---|---|---|---|
| 2 | 3 | 12 | 6 | 22 | 123457 |
| 2 | 4 | 13 | 6 | 38 | 356789 |
| 2 | 16 | 79 | 6 | 39 | 456789 |
| 2 | 17 | 89 | 7 | 28 | 1234567 |
| 3 | 6 | 123 | 7 | 29 | 1234568 |
| 3 | 7 | 124 | 7 | 41 | 2456789 |
| 3 | 23 | 689 | 7 | 42 | 3456789 |
| 3 | 24 | 78 9 | 8 | 36 | 12345678 |
| 4 | 10 | 1234 | 8 | 37 | 12345679 |
| 4 | 11 | 1235 | 8 | 38 | 12345689 |
| 4 | 29 | 5789 | 8 | 39 | 12345789 |
| 4 | 30 | 6789 | 8 | 40 | 12346789 |
| 5 | 15 | 12345 | 8 | 41 | 12356789 |
| 5 | 16 | 12346 | 8 | 42 | 12456789 |
| 5 | 34 | 46789 | 8 | 43 | 13456789 |
| 5 | 35 | 56789 | 8 | 44 | 23456789 |
| 6 | 21 | 123456 | 9 | 45 | 123456789 |

***Table 15.1.*** *The 34 "magic" combinations*

As these cases are the main starting points for the solution of many puzzles and they will play a particular role in our modelling choices, they will always be underlined in our graphical representations (as in Figure 15.1). Table 15.1 gives the full list of these thirty four "magic" cases, ordered by sector size. Notice that the 34 corresponding combinations constitute only a very small part of all the possible combinations (502), for sector-size p varying from 2 to 9 and sum S from 3 to 45.

Of course, the only "magic" here is no more than pure arithmetic – typical examples of propositions that Kant would have classified as *synthetic a priori*. In modern philosophy, especially after the development of formal logic since the beginning of the twentieth century, there has been a strong resistance to the idea that some mathematical propositions could be *synthetic a priori*. This is often based on both an implicit overly formalistic ideology and a misunderstanding of the meaning of these words in Kant's view.

For Kant, *synthetic* means that these propositions increase knowledge (with respect to the original concepts); *a priori* means that they are *anterior* to experience (i.e. they are logically anterior to observation or experimentation, they can be reached without them). From this logical anteriority, formalists argue that these propositions are *analytic*, because they can be (formally) deduced from the axioms. But what "analytic" means for Kant cannot be expressed in such anachronically formalistic terms as "provable from the definitions and axioms by a more or less complex proof". It means included in the very idea of the concepts involved, reachable by mentally analysing this very idea. [This is not to suggest that the "very idea" of these concepts should be construed as some eternal essence (a meaningless notion in our view); perhaps the best way of approximating it in modern terms is to say that it should be intuitively conceivable to choose it as an axiom in some reasonable axiom system.]

Depending on how arithmetic and addition are conceived, it may be debated whether commutativity of addition should be considered as analytic or synthetic. But why, for some values of S and p, there is only one possible combination of p different digits with sum S, and why there are exactly 34 such cases, this is undoubtedly not included in the "very idea" of addition, even though this can easily be proven from (any formalisation of) the definition of addition.

Similarly, the very idea of addition, in and of itself, does not include any reason why there are exactly one hundred and twenty (S, p) pairs such that there is at least one combination of p digits with sum S. Or why, for any value of q with $2 \leq q \leq 12$, there exist (S, p) pairs allowing exactly q different combinations of p different digits with sum S, *except for q = 10*, in which case there is no such (S, p) pair (see subsection 15.1.4). Generally speaking, theorems (or such exotic properties as above or as those that will appear when we study g-labels) have a cost (in terms of proof complexity) and the most interesting ones are generally not "mentally included" in the basic concepts and axioms of the theories.

*15.1.4. Non-magic sectors*

For our forthcoming modelling of Kakuro as a CSP, it must be noted that, beyond the above "magic" cases admitting only one combination, the number q of different combinations of p different digits having sum S remains bounded by 12 for any consistent (S, p) pair. The largest numbers of combinations are obtained when p = 5, 4 and 6; and the number q is 12 in only two cases:

(20, 4) → {1289 1379 1469 1478 1568 2369 2378 2459 2468 2567 3458 3467}
(25, 5)→ {12589 12679 13489 13579 13678 14569 14578 23479 23569 23578 24568 34567}

For $1 \leq q \leq 12$, Table 15.2 gives the number N(q) of (S, p) pairs allowing q different combinations of p different digits with sum S. The total $\Sigma_{(q=1,...,12)}$ q×N(q) is equal to 502, the number of possible digit combinations, for any S and p. This may give the impression that one has to deal with only small numbers of possibilities for each value of (S, p), but this would be forgetting that any (S, p) pair can appear in a puzzle, so that: 1) there are globally 502 possible combinations one may have to consider and 2) some of these give rise to huge numbers of permutations.

| q | 1 | 2 | 3 | 4 | 5 | 6 | 7 | 8 | 9 | 10 | 11 | 12 |
|---|---|---|---|---|---|---|---|---|---|----|----|----|
| Number N(q) of (S, p) pairs having q digit combinations | 34 | 16 | 16 | 10 | 8 | 4 | 8 | 10 | 4 | 0 | 8 | 2 |

**Table 15.2.** *Number N(q) of (S, p) pairs that are instantiated by q digit combinations.*

As for the sectors that have no clue, the relevant data appearing in Table 15.3 may seem much less enticing: in a sector of size n, there are C(9, n) possible combinations, i.e. upto 126 in the worst cases (which do not occur for the largest sector sizes). This is why most puzzles proposed as games have no sector deprived of a clue. However, possible combinations for such no-clue sectors convey no information beyond that defined by mutual exclusion within the boundaries and it will not be necessary to take them explicitly into consideration in our theoretical analyses.

| Sector size p | 1 | 2 | 3 | 4 | 5 | 6 | 7 | 8 | 9 |
|---|---|---|---|---|---|---|---|---|---|
| Number of combinations C(9, p) = 9! / p! / (9-p)! | 9 | 36 | 84 | 126 | 126 | 84 | 36 | 9 | 1 |

**Table 15.3.** *Number of combinations with non-predefined sum, as a function of sector size.*

*15.1.5. Pseudo-magic cases*

There are 34 magic cases and there are also twenty-eight cases, given in Table 15.4, of non-magic (S, p) pairs that have digits (up to five) common to all their combinations. This happens only when there are no more than five combinations.

These cases will also play a particular role in our modelling of Kakuro as a CSP, because digits common to all the (S, p)-compatible combinations are as good for many purposes as all the (S, p)-compatible digits in the "magic" cases. As far as we know, these "pseudo-magic" cases have never before been explicitly considered as forming a family worth of interest (although each may have been used implicitly in resolution, in the form: "this digit must be somewhere in this sector, therefore …").

| (p, S) | pseudo-magic digits | combinations |
|---|---|---|
| (3, 8) | 1 | 125  134 |
| (4, 12) | 1, 2 | 1236  1245 |
| (4, 13) | 1 | 1237  1246  1345 |
| (5, 17) | 1, 2, 3 | 12347  12356 |
| (5, 18) | 1, 2 | 12348  12357  12456 |
| (5, 19) | 1 | 12349  12358  12367  12457  13456 |
| (5, 31) | 9 | 16789  25789  34789  35689  45679 |
| (5, 32) | 9 | 26789  35789  45689 |
| (5, 33) | 9 | 36789  45789 |
| (6, 23) | 1, 2, 3, 4 | 123458  123467 |
| (6, 24) | 1, 2, 3 | 123459  123468  123567 |
| (6, 25) | 1, 2 | 123469  123478  123568  124567 |
| (6, 26) | 1 | 123479  123569  123578  124568  134567 |
| (6, 34) | 9 | 136789  145789  235789  245689  345679 |
| (6, 35) | 8, 9 | 146789  236789  245789  345689 |
| (6, 36) | 7, 8, 9 | 156789  246789  345789 |
| (6, 37) | 7, 8, 9 | 256789  346789 |
| (7, 30) | 1, 2, 3, 4, 5 | 1234569  1234578 |
| (7, 31) | 1, 2, 3, 4, 7 | 1234579  1234678 |
| (7, 32) | 1, 2, 3 | 1234589  1234679  1235678 |
| (7, 33) | 1, 2, 6 | 1234689  1235679  1245678 |
| (7, 34) | 1 | 1234789  1235689  1245679  1345678 |
| (7, 35) | 5 | 1235789  1245689  1345679  2345678 |
| (7, 36) | 9 | 1236789  1245789  1345689  2345679 |
| (7, 37) | 4, 8, 9 | 1246789  1345789  2345689 |
| (7, 38) | 7, 8, 9 | 1256789  1346789  2345789 |
| (7, 39) | 3, 6, 7, 8, 9 | 1356789  2346789 |
| (7, 40) | 5, 6, 7, 8, 9 | 1456789  2356789 |

**Table 15.4.** *The 28 "pseudo-magic" (sector size, sum) pairs with digits common to all their combinations*

It should now be noted that, if the analogy with crosswords had to be pushed further than mere grid structure, knowledge about words would have to be compared with knowledge about magic sectors (Table 15.1) and pseudo-magic cases (Table 15.4) [and also about g-combinations – see section 15.5]. This appears to us as the main limitation of Kakuro as a game: from a player's point of view, we can easily imagine that there is some pleasure in crosswords (because words are the stuff our lives are made of and good crosswords propose unexpected definitions of them); we can also imagine that there is some pleasure in finding complex patterns in a Sudoku grid, because such patterns rely on some fixed visible grid structure; but it is hard to imagine that there could be any pleasure in memorising so many combinations of digits or in spending time in consulting tables containing them (however, this may be due to our lack of imagination in this domain). In any case, this digression does not lessen the theoretical interest of Kakuro in itself or for the purposes of this book.

## 15.2. Modelling Kakuro as a CSP

In this section, we show how Kakuro can be modelled as a CSP according to the general principles of Part I, in spite of having non-binary constraints (they can indeed be very far from binary, as some of them can bear on up to nine variables).

### 15.2.1. Sorts and CSP variables of the Kakuro CSP

For Kakuro on a k×k grid, we adopt the same Row and Column sorts as for LatinSquare, but with domains adjusted to grid size, i.e. with respective sets of constant symbols $\{r1, …, rk\}$ and $\{c1, …, ck\}$. We also adopt a sort Number independent of grid size, with set of constant symbols $\{n1, n2, …, n9\}$. Depending on how we initialise the CSP variables, we can also introduce a sort $Compat_{S_H,p_H,S_V,p_V}$ instead of Number for each legitimate $(S_H, p_H)$ and $(S_V, p_V)$ pairs, with set of constant symbols $Compat(S_H, p_H, S_V, p_V)$, the set of $(S_H, p_H)$-compatible *and* $(S_V, p_V)$-compatible digits. In the sequel, we shall adopt this latter possibility.

We define a sort Combination, with set of constant symbols the set Comb of all the symbols $n_1n_2…$ made by glueing p (for any p with $1 \leq p \leq 9$) different digits in increasing order; see the examples for $(S, p) = (20, 4)$ and $(S, p) = (25, 5)$ in section 15.1.4.

In relation with sectors having no clue, we introduce a sub-sort $Combination_p$ of Combination, with set of constant symbols the set ***Comb(p)*** of all the symbols $n_1n_2…n_p$ made by glueing together exactly p different digits in increasing order (with no sum constraint).

Last but nor least, for each digit p and sum S, we also define a sub-sort $Combination_{S,p}$ of Combination (and of $Combination_p$), with set of constant symbols the set ***Comb(S, p)*** of all the symbols $n_1 n_2 \ldots n_p$ made by glueing together exactly p different (S, p)-compatible digits in increasing order and with sum S. The elements in Comb(S, p) represent all the possibilities for a sector controlled by parameters (S, p), notwithstanding the order of the digits.

Remarks:

– adopting an increasing order for the constant symbols of sorts Combination, Combination$_p$ and Combination$_{S,p}$ is a mere notational choice, unrelated with any consideration about permutations;

– we shall make an abuse of language by almost systematically identifying an abstract symbol in Comb, Comb(p) or Comb(S, p) with the subset of digits from which it is built.

*15.2.1.1. The "natural" Xrc CSP variables*

The "natural" CSP variables of k×k Kakuro, corresponding to the original problem formulation, are all the Xr°c° such that r° is in {r1, ..., rk}, c° is in {c1, ..., ck} *and* cell (r°, c°) is white. They are thus different for different patterns of black cells. The domain of variable Xrc is Compat($S_H$, $p_H$, $S_V$, $p_V$), where ($S_H$, $p_H$) and ($S_V$, $p_V$) are its horizontal and vertical parameters. It should be noticed that this entails in practice that most of the obvious initial domain restrictions of the white cells (which may be the main stuff for beginners to deal with) are supposed to be done before the start of the resolution process proper. Our main purpose in this choice is to avoid endless boring eliminations at the start.

*15.2.1.2. The Xrn and Xcn CSP variables*

Contrary to the previous Sudoku and Futoshiki examples, there are in general no Xr°n° or Xc°n° CSP variables associated with all the (Row, Number) or (Column, Number) pairs, even limited to sectors, because there is in general no constraint relative to the presence of each Number in each Row or Column.

However, there is a major exception and it is related to the "magic" sectors. Given a horizontal "magic" sector of size p, controlled by black cell (r°, c°) with horizontal clue S and associated with the unique combination C = {$n_1$, ..., $n_p$} of digits defined in Table 15.1, for each n° in C, and only for these Numbers, we introduce a CSP variable Hr°c°n°, with domain the set of columns in the magic sector. We introduce similar Vr°c°n° CSP variables for the "magic" vertical sectors. The reason why we have kept the special "45-in-9" magic case (contrary to usual conventions and although it conveys no information) should now be clear: we did not want to exclude it from generating such Hr°c°n° and/or Vr°c°n° variables. Notice that these variables cannot be called Hr°n° or Vc°n° as they would be in Sudoku or LatinSquare, because there may be several magic sectors in the same row (or column) and we need a means of distinguishing the associated variables.

There is also a secondary exception, related to the twenty-eight "pseudo-magic" cases given in Table 15.4. Given a horizontal "pseudo-magic" sector controlled by black cell (r°, c°) with parameters (S, p), for each n° in the set of digits common to all the (S, p)-compatible combinations, as defined in Table 15.4, and only for these Numbers, we introduce a CSP variable Xr°c°n°, with domain the set of columns in the magic sector. Of course, similar Xr°c°n° CSP variables are introduced for the vertical "pseudo-magic" sectors.

### 15.2.1.3. The Hrc and Vrc CSP variables

We must also introduce additional CSP variables that will allow to take the sum constraints into account. For each horizontal [resp. vertical] clue, we define a CSP variable representing the global content of the cells in the sector it controls, this content being considered as a set or a combination of different digits. We shall then also say that this new CSP variable controls the sector. More precisely:

– for each black cell (r°, c°) containing a horizontal clue S, if the sector it controls in row r° has length p, then we define CSP variable Hr°c°, with domain the set of combinations of p different Numbers with sum S, i.e. Combination$_{S,p}$;

– similarly, for each black cell (r°, c°) containing a vertical clue S, if the sector it controls in column c° has length p, then we define CSP variable Vr°c°, with domain Combination$_{S,p}$.

In less formal terms, each of these new CSP variables allows to manage the possible *combinations* of Numbers in the sector it controls. A candidate for such a CSP variable is a possible combination of digits for the sector it controls; along the resolution process, the number of these global possibilities for the sector will decrease in a way consistent with the possibilities remaining in the white cells. The horizontal [respectively vertical] magic sectors correspond to Hrv [resp. Vrc] CSP variables with domains having only one value.

### 15.2.1.4. Miscellaneous remarks

The Hrc and Vrc CSP variables are not "natural" in the sense that they would directly correspond to the original problem formulation: "find a value for each white cell such that …"; only the Xrc are "natural" in this sense. But they are natural in the larger sense that they are a mere formalisation of the classical idea that one must keep track of the combinations still possible for each sector. In this extended sense, the Xrn and Xcn variables are much less natural than the Hrc and Vrc.

The two new sets of Hrc and Vrc CSP variables are mutually disjoint, and disjoint from the sets of Xrc, Xrn and Xcn, thanks to the presence of prefix H or V in their names.

One can consider that there are 5 CSP-Variable-Types: rc, rn, cn, hrc, vrc.

Each of the possible values for each of the Hrc or Vrc CSP variables is a set (a combination); as each of these sets represents the whole set of values for cells in the sector it controls, these variables are highly redundant with the "natural" ones, as is usual in our approach.

After Table 15.2, the cardinalities of the domains of the new Hrc or Vrc CSP variables are not much larger than those of the "natural" Xrc ones (the maximum is 12); in most of the cases, they are even smaller. We have therefore avoided the complexity pitfall that generally goes with the replacement of non-binary constraints by binary ones.

No CSP variable is introduced for sectors defined only by their boundaries, with no sum constraint. This is first of all a natural modelling choice: such variables would not carry any useful information. But, considering the data in Table 15.3, this also has the fortunate consequence that we need not introduce any CSP variable with a domain much larger (up to 126) than those of the "natural" ones.

There is another implicit modelling choice: instead of choosing for domains of the new CSP variables the *sets* (i.e. combinations) of n different Numbers with sum S, one could have chosen the *sequences* (i.e. permutations) of n different Numbers. The cardinalities would have been much larger, upto 9! = 362,880. This would still have been manageable for a computer, although probably not for a human solver, but, as shown by theorem 15.1 below, this would have brought nothing more with respect to the expression of sum constraints. This choice is consistent with (and was inspired by) the way the usual resolution techniques are described on various websites, where it is mentioned that one must track *combinations* of digits.

These new CSP variables (together with the definition of their domains) depend in an essential way on the set of clues.

### 15.2.2. Labels of the Kakuro CSP

For a white cell $(r°, c°)$, labels for CSP variable $Xr°c°$ are defined as all the $(n°, r°, c°)$ triplets (also notated $n°r°c°$) with $r°$ in Row, $c°$ in Column and $n°$ in Compat($S°_H$, $p°_H$, $S°_V$, $p°_V$), similarly to the Sudoku or LatinSquare cases. But contrary to these CSPs, label $n°r°c°$ is generally the equivalence class of only one pre-label: $<Xr°c°, n°>$ because there are no $Xr°n°$ or $Xc°n°$ CSP variables. The main exception is for the "magic" sectors: if $(r°, c°)$ belongs to a horizontal magic sector controlled by cell $(r°, c'°)$ and/or a vertical magic sector controlled by cell $(r'°, c°)$, then there are pre-labels $<Hn°r°c'°, c°>$ and/or $<Vn°r'°c°, r°>$ equivalent to $<Xr°c°, n°>$. The secondary exception is for a "pseudo-magic" (S, p) pair and a digit common to all its combinations (and only such a digit).

Labels for CSP variable $Hr°c°$ controlling a horizontal sector with parameters (S, p) are defined as all the symbols $H[n_1 \ldots n_p]r°c°$, where $\{n_1, \ldots, n_p\}$ is a possible value for Hr°c°, i.e. a combination of p digits with sum S; for the purpose of having a well defined naming scheme for these labels, we always suppose that they are written with $n_1 < \ldots < n_p$. Informally, label $H[n_1, \ldots, n_p]r°c°$ represents the fact that $\{n_1, \ldots, n_p\}$ is exactly the set of values appearing somewhere (in any order) in the horizontal sector controlled by (r°, c°). Each of these labels is the equivalence class of only one pre-label $<Hr°c°, \{n_1, \ldots, n_p\}>$; there is therefore a one-to-one correspondence between labels for Hr°c° and elements of Comb(S, p).

Labels for CSP variable Vr°c° are defined similarly, using prefix V instead of H.

### 15.2.3. Constraints and Constraint-Types of the Kakuro CSP

We shall use seven Constraint-Types: rc, hrc, vrc, rn, cn, hS, vS.

Constraints of type rc, hrc and vrc are associated with the above-defined three classes of CSP variables (Xrc, Hrc and Vrc) and they mean as usual that two different values for the same CSP variable are incompatible. This distinction between three different types is not essential from the point of view of logic, but it may be useful if one wants to distinguish different types of Singles and assign the associated rules different priorities: e.g. Singles for Xrc variables may be considered as "easier" to spot on the grid than Singles for Hrc or Vrc variables.

Constraints of type rn [respectively cn] express that two white cells in the same horizontal [resp. vertical] sector cannot have the same value. As previously noticed, except in the "magic" or "pseudo-magic" cases, these constraints cannot generally be associated with "global" Xrn [resp. Xcn] CSP variables in the row [resp. column], not even with "local" Xrc'n [resp. Xr'cn] CSP variables restricted to the proper sectors. This will have concrete consequences, e.g. when we evoke Hidden or Super-Hidden Subset rules (see section 15.3.2). Notice that we use the same rn and cn constraint types, whether there is an underlying CSP variable or not, as this can introduce no confusion.

Constraints of type hS [respectively vS] link the labels for the "natural" CSP variables with the labels for the additional "horizontal" ones. They mean that the information conveyed by the two labels is inconsistent, i.e. that the value of the label for the white cell is not one of the values allowed by the combination in the label for its horizontal controller cell. More precisely, we introduce a (symmetric) constraint of type hS between label $H[n_1 \ldots n_p]rc$ for CSP variable Hrc and label n'r'c' for CSP variable Hr'c' whenever:

– r' = r,

– c' is in the horizontal sector controlled by (r, c),

– and $n' \notin \{n_1, \ldots, n_p\}$.

This constraint is expressed by predicate:

linked-by(H[$n_1…n_p$]rc, r'c', hS) ∧ linked-by(r'c', H[$n_1…n_p$]rc, hS).

Constraints of type vS linking the labels for the "natural" CSP variables with the labels for the additional "vertical" ones are defined similarly.

These different types of constraints may be used to assign a preference to ECP based on rn and cn constraints, with respect to ECP based on hS or vS constraints.

### 15.2.4. Givens and domain assignments in the Kakuro CSP

In the first section, we said that the clues of a Kakuro puzzle are horizontal or vertical sums in the black cells. In our formal CSP re-formulation, these correspond to assigning values for the additional CSP variables only in the case of "magic" sectors; in any other case, they only correspond to an initial restriction on the set of all the possible combinations of p digits (i.e. candidates) for these new variables. This does not change our model of resolution; the initial resolution state must only be defined in a less direct way than in the previous cases, by assigning each of the Hrc and Vrc variables a domain Combination$_{S,p}$ consistent with its (S, p) pair. The way domains are assigned to the Xrc variables has been discussed previously.

### 15.2.5. Re-formulation of Kakuro as a CSP

The motivation for the above detailed definitions lies in the following theorem:

***Theorem 15.1: a solution of a Kakuro puzzle is equivalent to a solution of the CSP defined by the natural Xrc and additional Xrn, Xcn, Hrc and Vrc CSP variables (together with their allowed domains of values) with all the above-defined constraints, namely (all the "strong" constraints and):***

*– in each sector, the constraints of mutual exclusion (along rn and cn links),*

*– for the additional CSP variables the constraints associated with their above defined contradiction links (hS and vS links) with the natural ones.*

Proof: the "natural ⇒ CSP" part is obvious. Let us prove the converse.

First, in a sector with no sum constraint, the only constraints for cells in this sector specified by the original puzzle data are the constraints of mutual exclusion and nothing needs be added.

Consider now any fixed clue in the original puzzle, with sum S for a sector of size n. Consider the associated additional CSP variable, say X. The value of X, which is a combination C of n different digits, means that these n digits have the required sum S. The fact that this CSP variable X satisfies the contradiction links with all the natural variables in the sector it controls means that each of the values for these variables is among those in C. The fact that the natural CSP variables in the sector satisfy the mutual exclusion constraints in the sector means that they are all different. As a result, they constitute a realisation of combination C, their sum is S and the original clue is satisfied. The same proof works for any clue. q.e.d.

The theorem says that, *as far as the problem formulation and solution are concerned*, the original non-binary arithmetic constraints can be completely replaced by the above-defined Hrc and Vrc variables together with their contradiction links with the Xrc variables (and the links between the latter). Although a little more complex, this is similar to the Futoshiki case, in which each inequality was replaced by an equivalent set of links. The intuitive meaning is that, *in theory, one needs do no more arithmetic, but only find a solution with consistent values for combinations and rc-cells*. However, there are two limitations to the practical interpretation of the theorem:

– it does not mean that a little more arithmetic may not make the resolution simpler in practice; we shall consider some aspects of this question later (see section 15.7);

– it does not mean that the hS and vS links of the new formulation are sufficient to express all the mathematical content of the sum constraints (see in section 15.3.2 how we shall palliate this limitation by adding six coupling rules).

Remarks:

– indirectly, the theorem also says that it would be useless to introduce CSP variables whose domains would be sets of permutations instead of sets of combinations; the next sections will show how to exploit concretely the interplay between the natural Xrc variables on the one hand and the additional Hrc and Vrc variables on the other hand;

– as far as can be seen from the existing websites, the idea of considering and tracking combinations of numbers for each sector is standard in Kakuro; but we have been able to find neither the origin of this idea, nor any formalisation of it, nor any mention that this could completely replace (in theory) the sum information;

– several Kakuro websites propose only instances whose non-magic sectors are restricted to $N(q) = 2$; this corresponds to having only bivalue additional CSP variables, which makes the puzzles much easier; as there are only sixteen (S, p) pairs with $N(q) = 2$ (see Table 15.2), this is a strong limitation on possible puzzles.

## 15.3. Elementary Kakuro resolution rules and theories

### 15.3.1. The Basic Kakuro Resolution Theory

There is nothing special to say about the Basic Kakuro Resolution Theory, except that, as the "magic" additional CSP variables have their values assigned at the start, they often allow obvious additional assertions.

Consider the case when a sum of 16 [respectively 17] bears on a sector with only two cells: the "magic" CSP variable has value {7 9} [resp. {8 9}] and all the candidate-Numbers with values other than 7 and 9 [resp. 8 and 9] are therefore absent from the initial resolution state. The opposite case is a clue with sum 45 bearing on nine cells: the only possible combination is {1 2 3 4 5 6 7 8 9} and it allows no elimination in its sector.

On some Kakuro websites, the following special "rule" is proposed: if a sector of two cells in a row has a sum of 16 and a sector of two cells in a column has a sum of 17 and if cell (r, c) belongs to the two sectors, then (r, c) = 9. But the result of this combination of magic sectors can be obtained by rules in BRT: as the horizontal controller has value {7, 9} and the vertical controller has value {8, 9}, the initial state can only contain one candidate (9) and rule S can conclude that (r, c) = 9.

### 15.3.2. Coupling rules between controller and controlled variables

As mentioned in our interpretation of theorem 15.1, having transformed the original arithmetic formulation into a binary one based on combinations and associated contradiction links between labels for controller and controlled CSP variables is not quite enough to make Kakuro fully amenable to our approach. One point is, when one introduces CSP variables that are not inherent in the formulation of the CSP, their relationship with the natural ones must somehow be specified. This question was almost hidden in our previous examples by the straightforward way equivalence relations between pre-labels for the different types (Xrc, Xrn, Xcn [and Xbn]) of CSP variables were defined to make labels. For Kakuro, a more explicit coupling between the CSP variables must be defined.

At this point, the links defined between "natural" and "controller" CSP variables allow only to eliminate via standard ECP:

– candidates in a white cell that are not compatible with the sum of one of its horizontal or vertical controllers, when one has been set;

– candidates in a controller cell that are not compatible with the value of a controlled cell, when one has been set.

But they do not allow any elimination before a variable value has been found. They can only be made fully operational if we consider the following four elementary resolution rules, that we shall call the $W_1$-coupling rules:

– **ctr-to-horiz-sector** (from horizontal controller to cells in the sector it controls): in any resolution state, if a candidate-Number is absent from all the combinations for a horizontal sector, then delete it from any cell in this sector; notice that, given our definitions of links and CSP variables, this elimination can be done by a whip[1] with the horizontal controller as its CSP variable;

– ***cell-to-horiz-ctr*** (from a cell to its horizontal controller): in any resolution state, if a cell contains no digit of a combination C, then delete C from its horizontal controller; here again, this elimination can be done by a whip[1] with the cell as its CSP variable;

– the corresponding two "vertical" rules.

At this point, it may be useful to notice that, if a solution to the CSP formulation is given, one never needs these rules in order to prove by standard mathematical means that it is a solution to the initial problem (as can be seen from the proof of theorem 15.1). However, they are required if a solution has to be built constructively from an initial state in which some of the variables have several possible combinations; this is the difference between checking a solution and building one.

It is easy to see that the four coupling rules are whips[1] and they are the only possible types of whips[1]. Adopting them systematically is thus equivalent to using $W_1$ as our minimal resolution theory instead of BRT.

One must also consider another type of coupling rule:

– ***horiz-sector-to-ctr*** (from horizontal sector to controller): in any resolution state, if a candidate-Number is absent from all the cells of a horizontal sector, then delete from the horizontal controller CSP variable any candidate-combination containing it; notice that, as there are in general no Hrcn CSP variables, this cannot be considered as a whip[1] (even a generalised one with missing $llc_1$), although it is akin to a whip[1] and it has the effect of a whip[1]; and, in the (magic and pseudo-magic) cases where there is an Hrcn CSP variable, the conditions of the rule can never be satisfied; as appears from the resolution paths, this rule is activated much less often than the previous ones if it is given lower priority. Of course, there is also a verti-sector-to-ctr rule.

One may wonder whether this introduces a new kind of rule; but it is easy to see that, for a sector of length p, it is equivalent to an $S_{p-1}$-braid[p] – although it appears as a much simpler and natural structure when considered as a coupling rule.

The sequel will show that the six coupling rules are enough to ensure the full resolution potential of the above defined CSP variables and links. We call **BRT$^+$** the union of BRT with these six coupling rules (BRT$^+$ is thus an extension of $W_1$); more generally, for any resolution theory T, we call **T$^+$** the union of T with these six rules. As the last two coupling rules are obviously stable for confluence, ***BRT$^+$ has the confluence property; similarly, if T has the confluence property, so has T$^+$***.

### 15.3.3. Subset rules in Kakuro

Kakuro has Subset rules, but, due to the presence of sectors, they are a little more complex than in our previous examples.

*15.3.3.1. Strict Subset rules*

Strictly speaking (i.e. according to the general definitions of Subsets in chapter 8), there are only three kinds of Subset rules:

– "Naked" Subset rules, both in rows and columns (based either on $Xrc_i$ CSP variables with transversal $rn_j$ constraints or on $Xr_ic$ CSP variables with transversal $cn_j$ constraints);

– "Hidden" Subset rules (based on $Xrn_i$ CSP variables, with transversal $rc_j$ constraints) in rows for horizontal magic sectors or for horizontal pseudo-magic sector-digit cases;

– "Hidden" Subset rules (based on $Xcn_i$ CSP variables, with transversal $rc_j$ constraints) in columns for vertical magic sectors or for vertical pseudo-magic sector-digit cases.

Given the definition of the rc, rn and cn constraints, all these rules can only be applied locally within a sector, not globally in a row or a column. Thus, given three cells in a sector in a row, if their candidate-Numbers belong to a same set of three (formally, they are related by three different contraints of type rn), these Numbers can be eliminated by Naked-Triplets-in-a-row from other cells in the same sector (but not from other cells in the same row outside this sector).

*15.3.3.2. Extended Hidden Subset rules*

Strictly speaking, there are no other Hidden Subset rules (in the sense that they would appear as mere Subset rules when considering the proper CSP variables) than those associated with magic or pseudo-magic sectors (and restricted to the appropriate digits in the latter case). We insist on this point, not for the theoretical reason of making formal distinctions, but mainly for the practical one that the way Hidden Subset rules are presented on Kakuro websites as similar to those of Sudoku may be very misleading.

In the non-magic and non-pseudo-magic cases, one can indeed define elimination rules similar to Hidden Subset rules, with the same sector restriction as in the Naked case. But, because there are no Xrn (or Xcn) CSP variables in these cases, even limited to the sector, such a rule must have the additional restriction that it must have already been proven that all the candidate-Numbers the pattern bears on must be present in the sector (this is never stated on any of the websites we have seen). This happens for instance when all the cells in the sector have been restricted by previous rules to the same p candidate-Numbers, where p is the size of the sector. But this condition is not necessary; it is enough to know that all the candidate-combinations remaining for the controller CSP-variable contain these candidate-Numbers. The problem is, this condition on combinations is more complex than the defining conditions of a Naked Subset; such a "Dynamic Hidden Subset" rule can therefore not be considered as a counterpart of the Naked one for Subsets of same size. The practical question is: in such cases, is it worth introducing these rules or is it better to rely on Naked vs Hidden complementarity and use only Naked Subsets? As we have not programmed such extensions in KakuRules and as Quads are already rare patterns, we leave it open.

*15.3.3.3. Extended Super Hidden Subset rules*

Strictly speaking, there are no Super-Hidden Subset (Fish) rules. One can introduce something similar, a "Dynamic" Super-Hidden Subset, although its application is still more restricted than in the Hidden case: for a "Dynamic" Super-Hidden Subset in rows [respectively columns], all the cells in each row *and* column involved in the pattern must be in the same sector [it is "in each row *and* column" for both horizontal and vertical cases, not respectively for each one]; moreover, it must already have been proven that the candidate-Number involved in the pattern must appear in each of the horizontal [resp. vertical] sectors containing cells of the pattern (the only way this can be granted is when all the combinations remaining for each horizontal [resp. vertical] sector all contain this candidate-Number – as in the Hidden Subset case). A target in a column [resp. a row] must be in the same vertical [resp. horizontal] sector as some cells in the pattern. However, nothing prevents blacks cells to appear in the convex hull of the pattern, outside its rows and columns. The direct proof of this rule is exactly as in Sudoku.

The additional conditions (with respect to the standard formulation of Subset rules) for this extension are the same as for Hidden Subsets: the presence of a candidate-Number in a sector can only be ascertained when all the combinations remaining for the sector all contain this Number.

**15.4. Bivalue-chains, whips and braids in Kakuro**

There are two main remarks about bivalue-chains, whips or braids in Kakuro:

– they entertwine "natural" Xrc and additional Hrc and Vrc CSP variables in an essential way;

– one can observe the same phenomenon as in Sudoku and Futoshiki: when whips and braids are both active, whips are given a higher priority than braids of same length (as in our standard complexity hierarchy) and the simplest-first strategy is used, non-whip braids rarely appear.

As mentioned in the Introduction, when we want to display the resolution path of an instance in a way that allows to consider it as a full proof of the solution, we face the problem of the trivial steps, in particular the boring sequences of eliminations due to direct constraint propagation. In Sudoku, the problem was easily dealt with because the eliminations done by ECP rules are quite obvious and can be omitted with no harm. In Futoshiki, in addition to ECP, we had to deal with ascending chains and the compromise between being boring and being unclear because of discarding too many steps was to "forget" such chains only at the start.

In Kakuro, in addition to ECP (which will never be displayed, even between different types of variables), the interplay between controller and controlled variables (formalised in the coupling rules) is an essential part of the resolution process, at any stage of it (as can be seen from the resolution paths below), and it would be queer not to display the corresponding steps, especially as all the puzzles proposed to beginners can be solved using only these rules.

The convention we shall adopt is that all the obvious eliminations whose effect is to restrict the domains of the natural CSP variables are already done in the initial resolution state $RS_1$, before any application of the Single rule. This can be viewed either as adopting the domain definitions given in section 15.2.4 or as allowing the ctr-to-horiz-sector and ctr-to-verti-sector coupling rules to apply before Singles (until $RS_1$ is reached), even though they do not belong to BRT (they are whips[1] and pseudo-whips[1]). As a result of this initialisation choice, Single assertions that could be available earlier (e.g. as in the case described at the end of section 15.3.1) will appear only after $RS_1$. In practice, this amounts to delaying any Single application a human player is likely to do during this initialisation phase.

We also adopt the convention that constraints in magic sectors are transformed from the start (after the above Xrc domain restrictions) into given values for the associated Hrc or Vrc variables and identified as such in the first two lines of the resolution path.

After this initialisation phase, it is natural to grant Singles a higher priority than coupling rules and it would be misleading not to display all of the instances of the latter. [As all the interesting resolution theories must contain the coupling rules and as only easy puzzles can be solved at such levels, this priority has no impact on our further analyses of harder puzzles and on their classifications.] The resolution path will generally start with a series of Singles for the Xrc, Hrc and Vrc variables.

### 15.4.1. Full resolution path of the puzzle in Figure 15.1

This section gives the full resolution path of the puzzle in Figure 15.1. Braids, whips, bivalue chains, Naked and Hidden Subsets (in magic sectors) are active. We have chosen this moderately difficult example because it has Naked and Hidden Subsets (although these instances could be replaced by whips). Although active, braids do not appear. The most interesting steps are in bold.

It should be noticed how Naked Singles for "natural" Xrc and for Hrc and Vrc variables are entertwined: in Kakuro, there is a permanent interplay between the two types of variables. It is thus worth for the reader to spend some time on checking (at least in the bivalue-chains[2] and whips[2]) how this general idea is materialised in most of the chains appearing in the resolution path.

***** KakuRules 1.2 based on CSP-Rules 1.2, config: B⁺ *****
**horizontal-magic-sectors ==> hr11c6 = 5789, hr9c3 = 124, hr5c1 = 356789, hr3c7 = 1235, hr3c1 = 56789**
**vertical-magic-sectors ==> vr7c11 = 124, vr2c9 = 123456789, vr6c8 = 13, vr2c8 = 13, vr9c7 = 79, vr4c6 = 46789, vr1c4 = 123456789, vr1c3 = 1235, vr6c2 = 13**
naked-singles ==> r11c8 = 5, r10c7 = 7, r11c7 = 9, r9c11 = 4, r9c6 = 4, r6c6 = 6, r5c11 = 6, r3c11 = 5, r3c3 = 5, r3c6 = 6, r5c3 = 3, vr1c6 = 16, r2c6 = 1, r2c3 = 2, r4c3 = 1, vr1c11 = 58, r2c11 = 8, hr2c9 = 78, r2c10 = 7, vr1c10 = 37, r3c10 = 3, r3c8 = 1, r3c9 = 2, r4c8 = 3, hr5c8 = 126, r5c9 = 1, r5c10 = 2, vr4c10 = 25, r6c10 = 5, vr4c11 = 69, r6c11 = 9, hr6c8 = 459, r6c9 = 4, hr6c3 = 1236, hr9c8 = 489, hr10c6 = 12347, r10c9 = 3, vr9c8 = 25, r10c8 = 2, r10c11 = 1, r10c10 = 4, r8c11 = 2
ctr-to-horiz-sector ==> r4c7 ≠ 8, r4c9 ≠ 8, r4c7 ≠ 4, r4c7 ≠ 5, r4c9 ≠ 5
ctr-to-horiz-sector ==> r4c2 ≠ 5, r4c4 ≠ 5 ; ctr-to-verti-sector ==> r8c10 ≠ 1, r8c10 ≠ 3
cell-to-horiz-ctr ==> hr7c1 ≠ 249, hr7c1 ≠ 258, hr7c1 ≠ 267 ; ctr-to-horiz-sector ==> r7c4 ≠ 2
cell-to-horiz-ctr ==> hr7c1 ≠ 456 ; cell-to-horiz-ctr ==> hr7c5 ≠ 2589, hr7c5 ≠ 2679
ctr-to-horiz-sector ==> r7c7 ≠ 2 ; cell-to-horiz-ctr ==> hr7c5 ≠ 4569, hr7c5 ≠ 4578
cell-to-horiz-ctr ==> hr11c1 ≠ 56 ; ctr-to-horiz-sector ==> r11c3 ≠ 5, r11c3 ≠ 6
cell-to-verti-ctr ==> vr1c5 ≠ 29 ; ctr-to-verti-sector ==> r3c5 ≠ 9 ; cell-to-verti-ctr ==> vr4c5 ≠ 34
ctr-to-verti-sector ==> r6c5 ≠ 3 ; cell-to-verti-ctr ==> vr1c5 ≠ 56
ctr-to-verti-sector ==> r2c5 ≠ 5, r2c5 ≠ 6 ; cell-to-verti-ctr ==> vr6c3 ≠ 67
ctr-to-verti-sector ==> r7c3 ≠ 6, r7c3 ≠ 7 ; verti-sector-to-ctr ==> vr2c2 ≠ 178, vr2c2 ≠ 169
biv-chain[2]: r8n3{c7 c8} – r8n1{c8 c7} ==> r8c7 ≠ 9, r8c7 ≠ 8, r8c7 ≠ 7, r8c7 ≠ 6, r8c7 ≠ 5, r8c7 ≠ 4
horiz-sector-to-ctr ==> hr8c5 ≠ 123468 ; ctr-to-horiz-sector ==> r8c6 ≠ 8, r8c9 ≠ 8, r8c10 ≠ 8
horiz-sector-to-ctr ==> hr8c5 ≠ 123459
naked-singles ==> hr8c5 = 123567, r8c6 = 7
biv-chain[2]: vr7c10{n4579 n4678} – r9c10{n9 n8} ==> r11c10 ≠ 8
naked-singles ==> r11c10 = 7, r11c9 = 8, r9c9 = 9, r9c10 = 8, vr7c10 = 4678, r8c10 = 6, r8c9 = 5
cell-to-horiz-ctr ==> hr7c5 ≠ 3489 ; ctr-to-horiz-sector ==> r7c7 ≠ 4
verti-sector-to-ctr ==> vr3c7 ≠ 13456, vr3c7 ≠ 12457, vr3c7 ≠ 12349
ctr-to-verti-sector ==> r5c7 ≠ 9, r7c7 ≠ 9
biv-chain[2]: vr8c5{n14 n23} – r9c5{n1 n2} ==> r10c5 ≠ 2
biv-chain[2]: vr6c3{n49 n58} – r8c3{n4 n5} ==> r7c3 ≠ 5
cell-to-horiz-ctr ==> hr7c1 ≠ 357 ; ctr-to-horiz-sector ==> r7c4 ≠ 7
biv-chain[2]: hr8c1{n125 n134} – r8c3{n5 n4} ==> r8c4 ≠ 4
biv-chain[2]: hr4c6{n137 n236} – r4c9{n7 n6} ==> r4c7 ≠ 6
biv-chain[2]: hr2c2{n1236 n1245} – r2c5{n3 n4} ==> r2c4 ≠ 4
biv-chain[2]: r9c5{n1 n2} – vr8c5{n14 n23} ==> r10c5 ≠ 1
cell-to-horiz-ctr ==> hr10c1 ≠ 1257
biv-chain[2]: r8c3{n4 n5} – vr6c3{n49 n58} ==> r7c3 ≠ 4
biv-chain[2]: r2c5{n3 n4} – hr2c2{n1236 n1245} ==> r2c4 ≠ 3
whip[2]: r4c9{n7 n6} – hr4c6{n137 .} ==> r4c7 ≠ 7
whip[2]: r7c2{n1 n3} – hr7c1{n168 .} ==> r7c4 ≠ 1
whip[2]: r7c2{n3 n1} – hr7c1{n348 .} ==> r7c4 ≠ 3
whip[2]: r7c3{n8 n9} – hr7c1{n348 .} ==> r7c4 ≠ 8

whip[2]: r7c3{n9 n8} – hr7c1{n159 .} ==> r7c4 ≠ 9
whip[2]: r7c8{n1 n3} – hr7c5{n1689 .} ==> r7c7 ≠ 1
whip[2]: r7c8{n3 n1} – hr7c5{n3678 .} ==> r7c7 ≠ 3
whip[2]: r8c3{n5 n4} – hr8c1{n125 .} ==> r8c4 ≠ 5
whip[2]: hr10c1{n1248 n1239} – r10c3{n4 .} ==> r10c4 ≠ 9
whip[2]: hr11c1{n29 n47} – r11c2{n2 .} ==> r11c3 ≠ 4
whip[2]: r5c2{n8 n9} – r3c2{n9 .} ==> vr2c2 ≠ 349
whip[2]: vr2c2{n457 n259} – r3c2{n7 .} ==> r5c2 ≠ 9
whip[2]: vr9c2{n23 n14} – r11c2{n2 .} ==> r10c2 ≠ 4
whip[2]: vr9c3{n49 n58} – r11c3{n7 .} ==> r10c3 ≠ 8
whip[2]: r10c5{n3 n4} – r10c3{n4 .} ==> hr10c1 ≠ 1248
ctr-to-horiz-sector  ==> r10c4 ≠ 8
whip[2]: vr9c3{n49 n67} – r11c3{n8 .} ==> r10c3 ≠ 7
whip[2]: vr9c3{n58 n49} – r11c3{n7 .} ==> r10c3 ≠ 9
cell-to-horiz-ctr  ==> hr10c1 ≠ 1239
**naked-triplets-in-a-column c4{r6 r8 r9}{n2 n3 n1} ==> r10c4 ≠ 3, r10c4 ≠ 2, r10c4 ≠ 1, r4c4 ≠ 3**
whip[2]: hr4c1{n128 n137} – r4c4{n8 .} ==> r4c2 ≠ 7
**naked-triplets-in-a-column c4{r6 r8 r9}{n2 n3 n1} ==> r4c4 ≠ 2**
whip[2]: hr4c1{n137 n128} – r4c4{n7 .} ==> r4c2 ≠ 8
whip[4]: r5c6{n8 n9} – c4n9{r5 r3} – r3c2{n9 n7} – vr2c2{n358 .} ==> r5c2 ≠ 8
whip[4]: vr3c7{n12367 n12358} – r5c7{n7 n8} – c6n8{r5 r7} – hr7c5{n3579 .} ==> r7c7 ≠ 5
horiz-sector-to-ctr  ==> hr7c5 ≠ 3579
whip[2]: vr3c7{n12367 n12358} – r7c7{n7 .} ==> r5c7 ≠ 8
**hidden-pairs-in-a-row r5{n8 n9}{c4 c6} ==> r5c4 ≠ 7, r5c4 ≠ 6, r5c4 ≠ 5**
**biv-chain[5]: c8n3{r7 r8} – r8n1{c8 c7} – r4c7{n1 n2} – hr4c6{n137 n236} – c9n7{r4 r7} ==> hr7c5 ≠ 1689**
naked-singles ==> hr7c5 = 3678, r7c8 = 3, r8c8 = 1, r8c7 = 3, r7c6 = 8, r5c6 = 9, r5c4 = 8
hidden-single-in-magic-verti-sector ==> r3c4 = 9
cell-to-horiz-ctr  ==> hr4c1 ≠ 128 ; ctr-to-horiz-sector  ==> r4c2 ≠ 2
cell-to-verti-ctr  ==> vr2c2 ≠ 259 ; cell-to-verti-ctr  ==> vr3c7 ≠ 12358
naked-single ==> vr3c7 = 12367
verti-sector-to-ctr  ==> vr2c2 ≠ 268
**naked-pairs-in-a-row r6{c5 c7}{n1 n2} ==> r6c4 ≠ 2, r6c4 ≠ 1**
naked-single ==> r6c4 = 3
whip[2]: r3c2{n7 n8} – vr2c2{n457 .} ==> r5c2 ≠ 7
hidden-single-in-magic-horiz-sector ==> r5c7 = 7
naked-singles ==> r7c7 = 6, r7c9 = 7, r4c9 = 6, hr4c6 = 236, r4c7 = 2, r6c7 = 1, r6c5 = 2, vr4c5 = 25, r5c5 = 5, r5c2 = 6, vr2c2 = 367, r4c2 = 3, r3c2 = 7, r3c5 = 8, vr1c5 = 38, r2c5 = 3, hr2c2 = 1236, r2c4 = 6, hr4c1 = 137, r4c4 = 7
cell-to-horiz-ctr  ==> hr7c1 ≠ 168 ; horiz-sector-to-ctr  ==> hr10c1 ≠ 1347
biv-chain[2]: hr10c1{n1356 n2346} – r10c4{n5 n4} ==> r10c5 ≠ 4
naked-singles ==> r10c5 = 3, vr8c5 = 23, r9c5 = 2, r9c4 = 1, r8c4 = 2, hr8c1 = 125, r8c2 = 1, r7c2 = 3, r8c3 = 5, vr6c3 = 58, r7c3 = 8, hr7c1 = 348, r7c4 = 4, r10c4 = 5, hr10c1 = 1356, r10c2 = 1, r10c3 = 6, vr9c3 = 67, r11c3 = 7, hr11c1 = 47, r11c2 = 4, vr9c2 = 14
Grid solved. Hardest step: Bivalue-Chain[5].

## 15.5. Theory of g-labels in Kakuro

Applying the general definition of a g-label to Kakuro is not as straightforward as in our previous CSP examples; in particular, we need investigate how the general condition of "saturation" or "local maximality" concretely appears when applied to sets of digits and sets of combinations.

As there can be no g-label in magic sectors, in all this section we suppose that (S, p) is a non-magic pair.

### 15.5.1. General preliminaries

For convenience, let us first repeat the definition of a g-label given in section 7.1.1.1. A *potential-g-label* is a pair <V, g>, where V is a CSP variable and g is a set of labels for V, such that:

– the cardinality of g is greater than one, but g is not the full set of labels for V;

– there is at least one label l such that l is not a label for V and l is linked (possibly by different constraints) to all the labels in g.

A *g-label* is a potential-g-label <V, g> that is "saturated" or "locally maximal" in the sense that, for any potential g-label <V, g'> with g' strictly larger than g (as sets of labels), there is a label l that is not a label for V and that is linked to all the elements of g but not to all the elements of g'.

The following three remarks show that the definition of g-labels is completely taken care of by the next sub-sections.

1) There is always a one-to-one correspondence between the labels <X, v> for a CSP variable X and the elements v of its domain (by the construction of pre-labels). We shall use it freely (i.e. we shall make no distinction at all between the corresponding elements) in the following two cases:

– for a fixed Xrc variable with parameters $(S_H, p_H)$ and $(S_V, p_V)$, the obvious correspondence between labels for Xrc and $(S_H, p_H)$-compatible and $(S_V, p_V)$-compatible digits;

– for a fixed Hrc [or Vrc] variable in a sector with parameters (S, p), the obvious label-to-combination correspondence, in which case we shall also use freely the obvious correspondences between symbols $n_1...n_p$ appearing in the labels, combinations in Comb(S, p) and subsets $\{n_1, ... n_p\}$ of p (S, p)-compatible digits.

2) For each sector, a g-label for the controller variable will be g-linked to a label for a cell in the sector, depending only on their values (respectively set of combinations and digit), not on the exact position of the cell in the sector. Similarly, a g-label for a cell in the sector will be g-linked to a label for the controller variable, depending only on their values (respectively set of digits and combination).

3) As mentioned in chapter 7, the saturation condition in the definition of a g-label is there mainly for reasons of efficiency. Too many useless g-labels would lead to too many redundant partial g-whips, many of which would differ only by g-labels that exclude the same candidates. When it was first introduced and illustrated by the Sudoku case, this condition did not make a spectacular difference. But we shall see that it is essential in practice for Kakuro.

### 15.5.2. Mutual exclusion between sets of combinations and sets of digits

For a legitimate (S, p) pair, we defined at the end of section 15.1.1 the set Comb(S, p) of all the (S, p)-compatible combinations, i.e. of all the combinations of p different digits with sum S. As can be seen from Table 15.2, the number of such combinations is always an integer in the range [1, …, 12]. Table 15.1 shows that there are thirty four "magic" (S, p) pairs that have only one combination and Table 15.4 shows that there are fifteen "pseudo-magic" (S, p) pairs that have digits (up to five) common to all their combinations. We shall now study more complex properties of Comb(S, p).

We shall be interested in particular subsets of Comb(S, p) and particular subsets of Compat(S, p) that exclude each other, the sets gComb(S, p) and gDig(S, p). They will play a major role in the definition of g-labels and their g-links.

#### 15.5.2.1. Mutual exclusion of digits and combinations

Definition: a digit $i \in$ Compat(S, p) and a combination $C \in$ Comb(S, p) exclude each other if $i \notin C$. We also say that C excludes i or that i excludes C, but this basic exclusion relation is fundamentally symmetric.

Definition: a set of digits gD $\subset$ Compat(S, p) excludes a combination C if every digit $i \in$ gD excludes C, i.e. if gD $\subset C^c$. A set of digits gD $\subset$ Compat(S, p) excludes a set of combinations gC $\subset$ Comb(S, p) if it excludes every combination $C \in$ gC, i.e. if gD $\subset \cap \{C^c, C \in gC\}$. Here, complementation is taken in Compat(S, p) and "⊂" is understood in the non-strict sense.

Definition: a set of combinations gC $\subset$ Comb(S, p) excludes a digit i if every combination $C \in$ gC excludes i, i.e. if $i \in \cap \{C^c, C \in gC\}$. A set of combinations gC $\subset$ Comb(S, p) excludes a set of digits gD $\subset$ Compat(S, p) if it excludes every digit $i \in$ gD, i.e. if gD $\subset \cap \{C^c, C \in gC\}$.

Exclusion between a set of digits and a set of combinations is obviously a symmetric relation, but in the context of g-labels we shall generally use it in unsymmetric ways, whence the separate definitions.

If gD $\subset$ Compat(S, p), we note D-Excl(gD) the set of combinations in Comb(S, p) excluded by gD. If gC $\subset$ Comb(S, p), we note C-Excl(gC) the set of digits in Compat(S, p) excluded by gC. D-Excl is thus a function from subsets of Compat(S, p) to subsets of Comb(S, p) and C-Excl a function from subsets of Comb(S, p) to subsets of Compat(S, p). As which of the two is concerned is obvious from the argument, we shall often write them loosely as Excl(gD) and Excl(gC).

*15.5.2.2. Envelopes*

It is obvious that D-Excl and C-Excl are decreasing functions: if $gD_1 \subset gD_2$, then $D\text{-Excl}(gD_2) \subset D\text{-Excl}(gD_1)$; if $gC_1 \subset gC_2$, then $C\text{-Excl}(gC_2) \subset C\text{-Excl}(gC_1)$. This remark justifies the following definitions.

Definiton: the envelope Env(gD) of a set of digits $gD \subset$ Compat(S, p) is the maximum superset of gD in Compat(S, p) that excludes the same combinations as gD. It is obviously the set of all the digits in Compat(S, p) that exclude Excl(gD).

Definition: the envelope Env(gC) of a set of combinations $gC \subset$ Comb(S, p) is the maximum superset of gC in Comb(S, p) that excludes the same digits as gC. It is obviously the set of all the combinations in Comb(S, p) that exclude Excl(gC).

It is obvious that mutual exclusion of a set of combinations $gC \subset$ Comb(S, p) and a set of digits $gD \subset$ Compat(S, p) entails mutual exclusion of their envelopes.

We now turn our attention to "saturated" or "locally maximum" subsets of digits and combinations.

*15.5.2.2. gDigs*

Definition: a potential-g-digit(S, p) is a subset gD of Compat(S, p):

– containing at least two elements of Compat(S, p) but not all of Compat(S, p),

– excluding at least one combination $C \in$ Comb(S, p).

Definition: a g-digit(S, p) is a potential g-digit(S, p) that is "saturated" or "locally maximal" in the sense that any strictly larger (with respect to set-theoretic inclusion) potential-g-digit(S, p), if any, excludes a strictly smaller subset of Comb(S, p). Equivalently: *a g-digit is a potential-g-digit that is equal to its envelope*. We call this the "saturation" or "local-maximality" property of g-digits. We define ***gDig(S, p)*** as the set of all the g-digits(S, p).

Remarks:

– any $C \in$ Comb(S, p), if considered as a subset of Compat(S, p), is a g-digit(S, p) as soon as the sector is not magical; but we shall see that there are many other cases of g-digits;

– any g-digit contains all the digits common to all the combinations in Comb(S, p).

**Theorem 15.2: *if gD $\in$ gDig(S, p), then Excl(Excl(gD)) = gD.***

Remark: as exclusion is a symmetric relation, we already know that any digit in gD is excluded by the set of combinations Excl(gD), i.e. that gD ⊂ Excl(Excl(gD)). What the theorem says is that there are no other digits excluded by Excl(gD).

Proof: by the saturation of gD, for any digit i ∈ Compat(S, p) such that i ∉ gD, i ∪ gD excludes a set of combinations strictly smaller than Excl(gD). There is therefore some combination C in Excl(gD) such that C is not excluded by i ∪ gD. As C is excluded by gD (i.e. by every digit in gD), it can only mean that C is not excluded by i. By the symmetry of exclusion, i is not excluded by C. Therefore i is not excluded by Excl(gD). qed.

*15.5.2.3. gCombs*

We can now repeat for sets of combinations all that was done for sets of digits.

Definition: a potential-g-combination(S, p) is a subset gC of Comb(S, p):
– containing at least two elements of Comb(S, p) but not all of Comb(S, p),
– excluding at least one digit i ∈ Compat(S, p).

Definition: a *g-combination(S, p)* is a potential-g-combination(S, p) such that any strictly larger (with respect to set-theoretic inclusion) potential-g-combination(S, p), if any, excludes a strictly smaller set of digits. Equivalently: *a g-combination is a potential-g-combination that is equal to its envelope*. We call this the "saturation" or "local-maximality" property of g-combinations. We define **gComb(S, p)** as the set of all the g-combinations(S, p).

**Theorem 15.3: if gC ∈ gComb(S, p), then Excl(Excl(gC)) = gC.**

Remark: as exclusion is a symmetric relation, we already know that any combination in gC is excluded by the set of digits Excl(gC), i.e. that gC ⊂ Excl(Excl(gC). What the theorem says is that there are no other combinations excluded by Excl(gC).

Proof: by the saturation of gC, for any combination D in Comb(S, p) such that D ∉ gC, D ∪ gC excludes a set of digits strictly smaller than C-Excl(gC). There is therefore some digit i in C-Excl(gC) such that i is not excluded by D ∪ gC. As i is excluded by gC (i.e. by every combination in gC), it can only mean that i is not excluded by D. By the symmetry of exclusion, D is not excluded by i. Therefore D is not excluded by C-Excl(gC). qed.

*15.2.3.4. Relationship between gDigs and gCombs*

The previous three sub-sections illustrate the duality between g-digits and g-combinations. The following theorem pushes it further.

We first need to set apart the cases in which only one label would be excluded. Let us therefore define gDig⁻(S, p) as the subset of elements gD of gDig(S, p) such that gD excludes at least two combinations from Comb(S, p). Similarly, define gComb⁻(S, p) as the subset of elements gC of gComb(S, p) such that gC excludes at least two digits from Compat(S, p).

***Theorem 15.4:** if gD ∈ gDig⁻(S, p), then Excl(gD) ∈ gComb⁻(S, p). If gC ∈ gComb⁻(S, p), then Excl(gC) ∈ gDig⁻(S, p). D-Excl defines a one-to-one correspondence between gDig⁻(S, p) and gComb⁻(S, p); C-Excl defines the inverse one-to-one correspondence between gComb⁻(S, p) and gDig⁻(S, p).*

Proof: we shall prove only the first part; the second part is easily obtained by duality; and the third is an obvious corollary to the first two. Suppose that gD is a g-digit(S, p) excluding at least two combinations $C_1$ and $C_2$ and consider the set of combinations Excl(gD). It contains at least two elements (namely $C_1$ and $C_2$) but it is not the full set Comb(S, p) because no digit in Compat(S, p) can exclude all of Comb(S, p). Excl(gD) excludes at least two digits in Compat(S, p), indeed it excludes all the digits in gD. There remains only to show that it is saturated. But, for any combination C excluding all of gD, i.e. C ∈ Excl(Excl(gD)), theorem 15.2 shows that C ∈ gD.

### 15.5.3. Representation of a g-combination as a number

The definition of a g-digit(S, p) leads to easy computations. However, a gComb(S, p), say gC, is a set of sets of digits and we still miss a simple way of representing it. This can easily be palliated by defining Env'(gC) as the set of digits compatible with gC [or, equivalently, with Env(gC)]. It is obvious that two different g-combs have different Env' values; we can therefore represent gC by Env'(gC) – more precisely by the number Env*(gC) obtained by glueing together, in ascending order, the elements of Env'(gC). This is convenient because the digits excluded by gC will be the complement of Env'(gC) in Compat(S, p).

### 15.5.4. More on gComb(S, p)

The definition of a gComb(S, p) leads to easy computations, showing that there are 63 (S, p) pairs (out of the 120 legitimate ones) that have g-combs. When an (S, p) pair has g-combs, it has at least 3 and at most 77. The latter happens in only four cases: (14, 3), (15, 3), (16, 3) and (20, 4). There are more than 10 g-combs in 49 cases. We cannot display all the possibilities here, but the following simple example illustrates the notion of saturation of g-combs in a concrete case.

Pair (p, S) = (3, 10 ) has 4 combs: {127 136 145 235} and 9 g-combs :
g-comb 12345 contains combs (145 235) and excludes digits (6 7)
g-comb 12356 contains combs (136 235) and excludes digits (4 7)

  g-comb 12357 contains combs (127 235) and excludes digits (4 6)
  g-comb 12367 contains combs (127 136) and excludes digits (4 5)
  g-comb 12457 contains combs (127 145) and excludes digits (3 6)
  g-comb 13456 contains combs (136 145) and excludes digits (2 7)
  g-comb 123456 contains combs (136 145 235) and excludes digit (7)
  g-comb 123457 contains combs (127 145 235) and excludes digit (6)
  g-comb 123567 contains combs (127 136 235) and excludes digit (4)

It is interesting to consider the two g-combs 12345 and 123456 (or 123457): they show that, in accordance with our general definition, saturation does not mean an absolute but a local maximum. 123456 (or 123457) contains more combs than 12345, but it excludes fewer digits.

Notice that, with this (p, S) = (3, 10) example, there are 4 combinations, which could lead to considering $2^4-4 = 12$ subsets of more than one combinations if we did not have the saturation condition, whereas it is useful to consider only 9 such subsets, namely the 9 g-combs.

The reduction is still more impressive with a pair such as (5, 25): it has 12 combinations and therefore $2^{12}-12 = 4084$ subsets of more than one combinations, but only 37 g-combs. This shows that, in Kakuro, the saturation condition is essential for the practical use of g-labels.

### 15.5.5. *Missing an example with g-bivalue-chains, g-whips and g-braids*

This sub-section will remain (almost) blank as a reminder that an example with g-whips is missing. Although we have programmed g-labels compliant with the above theory[16], we have found no Kakuro puzzle with a g-whip elimination in all those we have tried. This is almost certainly not due to some bug in our implementation: for any length, lots of partial g-whips that are not partial-whips are found (and we have checked that they are correct). We face here the problem evoked in the Introduction. Very little is known about CSPs other than Sudoku: no exceptionally hard cases, no instances with specific patterns, no forums where to submit problems… The same remarks will apply to the Numbrix® and Hidato® puzzles in the next chapter.

### 15.6. Application-specific rules in Kakuro: surface sums

The only type of application-specific rule we have met on all the Kakuro websites we have visited and in all the available literature we have seen is what we shall call "surface sums". But the simplest and most general way of expressing it is in terms of a cut in the graph underlying the puzzle (as defined below). It is a specificity of Kakuro, with respect to our previous examples, that some puzzles can be reduced, in rather straightforward ways, to several (easier) sub-puzzles. It raises the question of whether the condition of well-formedness should exclude reducibility.

---

[16] g-whips and g-braids are present in CSP-rules as generic rules, but they must be fed by the application-specific definition of g-labels.

### 15.6.1. *Graph underlying a Kakuro puzzle*

Definition: the (undirected) graph underlying a Kakuro puzzle P is composed of:

– a set of vertices (or nodes): one for each white cell of P;

– a set of edges (or arcs): there is an (undirected) edge between two nodes if and only if the corresponding white cells are (horizontally or vertically) adjacent.

Definitions (standard from graph theory): in an undirected graph, a path between two nodes $C_1$ and $C_2$ is a sequence of nodes starting in $C_1$ and ending in $C_2$, such that there is an arc between any two consecutive nodes in the sequence. Two nodes are connected if there is a path between them. A graph is connected if there is a path between any two nodes.

Definitions (standard from graph theory): a cut is a set of nodes whose removal makes the graph disconnected. A graph is k-connected if no cut of k-1 (or fewer) nodes can disconnect it. The connectivity of a graph is the smallest k such that there is a cut of size k disconnecting it.

The above definitions can be transferred to any Kakuro puzzle via its underlying graph: a Kakuro puzzle is connected if its underlying graph is connected. As already mentioned, if a puzzle is not connected, it is often reducible to several independent puzzles (see details in the forthcoming examples). But, even a connected puzzle can sometimes be decomposed into independent ones.

Definition: a cell C disconnects a Kakuro puzzle into two parts $S_1$ and $S_2$ if any path from any cell $C_1 \in S_1$ to any cell $C_2 \in S_2$ passes through C. In terms of graphs, this is equivalent to saying that C is a cut of the underlying graph.

A Kakuro puzzle cannot be disconnected by any cell if and only if its underlying graph is 2-connected.

### 15.6.2. *The "surface sum" rule*

Many websites mention a "surface sum" rule dealing with almost closed surfaces in which a cell C is included in the horizontal sums of the sectors making the surface but not in the vertical ones [or conversely]. Figure 15.4 shows two such situations, the simplest possible and one more complex. In these cases, the sum of the cells on the surface can be computed in two ways: sum of horizontal clues (including C) and sum of vertical clues (excluding C). The value of C is then obtained directly as the difference between these two sums. By transposing rows and columns, one obtains similar examples. In each case, the condition for the rule to work is that any sector completely included in the surface (i.e. all except at most one containing C) has a clue. In all the websites we have seen, this situation is described by examples but not formalised in the general and much simpler terms allowed by graph theory.

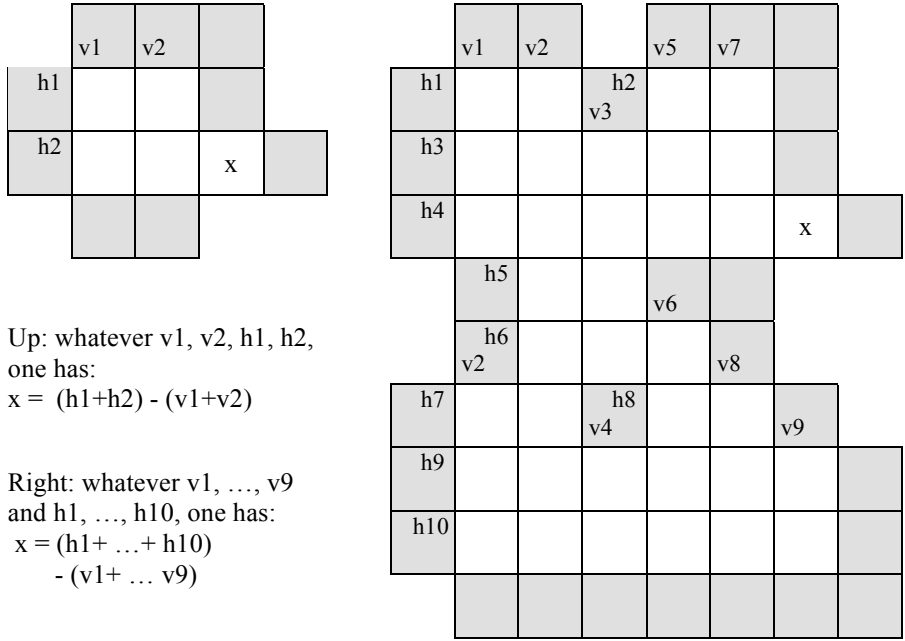

Up: whatever v1, v2, h1, h2, one has:
x = (h1+h2) - (v1+v2)

Right: whatever v1, ..., v9 and h1, ..., h10, one has:
x = (h1+ ...+ h10)
    - (v1+ ... v9)

**Figure 15.4.** *Two examples of the domain sum rule (only the relevant parts of the puzzles are shown; indicated h and v values and black cells are compulsory; black cells without explicit clues may have clues or not; all the rest of the puzzle, i.e. the part not shown here, is free; in particular, there may be white cells under x; inner cell with clues h8/v4 is not a problem).*

### 15.6.3. The "cut rule"

The above "surface rule" has a much more general counterpart, based on the notion of a cut. It has different conditions and conclusions, but it does not have to be restricted to unions of horizontal and vertical sectors that differ by only one cell.

**Theorem 15.6 (the strong cut rule):** *if in a Kakuro puzzle P there is a cell C that disconnects P (i.e. its underlying graph) into two parts $P_1$ and $P'_1$ such that:*
  – *all the sectors meeting $P_1$ have a clue,*

  – *one and only one of the horizontal or vertical sectors of C is entirely in $P_1$,*

  – *each of $P_1$ and $P'_1$ has a unique solution,*

then P is equivalent to two independent Kakuro sub-puzzles with respective white cells those of $P_1$ and $P'_1$; the clues for the newly created sector in each of $P_1$ and $P'_1$ are obtained by computing the differences in the vertical and horizontal sums of all the sectors at least partly in $P_1$.

Notice that this general graph-based formulation does not prevent black cells to appear between white ones in the surface (as in the rightmost part of Figure 15.4), provided that they have clues for all the sectors they control. Also, this puts no constraint on the place of C in its row or column.

It should be noticed that uniqueness of the sub-puzzles (without any reference to their origin) is not a consequence of uniqueness of the initial puzzle. One way to understand this is that some digits that were not compatible with the global sum may become compatible with either of the split sums. See the examples below.

In practice, as shown by the example below, the above theorem may still be too restrictive and *the uniqueness condition for $P_1$ and $P'_1$ can be relaxed*, if either one gives up full independence of the sub-puzzles or one adds some condition to ensure that their solutions are compatible (i.e. that the parts of the sector that has been split into two must have different digits). For instance, if only one of the two sub-puzzles has a unique solution, the information relative to the values of the cells of its half sector in the solution must be transferred to the complementary half sector in the other sub-puzzle or the solution of the values of the solved sub-puzzle cells can merely be re-injected into the global one.

Instead of giving a formal proof of the theorem, which would require boring technicalities but would not bring much insight into the elementary way this rule can be used, we shall illustrate how it works on the puzzle P in Figure 15.5. (To preserve the square grid hypothesis, one can always assume that five black columns are added to the right).

The generalised cut rule can be applied repeatedly six times (cells $C_1$ to $C_6$ are shown in Figure 15.5), leading to the situation represented in Figure 15.6:

  – $C_1$ disconnects P into two sub-puzzles: the first (small) part $P_1$ consists of the 4×4 sub-puzzle made of the first four rows and columns, with horizontal clue 32 in cell r4c1 replaced by clue (15+23+9) - (10+15) = 22; the second (and main) part $P'_1$ is obtained by replacing all the cells in $P_1$ by black cells with no clues, except an horizontal clue 10 = 32 - 22 in cell r4c4;

  – $C_2$ disconnects $P'_1$ into a small part $P_2$ with vertical clue (15+16) - (16) = 15 in r8c3 and a second sub-puzzle $P'_2$ with vertical clue 35 in r5c3 replaced 35-15 = 20;

  – $C_3$ disconnects $P'_2$ into a small part $P_3$ with horizontal clue 16 in r8c2 replaced by (10+20+6) - (8+19) = 9 and a third sub-puzzle $P'_3$ with horizontal clue 16-9 = 7 in r8c4;

– $C_4$ disconnect P'$_3$ into a small P$_4$ with horizontal clue (24+12+7) – (11+16) = 16 in r12c8 and a fourth sub-puzzle P'$_4$ with horizontal clue 33 in cell r12c6 replaced by 33-16 = 17;

– $C_5$ disconnect P'$_4$ into a small part P$_5$ with vertical clue 24 in r7c11 replaced by (11+5) - 13 = 3 and a fifth sub-puzzle P'$_5$ with vertical clue 24-3 = 21 in r9c10;

| $\mathcal{K}$ | 15 | 23 | 9 |  |  | 14 | 16 |  | 34 | 7 |
|---|---|---|---|---|---|---|---|---|---|---|
| 10 |  |  |  |  | 12\20 |  |  | 11\20 |  |  |
| 15 |  |  |  | 22\3 |  |  |  |  |  |  |
| 32 |  |  | $C_1$ |  |  | 23\41 |  |  |  | 4 |
|  | 10 | 35 | 18\6 |  |  |  | 14\10 |  |  |  |
| 8 |  |  |  |  | 15\31 |  |  |  |  |  |
| 19 |  |  |  | 26\7 |  |  |  |  | 24 | 13 |
|  | 16\16 |  | $C_3$ |  |  |  | 17 | 5 |  |  |
| 15 |  | $C_2$ | 29 |  |  |  |  | 11\7 | $C_5$ |  |
| 16 |  |  | 17 | 18\13 |  |  |  | $C_6$ |  | 9 |
|  | 17 | 27\26 |  |  |  |  | 21 |  |  |  |
| 23 |  |  |  |  |  | 17 | 11\12 |  |  |  |
| 24 |  |  |  | 13\21 |  |  |  |  | 24 | 12 | 7 |
|  | 11\8 |  |  |  | 33\4 |  |  | $C_4$ |  |  |
| 32 |  |  |  |  |  |  | 11 |  |  |  |
| 6 |  |  | 6 |  |  | 16 |  |  |  |  |

**Figure 15.5.** *A 16×11 puzzle (of unknown origin) with 6 cuts*

– $C_6$ disconnect $P'_5$ into a small part $P_6$ with horizontal clue (7+21+9)-(21+11) = 5 in r10c8 and a sixth sub-puzzle $P'_6$ with horizontal clue 18 in r10c5 replaced by 18-5 = 13.

*Figure 15.6. $P'_6$, the puzzle of Figure 15.5 after the 6 cuts have been applied*

Notice that, in Figure 15.6, each of $C_7$, $C_8$, $C_9$ and $C_{10}$ also disconnects the remaining puzzle $P'_6$ into two parts, but they do not satisfy the additional condition that one of the sectors containing C is totally in one of the two parts. [In such cases, the surface sum information could nevertheless be used, e.g. to introduce new Hrc and Vrc CSP variables in intermediate virtual cells together with equalities between their sums, but this is another topic.]

Notice also the particular situation of pairs of cells $(D_1, D'_1)$, $(D_2, D'_2)$ and $(D_2, D''_2)$: each pair has a potential of separating the puzzle into two sub-puzzles; but this information cannot be used as such, without additional conditions. However, in the two cases, as cells $C_7$, $C_8$, $C_9$ and $C_{10}$ or pairs $(D_1, D'_1)$, $(D_2, D'_2)$ and $(D_2, D''_2)$ are the key for splitting the puzzle into smaller ones, a reasonable heuristic would suggest to start by trying to find their values.

### 15.6.4. What is the effect of the cut rule on a whip based solution?

There now arises a natural question about the effect of the cut rule on the difficulty of solving a puzzle. Depending on how the original puzzle is split into pieces, it can have very different consequences.

In most of the cases we have seen, applying the cut rule at the start made it significantly simpler or it even turned it into an almost obvious instance. But no rigorous statistical meaning should be understood here: this remark is only based on the examples we could find on Kakuro websites and it is probably because they were intended for the enjoyment of human players and designed to do so.

However, the puzzle in Figure 15.5 presents an interesting case where the cut rule has no impact on the $W^+$ rating: it is 6 for both the original and the reduced puzzles. One may think that it is due to the fact that the remaining main part $P'_6$ still makes a long diagonal white stripe, but the examination of the resolution path shows that the whips[6] appear only after the upper part (above $C_8$) has been solved. Moreover, the two whips[6] appearing in the resolution paths of $P'_6$ and of the original puzzle lie completely in the lower part and are very similar.

We give only the path for the reduced puzzle $P'_6$ of Figure 15.6, where we track cells $C_7$, $C_8$ and $C_9$ (but we have programmed no special rule to focus search on them).

***** KakuRules 1.2 based on CSP-Rules 1.2, config: $W^+$ *****
**horizontal-magic-sectors: hr14c6 = 89, hr13c1 = 789, hr9c4 = 5789, hr8c4 = 124, hr6c6 = 12345, hr4c7 = 689, hr3c5 = 123457**
**vertical-magic-sectors: vr4c11 = 13, vr1c10 = 46789, vr8c8 = 89, vr4c7 = 2456789, vr14c6 = 13, vr3c5 = 12, vr11c2 = 89**
naked-singles ==> r16c6 = 1, r15c6 = 3, r13c4 = 7, r9c5 = 5, r6c10 = 4, **r3c10 = 7 (cell $C_7$)**, r3c7 = 5, vr1c7 = 59, r2c7 = 9, hr2c6 = 39, r2c8 = 3, vr7c5 = 25, r8c5 = 2, r8c7 = 4, r8c6 = 1, vr10c4 = 12347, hr16c4 = 15, r16c5 = 5, vr13c5 = 579, r14c5 = 7, r15c5 = 9, hr14c2 = 137
ctr-to-verti-sector  ==> r3c8 ≠ 1, r3c8 ≠ 2

naked-singles ==> r3c8 = 4, r3c6 = 3, vr2c6 = 389, vr1c8 = 349, r4c8 = 9
cell-to-horiz-ctr ==> hr4c4 ≠ 46 ; cell-to-horiz-ctr ==> hr5c4 ≠ 567
cell-to-horiz-ctr ==> hr4c4 ≠ 37 ; cell-to-horiz-ctr ==> hr11c3 ≠ 5679
ctr-to-horiz-sector ==> r11c5 ≠ 5, r11c6 ≠ 5, r11c7 ≠ 5 ; cell-to-horiz-ctr ==> hr2c9 ≠ 47
ctr-to-horiz-sector ==> r2c11 ≠ 4 ; cell-to-horiz-ctr ==> hr5c8 ≠ 257, hr5c8 ≠ 347
cell-to-horiz-ctr ==> hr5c4 ≠ 369, hr5c4 ≠ 378, hr5c4 ≠ 459 ; ctr-to-horiz-sector ==> r5c7 ≠ 5
cell-to-horiz-ctr ==> hr5c4 ≠ 468 ; ctr-to-horiz-sector ==> r5c7 ≠ 6
cell-to-horiz-ctr ==> hr5c8 ≠ 248 ; cell-to-horiz-ctr ==> hr10c5 ≠ 157, hr10c5 ≠ 247
ctr-to-horiz-sector ==> r10c7 ≠ 7, r10c6 ≠ 7 ; cell-to-horiz-ctr ==> hr10c5 ≠ 256
ctr-to-horiz-sector ==> r10c7 ≠ 5, r10c6 ≠ 5 ; cell-to-horiz-ctr ==> hr10c5 ≠ 346
ctr-to-horiz-sector ==> r10c7 ≠ 6, r10c6 ≠ 6
cell-to-horiz-ctr ==> hr12c1 ≠ 14567, hr12c1 ≠ 23567; cell-to-verti-ctr ==> vr2c9 ≠ 12359
ctr-to-verti-sector ==> r5c9 ≠ 9, r7c9 ≠ 9 ; cell-to-verti-ctr ==> vr1c11 ≠ 34
ctr-to-verti-sector ==> r2c11 ≠ 3 ; cell-to-horiz-ctr ==> hr2c9 ≠ 38
ctr-to-horiz-sector ==> r2c10 ≠ 8 ; cell-to-verti-ctr ==> vr11c3 ≠ 24569, vr11c3 ≠ 24578
cell-to-verti-ctr ==> vr5c8 ≠ 46 ; ctr-to-verti-sector ==> r7c8 ≠ 4, r7c8 ≠ 6
cell-to-verti-ctr ==> vr2c9 ≠ 13457 ; cell-to-verti-ctr ==> vr12c7 ≠ 467
cell-to-verti-ctr ==> vr12c8 ≠ 57 ; ctr-to-verti-sector ==> r13c8 ≠ 5, r13c8 ≠ 7
cell-to-horiz-ctr ==> hr13c5 ≠ 157, hr13c5 ≠ 256 ; ctr-to-horiz-sector ==> r13c7 ≠ 5, r13c6 ≠ 5
horiz-sector-to-ctr ==> hr10c5 ≠ 139 ; ctr-to-horiz-sector ==> r10c6 ≠ 9, r10c7 ≠ 9, r10c8 ≠ 9
naked-singles ==> r10c8 = 8, r10c7 = 2, r6c7 = 5, r9c8 = 9, hr10c5 = 238, r10c6 = 3
verti-sector-to-ctr ==> vr14c2 ≠ 35 ; ctr-to-verti-sector ==> r15c2 ≠ 5, r16c2 ≠ 5
biv-chain[2]: vr1c11{n16 n25} – r3c11{n1 n2} ==> r2c11 ≠ 2
cell-to-horiz-ctr ==> hr2c9 ≠ 29
naked-singles ==> hr2c9 = 56, **r2c10 = 6, r2c11 = 5**, r4c10 = 8, r4c9 = 6, r5c10 = 9, vr1c11 = 25, **r3c11 = 2**, r3c9 = 1
;;; Resolution state RS1

Although the value of $C_7$ (= r3c10) has been set long ago (almost at the beginning), those of r2c10, r2c11 and r3c11 are set only now and the small upper righmost graph that $C_7$ separates from the rest is solved only now. The solution has involved a bivalue-chain[2]. In the present case, if the focus had been set on the small subgraph, it could have been solved earlier in the path – but without changing its overall complexity.

ctr-to-horiz-sector ==> r5c9 ≠ 5, r5c9 ≠ 8, r5c9 ≠ 7 ; ctr-to-verti-sector ==> r7c9 ≠ 5
biv-chain[2]: hr5c8{n149 n239} – r5c11{n1 n3} ==> r5c9 ≠ 3
biv-chain[2]: vr12c8{n39 n48} – r14c8{n9 n8} ==> r13c8 ≠ 8
biv-chain[2]: vr14c2{n17 n26} – r16c2{n1 n2} ==> r15c2 ≠ 2
biv-chain[2]: hr16c1{n15 n24} – r16c2{n1 n2} ==> r16c3 ≠ 2
cell-to-verti-ctr ==> vr11c3 ≠ 23678
biv-chain[2]: hr11c3{n3789 n4689} – r11c4{n3 n4} ==> r11c6 ≠ 4, r11c5 ≠ 4
biv-chain[2]: r16c2{n1 n2} – vr14c2{n17 n26} ==> r15c2 ≠ 1
biv-chain[2]: r16c2{n1 n2} – hr16c1{n15 n24} ==> r16c3 ≠ 1
cell-to-verti-ctr ==> vr11c3 ≠ 12689, vr11c3 ≠ 13679
whip[2]: r14c7{n8 n9} – vr12c7{n458 .} ==> r13c7 ≠ 8
whip[2]: r14c7{n9 n8} – vr12c7{n359 .} ==> r13c7 ≠ 9

whip[2]: r14c8{n9 n8} – vr12c8{n39 .} ==> r13c8 ≠ 9
whip[2]: r14c7{n8 n9} – vr12c7{n458 .} ==> r15c7 ≠ 8
whip[2]: vr10c5{n49 n58} – r11c5{n6 .} ==> r12c5 ≠ 8
whip[2]: vr10c5{n58 n49} – r11c5{n6 .} ==> r12c5 ≠ 9
cell-to-horiz-ctr  ==> hr12c1 ≠ 12389
whip[2]: r12c2{n8 n9} – hr12c1{n23468 .} ==> r12c3 ≠ 8, r12c6 ≠ 8
whip[2]: r12c2{n9 n8} – hr12c1{n23459 .} ==> r12c3 ≠ 9, r12c6 ≠ 9
whip[2]: vr12c7{n179 n458} – r13c7{n1 .} ==> r15c7 ≠ 4
whip[2]: vr5c8{n19 n37} – r6c8{n1 .} ==> r7c8 ≠ 3
whip[2]: vr5c8{n19 n28} – r6c8{n1 .} ==> r7c8 ≠ 2
biv-chain[3]: r5c9{n2 n4} – vr2c9{n12368 n12467} – r6c9{n3 n2} ==> r7c9 ≠ 2
biv-chain[3]: vr2c9{n12368 n12467} – r6c9{n3 n2} – r5c9{n2 n4} ==> r7c9 ≠ 4
whip[3]: r6c9{n3 n2} – r5c9{n2 n4} – vr2c9{n12368 .} ==> r7c9 ≠ 3
horiz-sector-to-ctr  ==> hr7c5 ≠ 3689
whip[3]: vr6c6{n1234678 n1234579} – r11c6{n8 n7} – r9c6{n7 .} ==> r13c6 ≠ 9
cell-to-horiz-ctr  ==> hr13c5 ≠ 139
whip[3]: vr6c6{n1234678 n1234579} – r11c6{n8 n7} – r9c6{n7 .} ==> r7c6 ≠ 9
whip[3]: r7c9{n7 n8} – r7c8{n8 n9} – r7c7{n9 .} ==> hr7c5 ≠ 4589
whip[4]: r6n2{c8 c9} – r5c9{n2 n4} – hr5c8{n239 n149} – c11n3{r5 .} ==> r6c8 ≠ 3
cell-to-verti-ctr  ==> vr5c8 ≠ 37 ; ctr-to-verti-sector  ==> r7c8 ≠ 7
whip[4]: r7c8{n8 n9} – hr7c5{n5678 n2789} – r7c7{n6 n7} – r7c9{n7 .} ==> r7c6 ≠ 8
whip[4]: r4c6{n8 n9} – hr4c4{n28 n19} – c5n2{r4 r5} – hr5c4{n189 .} ==> r5c6 ≠ 8
naked-singles ==> r5c6 = 9, r4c6 = 8, hr4c4 = 28, r4c5 = 2, r5c5 = 1, hr5c4 = 189, r5c7 = 8, r9c7 = 7, r9c6 = 8, vr6c6 = 1234678
horiz-sector-to-ctr  ==> hr7c5 ≠ 5678 ; horiz-sector-to-ctr  ==> hr13c5 ≠ 148
ctr-to-horiz-sector  ==> r13c7 ≠ 1 ; horiz-sector-to-ctr  ==> hr13c5 ≠ 238
biv-chain[2]: hr11c3{n3789 n4689} – r11c6{n7 n6} ==> r11c7 ≠ 6
naked-singles ==> r11c7 = 9, **r7c7 = 6 (cell C$_8$)**, hr7c5 = 4679, r7c8 = 9, r7c9 = 7, r7c6 = 4, vr2c9 = 12467, r6c9 = 2, r6c8 = 1, r6c11 = 3, r5c11 = 1, r5c9 = 4, hr5c8 = 149, vr5c8 = 19
;;; Resolution state RS2

It is worth making a second pause here. As shown by the part of the resolution path upto RS$_2$, setting the value of cell C$_8$ has involved the two parts of the graph separated by C$_8$. The upper of the two parts is now completely solved.

cell-to-verti-ctr  ==> vr10c5 ≠ 49 ; ctr-to-verti-sector  ==> r12c5 ≠ 4
biv-chain[2]: hr11c3{n3789 n4689} – r11c6{n7 n6} ==> r11c5 ≠ 6
biv-chain[2]: vr10c5{n58 n67} – r11c5{n8 n7} ==> r12c5 ≠ 7
cell-to-horiz-ctr  ==> hr12c1 ≠ 12479 ; cell-to-horiz-ctr  ==> hr12c1 ≠ 13478
whip[2]: hr11c3{n4689 n3789} – r11c6{n6 .} ==> r11c5 ≠ 7
naked-singles ==> r11c5 = 8, vr10c5 = 58, r12c5 = 5
whip[2]: hr13c5{n247 n346} – r13c6{n2 .} ==> r13c7 ≠ 6
whip[3]: r14c3{n1 n3} – vr11c3{n14579 n13589} – r12c3{n7 .} ==> r15c3 ≠ 1
whip[4]: r13c7{n7 n3} – r13c8{n3 n4} – vr12c8{n39 n48} – r14n9{c8 .} ==> vr12c7 ≠ 368
whip[5]: r14n9{c7 c8} – vr12c8{n48 n39} – r13c8{n4 n3} – hr13c5{n247 n346} – r13c7{n2 .} ==> vr12c7 ≠ 278
whip[2]: vr12c7{n458 n269} – r13c7{n3 .} ==> r15c7 ≠ 2

whip[2]: r15c2{n7 n6} – r15c7{n6 .} ==> hr15c1 ≠ 234689
whip[2]: hr15c1{n134789 n235679} – r15c4{n1 .} ==> r15c3 ≠ 2
whip[2]: vr12c7{n458 n179} – r13c7{n2 .} ==> r15c7 ≠ 7
whip[3]: hr15c1{n235679 n134789} – r15c7{n5 n1} – r15c4{n1 .} ==> r15c3 ≠ 4
**whip[6]: hr12c1{n13568 n23459} – r12c6{n6 n2} – r12c3{n2 n3} – r14c3{n3 n1} – c4n1{r14 r15} – c4n2{r15 .} ==> r12c4 ≠ 4**
**whip[6]: r11c6{n6 n7} – hr11c3{n4689 n3789} – r11c4{n4 n3} – r14c4{n3 n1} – r12c4{n1 n2} – c6n2{r12 .} ==> r13c6 ≠ 6**
cell-to-horiz-ctr ==> hr13c5 ≠ 346
naked-singles ==> hr13c5 = 247, r13c8 = 4, vr12c8 = 48, r14c8 = 8, r14c7 = 9
cell-to-verti-ctr ==> vr12c7 ≠ 359 ; ctr-to-verti-sector ==> r15c7 ≠ 5
whip[3]: r15c2{n7 n6} – hr15c1{n134789 n235679} – r15c7{n1 .} ==> r15c3 ≠ 7
whip[2]: r16c3{n4 n5} – r15c3{n5 .} ==> vr11c3 ≠ 23579
whip[3]: r15c2{n6 n7} – hr15c1{n135689 n235679} – r15c7{n1 .} ==> r15c3 ≠ 6
biv-chain[2]: r15c3{n5 n8} – r13c3{n8 n9} ==> vr11c3 ≠ 14678
whip[3]: r15c7{n1 n6} – hr15c1{n134789 n135689} – r15c2{n7 .} ==> r15c4 ≠ 1
cell-to-horiz-ctr ==> hr15c1 ≠ 135689
biv-chain[2]: hr15c1{n134789 n235679} – r15c7{n1 n6} ==> r15c2 ≠ 6
naked-singles ==> r15c2 = 7, vr14c2 = 17, r16c2 = 1, hr16c1 = 15, r16c3 = 5, **r15c3 = 8 (cell C$_{10}$)**, r13c3 = 9, r13c2 = 8, r12c2 = 9, vr11c3 = 13589, hr15c1 = 134789, r15c4 = 4, r11c4 = 3, r14c4 = 1, r14c3 = 3, r12c3 = 1, r12c4 = 2, r15c7 = 1, vr12c7 = 179, r13c7 = 7, r13c6 = 2, hr12c1 = 12569, r12c6 = 6, **r11c6 = 7 (cell C$_9$)**, hr11c3 = 3789
Grid solved. Hardest step: Whip[6]

The separation potential of $C_9$ or $C_{10}$ has not been used in this resolution path. Whether there is another path in $W_6^+$ that would use these cells is an open question.

The small sub-puzzles $P_1$ to $P_6$ are easily solved once $P'_6$ is – provided that one uses the information obtained in the $P'_6$ solution. Considered as standalone puzzles, only $P_5$ and $P_6$ have a unique solution; $P_1$, $P_2$, $P_2+P_3$ and $P_4$ do not.

### 15.6.5. The cut rule is not subsumed by the coupling rules

Theorem 15.1 says that the original Kakuro problem is mathematically equivalent to our CSP re-formulation. However, as already noticed, this does not imply that our standard arsenal of resolution rules is enough to solve all the Kakuro puzzles, even with the coupling rules added. In this context, there naturally appears the question of whether the cut rule is subsumed by the coupling rules. It may seem that it should be so, but the first sub-graph $P_1$ of the example in Figure 15.5 provides an easy counter-example (Figure 15.7).

As the practical effect of the cut rule is to split the puzzle into several almost independent sub-puzzles, this situation does not present much interest from a theoretical point of view. It could therefore be assumed that a well-formed Kakuro puzzle is 2-connected. [Of course, from a new player's point of view, there may be some fun in finding such domains and easily solving apparently very large puzzles.

But, as detecting the cuts and then checking if they are valid is very easy, all the repetitive paper scratching it finally amounts to may also become very boring after some practice.]

| 𝒦 | 15 | 23 | 9 | | |
|---|----|----|---|---|---|
| 10 | 1  3 | 6 | 1  3 | | |
| 15 | 1  3<br>4 5 6<br>8 9 | 8 9 | 1  3<br>5 | | |
| 32 | 1  3<br>4 5 6<br>8 9 | 8 9 | 1  3<br>5 | | |
| | | | | | |

The question with this small sub-puzzle $P_1$ is: given the same information as used by the surface sums, i.e. given only the values of the *five* horizontal and vertical sums of the sectors completely inside $P_1$, can it be deduced that the sum of the first three cells in the fourth row is 22, *using only ECP, Singles and coupling rules*? But it cannot. All that can be deduced for the white cells inside $P_1$ is shown in the Figure.

**Figure 15.7.** *$P_1$, the first small part of puzzle of Figure 15.5*

# 16. Topological and geometric constraints: map colouring and path finding

In this chapter, we consider two kinds of Constraint Satisfaction Problems with constraints that can be considered as topological or geometrical in a broad sense of these words:

– the Map colouring problem is the simplest CSP of all those we shall study in this book; its constraints are obvious transcriptions of neighbourhood relations and are thus purely topological;

– in the path finding problem of the Numbrix® or Hidato® CSPs, where there is an underlying grid structure, one can choose whether they adopt only the obvious purely topological constraints derived from the relation of neighbourhood/ adjacency, thus implicitly forgetting much of the grid structure, or whether they rely on the notion of distance between two cells and they adopt a larger set of constraints derived from it; interestingly, it is easy to find concrete examples showing that the two views are not equivalent (i.e. they lead to different ratings).

## 16.1. Map colouring and the four-colour problem

Map colouring is interesting in the context of this book mainly because it will provide an example of a CSP in which, contrary to all our previous examples, there is no underlying grid structure at all (even distorted by "black cells" as in Kakuro, Numbrix® and Hidato®). This will illustrate our approach in its most basic form. [This section has no pretension of adding anything valuable to graph theory.]

### 16.1.1. The map colouring problem

Map colouring is a classical mathematical problem that became famous with the proof of the "four-colour theorem" in 1976. A map is defined as a partition of a plane (or a finite part of a plane) into a finite number of continuous domains with absolutely continuous boundaries called regions; contrary to countries in the real world, regions may not be made of separate parts (in this case, it is easy to find counterexamples to the theorem). Two regions are adjacent if they have a common boundary of positive length; two regions with only one point in common, or even with only isolated points in common, are not adjacent. A colouring is an assignement of a colour to each region (i.e. it is a function from the set of regions to the set of allowed colours) such that two adjacent regions have different colours. The theorem states that every map can be coloured with at most four colours.

The conditions of the theorem are strict. For regions on a 2D surface other than a plane, more than four colours can be required. Thus, if four are still enough on a sphere or a cylinder, six can be needed on a Möbius strip or a Klein bottle, and seven on a torus (there is a famous example of a partition of a torus into seven regions, each adjacent to all the others, and that require seven colours). Moreover, even on a plane, if the colours of some regions are pre-assigned, more than four colours can be required. Nevertheless, most of what we say below about resolution rules could easily be extended to such cases, with the appropriate number of colours.

We consider the adjacency constraints of map colouring as topological because they depend only on aspects of the "geometry" that are invariant by elastic transformations (and therefore independent of distances). Moreover, there are well-known results that associate the maximum number of colours required on a non-planar 2D surface of positive genus with its Euler characteristic or with its genus if it is orientable – both of these values being purely topological.

In the more formal view generally adopted in mathematical studies of the problem, a map is assimilated with a "planar graph" (a type of graph that can be given various purely graph-theoretic definitions, with no reference to geometry): a vertex is assigned to each region; there is an undirected edge between two vertices if and only if the corresponding regions are adjacent (there is only one edge even if the two regions are adjacent along several disjoint parts of their boundaries); conversely, it is easy to see that every planar graph originates in this way in a map (indeed, it can have many map representations). The colouring problem is then to assign a colour from a predefined set to each vertex in such a way that two vertices linked by an edge have different colours. The corresponding form of the theorem states that every planar graph is 4-colourable (i.e. that 4 colours are always enough).

Even though the theorem itself does not seem to have any practical applications in map production (real maps generally use more than four colours), it has become a topic of much debate, in relation with the way it was first proved: in 1976, the problem was reduced by Happel and Haken to a set of 1,936 particular cases (in 1996, this set was reduced to "only" 633); these cases had then to be tested individually by a computer program and the main objection from some mathematicians was that it was impossible to check the proof manually. Later, the whole proof was checked by the Coq automatic theorem checker, making it more "acceptable". In our view, the problem is not acceptability but the fact that it does not teach us anything, as explained in section 12.3.9.1.

From the standard graph-theoretic point of view, the minimum number p of colours necessary for colouring a map is the only problem and how many different such p-colourings are possible is more or less irrelevant. Given a map, it will generally have several possible 4-colourings if no region has a predefined colour; even if there is a definition of "Apollonian" graphs as the "uniquely 4-colourable" graphs (several of the equivalent geometric definitions could be considered as more basic), this uniqueness is meant only modulo a permutation of the colours.

The problem of colouring planar graphs can be extended to that of colouring graphs in general and any CSP could be considered as a graph colouring problem. Thus 9×9 Sudoku could be considered as a graph colouring problem with 81 regions (corresponding to the rc cells) and 9 colours (corresponding to the nine digits); it has a fixed, very specific and highly structured, but non-planar, network of edges, corresponding to the links between the cells.

In this section, we shall concentrate on the reverse view: map colouring will be seen as a CSP and we shall consider the colouring problem in the same way as we have done for Sudoku, Futoshiki or Kakuro: we shall deal with instances with sufficiently many "givens" to ensure that they have a unique solution with the allowed number of colours. For definiteness, we take this number to be four. As far as we know, this problem is not a standard one in graph theory.

### *16.1.2. Map colouring as a CSP*

Expressing the map colouring problem (or the equivalent planar-graph colouring problem) as a CSP is straightforward: each region/vertex is associated with a CSP variable (we call these generically X1, X2, …), with domain a predefined set of four colours – the same set for all the CSP variables, namely {Blue, Red, Yellow, Green} or {B, R, Y, G} for short. We use $X_1, X_2, \ldots$ and $c_1, c_2, \ldots$ for names of variables of respective sorts CSP-Variable and Colour.

Pre-labels are pairs < region, colour >. Labels are the same thing as pre-labels (each label has only one pre-label in its equivalence class). Two labels < $X_1, c_1$ > and < $X_2, c_2$ > are linked if and only if
- either: $X_1 = X_2$ and $c_1 \neq c_2$ (csp-links)
- or: $X_1 \neq X_2$, $X_1$ and $X_2$ are adjacent, and $c_1 = c_2$ (adjacency links).

There is no g-label and there are very limited possibilities for Subsets. As a result, map colouring does not seem to have much potential as a logic puzzle. Indeed, we have found only one website proposing a generator of map colouring instances ([Tatham www]; there are many map colouring games with hand made maps, but there seems to be no other generator). However, different global 2D topologies (i.e. maps on non-planar surfaces) may allow more subtle patterns.

### *16.1.3. A map colouring example and a whip-based solution*

Figure 16.1 shows a map with 30 regions and 12 givens. It is adapted from an example of the hardest level ("unreasonable") in the famous Tatham collection of generators of instances for various games [Tatham www]. As all the Sudoku and Futoshiki instances we have found on that website are relatively easy, even those classified as "extreme" or "unreasonable", we conjecture that this is also the case for the map examples – but we have no means of checking this. Anyway, the following example is the hardest one (with respect to the W rating) we could find in a set of 30 "unreasonable" ones we tested (some of which had upto 120 regions): it requires whips of length 7.

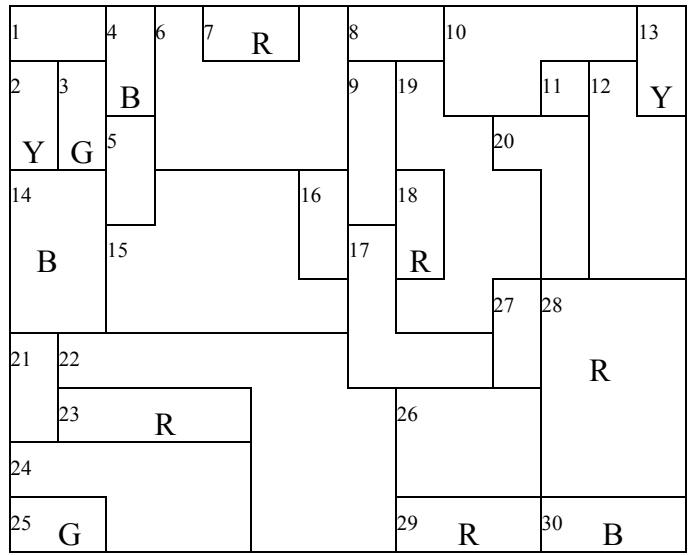

**Figure 16.1.** *A map with 30 regions and 12 givens (adapted from one on [Tatham www])*

The regions in the original puzzle have more complex shapes than in our Figure, which makes it harder and probably more interesting for a human solver; but, as mentioned at the end of section 4.1, this is typically one informal aspect that a formal resolution system can hardly tackle. To be more specific, let us mention briefly how this map is passed to the "solve" function of CSP-Rules:

(solve 4 30 ".YGB..R.....YB...R....R.G..RRB"
1 2 3 4 | 2 3 14 | 3 4 5 14 | 4 5 6 | 5 6 14 15 | 6 7 8 9 15 16 | 8 9 10 19 | 9 16 17 18 19 | 10 11 12 13 19 20 | 11 12 20 | 12 13 20 28 | 14 15 21 22 | 15 16 17 22 | 16 17 | 17 18 19 22 26 27 | 18 19 | 19 20 27 | 20 28 | 21 22 23 24 | 22 23 24 26 29 | 23 24 25 | 24 25 | 26 27 28 29 | 27 28 | 28 30 | 29 30 )

The first parameter (4) is the number of colours allowed; the second (30) is the number of CSP variables (i.e. of regions), the third is a string of length 30 (the same as the number of CSP variables) representing the series of givens (with a dot corresponding to no given, as in Sudoku); next comes a series of sequences of numbers separated by a vertical bar; each sequence between two bars represents the regions that are adjacent to its first element (as adjacency is a symmetric relation, only regions with a larger number than the first need be explicitly written): thus region X5 is linked to (and only to) X3, X4, X6, X14 and X15, as can be checked on Figure 16.1, and only the last three links need be written in the X5 sequence (the first two being written in the X3 and X4 sequences). As can be seen from this abstract graph representation, it provides no means of specifying the real shapes of regions or any other geometric detail. And it is not hard to imagine very different layouts for the same graph from the one in Figure 16.1.

\*\*\*\*\*  MapRules 1.2 based on CSP-Rules 1.2, config: W  \*\*\*\*\*
single ==> X1 = R
biv-chain[2]: X22{Y G} – X21{G Y} ==> X24 ≠ Y
single ==> X24 = B
whip[3]: X6{G Y} – X5{Y R} – X15{R .} ==> X16 ≠ G
whip[6]: X6{Y G} – X9{G B} – X19{B G} – X17{G Y} – X16{Y R} – X15{R .} ==> X8 ≠ Y
whip[6]: X6{G Y} – X5{Y R} – X15{R G} – X22{G Y} – X17{Y B} – X9{B .} ==> X8 ≠ G
whip[6]: X22{Y G} – X15{G R} – X5{R Y} – X6{Y G} – X9{G B} – X16{B .} ==> X17 ≠ Y
whip[3]: X17{B G} – X26{G Y} – X22{Y .} ==> X27 ≠ B
whip[3]: X17{B G} – X19{G Y} – X27{Y .} ==> X9 ≠ B
biv-chain[2]: X6{Y G} – X9{G Y} ==> X16 ≠ Y
**whip[7]: X6{G Y} – X9{Y G} – X17{G B} – X19{B Y} – X27{Y G} – X26{G Y} – X22{Y .} ==> X15 ≠ G**
biv-chain[2]: X5{Y R} – X15{R Y} ==> X6 ≠ Y
singles ==> X6 = G, X9 = Y
whip[2]: X17{G B} – X19{B .} ==> X27 ≠ G
single ==> X27 = Y
whip[2]: X17{G B} – X26{B .} ==> X22 ≠ G
singles to the end: X22 = Y, X15 = R, X5 = Y, X16 = B, X17 = G, X19 = B, X8 = R, X10 = G, X12 = B, X20 = Y, X11 = R, X26 = B, X21 = G

This resolution path is unchanged if braids are activated: both the W and the B ratings are 7.

### 16.2. Path finding: Numbrix® and Hidato®

Numbrix® and Hidato® are two closely related types of path-finding problems, invented respectively by Marylin vos Savant and Gyora Benayek. They are interesting in the context of this book mainly because they will lead us to introduce a kind of CSP variables we have not yet encountered (the Xn), they are based on a new kind of constraints and they allow an easy illustration of the consequences of different modelling choices for these constraints.

### 16.2.1. *Definition of Numbrix® and Hidato®*

We first give the broadest definitions before mentioning various (in our opinion unjustified) restrictions that are sometimes put on them.

Definition: A square grid of size n is given with two types of cells ("black" and "white", as in Kakuro); there are $N \leq n \times n$ white cells and some of them are filled in with a number not larger than N; the problem is to find a "continuous" path compatible with the clues, i.e. a sequence $(C_1, …, C_N)$ of white cells such that:

– for any $1 \leq p < N$, cell $C_{p+1}$ is "adjacent" to cell $C_p$,

– a clue indicates a forced passage of the path at a fixed time: more precisely, for any given number p in a white cell $D_p$, one must have $C_p = D_p$.

The difference between Numbrix® and Hidato® lies in the meaning of "adjacent": in Numbrix® two cells are adjacent if and only if they touch each other by one side in the same row or column; in Hidato®, they may also touch each other in diagonal, i.e. by a corner.

Remarks:

– as in Kakuro, the "real" grid (made of the white cells) can have any shape, provided only that it is path-connected (there is a "continuous" path between any two cells);

– the problem is to find a "continuous" path, but not necessarily to find it "in a continuous way", i.e. the successive steps do not have to be found in order;

– it is often supposed that the extremities of the solution path (i.e. numbers 1 and N) are given, but this is not necessary: one can deduce the value of N by counting the number of white cells; what could really change the problem is knowing *neither* the extreme values *nor* their positions (i.e. one would know only the length of the path, but the counting would start at some number k>1; of course, this would be a very artificial way of numbering the steps);

– many Numbrix® puzzles are proposed with no black cell;

– Hidato® is often presented as a King's Tour problem (an instance of the general Hamiltonian path problem in graph theory), due to the way the path must move from one place to the next, like a king in chess; however, there are so many more possibilities in this game that reducing it to this classical problem cannot be justified: the grid can have any size, its shape can be almost completely arbitrary, it can have inner holes, intermediate places are given, a well-formed puzzle is guaranteed to have a unique solution, …

– for each of these two problems, there are two equivalent ways of seeing it: either as finding a value for each white cell in the grid (the standard presentation) or as finding a place in the grid for each number in {1, …, N} (the dual presentation);

– as we shall see, and independently of the previous remark, there are also two natural ways of formalising their constraints.

### *16.2.2. Numbrix® and Hidato® as CSPs*

As the reader should now be used to our modelling principles and as their application to these two games is straightforward, we shall be a little sketchy, except for the definition of the constraints.

#### *16.2.2.1. Sorts, CSP-variables and labels*

We introduce the sorts Number, Row, Column and Cell with respective domains {1, …, N}, {r1, …, rn}, {c1, …, cn} and {(r, c) / (r, c) is white}.

As usual, we adopt a redundant set of CSP-variables, of two types, that naturally correspond to the dual ways of seeing the problem:

– for each Cell (r, c), we introduce a CSP-variable Xrc with domain Number: one must find a value for each white cell;

– for each Number n such that n is not in the set of clues, we introduce a CSP-variable Xn with domain Cell: one must find a place for each undecided Number. [It would be useless to introduce a CSP-variable for a decided value.]

We define a label as an (n, r, c) triplet, with the proper restrictions on n, r and c. As expected, (n, r, c) is the class of two pre-labels <Xrc, n> and <Xn, rc>.

#### *16.2.2.2. Constraints (topological vs geometric)*

In addition to the "strong" CSP constraints that automatically go with the CSP-variables, we introduce a unique (obviously symmetric) non-CSP constraint: "far". For each of the two CSPs, there are two ways of defining this constraint, somehow parallel to the Futoshiki example, although there is no transitivity involved in the present case. *The first approach corresponds to a purely topological view of the problem (based on adjacency relations), while the second is of a more geometric nature (based on distances). It is interesting that they are not equivalent* (they produce different ratings). Both are implemented in our Numbrix®/Hidato® solver based on CSP-Rules; which is chosen is passed as a parameter.

In the simplest and most obvious approach, two labels (n, r, c) and (n', r', c') are linked by constraint "far" if n' = n ± 1 but (r, c) and (r', c') are not adjacent (with the meaning of this word as specified above, depending on whether we speak of Numbrix® or Hidato®). The meaning of "far" as a contradiction should be clear: wherever n is in the grid, n±1 cannot be in a cell not adjacent to the cell where n is.

In the second approach, the distance between two cells (r, c) and (r', c') is first defined as the minimum number of steps necessary to pass from one to the other. Then, we say that two labels (n, r, c) and (n', r', c') are linked by constraint "far" if their distance is too large, i.e. if dist(r, c, r', c') > |n - n'|. In metaphoric terms, there is a contradiction between the two labels because one does not have enough time (measured by |n - n'|) for walking the distance from (r, c) to (r', c'). It is obvious that there are more links in this approach than in the first and it has therefore an *a priori* stronger resolution potential.

In all rigour, the distance should be computed as the length of the shortest path from (r, c) to (r', c') in the underlying graph whose vertices are the white cells and whose edges are the adjacency links specific to each game. One could even eliminate from this graph all the decided cells, which would make length grow with time and which could lead to the dynamical creation of links – but this remark is highly prospective, as we have found on the Web no instance of any of these problems that would justify doing so complicated things.

We have found convenient to use instead the following simple approximations that amount to "forgetting" the colours of the cells (and whether they are decided or not):

– for Numbrix®: dist(r, c, r', c') = |r - r'| + |c - c'|,
– for Hidato®: dist(r, c, r', c') = max(|r - r'|, |c - c'|).

Three questions immediately arise:

– can the topological and geometric approaches lead to different results? The next two sections will answer positively, even when the W rating is small: the Numbrix® puzzle in section 16.2.3 will become solvable by bivalue-chains[2] instead of whips[2], while the W rating of the Hidato® puzzle in section 16.2.4 will pass from 4 to 3. The part of the question that we shall leave unanswered (because we miss really hard instances) is: can any puzzle be solved (e.g. using whips,…, g-braids,…) with the geometric approach and not with the topological one?

– in the geometric approach, can the approximation (which leads to fewer links than the "real" distance and may thus reduce the resolution potential) lead to different results? We have no answer. But it seems unlikely in most instances, especially in Hidato®, unless very special patterns of black cells completely isolate a part of the white ones (e.g. by making long tubes).

– which approach is more realistic from a player's point of view? Undoubtedly the topological one for a beginner, but a more advanced player may want to use the geometric one together with the approximation. Using the real distance seems very unnatural as it requires to compute it each time it is needed or to remember it (in the normally rare cases) when it is not equal to its approximated value. An alternative is a restricted geometric approach, in which time and/or space differences considered in relation "far" are limited by some predefined value(s).

*16.2.2.3. Basic resolution theory*

We distinguish two types of Singles, corresponding to the two types of CSP variables: Naked-Single (a cell can only have one value) and Hidden-Single (a number can only be in one cell). The examples in the next sub-sections will illustrate the interplay between the two types of variables. They will also show that, in our approach, both of them are necessary. Somewhat arbitrarily, we give Naked-single a higher priority than Hidden-Single (the main purpose is to shorten the writing of the resolution paths, while keeping them distinct). As usual, the eliminations due to ECP will not be displayed (they may be different in the two approaches).

*16.2.2.4. Initial state*

In spite of our definition of the domains of the CSP variables, we must be careful with initialisation: if we merely started in a resolution state $RS_0$ with all the Numbers (or even with only all the undecided Numbers) as candidate-Numbers for all the white cells, there would be a huge number of candidates ($N^2$) and every resolution path would start with hundreds of trivial eliminations.

We shall therefore adopt the convention of starting with a resolution state $RS_1$ in which the most obvious whip[1] eliminations are already done. This is very far from enough to eliminate all the easy steps (in particular, this is not very efficient for instances with few clues), but this multitude of trivial eliminations is inherent in these types of puzzles and the vast majority of those proposed in newspapers are solvable by singles and whips[1].

We define $RS_1$ as the resolution state where all the givens are asserted as values and all and only the compatible labels are asserted as candidates, where compatible means non linked according to the second approach. This entails that the initial state is the same in the two approaches. Notice that, even when we adopt the first approach, the passage from $RS_0$ to $RS_1$ does not hide the use of any new rules; it amounts to doing a lot of ECP and whip[1] eliminations. (We leave the details of the easy proof, by recursion on $|n - n'|$, as an exercise for the reader.) The difference between the two approaches can appear only with longer chains – but the first example will show that it can already appear with whips[2].

*16.2.2.5. Warnings about the forthcoming resolution paths*

The number of eliminations increases like the number of initial candidates, i.e. approximately like $N^2$, most of which are really boring. Even with small-sized grids (and small N) and with the above-defined initialisation, the full resolution paths are very long in most cases, mainly due to the presence of innumerable whips[1]. In all our resolution paths, we shall suppose that the reader is able to find the whips[1] by himself when necessary and we shall skip almost all of them.

The length of the paths is in part the result of our goal of finding the "simplest" solution, of the associated simplest-first strategy and of the absence in CSP-Rules of "heuristic" rules for focusing on some candidate. A human solver is very unlikely to follow a similar path; instead, he would concentrate e.g. on finding some value close to the already known ones (not caring too much about the length of his chains of reasoning); but in the process he would somehow have to justify (part of) the same eliminations.

*16.2.2.6. Subsets in Numbrix® and Hidato®*

Subsets are very simple patterns in Numbrix® and Hidato®. The two types of CSP-variables are transversal. Associated with them, there are two kinds of Subsets; for each integer p > 2:

– *Naked-Subset[p]*: given p different rc-cells and p different Numbers such that each of these cells contains no other candidate-Number as these p Numbers (together with non-degeneracy conditions stated in chapter 8), then eliminate any of these candidate-Numbers from any other rc-cell. With respect to the general definitions of chapter 8, this corresponds to taking the p Xrc CSP-variables corresponding to the p rc-cells as the CSP-variables of the Subset and the p Xn CSP-variables (considered as constraints) corresponding to the p Numbers as its transversal sets.

– *Hidden-Subset[p]*: given p different Numbers and p different rc-cells such that these Numbers are candidates for no other rc-cell than these (together with non-degeneracy conditions stated in chapter 8), then eliminate any other candidate-Numbers from these rc-cells. With respect to the general definitions of chapter 8, this corresponds to taking the p Xn CSP-variables corresponding to the p Numbers as the CSP-variables of the Subset and the p Xrc CSP-variables (considered as constraints) corresponding to the p rc-cells as its transversal sets.

Remarks:

– in spite of the existence of a row-column grid structure, it plays strictly no role in the definition of Subsets;

– the distinction between "Naked" and "Hidden" corresponds to the standard presentation of these puzzles, on an rc-grid. But, considering the previous remark, these could be interchanged if one considers that the dual presentation, as a linear grid of n-cells of Undecided-Numbers, would be better;

– there is no limitation on the size of a Subset – other than the number of undecided cells and, from a practical point of view, the doubly exponential growth of complexity with size (be it for a human player or a computer program);

– in our Numbrix®/Hidato® solver, the implementation of Subsets is application-specific.

*16.2.3. A Numbrix® example*

The standard reference as the major source of Numbrix® puzzles is the "Parade" magazine [askmarilyn www], where a new one is published daily by the inventor of the game, with difficulty levels varying from "easy" to "expert". We had pre-selected the expert one from the 16[th] of October 2012 (Figure 16.2) because it is one of the hardest we had found there. (Here, as for most of the logic puzzles published in newspapers or journals, the notion of "hard" is very relative, as one has W = 2.) But the final reason for presenting it here is that the topological and geometric models lead to solutions with different hardest patterns, even for this easy puzzle.

The first steps of the resolution path are not very interesting; after Singles and whips[1], they lead to the "elaborated" puzzle displayed in the right part of Figure 16.2, from which we shall start.

|   |   |   |   |   |   |   |   |   |
|---|---|---|---|---|---|---|---|---|
|   |    |    |    |   |   |   |   |   |
|   | 25 | 22 |    | 16 |   | 8 | 1 |   |
|   | 24 |    |    |    |   |   | 2 |   |
|   |    |    |    |    |   |   |   |   |
|   | 36 |    |    |    |   |   | 54 |  |
|   |    |    |    |    |   |   |   |   |
|   | 42 |    |    |    |   |   | 58 |  |
|   | 41 | 78 |    | 68 |   | 70 | 59 | |
|   |    |    |    |    |   |   |   |   |
|   |    |    |    |    |   |   | 7 | 6 | 5 |
|   | 25 | 22 |    | 16 |   | 8 | 1 | 4 |
|   | 24 | 23 |    |    |   |   | 2 | 3 |
|   |    |    |    |    |   |   |   | 52 |
|   | 36 |    |    |    |   |   | 54 |   |
| 38 |   |    |    |    |   |   |    |   |
| 39 | 42 |  |    |    |   |   | 58 |   |
| 40 | 41 | 78 |  | 68 | 69 | 70 | 59 |  |
| 81 | 80 | 79 |  |    |   |   |    |   |

**Figure 16.2.** *A Numbrix® puzzle (clues of #20121016 expert, askmarylin) and its elaboration*

***** Numbrix-Rules 1.2 based on CSP-Rules 1.2, geometric-model, config: B *****
**biv-chain[2]: r8c9{n60 n62} – n64{r9c6 r9c8} ==> r9c8 ≠ 60**
naked-singles: r8c9 = 60, r9c8 = 62, r9c6 = 64, r9c7 = 63, r7c7 = 71, r9c5 = 65, r8c4 = 67, r7c3 = 77, r6c2 = 43, r6c3 = 44, r7c4 = 76, r5c1 = 37, r9c4 = 66, r9c9 = 61, r7c9 = 57, r6c9 = 56, r4c7 = 50, r4c8 = 51, r5c9 = 53, r6c8 = 55
whip[1]: n10{r1c6 .} ==> r5c7 ≠ 11 ; hidden-single: r5c7 = 49 ; more whips[1] ; singles to the end

It appears that the bivalue-chain[2] elimination r8c9 ≠ 60 is the key to the solution. It rests on the facts that, in the resolution state where it appears, cell r8c9 has only two possible values (60 and 62) and number 64 has only two possible places (r9c6 and r9c8); and this is true, after the long series of whips[1], in both the topological and geometric approaches. Moreover, the target is linked in both cases to the two ends of the chain. What makes this chain non valid in the topological model is the left-to-right link n62r8c9 – n64r9c6 because |64-62| ≠ 1; it is valid in the geometric approach because dist(r8c9, r9c6) = 4 > 2 = 64-62.

In the absence of this bivalue-chain[2] elimination, the resolution path for the topological model is longer and whips[2] are required:

***** Numbrix-Rules 1.2 based on CSP-Rules 1.2, topological-model, config: B *****
**biv-chain[2]: n53{r4c8 r5c9} – n51{r5c9 r4c8} ==> r4c8 ≠ 11**
**whip[2]: n51{r5c9 r4c8} – n53{r4c8 .} ==> r5c9 ≠ 55**
**whip[2]: n71{r7c7 r9c7} – n72{r7c6 .} ==> r7c7 ≠ 73**
**whip[2]: n71{r9c7 r7c7} – n72{r9c8 .} ==> r9c7 ≠ 73**
**whip[2]: n73{r6c6 r9c5} – n72{r7c6 .} ==> r9c6 ≠ 74**
**whip[2]: n74{r6c5 r9c4} – n73{r7c5 .} ==> r9c5 ≠ 75**
**whip[2]: n75{r6c4 r8c4} – n77{r8c4 .} ==> r9c4 ≠ 76**
**biv-chain[2]: n44{r6c3 r7c4} – n76{r7c4 r6c3} ==> r6c3 ≠ 32**
**biv-chain[2]: n44{r6c3 r7c4} – n76{r7c4 r6c3} ==> r6c3 ≠ 34**
**whip[2]: n44{r7c4 r6c3} – n76{r6c3 .} ==> r7c4 ≠ 74**
**whip[2]: n34{r4c1 r5c4} – n32{r5c4 .} ==> r6c4 ≠ 33**
**whip[2]: r6c4{n75 n45} – r7c5{n45 .} ==> r9c4 ≠ 74**
singles: r9c4 = 66, r9c5 = 65, r9c6 = 64, r9c7 = 63, r7c7 = 71, r9c8 = 62, r8c9 = 60, r9c9 = 61, r8c4 = 67, r7c3 = 77, r6c2 = 43, r6c3 = 44, r7c4 = 76, r5c1 = 37, r7c9 = 57, r6c9 = 56, r4c7 = 50, r4c8 = 51, r5c9 = 53, r6c8 = 55, r5c7 = 49
**whip[2]: n11{r3c5 r2c6} – n9{r2c6 .} ==> r1c6 ≠ 10**
singles: r3c6 = 10, r3c7 = 9, r1c6 = 18, r1c5 = 19, r1c4 = 20, r2c6 = 17
**whip[2]: n31{r3c1 r4c4} – n33{r4c4 .} ==> r3c4 ≠ 32**
whips[1] and singles to the end

### 16.2.4. Three Hidato® puzzles created by P. Mebane

The standard reference as the major source of Hidato® puzzles is the Smithsonian magazine [Smithsonian www]. However, here again, these puzzles are relatively easy (all those we have tested among those considered there as being at the hardest level could be solved by whips of maximum length 2, even in the topological model). We have found much harder instances in [Mebane 2012].

#### 16.2.4.1. First Hidato® example

An 8×8 puzzle with a very special pattern of black cells is reproduced in Figure 16.3. It is announced as the hardest in the Mebane collection, but we have seen that "hard" may have many meanings, depending on one's goals and on the CSP under consideration: with W = 3 (in both the topological and the geometric models), it would be considered as simple in Sudoku; however, what makes it hard here is the number of eliminations necessary at its hardest level $W_3$.

Notice that neither the starting point (Number 1) not the end of the path (Number 56, the number of white cells) are given; this is the first reason why we have chosen to present it here (the second being the small size of the grid). As before, we do not display the whips[1] in the following resolution paths.

| | | | 51 | | |
|---|---|---|---|---|---|
| | 15 | | | | |
| | | 31 | | | |
| | | | | 30 | 28 |
| 38 | | 12 | | | |
| | | | 6 | | |
| | | | | 20 | |
| | | 8 | | | |

| | 34 | 33 | 52 | 51 | 54 | 55 | 56 |
|---|---|---|---|---|---|---|---|
| 35 | | 15 | 32 | 53 | 50 | 25 | 26 |
| 36 | 14 | | 16 | 31 | 24 | 49 | 27 |
| 37 | 13 | 17 | | 23 | 30 | 48 | 28 |
| 38 | 39 | 12 | 18 | | 22 | 29 | 47 |
| 1 | 11 | 40 | 6 | 19 | | 21 | 46 |
| 2 | 10 | 5 | 41 | 7 | 20 | | 45 |
| 3 | 4 | 9 | 8 | 42 | 43 | 44 | |

***Figure 16.3.*** *A Hidato® puzzle and its solution (clues of # III.10, [Mebane 2012])*

\*\*\*\*\*  Hidato-Rules 1.2 based on CSP-Rules 1.2, topological-model, config: B   \*\*\*\*\*
biv-chain[2]: n16{r3c2 r3c4} – n14{r3c4 r3c2} ==> r3c2 ≠ 1, 2, 3, 34, 35, 36, 40, 41, 42, 43, 44, 45, 46, 47, 55, 56
whip[2]: n14{r3c4 r3c2} – n16{r3c2 .} ==> r3c4 ≠ 1, 2, 3, 24
whips[2]: n24{r3c6 r5c4} – n22{r5c4 .} ==> r4c3 ≠ 23
whip[2]: n14{r3c4 r3c2} – n16{r3c2 .} ==> r3c4 ≠ 32
whip[2]: n34{r5c4 r4c3} – n13{r4c3 .} ==> r4c2 ≠ 35
whip[2]: n14{r3c4 r3c2} – n16{r3c2 .} ==> r3c4 ≠ 33, 34
whip[2]: n32{r3c6 r2c5} – n34{r2c5 .} ==> r2c4 ≠ 33
whip[2]: n14{r3c4 r3c2} – n16{r3c2 .} ==> r3c4 ≠ 35
whip[2]: n14{r3c4 r3c2} – n16{r3c2 .} ==> r3c4 ≠ 41, 42, 43, 44, 45, 46, 47, 48, 49, 53
whip[2]: n55{r5c4 r4c3} – n13{r4c3 .} ==> r4c2 ≠ 56
whip[2]: n14{r3c4 r3c2} – n16{r3c2 .} ==> r3c4 ≠ 54, 55, 56
whip[2]:  n24{r3c6 r7c4} – n22{r7c4 .} ==> r8c3 ≠ 23, r7c3 ≠ 23
biv-chain[3]: n13{r4c2 r4c3} – n17{r4c3 r4c5} – r3c2{n16 n14} ==> r3c4 ≠ 14
naked-singles: r3c2 = 14, r3c4 = 16
whip[3]: n33{r5c4 r4c5} – n17{r4c5 r4c3} – n18{r5c6 .} ==> r5c4 ≠ 34
whip[3]: n35{r5c2 r5c4} – n18{r5c4 r5c6} – n17{r4c3 .} ==> r4c5 ≠ 34
biv-chain[3]: n32{r2c4 r4c5} – n17{r4c5 r4c3} – n18{r5c6 r5c4} ==> r5c4 ≠ 33
biv-chain[3]: n34{r1c2 r6c5} – n19{r6c5 r6c7} – n18{r5c4 r5c6} ==> r5c6 ≠ 33
naked-singles: r1c3 = 33, r1c2 = 34, r2c1 = 35, r3c1 = 36, r2c4 = 32
whips[2]: n13{r4c2 r4c3} – n46{r4c3 .} ==> r4c2 ≠ 45;   n56{r2c6 r2c5} – n54{r2c5 .} ==> r1c4 ≠ 55
    n55{r2c6 r2c5} – n53{r2c5 .} ==> r1c4 ≠ 54;   n54{r2c6 r2c5} – n52{r2c5 .} ==> r1c4 ≠ 53
    n50{r2c6 r2c5} – n48{r2c5 .} ==> r1c4 ≠ 49;   n49{r2c6 r2c5} – n47{r2c5 .} ==> r1c4 ≠ 48
    n48{r2c6 r2c5} – n46{r2c5 .} ==> r1c4 ≠ 47;   n47{r2c6 r2c5} – n45{r2c5 .} ==> r1c4 ≠ 46
    n46{r2c6 r2c5} – n44{r2c5 .} ==> r1c4 ≠ 45
whip[3]: n37{r4c1 r4c2} – n43{r4c2 r5c2} – n45{r5c2 .} ==> r4c1 ≠ 44, 43
whip[3]: n13{r4c2 r4c3} – n43{r4c3 r5c2} – n45{r5c2 .} ==> r4c2 ≠ 44
whip[3]: n37{r4c1 r4c2} – n43{r4c2 r5c2} – n41{r5c2 .} ==> r4c1 ≠ 42
whip[3]: n13{r4c2 r4c3} – n44{r4c3 r5c2} – n42{r5c2 .} ==> r4c2 ≠ 43
whip[3]: n42{r4c2 r5c4} – n18{r5c4 r5c6} – n17{r4c3 .} ==> r4c5 ≠ 43
whip[3]: n43{r5c2 r6c5} – n19{r6c5 r6c7} – n18{r5c4 .} ==> r5c6 ≠ 44
whip[3]: n44{r5c7 r5c4} – n18{r5c4 r5c6} – n17{r4c3 .} ==> r4c5 ≠ 45

whip[3]: n37{r4c1 r4c2} – n42{r4c2 r5c2} – n40{r5c2 .} ==> r4c1 ≠ 41
whip[3]: n13{r4c2 r4c3} – n43{r4c3 r5c2} – n41{r5c2 .} ==> r4c2 ≠ 42
whip[3]: n41{r4c2 r5c4} – n18{r5c4 r5c6} – n17{r4c3 .} ==> r4c5 ≠ 42
whip[3]: n42{r5c2 r6c5} – n19{r6c5 r6c7} – n18{r5c4 .} ==> r5c6 ≠ 43
whip[3]: n43{r6c1 r5c4} – n18{r5c4 r5c6} – n17{r4c3 .} ==> r4c5 ≠ 44
whip[3]: n44{r5c2 r6c5} – n19{r6c5 r6c7} – n18{r5c4 .} ==> r5c6 ≠ 45
whip[3]: n45{r6c2 r5c4} – n18{r5c4 r5c6} – n17{r4c3 .} ==> r4c5 ≠ 46
whip[3]: n54{r3c7 r3c6} – r1c7{n53 n1} – r1c8{n1 .} ==> r4c5 ≠ 55
whip[3]: n54{r4c7 r4c5} – r1c7{n53 n1} – r1c8{n1 .} ==> r5c4 ≠ 55
whip[3]: n55{r5c7 r5c6} – n18{r5c6 r5c4} – n19{r6c7 .} ==> r6c5 ≠ 56
whip[3]: n54{r1c6 r4c7} – r1c7{n53 n1} – r1c8{n1 .} ==> r5c7 ≠ 55, r5c8 ≠ 55
whip[3]: r1c8{n54 n1} – r1c7{n1 n53} – n54{r4c7 .} ==> r4c7 ≠ 55, r5c6 ≠ 55
whip[2]: n53{r3c7 r3c6} – n55{r3c6 .} ==> r4c5 ≠ 54
whip[3]: r1c8{n54 n1} – n2{r8c7 r2c7} – n3{r8c7 .} ==> r3c6 ≠ 55
whip[3]: n45{r6c2 r6c5} – n19{r6c5 r6c7} – n18{r5c4 .} ==> r5c6 ≠ 46
whip[3]: n46{r6c3 r5c4} – n47{r5c8 r4c5} – n17{r4c5 .} ==> r4c3 ≠ 45
whip[3]: n46{r6c3 r5c4} – n18{r5c4 r5c6} – n17{r4c3 .} ==> r4c5 ≠ 47
biv-chain[3]: n18{r5c4 r5c6} – n17{r4c3 r4c5} – n48{r4c5 r4c7} ==> r5c4 ≠ 47
whip[3]: n24{r3c7 r3c6} – n26{r3c6 r2c6} – n50{r2c6 .} ==> r2c5 ≠ 25
whip[3]: r1c8{n54 n1} – r1c7{n1 n53} – r1c4{n52 .} ==> r3c7 ≠ 55, r3c8 ≠ 55
whip[3]: n47{r5c7 r5c6} – n18{r5c6 r5c4} – n19{r6c7 .} ==> r6c5 ≠ 46
;;; more whips[1] and Singles
biv-chain[2]: n5{r6c3 r7c3} – n40{r7c3 r6c3} ==> r6c3 ≠ 1, 3, 4, 10, 11
biv-chain[2]: n39{r5c2 r6c2} – n11{r6c2 r5c2} ==> r5c2 ≠ 4
biv-chain[2]: r8c1{n1 n3} – r6c1{n3 n1} ==> r6c2 ≠ 1, r7c1 ≠ 1, r7c2 ≠ 1, r7c3 ≠ 1, r8c2 ≠ 1
whip[1]: r8c2{n3 .} ==> r6c1 ≠ 3
singles to the end

*16.2.4.2. Second Hidato® example: non equivalence of the topological and geometric models*

Puzzle # III.7 in [Melbane 2012] is interesting for four reasons:

– as before, the places of the first and the last values (36) are not given;

– it has only two givens and uniqueness of the solution is ensured by the very constrained pattern of black cells;

– it is a hard instance (relatively to all those we have seen) in a very compact design;

– above all, ***its W (or B) rating is 4 or 3, depending on whether one adopts the topological or the geometric model***.

This puzzle is given in Figure 16.4. In order to save space, whips[2] will not be written in the resolution paths. These should therefore be considered as giving only the main lines of a proof with blanks that must be filled by whips[1] and whips[2], when a single or a t-candidate in a longer whip must be justified.

|   |   |   |   |   |   | 2 |
|---|---|---|---|---|---|---|
|   |   |   |   |   |   |   |
|   |   |   |   |   |   |   |
|   |   |   |   |   |   |   |
|   |   |   |   |   |   |   |
|   |   |   |   |   |   |   |
|   |   |   |   |   |   |   |
| 22 |   |   |   |   |   |   |

|    |    | 35 | 36 |    | 7  | 1  | 2  |
|----|----|----|----|----|----|----|----|
| 33 | 34 |    |    | 8  | 6  |    | 3  |
| 32 |    | 30 |    | 9  |    | 5  | 4  |
|    | 31 | 29 |    | 10 |    |    |    |
|    |    |    | 28 |    | 11 | 13 |    |
| 24 | 25 |    | 27 |    | 12 |    | 14 |
| 23 |    | 26 | 19 |    |    | 16 | 15 |
| 22 | 21 | 20 |    | 18 | 17 |    |    |

***Figure 16.4.** A Hidato® puzzle and its solution (clues of # III.7, [Mebane 2012])*

Let us start with the geometric model:

```
*****  Hidato-Rules 1.2 based on CSP-Rules 1.2, geometric-model, config: B   *****
;;; lots of whips[1], biv-chains[2] and whips[2]
whip[3]: r3c8{n36 n4} – n6{r1c4 r1c6} – n5{r3c5 .} ==> r2c6 ≠ 36
whip[3]: r3c7{n5 n35} – n34{r1c3 r2c6} – n4{r2c6 .} ==> r5c4 ≠ 7
whip[3]: r3c7{n5 n35} – n34{r1c3 r2c6} – n4{r2c6 .} ==> r4c5 ≠ 6
whip[3]: n6{r1c4 r3c5} – r3c7{n5 n35} – r2c1{n36 .} ==> r8c6 ≠ 11, r8c5 ≠ 11, r8c3 ≠ 11, r7c7 ≠ 10, r7c4 ≠ 10, r7c3 ≠ 10
whip[2]: n8{r1c3 r5c4} – n10{r1c3 .} ==> r6c4 ≠ 9
whip[3]: n5{r3c5 r2c5} – n7{r2c5 r2c6} – n4{r2c6 .} ==> r1c6 ≠ 6
whip[3]: r3c7{n5 n35} – n34{r1c3 r2c6} – n4{r2c6 .} ==> r1c6 ≠ 5, r3c5 ≠ 5
whip[3]: n6{r2c6 r3c5} – r3c7{n5 n35} – n34{r4c2 .} ==> r2c6 ≠ 7
whip[3]: n8{r1c4 r2c2} – n11{r7c8 r3c1} – n9{r3c5 .} ==> r2c1 ≠ 10
whip[3]: r3c8{n4 n36} – n35{r1c3 r3c7} – n34{r4c2 .} ==> r2c6 ≠ 4
whip[3]: r8c2{n21 n23} – r6c1{n24 n36} – n35{r6c4 .} ==> r6c2 ≠ 20
whip[3]: n24{r8c3 r7c3} – n20{r7c3 r6c1} – n18{r6c4 .} ==> r8c3 ≠ 25
whip[3]: r7c1{n21 n23} – n25{r6c4 r6c2} – n26{r8c5 .} ==> r7c3 ≠ 20
whip[2]: n20{r8c3 r6c1} – n19{r7c3 .} ==> r8c3 ≠ 18
whip[3]: n6{r1c4 r3c5} – r3c7{n5 n35} – r2c1{n30 .} ==> r6c6 ≠ 9
whip[3]: r7c1{n23 n21} – n19{r7c4 r6c2} – n18{r6c4 .} ==> r7c3 ≠ 24
whip[3]: n35{r6c2 r7c3} – n19{r7c3 r7c4} – n25{r7c4 .} ==> r6c2 ≠ 36
whip[3]: n34{r6c4 r7c3} – n19{r7c3 r7c4} – n25{r7c4 .} ==> r6c2 ≠ 35
whip[3]: r6c1{n24 n20} – n18{r6c4 r7c3} – n17{r8c6 .} ==> r6c4 ≠ 26
whip[3]: r3c7{n5 n35} – n34{r1c3 r2c6} – n33{r2c1 .} ==> r3c5 ≠ 6
whip[3]: n6{r2c6 r1c4} – r3c7{n5 n35} – n34{r4c2 .} ==> r2c6 ≠ 8, r2c6 ≠ 9
whip[2]: n7{r3c5 r2c5} – n9{r3c1 .} ==> r1c6 ≠ 8
whip[3]: n6{r2c6 r1c4} – r3c7{n5 n35} – n34{r4c2 .} ==> r2c6 ≠ 10
whip[2]: n8{r3c5 r2c5} – n10{r3c1 .} ==> r1c6 ≠ 9
whip[3]: n10{r3c1 r2c5} – n13{r6c6 r2c6} – n12{r3c1 .} ==> r1c6 ≠ 11
whip[3]: n13{r4c2 r2c5} – n11{r2c1 r3c5} – n14{r3c5 .} ==> r2c6 ≠ 12
whip[3]: n6{r2c6 r1c4} – r3c7{n5 n35} – n34{r4c2 .} ==> r2c6 ≠ 11
whip[2]: n9{r3c1 r2c5} – n11{r3c1 .} ==> r1c6 ≠ 10
```

whip[3]: n6{r2c6 r1c4} – r3c7{n5 n35} – n34{r4c2 .} ==> r2c6 ≠ 13
whip[2]: n11{r3c1 r2c5} – n13{r3c1 .} ==> r1c6 ≠ 12
whip[3]: n6{r2c6 r1c4} – r3c7{n5 n35} – n34{r4c2 .} ==> r2c6 ≠ 31
whip[3]: n33{r3c1 r4c5} – n31{r3c1 r2c5} – n30{r3c3 .} ==> r3c5 ≠ 32
whip[3]: n32{r4c3 r5c6} – n30{r4c2 r3c5} – n29{r4c3 .} ==> r4c5 ≠ 31
whip[3]: n9{r1c3 r5c6} – n11{r4c3 r5c7} – n12{r4c5 .} ==> r6c6 ≠ 10
whip[3]: n10{r1c3 r5c6} – n12{r4c3 r6c6} – n13{r4c5 .} ==> r5c7 ≠ 11
whip[3]: r7c8{n29 n15} – r3c1{n13 n36} – r2c1{n36 .} ==> r1c4 ≠ 33
whip[2]: n31{r2c5 r2c2} – n33{r2c2 .} ==> r1c3 ≠ 32
whip[3]: n9{r1c3 r5c6} – n11{r1c3 r6c6} – n12{r4c5 .} ==> r5c7 ≠ 10
whip[2]: n8{r1c4 r4c5} – n10{r1c3 .} ==> r5c6 ≠ 9
whip[2]: n7{r1c3 r3c5} – n9{r2c1 .} ==> r4c5 ≠ 8
whip[3]: n10{r1c3 r5c6} – n12{r4c3 r5c7} – n13{r4c5 .} ==> r6c6 ≠ 11
whip[3]: n32{r4c3 r1c4} – n6{r1c4 r2c6} – n8{r2c2 .} ==> r2c5 ≠ 31
whip[3]: n11{r1c3 r5c6} – n13{r4c2 r6c6} – n14{r6c8 .} ==> r5c7 ≠ 12
whip[3]: n8{r2c2 r1c4} – n7{r1c6 r1c3} – n6{r2c6 .} ==> r2c5 ≠ 9
whip[3]: n9{r3c1 r1c3} – n8{r2c5 r2c2} – n7{r2c5 .} ==> r1c4 ≠ 10
whip[3]: r2c6{n6 n34} – r7c8{n36 n29} – r6c6{n29 .} ==> r2c1 ≠ 9
whip[3]: r2c6{n6 n34} – r7c8{n36 n29} – r6c6{n29 .} ==> r3c1 ≠ 9
whip[3]: n10{r1c3 r4c2} – n12{r6c6 r3c3} – n9{r3c3 .} ==> r4c3 ≠ 11
whip[3]: n9{r3c5 r3c3} – n11{r5c6 r3c1} – n14{r6c8 .} ==> r4c2 ≠ 10
whip[3]: n10{r4c3 r2c2} – n8{r2c5 r1c4} – n12{r6c6 .} ==> r1c3 ≠ 11
singles, whips[1], biv-chains[2] and whips[2] to the end

Let us now consider the topological model:

***** Hidato-Rules 1.2 based on CSP-Rules 1.2, topological-model, config: B *****
;;; lots of whips[1], biv-chains[2] and whips[2]
whip[3]: n6{r2c6 r3c5} – r3c7{n5 n35} – n34{r4c2 .} ==> r2c6 ≠ 7
whip[3]: n9{r2c5 r3c1} – n8{r3c5 r2c2} – n11{r2c2 .} ==> r2c1 ≠ 10
whip[3]: r3c8{n4 n36} – n35{r1c3 r3c7} – n34{r4c2 .} ==> r2c6 ≠ 4
whip[3]: r8c2{n21 n23} – r6c1{n24 n36} – n35{r6c4 .} ==> r6c2 ≠ 20
whip[3]: r8c2{n23 n21} – r6c1{n20 n36} – n35{r6c4 .} ==> r6c2 ≠ 24
**whip[4]: n36{r3c1 r2c5} – r3c7{n35 n5} – n6{r1c4 r2c6} – n34{r2c6 .} ==> r1c6 ≠ 35**
**whip[4]: r3c7{n35 n5} – n6{r1c4 r2c6} – n33{r2c6 r2c5} – n35{r2c5 .} ==> r1c6 ≠ 34**
**whip[4]: n7{r3c5 r2c5} – n5{r2c5 r3c7} – n6{r1c4 r2c6} – n9{r2c6 .} ==> r1c6 ≠ 8**
**whip[4]: n16{r3c3 r3c5} – n6{r3c5 r1c4} – r3c7{n5 n35} – n34{r4c2 .} ==> r2c6 ≠ 15**
**whip[4]: n15{r3c1 r2c5} – n5{r2c5 r3c7} – n6{r3c5 r2c6} – n13{r2c6 .} ==> r1c6 ≠ 14**
**whip[4]: n28{r8c6 r3c5} – n6{r3c5 r1c4} – r3c7{n5 n35} – n34{r4c2 .} ==> r2c6 ≠ 29**
**whip[4]: n29{r3c1 r2c5} – n5{r2c5 r3c7} – n6{r3c5 r2c6} – n31{r2c6 .} ==> r1c6 ≠ 30**
**whip[4]: n34{r6c4 r7c3} – n36{r7c3 r6c1} – n20{r6c1 r8c3} – n24{r8c3 .} ==> r6c2 ≠ 35**
whip[3]: r6c1{n24 n20} – n19{r6c4 r6c2} – n18{r6c4 .} ==> r7c3 ≠ 25
whip[3]: n25{r8c3 r7c4} – n24{r6c1 r7c3} – n27{r7c3 .} ==> r8c3 ≠ 26
whip[3]: r6c1{n24 n20} – n19{r6c4 r6c2} – n18{r6c4 .} ==> r7c3 ≠ 24
whip[3]: n24{r8c3 r6c1} – n20{r6c1 r7c3} – n26{r7c3 .} ==> r8c3 ≠ 27
whip[3]: n25{r6c2 r7c4} – r6c1{n24 n20} – n19{r6c4 .} ==> r6c2 ≠ 36

whip[2]: r6c2{n19 n25} – n26{r8c5 .} ==> r7c3 ≠ 19
whip[3]: n19{r8c3 r7c4} – n20{r6c1 r7c3} – n17{r7c3 .} ==> r8c3 ≠ 18
whip[3]: r6c2{n19 n25} – n26{r8c5 r7c3} – n20{r7c3 .} ==> r6c4 ≠ 19
whip[3]: r6c2{n19 n25} – n26{r8c5 r7c3} – n20{r7c3 .} ==> r8c3 ≠ 19
whip[3]: r6c2{n25 n19} – n18{r6c4 r7c3} – n17{r8c6 .} ==> r6c4 ≠ 26
biv-chain[3]: n27{r6c4 r8c6} – n26{r7c3 r8c5} – r6c2{n25 n19} ==> r6c4 ≠ 18
whip[3]: n5{r2c5 r3c7} – n6{r3c5 r2c6} – n13{r2c6 .} ==> r2c5 ≠ 12, 32
whip[3]: n32{r3c1 r2c6} – n31{r3c1 r2c5} – n34{r2c5 .} ==> r1c6 ≠ 33
whip[3]: n31{r4c2 r2c5} – n33{r2c5 r3c5} – n30{r3c5 .} ==> r2c6 ≠ 32
whip[3]: n12{r3c1 r2c6} – n13{r6c6 r2c5} – n10{r2c5 .} ==> r1c6 ≠ 11
whip[3]: n13{r4c2 r2c5} – n11{r2c5 r3c5} – n14{r3c5 .} ==> r2c6 ≠ 12
whip[3]: r5c4{n28 n16} – r7c8{n15 n36} – n35{r6c6 .} ==> r6c8 ≠ 29
whip[3]: r5c4{n16 n28} – r7c8{n29 n36} – n35{r6c6 .} ==> r6c8 ≠ 15
whip[3]: n8{r1c4 r5c6} – n10{r5c6 r6c6} – n11{r4c5 .} ==> r5c7 ≠ 9
whip[3]: n9{r6c6 r5c6} – n11{r5c6 r5c7} – n12{r4c5 .} ==> r6c6 ≠ 10
whip[3]: n10{r1c3 r5c6} – n12{r5c6 r6c6} – n13{r4c5 .} ==> r5c7 ≠ 11
whip[3]: n8{r1c4 r5c6} – n10{r5c6 r5c7} – n11{r4c5 .} ==> r6c6 ≠ 9
whip[3]: n9{r1c3 r5c6} – n11{r5c6 r6c6} – n12{r4c5 .} ==> r5c7 ≠ 10
whip[3]: n10{r1c3 r5c6} – n12{r5c6 r5c7} – n13{r4c5 .} ==> r6c6 ≠ 11
whip[3]: n11{r1c3 r5c6} – n13{r5c6 r6c6} – n14{r6c8 .} ==> r5c7 ≠ 12
whip[3]: n12{r6c6 r5c6} – n14{r5c6 r5c7} – n15{r7c8 .} ==> r6c6 ≠ 13
whip[3]: n6{r2c6 r1c4} – r3c7{n5 n35} – n34{r4c2 .} ==> r2c6 ≠ 8
whip[3]: n6{r2c6 r1c4} – r3c7{n5 n35} – n34{r4c2 .} ==> r2c6 ≠ 9
whip[2]: n7{r1c3 r2c5} – n9{r2c5 .} ==> r3c5 ≠ 8
whip[3]: n8{r2c2 r1c4} – n7{r1c6 r1c3} – n6{r2c6 .} ==> r2c5 ≠ 9
whip[3]: n9{r3c1 r1c3} – n8{r2c5 r2c2} – n7{r2c5 .} ==> r1c4 ≠ 10
whip[3]: n10{r2c2 r2c6} – n6{r2c6 r1c4} – n5{r3c7 .} ==> r2c5 ≠ 11
whip[3]: n14{r3c3 r3c5} – n12{r3c5 r1c6} – n11{r3c1 .} ==> r2c6 ≠ 13
whip[3]: n11{r3c1 r2c6} – n10{r3c1 r2c5} – n13{r2c5 .} ==> r1c6 ≠ 12
whip[3]: n10{r2c2 r2c5} – n5{r2c5 r3c7} – n6{r1c4 .} ==> r2c6 ≠ 11
whip[3]: n13{r5c6 r5c7} – n12{r4c5 r6c6} – n11{r4c3 .} ==> r5c6 ≠ 14
whip[3]: n6{r2c6 r1c4} – r3c7{n5 n35} – n34{r4c2 .} ==> r2c6 ≠ 10
whip[2]: n8{r1c4 r2c5} – n10{r2c5 .} ==> r1c6 ≠ 9
whip[3]: n6{r2c6 r1c4} – r3c7{n5 n35} – n34{r4c2 .} ==> r2c6 ≠ 31
whip[3]: n6{r2c6 r1c4} – r3c7{n5 n35} – n34{r4c2 .} ==> r2c6 ≠ 33
whip[2]: n31{r3c1 r2c5} – n33{r2c5 .} ==> r1c6 ≠ 32
whip[3]: r3c7{n35 n5} – r1c6{n4 n7} – n8{r2c2 .} ==> r2c5 ≠ 36
whip[3]: n5{r2c5 r3c7} – r1c6{n4 n36} – n35{r2c1 .} ==> r2c5 ≠ 7
whip[3]: n8{r2c5 r2c2} – n7{r3c5 r1c3} – r2c6{n6 .} ==> r2c5 ≠ 34
whip[3]: n15{r4c3 r7c8} – n14{r4c2 r6c8} – n13{r4c2 .} ==> r4c3 ≠ 12
**whip[4]: n31{r2c2 r2c5} – n33{r2c5 r1c3} – n34{r6c6 r2c2} – n8{r2c2 .} ==> r1c4 ≠ 32**
**whip[4]: r2c6{n34 n6} – r1c4{n6 n9} – n10{r4c5 r1c3} – n11{r5c6 .} ==> r2c2 ≠ 35**
**whip[4]: n32{r2c1 r2c2} – n34{r2c2 r1c4} – n35{r6c8 r2c5} – n8{r2c5 .} ==> r1c3 ≠ 33**
**whip[4]: n33{r2c1 r2c5} – n8{r2c5 r2c2} – n7{r3c5 r1c3} – n6{r2c6 .} ==> r1c4 ≠ 34**
**whip[4]: r3c7{n35 n5} – r1c4{n6 n9} – n10{r4c5 r1c3} – n11{r5c6 .} ==> r2c2 ≠ 36**
**whip[4]: r3c7{n35 n5} – r1c4{n6 n9} – n8{r2c2 r2c5} – n35{r2c5 .} ==> r1c6 ≠ 36**
biv-chain[3]: n5{r2c5 r3c7} – r1c6{n4 n7} – n8{r2c2 r2c5} ==> r2c5 ≠ 33

whip[2]: n31{r3c1 r4c5} – n33{r4c5 .} ==> r3c5 ≠ 32
whip[2]: n30{r4c2 r5c6} – n32{r5c6 .} ==> r4c5 ≠ 31
biv-chain[3]: n5{r2c5 r3c7} – r1c6{n4 n7} – n8{r2c2 r2c5} ==> r2c5 ≠ 35
whip[2]: n33{r3c1 r4c5} – n35{r4c5 .} ==> r3c5 ≠ 34
whip[2]: n32{r4c2 r5c6} – n34{r5c6 .} ==> r4c5 ≠ 33
biv-chain[3]: n8{r2c2 r2c5} – n5{r2c5 r3c7} – r1c6{n4 n7} ==> r3c5 ≠ 7
whip[3]: n36{r3c1 r4c5} – n34{r4c5 r2c6} – n33{r3c1 .} ==> r3c5 ≠ 35
whip[3]: n35{r3c7 r5c6} – n33{r5c6 r3c5} – n32{r4c2 .} ==> r4c5 ≠ 34
**whip[4]: r2c6{n34 n6} – r1c3{n7 n10} – n35{r1c3 r3c1} – r3c5{n36 .} ==> r2c1 ≠ 34**
**whip[4]: n34{r2c2 r5c7} – r2c6{n34 n6} – r1c3{n7 n10} – r3c5{n9 .} ==> r6c8 ≠ 35**
;;; singles, whips[1], biv-chains[2] and whips[2] to the end

### 16.2.4.3. Third Hidato® example

The reasons for choosing our last Hidato® example (Figure 16.5) should be obvious: with grid size 5, it is remarkably compact but it has an unexpectedly hard resolution path (in both the topological and the geometric models, W = B = 8), in spite of having both ends (Numbers 1 and 19) given. We show only the path for the topological model. As, contrary to the previous examples, there are few whips[1] before the first Single, we display them all.

|   |   |   |   |   |     |    |    |    |    |    |
|---|---|---|---|---|-----|----|----|----|----|----|
|   |   |   |   |   |     | 3  | 4  |    | 6  | 7  |
|   |   |   |   |   |     |    | 2  | 5  |    | 8  |
|   |   | 19| 1 | 10|     | 18 | 19 | 1  | 10 | 9  |
|   |   |   |   |   |     | 17 |    | 14 | 11 |    |
|   |   |   |   |   |     | 16 | 15 |    | 13 | 12 |

**Figure 16.5.** *A Hidato® puzzle and its solution (clues of # III.4, [Mebane 2012])*

***** Hidato-Rules 1.2 based on CSP-Rules 1.2, topological-model, config: B *****
whip[4]: n8{r1c2 r2c2} – n9{r2c5 r2c3} – n6{r2c3 r1c1} – n5{r1c4 .} ==> r1c2 ≠ 7
whip[3]: n7{r1c4 r2c2} – n8{r5c5 r1c2} – n5{r1c2 .} ==> r1c1 ≠ 6
whip[4]: n12{r1c2 r2c2} – n11{r4c4 r2c3} – n14{r2c3 r1c1} – n15{r5c5 .} ==> r1c2 ≠ 13
whip[3]: n13{r1c4 r2c2} – n12{r2c5 r1c2} – n15{r1c2 .} ==> r1c1 ≠ 14
whip[4]: n17{r5c4 r4c4} – n18{r2c2 r4c3} – n15{r4c3 r5c5} – n14{r5c2 .} ==> r5c4 ≠ 16
whip[3]: n16{r5c5 r4c4} – n17{r5c2 r5c4} – n14{r5c4 .} ==> r5c5 ≠ 15
whip[5]: n3{r3c5 r4c4} – n2{r4c4 r4c3} – n5{r4c3 r2c5} – n9{r2c5 r2c3} – n11{r2c3 .} ==> r3c5 ≠ 4
whip[5]: n17{r1c1 r4c4} – n18{r2c2 r4c3} – n15{r4c3 r2c5} – n9{r2c5 r2c3} – n11{r2c3 .} ==> r3c5 ≠ 16
whip[4]: n18{r2c2 r4c3} – n16{r4c3 r5c5} – n15{r5c2 r5c4} – n14{r5c2 .} ==> r4c4 ≠ 17
whip[6]: n17{r1c1 r5c4} – n18{r4c1 r4c3} – n15{r4c3 r3c5} – n14{r4c1 r2c5} – n9{r2c5 r2c3} – n11{r2c3 .} ==> r4c4 ≠ 16
whip[6]: n16{r5c1 r2c5} – n17{r5c4 r1c4} – n18{r4c3 r2c3} – n14{r2c3 r4c4} – n11{r4c4 r4c3} – n9{r4c3 .} ==> r3c5 ≠ 15

whip[3]: n17{r1c1 r1c4} – n15{r1c4 r1c5} – n14{r2c2 .} ==> r2c5 ≠ 16
whip[7]: n3{r1c1 r1c4} – n5{r1c4 r2c5} – n6{r2c5 r3c5} – n7{r3c5 r4c4} – n2{r4c4 r2c3} – n9{r2c3 r4c3} – n11{r4c3 .} ==> r1c5 ≠ 4
whip[7]: n18{r2c2 r4c3} – n16{r4c3 r5c5} – n15{r5c2 r4c4} – n14{r5c1 r3c5} – n13{r4c1 r2c5} – n9{r2c5 r2c3} – n11{r2c3 .} ==> r5c4 ≠ 17
whip[1]: n17{r1c1 .} ==> r5c5 ≠ 16
whip[7]: n3{r3c1 r2c2} – n2{r2c2 r2c3} – n5{r2c3 r4c1} – n18{r4c1 r4c3} – n17{r1c1 r5c2} – n16{r3c1 r5c1} – n15{r5c4 .} ==> r3c1 ≠ 4
whip[7]: n8{r1c2 r2c2} – n9{r2c5 r2c3} – n6{r2c3 r4c1} – n18{r4c1 r4c3} – n17{r1c1 r5c2} – n16{r3c1 r5c1} – n15{r5c4 .} ==> r3c1 ≠ 7
whip[4]: n9{r2c5 r2c3} – n7{r2c3 r1c1} – n6{r1c4 r1c2} – n5{r1c4 .} ==> r2c2 ≠ 8
whip[7]: n12{r1c2 r2c2} – n11{r4c4 r2c3} – n14{r2c3 r4c1} – n18{r4c1 r4c3} – n17{r1c1 r5c2} – n16{r3c1 r5c1} – n15{r5c4 .} ==> r3c1 ≠ 13
whip[4]: n11{r4c4 r2c3} – n13{r2c3 r1c1} – n14{r5c5 r1c2} – n15{r5c4 .} ==> r2c2 ≠ 12
whip[7]: n16{r3c1 r1c4} – n14{r1c4 r2c5} – n13{r4c1 r3c5} – n12{r4c3 r4c4} – n11{r2c5 r4c3} – n9{r4c3 r2c3} – n17{r2c3 .} ==> r1c5 ≠ 15
whip[7]: n18{r2c2 r2c3} – n16{r2c3 r1c5} – n15{r2c2 r2c5} – n14{r3c1 r3c5} – n13{r4c1 r4c4} – n9{r4c4 r4c3} – n11{r4c3 .} ==> r1c4 ≠ 17
whip[1]: n17{r2c2 .} ==> r1c5 ≠ 16
**whip[8]: n8{r1c2 r5c2} – n9{r2c3 r4c3} – n6{r4c3 r4c1} – n5{r4c1 r3c1} – n4{r1c1 r2c2} – n18{r2c2 r2c3} – n2{r2c3 r4c4} – n3{r5c2 .} ==> r5c1 ≠ 7**
whip[6]: n2{r2c2 r4c4} – n4{r4c4 r5c4} – n5{r5c4 r4c3} – n6{r4c3 r5c2} – n7{r5c2 r4c1} – n8{r5c5 .} ==> r5c5 ≠ 3
**whip[8]: n3{r1c1 r5c2} – n2{r2c2 r4c3} – n5{r4c3 r4c1} – n6{r4c1 r3c1} – n7{r1c1 r2c2} – n8{r5c5 r1c2} – n9{r2c5 r2c3} – n18{r2c3 .} ==> r5c1 ≠ 4**
whip[7]: n8{r1c4 r1c2} – n9{r4c4 r2c3} – n6{r2c3 r3c1} – n5{r3c5 r4c1} – n18{r4c1 r4c3} – n17{r5c1 r5c2} – n4{r5c2 .} ==> r2c2 ≠ 7
whip[8]: n9{r2c5 r2c3} – n7{r2c3 r1c1} – n6{r1c4 r2c2} – n5{r2c5 r3c1} – n4{r4c3 r4c1} – n18{r4c1 r4c3} – n17{r5c1 r5c2} – n3{r5c2 .} ==> r1c2 ≠ 8
whip[1]: n8{r1c4 .} ==> r1c1 ≠ 7
**whip[8]: n7{r2c3 r4c1} – n8{r5c5 r5c2} – n9{r4c4 r4c3} – n5{r4c3 r2c2} – n18{r2c2 r2c3} – n17{r3c1 r1c2} – n16{r1c4 r1c1} – n15{r1c4 .} ==> r3c1 ≠ 6**
whip[3]: n4{r5c4 r5c2} – n6{r5c2 r5c1} – n7{r5c5 .} ==> r4c1 ≠ 5
whip[3]: n5{r4c3 r5c2} – n7{r5c2 r4c1} – n8{r5c5 .} ==> r5c1 ≠ 6
whip[2]: n8{r1c4 r5c2} – n6{r5c2 .} ==> r4c1 ≠ 7
whip[2]: n9{r2c3 r4c3} – n7{r4c3 .} ==> r5c2 ≠ 8
whip[4]: n3{r5c2 r4c3} – n5{r4c3 r5c1} – n6{r5c5 r4c1} – n7{r5c5 .} ==> r5c2 ≠ 4
whip[4]: n2{r4c3 r4c4} – n4{r4c4 r5c4} – n5{r5c2 r5c5} – n6{r5c2 .} ==> r4c3 ≠ 3
whip[4]: n3{r5c2 r4c4} – n2{r2c3 r4c3} – n5{r4c3 r5c5} – n6{r5c2 .} ==> r5c4 ≠ 4
whip[4]: n2{r4c4 r4c3} – n4{r4c3 r5c5} – n5{r5c2 r5c4} – n6{r5c2 .} ==> r4c4 ≠ 3
whip[4]: n4{r5c5 r4c4} – n6{r4c4 r5c4} – n7{r5c2 r4c3} – n8{r5c5 .} ==> r5c5 ≠ 5
whip[5]: n15{r2c2 r1c4} – n16{r2c2 r2c3} – r5c1{n15 n5} – n4{r5c5 r4c1} – r1c1{n5 .} ==> r2c5 ≠ 14
whip[3]: n14{r3c1 r4c4} – n12{r4c4 r2c5} – n11{r4c3 .} ==> r3c5 ≠ 13
whip[5]: n16{r2c2 r2c3} – n14{r2c3 r1c5} – r5c1{n13 n5} – n4{r5c5 r4c1} – r1c1{n5 .} ==> r1c4 ≠ 15
whip[3]: n15{r2c2 r2c5} – n16{r5c2 r1c4} – n13{r1c4 .} ==> r1c5 ≠ 14

whip[5]: n16{r2c2 r1c4} − n14{r1c4 r3c5} − r5c1{n13 n5} − n4{r5c5 r4c1} − r1c1{n5 .} ==> r2c5 ≠ 15
whip[2]: n17{r1c1 r2c3} − n15{r2c3 .} ==> r1c4 ≠ 16
whip[4]: n18{r2c3 r2c2} − n16{r2c2 r1c2} − n15{r3c1 r1c1} − n14{r1c4 .} ==> r2c3 ≠ 17
whip[5]: n15{r3c1 r4c4} − n16{r5c2 r4c3} − n13{r4c3 r2c5} − n11{r2c5 r2c3} − n9{r2c3 .} ==> r3c5 ≠ 14
whip[7]: n18{r2c3 r2c2} − n16{r2c2 r1c2} − n15{r1c2 r2c3} − n14{r2c3 r1c4} − r5c1{n13 n5} − r3c1{n5 n3} − n2{r4c3 .} ==> r1c1 ≠ 17
whip[7]: n3{r5c2 r5c4} − n5{r5c4 r4c4} − n2{r4c4 r4c3} − n6{r4c3 r3c5} − n7{r5c5 r2c5} − n11{r2c5 r2c3} − n9{r2c3 .} ==> r5c5 ≠ 4
**whip[8]: n4{r1c1 r4c1} − n6{r4c1 r5c2} − n3{r5c2 r3c1} − n2{r4c3 r2c2} − n17{r2c2 r1c2} − n18{r4c3 r2c3} − n16{r2c3 r1c1} − n15{r2c3 .} ==> r5c1 ≠ 5**
whip[2]: n7{r4c4 r4c3} − n5{r4c3 .} ==> r5c2 ≠ 6
whip[3]: n15{r3c1 r2c3} − r5c1{n14 n17} − n16{r4c3 .} ==> r1c4 ≠ 14
whip[1]: n14{r2c2 .} ==> r1c5 ≠ 13, r2c5 ≠ 13
whip[3]: n16{r1c1 r4c3} − r5c1{n15 n13} − n14{r5c5 .} ==> r5c4 ≠ 15
whip[3]: n15{r1c1 r4c4} − r5c1{n14 n17} − n16{r4c3 .} ==> r5c5 ≠ 14
whip[3]: n7{r5c4 r5c2} − r1c5{n6 n12} − r5c5{n12 .} ==> r4c1 ≠ 6
whip[2]: n8{r1c4 r4c3} − n6{r4c3 .} ==> r5c2 ≠ 7
whip[3]: n7{r1c4 r2c3} − r1c5{n6 n12} − r5c5{n12 .} ==> r2c2 ≠ 6
whip[1]: n6{r2c3 .} ==> r3c1 ≠ 5
whip[3]: n2{r2c2 r4c3} − n4{r4c3 r4c1} − n5{r3c5 .} ==> r5c2 ≠ 3
whip[3]: n7{r1c4 r2c3} − r1c5{n6 n12} − r5c5{n12 .} ==> r1c2 ≠ 6
whip[1]: n6{r1c4 .} ==> r1c1 ≠ 5
whip[2]: n8{r1c5 r1c4} − n6{r1c4 .} ==> r2c3 ≠ 7
whip[3]: n4{r1c1 r4c3} − n3{r1c1 r5c4} − n2{r2c3 .} ==> r4c4 ≠ 5
whip[3]: n3{r1c1 r5c4} − n5{r5c4 r5c2} − n6{r5c5 .} ==> r4c3 ≠ 4
whip[3]: n5{r5c2 r5c4} − n7{r5c4 r4c4} − n4{r4c4 .} ==> r5c5 ≠ 6
whip[3]: n4{r1c1 r4c4} − n6{r4c4 r4c3} − n7{r5c5 .} ==> r5c4 ≠ 5
whip[3]: n9{r4c3 r4c4} − n7{r4c4 r5c4} − n6{r3c5 .} ==> r4c3 ≠ 8
whip[3]: n16{r1c1 r4c3} − n14{r4c3 r5c4} − r5c1{n15 .} ==> r4c4 ≠ 15
whip[2]: n17{r1c2 r5c2} − n15{r5c2 .} ==> r4c3 ≠ 16
whip[4]: n15{r5c1 r4c3} − r5c1{n14 n17} − r3c1{n18 n3} − r1c1{n4 .} ==> r5c4 ≠ 14
whip[4]: n14{r1c2 r4c4} − n12{r4c4 r5c4} − n11{r3c5 r4c3} − n15{r4c3 .} ==> r5c5 ≠ 13
whip[3]: r5c5{n7 n12} − r1c5{n12 n5} − n6{r5c4 .} ==> r4c3 ≠ 7
whip[4]: n15{r1c1 r4c3} − n13{r4c3 r5c4} − n12{r5c2 r5c5} − n11{r4c3 .} ==> r4c4 ≠ 14
whip[2]: n16{r5c1 r5c2} − n14{r5c2 .} ==> r4c3 ≠ 15
whip[3]: n18{r2c2 r4c1} − n16{r4c1 r5c2} − n15{r5c2 .} ==> r5c1 ≠ 17
whip[2]: r5c1{n13 n16} − n15{r3c1 .} ==> r2c3 ≠ 14
whips[1]: n14{r3c1 .} ==> r1c4 ≠ 13;   n13{r2c2 .} ==> r1c5 ≠ 12, r2c5 ≠ 12
whip[2]: r1c5{n5 n8} − n7{r3c5 .} ==> r4c3 ≠ 6
whips[1]: n6{r4c4 .} ==> r5c2 ≠ 5;   n5{r1c2 .} ==> r4c1 ≠ 4
whip[2]: n2{r2c3 r2c2} − n4{r2c2 .} ==> r3c1 ≠ 3
whip[2]: n6{r5c4 r4c4} − r1c5{n7 .} ==> r5c4 ≠ 7
whip[2]: n9{r2c3 r4c4} − n7{r4c4 .} ==> r5c5 ≠ 8
whip[2]: r5c5{n7 n12} − n11{r4c3 .} ==> r4c4 ≠ 7
whip[2]: n5{r1c2 r2c5} − n7{r2c5 .} ==> r3c5 ≠ 6

whip[2]: n7{r1c4 r5c5} – r1c5{n8 .} ==> r5c4 ≠ 6
whip[2]: r5c5{n12 n7} – n6{r1c4 .} ==> r4c4 ≠ 12
whip[1]: n12{r5c2 .} ==> r3c5 ≠ 11
whip[2]: n4{r1c1 r4c4} – n6{r4c4 .} ==> r4c3 ≠ 5
whip[2]: n6{r1c4 r4c4} – r1c5{n7 .} ==> r5c5 ≠ 7
;;; Until now there has been no Single
singles and whips[1] to the end

In all these examples, one may wonder whether these long resolution paths could be simplified. By keeping all the assertion steps and, moving backwards from the end of the path, keeping only the elimination steps necessary to justify the assertions and eliminations that have been kept in the previous (from the end) elimination and assertion steps, it is likely that some intermediate eliminations could be avoided. But, as the first value assertions appear only near the end of the path, it is unlikely that this would lead to drastic simplifications. And, in any case, the B or W rating would not be changed.

# 17. Final remarks

In these final, partly retrospective remarks, which are intended neither as a summary nor as a conclusion, we shall highlight and comment some overlapping facets of what has been achieved for the pattern-based solution of the general finite Constraint Satisfaction Problem (with a few open questions). As for the practical applicability of the approach developed in this book, we merely refer to the many Sudoku examples and to the chapters dedicated to other logic puzzles.

## 17.1. About our approach to the finite CSP

### 17.1.1. About the general distinctive features of our approach

There are five main inter-related reasons why this book diverges radically from the current literature on the finite CSP[17]:

– almost everything in our approach, in particular all our definitions and theorems, is formulated in terms of *mathematical logic*, independently of any algorithmic implementation; (apart from the obvious logical re-formulation of a CSP, the current literature on CSPs is mainly about algorithms for solving them and comparisons of such algorithms); however, by effectively implementing them and applying them to various types of constraints, we have shown that these logical definitions are not mere abstractions and that they can be made fully operational;

– we systematically use redundant (but not overly redundant) sets of CSP variables; correlatively, we do not define labels as <variable, value> pairs but as equivalence classes of such pairs;

– we fix the main parameter defining the "size" of a CSP and we are not (or not directly) concerned with the usual theoretical perspectives of complexity, such as NP-completeness of a CSP with respect to its size;

– we nevertheless tackle questions of complexity, in terms of the statistical distribution of the *minimal instances* of a fixed size CSP; although all our resolution rules are valid for all the instances of a CSP, without any kind of restriction, we grant minimal instances a major role in all our statistical analyses and classification results; the thin layer of instances they define in the whole forest of possible instances (see chapter 6 for this view) allows to discard secondary problems that multi-solution or over-constrained instances would raise for statistics; (by contrast, the notion of minimality is almost unknown in the CSP world);

---

[17] We are not suggesting that our approach is better than the usual ones; we are aware that our purposes are non-standard and they may be irrelevant when speed of resolution is the main criterion; this is why we have stated our motivations with some detail in the Foreword.

– last but not least, our *purposes* lie much beyond the usual ones of finding a solution or defining the fastest algorithms for this. Here, *instead of the solution as a result, we are interested in the solution as a proof of the result, i.e. in the resolution path*. Accordingly, we have concentrated on finding *no-guessing, constructive, pure logic, pattern-based, rule-based, understandable, meaningful* resolution paths – though these words did not have a clear pre-assigned meaning.

We have taken this purpose into account in Part I by interpreting the "pure logic" requirement literally – i.e. as a solution completely defined in terms of mathematical logic (with no reference to any algorithmic notions). Thus, we have introduced a general resolution paradigm based on progressive candidate elimination. This amounts to progressive domain restriction, a classical idea in the CSP community. But, in our approach, each of these eliminations is justified by a single pattern – more precisely by a well defined *resolution rule* of a given *resolution theory* – and is interpreted in modal (non algorithmic) terms. We have established a clear logical (intuitionistic) status for the notion of a candidate (a notion that does not *a priori* pertain to the CSP Theory). Moreover, we have shown that the modal operator that naturally appears when one tries to provide a formal definition of a candidate can be "forgotten" when we state resolution rules, provided that we work with intuitionistic (or constructive) instead of classical logic (which is not a restriction in practice).

Once this logical framework is set, a more precise purpose can be examined, not completely independent from the vague "understandable" and "meaningful" original ones: one may want the *simplest* pure logic (or "rule-based" or "pattern-based") solution. As is generally understood without saying when one speaks of the simplest solution to a mathematical problem, we mean neither easiest to discover for a human being nor computationally cheapest, but simplest to understand for the reader. Even with such precisions, we have shown that "simplest" may still have many different, all logically grounded, meanings, associated with different (purely logical) ratings of instances.

Taking for granted that hard minimal instances of most fixed size CSPs cannot be solved by elementary rules but they require some kind of chain rules (with the classical xy-chains of Sudoku as our initial inspiration), we have refined our general paradigm by defining families of resolution rules of increasing logical (and computational) complexity, valid for any CSP: some reversible (Bivalue-Chains, g-Bivalue-Chains, Reversible-Subset-Chains, Reversible-g-Subset-Chains) and some orientated, much more powerful ones (whips, g-whips, $S_p$-whips, $gS_p$-whips, $W_p$-whips and similar braid families).

The different resolution paths obtained with each of these families when the simplest-first strategy is adopted correspond to different legitimate meanings of "simplest solution" (when they lead to a solution) and, in spite of strong subsumption relationships, we have shown (in several chapters, by examples of instances that have different ratings) that none of them can be completely reduced to another in a way that would preserve the ratings. Said otherwise: there does not seem to be any universal notion of (logical) simplicity for the resolution of a CSP.

### *17.1.2. About our resolution rules (whips, braids, …)*

Regarding these new families of chain rules, now reversing the history of our theoretical developments, four main points should be recalled:

– We have introduced a formal definition of Trial-and-Error (T&E), a procedure that, in noticeable contrast with the well known structured search algorithms (breadth-first, depth-first, …) and with all their CSP specific variants implementing some form of constraint propagation (arc-consistency, path-consistency, MAC, …), allows no "guessing", in the sense that it accepts no solution found by sheer chance during the search process: a value for a CSP variable is accepted only if all its other possible values have been tested and each of them has been constructively proven to lead to a contradiction.

– With the "T&E vs braids" theorem and its "T&E(T) vs T-braids" extensions to various resolution theories T, we have proven that a solution obtained by the T&E(T) procedure can always be replaced by a "pure logic" solution based on T-braids, i.e. on sequential patterns with no OR-branching accepting simpler patterns taken from the rules in T as their building blocks.

– Because its importance could not be over-estimated, we have proven in great detail that all our generalised braid resolution theories (braids, g-braids, $S_p$-braids, $gS_p$-braids, $B_p$-braids, B*-braids, …) have the *confluence property*. Thanks to this property, we have justified the idea that these types of logical theories can be supplemented by a "simplest first" strategy, defined by assigning in a natural way a different priority to each of their rules. When one tries to compute the rating of an instance and to find the simplest, pure logic solution for it, in the sense that it has a resolution path with the shortest possible braids in the family (which the T&E procedure alone is unable to provide), this strategy allows to consider only one resolution path; without this property, all of them should *a priori* be examined, which would add an exponential factor to computational complexity[18]. Even if the goal of maximum simplicity is not retained, the property of stability for confluence of these T-braid resolution theories remains very useful in practice, because it guarantees that valid eliminations and assertions occasionally found by any other consistent opportunistic solving methods (or any application-specific heuristics or any other search strategy) cannot introduce any risk of missing a solution based on T-braids or of finding only ones with unnecessarily long braids.

---

[18] The confluence property of a resolution theory T should not be interpreted beyond what it means. In particular, it does not allow to assign a rating to each candidate of an instance P: different resolution paths for P within T will always have the same rating of their hardest step,

– With the statistical results of chapter 6, we have also shown that, in spite of a major structural difference between whips and braids (the "continuity" condition), whips (even if restricted to the no-loop ones) are a very good approximation of braids[19], in the double sense that: 1) the associated W and B ratings are rarely different when the W rating is finite and 2) the same "simplest first" strategy, *a priori* justified for braids but not for whips, can be applied to whips, with the result that a good approximation of the W rating is obtained after considering only one resolution path (i.e. the concrete effects of non confluence of the whip resolution theories appear only rarely). This is the best situation one can desire for a restriction: it reduces structural (and computational) complexity but it entails little difference in classification results.[20] Of course, much work remains to be done to check whether this proximity of whips and braids is true for all the types of extended whips and braids defined in this book (it seems to be true for g-whips) and for CSPs other than Sudoku (it seems to be true also for Futoshiki, Kakuro, Map colouring, Numbrix® and Hidato®, as can be seen by the small number of occurrences of braids appearing in the resolution paths).

### 17.1.3. *About human solving based on these rules*

The four above-mentioned points have their correlates regarding a human trying to solve an instance of a CSP "manually" (or should we say "neuronally"?), as may be the "standard" situation for some CSPs, such as logic puzzles:

– It should first be noted that T&E is the most natural and universal resolution method for a human who is unaware of more complex possibilities and who does

---

but these hardest steps may correspond to the elimination of different candidates. This is not an abstract view; it happens very often.

[19] We have shown this in great detail for Sudoku, but the resolution paths we have obtained for most of the Futoshiki, Kakuro, Map colouring, Numbrix® and Hidato® examples confirm a similar behaviour.

[20] By contrast, the "reversibility" condition often imposed on chains in some Sudoku circles (never clearly formulated before *HLS*) is very restrictive and it leads some players to reject solutions based on non-reversible (or "orientated") chains (such as whips and braids) and to the (in our opinion, hopeless for hard instances) search for extremely complex patterns (such as all kinds of what we would call extended g-Fish patterns: finned, sashimi, chains of g-Fish, …). This said, we acknowledge that Reversible-Subset-Chains (Nice Loops, AICs) may have some appeal for moderately complex instances.

not accept guessing. This was initially only a vague intuition. But, with time, it has received very concrete confirmations from our experience in the Sudoku microworld (with friends, students, contacts, or from questions of newcomers on forums), considering the way new players spontaneously re-invent it without even having to think of it consciously. Indeed, it does not seem that they reject guessing *a priori*; they start by using it and they feel unsatisfied about it after some time, as soon as they understand that it is an arbitrary step in their solution; "no-guessing" then appears as an additional *a posteriori* requirement. Websites dedicated to the other logic puzzles studied in this book are another source of confirmation: T&E (in various names and usually in informal guises, but always in a form compatible with our formal definition) always appears as the most widely used resolution method, except of course for the easiest puzzles.

— The "T&E vs braids" theorem means that the most natural T&E solving technique, in spite of being strongly anathemised by some Sudoku experts, is not so far from being compatible with the abstract "pure logic" requirement. Moreover, its proof shows that a human solver can always easily modify a T&E solution in order to present it as a braid solution. Thanks to the subsumption theorems or to the more general "T&E(T) vs T-braids" theorem, this remains true when he learns more elaborate techniques (such as Subset or g-Subset rules) and he starts to combine them with T&E.

— Finding the shortest braid solution is a much harder goal than finding any solution based on braids and this is where the main divergence with a solution obtained by mere T&E occurs. For the human solver who started with T&E, it is nevertheless a natural step to try to find a shorter (even if not the shortest) solution. An obvious possibility consists of excising the useless branches of what he has first found; but he can also look for alternative braids, either for the same elimination or for a different one.

— As for the fourth point, a human solver is very likely to have spontaneously the idea of using the continuity condition of whips to guide his search for a contradiction on some target Z: it means giving a preference to pushing further the last tried step rather than a previous one. It is so natural that he may even apply it without being aware of it.

Finally, for a human solver, the transition from the spontaneous T&E procedure to the search for whips can be considered as a very natural process. Learning about Subsets and g-Subsets and looking for them can also be considered as a natural, though different, evolution. And the two can be combined. Once more, there is no unique way of defining what "the best solution" may mean.

Of course, a human player can also follow a very different learning path, starting with application specific rules, such as xy-chains in Sudoku and progressively trying to spot patterns from the ascending sequence of more complex rules following a discovery path similar to that in *HLS*. But, unless he limits himself to moderately complex instances, he cannot avoid the kind of non-reversible chain patterns introduced in this book.

### 17.1.4. About a strategic level

We have used the confluence property to justify the definition of a "simplest-first" strategy for all the braid and generalised braid (and, by extension, all the whip and generalised whip) resolution theories. This strategy perfectly fits the goals of finding the simplest solution (keeping the above comments on "simplest" in mind) and of rating an instance.

What the "simplest-first" strategy guarantees should be clear: for a resolution theory T with the confluence property, it finds a solution with the smallest T-rating (if there is one); in any case, at each step in any resolution path within T, the available assertion or elimination with the lowest T-rating is applied (or, when there are several, one of the possible assertions or eliminations with this rating is randomly chosen and applied). One thing it does *not* guarantee is that all these steps are necessary for justifying the next ones or that there is no other resolution path with fewer eliminations (not counting Elementary Constraints Propagation).

Other systematic strategies can also be imagined. One of them consists of considering subsets of CSP variables of "same type" and defining special cases of all the rules by restricting them to such subsets of variables and by assigning these cases higher priorities than their initial full version. This is what we have done for Sudoku in *HLS1*, with the 2D rules. It is easy to see that, as the "2D" rules are the various 2D projections (on the rc-, rn-, cn- and bn- spaces) of the "3D" ones presented here, all the 2D-braid theories (in each of these four 2D spaces) are stable for confluence and have the confluence property; it is therefore also true of their union. In *HLS1*, we have shown that 97% of the puzzles in the random Sudogen0 collection can be solved by such 2D rules (the real percentage may be a little less for an unbiased sample). We still consider these rules as interesting special cases that have an obvious place in the "simplest-first" strategy and that may be easier to find and/or to understand for a human player.

Now, it is very unlikely that any human solver would proceed in such a systematic way as described in any of the above two strategies. He may prefer to concentrate on some aspect of the puzzle and try to eliminate a candidate from a chosen cell (or group of cells). As soon as he has found a pattern justifying an elimination, he applies it. This could be called the opportunistic "first-found-first-applied" strategy. And, thanks to stability for confluence, it is justified in all the generalised braid resolution theories defined in this book. In simple terms, there can be no "bad" move able to block the way to the solution. This conclusion is in strong opposition to claims often made in some Sudoku circles that adding a clue (or asserting a value) may make a puzzle harder; such views can only rely on forgetting a few facts: 1) such cases arise only when rules of uniqueness are involved; 2) they arise only when hardness is measured by the SER; 3) if added to a resolution theory with the confluence property, a rule for uniqueness destroys it, unless it is given higher priority than all the other rules; 4) there is a confusion in SER between the priority of a rule and its rating; 5) this confusion prevents rules for uniqueness to apply as soon as they should; 6) as a result, the SER rating of rules for uniqueness is inconsistent.

What may be missing however in our approach is more general "strategic" knowledge for orientating the search: when should one look for such or such pattern? This would be meta-knowledge about how to use the knowledge included in the resolution theories. It would very likely have to be application-specific[21].

But the fact is, we have no idea of which criteria could constitute a basis for such meta-knowledge. Worse, even in the most studied Sudoku CSP, whereas there is a plethora of literature on resolution techniques (sometimes misleadingly called strategies), nothing has ever been written on the ways they should be used, i.e. on what might legitimately be called strategies. In particular, one common prejudice is that one should first try to eliminate bivalue/bilocal candidates (i.e., in our vocabulary, candidates in bivalue rc, rn, cn or bn cells). Whereas this may work for simple puzzles, it is almost never possible for complex ones. This can easily be seen by examining the hard examples of this book (for any of the CSPs we have studied), with the long sequences of whip eliminations necessary before a Single is found: if any of these eliminations had occurred for a bivalue CSP variable, then it would have been immediately followed by a Single.

## 17.2. About minimal instances and uniqueness

### 17.2.1. Minimal instances and uniqueness

Considering that, most of the time, we restrict our attention to minimal instances that (by definition) have a unique solution, one may wonder why we do not introduce any "axiom" of uniqueness. Indeed, there are many reasons:

– it is true that we restrict all our statistical analyses of resolution rules to minimal instances, for reasons that have been explained in the Introduction; but it does not entail that validity of resolution rules should be limited *per se* to minimal instances; on the contrary, they should apply to any instance; in a few examples in this book, our rules have even been used to prove non-uniqueness or non-existence of solutions;

– as mentioned in the Introduction, from the point of view of Mathematical Logic, uniqueness cannot be an *axiom*, at least not an axiom that could impose uniqueness of a solution; for any instance, it can only be an *assumption*; moreover, when incorrectly applied to a multi-solution instance, the *assumption* of uniqueness can lead, via a vicious circle, to the erroneous *conclusion* that an instance has a unique solution; we have given an example in *HLS1*, section XXII.3.1 (section 3.1 of chapter "Miscellanea" in *HLS2*);

---

[21] [Laurière 1978] presents a different perspective, based on general-purpose heuristics.

– uniqueness is not a constraint the CSP solver (be he human or machine) is expected or can choose to satisfy; in some CSPs or some situations (such as for statistical analyses or for logic puzzles like Sudoku), uniqueness may be a requirement to the provider of instances (he should provide only "well formed" instances, i.e. minimal instances or, at least, instances with a unique solution); the CSP solver can then decide to trust his provider or not; if he does and he uses rules based on it in his resolution paths, then uniqueness can best be described as an oracle; for this reason, in all the solutions we have given, uniqueness is never assumed, but it is proven constructively from the givens;

– the fact is, there is no known way of exploiting the assumption of uniqueness for writing any general resolution rule for uniqueness; and we can take no inspiration in the Sudoku case, because all the known techniques based on the assumption of uniqueness are Sudoku specific;

– in the Sudoku case, if any of the known rules of uniqueness is added in its usual form to a resolution theory with the confluence property, it destroys confluence (see *HLS* for an example); however, we have not explored the possibility of other (more complex) formulations that could preserve it;

– still in the Sudoku case, it does not seem that the known rules for uniqueness have much resolution power; there is no known example that could be solved if they were added to "standard" resolution rules but that could not otherwise.

Of course, we are not trying to deter anyone from using uniqueness in practice, if they like it, in CSPs for which it allows to formulate specific resolution rules, such as Sudoku (where it has always been a very controversial topic, but it has also led to the definition of smart techniques); in some rare cases, it can simplify the resolution paths. We are only explaining why we chose not to use it in our theoretical approach. One should always keep in mind that theory often requires more stringent constraints than practice.

### 17.2.2. *Minimal instances vs density and tightness of constraints*

Two global parameters of a CSP, its "density of constraints" and its "tightness", have been identified in the classical CSP literature. Their influence on the behaviour of general-purpose CSP solving algorithms has been studied extensively and they have also been used to compare such algorithms. (As far as we know, these studies have been about unrestricted CSP instances; we have been unable to find any reference to the notion of a minimal instance in the CSP literature.)

Definitions (classical in CSPs): the *density of constraints* of a CSP is the ratio between the number of label pairs linked by some constraint (supposing that all the constraints are binary) and the total number of label pairs; the *tightness* of a CSP is the ratio between the number of label pairs linked by some "strong" constraint (i.e. some constraint due to a CSP variable) and the number of label pairs linked by some constraint.

Density reflects the intuitive idea that the vertices of an undirected graph (here, the graph of labels) can be more or less tightly linked by the edges (here the direct binary contradictions); it also evokes a few general theorems relating the density and the diameter of a random graph (a topic that has recently become very attractive because of communication networks). Tightness evokes the difference we have mentioned between Sudoku or LatinSquare (tightness 100%, for any grid size) and N-Queens (tightness ~ 50%, depending on n).

In the context of this book, relevant questions related to these parameters should be about their influence on the scope of the various types of resolution rules with respect to the set of minimal instances of the CSP. However, how the definitions of these two parameters should be adapted to this context is less obvious than it may seem at first sight. The question is, should one compute these parameters using all the labels of the CSP or only the actual candidates? In the latter case, they would change with each step of the resolution process.

Taking the 9×9 Sudoku example, the computation is easy for labels: there are 729 labels (all the nrc triplets) and each label is linked by some constraint to 8 different labels on each of the n, r, c axes, plus 4 remaining labels on the b axis. Each label is thus linked by some constraint to the same number (28) of other labels and one gets a density equal to $28/728 = 3.846\%$. More generally, for n×n Sudoku with $n = m^2$, density is: $(4m^2-2m-2)/(m^6-1)$; it tends rapidly to zero (as fast as $4/n^2$) as the size n of the grid increases.

However, considering the first line of each Sudoku resolution path in this book, one can check that for a minimal puzzle, after the Elementary Constraint Propagation rules have been applied (i.e. after the straightforward initial domain restrictions), the number of candidates remaining in the initial resolution state $RS_P$ of an instance P is much smaller. As all that happens in a resolution path depends only on $RS_P$, a definition of density based on the candidates in $RS_P$ can be expected to be more relevant. But, the analysis of the first series of 21,375 puzzles produced by the controlled-bias generator, leads to the following conclusions, showing that neither the number of candidates in $RS_P$ nor the density of constraints in $RS_P$ have any significant correlation with the difficulty of a puzzle P (measured by its W rating):

– the number of candidates in $RS_P$ has mean 206.1 (far less than the 729 labels) and standard deviation 10.9; it has correlation coefficient -0.20 with the W rating;

– the density of constraints in $RS_P$ has mean 1.58% (much less than when computed on all the labels) and standard deviation 0.05%; its has correlation coefficients -0.16 with the number of candidates in $RS_P$ and -0.06 with the W rating.

One (seemingly more interesting) open question is: is there a correlation between the rating of the current "simplest" possible elimination and the current density (based on the current set of candidates before the elimination). In the instances with a hard first step that we checked, there was no significant deviation from the mean; but the question may be worth more systematic investigation.

Can tightness give better or different insights? This parameter plays a major role in the left to right extension steps of the partial chains of all the types defined in this book. In n×n Sudoku or n×n LatinSquare, tightness is 100%, whatever the value of n; these examples can therefore not be used to investigate this parameter. If there are few CSP variables, there may be few chains. In this context, it should however be noticed that, from the millions of Sudoku puzzles we have solved, problems that appear for the hardest ones solvable by whips or g-whips arise from two opposite causes: not only because there are too few partial whips or g-whips (and no complete ones), but also because there are too many useless partial whips or g-whips (eventually leading to computational problems due to memory overflow).

One idea that needs be explored in more detail is that the possible statistical effects of initial density or tightness of constraints on complexity are minimised (as is the case for the number of givens) by considering the thin layer of minimal instances (because they have a unique solution). But the 16×16 and 25×25 Sudoku examples in section 11.5 show that they cannot be minimised to the point of limiting the depth of T&E in a way independent of density (or grid size).

**17.3. About ratings, simplicity, patterns of proof**

Our initial motivations included three broad categories of (vague) requirements:

– a "pure logic", "pattern-based", "rule-based", "constructive" solution with "no guessing",

– an "understandable", "explainable" solution,

– and a "simplest" solution.

If the first type has been given a precise meaning and has been satisfied in Part I, and if the second can be considered as a more or less subjective mix of the other two, one may wonder what the third has become or rather how it had to be refined.

*17.3.1. About general ratings and the requirement for the "simplest" solution*

For any instance P of any CSP, several ratings of P have been introduced: W, B, gW, gB, S+W, S+B, SW, SB,… All of them have been defined in pure logic terms, they are invariant under the symmetries of the CSP (if its constraints are properly modelled) and they are intrinsic properties of P. They have also been shown to be largely mutually consistent, i.e. they assign the same finite ratings "most of the time" to instances in T&E(1)[22] – which probably already includes much more than what can be solved "manually" by normal human beings.

Moreover, if one nevertheless wants to go further, we have defined the WW, BB, W*W, B*B ratings and we have shown that the BB rating is finite for any instance in T&E(2), i.e. that can be solved with at most two levels of Trial-and-Error.

What the multiplicity of these logically grounded ratings also shows is that there is one thing all our formal analyses cannot do in our stead: choosing what should be considered as "simplest". And we strongly believe that there can be no universal *a priori* definition of simplicity of a resolution path, even when one adopts a hardest-step view of simplicity and even for a problem as "simple" as Sudoku, let alone for the general finite CSP. Simplicity can only depend on one's specific goals. For definiteness, let us illustrate this with the Sudoku CSP.

If one is interested in providing examples of some particular set of techniques or promoting them, then a solution considered as the simplest must (tautologically) use only these techniques; the job will then be to provide nice handcrafted examples of such puzzles (and, sometimes, to carefully hide the fact that they are exceptional in the set of all the minimal puzzles); this is the approach implicitly taken by most Sudoku puzzle providers and most databases of "typical examples" associated with computerised solvers. Unfortunately, apart from those here and in *HLS*, we lack both formal studies of such sets of techniques and statistical analyses of their scopes.

If one is interested in the simplest pattern-based solution for all the minimal puzzles, then, considering the statistical results of chapter 6, a whip solution could certainly be considered as the simplest one, *statistically*; a g-whip solution would be a good alternative, as the structural complexity of g-whips is not much greater than that of whips. "Statistically" means that, in rare cases, a better solution including Subsets or g-Subsets or Reversible-Subset-Chains or S-whips or W-whips could be found – "better" in the sense that it would provide a smaller rating (at the cost of using more complex patterns). Although it is hard to imagine a motivation for this when whips or g-whips would be enough, one could also use $W_p^*$-whips or $B^*$-braids, i.e. rely on T&E(2) contradictions as if they were ordinary constraints; doing this may ultimately be only a matter of personal taste [provided that confusion is not created by comparing without caution ratings that involve these derived constraints with those that do not].

---

[22] Strictly speaking, this has been shown in precise terms only for 9×9 Sudoku, but there are serious indications that it remains true for the other logic puzzles we have examined.

If one is interested in the "hardest" instances, then it should first be specified precisely what is meant by "hardest" (in particular with respect to which rating); this may seem obvious, but it remains frequent on Sudoku forums to see (implicit) references to two different ratings in the same sentence. In Sudoku, puzzles harder than the "hardest" known ones with respect to the prevailing SER rating keep being discovered. One can consider that Part III of this book (apart from chapter 8) is dedicated to resolution rules for the hardest puzzles (not in the sense of the SER, but in the broader sense that they are not solvable by braids or g-braids, or equivalently by at most one level or T&E or gT&E). Much depends on two parameters: the maximal depth d of Trial-and-Error necessary to solve these instances and the maximal look-ahead p necessary to solve them at depth d-1. [Even for 9×9 Sudoku, although we have shown that there are very strong reasons to conjecture that d = 2 and p = 7, i.e. that every puzzle can be solved by $B_7$-braids, we have no formal proof of this.]

The T&E(2) land is where many different possibilities appear. For instances there, instead of looking for the simplest solution with respect to the universal BB rating, one can consider two simpler approaches: 1) the $B_?B$ classification, possibly followed by a $B_p$-braids solution, and 2) the $B_p$*-braids view. As an illustration of the latter, the solution given for EasterMonster in section 12.3.3.1 proceeds in two steps: the first step provides the main lines of the proof as a sequence of B*-whips[1] eliminations; the second step should contain the "details" of the proof by exhibiting the bi-braids justifying each of these B*-whips[1]. This led us to introduce the general notion of a pattern of proof, but this is a vast topic and we have only skimmed it.

As shown by the sk-loop examples in chapter 13, it may occasionally happen that application-specific patterns (often tightly related to patterns of givens enjoying very particular symmetries or quasi-symmetries) reduce the complexity of an instance (measured in this case by the $B_?B$ classification). However, for the very hardest instances, it may also happen that the whole requirement of simplicity becomes merely meaningless: the existence of extremely rare but very hard instances that cannot be solved by any "simple" rules (in a vague intuitive sense of "simple") is a fact that cannot be ignored.

### 17.3.2. About adapting the general ratings to an application

The Futoshiki CSP allows two additional comments about how the general ratings introduced in this book can easily be adapted to a particular CSP in order to better take into account any "natural" notion of simplicity in specific applications:

– although "ascending chains" of any size are equivalent to series of whips of length one, they are so natural that presenting them as whips would make the resolution paths look unnecessarily complicated, with lots of elementary and boring steps; this means that, in some cases, our requirement of simplicity cannot be defined based only on formal criteria but it may have to take into account matters of presentation; however, from a technical point of view, this is more a cosmetic than a deep matter;

– "hills" and "valleys" raise a much more interesting question; they are almost as natural and obvious patterns as ascending chains, whatever their size; although they can always be considered as Subsets or as S-whips and their complexity in terms of the equivalent Subsets or S-whips would be much higher than that of ascending chains, it would be intuitively absurd to assign them a much greater complexity, because there is not much difference between finding or understanding hills and valleys and finding or understanding ascending chains, and this does not depend on their size; fortunately, stability for confluence allows to combine any $B_n$ or $gB_n$ theory with hills and valleys of unrestricted size without loosing confluence; this means that hills and valleys can consistently be assigned any rating one wants in the $B_n$ or $gB_n$ hierarchy; said otherwise, one can refine the notion of simplicity in such a way that it becomes adapted to the specificities of the Futoshiki CSP, without loosing the benefits of the general theory; if needed, this illustrates again the importance of the confluence property.

The above remarks can be transposed to Kakuro and to the coupling rules: any resolution theory should include them (and we have accordingly defined the $^+$ variants of all the theories introduced in this book: $BRT^+$, $W_1^+$, …).

### 17.3.3. Similarity between Subset and whip/braid patterns of same size

We have noticed a remarkable formal similarity between the Subset and the whip/braid patterns of same size (see Figure 11.3 and comments there). It has appeared in very explicit ways in the proofs of the confluence property and of the generalised "T&E(T) vs T-braids" theorems for the $S_p$-braids and $B_p$-braids. But the general subsumption theorems in section 8.7 and the Sudoku-specific statistical results in Table 8.1 suggest that whips/braids have a much greater resolution power than Subsets of same size. As mentioned in section 8.7.3, these results indicate that the definition of Subsets is much more restrictive than the definition of whips/braids. And Table 11.1 shows that the same kind of very large difference in resolution power remains true for the generalised braids including these patterns as right-linking elements, at least for the Sudoku CSP.

In Subsets, transversal sets are defined by a single constraint. In whips, the fact of being linked to the target or to a given previous right-linking candidate plays a role very similar to each of these transversal sets. But being linked to a candidate is much less restrictive than being linked to it via a pre-assigned constraint; in this respect, the three elementary examples for whips of length 2 in sections 8.7.1.1 and 8.8.1 are illuminating. As shown by the subsumption and almost-subsumption results in section 8.7, the few cases of Subsets not covered by whips because of the restrictions related to sequentiality are too rarely met in practice to be able to compensate for this.

For the above reasons, we conjecture that, in any CSP, whips/braids have a much greater resolution potential than Subsets of same length p, at least for small values of p; and $B_p$-braids have a much greater resolution potential than $S_p$-braids. For large values of p, it is likely true also, but it is less clear because there may be an increasing number of cases of non-subsumption but there may also be more ways of being linked to a candidate. Much depends on how many different constraints a given candidate can participate in. This is an area where more work is necessary.

**17.4. About CSP-Rules**

As mentioned in the Foreword and as can be checked by a quick browsing of this book, it is almost completely written at the logic level; it does not say much about the algorithmic or the implementation levels – beyond the fact that our detailed definitions provide unambiguous specifications for them, whichever computer language one finally chooses. However, a few general indications on CSP-Rules may be welcome.

In this section, it may be useful for the reader not yet familiar with the basic principles of expert systems and/or inference engines to read one of the quick introductions that are widely available on the Web (in particular the notions of a rule base and a fact base); the CLIPS documentation can be browsed, but this is not essential for reading what follows.

*17.4.1. CSP-Rules*

Almost all[23] the resolution paths appearing in this book were obtained with the current last version of CSP-Rules (version 1.2), the generic finite CSP solver we wrote in the rule-based language of the CLIPS[24] inference engine.

---

[23] The only exceptions are the few N-Queens examples, for which we did not implement the necessary interface (mainly because we could not find any generator of N-Queen instances and we did not want to spend time on writing one, so that we finally have only very easy instances). Two other exceptions are mentioned explicitly in the text.

[24] CLIPS for Mac OSX, version 6.30. CLIPS is the acronym for "C Language Integrated Production System"; it is a distant descendant of OPS (the Official Production System) but its syntax (inherited from ART, a commercial expert system shell) is much better. CLIPS is free,

In principle, CSP-Rules can also be run on JESS[25] (all the rules we have implemented use only the part of the syntax ensuring compatibility). But JESS is slower and we have given up trying to fill up the compatibility issues when coding the application-specific parts of the various CSPs or to deal with Java-specific memory management problems.

CSP-Rules was designed from the start as a *research tool*, with the main purpose of proving concretely that the general resolution rules and the simplest-first strategy defined in this book can be implemented in a generic way and can lead in practice to real solutions for different CSPs, even for their hard instances. Another purpose was to allow quick implementation of tentative rules and to test their resolution potential with respect to those we had already defined. Finally, we also wanted to make it easy to add application-specific rules (such as sk-loops in Sudoku, hills and valleys in Futoshiki or coupling rules in Kakuro) or to code alternative strategies without having to deal with a programming language like C.

Saying that we conceive CSP-Rules as a research tool means in particular that it was not designed with high speed or low memory purposes in mind, although it includes a few standard tricks to avoid too fast memory explosion and it has been used several times to solve millions of instances. It seems obvious to us that a direct implementation in C or any other procedural language could lead to large improvements in computation times and memory requirements, especially for hard instances – although the exponential increase of the number of partial patterns (with respect to their length) before a full one can be used to produce an elimination is inherent in some instances. The reference to g-labels and S-labels instead of g-candidates and Subsets in g-whips and S-whips is a key for many optimisations of memory.

CSP-Rules is a descendant of SudoRules, the Sudoku solver we originally developed in parallel with the writing of *HLS*. As the main parts of the later versions of SudoRules were already written in an almost application independent way, it was easy to maximally reduce and to isolate the unavoidably application-specific parts. The version of SudoRules (16.2) based on CSP-Rules that was used in the Sudoku examples presented in this book is 100% equivalent to (i.e. it produces exactly the same resolution paths as) the last version before the split (namely 15b.1.12, which has been our version of reference at the time of writing *CRT*), when the same rules are enabled.

---

which probably largely contributed to make it one of the most widely adopted shells. Another reason is that CLIPS implements the RETE algorithm that made OPS famous, with all the improvements that appeared since that time, making it one of the most efficient shells.

[25] Current version as of this writing, i.e. 6.1p2. JESS is the acronym for "Java Expert System Shell"; it was initially the Java version of CLIPS; but, due to the underlying language, it has grown up differently and there are now compatibility issues.

The current version of CSP-Rules implements the following sets of rules (we have also implemented other tentative rules but they are not mentioned in this book because they did not lead to interesting results):

– BRT (i.e. ECP + Single + Contradiction detection + Solution detection),
– bivalue-chains, whips, braids,
– g-bivalue-chains, g-whips, g-braids,
– forcing whips, forcing braids,
– bi-whips, bi-braids,
– forcing bi-whips, forcing bi-braids,
– W*-whips, B*-braids.

For each of these patterns and for each possible length, CSP-Rules has two or three rules (one or two for building the partial patterns, one for detecting the full ones and doing the eliminations), plus an activation rule (used mainly for memory optimisation) and a tracking rule (as they are mainly used for tracking the numbers of partial patterns and for statistics, their output does not appear in the resolution paths given here). All these rules are written only in the generic terms of candidates, g-candidates, CSP-variables, links and g-links. Their effective output (what we want to appear in a resolution path) is controlled by a set of global variables.

CSP-Rules also implements the generic parts of functions used in the left-hand side of rules (when it is both possible and more efficient to make a test [linked, glinked, …] than to write an additional explicit condition pattern) or for the interfacing with specific applications (e.g. for printing the different steps of the resolution path – although it already implements the generic parts of the output functions). Any application must provide the specific parts of these functions.

CSP-Rules also provides the possibility of computing T&E(T) and bi-T&E(T) for any resolution theory T whose rules are programmed in CSP-Rules.

Because it was too hard to do this in sufficiently efficient ways, CSP-Rules does not implement a generic version of Subsets (let alone of g-Subsets). Instead, it has a standard version of Subsets (upto size four) valid for CSPs based on a square (or rectangular) grid (like most of the examples in this book), with a sub-version with blocks as in Sudoku. In the Kakuro CSP, its adaptation to the case of Subsets restricted to sectors was straightforward.

The generation of instances is not part of CSP-Rules.

### 17.4.2. Configuration of an application for solving an instance

Any application (any particular CSP) has a configuration file allowing to choose the resolution theory one wants to use, i.e. which patterns should be enabled and up to which size. Technically, "enabled" means loaded into the rule base; it does not mean "activated". An enabled pattern gets activated only if necessary (i.e. if shorter ones are not enough to solve the instance under consideration).

Consistency of the chosen parameters is ensured automatically, e.g.

– for any pattern P[n] depending on a size or length parameter n, if P[n] is explicitly enabled, then P[n-1], … P[1] are automatically enabled;

– if g-braids of length upto n are enabled, then braids and g-whips of length upto n are enabled if they have not been explicitly enabled with a larger length;

– if g-whips of length upto n are enabled, then whips of length upto n are enabled if they have not been explicitly enabled with a larger length;

– if braids of length upto n are enabled, then whips of length upto n are enabled if they have not been explicitly enabled with a larger length…

However, bivalue-chains are not automatically enabled when whips are enabled. This may be changed in the future. But we have found it useful to keep this degree of freedom, as enabling special types of whips sometimes allows to find different whip resolution paths (see an example in section 5.10.3).

### *17.4.3. Resolution strategies predefined in CSP-Rules*

The current version of CSP-Rules has only one resolution strategy, the "simplest-first", with the priorities as described in section 7.5.2:
ECP > S >
biv-chain[1] > whip[1] > g-whip[1] > braid[1] > g-braid[1] >
… > …
biv-chain[k] > whip[k] > g-whip[k] > braid[k] > g-braid[k] >
biv-chain[k+1] > whip[k+1] > g-whip[k+1] > braid[k+1] > g-braid[k+1] > …

A few things are easy to change, such as assigning braids[k] a higher priority than g-whips[k] or introducing more special cases of whips. For radically different strategies, the main problem would not be to code them in CSP-Rules, but to first define them (see the remarks in section 17.1.4).

### *17.4.4. Applications already interfaced to CSP-Rules*

As of this writing, the current version of CSP-Rules has application-specific interfaces (and in some cases a few application-specific resolution rules, possibly including alternative versions of the rules in BRT, e.g. different rules for Naked and Hidden Singles) for the following CSPs: *LatinSquare, Sudoku, Futoshiki, Kakuro, Map-colouring, Numbrix® and Hidato®*. For each of them, the volume of the source code of the application-specific part (including mainly input-output functions) is between 3% and 5% of the total generic CSP-Rules part. For Sudoku, more functions had been written in the previous versions of SudoRules, but they were mainly intended for statistical analyses and cannot be considered as necessary for the normal resolution of instances; moreover, with a little more adaptation work, they could also be made generic, if needed.

# 18. References

***Books and articles***

[Apt 2003]: APT K., *Principles of Constraint Programming*, Cambridge University Press, 2003.

[Barcan 1946a]: BARCAN M., A Functional Calculus of First Order Based on Strict Implication, Journal of Symbolic Logic, Vol. 11 n°1, pp. 1-16, 1946.

[Barcan 1946b]: BARCAN M., The Deduction Theorem in a Functional Calculus of First Order Based on Strict Implication, Journal of Symbolic Logic, Vol. 12 n°4, pp. 115-118, 1946.

[Berthier 2007a]: BERTHIER D., *The Hidden Logic of Sudoku*, First Edition, Lulu.com Publishers, May 2007.

[Berthier 2007b]: BERTHIER D., *The Hidden Logic of Sudoku, Second Edition*, Lulu.com Publishers, November 2007.

[Berthier 2008a]: BERTHIER D., From Constraints to Resolution Rules, Part I: Conceptual Framework, *International Joint Conferences on Computer, Information, Systems Sciences and Engineering (CISSE 08)*, December 5-13, 2008, Springer. Published as a chapter of *Advanced Techniques in Computing Sciences and Software Engineering*, Khaled Elleithy Editor, pp. 165-170, Springer, 2010.

[Berthier 2008b]: BERTHIER D., From Constraints to Resolution Rules, Part II: chains, braids, confluence and T&E, *International Joint Conferences on Computer, Information, Systems Sciences and Engineering (CISSE 08)*, December 5-13, 2008, Springer. Published as a chapter of *Advanced Techniques in Computing Sciences and Software Engineering*, Khaled Elleithy Editor, pp. 171-176, Springer, 2010.

[Berthier 2009]: BERTHIER D., Unbiased Statistics of a CSP - A Controlled-Bias Generator, *International Joint Conferences on Computer, Information, Systems Sciences and Engineering (CISSE 09)*, December 4-12, 2009, Springer. Published as a chapter of *Innovations in Computing Sciences and Software Engineering*, Khaled Elleithy Editor, pp. 11-17, Springer, 2010.

[Berthier 2011]: BERTHIER D., *Constraint Resolution Theories*, Lulu.com Publishers, November 2011.

[Bridges et al. 2006]: BRIDGES D. & VITA L., *Techniques of Constructive Analysis*, Springer, 2006.

[Dechter 2003]: DECHTER R., *Constraint Processing*, Morgan Kaufmann, 2003.

[Feys 1965]: FEYS R., *Modal Logics*, Fondation Universitaire de Belgique, 1965.

[Fitting 1969]: FITTING M., *Intuitionistic Logic, Model Theory and Forcing*, North Holland, 1969.

[Fitting et al. 1999]: FITTING M. & MENDELSOHN R., *First-Order Modal Logic*, Kluwer Academic Press, 1999.

[Freuder et al. 1994]: FREUDER E. & MACKWORTH A., *Constraint-Based Reasoning*, MIT Press, 1994.

[Früwirth et al. 2003]: FRÜWIRTH T. & SLIM A., *Essentials of Constraint Programming*, Springer, 2003.

[Garson 2003]: GARSON J., Modal Logic, *Stanford Encyclopedia of Philosophy*, 2003, available at http://plato.stan ford. edu/entries/logic-modal.

[Gary et al. 1979]: GARY M. & JOHNSON D., *Computers and Intractability: A Guide to the Theory of NP-Completeness*, Freeman, 1979.

[Gentzen 1934], GENTZEN G., Untersuchungen über das logische Schlieβen I, *Mathematische Zeitschrift*, vol. 39, pp. 176-210, 1935.

[Hendricks et al. 2006]: HENDRICKS V. & SYMONS J., Modal Logic, *Stanford Encyclopedia of Philosophy*, 2006, available at http://plato.stan ford. edu/entries/logic-modal.

[Hintikka 1962]: HINTIKKA J., *Knowledge and Belief: An Introduction to the Logic of the Two Notions*, Cornell University Press, 1962.

[HLS1, HLS2, HLS]: respectively, abbreviations for [Berthier 2007a], [Berthier 2007b] or for any of the two.

[Guesguen et al. 1992]: GUESGUEN H.W. & HETZBERG J., *A Perspective of Constraint-Based Reasoning*, Lecture Notes in Artificial Intelligence, Springer, 1992.

[Kripke 1963]: KRIPKE S., Semantical Analysis of Modal Logic, *Zeitschrift für Mathematische Logik und Grundlagen der Matematik*, Vol. 9, pp. 67-96, 1963.

[Kumar 1992]: KUMAR V., Algorithms for Constraint Satisfaction Problems: a Survey, *AI Magazine*, Vol. 13 n° 1, pp. 32-44, 1992.

[Laurière 1978]: LAURIERE J.L., A language and a program for stating and solving combinatorial problems, *Artificial Intelligence*, Vol. 10, pp. 29-117, 1978.

[Lecoutre 2009]: LECOUTRE C., *Constraint Networks: Techniques and Algorithms*, ISTE/Wiley, 2009.

[Lemmon et al. 1977]: LEMMON E. & SCOTT D., *An introduction to Modal Logic*, Blackwell, 1977.

[Marriot et al. 1998]: MARRIOT K. & STUCKEY P., *Programming with Constraints: an Introduction*, MIT Press, 1998.

[Meinke et al. 1993]: MEINKE K. & TUCKER J., eds., *Many-Sorted Logic and its Applications*, Wiley, 1993.

[Moschovakis 2006]: MOSCHOVAKIS J., Intuitionistic Logic, *Stanford Encyclopedia of Philosophy*, 2006, available at http://plato.stanford.edu/entries/logic-intuitionistic.

[Newell 1982]: NEWELL A., The Knowledge Level, *Artificial Intelligence*, Vol. 59, pp 87-127, 1982.

[Riley 2008]: RILEY G., *CLIPS documentation*, 2008, available at http://clipsrules.sourceforge.net/OnlineDocs.html.

[Rossi et al. 2006]: ROSSI F., VAN BEEK P. & WALSH T., *Handbook of Constraint Programming*, Foundations of Artificial Intelligence, Elsevier, 2006.

[Schank 1986]: SCHANCK R., *Explanation Patterns, Understanding Mechanically and Creatively*, Lawrence Erlbaum Associates Publishers, 1986.

[Stuart 2007]: STUART A., *The Logic of Sudoku*, Michael Mepham Publishing, 2007.

[Van Hentenryck 1989]: VAN HENTENRYCK P., *Constraint Satisfaction in Logic Programming*, MIT Press, 1989.

*Websites*

[Angus www]: ANGUS J. (Simple Sudoku), http://www.angusj.com/sudoku/, 2005-2007 [the main reference for the basic Sudoku techniques].

[Armstrong www]: ARMSTRONG S. (Sadman Software Sudoku, Solving Techniques), http://www.sadmansoftware.com/sudoku/techniques.htm, 2000-2007.

[askmarilyn www]: http://www.parade.com/askmarilyn/index.html [the "official" place for Numbrix® puzzles].

[atksolutions www]: http://www.atksolutions.com [the most interesting source we have found for Futoshiki and Kakuro puzzles].

[Barker 2006]: BARKER M., Sudoku Players Forum, Advanced solving techniques, post 362, *in* http://www.sudoku.com/forums/viewtopic.php?t=3315

[Berthier www]: BERTHIER D., http://www.carva.org/denis.berthier (permanent URL). This is where supplements to this book and to *HLS* can be found.

[Brouwer 2006]: BROUWER A., Solving Sudokus, http://homepages.cwi.nl/~aeb/games/sudoku/, 2006.

[CLIPS www]: http://clipsrules.sourceforge.net

[Davis 2006]: DAVIS T., The Mathematics of Sudoku, www.geometer.org/mathcircles/sudoku.pdf, 2006.

[edhelper www]: http://www.edhelper.com/puzzles.htm [a website with instances of various difficulty levels for many different logic puzzles].

[Eleven www]: https://sites.google.com/site/sudoeleven/, 08/07/2011.

[Eleven 2011]: https://sites.google.com/site/sudoeleven/elevens_hardest_V2.zip?attredirects=0, 08/07/2011.

[Felgenhauer et al. 2005]: FELGENHAUER B. & JARVIS F., Enumerating possible Sudoku grids, http://www.afjarvis.staff.shef.ac.uk/sudoku/sudgroup.html, 2005.

[gsf www]: FOWLER G. (alias gsf), http://www2.research.att.com/~gsf/sudoku

[Hodoku www]: http://hodoku.sourceforge.net

[Jarvis 2006]: JARVIS F., Sudoku enumeration problems, http://www.afjarvis.staff.shef.ac.uk/ sudoku/, 2006.

[JESS www]: http://herzberg.ca.sandia.gov/jess

[Juillerat www]: JUILLERAT N., http://diuf.unifr.ch/people/juillera/Sudoku/Sudoku.html

[Mebane 2012]: MEBANE P., http://mellowmelon.files.wordpress.com/2012/05/pack03 hidato_v3.pdf [the hardest and most interesting Hidato® puzzles we have found].

[Nikoli www]: http://www.nikoli.com/ [Probably the most famous reference in logic puzzles].

[Penet 2012]: PENET G. (alias champagne), http://gpenet.pagesperso-orange.fr/downloads/hard11.zip, 2012.

[Russell et al. 2005]: RUSSELL E. & JARVIS F., There are 5,472,730,538 essentially different Sudoku grids … and the Sudoku symmetry group, http://www.afjarvis.staff.shef.ac.uk/ sudoku/sudgroup.html, 2005.

[Smithsonian www]: http://www.smithsonianmag.com/games/hidato.html [the "official" place for Hidato® puzzles].

[SPlF]: the late Sudoku Player's Forums, http://www.sudoku.com/forums/index.php

[SPrF]: Sudoku Programmers Forums, http://www.setbb.com/sudoku/index.php?mforum= sudoku

[Sterten www]: STERTEN (alias dukuso), http://magictour.free.fr/sudoku.htm

[Sterten 2005]: STERTEN (alias dukuso), *suexg*, http://www.setbb.com/phpbb/viewtopic.php?t=206&mforum= sudoku, 2005.

[Sudopedia]: Sudopedia, http://www.sudopedia.org/wiki/Main_Page

[Tatham www]: http://www.chiark.greenend.org.uk/~sgtatham/puzzles/ [One of the classical references in logic puzzles, with easy instances].

[Werf www]: van der WERF R., Sudocue, Sudoku Solving Guide, http://www.sudocue.net/guide.php, 2005-2007.

[Yato et al. 2002]: YATO T. & SETA T., Complexity and completeness of finding another solution and its application to puzzles, IPSG SIG Notes 2002-AL-87-2, http://www-imai.is.s.u-tokyo.ac.jp/~yato/data2/SIGAL87-2.pdf, 2002.